Zhiyuan Liu · Yankai Lin · Maosong Sun

# Representation Learning for Natural Language Processing

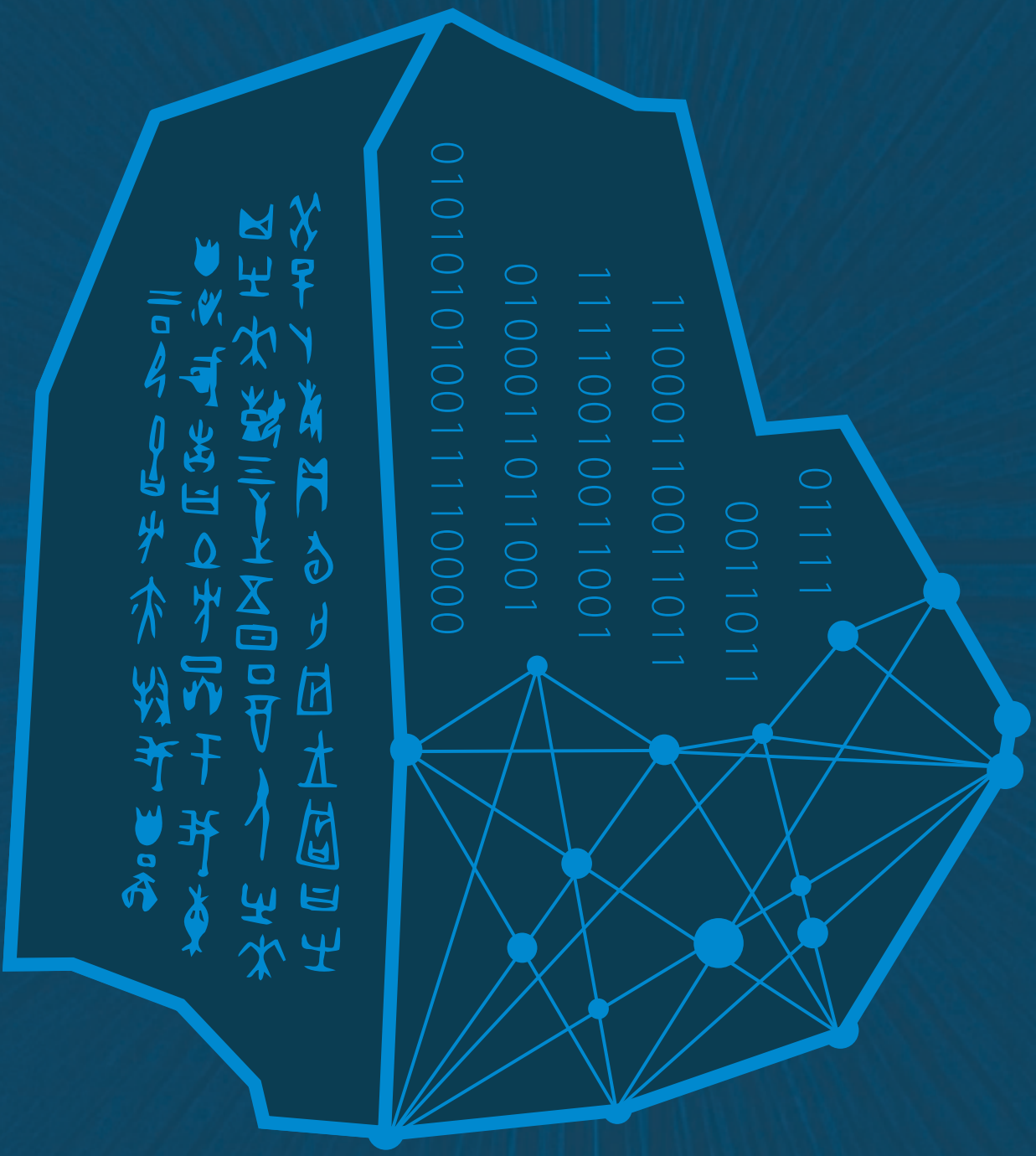

OPEN ACCESS

# Representation Learning for Natural Language Processing

Zhiyuan Liu · Yankai Lin · Maosong Sun

# Representation Learning for Natural Language Processing

Zhiyuan Liu
Tsinghua University
Beijing, China

Maosong Sun
Department of Computer Science
and Technology
Tsinghua University
Beijing, China

Yankai Lin
Pattern Recognition Center
Tencent Wechat
Beijing, China

# Preface

In traditional Natural Language Processing (NLP) systems, language entries such as words and phrases are taken as distinct symbols. Various classic ideas and methods, such as *n*-gram and bag-of-words models, were proposed and have been widely used until now in many industrial applications. All these methods take words as the minimum units for semantic representation, which are either used to further estimate the conditional probabilities of next words given previous words (e.g., *n*-gram) or used to represent semantic meanings of text (e.g., bag-of-words models). Even when people find it is necessary to model word meanings, they either manually build some linguistic knowledge bases such as WordNet or use context words to represent the meaning of a target word (i.e., distributional representation). All these semantic representation methods are still based on symbols!

With the development of NLP techniques for years, it is realized that many issues in NLP are caused by the symbol-based semantic representation. First, the symbol-based representation always suffers from the data sparsity problem. Take statistical NLP methods such as *n*-gram with large-scale corpora, for example, due to the intrinsic power-law distribution of words, the performance will decay dramatically for those few-shot words, even many smoothing methods have been developed to calibrate the estimated probabilities about them. Moreover, there are multiple-grained entries in natural languages from words, phrases, sentences to documents, it is difficult to find a unified symbol set to represent the semantic meanings for all of them simultaneously. Meanwhile, in many NLP tasks, it is required to measure semantic relatedness between these language entries at different levels. For example, we have to measure semantic relatedness between words/phrases and documents in Information Retrieval. Due to the absence of a unified scheme for semantic representation, there used to be distinct approaches proposed and explored for different tasks in NLP, and it sometimes makes NLP does not look like a compatible community.

As an alternative approach to symbol-based representation, distributed representation was originally proposed by Geoffrey E. Hinton in a technique report in 1984. The report was then included in the well-known two-volume book *Parallel Distributed Processing (PDP)* that introduced neural networks to model human cognition and intelligence. According to this report, distributed representation is inspired by the neural computation scheme of humans and other animals, and the essential idea is as follows:

> Each entity is represented by a pattern of activity distributed over many computing elements, and each computing element is involved in representing many different entities.

It means that each entity is represented by multiple neurons, and each neuron involves in the representation of many concepts. This also indicates the meaning of *distributed* in distributed representation. As opposed to distributed representation, people used to assume one neuron only represents a specific concept or object, e.g., there exists a single neuron that will only be activated when recognizing a person or object, such as his/her grandmother, well known as the grandmother-cell hypothesis or local representation. We can see the straightforward connection between the grandmother-cell hypothesis and symbol-based representation.

It was about 20 years after distributed representation was proposed, neural probabilistic language model was proposed to model natural languages by Yoshua Bengio in 2003, in which words are represented as low-dimensional and real-valued vectors based on the idea of distributed representation. However, it was until 2013 that a simpler and more efficient framework *word2vec* was proposed to learn word distributed representations from large-scale corpora, we come to the popularity of distributed representation and neural network techniques in NLP. The performance of almost all NLP tasks has been significantly improved with the support of the distributed representation scheme and the deep learning methods.

This book aims to review and present the recent advances of distributed representation learning for NLP, including why representation learning can improve NLP, how representation learning takes part in various important topics of NLP, and what challenges are still not well addressed by distributed representation.

## Book Organization

This book is organized into 11 chapters with 3 parts. The first part of the book depicts key components in NLP and how representation learning works for them. In this part, Chap. 1 first introduces the basics of representation learning and why it is important for NLP. Then we give a comprehensive review of representation learning techniques on multiple-grained entries in NLP, including word representation (Chap. 2), phrase representation as known as compositional semantics (Chap. 3), sentence representation (Chap. 4), and document representation (Chap. 5).

The second part presents representation learning for those components closely related to NLP. These components include sememe knowledge that describes the commonsense knowledge of words as human concepts, world knowledge (also known as knowledge graphs) that organizes relational facts between entities in the real world, various network data such as social networks, document networks, and cross-modal data that connects natural languages to other modalities such as visual data. A deep understanding of natural languages requires these complex components as a rich context. Therefore, we provide an extensive introduction to these components, i.e., sememe knowledge representation (Chap. 6), world knowledge representation (Chap. 7), network representation (Chap. 8), and cross-modal representation (Chap. 9).

In the third part, we will further provide some widely used open resource tools on representation learning techniques (Chap. 10) and finally outlook the remaining challenges and future research directions of representation learning for NLP (Chap. 11).

Although the book is about representation learning for NLP, those theories and algorithms can be also applied in other related domains, such as machine learning, social network analysis, semantic web, information retrieval, data mining, and computational biology.

Note that, some parts of this book are based on our previous published or pre-printed papers, including [1, 11] in Chap. 2, [32] in Chap. 3, [10, 5, 29] in Chap. 4, [12, 7] in Chap. 5, [17, 14, 24, 30, 6, 16, 2, 15] in Chap. 6, [9, 8, 13, 21, 22, 23, 3, 4, 31] in Chap. 7, and [25, 19, 18, 20, 26, 27, 33, 28] in Chap. 8.

## Book Cover

The book cover shows an oracle bone divided into three parts, corresponding to three revolutionized stages of cognition and representation in human history.

The left part shows oracle scripts, the earliest known form of Chinese writing characters used on oracle bones in the late 1200 BC. It is used to represent the emergence of human languages, especially writing systems. We consider this as the first representation revolution for human beings about the world.

The upper right part shows the digitalized representation of information and signals. After the invention of electronic computers in the 1940s, big data can be efficiently represented and processed in computer programs. This can be regarded as the second representation revolution for human beings about the world.

The bottom right part shows the distributed representation in artificial neural networks originally proposed in the 1980s. As the representation basis of deep learning, it has extensively revolutionized many fields in artificial intelligence, including natural language processing, computer vision, and speech recognition ever since the 2010s. We consider this as the third representation revolution about the world. This book focuses on the theory, methods, and applications of distributed representation learning in natural language processing.

## Prerequisites

This book is designed for advanced undergraduate and graduate students, postdoctoral fellows, researchers, lecturers, and industrial engineers, as well as anyone interested in representation learning and NLP. We expect the readers to have some prior knowledge in Probability, Linear Algebra, and Machine Learning. We recommend the readers who are specifically interested in NLP to read the first part (Chaps. 1–5) which should be read sequentially. The second and third parts can be read in selected order according to readers' interests.

## Contact Information

We welcome any feedback, corrections, and suggestions on the book, which may be sent to liuzy@tsinghua.edu.cn. The readers can also find updates about the book from the personal homepage http://nlp.csai.tsinghua.edu.cn/~lzy/.

Beijing, China
March 2020

Zhiyuan Liu
Yankai Lin
Maosong Sun

# Acknowledgements

The authors are very grateful to the contributions of our students and research collaborators, who have prepared initial drafts of some chapters or have given us comments, suggestions, and corrections. We list main contributors for preparing initial drafts of each chapter as follows,

- Chapter 1: Tianyu Gao, Zhiyuan Liu.
- Chapter 2: Lei Xu, Yankai Lin.
- Chapter 3: Yankai Lin, Yang Liu.
- Chapter 4: Yankai Lin, Zhengyan Zhang, Cunchao Tu, Hongyin Luo.
- Chapter 5: Jiawei Wu, Yankai Lin, Zhenghao Liu, Haozhe Ji.
- Chapter 6: Fanchao Qi, Chenghao Yang.
- Chapter 7: Ruobing Xie, Xu Han.
- Chapter 8: Cheng Yang, Jie Zhou, Zhengyan Zhang.
- Chapter 9: Ji Xin, Yuan Yao, Deming Ye, Hao Zhu.
- Chapter 10: Xu Han, Zhengyan Zhang, Cheng Yang.
- Chapter 11: Cheng Yang, Zhiyuan Liu.

For the whole book, we thank Chaojun Xiao and Zhengyan Zhang for drawing model figures, thank Chaojun Xiao for unifying the styles of figures and tables in the book, thank Shengding Hu for making the notation table and unifying the notations across chapters, thank Jingcheng Yuzhi and Chaojun Xiao for organizing the format of reference, thank Jingcheng Yuzhi, Jiaju Du, Haozhe Ji, Sicong Ouyang, and Ayana for the first-round proofreading, and thank Weize Chen, Ganqu Cui, Bowen Dong, Tianyu Gao, Xu Han, Zhenghao Liu, Fanchao Qi, Guangxuan Xiao, Cheng Yang, Yuan Yao, Shi Yu, Yuan Zang, Zhengyan Zhang, Haoxi Zhong and Jie Zhou for the second-round proofreading. We also thank Cuncun Zhao for designing the book cover.

In this book, there is a specific chapter talking about sememe knowledge representation. Many works in this chapter are carried out by our research group. These works have received great encouragement from the inventor of HowNet, Mr. Zhendong Dong, who died at 82 on February 28, 2019. HowNet is the great linguistic and commonsense knowledge base composed by Mr. Dong for about 30 years. At the end of his life, he and his son Mr. Qiang Dong decided to collaborate with us and released the open-source version of HowNet, OpenHowNet. As a pioneer of machine translation in China, Mr. Zhendong Dong devoted his whole life to natural language processing. He will be missed by all of us forever.

We thank our colleagues and friends, Yang Liu and Juanzi Li at Tsinghua University, and Peng Li at Tencent Wechat, who offered close and frequent discussions which substantially improved this book. We also want to express our special thanks to Prof. Bo Zhang. His insights to deep learning and representation learning, and sincere encouragements to our research of representation learning on NLP, have greatly stimulated us to move forward with more confidence and passion.

We proposed the plan of this book in 2015 after discussing it with the Springer Senior Editor, Dr. Celine Lanlan Chang. As the first of the time of preparing a technical book, we were not expecting it took so long to finish this book. We thank Celine for providing insightful comments and incredible patience to the preparation of this book. We are also grateful to Springer's Assistant Editor, Jane Li, for offering invaluable help during manuscript preparation.

Finally, we give our appreciations to our organizations, Department of Computer Science and Technology at Tsinghua University, Institute for Artificial Intelligence at Tsinghua University, Beijing Academy of Artificial Intelligence (BAAI), Chinese Information Processing Society of China, and Tencent Wechat, who have provided outstanding environment, supports, and facilities for preparing this book.

This book is supported by the Natural Science Foundation of China (NSFC) and the German Research Foundation (DFG) in Project Crossmodal Learning, NSFC 61621136008/DFG TRR-169.

# Contents

# Acronyms

| | |
|---|---|
| ACNN | Anisotropic Convolutional Neural Network |
| AI | Artificial Intelligence |
| AUC | Area Under the Receiver Operating Characteristic Curve |
| BERT | Bidirectional Encoder Representations from Transformers |
| BFS | Breadth-First Search |
| BiDAF | Bi-Directional Attention Flow |
| BRNN | Bidirectional Recurrent Neural Network |
| CBOW | Continuous Bag-of-Words |
| ccDCLM | Context-to-Context Document-Context Language Model |
| CIDEr | Consensus-based Image Description Evaluation |
| CLN | Column Network |
| CLSP | Cross-Lingual Lexical Sememe Prediction |
| CNN | Convolutional Neural Network |
| CNRL | Community-enhanced Network Representation Learning |
| COCO-QA | Common Objects in COntext Question Answering |
| ConSE | Convex Combination of Semantic Embeddings |
| Conv-KNRM | Convolutional Kernel-based Neural Ranking Model |
| CSP | Character-enhanced Sememe Prediction |
| CWE | Character-based Word Embeddings |
| DCNN | Diffusion-Convolutional Neural Network |
| DeViSE | Deep Visual-Semantic Embedding Model |
| DFS | Depth-First Search |
| DGCN | Dual Graph Convolutional Network |
| DGE | Directed Graph Embedding |
| DKRL | Description-embodied Knowledge Graph Representation Learning |
| DRMM | Deep Relevance Matching Model |
| DSSM | Deep Structured Semantic Model |
| ECC | Edge-Conditioned Convolution |
| ERNIE | Enhanced Language Representation Model with Informative Entities |
| FM-IQA | Freestyle Multilingual Image Question Answering |
| GAAN | Gated Attention Network |
| GAT | Graph Attention Networks |
| GCN | Graph Convolutional Network |
| GCNN | Geodesic Convolutional Neural Network |
| GEAR | Graph-based Evidence Aggregating and Reasoning |
| GENQA | Generative Question Answering Model |
| GGNN | Gated Graph Neural Network |
| GloVe | Global Vectors for Word Representation |
| GNN | Graph Neural Networks |
| GRN | Graph Recurrent Network |
| GRU | Gated Recurrent Unit |
| HAN | Heterogeneous Graph Attention Network |
| HMM | Hidden Markov Model |
| HOPE | High-Order Proximity preserved Embeddings |
| IDF | Inverse Document Frequency |
| IE | Information Extraction |
| IKRL | Image-embodied Knowledge Graph Representation Learning |
| IR | Information Retrieval |
| KALM | Knowledge-Augmented Language Model |
| KB | Knowledge Base |
| KBC | Knowledge Base Completion |
| KG | Knowledge Graph |
| KL | Kullback-Leibler |
| KNET | Knowledge-guided Attention Neural Entity Typing |
| K-NRM | Kernel-based Neural Ranking Model |
| KR | Knowledge Representation |
| LBSN | Location-Based Social Network |
| LDA | Latent Dirichlet Allocation |
| LIWC | Linguistic Inquiry and Word Count |
| LLE | Locally Linear Embedding |
| LM | Language Model |
| LSA | Latent Semantic Analysis |
| LSHM | Latent Space Heterogeneous Model |
| LSTM | Long Short-Term Memory |
| MAP | Mean Average Precision |
| METEOR | Metric for Evaluation of Translation with Explicit ORdering |
| MMD | Maximum Mean Discrepancy |
| MMDW | Max-Margin DeepWalk |
| M-NMF | Modularized Nonnegative Matrix Factorization |
| movMF | mixture of von Mises-Fisher distributions |
| MRF | Markov Random Field |
| MSLE | Mean-Square Log-Transformed Error |
| MST | Minimum Spanning Tree |
| MV-RNN | Matrix-Vector Recursive Neural Network |
| NEU | Network Embedding Update |
| NKLM | Neural Knowledge Language Model |
| NLI | Natural Language Inference |
| NLP | Natural Language Processing |
| NRE | Neural Relation Extraction |
| OOKB | Out-of-Knowledge-Base |
| PCNN | Piece-wise Convolution Neural Network |
| pLSI | Probabilistic Latent Semantic Indexing |
| PMI | Point-wise Mutual Information |
| POS | Part-of-Speech |
| PPMI | Positive Point-wise Mutual Information |
| PTE | Predictive Text Embedding |
| PV-DBOW | Paragraph Vector with Distributed Bag-of-Words |
| PV-DM | Paragraph Vector with Distributed Memory |
| QA | Question Answering |
| RBF | Restricted Boltzmann Machine |
| RC | Relation Classification |
| R-CNN | Region-based Convolutional Neural Network |
| RDF | Resource Description Framework |
| RE | Relation Extraction |
| RMSE | Root Mean Squared Error |
| RNN | Recurrent Neural Network |
| RNTN | Recursive Neural Tensor Network |
| RPN | Region Proposal Network |
| SAC | Sememe Attention over Context Model |
| SAT | Sememe Attention over Target Model |
| SC | Semantic Compositionality |
| SCAS | Semantic Compositionality with Aggregated Sememe |
| SCMSA | Semantic Compositionality with Mutual Sememe Attention |
| SDLM | Sememe-Driven Language Model |
| SDNE | Structural Deep Network Embeddings |
| SE-WRL | Sememe-Encoded Word Representation Learning |
| SGD | Stochastic Gradient Descent |
| SGNS | Skip-gram with Negative Sampling Model |
| S-LSTM | Sentence Long Short-Term Memory |
| SPASE | Sememe Prediction with Aggregated Sememe Embeddings |
| SPCSE | Sememe Prediction with Character and Sememe Embeddings |
| SPICE | Semantic Propositional Image Caption Evaluation |
| SPSE | Sememe Prediction with Sememe Embeddings |
| SPWCF | Sememe Prediction with Word-to-Character Filtering |
| SPWE | Sememe Prediction with Word Embeddings |
| SSA | Simple Sememe Aggregation Model |
| SSWE | Sentiment-Specific Word Embeddings |
| SVD | Singular Value Decomposition |
| SVM | Support Vector Machine |
| TADW | Text-associated DeepWalk |
| TF | Term Frequency |
| TF-IDF | Term Frequency–Inverse Document Frequency |
| TKRL | Type-embodied Knowledge Graph Representation Learning |
| TSP | Traveling Salesman Problem |
| TWE | Topical Word Embeddings |
| VQA | Visual Question Answering |
| VSM | Vector Space Model |
| WRL | Word Representation Learning |
| WSD | Word Sense Disambiguation |
| YAGO | Yet Another Great Ontology |

# Symbols and Notations

| | |
|---|---|
| `Tokyo` | Word example |
| $\circledast$ | Convolution operator |
| $\triangleq$ | Defined as |
| $\odot$ | Element-wise multiplication (Hadamard product) |
| $\Rightarrow$ | Induces |
| $\propto$ | Proportional to |
| $\sum$ | Summation operator |
| min | Minimize |
| max | Maximize/max pooling |
| $\arg\min_k$ | The parameter that minimizes a function |
| $\arg\max_k$ | The parameter that maximizes a function |
| sim | Similarity |
| exp | Exponential function |
| Att | Attention function |
| Avg | Average function |
| F1-Score | F1 score |
| PMI | Pair-wise mutual information |
| ReLU | ReLU activation function |
| Sigmoid | Sigmoid function |
| Softmax | Softmax function |
| $V$ | Vocabulary set |
| $w$;$\mathbf{w}$ | Word; word embedding vector |
| $\mathbf{E}$ | Word embedding matrix |
| $R$ | Relation set |
| $r$;$\mathbf{r}$ | Relation; relation embedding vector |
| $a$ | Answer to question |
| $q$ | Query |
| $\mathbf{W}$;$\mathbf{M}$;$\mathbf{U}$ | Weight matrix |
| $\mathbf{b}$;$\mathbf{d}$ | bias vector |
| $\mathbf{M}_{i,:}$;$\mathbf{M}_{:,j}$ | Matrix's $i$th row; matrix's $j$th column |
| $\mathbf{a}^{\mathrm{T}};\mathbf{M}^{\mathrm{T}}$ | Transpose of a vector or a matrix |
| $\mathrm{tr}(\cdot)$ | Trace of a matrix |
| $\overrightarrow{\mathbf{M}}$ | Tensor |
| $\alpha; \beta; \lambda$ | Hyperparameters |
| $g; f; \phi; \Phi; \delta$ | Functions |
| $N_a$ | Number of occurrences of $a$ |
| $d(\cdot, \cdot)$ | Distance function |
| $s(\cdot, \cdot)$ | Similarity function |
| $P(\cdot); p(\cdot)$ | Probability |
| $I(\cdot, \cdot)$ | Mutual information |
| $H(\cdot)$ | Entropy |
| $O(\cdot)$ | Time complexity |
| $D_{\mathrm{KL}}(\cdot\|\|\cdot)$ | KL divergence |
| $\mathscr{L}$ | Loss function |
| $\mathscr{O}$ | Objective function |
| $\mathscr{E}$ | Energy function |
| $\alpha$ | Attention score |
| $\phi^*$ | Optimal value of a variable/function $\phi$ |
| $\vert \cdot \vert$ | Vector length; set size |
| $\Vert\cdot\Vert_a$ | a -norm |
| $\mathbf{1}(\cdot)$ | Indicator function |
| $\theta$ | Parameter in a neural network |
| $\mathbb{E}$ | Expectation of a random variable |
| $\cdot^+; \cdot^-$ | Positive sample; negative sample |
| $\gamma$ | Margin |
| $\mu$ | Mean of normal distribution |
| $\mu$ | Mean vector of Gaussian distribution |
| $\sigma$ | Standard error of normal distribution |
| $\Sigma$ | Covariance matrix of Gaussian distribution |
| $\mathbf{I}_n$ | $n$-dimensional identity matrix |
| $\mathcal{N}$ | Normal distribution |

# Chapter 1
# Representation Learning and NLP

**Abstract** Natural languages are typical unstructured information. Conventional Natural Language Processing (NLP) heavily relies on feature engineering, which requires careful design and considerable expertise. Representation learning aims to learn representations of raw data as useful information for further classification or prediction. This chapter presents a brief introduction to representation learning, including its motivation and basic idea, and also reviews its history and recent advances in both machine learning and NLP.

## 1.1 Motivation

Machine learning addresses the problem of automatically learning computer programs from data. A typical machine learning system consists of three components [5]:

$$\text{Machine Learning} = \text{Representation} + \text{Objective} + \text{Optimization}. \tag{1.1}$$

That is, to build an effective machine learning system, we first transform useful information on raw data into internal representations such as feature vectors. Then by designing appropriate objective functions, we can employ optimization algorithms to find the optimal parameter settings for the system.

Data representation determines how much useful information can be extracted from raw data for further classification or prediction. If there is more useful information transformed from raw data to feature representations, the performance of classification or prediction will tend to be better. Hence, data representation is a crucial component to support effective machine learning.

Conventional machine learning systems adopt careful feature engineering as preprocessing to build feature representations from raw data. Feature engineering needs careful design and considerable expertise, and a specific task usually requires customized feature engineering algorithms, which makes feature engineering labor intensive, time consuming, and inflexible.

Representation learning aims to learn informative representations of objects from raw data automatically. The learned representations can be further fed as input to machine learning systems for prediction or classification. In this way, machine learning algorithms will be more flexible and desirable while handling large-scale and noisy unstructured data, such as speech, images, videos, time series, and texts.

Deep learning [9] is a typical approach for representation learning, which has recently achieved great success in speech recognition, computer vision, and natural language processing. Deep learning has two distinguishing features:

- **Distributed Representation**. Deep learning algorithms typically represent each object with a low-dimensional real-valued dense vector, which is named as *distributed representation*. As compared to one-hot representation in conventional representation schemes (such as bag-of-words models), distributed representation is able to represent data in a more compact and smoothing way, as shown in Fig. 1.1, and hence is more robust to address the sparsity issue in large-scale data.
- **Deep Architecture**. Deep learning algorithms usually learn a *hierarchical deep architecture* to represent objects, known as multilayer neural networks. The deep architecture is able to extract abstractive features of objects from raw data, which is regarded as an important reason for the great success of deep learning for speech recognition and computer vision.

Currently, the improvements caused by deep learning for NLP may still not be so significant as compared to speech and vision. However, deep learning for NLP has been able to significantly reduce the work of feature engineering in NLP in the meantime of performance improvement. Hence, many researchers are devoting to developing efficient algorithms on representation learning (especially deep learning) for NLP.

In this chapter, we will first discuss why representation learning is important for NLP and introduce the basic ideas of representation learning. Afterward, we will briefly review the development history of representation learning for NLP, introduce typical approaches of contemporary representation learning, and summarize existing and potential applications of representation learning. Finally, we will introduce the general organization of this book.

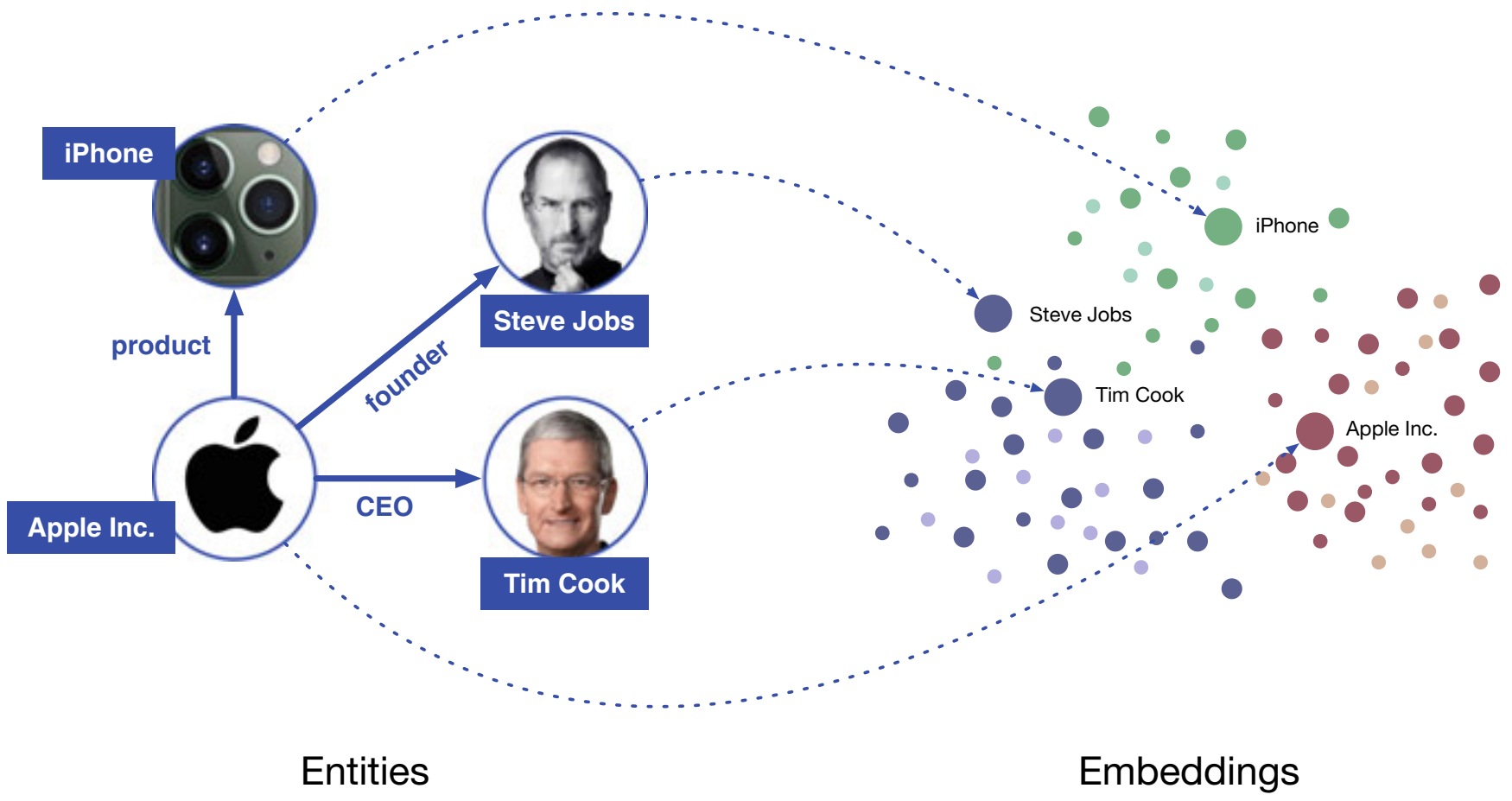

**Fig. 1.1** Distributed representation of words and entities in human languages

## 1.2 Why Representation Learning Is Important for NLP

NLP aims to build linguistic-specific programs for machines to understand languages. Natural language texts are typical unstructured data, with multiple granularities, multiple tasks, and multiple domains, which make NLP challenging to achieve satisfactory performance.

**Multiple Granularities**. NLP concerns about multiple levels of language entries, including but not limited to characters, words, phrases, sentences, paragraphs, and documents. Representation learning can help to represent the semantics of these language entries in a unified semantic space, and build complex semantic relations among these language entries.

**Multiple Tasks**. There are various NLP tasks based on the same input. For example, given a sentence, we can perform multiple tasks such as word segmentation, part-of-speech tagging, named entity recognition, relation extraction, and machine translation. In this case, it will be more efficient and robust to build a unified representation space of inputs for multiple tasks.

**Multiple Domains**. Natural language texts may be generated from multiple domains, including but not limited to news articles, scientific articles, literary works, and online user-generated content such as product reviews. Moreover, we can also regard texts in different languages as multiple domains. Conventional NLP systems have to design specific feature extraction algorithms for each domain according to its characteristics. In contrast, representation learning enables us to build representations automatically from large-scale domain data.

In summary, as shown in Fig. 1.2, representation learning can facilitate knowledge transfer across multiple language entries, multiple NLP tasks, and multiple application domains, and significantly improve the effectiveness and robustness of NLP performance.

## 1.3 Basic Ideas of Representation Learning

In this book, we focus on the distributed representation scheme (i.e., embedding), and talk about recent advances of representation learning methods for multiple language entries, including words, phrases, sentences, and documents, and their closely related objects including sememe-based linguistic knowledge, entity-based world knowledge, networks, and cross-modal entries.

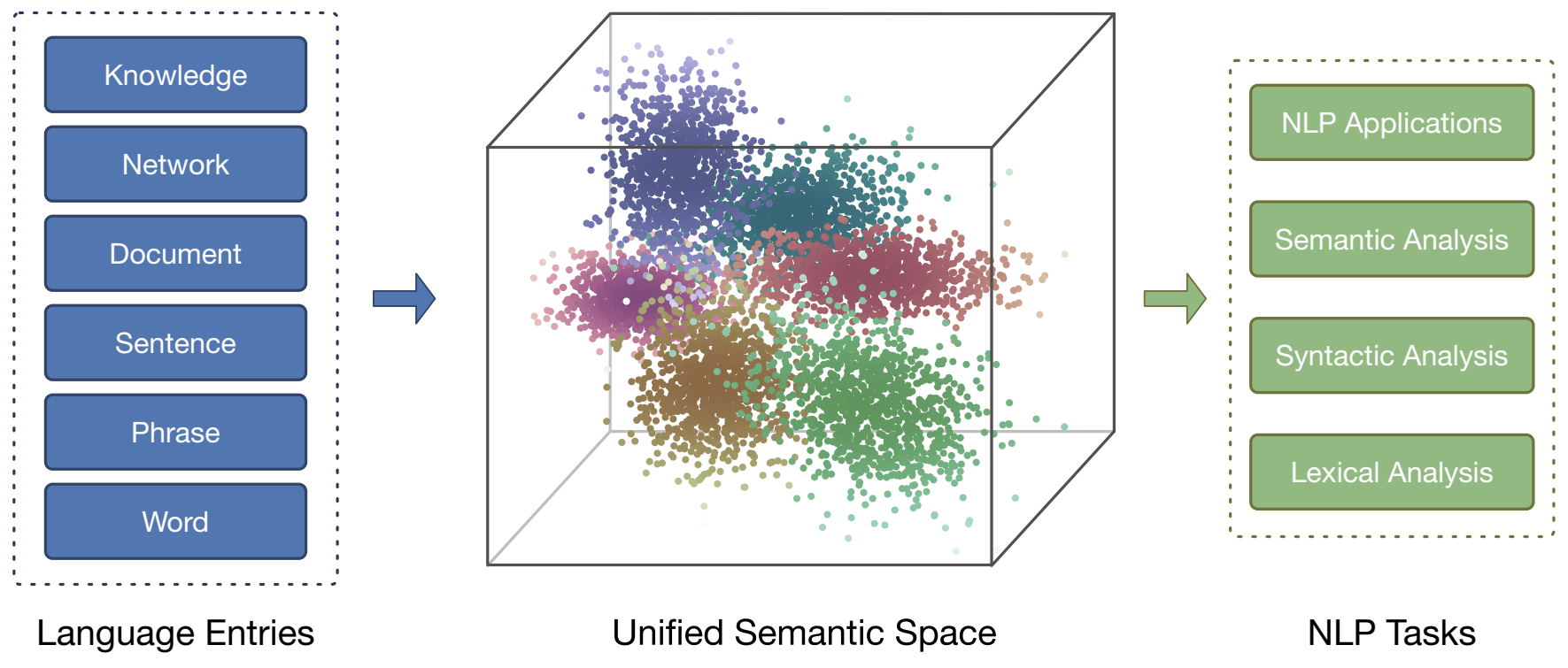

**Fig. 1.2** Distributed representation can provide unified semantic space for multi-grained language entries and for multiple NLP tasks

By distributed representation learning, all objects that we are interested in are projected into a unified low-dimensional semantic space. As demonstrated in Fig. 1.1, the geometric distance between two objects in the semantic space indicates their semantic relatedness; the semantic meaning of an object is related to which objects are close to it. In other words, it is the relative closeness with other objects that reveals an object's meaning rather than the absolute position.

## 1.4 Development of Representation Learning for NLP

In this section, we introduce the development of representation learning for NLP, also shown in Fig. 1.3. To study representation schemes in NLP, words would be a good start, since they are the minimum units in natural languages. The easiest way to represent a word in a computer-readable way (e.g., using a vector) is **one-hot vector**, which has the dimension of the vocabulary size and assigns 1 to the word's corresponding position and 0 to others. It is apparent that one-hot vectors hardly contain any semantic information about words except simply distinguishing them from each other.

One of the earliest ideas of word representation learning can date back to $n$**-gram models** [15]. It is easy to understand: when we want to predict the next word in a sequence, we usually look at some previous words (and in the case of $n$-gram, they are the previous $n - 1$ words). And if going through a large-scale corpus, we can count and get a good probability estimation of each word under the condition of all combinations of $n - 1$ previous words. These probabilities are useful for predicting words in sequences, and also form vector representations for words since they reflect the meanings of words.

The idea of $n$-gram models is coherent with the **distributional hypothesis**: linguistic items with similar distributions have similar meanings [7]. In another phrase, "a word is characterized by the company it keeps" [6]. It became the fundamental idea of many NLP models, from word2vec to BERT.

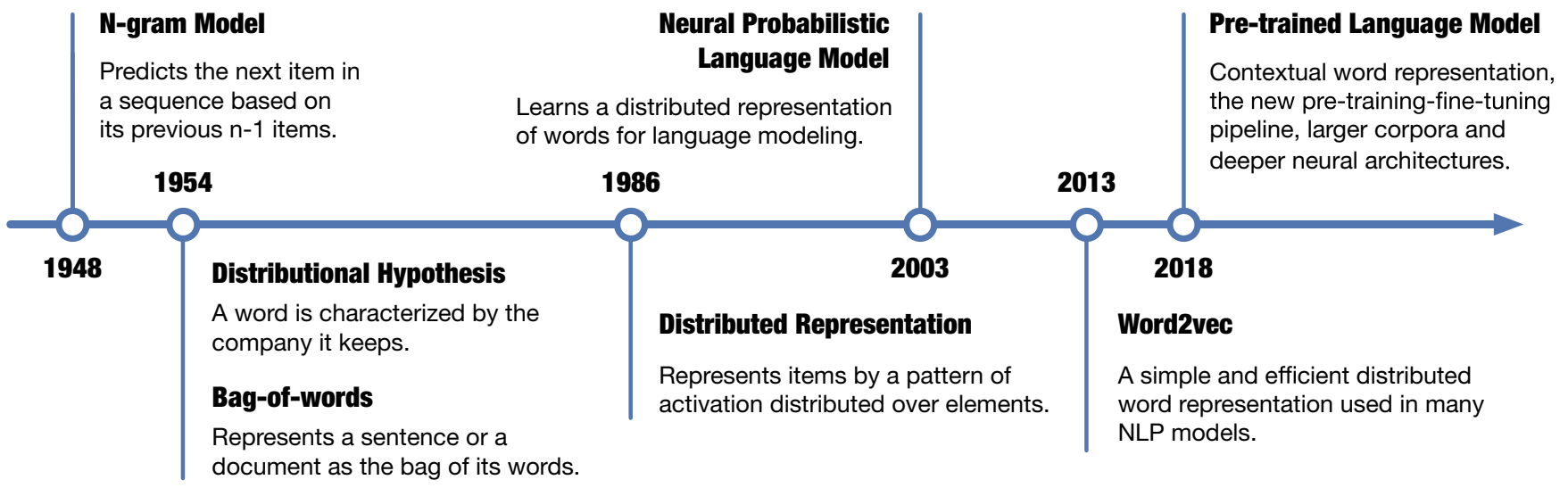

**Fig. 1.3** The timeline for the development of representation learning in NLP. With the growing computing power and large-scale text data, distributed representation trained with neural networks and large corpora has become the mainstream

Another example of the distributional hypothesis is **Bag-Of-Words (BOW) models** [7]. BOW models regard a document as a bag of its words, disregarding the orders of these words in the document. In this way, the document can be represented as a vocabulary-size vector, in which each word that has appeared in the document corresponds to a unique and nonzero dimension. Then a score can be further computed for each word (e.g., the numbers of occurrences) to indicate the weights of these words in the document. Though very simple, BOW models work great in applications like spam filtering, text classification, and information retrieval, proving that the distributions of words can serve as a good representation for text.

In the above cases, each value in the representation clearly matches one entry (e.g., word scores in BOW models). This one-to-one correspondence between concepts and representation elements is called **local representation** or **symbol-based representation**, which is natural and simple.

In **distributed representation**, on the other hand, each entity (or attribute) is represented by a pattern of activation distributed over multiple elements, and each computing element is involved in representing multiple entities [11]. Distributed representation has been proved to be more efficient because it usually has low dimensions that can prevent the sparsity issue. Useful hidden properties can be learned from large-scale data and emerged in distributed representation. The idea of distributed representation was originally inspired by the neural computation scheme of humans and other animals. It comes from neural networks (activations of neurons), and with the great success of deep learning, distributed representation has become the most commonly used approach for representation learning.

One of the pioneer practices of distributed representation in NLP is **Neural Probabilistic Language Model (NPLM)** [1]. A language model is to predict the joint probability of sequences of words ($n$-gram models are simple language models). NPLM first assigns a distributed vector for each word, then uses a neural network to predict the next word. By going through the training corpora, NPLM successfully learns how to model the joint probability of sentences, while brings **word embeddings** (i.e., low-dimensional word vectors) as learned parameters in NPLM. Though it is hard to tell what each element of a word embedding actually means, the vectors indeed encode semantic meanings about the words, verified by the performance of NPLM.

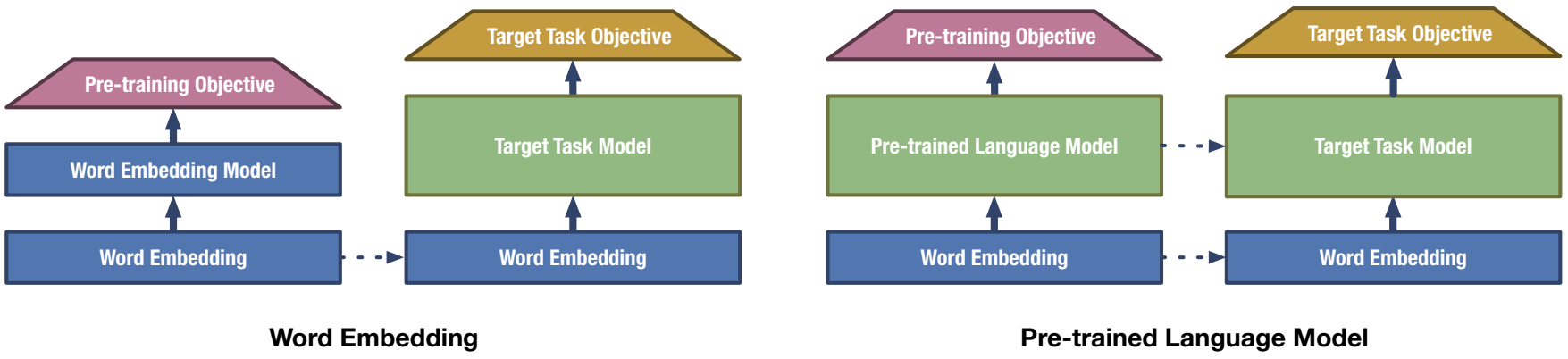

**Fig. 1.4** This figure shows how word embeddings and pre-trained language models work in NLP pipelines. They both learn distributed representations for language entries (e.g., words) through pretraining objectives and transfer them to target tasks. Furthermore, pre-trained language models can also transfer model parameters

Inspired by NPLM, there came many methods that embed words into distributed representations and use the language modeling objective to optimize them as model parameters. Famous examples include **word2vec** [12], **GloVe** [13], and **fastText** [3]. Though differing in detail, these methods are all very efficient to train, utilize large-scale corpora, and have been widely adopted as word embeddings in many NLP models. Word embeddings in the NLP pipeline map discrete words into informative low-dimensional vectors, and help to shine a light on neural networks in computing and understanding languages. It makes representation learning a critical part of natural language processing.

The research on representation learning in NLP took a big leap when **ELMo** [14] and **BERT** [4] came out. Besides using larger corpora, more parameters, and more computing resources as compared to word2vec, they also take complicated context in text into consideration. It means that instead of assigning each word with a fixed vector, ELMo and BERT use multilayer neural networks to calculate dynamic representations for the words based on their context, which is especially useful for the words with multiple meanings. Moreover, BERT starts a new fashion (though not originated from it) of the pretrained fine-tuning pipeline. Previously, word embeddings are simply adopted as input representation. But after BERT, it becomes a common practice to keep using the same neural network structure such as BERT in both pretraining and fine-tuning, which is taking the parameters of BERT for initialization and fine-tuning the model on downstream tasks (Fig. 1.4).

Though not a big theoretical breakthrough, BERT-like models (also known as **Pre-trained Language Models (PLM)**, for they are pretrained through language modeling objective on large corpora) have attracted wide attention in the NLP and machine learning community, for they have been so successful and achieved state-of-the-art on almost every NLP benchmarks. These models show what large-scale data and computing power can lead to, and new research works on the topic of Pre-Trained language Models (PLMs) emerge rapidly. Probing experiments demonstrate that PLMs implicitly encode a variety of linguistic knowledge and patterns inside their multilayer network parameters [8, 10]. All these significant performances and interesting analyses suggest that there are still a lot of open problems to explore in PLMs, as the future of representation learning for NLP.

Based on the distributional hypothesis, representation learning for NLP has evolved from symbol-based representation to distributed representation. Starting from word2vec, word embeddings trained from large corpora have shown significant power in most NLP tasks. Recently, emerged PLMs (like BERT) take complicated context into word representation and start a new trend of the pretraining fine-tuning pipeline, bringing NLP to a new level. What will be the next big change in representation learning for NLP? We hope the contents of this book can give you some inspiration.

## 1.5 Learning Approaches to Representation Learning for NLP

People have developed various effective and efficient approaches to learn semantic representations for NLP. Here we list some typical approaches.

**Statistical Features**: As introduced before, semantic representations for NLP in the early stage often come from statistics, instead of emerging from the optimization process. For example, in $n$-gram or bag-of-words models, elements in the representation are usually frequencies or numbers of occurrences of the corresponding entries counted in large-scale corpora.

**Hand-craft Features**: In certain NLP tasks, syntactic and semantic features are useful for solving the problem. For example, types of words and entities, semantic roles and parse trees, etc. These linguistic features may be provided with the tasks or can be extracted by specific NLP systems. In a long period before the wide use of distributed representation, researchers used to devote lots of effort into designing useful features and combining them as the inputs for NLP models.

**Supervised Learning**: Distributed representations emerge from the optimization process of neural networks under supervised learning. In the hidden layers of neural networks, the different activation patterns of neurons represent different entities or attributes. With a training objective (usually a loss function for the target task) and supervised signals (usually the gold-standard labels for training instances of the target tasks), the networks can learn better parameters via optimization (e.g., gradient descent). With proper training, the hidden states will become informative and generalized as good semantic representations of natural languages.

For example, to train a neural network for a sentiment classification task, the loss function is usually set as the cross-entropy of the model predictions with respect to the gold-standard sentiment labels as supervision. While optimizing the objective, the loss gets smaller, and the model performance gets better. In the meantime, the hidden states of the model gradually form good sentence representations by encoding the necessary information for sentiment classification inside the continuous hidden space.

**Self-supervised Learning**: In some cases, we simply want to get good representations for certain elements, so that these representations can be transferred to other tasks. For example, in most neural NLP models, words in sentences are first mapped to their corresponding word embeddings (maybe from word2vec or GloVe) before sent to the networks. However, there are no human-annotated "labels" for learning word embeddings. To acquire the training objective necessary for neural networks, we need to generate "labels" intrinsically from existing data. This is called self-supervised learning (one way for unsupervised learning).

For example, language modeling is a typical "self-supervised" objective, for it does not require any human annotations. Based on the distributional hypothesis, using the language modeling objective can lead to hidden representations that encode the semantics of words. You may have heard of a famous equation: $\mathbf{w}(\texttt{king}) - \mathbf{w}(\texttt{man}) + \mathbf{w}(\texttt{woman}) = \mathbf{w}(\texttt{queen})$, which demonstrates the analogical properties that the word embeddings have possessed through self-supervised learning.

We can see another angle of self-supervised learning in autoencoders. It is also a way to learn representations for a set of data. Typical autoencoders have a reduction (encoding) phase and a reconstruction (decoding) phase. In the reduction phase, an item from the data is encoded into a low-dimensional representation, and in the reconstruction phase, the model tries to reconstruct the item from the intermediate representation. Here, the training objective is the reconstruction loss, derived from the data itself. During the training process, meaningful information is encoded and kept in the latent representation, while noise signals are discarded.

Self-supervised learning has made a great success in NLP, for the plain text itself contains abundant knowledge and patterns about languages, and self-supervised learning can fully utilize the existing large-scale corpora. Nowadays, it is still the most exciting research area of representation learning for natural languages, and researchers continue to put their efforts into this direction.

Besides, many other machine learning approaches have also been explored in representation learning for NLP, such as adversarial training, contrastive learning, few-shot learning, meta-learning, continual learning, reinforcement learning, et al. How to develop more effective and efficient approaches of representation learning for NLP and to better take advantage of large-scale and complicated corpora and computing power, is still an important research topic.

## 1.6 Applications of Representation Learning for NLP

In general, there are two kinds of applications of representation learning for NLP. In one case, the semantic representation is trained in a pretraining task (or designed by human experts) and is transferred to the model for the target task. Word embedding is an example of the application. It is trained by using language modeling objective and is taken as inputs for other down-stream NLP models. In this book, we will also introduce sememe knowledge representation and world knowledge representation, which can also be integrated into some NLP systems as additional knowledge augmentation to enhance their performance in certain aspects.

In other cases, the semantic representation lies within the hidden states of the neural model and directly aims for better performance of target tasks as an end-to-end fashion. For example, many NLP tasks want to semantically compose sentence or document representation: tasks like sentiment classification, natural language inference, and relation extraction require sentence representation and the tasks like question answering need document representation. As shown in the latter part of the book, many representation learning methods have been developed for sentences and documents and benefit these NLP tasks.

## 1.7 The Organization of This Book

We start the book from word representation. By giving a thorough introduction to word representation, we hope the readers can grasp the basic ideas for representation learning for NLP. Based on that, we further talk about how to compositionally acquire the representation for higher level language components, from sentences to documents.

As shown in Fig. 1.5, representation learning will be able to incorporate various types of structural knowledge to support a deep understanding of natural languages, named as knowledge-guided NLP. Hence, we next introduce two forms of knowledge representation that are closely related to NLP. On the one hand, sememe representation tries to encode linguistic and commonsense knowledge in natural languages. Sememe is defined as the minimum indivisible unit of semantic meaning [2]. With the help of sememe representation learning, we can get more interpretable and more robust NLP models. On the other hand, world knowledge representation studies how to encode world facts into continuous semantic space. It can not only help with knowledge graph tasks but also benefit knowledge-guided NLP applications.

Besides, the network is also a natural way to represent objects and their relationships. In the network representation section, we study how to embed vertices and edges in a network and how these elements interact with each other. Through the applications, we further show how network representations can help NLP tasks.

Another interesting topic related to NLP is the cross-modal representation, which studies how to model unified semantic representations across different modalities (e.g., text, audios, images, videos, etc.). Through this section, we review several cross-modal problems along with representative models.

At the end of the book, we introduce some useful resources to the readers, including deep learning frameworks and open-source codes. We also share some views about the next big topics in representation learning for NLP. We hope that the resources and the outlook can help our readers have a better understanding of the content of the book, and inspire our readers about how representation learning in NLP would further develop.

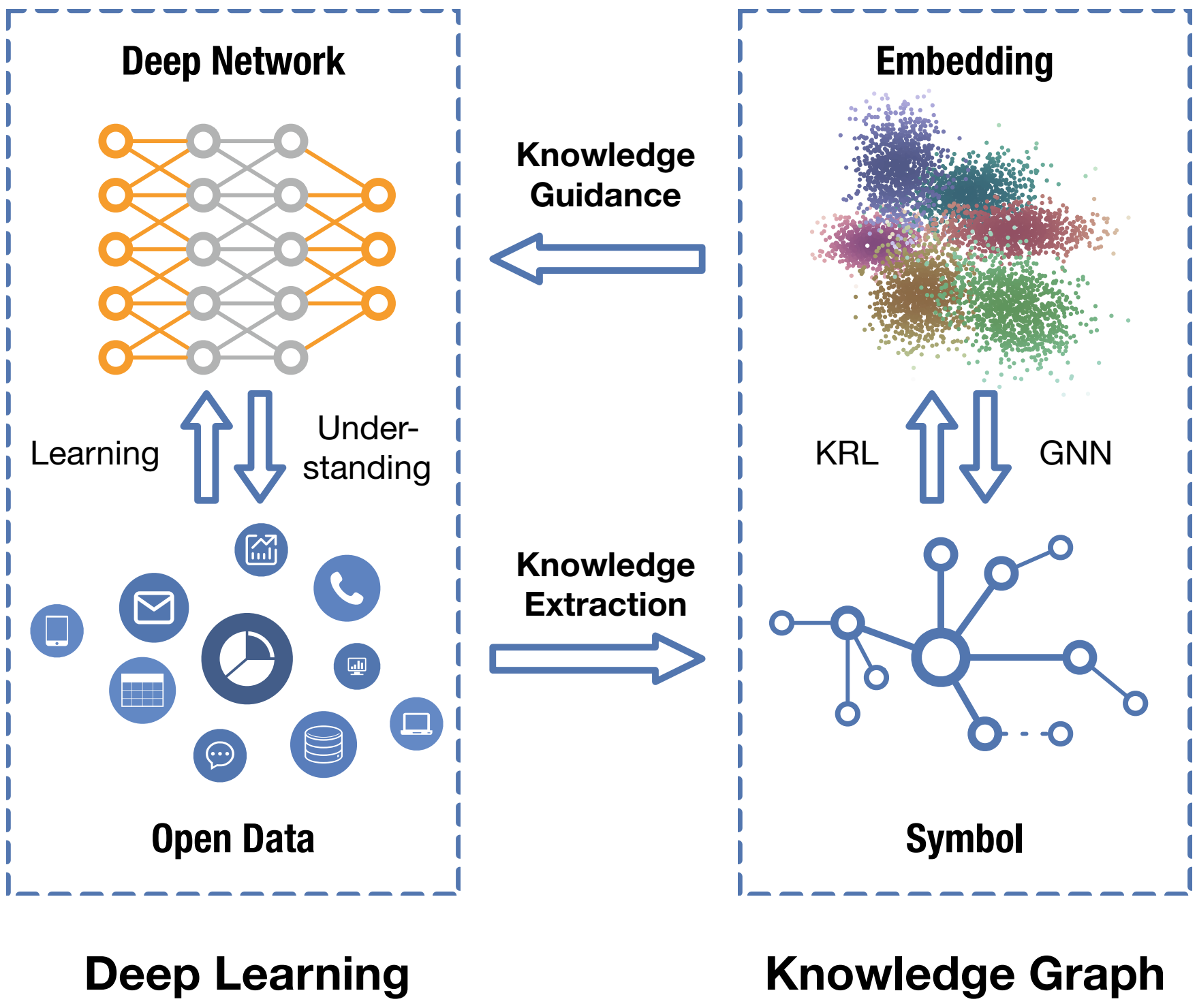

**Fig. 1.5** The architecture of knowledge-guided NLP

# Chapter 2
# Word Representation

**Abstract** Word representation, aiming to represent a word with a vector, plays an essential role in NLP. In this chapter, we first introduce several typical word representation learning methods, including one-hot representation and distributed representation. After that, we present two widely used evaluation tasks for measuring the quality of word embeddings. Finally, we introduce the recent extensions for word representation learning models.

## 2.1 Introduction

Words are usually considered as the smallest meaningful units of speech or writing in human languages. High-level structures in a language, such as phrases and sentences, are further composed of words. For human beings, to understand a language, it is crucial to understand the meanings of words. Therefore, it is essential to accurately represent words, which could help models better understand, categorize, or generate text in NLP tasks.

A word can be naturally represented as a sequence of several characters. However, it is very inefficient and ineffective only to use raw character sequences to represent words. First, the variable lengths of words make it hard to be processed and used in machine learning methods. Moreover, it is very sparse, because only a tiny proportion of arrangements are meaningful. For example, English words are usually character sequences which are composed of 1–20 characters in the English alphabet, but most of these character sequences such as “aaaaa” are meaningless.

One-hot representation is another natural approach to represent words, which assigns a unique index to each word. It is also not good enough to represent words with one-hot representation. First, one-hot representation could not capture the semantic relatedness among words. Second, one-hot representation is a high-dimensional sparse representation, which is very inefficient. Third, it is very inflexible for one-hot representation to deal with new words, which requires assigning new indexes for new words and would change the dimensions of the representation. The change may lead to some problems for existing NLP systems.

Recently, distributed word representation approaches are proposed to address the problem of one-hot word representation. The distributional hypothesis [23, 30] that linguistic objects with similar distributions have similar meanings is the basis for distributed word representation learning. Based on the distributional hypothesis, various word representation models, such as CBOW and Skip-gram, have been proposed and applied in different areas.

In the remaining part of this chapter, we start with one-hot word representation. Further, we introduce distributed word representation models, including Brown Cluster, Latent Semantic Analysis, Word2vec, and GloVe in detail. Then we introduce two typical evaluation tasks for word representation. Finally, we discuss various extensions of word representation models.

## 2.2 One-Hot Word Representation

In this section, we will introduce one-hot word representation in details. Given a fixed set of vocabulary $V = \{w_1, w_2, \ldots, w_{|V|}\}$, one very intuitive way to represent a word $w$ is to encode it with a $|V|$-dimensional vector $\mathbf{w}$, where each dimension of $w$ is either 0 or 1. Only one dimension in $\mathbf{w}$ can be 1 while all the other dimensions are 0. Formally, each dimension of $\mathbf{w}$ can be represented as

$$\mathbf{w}_i = \begin{cases} 1 & \text{if } w = w_i \\ 0 & \text{otherwise.} \end{cases} \tag{2.1}$$

One-hot word representation, in essence, maps each word to an index of the vocabulary, which can be very efficient for storage and computation. However, it does not contain rich semantic and syntactic information of words. Therefore, one-hot representation cannot capture the relatedness among words. The difference between `cat` and `dog` is as much as the difference between `cat` and `bed` in one-hot word representation. Besides, one-hot word representation embeds each word into a $|V|$-dimensional vector, which can only work for a fixed vocabulary. Therefore, it is inflexible to deal with new words in a real-world scenario.

## 2.3 Distributed Word Representation

Recently, distributed word representation approaches are proposed to address the problem of one-hot word representation. The distributional hypothesis [23, 30] that linguistic objects with similar distributions have similar meanings is the basis for semantic word representation learning.

Based on the distributional hypothesis, Brown Cluster [9] groups words into hierarchical clusters where words in the same cluster have similar meanings. The cluster label can roughly represent the similarity between words, but it cannot precisely compare words in the same group. To address this issue, distributed word representation[1] aims to embed each word into a continuous real-valued vector. It is a dense representation, and "dense" means that one concept is represented by more than one dimension of the vector, and each dimension of the vector is involved in representing multiple concepts. Due to its continuous characteristic, distributed word representation can be easily applied in deep neural models for NLP tasks. Distributed word representation approaches such as Word2vec and GloVe usually learn word vectors from a large corpus based on the distributional hypothesis. In this section, we will introduce several distributed word representation approaches in detail.

### *2.3.1 Brown Cluster*

Brown Cluster classifies words into several clusters that have similar semantic meanings. Detailedly, Brown Cluster learns a binary tree from a large-scale corpus, in which the leaves of the tree indicate the words and the internal nodes of the tree indicate word hierarchical clusters. This is a hard clustering method since each word belongs to exactly one group.

The idea of Brown Cluster to cluster the words comes from the $n$-gram language model. A language model evaluates the probability of a sentence. For example, the sentence `have a nice day` should have a higher probability than a random sequence of words. Using a $k$-gram language model, the probability of a sentence $s = \{w_1, w_2, w_3, \ldots, w_n\}$ can be represented as

$$P(s) = \prod_{i=1}^{n} P(w_i|\mathbf{w}_{i-k}^{i-1}). \tag{2.2}$$

It is easy to estimate $P(w_i|\mathbf{w}_{i-k}^{i-1})$ from a large corpus, but the model has $|V|^k - 1$ independent parameters which is a huge number for computers in the 1990s. Even if $k$ is 2, the number of parameters is considerable. Moreover, the estimation is inaccurate for rare words. To address these problems, [9] proposes to group words into clusters and train a cluster-level $n$-gram language model rather than a word-level model. By assigning a cluster to each word, the probability can be written as

$$P(s) = \prod_{i=1}^{n} P(c_i|\mathbf{c}_{i-k}^{i-1})P(w_i|c_i), \tag{2.3}$$

[1]We emphasize that distributed representation and distributional representation are two completely different aspects of representations. A word representation method may belong to both categories. Distributed representation indicates that the representation is a real-valued vector, while distributional representation indicates that the meaning of a word is learned under the distributional hypothesis.

where $c_i$ is the corresponding cluster of $w_i$. In cluster-level language model, there are only $|C^k| - 1 + |V| - |C|$ independent parameters, where $C$ is the cluster set which is usually much smaller than the vocabulary $|V|$.

The quality of the cluster affects the performance of the language model. Given a training text $s$, for a 2-gram language model, the quality of a mapping $\pi$ from words to clusters is defined as

$$Q(\pi) = \frac{1}{n} \log P(s) \tag{2.4}$$

$$= \frac{1}{n} \sum_{i=1}^{n} \big( \log P(c_i|c_{i-1}) + \log P(w_i|c_i) \big). \tag{2.5}$$

Let $N_w$ be the number of times word $w$ appears in corpus $s$, $N_{w_1 w_2}$ be the number of times bigram $w_1 w_2$ appears, and $N_{\pi(w)}$ be the number of times a cluster appears. Then the quality function $Q(\pi)$ can be rewritten in a statistical way as follows:

$$Q(\pi) = \frac{1}{n} \sum_{i=1}^{n} \big( \log P(c_i|c_{i-1}) + \log P(w_i|c_i) \big) \tag{2.6}$$

$$= \sum_{w_1, w_2} \frac{N_{w_1 w_2}}{n} \log P(\pi(w_2)|\pi(w_1)) P(w_2|\pi(w_2)) \tag{2.7}$$

$$= \sum_{w_1, w_2} \frac{N_{w_1 w_2}}{n} \log \frac{N_{\pi(w_1)\pi(w_2)}}{N_{\pi(w_1)}} \frac{N_{w_2}}{N_{\pi(w_2)}} \tag{2.8}$$

$$= \sum_{w_1, w_2} \frac{N_{w_1 w_2}}{n} \log \frac{N_{\pi(w_1)\pi(w_2)} n}{N_{\pi(w_1)} N_{\pi(w_2)}} + \sum_{w_1, w_2} \frac{N_{w_1 w_2}}{n} \log \frac{N_{w_2}}{n} \tag{2.9}$$

$$= \sum_{c_1, c_2} \frac{N_{c_1 c_2}}{n} \log \frac{N_{c_1 c_2} n}{N_{c_1} N_{c_2}} + \sum_{w_2} \frac{N_{w_2}}{n} \log \frac{N_{w_2}}{n}. \tag{2.10}$$

Since $P(w) = \frac{N_w}{n}$, $P(c) = \frac{N_c}{n}$, and $P(c_1 c_2) = \frac{N_{c_1 c_2}}{n}$, the quality function can be rewritten as

$$Q(\pi) = \sum_{c_1, c_2} P(c_1 c_2) \log \frac{P(c_1 c_2)}{P(c_1) P(c_2)} + \sum_{w} P(w) \log P(w) \tag{2.11}$$

$$= I(C) - H(V), \tag{2.12}$$

where $I(C)$ is the mutual information between clusters and $H(V)$ is the entropy of the word distribution, which is a constant value. Therefore, to optimize $Q(\pi)$ equals to optimize the mutual information.

There is no practical method to obtain optimum partitions. Nevertheless, Brown Cluster uses a greedy strategy to obtain a suboptimal result. Initially, it assigns a distinct class for each word. Then it merges two classes with the least average mutual information. After $|V| - |C|$ mergences, the partition is generated. Keeping the $|C|$ clusters, we can continuously perform $|C| - 1$ mergences to get a binary tree. With certain care in implementation, the complexity of this algorithm is $O(|V|^3)$.

**Table 2.1** Some clusters of Brown Cluster

| | | | | | | |
|---|---|---|---|---|---|---|
| Cluster #1 | Friday | Monday | Thursday | Wednesday | Tuesday | Saturday |
| Cluster #2 | June | March | July | April | January | December |
| Cluster #3 | Water | Gas | Coal | Liquid | Acid | Sand |
| Cluster #4 | Great | Big | Vast | Sudden | Mere | Sheer |
| Cluster #5 | Man | Woman | Boy | Girl | Lawyer | Doctor |
| Cluster #6 | American | Indian | European | Japanese | German | African |

We show some clusters in Table 2.1. From the table, we can find that each cluster relates to a sense in the natural language. The words in the same cluster tend to express similar meanings or could be used exchangeably.

### *2.3.2 Latent Semantic Analysis*

Latent Semantic Analysis (LSA) is a family of strategies derived from vector space models, which could capture word semantics much better. LSA aims to explore latent factors for words and documents by matrix factorization to improve the estimation of word similarities. Reference [14] applies Singular Value Decomposition (SVD) on the word-document matrix and exploits uncorrelated factors for both words and documents. The SVD of word-document matrix $\mathbf{M}$ yields three matrices $\mathbf{E}$, $\Sigma$ and $\mathbf{D}$ such that

$$\mathbf{M} = \mathbf{E}\Sigma\mathbf{D}^\top, \tag{2.13}$$

where $\Sigma$ is the diagonal matrix of singular values of $\mathbf{M}$, each row vector $\mathbf{w}_i$ in matrix $\mathbf{E}$ corresponds to word $w_i$, and each row vector $\mathbf{d}_i$ in matrix $\mathbf{D}$ corresponds to document $d_i$. Then the similarity between two words could be

$$\text{sim}(w_i, w_j) = \mathbf{M}_{i,:}\mathbf{M}_{j,:}^\top = \mathbf{w}_i\Sigma^2\mathbf{w}_j. \tag{2.14}$$

Here, the number of singular values $k$ included in $\Sigma$ is a hyperparameter that needs to be tuned. With a reasonable amount of the largest singular values used, LSA could capture much useful information in the word-document matrix and provide a smoothing effect that prevents large variance.

With a relatively small $k$, once the matrices $\mathbf{E}$, $\Sigma$ and $\mathbf{D}$ are computed, measuring word similarity could be very efficient because there are often fewer nonzero dimensions in word vectors. However, the computation of $\mathbf{E}$ and $\mathbf{D}$ can be costly because full SVD on a $n \times m$ matrix takes $O(\min\{m^2n, mn^2\})$ time, while the parallelization of SVD is not trivial.

Another algorithm for LSA is Random Indexing [34, 55]. It overcomes the difficulty of SVD-based LSA, by avoiding costly preprocessing of a huge *word-document* matrix. In random indexing, each document is assigned with a randomly generated high-dimensional sparse ternary vector (called *index vector*). Then for each word in the document, the *index vector* is added to the word's vector. The *index vectors* are supposed to be orthogonal or nearly orthogonal. This algorithm is simple and scalable, which is easy to parallelize and implemented incrementally. Moreover, its performance is comparable with the SVD-based LSA, according to [55].

### *2.3.3 Word2vec*

Google's word2vec[2] toolkit was released in 2013. It can efficiently learn word vectors from a large corpus. The toolkit has two models, including Continuous Bag-Of-Words (CBOW) and Skip-gram. Based on the assumption that the meaning of a word can be learned from its context, CBOW optimizes the embeddings so that they can predict a target word given its context words. Skip-gram, on the contrary, learns the embeddings that can predict the context words given a target word. In this section, we will introduce these two models in detail.

#### 2.3.3.1 Continuous Bag-of-Words

CBOW predicts the center word given a window of context. Figure 2.1 shows the idea of CBOW with a window of 5 words.

Formally, CBOW predicts $w_i$ according to its contexts as

$$P(w_i|w_{j(|j-i|\leq l, j\neq i)}) = \text{Softmax}\left(\mathbf{M}\left(\sum_{|j-i|\leq l, j\neq i}\mathbf{w}_j\right)\right), \tag{2.15}$$

where $P(w_i|w_{j(|j-i|\leq l, j\neq i)})$ is the probability of word $w_i$ given its contexts, $l$ is the size of training contexts, $\mathbf{M}$ is the weight matrix in $\mathbb{R}^{|V|\times m}$, $V$ indicates the vocabulary, and $m$ is the dimension of the word vector.

The CBOW model is optimized by minimizing the sum of negative log probabilities:

$$\mathscr{L} = -\sum_i \log P(w_i|w_{j(|j-i|\leq l, j\neq i)}). \tag{2.16}$$

Here, the window size $l$ is a hyperparameter to be tuned. A larger window size may lead to a higher accuracy as well as the more expense of the training time.

[2] https://code.google.com/archive/p/word2vec/.

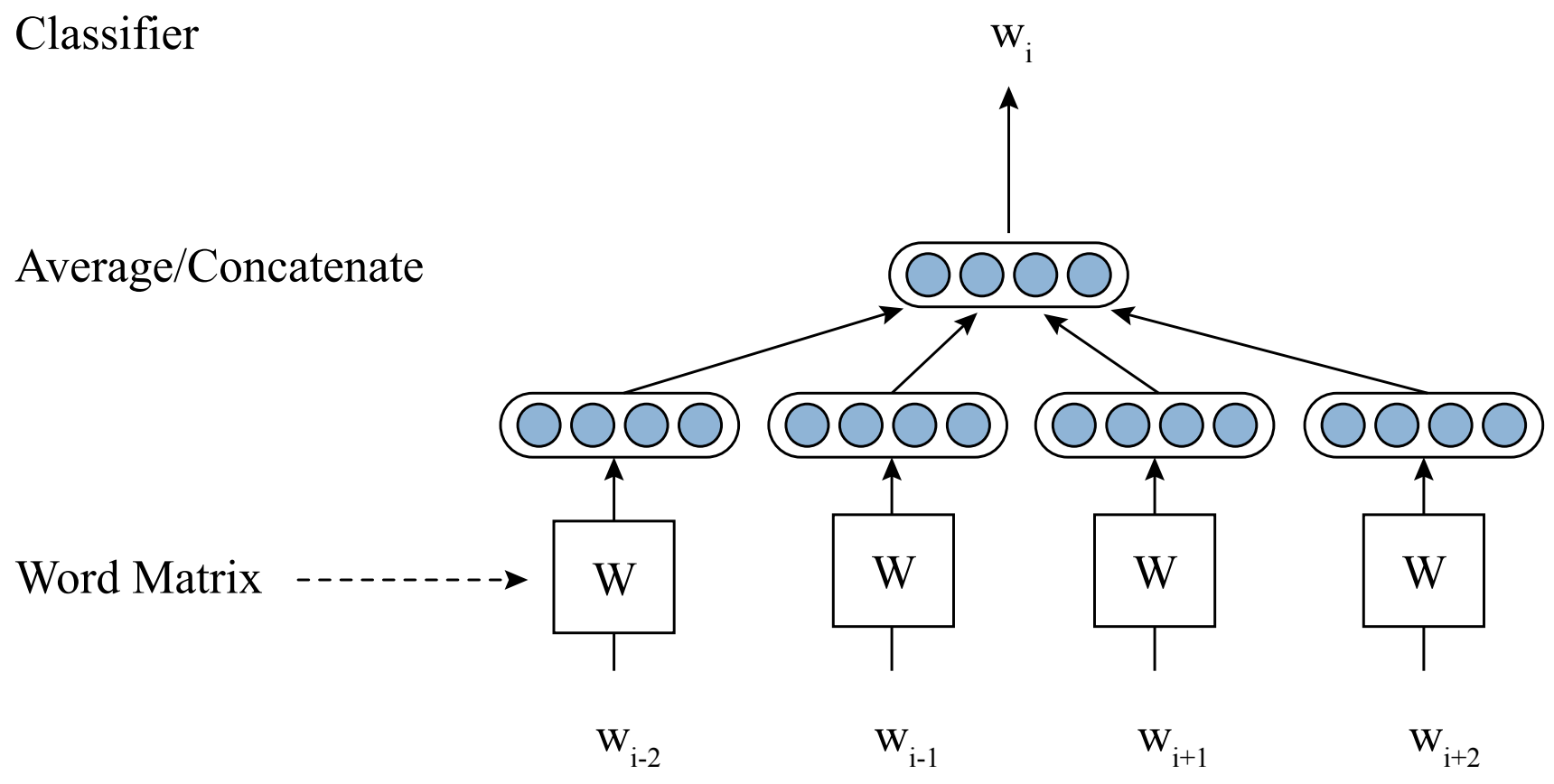

**Fig. 2.1** The architecture of CBOW model

#### 2.3.3.2 Skip-Gram

On the contrary to CBOW, Skip-gram predicts the context given the center word. Figure 2.2 shows the model.

Formally, given a word $w_i$, Skip-gram predicts its context as

$$P(w_j|w_i) = \text{softmax}(\mathbf{M}\mathbf{w}_i)\big(|j-i| \leq l, j \neq i\big), \tag{2.17}$$

where $P(w_j|w_i)$ is the probability of context word $w_j$ given $w_i$, and $\mathbf{M}$ is the weight matrix. The loss function is defined similar to CBOW as

$$\mathscr{L} = -\sum_i \sum_{j(|j-i|\leq l, j\neq i)} P(w_j|w_i). \tag{2.18}$$

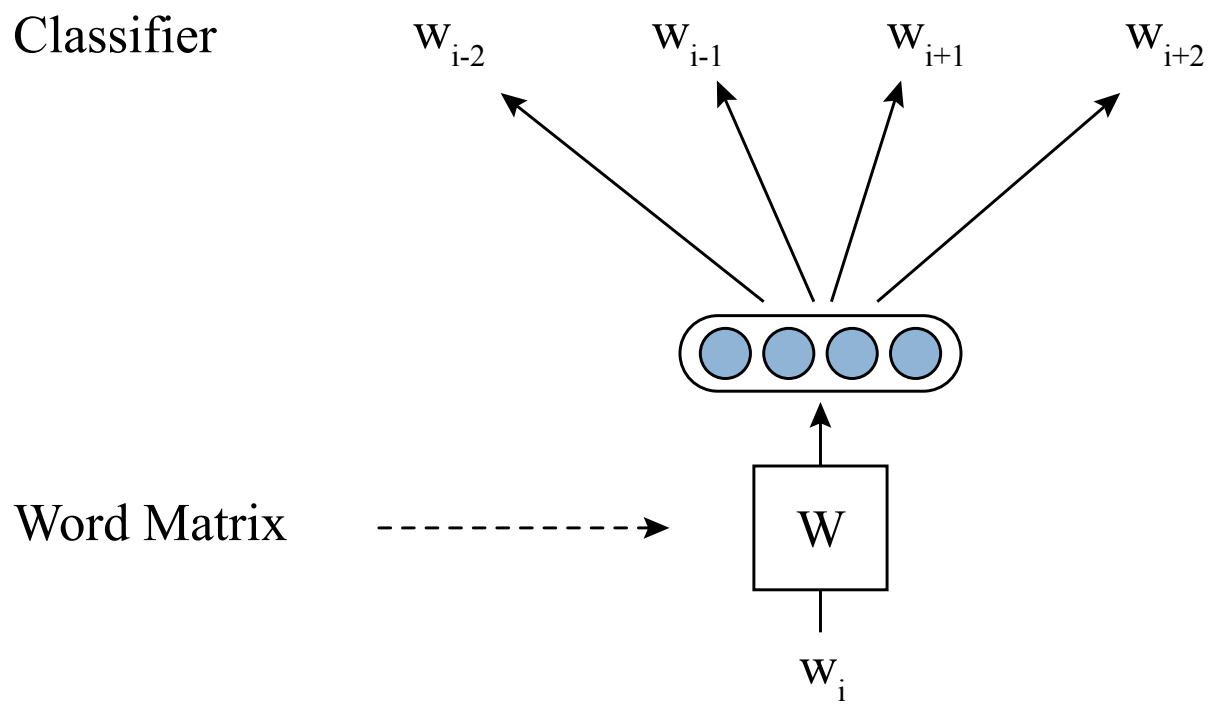

**Fig. 2.2** The architecture of skip-gram model

#### 2.3.3.3 Hierarchical Softmax and Negative Sampling

To train CBOW or Skip-gram directly is very time consuming. The most time-consuming part is the softmax layer. The conventional softmax layer needs to obtain the scores of all words even though only one word is used in computing the loss function. An intuitive idea to improve efficiency is to get a reasonable but much faster approximation of that word. Here, we will introduce two typical approximation methods which are included in the toolkit, including hierarchical softmax and negative sampling. We explain these two methods using CBOW as an example.

The idea of hierarchical softmax is to build hierarchical classes for all words and to estimate the probability of a word by estimating the conditional probability of its corresponding hierarchical class. Figure 2.3 gives an example. Each internal node of the tree indicates a hierarchical class and has a feature vector, while each leaf node of the tree indicates a word. In this example, the probability of word `the` is $p_0 \times p_{01}$ while the probability of `cat` is $p_0 \times p_{00} \times p_{001}$. The conditional probability is computed by the feature vector of each node and the context vector. For example,

$$p_0 = \frac{\exp(\mathbf{w}_0 \cdot \mathbf{w}_c)}{\exp(\mathbf{w}_0 \cdot \mathbf{w}_c) + \exp(\mathbf{w}_1 \cdot \mathbf{w}_c)}, \tag{2.19}$$

$$p_1 = 1 - p_0, \tag{2.20}$$

where $\mathbf{w}_c$ is the context vector, $\mathbf{w}_0$ and $\mathbf{w}_1$ are the feature vectors.

Hierarchical softmax generates the hierarchical classes according to the word frequency, i.e., a Huffman tree. By the approximation, it can compute the probability of each word much faster, and the complexity of calculating the probability of each word is $O(\log |V|)$.

Negative sampling is more straightforward. To calculate the probability of a word, negative sampling directly samples $k$ words as negative samples according to the word frequency. Then, it computes a softmax over the $k + 1$ words to approximate the probability of the target word.

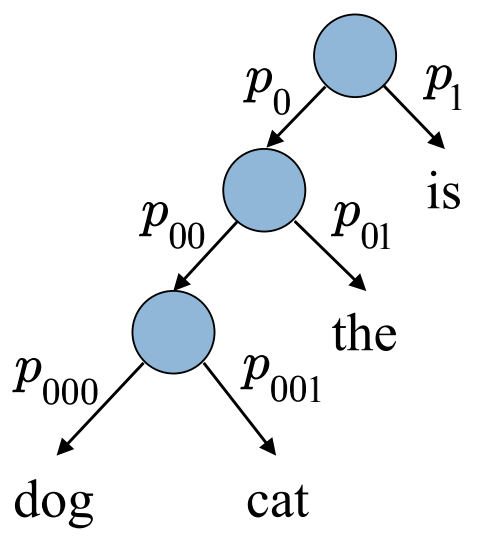

**Fig. 2.3** An illustration of hierarchical softmax

**Table 2.2** Co-occurrence probabilities and the ratio of probabilities for target words `ice` and `steam` with context word `solid`, `gas`, `water`, and `fashion`

| Probability and ratio | $k = solid$ | $k = gas$ | $k = water$ | $k = fashion$ |
|---|---|---|---|---|
| $P(k\|ice)$ | $1.9e-4$ | $6.6e-5$ | $3e-3$ | $1.7e-5$ |
| $P(k\|steam)$ | $2.2e-5$ | $7.8e-4$ | $2.2e-3$ | $1.8e-5$ |
| $P(k\|ice)/P(k\|steam)$ | 8.9 | $8.5e-2$ | 1.36 | 0.96 |

## 2.3.4 GloVe

Methods like Skip-gram and CBOW are shallow window-based methods. These methods scan a context window across the entire corpus, which fails to take advantage of some global information. Global Vectors for Word Representation (GloVe), on the contrary, can capture corpus statistics directly.

As shown in Table 2.2, the meaning of a word can be learned from the co-occurrence matrix. The ratio of co-occurrence probabilities can be especially useful. In the example, the meaning of `ice` and `water` can be examined by studying the ratio of their co-occurrence probabilities with various probe words. For words related to `ice` but not `steam`, for example, `solid`, the ratio $P(\texttt{solid}|\texttt{ice})/P(\texttt{solid}|\texttt{steam})$ will be large. Similarly, `gas` is related to `steam` but not `ice`, so $P(\texttt{gas}|\texttt{ice})/P(\texttt{gas}|\texttt{steam})$ will be small. For words that are relevant or irrelevant to both words, the ratio is close to 1.

Based on this idea, GloVe models

$$F(\mathbf{w}_i, \mathbf{w}_j, \tilde{\mathbf{w}}_k) = \frac{P_{ik}}{P_{jk}}, \tag{2.21}$$

where $\tilde{\mathbf{w}} \in \mathbb{R}^d$ are separate context word vectors, and $P_{ij}$ is the probability of word $j$ to be in the context of word $i$, formally

$$P_{ij} = \frac{N_{ij}}{N_i}, \tag{2.22}$$

where $N_{ij}$ is the number of occurrences of word $j$ in the context of word $i$, and $N_i = \sum_k N_{ik}$ is the number of times any word appears in the context of word $j$.

$F(\cdot)$ is supposed to encode the information presented in the ratio $P_{ik}/P_{jk}$ in the word vector space. To keep the inherently linear structure, $F$ should only depend on the difference of two target words

$$F(\mathbf{w}_i - \mathbf{w}_j, \tilde{\mathbf{w}}_k) = \frac{P_{ik}}{P_{jk}}. \tag{2.23}$$

The arguments of F are vectors while the right side of the equation is a scalar, to avoid $F$ obfuscating the linear structure, a dot product is used:

$$F\left((\mathbf{w}_i - \mathbf{w}_j)^\top \tilde{\mathbf{w}}_k\right) = \frac{P_{ik}}{P_{jk}}. \tag{2.24}$$

The model keeps the invariance under relabeling the target word and context word. It requires $F$ to be a homomorphism between the groups $(\mathbb{R}, +)$ and $(\mathbb{R}_{>0}, \times)$. The solution is $F = \exp$. Then

$$\mathbf{w}_i^\top \tilde{\mathbf{w}}_k = \log N_{ik} - \log N_i. \tag{2.25}$$

To keep exchange symmetry, $\log N_i$ is eliminated by adding biases $b_i$ and $\tilde{b}_k$. The model becomes

$$\mathbf{w}_i^\top \tilde{\mathbf{w}}_k + b_i + \tilde{b}_k = \log N_{ik}, \tag{2.26}$$

which is significantly simpler than Eq. (2.21).

The loss function is defined as

$$\mathcal{L} = \sum_{i,j=1}^{|V|} f(N_{ij})(\mathbf{w}_i^\top \tilde{\mathbf{w}}_j + b_i + \tilde{b}_j - \log N_{ij}), \tag{2.27}$$

where $f(\cdot)$ is a weighting function:

$$f(x) = \begin{cases} (x/x_{max})^\alpha & \text{if } x < x_{max}, \\ 1 & \text{otherwise.} \end{cases} \tag{2.28}$$

## 2.4 Contextualized Word Representation

In natural language, the meaning of an individual word usually relates to its context in a sentence. For example,

- `The central bank has slashed its forecast for economic growth this year from 4.1 to 2.6%.`
- `More recently, on a blazing summer day, he took me back to one of the den sites, in a slumping bank above the South Saskatchewan River.`

In these two sentences, although the word `bank` is always the same, their meanings are different. However, most of the traditional word embeddings (CBOW, Skip-gram, GloVe, etc.) cannot well understand the different nuances of the meanings of words with the different surrounding texts. The reason is that these models only learn a unique representation for each word, and therefore it is impossible for these models to capture how the meanings of words change based on their surrounding contexts.

To address this issue, [48] proposes ELMo, which uses a deep, bidirectional LSTM model to build word representations. ELMo could represent each word depending on the entire context in which it is used. More specifically, rather than having a look-up table of word embedding matrix, ELMo converts words into low-dimensional vectors on-the-fly by feeding the word and its surrounding text into a deep neural network. ELMo utilizes a bidirectional language model to conduct word representation. Formally, given a sequence of $N$ words, $(w_1, w_2, \ldots, w_N)$, a forward language model (LM, the details of language model are in Sect. 4) models the probability of the sequence by predicting the probability of each word $t_k$ according to the historical context:

$$P(w_1, w_2, \ldots, w_N) = \prod_{k=1}^{N} P(w_k \mid w_1, w_2, \ldots, w_{k-1}). \tag{2.29}$$

The forward LM in ELMo is a multilayer LSTM, and the $j$th layer of the LSTM-based forward LM will generate the context-dependent word representation $\overrightarrow{\mathbf{h}}_{k,j}^{LM}$ for the word $w_k$. The backward LM is similar to the forward LM. The only difference is that it reverses the input word sequence to $(w_N, w_{N-1}, \ldots, w_1)$ and predicts each word according to the future context:

$$P(w_1, w_2, \ldots, w_N) = \prod_{k=1}^{N} P(w_k \mid w_{k+1}, w_{k+2}, \ldots, w_N). \tag{2.30}$$

As the same as the forward LM, the $j$th backward LM layer generates the representations $\overleftarrow{\mathbf{h}}_{k,j}^{LM}$ for the word $w_k$.

ELMo generates a task-specific word representation, which combines all layer representations of the bidirectional LM. Formally, it computes a task-specific weighting of all bidirectional LM layers:

$$\mathbf{w}_k = \alpha^{task} \sum_{j=0}^{L} s_j^{task} \mathbf{h}_{k,j}^{LM}, \tag{2.31}$$

where $\mathbf{s}^{task}$ are softmax-normalized weights and $\alpha^{task}$ is the weight of the entire word vector for the task.

## 2.5 Extensions

Besides those very popular toolkits, such as word2vec and GloVe, various works are focusing on different aspects of word representation, contributing to numerous extensions. These extensions usually focus on the following directions.

### 2.5.1 Word Representation Theories

With the success of word representation, researchers begin to explore the theories of word representation. Some works attempt to give more theoretical analysis to prove the reasonability of existing tricks on word representation learning [39, 45], while some works try to discuss the new learning methods [26, 61].

**Reasonability**. Word2vec and other similar tools are empirical methods of word representation learning. Many tricks are proposed in [43] to learn the representation of words from a large corpus efficiently, for example, negative sampling. Considering the effectiveness of these methods, a more theoretical analysis should be done to prove the reasonability of these tricks. Reference [39] gives some theoretical analysis of these tricks. They formalize the Skip-gram model with negative sampling as an implicit matrix factorization process. The Skip-gram model generates a word embedding matrix $\mathbf{E}$ and a context matrix $\mathbf{C}$. The size of the word embedding matrix $\mathbf{E}$ is $|V| \times m$. Each row of context matrix $\mathbf{C}$ is a context word's $m$-dimensional vector. The training process of Skip-gram is an implicit factorization of $\mathbf{M} = \mathbf{E}\mathbf{C}^\top$. $\mathbf{C}$ is not explicitly considered in word2vec. This work further analyzes that the matrix $\mathbf{M}$ is

$$\mathbf{M}_{ij} = \mathbf{w}_i \cdot \mathbf{c}_j = \mathrm{PMI}(w_i, c_j) - \log k, \tag{2.32}$$

where $k$ is the number of negative samples, $\mathrm{PMI}(w, c)$ is the point-wise mutual information

$$\mathrm{PMI}(w, c) = \log \frac{P(w, c)}{P(w)P(c)}. \tag{2.33}$$

The shifted PMI matrix can directly be used to compare the similarity of words. Another intuitive idea is to factorize the shifted PMI matrix directly. Reference [39] evaluates the performance of using the SVD matrix factorization method on the implicit matrix $\mathbf{M}$. Matrix factorization achieves significantly better objective value when the embedding size is smaller than 500 dimensions and the number of negative samples is 1. With more negative samples and higher embedding dimensions, Skip-gram with negative sampling gets better objective value. This is because when the number of zeros increases in $\mathbf{M}$, and SVD prefers to factorize a matrix with minimum values. With 1,000 dimensional embeddings and different numbers of negative samples in $\{1, 5, 15\}$, SVD achieves slightly better performance on word analogy and word similarity. In contrast, Skip-gram with negative sampling achieves 2% better performance on syntactical analogy.

**Interpretability**. Most existing distributional word representation methods could generate a dense real-valued vector for each word. However, the word embeddings obtained by these models are hard to be interpreted. Reference [26] introduces non-negative and sparsity embeddings, where the models are interpretable and each dimension indicates a unique concept. This method factorizes the corpus statistics matrix $\mathbf{X} \in \mathbb{R}^{|V| \times |D|}$ into a word embedding matrix $\mathbf{E} \in \mathbb{R}^{|V| \times m}$ and a document statistics matrix $\mathbf{D} \in \mathbb{R}^{m \times |D|}$. Its training objective is

$$\begin{aligned}&\arg\min_{\mathbf{E},\mathbf{D}} \frac{1}{2}\sum_{i=1}^{|V|} \|\mathbf{X}_{i,:} - \mathbf{E}_{i,:}\mathbf{D}\|_2 + \lambda\|\mathbf{E}_{i,:}\|_1,\\ &\text{s.t. } \mathbf{D}_{i,:}\mathbf{D}_{i,:}^\top \le 1, \forall 1 \le i \le m,\\ &\quad \mathbf{E}_{i,j} \ge 0, 1 \le i \le |V|, 1 \le j \le m.\end{aligned} \tag{2.34}$$

By iteratively optimizing **E** and **D** via gradient descent, this model can learn nonnegative and sparse embeddings for words. Since the embeddings are sparse and nonnegative, words with the highest scores in each dimension show high similarity, which can be viewed as a concept of this dimension. To further improve the embeddings, this work also proposes phrasal-level constraints into the loss function. With new constraints, it could achieve both interpretability and compositionality.

### *2.5.2 Multi-prototype Word Representation*

Using only one single vector to represent a word is problematic due to the ambiguity of words. A single vector cannot represent multiple meanings of a word well because it may lead to semantic confusion among the different senses of this word.

The multi-prototype vector space model [51] is proposed to better represent different meanings of a word. In multi-prototype vector space model, a mixture of von Mises-Fisher distributions (movMF) clustering method with first-order unigram contexts [5] is used to cluster different meanings of a word. Formally, it assigns a different word representation $\mathbf{w}_i(x)$ to the same word $x$ in each different cluster $i$. When the multi-prototype embedding is used, the similarity between two words $x$, $y$ is computed straightforwardly. If contexts of words are not available, the similarity between two words is defined as

$$\text{AvgSim}(x, y) = \frac{1}{K^2}\sum_{i=1}^{K}\sum_{j=1}^{K} s(\mathbf{w}_i(x), \mathbf{w}_j(y)), \tag{2.35}$$

$$\text{MaxSim}(x, y) = \max_{1\le i,j\le K} s(\mathbf{w}_i(x), \mathbf{w}_j(y)), \tag{2.36}$$

where $K$ is a hyperparameter indicating the number of the clusters and $s(\cdot)$ is a similarity function of two vectors such as cosine similarity. When contexts are available, the similarity can be computed more precisely as:

$$\text{AvgSimC}(x, y) = \frac{1}{K^2}\sum_{i=1}^{K}\sum_{j=1}^{K} s_{c,x,i}s_{c,y,j}s(\mathbf{w}_i(x), \mathbf{w}_j(y)), \tag{2.37}$$

$$\text{MaxSimC}(x, y) = s(\hat{\mathbf{w}}(x), \hat{\mathbf{w}}(y)), \tag{2.38}$$

where $s_{c,x,i} = s(\mathbf{w}_i(c), \mathbf{w}_i(x))$ is the likelihood of context $c$ belonging to cluster $i$, and $\hat{\mathbf{w}}(x) = \mathbf{w}_{\arg\max_{1\leq i\leq K} s_{c,x,i}}(x)$ is the maximum likelihood cluster for $x$ in context $c$. With multi-prototype embeddings, the accuracy on the word similarity task is significantly improved, but the performance is still sensitive to the number of clusters.

Although the multi-prototype embedding method can effectively cluster different meanings of a word via its contexts, the clustering is offline, and the number of clusters is fixed and needs to be predefined. It is difficult for a model to select an appropriate amount of meanings for different words, to adapt to new senses, new words, or new data, and to align the senses with prototypes. To address these problems, [12] proposes a unified model for word sense representation and word sense disambiguation. This model uses available knowledge bases such as WordNet [46] to determine the senses of a word. Each word and each sense had a single vector and are trained jointly. This model can learn representations of both words and senses, and two simple methods are proposed to do disambiguation using the word and sense vectors.

## 2.5.3 *Multisource Word Representation*

There is much information about words that can be leveraged to improve the quality of word representations. We will introduce other kinds of word representation learning methods utilizing multisource information.

### 2.5.3.1 Word Representation with Internal Information

There is much information locating inside words, which can be utilized to improve the quality of word representations further.

**Using Character Information.** Many languages such as Chinese and Japanese have thousands of characters, and the words in these languages are composed of several characters. Characters in these languages have richer semantic information comparing with other languages containing only dozens of characters. Hence, the meaning of a word can not only be learned from its contexts but also the composition of characters. Driven by this intuitive idea, [13] proposes a joint learning model for Character and Word Embeddings (CWE). In CWE, a word is a composition of a word embedding and its character embeddings. Formally,

$$\mathbf{x} = \mathbf{w} + \frac{1}{|w|}\sum_{i}\mathbf{c}_i, \tag{2.39}$$

where $\mathbf{x}$ is the representation of a word, which is the composition of a word vector $\mathbf{w}$ and several character vectors $\mathbf{c}_i$, and $|w|$ is the number of characters in the word. Note that this model can be integrated with various models such as Skip-gram, CBOW, and GloVe.

Further, position-based and cluster-based methods are proposed to address this issue that characters are highly ambiguous. In position-based approach, each character is assigned three vectors which appear in *begin*, *middle* and *end* of a word respectively. Since the meaning of a character varies when it appears in the different positions of a word, this method can significantly resolve the ambiguity problem. However, characters that appear in the same position may also have different meanings. In the cluster-based method, a character is assigned $K$ different vectors for its different meanings, in which a word's context is used to determine which vector to be used.

By introducing character embeddings, the representation of low-frequency words can be significantly improved. Besides, this method can deal with new words while other methods fail. Experiments show that the joint learning method can achieve better performance on both word similarity and word analogy tasks. By disambiguating characters using the position-based and cluster-based method, it can further improve the performance.

**Using Morphology Information.** Many languages such as English have rich morphology information and plenty of rare words. Most word representation models assign a distinct vector to each word ignoring the rich morphology information. This is a limitation because the affixes of a word can help infer the meaning of a word and the morphology information of word is essential especially when facing rare contexts.

To address this issue, [8] proposes to represent a word as a bag of morphology $n$-grams. This model substitutes word vectors in Skip-gram with the sum of morphology $n$-gram vectors. When creating the dictionary of $n$-grams, they select all $n$-grams with a length greater or equal than 3 and smaller or equal than 6. To distinguish prefixes and suffixes with other affixes, they also add special characters to indicate the beginning and the end of a word. This model is efficient and straightforward, which achieves good performance on word similarity and word analogy tasks especially when the training set is small.

Reference [41] further uses a bidirectional LSTM to generate word representation by composing morphologies. This model does not use a look-up table to assign a distinct vector to each word like what those independent word embedding methods are doing. Hence, this model not only significantly reduces the number of parameters but also addresses some disadvantages of independent word embeddings. Moreover, the embeddings of words in this model could affect each other.

#### 2.5.3.2 Word Representation with External Knowledge

Besides internal information of words, there is much external knowledge that could help us learn the word representations.

**Using Knowledge Base.** Some languages have rich internal information, whereas people have also annotated lots of knowledge bases which can be used in word representation learning to constrain embeddings. Reference [62] introduces relation constraints into the CBOW model. With these constraints, the embeddings can not only predict its contexts, but also predict words with relations. The objective is to maximize the sum of log probability of all relations as

$$\mathscr{O} = \frac{1}{N} \sum_{i=1}^{N} \sum_{w \in R_{w_i}} \log P(w|w_i), \tag{2.40}$$

where $R_{w_i}$ indicates a set of words which have relation with $w_i$. Then the joint objective is defined as

$$\mathscr{O} = \frac{1}{N} \sum_{i=1}^{N} \log P(w_i | w_{j(|j-i|<l, j \neq i)}) + \frac{\beta}{N} \sum_{i=1}^{N} \sum_{w \in R_{w_i}} \log p(w|w_i), \tag{2.41}$$

where $\beta$ is a hyperparameter. The external information helps to train a better word representation, which shows significant improvements on word similarity benchmarks.

Moreover, Retrofitting [19] introduces a post-processing step which can introduce knowledge bases into word representation learning. It is more modular than other approaches which consider knowledge base during training. Let the word embeddings learned by existing word representation approaches be $\mathbf{E}$. Retrofitting attempts to find another embedding space $\hat{\mathbf{E}}$, which is close to $\mathbf{E}$ but considers the relations in the knowledge base. Formally,

$$\mathscr{L} = \sum_{i} \big( \alpha_i \|\mathbf{w}_i - \hat{\mathbf{w}}_i\|_2 + \sum_{(i,j) \in R} \beta_{ij} \|\mathbf{w}_i - \mathbf{w}_j\|_2 \big), \tag{2.42}$$

where $\alpha$ and $\beta$ are hyperparameters indicating the strength of the associations, and $R$ is a set of relations in the knowledge base. The adapted embeddings $\hat{\mathbf{E}}$ can be optimized by several iterations of the following online updates:

$$\hat{\mathbf{w}}_i = \frac{\sum_{\{j|(i,j) \in R\}} \beta_{ij} \hat{\mathbf{w}}_j + \alpha_i \mathbf{w}_i}{\sum_{\{j|(i,j) \in R\}} \beta_{ij} + \alpha_i}, \tag{2.43}$$

where $\alpha$ is usually set to 1 and $\beta_{ij}$ is $\deg(i)^{-1}$ ($\deg(\cdot)$ is a node's degree in a knowledge graph). With knowledge bases such as the paraphrase database [27], WordNet [46] and FrameNet [3], this model can achieve consistent improvement on word similarity tasks. But it also may significantly reduce the performance on the analogy of syntactic relations. Since this module is a post-processing of word embeddings, it is compatible with various distributed representation models.

In addition to the aforementioned synonym-based knowledge bases, there are also sememe-based knowledge bases, in which the sememe is defined as the minimum semantic unit of word meanings. HowNet [16] is one of such knowledge bases, which annotates each Chinese word with one or more relevant sememes. General knowledge injecting methods could not apply to HowNet. As a result, [47] proposes a specific model to introduce HowNet into word representation learning.

Bases on Skip-gram model, [47] introduces sense and sememe embeddings to represent target word $w_i$. More specifically, this model leverages context words, which are represented with original word embeddings, as attention over multiple senses of target word $w_i$ to obtain its new embeddings.

$$w_i = \sum_{k=1}^{|S^{(w_i)}|} \text{Att}(\mathbf{s}_k^{(w_i)})\mathbf{s}_k^{(w_i)}, \tag{2.44}$$

where $\mathbf{s}_k^{(w_i)}$ denotes the $k$th sense embedding of $w_i$ and $S^{(w_i)}$ is the sense set of $w_i$. The attention term is as follows:

$$Att(\mathbf{s}_k^{(w_i)}) = \frac{\exp(\mathbf{w}'_c \cdot \hat{\mathbf{s}}_k^{(w_i)})}{\sum_{n=1}^{|S^{(w_i)}|} \exp(\mathbf{w}'_c \cdot \hat{\mathbf{s}}_n^{(w_i)})}, \tag{2.45}$$

where $\hat{\mathbf{s}}_k^{(w_i)}$ stands for the average of sememe embeddings $\mathbf{x}$, $\hat{\mathbf{s}}_k^{(w_i)} = \text{Avg}(\mathbf{x}^{(s_k)})$ and $\mathbf{w}'_c$ is the average of context word embeddings, $\mathbf{w}'_c = \text{Avg}(\mathbf{w}_j)(|j-i| \leq l, j \neq i)$.

This model shows a substantial advance in both word similarity and analogy tasks. Moreover, the introduction of sense embeddings can also be used in word sense disambiguation.

**Considering Document Information**. Word embedding methods like Skip-gram simply consider the context information within a window to learn word representation. However, the information in the whole document could help our word representation learning. Topical Word Embeddings (TWE) [42] introduces topic information generated by Latent Dirichlet Allocation (LDA) to help distinguish different meanings of a word. The model is defined to maximize the following average log probability:

$$\mathcal{O} = \frac{1}{N}\sum_{i=1}^{N} \sum_{-k \leq c \leq k, c \neq 0} \left(\log P(\mathbf{w}_{i+c}|\mathbf{w}_i) + \log P(\mathbf{w}_{i+c}|\mathbf{z}_i)\right), \tag{2.46}$$

where $\mathbf{w}_i$ is the word embedding and $\mathbf{z}_i$ is the topic embedding of $w_i$. Each word $w_i$ is assigned a unique topic, and each topic has a topic embedding. The topical word embedding model shows advantages of contextual word similarity and document classification tasks.

However, TWE simply combines the LDA with word embeddings and lacks statistical foundations. The LDA topic model needs numerous documents to learn semantically coherent topics. Reference [40] further proposes the TopicVec model, which encodes words and topics in the same semantic space. TopicVec outperforms TWE and other word embedding methods on text classification datasets. It can learn coherent topics on only one document which is not possible for other topic models.

#### 2.5.3.3 Word Representation with Hierarchical Structure

Human knowledge is in a hierarchical structure. Recently, many works also introduce a hierarchical structure of texts into word representation learning.

**Dependency-based Word Representation.** Continuous word embeddings are combinations of semantic and syntactic information. However, existing word representation models depend solely on linear contexts and show more semantic information than syntactic information. To make the embeddings show more syntactic information, the dependency-based word embedding [38] uses the dependency-based context. The dependency-based embeddings are less topical and exhibit more functional similarity than the original Skip-gram embeddings. It takes the information of dependency parsing tree into consideration when learning word representations. The contexts of a target word $w$ are the modifiers of this word, i.e., $(m_1, r_1), \ldots, (m_k, r_k)$, where $r_i$ is the type of the dependency relation between the head node and the modifier. When training, the model optimizes the probability of dependency-based contexts rather than neighboring contexts. This model gains some improvements on word similarity benchmarks compared with Skip-gram. Experiments also show that words with syntactic similarity are more similar in the vector space.

**Semantic Hierarchies.** Because of the linear substructure of the vector space, it is proven that word embeddings can make simple analogies. For example, the difference between `Japan` and `Tokyo` is similar to the difference between `China` and `Beijing`. But it has trouble identifying hypernym-hyponym relations since these relationships are complicated and do not necessarily have linear substructure.

To address this issue, [25] tries to identify hypernym-hyponym relationships using word embeddings. The basic idea is to learn a linear projection rather than simply use the embedding offset to represent the relationship. The model optimizes the projection as

$$\mathbf{M}^* = \arg\min_{\mathbf{M}} \frac{1}{N} \sum_{(i,j)} \|\mathbf{M}\mathbf{x}_i - \mathbf{y}_j\|_2, \tag{2.47}$$

where $\mathbf{x}_i$ and $\mathbf{y}_j$ are hypernym and hyponym embeddings.

To further increase the capability of the model, they propose to first cluster word pairs into several groups and learn a linear projection for each group. The linear projection can help identify various hypernym-hyponym relations.

### *2.5.4 Multilingual Word Representation*

There are thousands of languages in the world. In word level, how to represent words from different languages in a unified vector space is an interesting problem. The bilingual word embedding model [64] uses machine translation word alignments as constraining translational evidence and embeds words of two languages into a single vector space. The basic idea is (1) to initialize each word according to its aligned words in another language and (2) to constrain the distance between two languages during the training using translation pairs.

When learning bilingual word embeddings, it firstly trains source word embeddings. Then they use aligned sentence pairs to count the co-occurrence of source and target words. The target word embeddings can be initialized as

$$\mathbf{E}_{t-init} = \sum_{s=1}^{S} \frac{N_{ts} + 1}{N_t + S} \mathbf{E}_s, \tag{2.48}$$

where $\mathbf{E}_s$ and $\mathbf{E}_{t-init}$ are the trained embeddings of the source word and the initial embedding of the target word, respectively. $N_{ts}$ is the number of target words being aligned with source word. $S$ is all the possible alignments of word $t$. So $N_t + S$ normalizes the weights as a distribution. During the training, they jointly optimize the word embedding objective as well as the bilingual constraint. The constraint is defined as

$$\mathscr{L}_{cn \to en} = \|\mathbf{E}_{en} - N_{en \to cn} \mathbf{E}_{cn}\|^2, \tag{2.49}$$

where $N_{en \to cn}$ is the normalized align counts.

When given a lexicon of bilingual word pairs, [44] proposes a simple model that can learn bilingual word embeddings in a unified space. Based on the distributional geometric similarities of word vectors of two languages, this model learns a linear transformation matrix $\mathbf{T}$ that transforms the vector space of source language to that of the target language. The training loss is

$$\mathscr{L} = \|\mathbf{T}\mathbf{E}_s - \mathbf{E}_t\|^2, \tag{2.50}$$

where $\mathbf{E}_t$ is the word vector matrix of aligned words in target language.

However, this model performs badly when the seed lexicon is small. To tackle this limitation, some works introduce the idea of bootstrapping into bilingual word representation learning. Let's take [63] for example. In this work, in addition to monolingual word embedding learning and bilingual word embedding alignment based on seed lexicon, a new matching mechanism is introduced. The main idea of matching is to find the most probably matched source (target) word for each target (source) word and make their embeddings closer. Next, we explain the target-to-source matching process formally, and the source-to-target side is similar.

The target-to-source matching loss function is defined as

$$\mathscr{L}_{T2S} = -\log P\left(C^{(T)} | \mathbf{E}^{(S)}\right) = -\log \sum_{\mathbf{m}} P\left(C^{(T)}, \mathbf{m} | \mathbf{E}^{(S)}\right), \tag{2.51}$$

where $C^{(T)}$ denotes the target corpus and $\mathbf{m}$ is a latent variable specifying the matched source word for each target word. On independency assumption, it has

$$P\left(C^{(T)}, \mathbf{m}|\mathbf{E}^{(S)}\right) = \prod_{w_i^{(T)} \in C^{(T)}} P\left(w_i^{(T)}, \mathbf{m}|\mathbf{E}^{(S)}\right) = \prod_{i=1}^{|V^{(T)}|} P\left(w_i^{(T)}|w_{\mathbf{m}_i}^{(S)}\right)^{N_{w_i^{(T)}}}, \quad (2.52)$$

where $N_{w_i^{(T)}}$ is the number of $w_i^{(T)}$ occurrences in the target corpus. By training using Viterbi EM algorithm, this method can improve bilingual word embeddings on its own and address the limitation of a small seed lexicon.

### 2.5.5 *Task-Specific Word Representation*

In recent years, word representation learning has achieved great success and played a crucial role in NLP tasks. People find that word representation learning of the general field is still a limitation in a specific task and begin to explore the learning of task-specific word representation. In this section, we will take sentiment analysis as an example.

**Word Representation for Sentiment Analysis**. Most word representation methods capture syntactic and semantic information while ignoring sentiment of text. This is problematic because words with similar syntactic polarity but opposite sentiment polarity obtain closed word vectors. Reference [58] proposes to learn Sentiment-Specific Word Embeddings (SSWE) by integrating the sentiment information. An intuitive idea is to jointly optimize the sentiment classification model using word embeddings as its feature and SSWE minimizes the cross-entropy loss to achieve this goal. To better combine the unsupervised word embedding method and the supervised discriminative model, they further use the words in a window rather than a whole sentence to classify sentiment polarity. They propose the following ranking-based loss:

$$\mathscr{L}_r(t) = \max(0, 1 - \mathbf{1}_s(t) f_0^r(t) + \mathbf{1}_s(t) f_1^r(t)), \quad (2.53)$$

where $f_0^r$, $f_1^r$ are the predicted positive and negative scores. $\mathbf{1}_s(t)$ is an indicator function:

$$\mathbf{1}_s(t) = \begin{cases} 1 & \text{if t is positive,} \\ -1 & \text{if t is negative.} \end{cases} \quad (2.54)$$

This loss function only punishes the model when the model gives an incorrect result.

To get massive training data, they use distant-supervision technology to generate sentiment labels for a document. The increase of labeled data can improve the sentiment information in word embeddings. On sentiment classification tasks, sentiment embeddings outperform other strong baselines including SVM and other word embedding methods. SSWE also shows strong polarity consistency, where the closest words of a word are more likely to have the same sentiment polarity compared with existing word representation models. This sentiment specific word embedding method provides us a general way to learn task-specific word embeddings, which is to design a joint loss function and to generate massive labeled data automatically.

### *2.5.6 Time-Specific Word Representation*

The meaning of a word changes during the time. Analyzing the changing meaning of a word is an exciting topic in both linguistic and NLP research. With the rise of word embedding methods, some works [29, 35] use embeddings to analyze the change of words' meanings. They separate corpus into bins with respect to years to train time-specific word embeddings and compare embeddings of different time series to analyze the change of word semantics. This method is intuitive but has some problems. Dividing corpus into bins causes the data sparsity issue. The objective of word embedding methods is nonconvex so that different random initialization leads to different results, which makes comparing word embeddings difficult. Embeddings of a word in different years are in different semantic spaces and cannot be compared directly. Most work indirectly compares the meanings of a word in a different time by the changes of a word's closest words in the semantic space.

To address these issues, [4] proposes a dynamic Skip-gram model which connects several Bayesian Skip-gram models [6] using Kalman filters [33]. In this model, the embeddings of words in different periods could affect each other. For example, a word that appears in the 1990s' document can affect the embeddings of that word in the 1980s and 2000s. Moreover, it also trains the embedding in different periods by the whole corpus to reduce the sparsity issue. This model also puts all the embeddings into the same semantic space, which is a significant improvement against other methods and makes word embeddings in different periods comparable. Therefore, the change of word embeddings in this model is continuous and smooth. Experimental results show that the cosine distance between two words changes much more smoothly in this model than those models which simply divide the corpus into bins.

## 2.6 Evaluation

In recent years, various methods to embed words into a vector space have been proposed. Hence, it is essential to evaluate different methods. There are two general evaluations of word embeddings, including word similarity and word analogy. They both aim to check if the word distribution is reasonable. These two evaluations sometimes give different results. For example, CBOW achieves better performance on word similarity, whereas Skip-gram outperforms CBOW on word analogy. Therefore, which method to choose depends on the high-level application. Task-specific word embedding methods are usually designed for specific high-level tasks and achieve significant improvement on these tasks compared with baselines such as CBOW and Skip-gram. However, they only marginally outperform baselines on two general evaluations.

### *2.6.1 Word Similarity/Relatedness*

The dynamics of words are very complex and subtle. There is no static, finite set of relations that can describe all interactions between two words. It is also not trivial for downstream tasks to leverage different kinds of word relations. A more practical way is to assign a score to a pair of words to represent to what extent they are related. This measurement is called *word similarity*. When talking about the term *word similarity*, the precise meaning may vary a lot in different situations. There are several kinds of *similarity* that may be referred to in various literature.

**Morphological similarity**. Many languages including English define morphology. The same morpheme can have multiple surface forms according to the syntactical function. For example, the word `active` is an adjective and `activeness` is its noun version. The word `activate` is a verb and `activation` is its noun version. The morphology is an important dimension when considering the meaning and usage of words. It defines some relations between words from a syntactical view. Some relations are used in the Syntactic Word Relationship test set [43], including adjectives to adverbs, past tense, and so on. However, in many higher level applications and tasks, the words are often morphologically normalized by the base form (this process is also known as lemmatization). One widely used technique is the Porter stemming algorithm [49]. This algorithm converts `active`, `activeness`, `activate`, and `activation` to the same root format `activ`. By removing morphological features, the semantic meaning of words is more emphasized.

**Semantic Similarity**. Two words are semantically similar if they can express the same concept, or sense, like `article` and `document`. One word may have different senses, and each of its synonyms is associated with one or more of its senses. WordNet [46] is a lexical database that organizes the words as groups according to the senses. Each group of words is called a *synset*, which contains all synonymous words sharing the same specific sense. The words within the same synset are considered semantically similar. Words from two synsets that are linked by some certain relation (such as *hyponym*) are also considered semantically similar to some degree, like `bank(river)` and `bank` (`bank(river)` is the hyponym of `bank`).

**Semantic relatedness**. Most modern literature that considers word similarity refers to the semantic relatedness of words. Semantic relatedness is more general than semantic similarity. Words that are not semantically similar could still be related in many ways such as meronymy (`car` and `wheel`) or antonymy (`hot` and `cold`). Semantic relatedness often yields co-occurrence, but they are not equivalent. The syntactic structure could also yield co-occurrence. Reference [10] argues that distributional similarity is not an adequate proxy for semantic relatedness.

**Table 2.3** Datasets for evaluating word similarity/relatedness

| Dataset | Similarity Type |
|---|---|
| RG-65 [52] | Word Similarity |
| WordSim-353 [22] | Mixed |
| WordSim-353 REL [1] | Word Relatedness |
| WordSim-353 SIM [1] | Word Similarity |
| MTurk-287 [50] | Word Relatedness |
| SimLex-999 [31] | Word Similarity |

To evaluate the word representation system intrinsically, the most popular approach is to collect a set of word pairs and compute the correlation between human judgment and system output. So far, many datasets are collected and made public. Some datasets focus on the word similarity, such as RG-65 [52] and SimLex-999 [31]. Other datasets concern word relatedness, such as MTurk [50]. WordSim-353 [22] is a very popular dataset for word representation evaluation, but its annotation guideline does not differentiate similarity and relatedness very clearly. Reference [1] conducts another round of annotation based on WordSim-353 and generates two subsets, one for similarity and the other for relatedness. Some information about these datasets is summarized in Table 2.3.

To evaluate the similarity of two distributed word vectors, researchers usually select cosine similarity as an evaluation metric. The cosine similarity of word $w$ and word $v$ is defined as

$$\mathrm{sim}(w, v) = \frac{\mathbf{w} \cdot \mathbf{v}}{\|\mathbf{w}\| \|\mathbf{v}\|}. \tag{2.55}$$

When evaluating a word representation approach, the similarity of each word pair is computed in advance using cosine similarity. After that, Spearman's correlation coefficient $\rho$ is then used to evaluate the similarity between human annotator and word representation model as

$$\rho = 1 - \frac{6 \sum d_i^2}{n^3 - n}, \tag{2.56}$$

where a higher Spearman's correlation coefficient indicates they are more similar.

Reference [10] describes a series of methods based on WordNet to evaluate the similarity of a pair of words. After the comparison between the traditional WordNet-based methods and distributed word representations, [1] addresses that relatedness and similarity are two different concerns. They point out that WordNet-based methods perform better on similarity than on relatedness, while distributed word representation shows similar performance on both. A series of distributed word representations are compared on a wide variety of datasets in [56]. The state-of-the-art on both similarity and relatedness is achieved by distributed representation, without a doubt.

This evaluation method is simple and straightforward. However, as stated in [20], there are several problems with this evaluation. Since the datasets are small (less than 1,000 word pairs in each dataset), one system may yield many different scores on different partitions. Testing on the whole dataset makes it easier to overfit and hard to compute the statistical significance. Moreover, the performance of a system on these datasets may not be very correlated to its performance on downstream tasks.

The word similarity measurement can come in an alternative format, the TOEFL synonyms test. In this test, a cue word is given, and the test is required to choose one from four words that are the synonym of the cue word. The exciting part of this task is that the performance of a system could be compared with human beings. Reference [37] evaluates the system with the TOEFL synonyms test to address the knowledge inquiring and representing of LSA. The reported score is 64.4%, which is very close to the average rating of the human test-takers. On this test set with 80 queries, [54] reported a score of 72.0%. Reference [24] extends the original dataset with the help of WordNet and generates a new dataset[3] (named WordNet-based synonymy test) containing thousands of queries.

### 2.6.2 *Word Analogy*

Besides word similarity, the word analogy task is an alternative way to measure how well representations capture semantic meanings of words. This task gives three words $w_1$, $w_2$, and $w_3$, then it requires the system to predict a word $w_4$ such that the relation between $w_1$ and $w_2$ is the same as that between $w_3$ and $w_4$. This task is used since [43, 45] to exploit the structural relationships among words. Here, the word relations could be divided into two categories, including semantic relations and syntactic relations. This is a relatively novel method for word representation evaluation but quickly becomes a standard evaluation metric since the dataset is released. Unlike the TOEFL synonyms test, most words in this dataset are frequent across all kinds of the corpus, but the fourth word is chosen from the whole vocabulary instead of four options. This test favors distributed word representations because it emphasizes the structure of word space.

The comparison between different models on the word analogy task measured by accuracy could be found in [7, 56, 57, 61].

## 2.7 Summary

In this chapter, we first introduce word representation methods, including one-hot representation and various distributed representation methods. These classical methods are the important foundation of various NLP models, and meanwhile present the major concepts and mechanisms of word representation learning for the reader. Next, considering classical word representation methods often suffer from the word polysemy, we further introduce the effective contextualized word representation methods ELMo, to show the approach to capture complex word features across different linguistic contexts. As word representation methods are widely utilized in various downstream tasks, we then overview numerous extensions toward some representative directions and discuss how to adapt word representations for specific scenarios. Finally, we introduce several evaluation tasks of word representation, including word similarity and word analogy, which are the basic experimental settings for researching word representation methods.

[3] http://www.cs.cmu.edu/~dayne/wbst-nanews.tar.gz.

In the past decade, learning methods and applications of word representation have been studied in depth. Here we recommend some surveys and books on word representative learning for reading:

- Erk. Vector Space Models of Word Meaning and Phrase Meaning: A Survey [18].
- Lai et al. How to Generate a Good Word Embedding [36].
- Camacho et al. From Word to Sense Embeddings: A Survey on Vector Representations of Meaning [11].
- Ruder et al. A Survey of Cross-lingual Word Embedding Models [53].
- Bakarov. A Survey of Word Embeddings Evaluation Methods [2].

In the future, toward more effective word representation learning, some directions are requiring further efforts:

(1) **Utilizing More Knowledge**. Current word representation learning models focus on representing words based on plain textual corpora. In fact, besides rich semantic information in text, there are also various kinds of word-related information hidden in heterogeneous knowledge in the real world, such as visual knowledge, factual knowledge, and commonsense knowledge. Some preliminary explorations have attempted [59, 60] to utilize heterogeneous knowledge for learning better word representations, and these explorations indicate that utilizing more knowledge is a promising direction toward enhancing word representations. There remain open problems for further explorations.

(2) **Considering More Contexts**. As shown in this chapter, those word representation learning methods considering contexts can achieve more expressive word embeddings, which can grasp richer semantic information and further benefit downstream NLP tasks than classical distributed methods. Context-aware word representations have been systematically verified for their effectiveness in existing works [32, 48], and adopting those context-aware word representations has also become a necessary and mainstream operation for various NLP tasks. After BERT [15] has been proposed, language models pretrained on large-scale corpora have entered the public vision and their fine-tuning models have also achieved the state-of-the-art performance on specific NLP tasks. These new explorations based on large-scale textual corpora and pretrained fine-tuning language representation architectures indicate a promising direction to consider more contexts with more powerful representation architectures, and we will discuss them more in the next chapter.

(3) **Orienting Finer Granularity**. Polysemy is a widespread phenomenon for words. Hence, it is essential and meaningful to consider the finer granulated semantic information than the words themselves. As some linguistic knowledge bases have been developed, such as synonym-based knowledge bases WordNet [21] and sememe-based knowledge bases HowNet [17], we thus have ways to study the atomic semantics of words. The current work on word representations learning is coarse-grained, and mainly focuses on shallow semantics of the words themselves in text, and ignores the rich semantic information inside the words, which is also an important resource for achieving better word embeddings. Reference [28] explores to inject finer granulated atomic semantics of words into word representations and performs much better language understanding. Although these explorations are still preliminary, orienting finer granularity of word representations is important. In the next chapter, we will also introduce more details in this part.

In the past decade, learning methods and applications of distributed representation have been studied in depth. Because of its efficiency and effectiveness, lots of task-specific models have been proposed for various tasks. Word representation learning has become a popular and important topic in NLP. However, word representation learning is still challenging due to its ambiguity, data sparsity, and interpretability. In recent years, word representation learning has been no longer studied in isolation, but explored together with sentence or document representation learning using pretrained language models. Readers are recommended to refer to the following chapters to further learn the integration of word representations in other scenarios.

# Chapter 3
# Compositional Semantics

**Abstract** Many important applications in NLP fields rely on understanding more complex language units such as phrases, sentences, and documents beyond words. Therefore, compositional semantics has remained a core task in NLP. In this chapter, we first introduce various models for binary semantic composition, including additive models and multiplicative models. After that, we present various typical models for N-ary semantic composition including recurrent neural network, recursive neural network, and convolutional neural network.

## 3.1 Introduction

From the previous chapter, following the distributed hypothesis, one could project the semantic meaning of a word into a low-dimensional real-valued vector according to its context information, which is named as word vectors. Here comes a further problem: how to compress a higher semantic unit into a vector or other kinds of mathematical representations like a matrix or a tensor. In other words, using representation learning to model a semantic composition function remains an unsolved but surging research topic recently.

Compositionality enables natural languages to construct complex semantic meanings from the combinations of simpler semantic elements. This property is often captured with the following principle: the semantic meaning of a whole is a function of the semantic meanings of its several parts. Therefore, the semantic meanings of complex structures will depend on how their semantic elements combine.

Here we express the composition of two semantic units, which are denoted as $\mathbf{u}$ and $\mathbf{v}$, respectively, and the most intuitive way to define the joint representation could be formulated as follows:

$$\mathbf{p} = f(\mathbf{u}, \mathbf{v}), \tag{3.1}$$

where $\mathbf{p}$ corresponds to the representation of the joint semantic unit $(\mathbf{u}, \mathbf{v})$. It should be noted that here $\mathbf{u}$ and $\mathbf{v}$ could denote words, phrases, sentences, paragraphs, or even higher level semantic units.

However, given the representations of two semantic constituents, it is not enough to derive their joint embeddings with the lack of syntactic information. For instance, although the phrase `machine learning` and `learning machine` have the same vocabulary, they contain different meanings: `machine learning` refers to a research field in artificial intelligence while `learning machine` means some specific learning algorithms. This phenomenon stresses the importance of syntactic and order information in a compositional sentence. Reference [12] takes the role of syntactic and order information into consideration and suggests a further refinement of the above principle: the meaning of a whole is a function of the meaning of its several parts and the way they are syntactically combined. Therefore, the composition function in Eq. (3.1) is redefined to combine the syntactic relationship rule $\mathscr{R}$ between the semantic units $\mathbf{u}$ and $\mathbf{v}$:

$$\mathbf{p} = f(\mathbf{u}, \mathbf{v}, \mathscr{R}), \tag{3.2}$$

where $\mathscr{R}$ denotes the syntactic relationship rule between two constituent semantic units.

Unfortunately, even this formulation may not be fully adequate. Therefore, [7] claims that the meaning of a whole is greater than the meanings of its several parts. It implies that people may suffer from the problem of constructing complex meanings rather than simply understanding the meanings of several parts and their syntactic relations. In real language composition, in different contexts, the same sentence could have different meanings, which means that some sentences are hard to understand without any background information. For example, the sentence `Tom and Jerry is one of the most popular comedies in that style.` needs two main backgrounds: Firstly, `Tom and Jerry` is a special noun phrase or knowledge entity which indicates a cartoon comedy, rather than two ordinary people. The other prior knowledge should be `that style`, which needs further explanation in the previous sentences. Hence, a full understanding of the compositional semantics needs to take existing knowledge into account. Here, the argument $\mathscr{K}$ is added into the composition function, incorporating knowledge information as a prior in the compositional process:

$$\mathbf{p} = f(\mathbf{u}, \mathbf{v}, \mathscr{R}, \mathscr{K}), \tag{3.3}$$

where $\mathscr{K}$ represents the background knowledge.

Reference [4] claims that we should ask for the meaning of a word in isolation but only in the context of a statement. That is, the meaning of a whole is constructed from its parts, and the meanings of the parts are meanwhile derived from the whole. Moreover, compositionality is a matter of degree rather than a binary notion. Linguistic structures range from fully compositional (e.g., black hair), to partly compositional syntactically fixed expressions, (e.g., take advantage), in which the constituents can still be assigned separate meanings, and non-compositional idioms (e.g., kick the bucket) or multi-word expressions (e.g., by and large), whose meaning cannot be distributed across their constituents [11].

From the above three equations formulating composition function, it could be concluded that composition could be viewed as a specific binary operation but beyond this. The syntactic message could help to indicate a particular approach while background knowledge helps to explain some obscure words or specific context-dependent entities such as pronouns. Beyond binary compositional operations, one could build the sentence-level composition by applying binary composition operations recursively. In this chapter, we will first explain some sorts of basic binary composition functions in both the semantic vector space and matrix-vector space. After, we will climb up to more complex composition scenarios and introduce several approaches to model sentence-level composition.

## 3.2 Semantic Space

### *3.2.1 Vector Space*

In general, the central task in semantic representation is projecting words from an abstract semantic space to a mathematical low-dimensional space. As introduced in the previous chapters, to make the transformation reasonable, the purpose is to maintain the word similarity in this new projected space. In other words, the more similar the words are, the closer their vectors should be. For instance, we hope the word vectors $\mathbf{w}(book)$ and $\mathbf{w}(magazine)$ are close while the word vectors $\mathbf{w}(apple)$ and $\mathbf{w}(computer)$ are far away. In this chapter, we will introduce several widely used typical semantic vector space including one-hot representation, distributed representation, and distributional representation.

### *3.2.2 Matrix-Vector Space*

Despite the wide use of semantic vector spaces, an alternative semantic space is proposed to be a more powerful and general compositional semantic framework. Different from conventional vector spaces, matrix-vector semantic space utilizes a matrix to represent the word meaning rather than a skinny vector. The motivation behind this is when modeling the semantic meaning under a specific context, one is wondering not only what is the meaning of each word, but also the holistic meaning of the whole sentence. Thus, we concern about the semantic transformation between adjacent words inside each sentence. However, the semantic vector space could not characterize the semantic transformation of one word on the others explicitly.

Driven by the idea of modeling semantic transformation, some researchers have proposed to use a matrix to represent the transformation operation of one word on the others. Different from those vector space models, it could incorporate some structural information like the word order and syntax composition.

## 3.3 Binary Composition

The goal is to construct vector representations for phrases, sentences, paragraphs, and documents. Without loss of generality, we assume that each constituent of a phrase (sentence, paragraph, or document) is embedded into a vector which will be subsequently combined in some way to generate a representation vector for the phrase (sentence, paragraph, or document).[1]

In this section, we focus on binary composition. We will take phrases consisting of a head and a modifier or complement as an example. If we cannot model the binary composition (or phrase representation), there is little hope that we can construct more complex compositional representations for sentences or even documents. Therefore, given a phrase such as "machine learning" and the vectors **u** and **v** representing the constituents "machine" and "learning", respectively, we aim to produce a representation vector **p** of the whole phrase. Let the hypothetical vectors for `machine` and `learning` be [0, 3, 1, 5, 2] and [1, 4, 2, 2, 0], respectively. This simplified semantic space will serve to illustrate examples of the composition functions which we consider in this section.

The fundamental problem of semantic composition modeling in representing a two-word phrase is designing a primitive composition function as a binary operator. Based on this function, one could apply it on a word sequence recursively and derive sentence-level composition. Here a word sequence could be any level of the semantic units, such as a phrase, a sentence, a paragraph, a knowledge entity, or even a document.

From the previous section, one of the basic formulae is to formulate semantic composition $f$ in the following equation:

$$\mathbf{p} = f(\mathbf{u}, \mathbf{v}, \mathscr{R}, \mathscr{K}), \tag{3.4}$$

where $\mathbf{u}$, $\mathbf{v}$ denote the representations of the constituent parts in this semantic unit, $\mathbf{p}$ denotes the joint representation, $R$ indicates the relationship while $\mathscr{K}$ indicates the necessary background knowledge. The expression defines a wide class of composition functions. For easier discussion, we give some appropriate constraints to narrow the space of our considering function. First, we will ignore the background knowledge $\mathscr{K}$ to explore what can be achieved without any utilization of background or world knowledge. Second, for the consideration of the syntactic relation $\mathscr{R}$, we can proceed by investigating only one relation at a time. And then we can remove any explicit dependence on $\mathscr{R}$ which allows us to explore any possible distinct composition function for various syntactic relations. That is, we simplify the formula $\mathbf{p} = f(\mathbf{u}, \mathbf{v})$ by simply ignoring the background knowledge and relationship.

[1]Note that, the problem of combining semantic vectors of small units to make a representation for a multi-word sequence is different from the problem of incorporating information about multi-word contexts into a distributional representation for a single target word.

In recent years, modeling the binary composition function is a well-studied but still challenging problem. There are mainly two perspectives toward this question, including the additive model and the multiplicative model.

### *3.3.1 Additive Model*

The additive model has a constraint in which it assumes that **p**, **u**, and **v** lie in the same semantic space. This essentially means that all syntactic types have the same dimension. One of the simplest ways is to directly use the sum to represent the joint representation:

$$\mathbf{p} = \mathbf{u} + \mathbf{v}. \tag{3.5}$$

According to Eq. (3.5), the sum of the two vectors representing `machine` and `learning` would be $\mathbf{w}(machine) + \mathbf{w}(learning) = [1, 7, 3, 7, 2]$. It assumes that the composition of different constituents is a symmetric function of them; in other words, it does not consider the order of constituents. Although having lots of drawbacks such as lack of the ability to model word orders and absence from background syntactic or knowledge information, this approach still provides a relatively strong baseline [9].

To overcome the word order issue, one easy variant is applying a weighted sum instead of uniform weights. This is to say, the composition has the following form:

$$\mathbf{p} = \alpha\mathbf{u} + \beta\mathbf{v}, \tag{3.6}$$

where $\alpha$ and $\beta$ correspond to different weights for two vectors. Under this setting, two sequences $(u, v)$ and $(v, u)$ have different representations, which is consistent with real language phenomena. For example, "machine learning" and "learning machine" have different meanings which requires different representations. In this setting, we could give greater emphasis to heads than other constituents. As an example, if we set $\alpha$ to 0.3 and $\beta$ to 0.7, the $0.3 \times \mathbf{w}(machine) = [0, 0.9, 0.3, 1.5, 0.6]$ and $0.7 \times \mathbf{w}(learning) = [0.7, 2.8, 1.4, 1.4, 0]$, and "machine learning" is represented by their addition $0.3 \times \mathbf{w}(machine) + 0.7 \times \mathbf{w}(learning) = [0.7, 3.6, 1.7, 2.9, 0.6]$.

However, this model could not consider prior knowledge and syntax information. To incorporate prior information into the additive model, one method combines nearest neighborhood semantics into composition, deriving

$$\mathbf{p} = \mathbf{u} + \mathbf{v} + \sum_{i=1}^{K} \mathbf{n}_i, \tag{3.7}$$

where $n_1, n_2, \ldots, n_K$ denote all semantic neighbors of **v**. Therefore, this method could ensemble all synonyms of the component as a smoothing factor into composition function, which reduces the variance of language. For example, if in the composition of "machine" and "learning", the chosen neighbor is "optimizing", with $\mathbf{w}(optimizing) = [1, 5, 3, 2, 1]$, then this leads to the situation that the representation of "machine learning" becomes $\mathbf{w}(machine) + \mathbf{w}(learning) + \mathbf{w}(optimizing) = [2, 12, 6, 9, 3]$.

Since the joint representations of one additive model still lie in the same semantic space with their original component vectors, it is natural to conduct cosine similarity to measure their semantic relationships. Thus, under a naive additive model, we have the following similarity equation:

$$s(\mathbf{p}, \mathbf{w}) = \frac{\mathbf{p} \cdot \mathbf{w}}{\|\mathbf{p}\| \cdot \|\mathbf{w}\|} = \frac{(\mathbf{u} + \mathbf{v})\mathbf{w}}{\|\mathbf{u} + \mathbf{v}\| \|\mathbf{w}\|} \tag{3.8}$$

$$= \frac{\|\mathbf{u}\|}{\|\mathbf{u} + \mathbf{v}\|} s(\mathbf{u}, \mathbf{w}) + \frac{\|\mathbf{v}\|}{\|\mathbf{u} + \mathbf{v}\|} s(\mathbf{v}, \mathbf{w}), \tag{3.9}$$

where $\mathbf{w}$ denotes any other word in the vocabulary and $s$ indicates the similarity function. From derivation ahead, it could be concluded that this composition function composes both magnitude and directions of two component vectors. In other words, if one vector dominates the magnitude, it will also dominate the similarity. Furthermore, we have

$$\|\mathbf{p}\| = \|\mathbf{u} + \mathbf{v}\| \le \|\mathbf{u}\| + \|\mathbf{v}\|. \tag{3.10}$$

This lemma suggests that the semantic unit with a deeper-rooted parsing tree could determine the joint representation when combining with a shallow unit. Because the deeper the semantic unit is, the larger the magnitude it has.

Moreover, incorporating geometry insight, we can observe that the additive model builds a more solid understanding of semantic composition. Supposing that our component vectors are $\mathbf{u}$ and $\mathbf{v}$, the additive model aims to project them to $\mathbf{x}$ and $\mathbf{y}$, where $\mathbf{x}$ follows the direction of $\mathbf{u}$ while $\mathbf{y}$ is orthogonal to $\mathbf{u}$. The following figure could clearly illustrate this issue (Fig. 3.1).

**Fig. 3.1** An illustration of the additive model

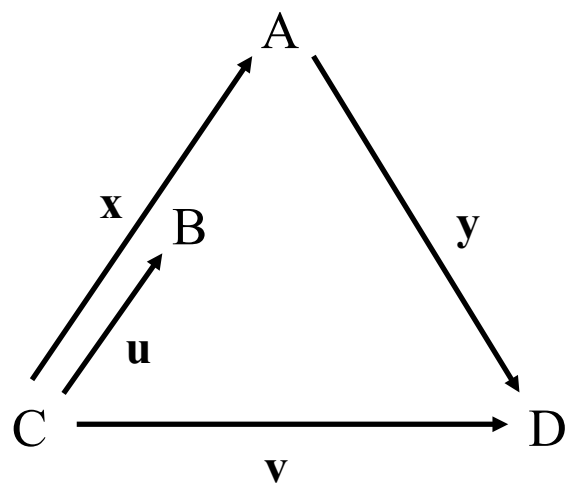

From the figure, the vector $\mathbf{x}$ and the vector $\mathbf{y}$ could be represented as

$$\mathbf{x} = \frac{\mathbf{u} \cdot \mathbf{v}}{\mathbf{u} \cdot \mathbf{u}} \cdot \mathbf{u},$$
$$\mathbf{y} = \mathbf{v} - \mathbf{x} = \mathbf{v} - \frac{\mathbf{u} \cdot \mathbf{v}}{\mathbf{u} \cdot \mathbf{u}} \cdot \mathbf{u}. \tag{3.11}$$

Then, using the linear combination of these two new vectors $\mathbf{x}, \mathbf{y}$ yields a new additive model:

$$\mathbf{p} = \alpha \mathbf{x} + \beta \mathbf{y} \tag{3.12}$$
$$= \alpha \frac{\mathbf{u} \cdot \mathbf{v}}{\mathbf{u} \cdot \mathbf{u}} \cdot \mathbf{u} + \beta \left( \mathbf{v} - \frac{\mathbf{u} \cdot \mathbf{v}}{\mathbf{u} \cdot \mathbf{u}} \cdot \mathbf{u} \right) \tag{3.13}$$
$$= (\alpha - \beta) \cdot \frac{\mathbf{u} \cdot \mathbf{v}}{\mathbf{u} \cdot \mathbf{u}} \cdot \mathbf{u} + \beta \mathbf{v}. \tag{3.14}$$

Furthermore, using cosine similarity measurement, the relationship could be written as follows:

$$s(\mathbf{p}, \mathbf{w}) = \frac{|\alpha - \beta|}{|\alpha|} s(\mathbf{u}, \mathbf{w}) + \frac{|\beta|}{|\alpha|} s(\mathbf{v}, \mathbf{w}). \tag{3.15}$$

From similarity measurement derivation, it is indicated that with this projection method, the composition similarity could be viewed as a linear combination of the similarities of two components, which means that combining semantic units with different semantic depths, the deeper one will not dominate the representation.

### *3.3.2 Multiplicative Model*

Though the additive model achieves great success in semantic composition, the simplification it adopted may be too restrictive because it assumes all words, phrases, sentences, and documents are substantially similar enough to be represented in a unified semantic space. Different from the additive model which regards composition as a simple linear transformation, the multiplicative model aims to make higher order interaction. Among all models from this perspective, the most intuitive approach tried to apply the pair-wise product as a composition function approximation. In this method, the composition function is shown as the following:

$$\mathbf{p} = \mathbf{u} \odot \mathbf{v}, \tag{3.16}$$

where, $\mathbf{p}_i = \mathbf{u}_i \cdot \mathbf{v}_i$, which implies each dimension of the output only depends on the corresponding dimension of two input vectors. However, similar to the simplest additive model, this model is also suffering from the lack of the ability to model word order, and the absence from background syntactic or knowledge information.

In the additive model, we have $\mathbf{p} = \alpha\mathbf{u} + \beta\mathbf{v}$ to alleviate the word order issue. Note that here $\alpha$ and $\beta$ are two scalars, which could be easily changed to two matrices. Therefore, the composition function could be represented as

$$\mathbf{p} = \mathbf{W}_\alpha \cdot \mathbf{u} + \mathbf{W}_\beta \cdot \mathbf{v}, \tag{3.17}$$

where $\mathbf{W}_\alpha$ and $\mathbf{W}_\beta$ are matrices which determine the importance of $\mathbf{u}$ and $\mathbf{v}$ to $\mathbf{p}$. With this expression, the composition could be more expressive and flexible although much harder to train.

Generalizing multiplicative model ahead, another approach is to utilize tensors as multiplicative descriptors and the composition function could be viewed as

$$\mathbf{p} = \overrightarrow{\mathbf{W}} \cdot \mathbf{uv}, \tag{3.18}$$

where $\overrightarrow{\mathbf{W}}$ denotes a 3-order tensor, i.e., the formula above could be written as $\mathbf{p}_k = \sum_{i,j} \mathbf{W}_{ijk} \cdot \mathbf{u}_i \cdot \mathbf{v}_j$. Hence, this model makes that each element of $\mathbf{p}$ could be influenced by all elements of both $\mathbf{u}$ and $\mathbf{v}$, with a relationship of linear combination by assigning each $(i, j)$ a unique weight.

Starting from this simple but general baseline, some researchers proposed to make the function not symmetric to consider word order in the sequence. Paying more attention to the first element, the composition function could be

$$\mathbf{p} = \overrightarrow{\mathbf{W}} \cdot \mathbf{uuv}, \tag{3.19}$$

where $\overrightarrow{\mathbf{W}}$ denotes a 4-order tensor. This method could be understood as replacing linear transformation of $\mathbf{u}$ and $\mathbf{v}$ to a quadratic in $\mathbf{u}$ asymmetrically. So this is a variant of the tensor multiplicative compositional model.

Different from expanding a simple multiplicative model to complex ones, other kinds of approaches are proposed to reduce the parameter space. With the reduction of parameter size, people could make compositions much more efficient rather than have an $O(n^3)$ time complexity in the tensor-based model. Thus, some compression techniques could be applied in the original tensor model. One representative instance is the circular convolution model, which could be shown as

$$\mathbf{p} = \mathbf{u} \circledast \mathbf{v}, \tag{3.20}$$

where $\circledast$ represents the circular convolution operation with the following definition:

$$\mathbf{p}_i = \sum_j \mathbf{u}_j \cdot \mathbf{v}_{i-j}. \tag{3.21}$$

If we assign each pair with unique weights, the composition function will be

$$\mathbf{p}_i = \sum_j \mathbf{W}_{ij} \cdot \mathbf{u}_j \cdot \mathbf{v}_{i-j}. \tag{3.22}$$

Note that the circular convolution model could be viewed as a special instance of a tensor-based composition model. If we write the circular convolution in the tensor form, we have $\mathbf{W}_{ijk} = 0$, where $k \neq i + j$. Thus, the parameter number could be reduced from $n^3$ to $n^2$, while maintaining the interactions between each pair of dimensions in the input vectors.

Both in the additive and multiplicative models, the basic condition is all components lie in the same semantic space as the output. Nevertheless, different modeling types of words in different semantic spaces could bring us a different perspective. For instance, given $(u, v)$, the multiplicative model could be reformulated as

$$\mathbf{p} = \mathbf{W} \cdot (\mathbf{u} \cdot \mathbf{v}) = \mathbf{U} \cdot \mathbf{v}. \tag{3.23}$$

This implies that each left unit could be treated as an operation on the representation of the right one. In other words, each remaining unit could be formulated as a transformation matrix, while the right one should be represented as a semantic vector. This argument could be meaningful, especially for some kinds of phrase compositions. Reference [2] argues that for ADJ-NOUN phrases, the joint semantic information could be viewed as the conjunction of the semantic meanings of two components. Given a phrase `red car`, its semantic meaning is the conjunction of all red things and all different kinds of cars. Thus, `red` could be formulated as an operator on the vector of `car`, deriving the new semantic vector, which expressed the meaning of `red car`. These observations lead to another genre of semantic compositional modeling: semantic matrix-composition space.

## 3.4 N-Ary Composition

In real-world NLP tasks, the input is usually a sequence of multiple words rather than just a pair of words. Therefore, besides designing a suitable binary compositional operator, the order to apply binary operations is also important. In this section, we will introduce three mainstream strategies in N-ary composition by taking language modeling as an example.

To illustrate the language modeling task more clearly, the composition problem to model a sentence or even a document could be formulated as

*Given a sentence/document consisting of a word sequence* $\{w_0, w_1, w_2, \ldots, w_n\}$, *we aim to design following functions to obtain the joint semantic representation of the whole sentence/document:*

1. A semantic representation method like semantic vector space or compositional matrix space.
2. A binary compositional operation function $f(u, v)$ like we introduced in the previous sections. Here the input $u$ and $v$ denote the representations of two constitute semantic units, while the output is also the representation in the same space.

3. A sequential order to apply the binary function in step 2. To describe in detail, we could use a bracket to identify the order to apply the composition function. For instance, we could use $((w_1, w_2), w_3)$ to represent the sequential order from beginning to end.

In this section, we will introduce several systematic strategies to model sentence semantics by describing the solutions for the three problems above. We will classify the methods by word-level order: sequential order, recursive order (following parsing trees), and convolution order.

### *3.4.1 Recurrent Neural Network*

To design orders to apply binary compositional functions, the most intuitive method is utilizing sequentiality. Namely, the sequence order should be $s_n = (s_{n-1}, w_n)$, where $s_{n-1}$ is the order of the first $n-1$ words. Motivated by this thought, the neural network model used is the Recurrent Neural Network (RNN).

An RNN applies the composition function sequentially and derives the representations of hidden semantic units. Based on these hidden semantic units, we could use them on some specific NLP tasks like sentiment analysis or text classification. Also, note that the basic RNN only utilizes the sequential information from head to tail of a sentence/document. To improve its representation ability, the RNN could be enhanced as bi-directional RNN by considering sequential and reverse-sequential information.

After deciding sequential order to model sentence-level semantics, the next question is determining the binary composition functions. In detail, supposing that $\mathbf{h}_t$ denotes the representation of the first $t$ words and $\mathbf{w}_t$ represents the $t$th word, the general composition could be formulated as

$$\mathbf{h_t} = f(\mathbf{h}_{t-1}, \mathbf{x}_t), \tag{3.24}$$

where $f$ is a well-designed binary composition function.

From the definition of the RNN, the composition function could be formulated as follows:

$$\mathbf{h}_t = \tanh(\mathbf{W}_1\mathbf{h}_{t-1} + \mathbf{W}_2\mathbf{w}_t), \tag{3.25}$$

where $\mathbf{W}_1$ and $\mathbf{W}_2$ are two weighted matrices.

We could see that here we use a matrix-weighted summation to represent binary semantic composition:

$$\mathbf{p} = \mathbf{W}_\alpha\mathbf{u} + \mathbf{W}_\beta\mathbf{v}. \tag{3.26}$$

**LSTM.** Since the raw RNN only utilizes the simple tangent function, it is hard to obtain the long-term dependency of a long sentence/document. Reference [5] reinvents Long Short-Term Memory (LSTM) networks to strengthen the ability to model long-term semantic dependency in RNN. In detail, the composition function of the LSTM allows information from previous layers to flow directly to their following layers. The composition function could be defined as

$$\mathbf{f}_t = \text{Sigmoid}(\mathbf{W}_f^h \mathbf{h}_{t-1} + \mathbf{W}_f^x \mathbf{x}_t + \mathbf{b}_f), \tag{3.27}$$

$$\mathbf{i}_t = \text{Sigmoid}(\mathbf{W}_i^h \mathbf{h}_{t-1} + \mathbf{W}_i^x \mathbf{x}_t + \mathbf{b}_i), \tag{3.28}$$

$$\mathbf{o}_t = \text{Sigmoid}(\mathbf{W}_o^h \mathbf{h}_{t-1} + \mathbf{W}_o^x \mathbf{x}_t + \mathbf{b}_o), \tag{3.29}$$

$$\hat{\mathbf{c}}_t = \tanh(\mathbf{W}_c^h \mathbf{h}_{t-1} + \mathbf{W}_c^x \mathbf{x}_t + \mathbf{b}_c), \tag{3.30}$$

$$\mathbf{c}_t = \mathbf{f}_t \odot \mathbf{c}_{t-1} + \mathbf{i}_t \odot \hat{\mathbf{c}}_t, \tag{3.31}$$

$$\mathbf{h}_t = \mathbf{o}_t \odot \mathbf{c}_t. \tag{3.32}$$

**Variants of LSTM.** To simplify LSTM and obtain more efficient algorithms, [3] proposes to utilize a simple but comparable RNN architecture, named Gated Recurrent Unit (GRU). Compared with LSTM, GRU has fewer parameters, which bring higher efficiency. The composition function is showed as

$$\mathbf{z}_t = \text{Sigmoid}(\mathbf{W}_z^h \mathbf{h}_{t-1} + \mathbf{W}_z^x \mathbf{x}_t + \mathbf{b}_z), \tag{3.33}$$

$$\mathbf{r}_t = \text{Sigmoid}(\mathbf{W}_r^h \mathbf{h}_{t-1} + \mathbf{W}_r^x \mathbf{x}_t + \mathbf{b}_r), \tag{3.34}$$

$$\hat{\mathbf{h}}_t = \tanh(\mathbf{W}_h(\mathbf{r}_t \odot \mathbf{h}_{t-1}) + \mathbf{W}_h^x \mathbf{x}_t + \mathbf{b}_h), \tag{3.35}$$

$$\mathbf{h}_t = (1 - \mathbf{z}_t) \odot \mathbf{h}_{t-1} + \mathbf{z}_t \odot \hat{\mathbf{h}}_t. \tag{3.36}$$

### *3.4.2 Recursive Neural Network*

Besides the recurrent neural network, another strategy to apply binary compositional function follows a parsing tree instead of sequential word order. Based on this philosophy, [15] proposes a recursive neural network to model different levels of semantic units. In this subsection, we will introduce some algorithms following the recursive parsing tree with different binary compositional functions.

Since all the recursive neural networks are binary trees, the basic problem we need to consider is how to derive the representation of the father component on the tree given its two children semantic components. Reference [15] proposes a recursive matrix-vector model (MV-RNN) which captures constituent parsing tree structure information by assigning a matrix-vector representation for each constituent. The vector captures the meaning of the constituent itself, and the matrix represents how it modifies the meaning of the word it combines with. Suppose we have two children components $a$, $b$ and their father component $p$, the composition can be formulated as follows:

$$\mathbf{p} = f_{vec}(a, b) = g\left(\mathbf{W}_1 \begin{bmatrix} \mathbf{Ba} \\ \mathbf{Ab} \end{bmatrix}\right), \tag{3.37}$$

$$\mathbf{P} = f_{matrix}(a, b) = \mathbf{W}_2 \begin{bmatrix} \mathbf{A} \\ \mathbf{B} \end{bmatrix}, \tag{3.38}$$

where $\mathbf{a}, \mathbf{b}, \mathbf{p}$ are the embedding vectors for each component and $\mathbf{A}, \mathbf{B}, \mathbf{P}$ are the matrices, $\mathbf{W}_1$ is a matrix that maps the transformed words into another semantic space, the element-wise function $g$ is an activation function, and $\mathbf{W}_2$ is a matrix that maps the two matrices into one combined matrix $\mathbf{P}$ with the same dimension. The whole process is illustrated in Fig. 3.2. And then MV-RNN selects the highest node of the path in the parse tree between the two target entities to represent the input sentence.

In fact, the composition operation used in the above recursive network is similar to an RNN unit introduced in the previous subsection. And the RNN unit here can be replaced by LSTM units or GRU units. Reference [16] proposes two types of tree-structured LSTMs including the Child-Sum Tree-LSTM and the N-ary Tree-LSTM to capture constituent or dependency parsing tree structure information. For the Child-Sum Tree-LSTM, given a tree, let $C(t)$ denote the children set of the node $t$. Its transition equations are defined as follows:

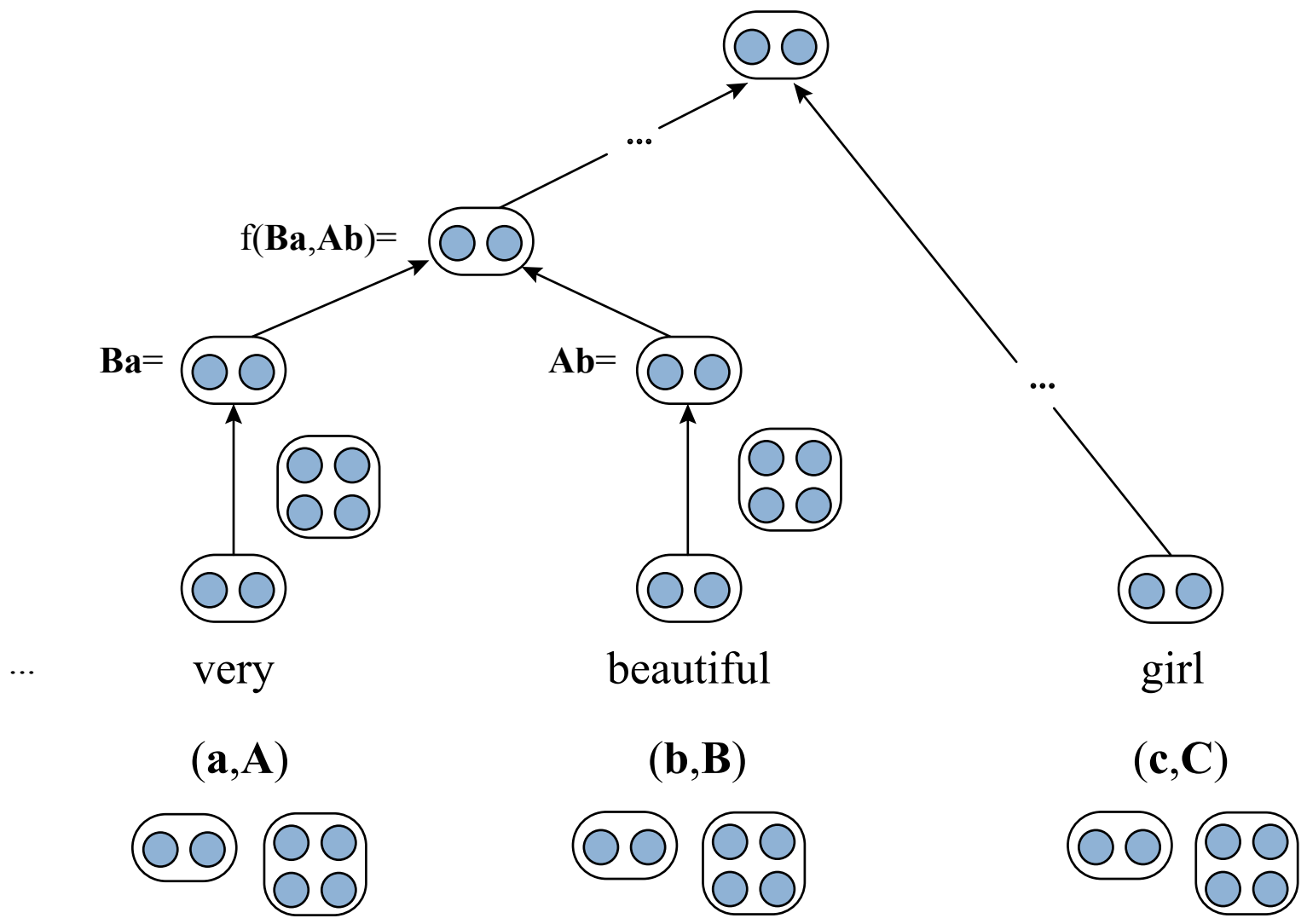

**Fig. 3.2** The architecture of the matrix-vector recursive encoder

$$\hat{\mathbf{h}}_t = \sum_{k \in C(t)} \mathbf{h}_k, \tag{3.39}$$

$$\mathbf{i}_t = \text{Sigmoid}(\mathbf{W}^{(i)}\mathbf{w}_t + \mathbf{U}^i\hat{\mathbf{h}}_t + \mathbf{b}^{(i)}), \tag{3.40}$$

$$\mathbf{f}_{tk} = \text{Sigmoid}(\mathbf{W}^{(f)}\mathbf{w}_t + \mathbf{U}^f\hat{\mathbf{h}}_k + \mathbf{b}^{(f)}) \;\; (k \in C(t)), \tag{3.41}$$

$$\mathbf{o}_t = \text{Sigmoid}(\mathbf{W}^{(o)}\mathbf{w}_t + \mathbf{U}^o\hat{\mathbf{h}}_t + \mathbf{b}^{(o)}), \tag{3.42}$$

$$\mathbf{u}_t = \tanh(\mathbf{W}^{(u)}\mathbf{w}_t + \mathbf{U}^u\hat{\mathbf{h}}_t + \mathbf{b}^{(u)}), \tag{3.43}$$

$$\mathbf{c}_t = \mathbf{i}_t \odot \mathbf{u}_t + \sum_{k \in C(t)} \mathbf{f}_{tk} \odot \mathbf{c}_{t-1}, \tag{3.44}$$

$$\mathbf{h}_t = \mathbf{o}_t \odot \tanh(\mathbf{c}_t). \tag{3.45}$$

The N-ary Tree-LSTM has similar transition equations as the Child-Sum Tree-LSTM. The only difference is that it limits the tree structures to have at most N branches.

### *3.4.3 Convolutional Neural Network*

Reference [6] proposes to embed an input sentence using a Convolutional Neural Network (CNN) which extracts local features by a convolution layer and combines all local features via a max-pooling operation to obtain a fixed-sized vector for the input sentence.

Formally, the convolution operation is defined as a matrix multiplication between a sequence of vectors, a convolution matrix $\mathbf{W}$, and a bias vector $\mathbf{b}$ with a sliding window. Let us define the vector $\mathbf{q}_i$ as the concatenation of the subsequence of input representations in the $i$th window, we have

$$\mathbf{h}_j = \max_i [f(\mathbf{W}\mathbf{q}_i + \mathbf{b})]_j, \tag{3.46}$$

where $f$ indicates a nonlinear function such as sigmoid or tangent function, and $\mathbf{h}$ indicates the final representation of the sentence.

## 3.5 Summary

In this chapter, we first introduce the semantic space for compositional semantics. Afterwards, we take phrase representation as an example to introduce various models for binary semantic composition, including additive models and multiplicative models. Finally, we introduce typical models for N-ary semantic composition including recurrent neural network, recursive neural network, and convolutional neural network. Compositional semantics allows languages to construct complex meanings from the combinations of simpler elements, and its binary semantic composition and N-ary semantic composition is the foundation of multiple NLP tasks including sentence representation, document representation, relational path representation, etc. We will give a detailed introduction to these scenarios in the following chapters.

For further understanding of compositional semantics, there are also some recommended surveys and books:

- Pelletier et al., The principle of semantic compositionality [13].
- Jeff et al., Composition in distributional models of semantics [10].

For better modeling compositional semantics, some directions require further efforts in the future:

(1) **Neurobiology-inspired Compositional Semantics**. What is the neurobiology for dealing with compositional semantics in human language? Recently, [14] finds that the human combinatory system is related to rapidly peaking activity in the left anterior temporal lobe and later engagement of the medial prefrontal cortex. The analysis of how language builds meaning and lays out directions in neurobiological research may bring some instructive reference for modeling compositional semantics in representation learning. It is valuable to design novel compositional forms inspired by recent neurobiological advances.

(2) **Combination of Symbolic and Distributed Representation**. Human language is inherently a discrete symbolic representation of knowledge. However, we represent the semantics of discrete symbols with distributed/distributional representations when dealing with natural language in deep learning. Recently, there are some approaches such as neural module networks [1] and neural symbolic machine [8] attempting to consider discrete symbols in neural networks. How to take advantage of these symbolic neural models to represent the composition of semantics is an open problem to be explored.

# Chapter 4
# Sentence Representation

**Abstract** Sentence is an important linguistic unit of natural language. Sentence Representation has remained as a core task in natural language processing, because many important applications in related fields lie on understanding sentences, for example, summarization, machine translation, sentiment analysis, and dialogue system. Sentence representation aims to encode the semantic information into a real-valued representation vector, which will be utilized in further sentence classification or matching tasks. With large-scale text data available on the Internet and recent advances on deep neural networks, researchers tend to employ neural networks (e.g., convolutional neural networks and recurrent neural networks) to learn low-dimensional sentence representations and achieve great progress on relevant tasks. In this chapter, we first introduce the one-hot representation for sentences and the $n$-gram sentence representation (i.e., probabilistic language model). Then we extensively introduce neural-based models for sentence modeling, including feedforward neural network, convolutional neural network, recurrent neural network, and the latest Transformer, and pre-trained language models. Finally, we introduce several typical applications of sentence representations.

## 4.1 Introduction

Natural language sentences consist of words or phrases, follow grammatical rules, and convey complete semantic information. Compared with words and phrases, sentences have more complex structures, including both sequential and hierarchical structures, which are essential for understanding sentences. In NLP, how to represent sentences is critical for related applications, such as sentence classification, sentiment analysis, sentence matching, and so on.

Before deep learning took off, sentences were usually represented as one-hot vectors or TF-IDF vectors, following the assumption of bag-of-words. In this case, a sentence is represented as a vocabulary-sized vector, in which each element represents the importance of a specific word (either term frequency or TF-IDF) to the sentence. However, this method confronts two issues. Firstly, the dimension of such representation vectors is usually up to thousands or millions. Thus, they usually face sparsity problem and bring in computational efficiency problem. Secondly, such a representation method follows the bag-of-words assumption and ignores the sequential and structural information, which can be crucial for understanding the semantic meanings of sentences.

Inspired by recent advances of deep learning models in computer vision and speech, researchers proposed to model sentences with deep neural networks, such as convolutional neural network, recurrent neural network, and so on. Compared with conventional word frequency-based sentence representations, deep neural networks can capture the internal structures of sentences, e.g., sequential and dependency information, through convolutional or recurrent operations. Thus, neural network-based sentence representations have achieved great success in sentence modeling and NLP tasks.

## 4.2 One-Hot Sentence Representation

One-hot representation is the most simple and straightforward method for word representation tasks. This method represents each word with a fixed length binary vector. Specifically, for a vocabulary $V = \{w_1, w_2, \ldots, w_{|V|}\}$, the one-hot representation of word $w$ is $\mathbf{w} = [0, \ldots, 0, 1, 0, \ldots, 0]$. Based on the one-hot word representation and the vocabulary, it can be extended to represent a sentence $s = \{w_1, w_2, \ldots, w_l\}$ as

$$\mathbf{s} = \sum_{k=1}^{l} \mathbf{w}_i, \tag{4.1}$$

where $l$ indicates the length of the sentence $s$. The sentence representation $\mathbf{s}$ is the sum of the one-hot representations of $n$ words within the sentence, i.e., each element in $\mathbf{s}$ represents the Term Frequency (TF) of the corresponding word.

Moreover, researchers usually take the importance of different words into consideration, rather than treat all the words equally. For example, the function words such as "a", "an", and "the" usually appear in different sentences, and reserve little meanings. Therefore, the Inverse Document Frequency (IDF) is employed to measure the importance of $w_i$ in $V$ as follows:

$$\mathrm{idf}_{w_i} = \log \frac{|D|}{\mathrm{df}_{w_i}}, \tag{4.2}$$

where $|D|$ is the number of all documents in the corpus $D$ and $\mathrm{df}_{w_i}$ represents the Document Frequency (DF) of $w_i$.

With the importance of each word, the sentences are represented more precisely as follows:

$$\hat{\mathbf{s}} = \mathbf{s} \otimes \mathrm{idf}, \tag{4.3}$$

where $\otimes$ is the element-wise product.

Here, $\hat{\mathbf{s}}$ is the TF-IDF representation of the sentence $s$.

## 4.3 Probabilistic Language Model

One-hot sentence representation usually neglects the structure information in a sentence. To address this issue, researchers propose probabilistic language model, which treats $n$-grams rather than words as the basic components. An $n$-gram means a subsequence of words in a context window of length $n$, and probabilistic language model defines the probability of a sentence $s = [w_1, w_2, \ldots, w_l]$ as

$$P(s) = \prod_{i=1}^{l} P(w_i | w_1^{i-1}). \tag{4.4}$$

Actually, model indicated in Eq. (4.4) is not practicable due to its enormous parameter space. In practice, we simplify the model and set an $n$-sized context window, assuming that the probability of word $w_i$ only depends on $[w_{i-n+1} \cdots w_{i-1}]$. More specifically, an $n$-gram language model predicts word $w_i$ in the sentence $s$ based on its previous $n - 1$ words. Therefore, the simplified probability of a sentence is formalized as

$$P(s) = \prod_{i=1}^{l} P(w_i | w_{i-n+1}^{i-1}), \tag{4.5}$$

where the probability of selecting the word $w_i$ can be calculated from $n$-gram model frequency counts:

$$P(w_i | w_{i-n+1}^{i-1}) = \frac{P(w_{i-n+1}^{i})}{P(w_{i-n+1}^{i-1})}. \tag{4.6}$$

Typically, the conditional probabilities in $n$-gram language models are not calculated directly from the frequency counts, since it suffers severe problems when confronted with any $n$-grams that have not explicitly been seen before. Therefore, researchers proposed several types of smoothing approaches, which assign some of the total probability mass to unseen words or $n$-grams, such as “add-one” smoothing, Good-Turing discounting, or back-off models.

$n$-gram model is a typical probabilistic language model for predicting the next word in an $n$-gram sequence, which follows the Markov assumption that the probability of the target word only relies on the previous $n - 1$ words. The idea is employed by most of current sentence modeling methods. $n$-gram language model is used as an approximation of the true underlying language model. This assumption is crucial because it massively simplifies the problem of learning the parameters of language models from data. Recent works on word representation learning [3, 40, 43] are mainly based on the $n$-gram language model.

## 4.4 Neural Language Model

Although smoothing approaches could alleviate the sparse problem in the probabilistic language model, it still performs poorly for those unseen or uncommon words and $n$-grams. Moreover, since probabilistic language models are constructed on larger and larger texts, the number of unique words (the vocabulary) increases and the number of possible sequences of words increases exponentially with the size of the vocabulary, causing a data sparsity problem. Thus statistics are needed to estimate probabilities accurately.

To address this issue, researchers propose neural language models which use continuous representations or embeddings of words and neural networks to make their predictions, in which embeddings in the continuous space help to alleviate the curse of dimensionality in language modeling, and neural networks avoid this problem by representing words in a distributed way, as nonlinear combinations of weights in a neural net [2]. An alternate description is that a neural network approximates the language function. The neural net architecture might be feedforward or recurrent, and while the former is simpler, the latter is more common.

Similar to probabilistic language models, neural language models are constructed and trained as probabilistic classifiers that learn to predict a probability distribution:

$$P(s) = \prod_{i=1}^{l} P(w_i|\mathbf{w}_1^{i-1}), \tag{4.7}$$

where the conditional probability of the selecting word $w_i$ can be calculated by various kinds of neural networks such as feedforward neural networks, recurrent neural networks, and so on. In the following sections, we will introduce these neural language models in detail.

### *4.4.1 Feedforward Neural Network Language Model*

The goal of neural network language model is to estimate the conditional probability $P(w_i|w_1, \ldots, w_{i-1})$. However, the feedforward neural network (FNN) lacks an effective way to represent the long-term historical context. Therefore, it adopts the idea of $n$-gram language models to approximate the conditional probability, which assumes that each word in a word sequence more statistically depends on those words closer to it, and only $n - 1$ context words are used to calculate the conditional probability, i.e., $P(w_i|\mathbf{w}_1^{i-1}) \approx P(w_i|\mathbf{w}_{i-n+1}^{i-1})$.

The overall architecture of the FNN language model is proposed by [3]. To evaluate the conditional probability of the word $w_i$, it first projects its $n - 1$ context-related words to their word vector representations $\mathbf{x} = [\mathbf{w}_{i-n+1}, \ldots, \mathbf{w}_{i-1}]$, and then feeds them into an FNN, which can be generally represented as

$$\mathbf{y} = \mathbf{M}f(\mathbf{W}\mathbf{x} + \mathbf{b}) + \mathbf{d}, \tag{4.8}$$

where $\mathbf{W}$ is a weighted matrix to transform word vectors to hidden representations, $\mathbf{M}$ is a weighted matrix for the connections between the hidden layer and the output layer, and $\mathbf{b}$, $\mathbf{d}$ are bias vectors. And then the conditional probability of the word $w_i$ can be calculated as

$$P(w_i|\mathbf{w}_{i-n}^{i-1}) = \frac{\exp(\mathbf{y}_{w_i})}{\sum_j \exp(\mathbf{y}_j)}. \tag{4.9}$$

### 4.4.2 Convolutional Neural Network Language Model

The Convolutional Neural Network (CNN) is the family of neural network models that features a type of layer known as the convolutional layer. This layer can extract features by a learnable filter (or kernel) at the different positions of an input. Pham et al. [47] propose the CNN language model to enhance the FNN language model. The proposed CNN network is produced by injecting a convolutional layer after the word input representation $\mathbf{x} = [\mathbf{w}_{i-n}, \ldots, \mathbf{w}_{i-1}]$. Formally, the convolutional layer involves a sliding window of the input vectors centered on each word vector using a parameter matrix $W_c$, which can be generally represented as

$$\mathbf{y} = \mathbf{M}\big(\max(\mathbf{W_c}\mathbf{x})\big), \tag{4.10}$$

where $\max(\cdot)$ indicates a max-pooling layer. The architecture of CNN is shown in Fig. 4.1.

Moreover, [12] also introduces a convolutional neural network for language modeling with a novel gating mechanism.

### 4.4.3 Recurrent Neural Network Language Model

To address the lack of ability for modeling long-term dependency in the FNN language model, [41] proposes a Recurrent Neural Network (RNN) language model which applies RNN in language modeling. RNNs are fundamentally different from FNNs in the sense that they operate on not only an input space but also an internal state space, and the internal state space enables the representation of sequentially extended dependencies. Therefore, the RNN language model can deal with those sentences of arbitrary length. At every time step, its input is the vector of its previous word instead of the concatenation of vectors of its $n$ previous words, and the information of all other previous words can be taken into account by its internal state. Formally, the RNN language model can be defined as

$$\mathbf{h}_i = f(\mathbf{W}_1\mathbf{h}_{i-1} + \mathbf{W}_2\mathbf{w}_i + \mathbf{b}), \tag{4.11}$$

$$\mathbf{y} = \mathbf{M}\mathbf{h}_{i-1} + \mathbf{d}, \tag{4.12}$$

where $\mathbf{W}_1$, $\mathbf{W}_2$, $\mathbf{M}$ are weighted matrices and $\mathbf{b}$, $\mathbf{d}$ are bias vectors. Here, the RNN unit can also be implemented by LSTM or GRU. The architecture of RNN is shown in Fig. 4.2.

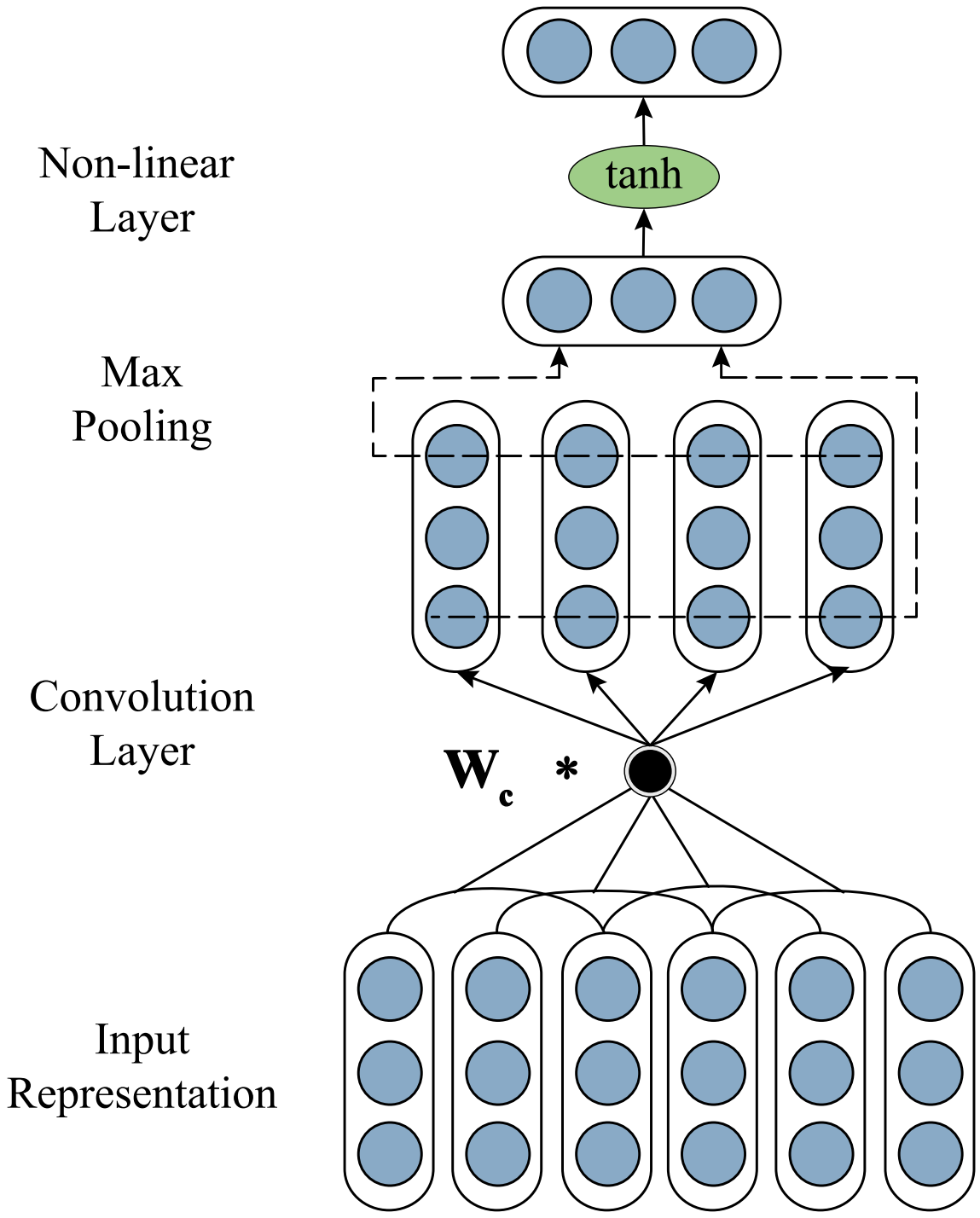

**Fig. 4.1** The architecture of CNN

Recently, researchers make some comparisons among neural network language models with different architectures on both small and large corpora. The experimental results show that, generally, the RNN language model outperforms the CNN language model.

### 4.4.4 Transformer Language Model

In 2018, Google proposed a pre-trained language model (PLM), called BERT, which achieved state-of-the-art results on a variety of NLP tasks. At that time, it was very big news. Since then, all the NLP researchers began to consider how PLMs can benefit their research tasks.

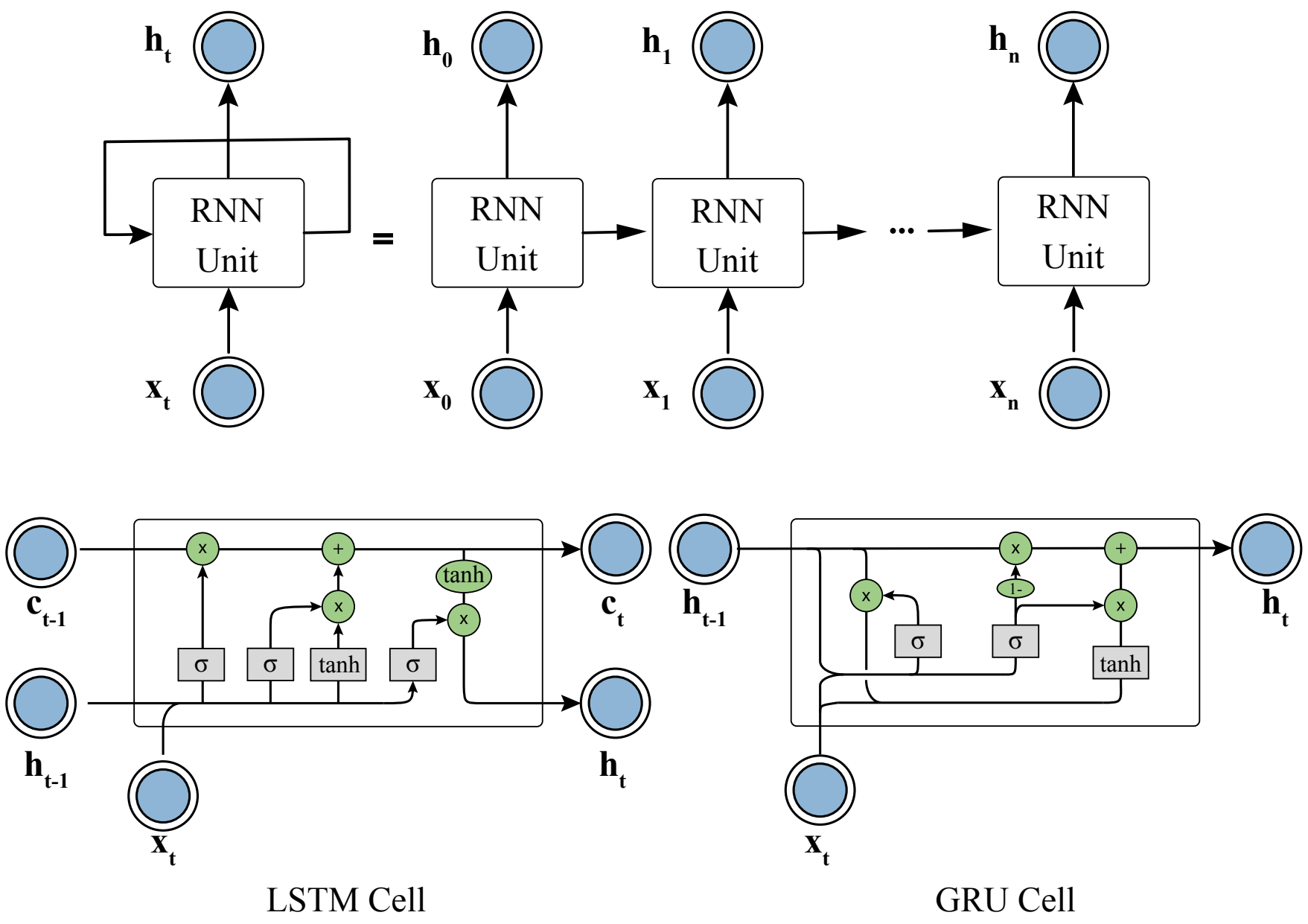

**Fig. 4.2** The architecture of RNN

In this section, we will first introduce the Transformer architecture and then talk about BERT and other PLMs in detail.

#### 4.4.4.1 Transformer

Transformer [65] is a nonrecurrent encoder-decoder architecture with a series of attention-based blocks. For the encoder, there are 6 layers and each layer is composed of a multi-head attention sublayer and a position-wise feedforward sublayer. And there is a residual connection between sublayers. The architecture of the Transformer is as shown in Fig. 4.3.

There are several attention heads in the multi-head attention sublayer. A *head* represents a scaled dot-product attention structure, which takes the query matrix **Q**, the key matrix **K**, and the value matrix **V** as the inputs, and the output is computed by

$$\text{Attention}(\mathbf{Q}, \mathbf{K}, \mathbf{V}) = \text{Softmax}\left(\frac{\mathbf{Q}\mathbf{K}^T}{\sqrt{d_k}}\right)\mathbf{V}, \tag{4.13}$$

where $d_k$ is the dimension of query matrix.

The multi-head attention sublayer linearly projects the input hidden states $H$ several times into the query matrix, the key matrix, and the value matrix for $h$ heads. The dimensions of the query, key, and value vectors are $d_k$, $d_k$, and $d_v$, respectively.

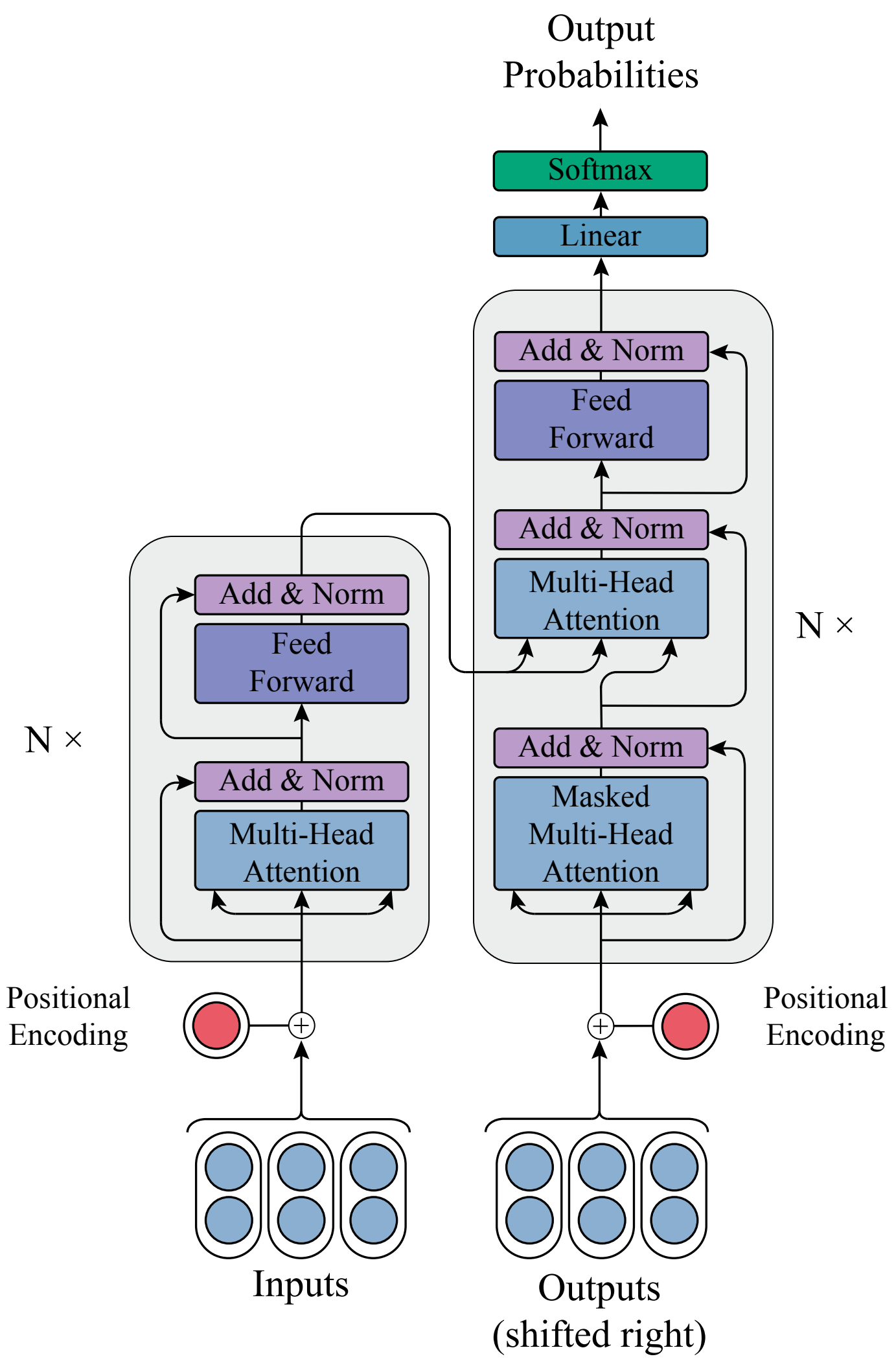

**Fig. 4.3** The architecture of Transformer

The multi-head attention sublayer could be formulated as

$$\text{Multihead}(H) = [head_1, head_2, \ldots, head_h]\mathbf{W}^O, \tag{4.14}$$

where $head_i = \text{Attention}(\mathbf{HW}_i^Q, \mathbf{HW}_i^K, \mathbf{HW}_i^V)$, and $\mathbf{W}_i^Q$, $\mathbf{W}_i^K$ and $\mathbf{W}_i^V$ are linear projections. $\mathbf{W}^O$ is also a linear projection for the output. Here, the fully connected position-wise feedforward sublayer contains two linear transformations with ReLU activation:

$$\mathrm{FFN}(x) = \mathbf{W}_2 \max(0, \mathbf{W}_1 x + \mathbf{b}_1) + \mathbf{b}_2. \tag{4.15}$$

Transformer is better than RNNs for modeling the long-term dependency, where all tokens will be equally considered during the attention operation. The Transformer was proposed to solve the problem of machine translation. Since Transformer has a very powerful ability to model sequential data, it becomes the most popular backbone of NLP applications.

#### 4.4.4.2 Transformer-Based PLM

Neural models can learn large amounts of language knowledge from language modeling. Since the language knowledge covers the demands of many downstream NLP tasks and provides powerful representations of words and sentences, some researchers found that knowledge can be transferred to other NLP tasks easily. The transferred models are called Pre-trained Language Models (PLMs).

Language modeling is the most basic and most important NLP task. It contains a variety of knowledge for language understanding, such as linguistic knowledge and factual knowledge. For example, the model needs to decide whether it should add an article before a noun. This requires linguistic knowledge about articles. Another example is the question of what is the following word after "Trump is the president of". The answer is "America", which requires factual knowledge. Since language modeling is very complex, the models can learn a lot from this task.

On the other hand, language modeling only requires plain text without any human annotation. With this feature, the models can learn complex NLP abilities from a very large-scale corpus. Since deep learning needs large amounts of data and language modeling can make full use of all texts in the world, PLMs significantly benefit the development of NLP research.

Inspired by the success of the Transformer, GPT [50] and BERT [14] begin to adopt the Transformer as the backbone of the pre-trained language models. GPT and BERT are the most representative Transformer-based pre-trained language models (PLMs). Since they achieved state-of-the-art performance on various NLP tasks, nearly all PLMs after them are based on the Transformer. In this subsection, we will talk about GPT and BERT in more detail.

GPT is the first work to pretrain a PLM based on the Transformer. The training procedure of GPT [50] contains two classic stages: generative pretraining and discriminative fine-tuning.

In the pretraining stage, the input of the model is a large-scale unlabeled corpus denoted as $\mathscr{U} = \{u_1, u_2, \ldots, u_n\}$. The pretraining stage aims to optimize a language model. The learning objective over the corpus is to maximize a conditional likelihood in a fixed-size window:

$$\mathscr{L}_1(\mathscr{U}) = \sum_i \log P(u_i | u_{i-k}, \ldots, u_{i-1}; \Theta), \tag{4.16}$$

where $k$ represents the size of the window, the conditional likelihood $P$ is modeled by a neural network with parameters $\Theta$.

For a supervised dataset $\chi$, the input is a sequence of words $s = (w_1, w_2, .., w_l)$ and the output is a label $y$. The pretraining stage provides an advantageous start point of parameters that can be used to initialize subsequent supervised tasks. At this occasion, the objective is a discriminative task that maximizes the conditional possibility distribution:

$$\mathscr{L}_2(\chi) = \sum_{(s,y)} \log P(y|w_1, \ldots, w_l), \tag{4.17}$$

where $P(y|w_1, \ldots, w_l)$ is modeled by a K-layer Transformer. After the input tokens pass through the pretrained GPT, a hidden vector of the final layer $\mathbf{h}_l^K$ will be produced. To obtain the output distribution, a linear transformation layer is added, which has the same size as the number of labels:

$$P(y|w_1, \ldots, w_m) = \text{Softmax}(\mathbf{W}_y \mathbf{h}_l^K). \tag{4.18}$$

The final training objective is combined with a language modeling $\mathscr{L}_1$ for better generalization:

$$\mathscr{L}(\chi) = \mathscr{L}_2(\chi) + \lambda * \mathscr{L}_1(\chi), \tag{4.19}$$

where $\lambda$ is a weight hyperparameter.

BERT [14] is a milestone work in the field of PLM. BERT achieved significant empirical results on 17 different NLP tasks, including SQuAD (outperform human being), GLUE (7.7% point absolute improvement), MultiNLI (4.6% point absolute improvement), etc. Compared to GPT, BERT uses a bidirectional deep Transformer as the model backbone. As illustrated in Fig. 4.4, BERT contains pretraining and fine-tuning stages.

In the pretraining stage, two objectives are designed: *Masked Language Model (MLM)* and *Next Sentence Prediction (NSP)*. (1) For MLM, tokens are randomly masked with a special token [MASK]. The training objective is to predict the masked tokens based on the contexts. Compared with the standard unidirectional conditional language model, which can only be trained in one direction, MLM aims to train a deep bidirectional representation model. This task is inspired by *Cloze* [64]. (2) The objective of NSP is to capture relationships between sentences for some sentence-based downstream tasks such as natural language inference (NLI) and question answering (QA). In this task, a binary classifier is trained to predict whether the sentence is the next sentence for the current. This task effectively captures the deep relationship between sentences, exploring semantic information from a different level.

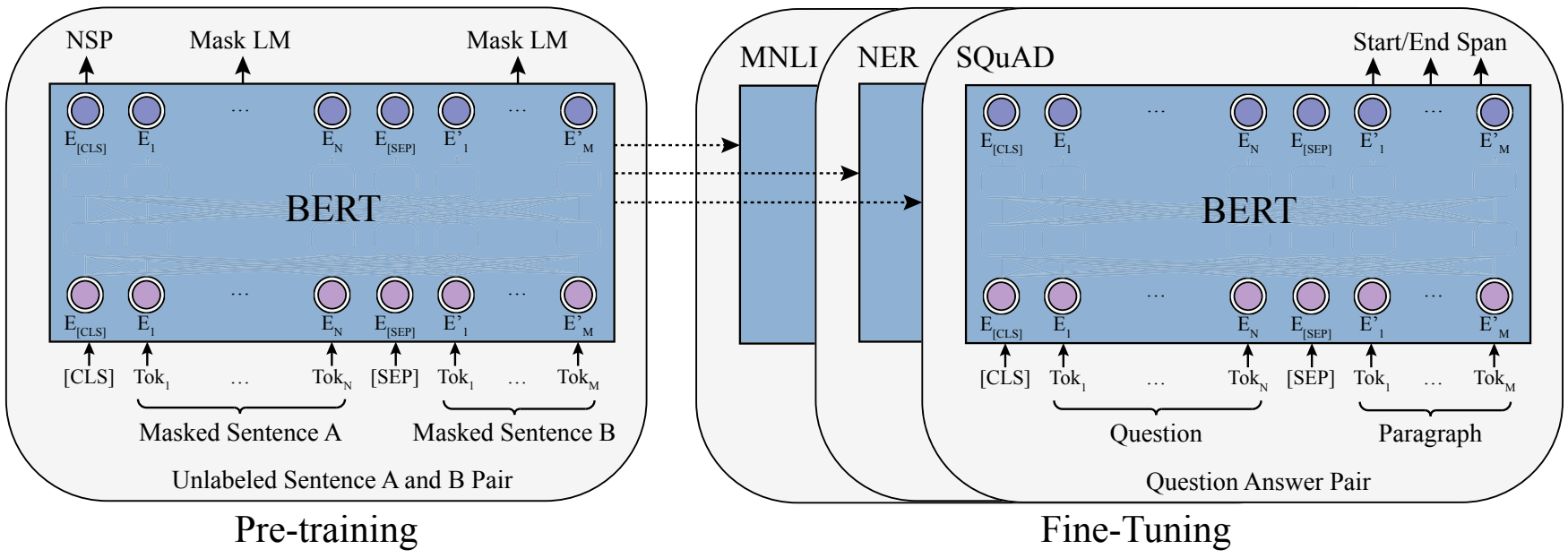

**Fig. 4.4** The pretraining and fine-tuning stages for BERT

After pretraining, BERT can capture various language knowledge for downstream supervised tasks. By modifying inputs and outputs, BERT can be fine-tuned for any NLP tasks, which contain the applications with the input of single text or text pairs. The input consists of sentence A and sentence B, which can represent (1) sentence pairs in paraphrase, (2) hypothesis-premise pairs in entailment, (3) question-passage pairs in QA, and (4) text-Ø for text classification task or sequence tagging. For the output, BERT can produce the token-level representation for each token, which is used to sequence tagging task or question answering. Besides, the special token [CLS] in BERT is fed into the classification layer for sequence classification.

#### 4.4.4.3 PLM Family

Pre-trained language models have rapid progress after BERT. We summarize several important directions of PLMs and show some representative models and their relationship in Fig. 4.5.

Here is a brief introduction of the PLMs after BERT. Firstly, there are some variants of BERT for better general language representation, such as RoBERTa [38] and XLNet [70]. These models mainly focus on the improvement of pretraining tasks. Secondly, some people work on pretrained generation models, such as MASS [57] and UniLM [15]. These models achieve promising results on the generation tasks instead of the Natural Language Understanding (NLU) tasks used by BERT. Thirdly, the sentence pair format of BERT inspired works on the cross-lingual and cross-modal fields. XLM [8], ViLBERT [39], and VideoBERT [59] are the important works in this direction. Lastly, there are some works [46, 81] that explore to incorporate external knowledge into PLMs since some low-frequency knowledge cannot be efficiently learned by PLMs.

### *4.4.5 Extensions*

#### 4.4.5.1 Importance Sampling

Inspired by the contrastive divergence model, [4] proposes to adopt importance sampling to accelerate the training of neural language models. They first normalize the outputs of neural network language model and view neural network language models as a special case of energy-based probability models as following:

$$P(w_i|\mathbf{w}_{i-n}^{i-1}) = \frac{\exp(-y_{w_i})}{\sum_j \exp(-y_j)}. \tag{4.20}$$

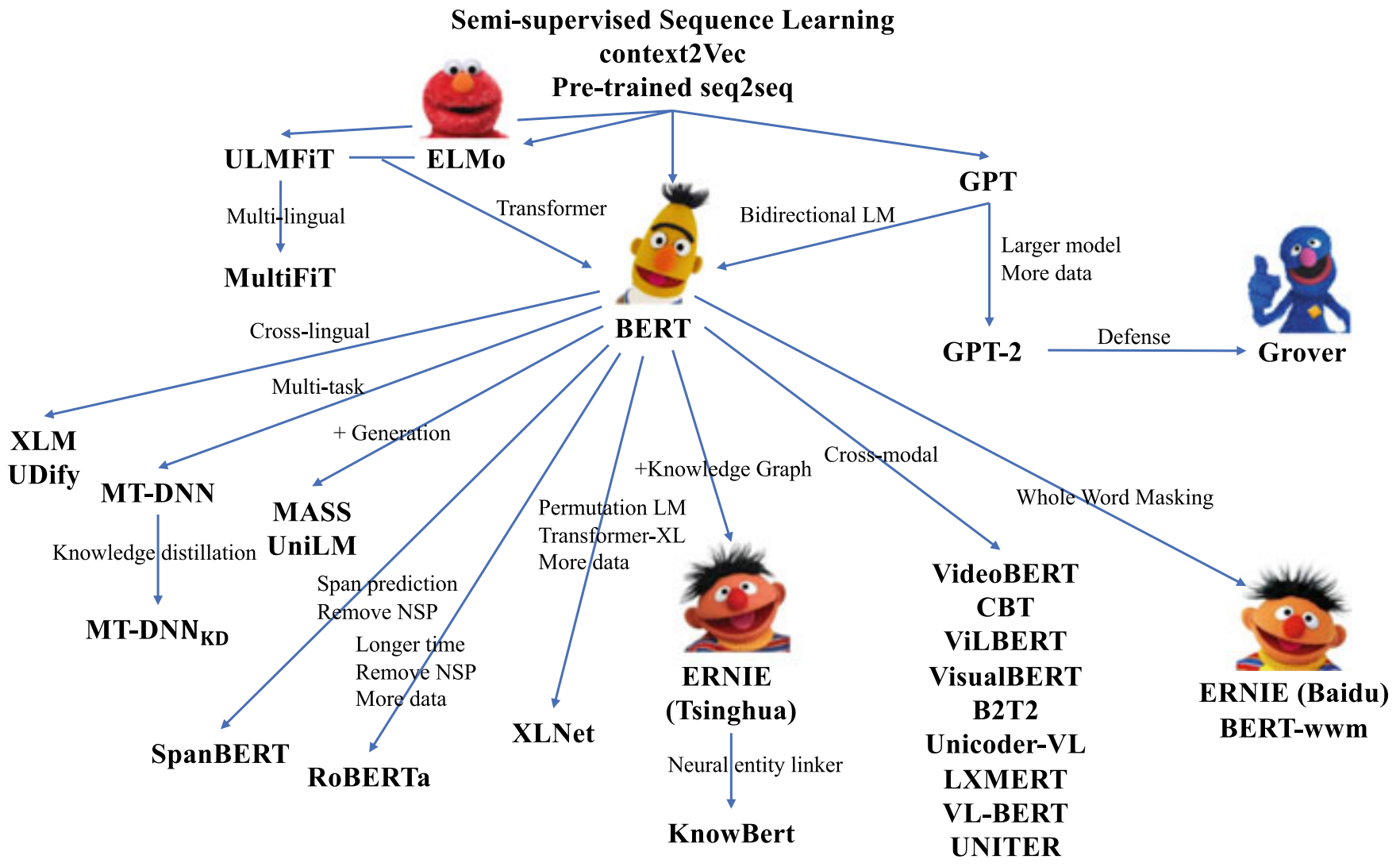

**Fig. 4.5** The Pre-trained language model family

The key idea of importance sampling is to approximate the mean of log-likelihood gradient of the loss function of neural network language model by sampling several important words instead of calculating the explicit gradient. Here, the log-likelihood gradient of the loss function of neural network language model can be generally represented as

$$\begin{aligned}\frac{\partial P(w_i|\mathbf{w}_{i-n}^{i-1})}{\partial \theta} &= -\frac{\partial y_{w_i}}{\partial \theta} + \sum_{j=1}^{|V|} P(w_j|\mathbf{w}_{i-n}^{i-1})\frac{\partial y_j}{\partial \theta} \\ &= -\frac{\partial y_i}{\partial \theta} + \mathbb{E}_{w_k \sim P}\left[\frac{\partial y_k}{\partial \theta}\right],\end{aligned} \tag{4.21}$$

where $\theta$ indicates all parameters of the neural network language model. Here, the log-likelihood gradient of the loss function consists of two parts including positive gradient for target word $w_i$ and negative gradient for all words $w_j$, i.e., $\mathbb{E}_{w_i \sim P}[\frac{\partial y_j}{\partial \theta}]$. Here, the second part can be approximated by sampling important words following the probability distribution $P$:

$$\mathbb{E}_{w_k \sim P}\left[\frac{\partial y_k}{\partial \theta}\right] \approx \sum_{w_k \in V'} \frac{1}{|V'|}\frac{\partial y_k}{\partial \theta}, \tag{4.22}$$

where $V'$ is the word set sampled under $P$.

However, since we cannot obtain probability distribution $P$ in advance, it is impossible to sample important words following the probability distribution $P$. Therefore, importance sampling adopts a Monte Carlo scheme which uses an existing proposal distribution $Q$ to approximate $P$, and then we have

$$\mathbb{E}_{w_k \sim P}\left[\frac{\partial y_k}{\partial \theta}\right] \approx \frac{1}{|V''|}\sum_{w_l \in V''} \frac{\partial y_l}{\partial \theta} P(w_l|\mathbf{w}_{i-n}^{i-1})/Q(w_l), \tag{4.23}$$

where $V''$ is the word set sampled under $Q$. Moreover, the sample size of importance sampling approach should be increased as training processes in order to avoid divergence, which aims to ensure its effective sample size $S$:

$$S = \frac{(\sum_{w_l \in V''} r_l)^2}{\sum_{w_l \in V''} r_l^2}, \tag{4.24}$$

where $r_l$ is further defined as

$$r_l = \frac{P(w_l|\mathbf{w}_{i-n}^{i-1})/Q(w_l)}{\sum_{w_j \in V''} P(w_j|\mathbf{w}_{i-n}^{i-1})/Q(w_j)}. \tag{4.25}$$

#### 4.4.5.2 Word Classification

Besides important sampling, researchers [7, 22] also propose class-based language model, which adopts word classification to improve the performance and speed of a language model. In class-based language model, all words are assigned to a unique class, and the conditional probability of a word given its context can be decomposed into the probability of the word's class given its previous words and the probability of the word given its class and history, which is formally defined as

$$P(w_i|\mathbf{w}_{i-n}^{i-1}) = \sum_{c(w_i) \in C} P(w_i|c(w_i))P(c(w_i)|\mathbf{w}_{i-n}^{i-1}), \tag{4.26}$$

where $C$ indicates the set of all classes and $c(w_i)$ indicates the class of word $w_i$.

Moreover, [44] proposes a hierarchical neural network language model, which extends word classification to a hierarchical binary clustering of words in a language model. Instead of simply assigning each word with a unique class, it first builds a hierarchical binary tree of words according to the word similarity obtained from WordNet. Next, it assigns a unique bit vector $c(w_i) = [c_1(w_i), c_2(w_i), \ldots, c_l(w_i)]$ for each word, which indicates the hierarchical classes of them. And then the conditional probability of each word can be defined as

$$P(w_i|\mathbf{w}_{i-n}^{i-1}) = \prod_{j=0}^{l} P(c_j(w_i)|c_1(w_i), c_2(w_i), \ldots, c_{j-1}(w_i), \mathbf{w}_{i-n}^{i-1}). \tag{4.27}$$

The hierarchical neural network language model can achieve $O(k/\log k)$ speed up as compared to a standard language model. However, the experimental results of [44] show that although the hierarchical neural network language model achieves an impressive speed up for modeling sentences, it has worse performance than the standard language model. The reason is perhaps that the introduction of hierarchical architecture or word classes imposes a negative influence on the word classification by neural network language models.

#### 4.4.5.3 Caching

Caching is also one of the important extensions in language model. A type of cache-based language model assumes that each word in recent context is more likely to appear again [58]. Hence, the conditional probability of a word can be calculated by the information from history and caching:

$$P(w_i|\mathbf{w}_{i-n}^{i-1}) = \lambda P_s(w_i|\mathbf{w}_{i-n}^{i-1}) + (1-\lambda) P_c(w_i|\mathbf{w}_{i-n}^{i-1}), \tag{4.28}$$

where $P_s(w_i|\mathbf{w}_{i-n}^{i-1})$ indicates the conditional probability generated by standard language and $P_c(w_i|\mathbf{w}_{i-n}^{i-1})$ indicates the conditional probability generated by caching, and $\lambda$ is a constant.

Another cache-based language model is also used to speed up the RNN language modeling [27]. The main idea of this approach is to store the outputs and states of language models for future predictions given the same contextual history.

## 4.5 Applications

In this section, we will introduce two typical sentence-level NLP applications including text classification and relation extraction, as well as how to utilize sentence representation for these applications.

### 4.5.1 Text Classification

Text classification is a typical NLP application and has lots of important real-world tasks such as parsing and semantic analysis. Therefore, it has attracted the interest of many researchers. The conventional text classification models (e.g., the LDA [6] and tree kernel [48] models) focus on capturing more contextual information and correct word order by extracting more useful and distinct features, but still expose a few issues (e.g., data sparseness) which has the significant impact on the classification accuracy. Recently, with the development of deep learning in the various fields of artificial intelligence, neural models have been introduced into the text classification field due to their abilities of text representation learning. In this section, we will introduce the two typical tasks of text classification, including sentence classification and sentiment classification.

#### 4.5.1.1 Sentence Classification

Sentence classification aims to assign a sentence an appropriate category, which is a basic task of the text classification application.

Considering the effectiveness of the CNN models in capturing sentence semantic meanings, [31] first proposes to utilize the CNN models trained on the top of pre-trained word embeddings to classify sentences, which achieved promising results on several sentence classification datasets. Then, [30] introduces a dynamic CNN model to model the semantic meanings of sentences. This model handles sentences of varying lengths and uses dynamic max-pooling over linear sequences, which could help the model capture both short-range and long-range semantic relations in sentences. Furthermore, [9] proposes a novel CNN-based model named as Very Deep CNN, which operates directly at the character level. It shows that those deeper models have better results on sentence classification and can capture the hierarchical information from scattered characters to whole sentences. Yin and Schütze [74] also propose MV-CNN, which utilizes multiple types of pretrained word embeddings and extracts features from multi-granular phrases with variable-sized convolutional layers. To address the drawbacks of MV-CNN such as model complexity and the requirement for the same dimension of embeddings, [80] proposes a novel model called MG-CNN to capture multiple features from multiple sets of embeddings that are concatenated at the penultimate layer. Zhang et al. [79] present RA-CNN to jointly exploit labels on documents and their constituent sentences, which can estimate the probability that a given sentence is informative and then scales the contribution of each sentence to aggregate a document representation in proportion to the estimates.

The RNN model which aims to capture the sequential information of sentences is also widely used in sentence classification. Lai et al. [32] propose a neural network for text classification, which applies a recurrent structure to capture contextual information. Moreover, [37] introduces a multitask learning framework based on the RNN to jointly learn across multiple sentence classification tasks, which employs three different mechanisms of sharing information to model sentences with both task-specific and shared layers. Yang et al. [71] introduce word-level and sentence-level attention mechanisms into an RNN-based model as well as a hierarchical structure to capture the hierarchical information of documents for sentence classification.

#### 4.5.1.2 Sentiment Classification

Sentiment classification is a special task of the sentence classification application, whose objective is to classify the sentimental polarities of opinions a piece of text contains, e.g., favorable or unfavorable, positive or negative. This task appeals the NLP community since it has lots of potential downstream applications such as movie review suggestions.

Similar to text classification, the sentence representation based on neural models has also been widely explored for sentiment classification. Glorot et al. [20] use a stacked denoising autoencoder in sentiment classification for the first time. Then, a series of recursive neural network models based on the recursive tree structure of sentences are conducted to learn sentence representations for sentiment classification, including the recursive autoencoder (RAE) [55], matrix-vector recursive neural network (MV-RNN) [54], and recursive neural tensor network (RNTN) [56]. Besides, [29] adopts a CNN to learn sentence representations and achieves promising performance in sentiment classification.

The RNN models also benefit sentiment classification as they are able to capture the sequential information. Li et al. [35] and Tai et al. [62] investigate a tree-structured LSTM model on text classification. There are also some hierarchical models proposed to deal with document-level sentiment classification [5, 63], which generate semantic representations at different levels (e.g., phrase, sentence, or document) within a document. Moreover, the attention mechanism is also introduced into sentiment classification, which aims to select important words from a sentence or important sentences from a document [71].

### *4.5.2 Relation Extraction*

To enrich existing KGs, researchers have devoted many efforts to automatically finding novel relational facts in text. Therefore, relation extraction (RE), which aims at extracting relational facts according to semantic information in plain text, has become a crucial NLP application. As RE is also an important downstream application of sentence representation, we will, respectively, introduce the techniques and extensions to show how to utilize sentence representation for different RE scenarios. Considering neural networks have become the backbone of the recent NLP research, we mainly focus on Neural RE (NRE) models in this section.

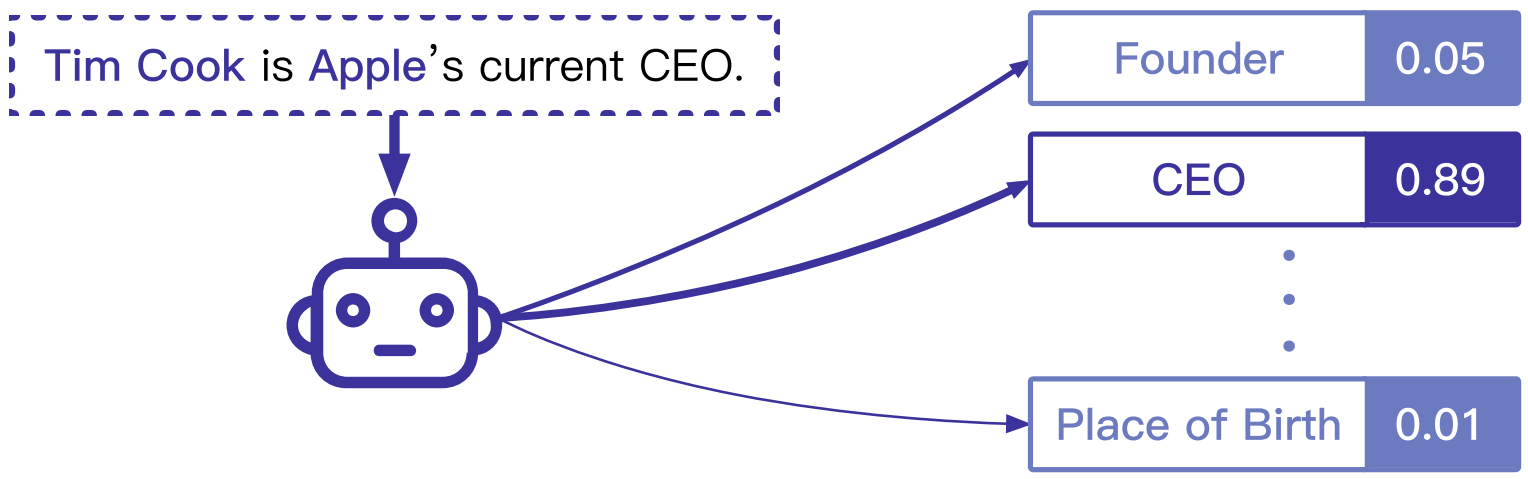

**Fig. 4.6** An example of sentence-level relation extraction

#### 4.5.2.1 Sentence-Level NRE

Sentence-level NRE aims at predicting the semantic relations between the given entity (or nominal) pair in a sentence. As shown in Fig. 4.6, given the input sentence $s$ which consists of $n$ words $s = \{w_1, w_2, \ldots, w_n\}$ and its corresponding entity pair $e_1$ and $e_2$ as input, sentence-level NRE wants to obtain the conditional probability $P(r|s, e_1, e_2)$ of relation $r$ ($r \in \mathscr{R}$) via a neural network, which can be formalized as

$$P(r|s, e_1, e_2) = P(r|s, e_1, e_2, \theta), \tag{4.29}$$

where $\theta$ is all parameters of the neural network and $r$ is a relation in the relation set $\mathscr{R}$.

A basic form of sentence-level NRE consists of three components: (a) an input encoder to give a representation for each input word, (b) a sentence encoder which computes either a single vector or a sequence of vectors to represent the original sentence, and (c) a relation classifier which calculates the conditional probability distribution of all relations.

**Input Encoder**. First, a sentence-level NRE system projects the discrete words of the source sentence into a continuous vector space, and obtains the input representation $\mathbf{w} = \{\mathbf{w}_1, \mathbf{w}_2, \ldots, \mathbf{w}_m\}$ of the source sentence.

(1) Word Embeddings. Word embeddings aim to transform words into distributed representations to capture the syntactic and semantic meanings of the words. In the sentence $s$, every word $w_i$ is represented by a real-valued vector. Word representations are encoded by column vectors in an embedding matrix $\mathbf{E} \in \mathbb{R}^{d^a \times |V|}$ where $V$ is a fixed-sized vocabulary. Although word embeddings are the most common way to represent input words, there are also efforts made to utilize more complicated information of input sentences for RE.

(2) Position Embeddings. In RE, the words close to the target entities are usually informative to determine the relation between the entities. Therefore, position embeddings are used to help models keep track of how close each word is to the head or tail entities. It is defined as the combination of the relative distances from the current word to the head or tail entities. For example, in the sentence `Bill_Gates is the founder of Microsoft.`, the relative distance from the word `founder` to the head entity `Bill_Gates` is −3 and the tail entity `Microsoft` is 2. Besides word position embeddings, more linguistic features are also considered in addition to the word embeddings to enrich the linguistic features of the input sentence.

(3) Part-of-speech (POS) Tag Embeddings. POS tag embeddings are to represent the lexical information of the target word in the sentence. Because word embeddings are obtained from a large-scale general corpus, the general information they contain may not be in accordance with the meaning in a specific sentence. Hence, it is necessary to align each word with its linguistic information considering its specific context, e.g., noun and verb. Formally, each word $w_i$ is encoded by the corresponding column vector in an embedding matrix $\mathbf{E}^p \in \mathbb{R}^{d^p \times |V^p|}$, where $d^p$ is the dimension of embedding vector and $V^p$ indicates a fixed-sized POS tag vocabulary.

(4) WordNet Hypernym Embeddings. WordNet hypernym embeddings aim to take advantages of the prior knowledge of hypernym to help RE models. When given the hypernym information of each word in WordNet (e.g., noun.food and verb.motion), it is easier to build the connections between different but conceptually similar words. Formally, each word $w_i$ is encoded by the corresponding column vector in an embedding matrix $\mathbf{E}^h \in \mathbb{R}^{d^h \times |V^h|}$, where $d^h$ is the dimension of embedding vector and $V^h$ indicates a fixed-sized hypernym vocabulary.

For each word, the NRE models often concatenate some of the above four feature embeddings as their input embeddings. Therefore, the feature embeddings of all words are concatenated and denoted as a final input sequence $\mathbf{w} = \{\mathbf{w}_1, \mathbf{w}_2, \ldots, \mathbf{w}_m\}$, where $\mathbf{w}_i \in \mathbb{R}^d$, $d$ is the total dimension of all feature embeddings concatenated for each word.

**Sentence Encoder**. The sentence encoder is the core for sentence representation, which encodes input representations into either a single vector or a sequence of vectors $\mathbf{x}$ to represent sentences. We will introduce the different sentence encoders in the following.

(1) Convolutional Neural Network Encoder. Zeng et al. [76] propose to encode input sentences using a CNN model, which extracts local features by a convolutional layer and combines all local features via a max-pooling operation to obtain a fixed-sized vector for the input sentence. Formally, a convolutional layer is defined as an operation on a vector sequence $\mathbf{w}$:

$$\mathbf{p} = \text{CNN}(\mathbf{w}), \tag{4.30}$$

where CNN indicates the convolution operation inside the convolutional layer.

And the $i$th element of the sentence vector $\mathbf{x}$ can be calculated as follows:

$$[\mathbf{x}]_i = f(\max(\mathbf{p}_i)), \tag{4.31}$$

where $f$ is a nonlinear function applied at the output, such as the hyperbolic tangent function.

Further, PCNN [75], which is a variation of CNN, adopts a piece-wise max-pooling operation. All hidden vectors $\{\mathbf{p}_1, \mathbf{p}_2, \ldots\}$ are divided into three segments by the head and tail entities. The max-pooling operation is performed over the three segments separately, and the $\mathbf{x}$ is the concatenation of the pooling results over the three segments.

(2) Recurrent Neural Network Encoder. Zhang and Wang [78] propose to embed input sentences using an RNN model which can learn the temporal features. Formally, each input word representation is put into recurrent layers step by step. For each step $i$, the network takes the $i$th word representation vector $\mathbf{w}_i$ and the output of the previous $i-1$ steps $\mathbf{h}_{i-1}$ as input:

$$\mathbf{h}_i = \text{RNN}(\mathbf{w}_i, \mathbf{h}_{i-1}), \tag{4.32}$$

where RNN indicates the transform function inside the RNN cell, which can be the LSTM units or the GRU units mentioned before.

The conventional RNN models typically deal with text sequences from start to end, and build the hidden state of each word only considering its preceding words. It has been verified that the hidden state considering its following words is more effective. Hence, the bi-directional RNN (BRNN) [52] is adopted to learn hidden states using both preceding and following words.

Similar to the previous CNN models in RE, the RNN model combines the output vectors of the recurrent layer as local features, and then uses a max-pooling operation to extract the global feature, which forms the representation of the whole input sentence. The max-pooling layer could be formulated as

$$[\mathbf{x}]_j = \max_i [\mathbf{h}_i]_j. \tag{4.33}$$

Besides max-pooling, word attention can also combine all local feature vectors together. The attention mechanism [1] learns attention weights on each step. Supposing $\mathbf{H} = [\mathbf{h}_1, \mathbf{h}_2, \ldots, \mathbf{h}_m]$ is the matrix consisting of all output vectors produced by the recurrent layer, the feature vector of the whole sentence $\mathbf{x}$ is formed by a weighted sum of these output vectors:

$$\alpha = \text{Softmax}(\mathbf{s}^\top \tanh(\mathbf{H})), \tag{4.34}$$

$$\mathbf{x} = \mathbf{H}\alpha^\top, \tag{4.35}$$

where $\mathbf{s}$ is a trainable query vector.

Besides, [42] proposes a model that captures information from both word sequence and tree-structured dependency by stacking bidirectional path-based LSTM-RNNs (i.e., bottom-up and top-down). More specifically, it focuses on the shortest path between the two target entities in the dependency tree, and utilizes the stacked layers to encode the shortest path for the whole sentence representation. In fact, some preliminary work [69] has shown that these paths are useful in RE, and various recursive neural models are also proposed for this. Next, we will introduce these recursive models in detail.

(3) Recursive Neural Network Encoder. The recursive encoder aims to extract features from the information of syntactic parsing trees, considering the syntactic information is beneficial for extracting relations from sentences. Generally, these encoders treat the tree structure inside syntactic parsing trees as a strategy of composition as well as a direction to combine each word feature.

Socher et al. [54] propose a recursive matrix-vector model (MV-RNN) which captures the structure information by assigning a matrix-vector representation for each constituent of the constituents in parsing trees. The vector captures the meaning of the constituent itself and the matrix represents how it modifies the meaning of the word it combines with. Tai et al. [62] further propose two types of tree-structured LSTMs including the Child-Sum Tree-LSTM and the N-ary Tree-LSTM to capture tree structure information. For the Child-Sum Tree-LSTM, given a tree, let $C(t)$ denote the set of children of node $t$. Its transition equations are defined as follows:

$$\hat{\mathbf{h}}_t = \sum_{k \in C(t)} \text{TLSTM}(\mathbf{h}_k), \tag{4.36}$$

where $\text{TLSTM}(\cdot)$ indicates a Tree-LSTM cell, which is simply modified from LSTM cell. The N-ary Tree-LSTM has similar transition equations as the Child-Sum Tree-LSTM. The only difference is that it limits the tree structures to have at most $N$ branches.

**Relation Classifier**. When obtaining the representation $\mathbf{x}$ of the input sentence, relation classifier calculates the conditional probability $P(r|x, e_1, e_2)$ via a softmax layer as follows:

$$P(r|x, e_1, e_2) = \text{Softmax}(\mathbf{M}\mathbf{x} + \mathbf{b}), \tag{4.37}$$

where $\mathbf{M}$ indicates the relation matrix and $\mathbf{b}$ is a bias vector.

#### 4.5.2.2 Bag-Level NRE

Although existing neural models have achieved great success for extracting novel relational facts, it always suffers the lack of training data. To address this issue, researchers proposed a distant supervision assumption to generate training data via aligning KGs and plain text automatically. The intuition of distant supervision assumption is that all sentences that contain two entities will express their relations in KGs. For example, (`New York`, `city_of`, `United States`) is a relational fact in a KG, distant supervision assumption will regard all sentences that contain these two entities as positive instances for the relation `city_of`. It offers a natural way of utilizing information from multiple sentences (bag-level) rather than a single sentence (sentence-level) to decide if a relation holds between two entities.

Therefore, bag-level NRE aims to predict the semantic relations between an entity pair using all involved sentences. As shown in Fig. 4.7, given the input sentence set $S$ which consists of $n$ sentences $S = \{s_1, s_2, \ldots, s_n\}$ and its corresponding entity pair $e_1$ and $e_2$ as inputs, bag-level NRE wants to obtain the conditional probability $P(r|S, e_1, e_2)$ of relation $r$ ($r \in \mathbb{R}$) via a neural network, which can be formalized as

$$P(r|S, e_1, e_2) = P(r|S, e_1, e_2, \theta). \tag{4.38}$$

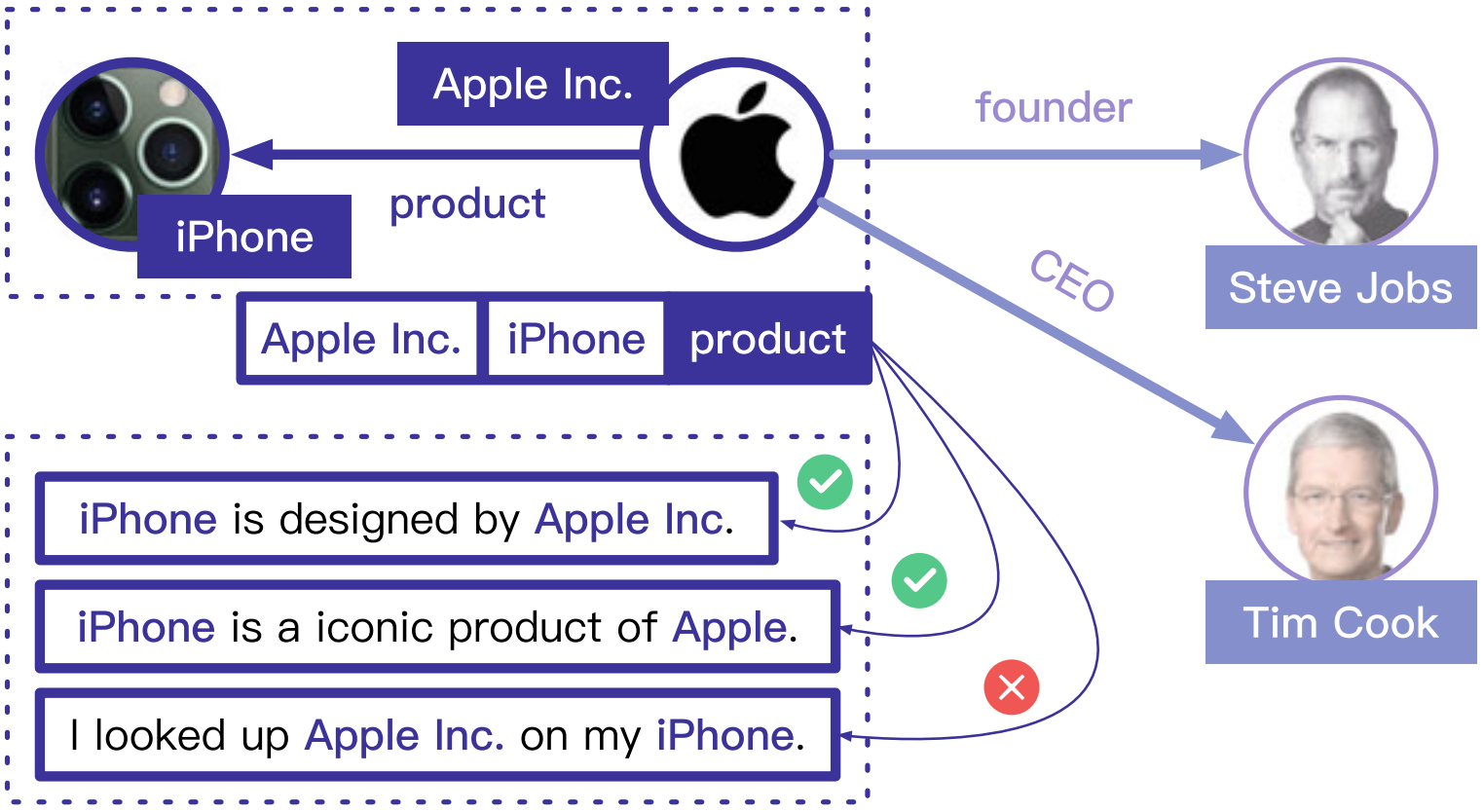

**Fig. 4.7** An example of bag-level relation extraction

A basic form of bag-level NRE consists of four components: (a) an input encoder similar to sentence-level NRE, (b) a sentence encoder similar to sentence-level NRE, (c) a bag encoder which computes a vector representing all related sentences in a bag, and (d) a relation classifier similar to sentence-level NRE which takes bag vectors as input instead of sentence vectors. As the input encoder, sentence encoder, and relation classifier of bag-level NRE are similar to the ones of sentence-level NRE, we will thus mainly focus on introducing the bag encoder in detail.

**Bag Encoder**. The bag encoder encodes all sentence vectors into a single vector $\mathbf{S}$. We will introduce the different bag encoders in the following:

(1) Random Encoder. It simply assumes that each sentence can express the relation between two target entities and randomly select one sentence to represent the bag. Formally, the bag representation is defined as

$$\mathbf{S} = \mathbf{s}_i \ (i \in \{1, 2, \ldots, n\}), \tag{4.39}$$

where $\mathbf{s}_i$ indicates the sentence representation of $s_i \in S$ and $i$ is a random index.

(2) Max Encoder. As introduced above, not all sentences containing two target entities can express their relations. For example, the sentence `New York City is the premier gateway for legal immigration to the United States` does not express the relation `city of`. Hence, in [75], they follow the at-least-one assumption which assumes that at least one sentence that contains these two target entities can express their relations, and select the sentence with the highest probability for the relation to represent the bag. Formally, bag representation is defined as

$$\mathbf{S} = \mathbf{s}_i \ (i = \arg\max_i P(r|s_i, e_1, e_2)). \tag{4.40}$$

(3) Average Encoder. Both random encoder or max encoder use only one sentence to represent the bag, which ignores the rich information of different sentences. To exploit the information of all sentences, [36] believes that the representation $\mathbf{S}$ of the bag depends on all sentences' representations. Each sentence representation $\mathbf{s}_i$ can give the relation information about two entities to a certain extent. The average encoder assumes that all sentences contribute equally to the representation of the bag. It means the embedding $\mathbf{S}$ of the bag is the average of all the sentence vectors:

$$\mathbf{S} = \sum_i \frac{1}{n}\mathbf{s}_i. \tag{4.41}$$

(4) Attentive Encoder. Due to the wrong label issue brought by distant supervision assumption inevitably, the performance of average encoder will be influenced by those sentences that contain no relation information. To address this issue, [36] further proposes to employ a selective attention to reduce those noisy sentences. Formally, the bag representation is defined as a weighted sum of sentence vectors:

$$\mathbf{S} = \sum_i \alpha_i \mathbf{s}_i, \tag{4.42}$$

where $\alpha_i$ is defined as

$$\alpha_i = \frac{\exp(\mathbf{s}_i^\top \mathbf{A}\mathbf{r})}{\sum_j \exp(\mathbf{x}_j^\top \mathbf{A}\mathbf{r})}, \tag{4.43}$$

where $\mathbf{A}$ is a diagonal matrix and $\mathbf{r}$ is the representation vector of relation $r$.

**Relation Classifier**. Similar to sentence-level NRE, when obtaining the bag representation $\mathbf{S}$, relation classifier also calculates the conditional probability $P(r|S, e_1, e_2)$ via a softmax layer as follows:

$$P(r|S, e_1, e_2) = \text{Softmax}(\mathbf{M}\mathbf{S} + \mathbf{b}), \tag{4.44}$$

where $\mathbf{M}$ indicates the relation matrix and $\mathbf{b}$ is a bias vector.

#### 4.5.2.3 Extensions

Recently, NRE systems have achieved significant improvements in both, the supervised and distantly supervised scenarios. However, there are still many challenges in the task of RE, and many researchers have been focusing on other aspects to improve the performance of NRE as well. In this section, we will introduce these extensions in detail.

**Utilization of External Information**. Most existing NRE systems stated above only concentrate on the sentences which are extracted, regardless of the rich external information such as KGs. This heterogeneous information could provide additional knowledge from KG and is essential when extracting new relational facts.

Han et al. [24] propose a novel joint representation learning framework for knowledge acquisition. The key idea is that the joint model learns knowledge and text representations within a unified semantic space via KG-text alignments. For the text part, the sentence with two entities `Mark Twain` and `Florida` is regarded as the input for a CNN encoder, and the output of CNN is considered to be the latent relation `PlaceOfBirth` of this sentence. For the KG part, entity and relation representations are learned via translation-based methods. The learned representations of KG and text parts are aligned during training. Besides this preliminary attempt, many efforts have been devoted to this direction [25, 28, 51, 67, 68].

**Incorporating Relational Paths**. Although existing NRE systems have achieved promising results, they still suffer a major problem: the models can only directly learn from those sentences which contain both two-target entities. However, those sentences containing only one of the entities could also provide useful information and help build inference chains. For example, if we know that "A is the son of B" and "B is the son of C", we can infer that A is the grandson of C.

To utilize the information of both direct and indirect sentences, [77] introduces a path-based NRE model that incorporates textual relational paths. The model first employs a CNN encoder to embed the semantic meanings of sentences. Then, the model builds a relation path encoder, which measures the probability of relations given an inference chain in the text. Finally, the model combines information from both direct sentences and relational paths, and then predicts the confidence of each relationship. This work is the first effort to consider the knowledge of relation path in text for NRE, and there are also several methods later to consider the reasoning path of sentence semantic meanings for RE [11, 19].

**Document-level Relation Extraction**. In fact, not all relational facts can be extracted by sentence-level RE, i.e., a large number of relational facts are expressed in multiple sentences. Taking Fig. 4.9 as an example, multiple entities are mentioned in the document and exhibit complex interactions. In order to identify the relational fact (`Riddarhuset`, `country`, `Sweden`), one has to first identify the fact that `Riddarhuset` is located in `Stockholm` from Sentence 4, then identify the facts `Stockholm` is the capital of `Sweden` and `Sweden` is a country from Sentence 1. With the above facts, we can finally infer that the sovereign state of `Riddarhuset` is `Sweden`. This process requires reading and reasoning over multiple sentences in a document, which is intuitively beyond the reach of sentence-level RE methods. According to the statistics on a human-annotated corpus sampled from Wikipedia documents [72], at least 40.7% relational facts can only be extracted from multiple sentences, which is not negligible. Swampillai and Stevenson [61] and Verga et al. [66] also report similar observations. Therefore, it is necessary to move RE forward from the sentence level to the document level. Figure 4.8 is an example for document-level RE.

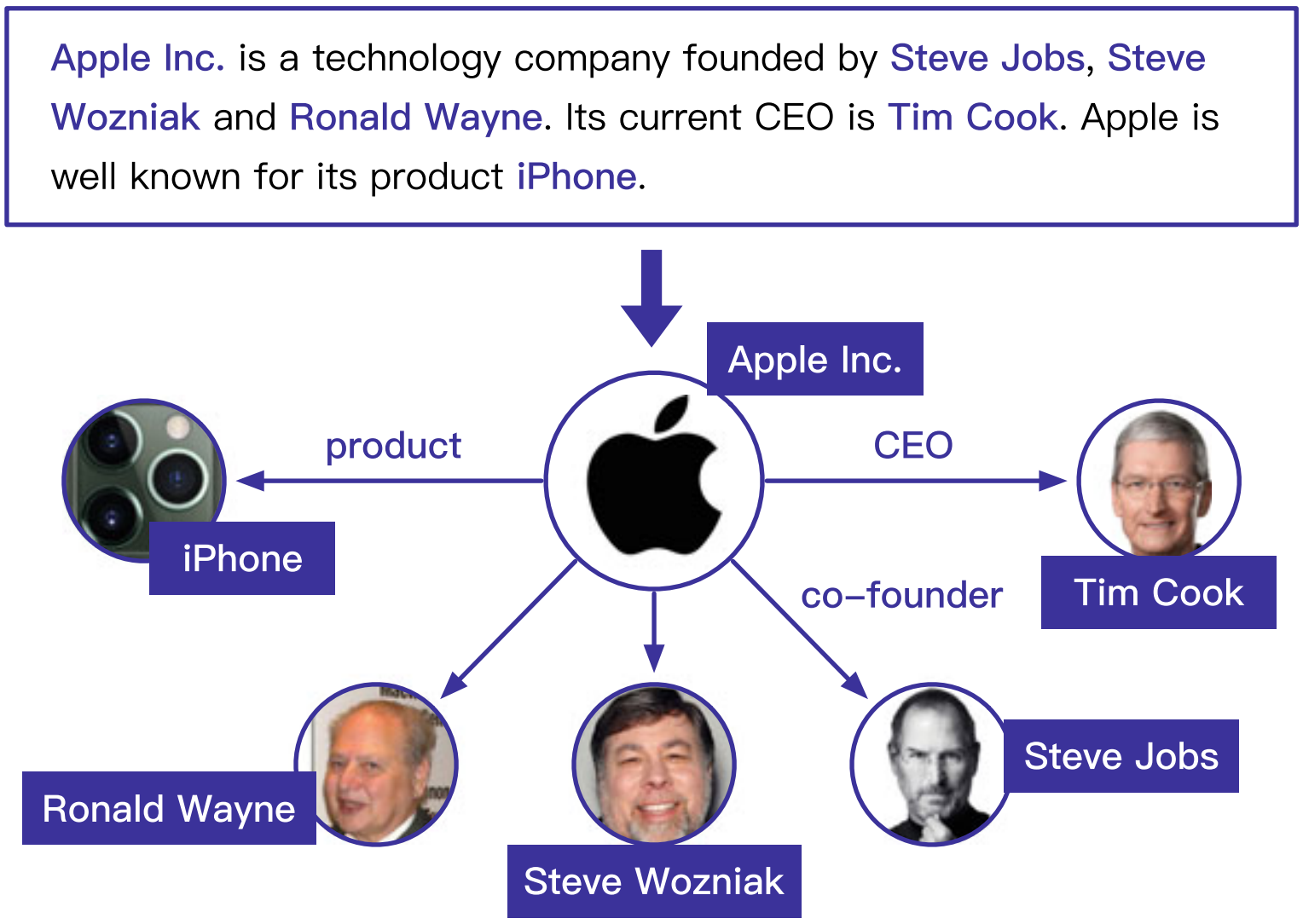

**Fig. 4.8** An example of document-level relation extraction

**Kungliga Hovkapellet**

[1] ***Kungliga Hovkapellet*** (The ***Royal Court Orchestra***) is a *Swedish* orchestra, originally part of the *Royal Court* in ***Sweden***'s capital *Stockholm*. [2] The orchestra originally consisted of both musicians and singers. [3] It had only male members until *1727*, when *Sophia Schröder* and *Judith Fischer* were employed as vocalists; in the *1850s*, the harpist *Marie Pauline Åhman* became the first female instrumentalist. [4] From *1731*, public concerts were performed at ***Riddarhuset*** in *Stockholm*. [5] Since *1773*, when the ***Royal Swedish Opera*** was founded by *Gustav III* of ***Sweden***, the ***Kungliga Hovkapellet*** has been part of the opera's company.

**Subject:** ***Kungliga Hovkapellet; Royal Court Orchestra***
**Object:** ***Royal Swedish Opera***
**Relation:** `part_of` **Supporting Evidence: 5**

**Subject:** ***Riddarhuset***
**Object:** ***Sweden***
**Relation:** `country` **Supporting Evidence: 1, 4**

**Fig. 4.9** An example from DocRED [72]

However, existing datasets for document-level RE either only have a small number of manually annotated relations and entities [34], or exhibit noisy annotations from distant supervision [45, 49], or serve specific domains or approaches [33]. To address this issue, [72] constructs a large-scale, manually annotated, and general-purpose document-level RE dataset, named as DocRED. DocRED is constructed from Wikipedia and Wikidata, and has two key features. First, DocRED contains $132,375$ entities and $56,354$ relational facts annotated on $5,053$ Wikipedia documents, which is the largest human-annotated document-level RE dataset now. Second, over 40% of the relational facts in DocRED can only be extracted from multiple sentences. This makes DocRED require reading multiple sentences in a document to recognize entities and inferring their relations by synthesizing all information of the document.

The experimental results on DocRED show that the performance of existing sentence-level RE methods declines significantly on DocRED, indicating the task document-level RE is more challenging than sentence-level RE and remains an open problem. It also relates to the document representation which will be introduced in the next chapter.

**Few-shot Relation Extraction**.

As we mentioned before, the performance of the conventional RE models [23, 76] heavily depend on time-consuming and labor-intensive annotated data, which make themselves hard to generalize well. Although adopting distant supervision is a primary approach to alleviate this problem, the distantly supervised data also exhibits a long-tail distribution, where most relations have very limited instances. Furthermore, distant supervision suffers the wrong labeling problem, which makes it harder to classify long-tail relations. Hence, it is necessary to study training RE models with insufficient training instances. Figure 4.10 is an example for few-shot RE.

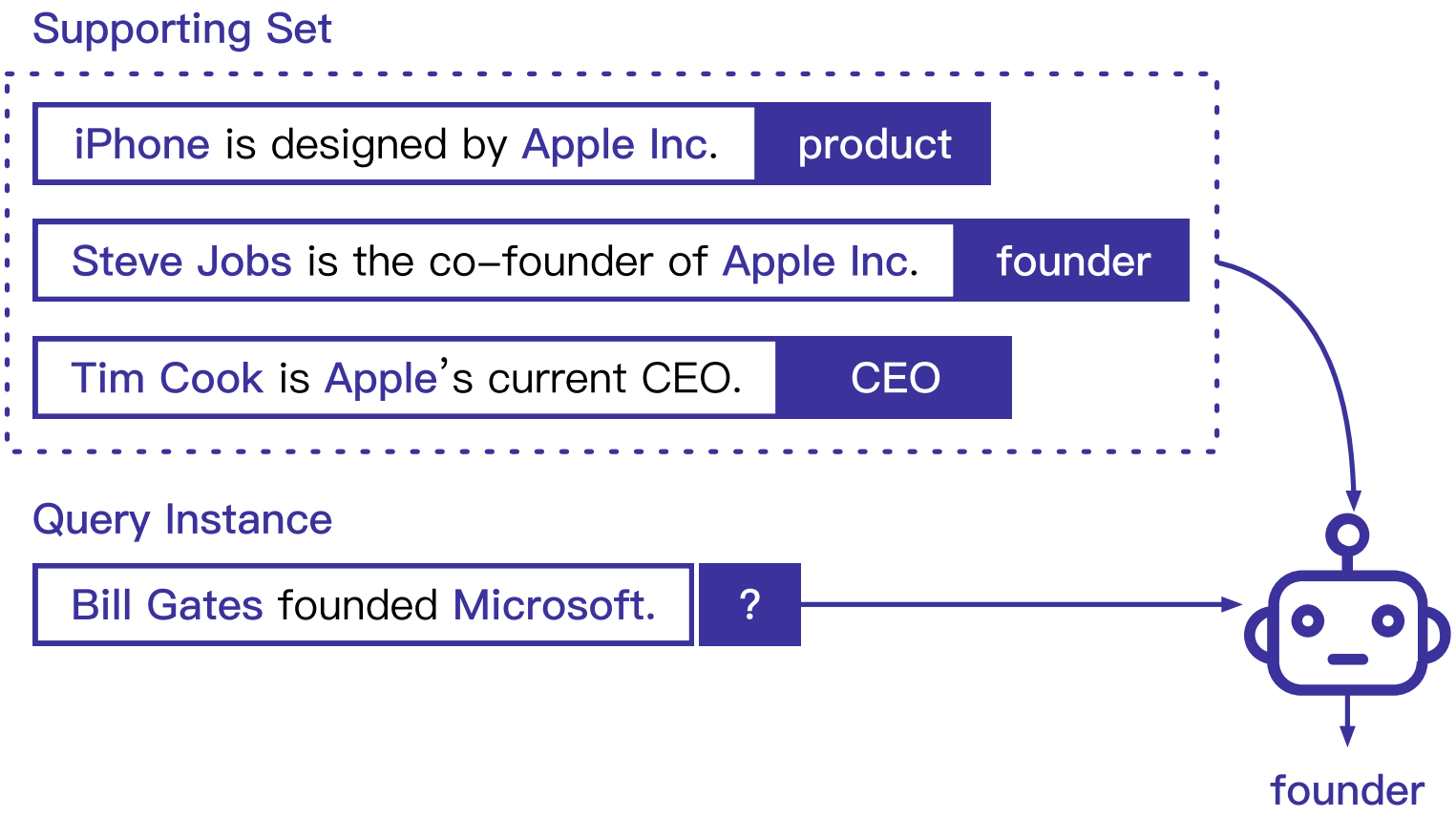

**Fig. 4.10** An example of few-shot relation extraction

**Table 4.1** An example for a 3 way 2 shot scenario. Different colors indicate different entities, underline for head entity, and *emphasize* for tail entity

| Supporting set | |
|---|---|
| (A) capital_of | (1) London is the capital of *the U.K*<br>(2) Washington is the capital of *the U.S.A* |
| (B) member_of | (1) Newton served as the president of *the Royal Society*<br>(2) Leibniz was a member of *the Prussian Academy of Sciences* |
| (C) birth_name | (1) *Samuel Langhorne Clemens*, better known by his pen name Mark Twain, was an American writer<br>(2) Alexei Maximovich Peshkov, primarily known as *Maxim Gorky*, was a Russian and Soviet writer |
| Test instance | |
| (A) or (B) or (C) | Euler was elected a foreign member of *the Royal Swedish Academy of Sciences* |

FewRel [26] is a new large-scale supervised few-shot RE dataset, which requires models capable of handling classification task with a handful of training instances, as shown in Table 4.1. Benefiting from the FewRel dataset, there are some efforts to exploring few-shot RE [17, 53, 73] and achieve promising results. Yet, few-shot RE still remains a challenging problem for further research [18].

## 4.6 Summary

In this chapter, we introduce sentence representation learning. Sentence representation encodes the semantic information of a sentence into a real-valued representation vector, and can be utilized in further sentence classification or matching tasks. First, we introduce the one-hot representation for sentences and probabilistic language models. Secondly, we extensively introduce several neural language models, including adopting the feedforward neural networks, the convolutional neural networks, the recurrent neural networks, and the Transformer for language models. These neural models can learn rich linguistic and semantic knowledge from language modeling. Benefiting from this, the pre-trained language models trained with large-scale corpora have achieved state-of-the-art performance on various downstream NLP tasks by transferring the learned semantic knowledge from general corpora to the target tasks. Finally, we introduce several typical applications of sentence representation including text classification and relation extraction.

For further understanding of sentence representation learning and its applications, there are also some recommended surveys and books including

- Yoav, Neural network methods for natural language processing [21].
- Deng & Liu, Deep learning in natural language processing [13].

In the future, for better sentence representation, some directions are requiring further efforts:

(1) **Exploring Advanced Architectures.** The improvement of model architectures is the key factor in the success of sentence representation. From the feedforward neural networks to the Transformer, people are designing more suitable neural models for sequential inputs. Based on the Transformer, some researchers are working on new NLP architectures. For instance, Transformer-XL [10] is proposed to solve the problem of fixed-length context in the Transformer. Since the Transformer is the state-of-the-art NLP architecture, current works mainly adopt attention mechanisms. Beyond these works, is it possible to introduce more human cognitive mechanisms to neural models?

(2) **Modeling Long Documents.** The representation of long documents is an important extension of sentence representation. There are some new challenges during modeling long documents, such as discourse analysis and co-reference resolution. Although some existing works already provide document-level NLP tasks (e.g., DocRED [72]), the model performance on these tasks is still much lower than the human performance. We will also introduce the advances in document representation learning in the following chapter.

(3) **Performing Efficient Representation.** Although the combination of Transformer and large-scale data leads to very powerful sentence representation, these representation models require expensive computational cost, which limits the applications in downstream tasks. Some existing works explore to use model compression techniques for more efficient models. These techniques include knowledge distillation [60], parameter pruning [16], etc. Beyond these works, there remain lots of unsolved problems for developing better representation models, which can efficiently learn from large-scale data and provide effective vectors in downstream tasks.

# Chapter 5
# Document Representation

**Abstract** A document is usually the highest linguistic unit of natural language. Document representation aims to encode the semantic information of the whole document into a real-valued representation vector, which could be further utilized in downstream tasks. Recently, document representation has become an essential task in natural language processing and has been widely used in many document-level real-world applications such as information retrieval and question answering. In this chapter, we first introduce the one-hot representation for documents. Next, we extensively introduce topic models that learn the topic distribution of words and documents. Further, we give an introduction to distributed document representation, including paragraph vector and neural document representations. Finally, we introduce several typical real-world applications of document representation, including information retrieval and question answering.

## 5.1 Introduction

Advances in information and communication technologies offer ubiquitous access to vast amounts of information and are causing an exponential increase in the number of documents available online. While more and more textual information is available electronically, effective retrieval and mining are getting more and more difficult without the efficient organization, summarization, and indexing of document content. Therefore, document representation is playing an important role in many real-world applications, e.g., document retrieval, web search, and spam filtering. Document representation aims to represent document input into a fixed-length vector, which could describe the contents of the document, to reduce the complexity of the documents and make them easier to handle. Traditional document representation models such as one-hot document representation have achieved promising results in many document classification and clustering tasks due to their simplicity, efficiency, and often surprising accuracy.

However, the one-hot document representation model has many disadvantages. First, it loses the word order, and thus, different documents can have the same representation, as long as the same words are used. Second, it usually suffers data sparsity and high dimensionality. One-hot document representation model has very little sense about the semantics of the words or, more formally, the distances between the words. Hence, the approach for representing text documents uses multi-word terms as vector components, which are noun phrases extracted using a combination of linguistic and statistical criteria. This representation is motivated by the notion of topic models that terms should contain more semantic information than individual words. And another advantage of using terms for representing a document is its lower dimensionality compared with the traditional one-hot document representation.

Nevertheless, applying these to generation tasks remains difficult. To understand how discourse units are connected, one has to understand the communicative function of each unit, and the role it plays within the context that encapsulates it, recursively all the way up for the entire text. Identifying increasingly sophisticated human-developed features may be insufficient for capturing these patterns, but developing representation-based alternatives has also been difficult. Although document representation can capture aspects of coherent sentence structure, it is not clear how it could help in generating more broadly cohesive text.

Recently, neural network models have shown compelling results in generating meaningful and grammatical documents in sequence generation tasks like machine translation or parsing. It is partially attributed to the ability of these systems to capture local compositionally: the way neighboring words are combined semantically and syntactically to form meanings that they wish to express. Based on neural network models, many research works have developed a variety of ways to incorporate document-level contextual information. These models are all hybrid architectures in that they are recurrent at the sentence level, but use a different structure to summarize the context outside the sentence. Furthermore, some models explore multilevel recurrent architectures for combining local and global information in language modeling.

In this chapter, we first introduce the one-hot representation for documents. Next, we extensively introduce topic models that aim to learn latent topic distributions of words and documents. Further, we give an introduction on distributed document representation including paragraph vector and neural document representations. Finally, we introduce several typical real-world applications of document representations, including information retrieval and question answering.

## 5.2 One-Hot Document Representation

Majority of machine learning algorithms take a fixed-length vector as the input, so documents are needed to be represented as vectors. The bag-of-words model is the most common and simple representation method for documents. Similar to one-hot sentence representation, for a document $d = \{w_1, w_2, \ldots, w_l\}$, a bag-of-word representation $\mathbf{d}$ can be used to represent this document. Specifically, for a vocabulary $V = [w_1, w_2, \ldots, w_{|V|}]$ , the one-hot representation of word $w$ is $\mathbf{w} = [0, 0, \ldots, 1, \ldots, 0]$. Based on the one-hot word representation and a vocabulary $V$, it can be extended to represent a document as

$$\mathbf{d} = \sum_{k=1}^{l} \mathbf{w}_i, \tag{5.1}$$

where $l$ is the length of the document $d$. And similar to one-hot sentence representation, the TF-IDF method is also proposed to enhance the ability of bag-of-words representation in reflecting how important a word is to a document in a corpus.

Actually, the bag-of-words representation is mainly used as a tool of feature generation, and the most common type of features calculated from this method is word frequency appearing in the documents. This method is simple but efficient and sometimes can reach excellent performance in many real-world applications. However, the bag-of-words representation still ignores entirely the word order information, which means different documents can have the same representation as long as the same words are used. Furthermore, bag-of-words representation has little sense about the semantics of the words or, more formally, the distances between words, which means this method cannot utilize rich information hidden in the word representations.

## 5.3 Topic Model

As our collective knowledge continues to be digitized and stored in the form of news, blogs, web pages, scientific articles, books, images, audio, videos, and social networks, it becomes more difficult to find and discover what we are looking for. We need new computational tools to help organize, search, and understand these vast amounts of information.

Right now, we work with online information using two main tools—search and links. We type keywords into a search engine and find a set of documents related to them. We look at the documents in that set, possibly navigating to other linked documents. This is a powerful way of interacting with our online archive, but something is missing.

Imagine searching and exploring documents based on the themes that run through them. We might "zoom in" and "zoom out" to find specific or broader themes; we might look at how those themes changed through time or how they are connected. Rather than finding documents through keyword search alone, we might first find the theme that we are interested in, and then examine the documents related to that theme.

For example, consider using themes to explore the complete history of the New York Times. At a broad level, some of the themes might correspond to the sections of the newspaper, such as foreign policy, national affairs, and sports. We could zoom in on a theme of interest, such as foreign policy, to reveal various aspects of it, such as Chinese foreign policy, the conflict in the Middle East, and the United States' relationship with Russia. We could then navigate through time to reveal how these specific themes have changed, tracking, for example, the changes in the conflict in the Middle East over the last 50 years. And, in all of this exploration, we would be pointed to the original articles relevant to the themes. The thematic structure would be a new kind of window through which to explore and digest the collection.

But we do not interact with electronic archives in this way. While more and more texts are available online, we do not have the human power to read and study them to provide the kind of browsing experience described above. To this end, machine learning researchers have developed **probabilistic topic modeling**, a suite of algorithms that aim to discover and annotate vast archives of documents with thematic information. Topic modeling algorithms are statistical methods that analyze the words of the original texts to explore the themes that run through them, how those themes are connected, and how they change over time. Topic modeling algorithms do not require any prior annotations or labeling of the documents. The topics emerge from the analysis of the original texts. Topic modeling enables us to organize and summarize electronic archives at a scale that would be impossible by human annotation.

### *5.3.1 Latent Dirichlet Allocation*

A variety of probabilistic topic models have been used to analyze the content of documents and the meaning of words. Hofmann first introduced the probabilistic topic approach to document modeling in his Probabilistic Latent Semantic Indexing method (pLSI). The pLSI model does not make any assumptions about how the mixture weights are generated, making it difficult to test the generalization ability of the model to new documents. Thus, Latent Dirichlet Allocation (LDA) was extended from this model by introducing a Dirichlet prior to the model. LDA is believed as a simple but efficient topic model. We first describe the basic ideas of LDA [6].

The intuition behind LDA is that documents exhibit multiple topics. LDA is a statistical model of document collections that tries to capture this intuition. It is most easily described by its generative process, the imaginary random process by which the model assumes the documents arose.

We formally define a *topic* to be a distribution over a fixed vocabulary. We assume that these topics are specified before any data has been generated. Now for each document in the collection, we generate the words in a two-stage process.

1. Randomly choose a distribution over topics.
2. For each word in the document,
    - Randomly choose a topic from the distribution over topics in step #1.
    - Randomly choose a word from the corresponding distribution over the vocabulary.

This statistical model reflects the intuition that documents exhibit multiple topics. Each document exhibits the topics with different proportions (step #1); each word in each document is drawn from one of the topics (step #2b), where the selected topic is chosen from the per-document distribution over topics (step #2a).

We emphasize that the algorithms have no information about these subjects and the articles are not labeled with topics or keywords. The interpretable topic distributions arise by computing the hidden structure that likely generated the observed collection of documents.

#### 5.3.1.1 LDA and Probabilistic Models

LDA and other topic models are part of the broader field of *probabilistic modeling*. In generative probabilistic modeling, we treat our data as arising from a generative process that includes *hidden variables*. This generative process defines a joint probability distribution over both the observed and hidden random variables. Given the observed variables, we perform data analysis by using that joint distribution to compute the conditional distribution of the hidden variables. This conditional distribution is also called the *posterior distribution*.

LDA falls precisely into this framework. The observed variables are the words of the documents, the hidden variables are the topic structure, and the generative process is as described above. The computational problem of inferring the hidden topic structure from the documents is the problem of computing the posterior distribution, the conditional distribution of the hidden variables given the documents.

We can describe LDA more formally with the following notation. The topics are $\beta_{1:K}$, where each $\beta_k$ is a distribution over the vocabulary. The topic proportions for the $d$th document are $\theta_d$, where $\theta_{dk}$ is the topic proportion for topic $k$ in document $d$. The topic assignments for the $d$th document are $z_d$, where $z_{d,n}$ is the topic assignment for the $n$th word in document $d$. Finally, the observed words for document $d$ are $w_d$, where $w_{d,n}$ is the $n$th word in document $d$, which is an element from the fixed vocabulary.

With this notation, the generative process for LDA corresponds to the following joint distribution of the hidden and observed variables:

$$P(\beta_{1:K}, \theta_{1:D}, z_{1:D}, w_{1:D}) = \prod_{i=1}^{K} P(\beta_i) \prod_{d=1}^{D} P(\theta_d)(\prod_{n=1}^{N} P(z_{d,n}|\theta_d) P(w_{d,n}|\beta_{1:K}, z_{d,n}). \tag{5.2}$$

Notice that this distribution specifies the number of dependencies. For example, the topic assignment $z_{d,n}$ depends on the per-document topic proportions $\theta_d$. As another example, the observed word $w_{d,n}$ depends on the topic assignment $z_{d,n}$ and all of the topics $\beta_{1:K}$.

These dependencies define LDA. They are encoded in the statistical assumptions behind the generative process, in the particular mathematical form of the joint distribution, and in a third way, in the probabilistic graphical model for LDA. Probabilistic graphical models provide a graphical language for describing families of probability distributions. The graphical model for LDA is in Fig. 5.1. Each node is a random variable and is labeled according to its role in the generative process. The hidden nodes, the topic proportions, assignments, and topics are unshaded. The observed nodes and the words of the documents, are shaded. We use rectangles as plate notation to denote replication. The *N* plate denotes the collection of words within documents; the *D* plate denotes the collection of documents within the collection. These three representations are equivalent ways of describing the probabilistic assumptions behind LDA.

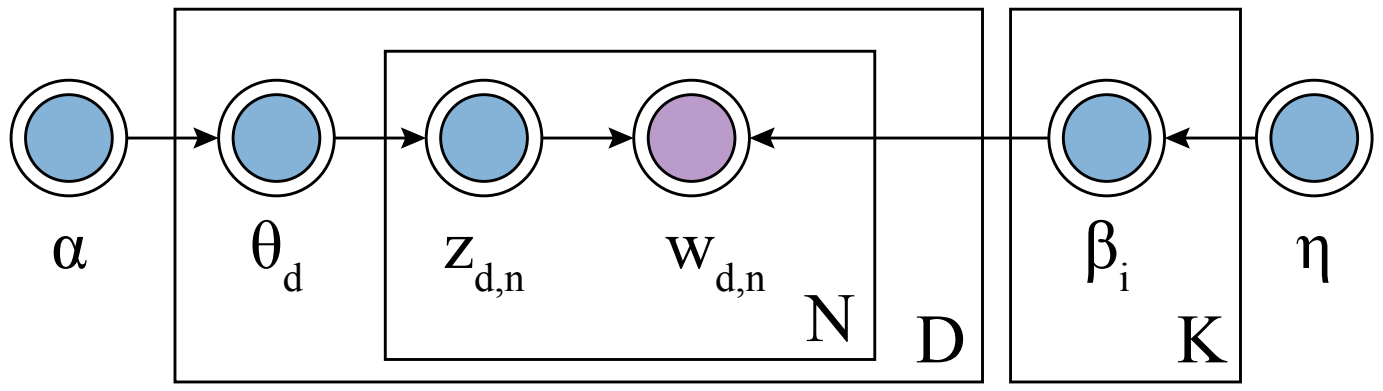

**Fig. 5.1** The architecture of graphical model for Latent Dirichlet Allocation

#### 5.3.1.2 Posterior Computation for LDA

We now turn to the computational problem, computing the conditional distribution of the topic structure given the observed documents. (As we mentioned above, this is called the *posterior*.) Using our notation, the posterior is

$$P(\beta_{1:K}, \theta_{1:D}, z_{1:D}|v_{1:D}) = \frac{P(\beta_{1:K}, \theta_{1:D}, z_{1:D}, v_{1:D})}{P(v_{1:D})}. \tag{5.3}$$

The numerator is the joint distribution of all the random variables, which can be easily computed for any setting of the hidden variables. The denominator is the marginal probability of the observations, which is the probability of seeing the observed corpus under any topic model. In theory, it can be computed by summing the joint distribution over every possible instantiation of the hidden topic structure.

Topic modeling algorithms form an approximation of the above equation by forming an alternative distribution over the latent topic structure that is adapted to be close to the true posterior. Topic modeling algorithms generally fall into two categories: sampling-based algorithms and variational algorithms.

Sampling-based algorithms attempt to collect samples from the posterior by approximating it with an empirical distribution. The most commonly used sampling algorithm for topic modeling is Gibbs sampling, where we construct a Markov chain, a sequence of random variables, each dependent on the previous—whose limiting distribution is posterior. The Markov chain is defined on the hidden topic variables for a particular corpus, and the algorithm is to run the chain for a long time, collect samples from the limiting distribution, and then approximate the distribution with the collected samples.

Variational methods are a deterministic alternative to sampling-based algorithms. Rather than approximating the posterior with samples, variational methods posit a parameterized family of distributions over the hidden structure and then find the member of that family that is closest to the posterior. Thus, the inference problem is transformed into an optimization problem. Variational methods open the door for innovations in optimization to have a practical impact on probabilistic modeling.

## *5.3.2 Extensions*

The simple LDA model provides a powerful tool for discovering and exploiting the hidden thematic structure in large archives of text. However, one of the main advantages of formulating LDA as a probabilistic model is that it can easily be used as a module in more complicated models for more complex goals. Since its introduction, LDA has been extended and adapted in many ways.

### 5.3.2.1 Relaxing the Assumptions of LDA

LDA is defined by the statistical assumptions it makes about the corpus. One active area of topic modeling research is how to relax and extend these assumptions to uncover a more sophisticated structure in the texts.

One assumption that LDA makes is the bag-of-words assumption that the order of the words in the document does not matter. While this assumption is unrealistic, it is reasonable if our only goal is to uncover the coarse semantic structure of the texts. For more sophisticated goals, such as language generation, it is patently not appropriate. There have been many extensions to LDA that model words non-exchangeable. For example, [59] develops a topic model that relaxes the bag-of-words assumption by assuming that the topics generate words conditional on the previous word; [22] develops a topic model that switches between LDA and a standard HMM. These models expand the parameter space significantly but show improved language modeling performance.

Another assumption is that the order of documents does not matter. Again, this can be seen by noticing that Eq. 5.3 remains invariant to permutations of the ordering of documents in the collection. This assumption may be unrealistic when analyzing long-running collections that span years or centuries. In such collections, we may want to assume that the topics change over time. One approach to this problem is the dynamic topic model [5], a model that respects the ordering of the documents and gives a more productive posterior topical structure than LDA.

The third assumption about LDA is that the number of topics is assumed known and fixed. The Bayesian nonparametric topic model provides an elegant solution: The collection determines the number of topics during posterior inference, and new documents can exhibit previously unseen topics. Bayesian nonparametric topic models have been extended to hierarchies of topics, which find a tree of topics, moving from more general to more concrete, whose particular structure is inferred from the data [4].

#### 5.3.2.2 Incorporating Meta-Data to LDA

In many text analysis settings, the documents contain additional information such as author, title, geographic location, links, and others that we might want to account for when fitting a topic model. There has been a flurry of research on adapting topic models to include meta-data.

The author-topic model [51] is an early success story for this kind of research. The topic proportions are attached to authors; papers with multiple authors are assumed to attach each word to an author, drawn from a topic drawn from his or her topic proportions. The author-topic model allows for inferences about authors as well as documents.

Many document collections are linked. For example, scientific papers are linked by citations, or web pages are connected by hyperlinks. And several topic models have been developed to account for those links when estimating the topics. The relational topic model of [9] assumes that each document is modeled as in LDA and that the links between documents depend on the distance between their topic proportions. This is both a new topic model and a new network model. Unlike traditional statistical models of networks, the relational topic model takes into account node attributes in modeling the links.

Other work that incorporates meta-data into topic models includes models of linguistic structure [8], models that account for distances between corpora [60], and models of named entities [42]. General-purpose methods for incorporating meta-data into topic models include Dirichlet-multinomial regression models [39] and supervised topic models [37].

#### 5.3.2.3 Acceleration

In the existing fast algorithms, it is difficult to decouple the access to $C_d$ and $C_w$ because both counts need to be updated instantly after the sampling of every token. Many algorithms have been proposed to accelerate LDA based on this equation. WarpLDA [13] is built based on a new Monte Carlo Expectation Maximization (MCEM) algorithm, which is similar to CGS, but both counts are fixed until the sampling of all tokens is finished. This scheme can be used to develop a reordering strategy to decouple the accesses to $C_d$ and $C_w$, and minimize the size of randomly accessed memory.

Specifically, WarpLDA seeks a MAP solution of the latent variables $\Theta$ and $\Phi$, with the latent topic assignments $Z$ integrated out: where $\alpha'$ and $\beta'$ are the Dirichlet hyperparameters. Reference [2] has shown that this MAP solution is almost identical with the solution of CGS, with proper hyperparameters.

Computing $\log P(\Theta, \Phi | W, \alpha', \beta')$ directly is expensive because it needs to enumerate all the $K$ possible topic assignments for each token. We, therefore, optimize its lower bound as a surrogate. Let $Q(Z)$ be a variational distribution. Then, by Jensen's inequality, the lower bound can be $\mathcal{J}(\Theta, \Phi, Q(Z))$:

$$\begin{aligned}\log P(\Theta, \Phi | W, \alpha', \beta') \geq & \mathbb{E}_Q[\log P(W, Z|\Theta, \Phi) - \log Q(Z)] + \log P(\Theta|\alpha') + \log P(\Phi|\beta') \\ \triangleq & \mathcal{J}(\Theta, \Phi, Q(Z)).\end{aligned} \tag{5.4}$$

An Expectation Maximization (EM) algorithm is implemented to find a local maximum of the posterior $P(\Theta, \Phi | W, \alpha', \beta')$, where the E-step maximizes $\mathcal{J}$ with respect to the variational distribution $Q(Z)$ and the M-step maximizes $\mathcal{J}$ with respect to the model parameters $(\Theta, \Phi)$, while keeping $Q(Z)$ fixed. One can prove that the optimal solution at E-step is $Q(Z) = P(Z|W, \Theta, \Phi)$ without further assumption on $Q$. We apply Monte Carlo approximation on the expectation in Eq. 5.4,

$$\mathbb{E}_Q[\log P(W, Z|\Theta, \Phi) - \log Q(Z)] \approx \frac{1}{S} \sum_{s=1}^{S} \log P(W, Z^{(s)}|\Theta, \Phi) - \log Q(Z^{(s)}), \tag{5.5}$$

where $Z^{(1)}, \ldots, Z^{(S)} \sim Q(Z) = P(Z|W, \Theta, \Phi)$. The sample size is set as $S = 1$ and the model uses $Z$ as an abbreviation of $Z^{(1)}$.

**Sampling** $Z$: Each dimension of $Z$ can be sampled independently:

$$Q(z_{d,n} = k) \propto P(W, Z|\Theta, \Phi) \propto \theta_{dk} \phi_{w_{d,n},k}. \tag{5.6}$$

**Optimizing** $\Theta, \Phi$: With the Monte Carlo approximation, we have

$$\begin{aligned}\mathcal{J} &\approx \log P(W, Z|\Theta, \Phi) + \log P(\Theta|\alpha') + \log P(\Phi|\beta') + \text{const.} \\ &= \sum_{d,k} (C_{dk} + \alpha'_k - 1) \log \theta_{dk} + \sum_{k,w} (C_{kw} + \beta' - 1) \log \phi_{kw} + \text{const.},\end{aligned} \tag{5.7}$$

and with the optimal solutions, we have

$$\hat{\theta}_{dk} \propto C_{dk} + \alpha'_k - 1, \quad \hat{\phi}_{wk} = \frac{C_{wk} + \beta' - 1}{C_k + \bar{\beta}' - V}. \tag{5.8}$$

Instead of computing and storing $\hat{\Theta}$ and $\hat{\Phi}$, we compute and store $C_d$ and $C_w$ to save memory because the latter are sparse. Plug Eqs. 5.8–5.6, and let $\alpha = \alpha' - 1$, $\beta = \beta' - 1$, we get the full MCEM algorithm, which iteratively performs the following two steps until a given iteration number is reached:

- E-step: We can sample $z_{d,n} \sim Q(z_{d,n} = k)$ according to

$$Q(z_{d,n} = k) \propto (C_{dk} + \alpha_k)\frac{C_{wk} + \beta_w}{C_k + \bar{\beta}}. \tag{5.9}$$

- M-step: Compute $C_d$ and $C_w$ by $Z$.

Note the resemblance intuitively justifies why MCEM leads to similar results with CGS. The difference between MCEM and CGS is that MCEM updates the counts $C_d$ and $C_w$ after sampling all $z_{d,n}$s, while CGS updates the counts instantly after sampling each $z_{d,n}$. The strategy that MCEM updates the counts after sampling all $z_{d,n}$s is called *delayed count update*, or simply *delayed update*. MCEM can be viewed as a CGS with a delayed update, which has been widely used in other algorithms [1, 41]. While previous work uses the delayed update as a trick, we at this moment present a theoretical guarantee to converge to a MAP solution. The delayed update is essential for us to decouple the accesses of $C_d$ and $C_w$ to improve cache locality, without affecting the correctness.

## 5.4 Distributed Document Representation

To address the disadvantages of bag-of-words document representation, [31] proposes paragraph vector models, including the version with Distributed Memory (PV-DM) and the version with Distributed Bag-of-Words (PV-DBOW). Moreover, researchers also proposed several hierarchical neural network models to represent documents. In this section, we will introduce these models in detail.

### *5.4.1 Paragraph Vector*

As shown in Fig. 5.2, paragraph vector maps every paragraph to a unique vector, represented by a column in the matrix **P** and maps every word to a unique vector, represented by a column in word embedding matrix **E**. The paragraph vector and word vectors are averaged or concatenated to predict the next word in a context. More formally, compared to the word vector framework, the only change in this model is in the following equation, where $h$ is constructed from **E** and **P**.

$$y = \text{Softmax}(h(w_{t-k}, \ldots, w_{t+k}; \mathbf{E}, \mathbf{P})), \tag{5.10}$$

where $h$ is constructed by the concatenation or average of word vectors extracted from **E** and **P**.

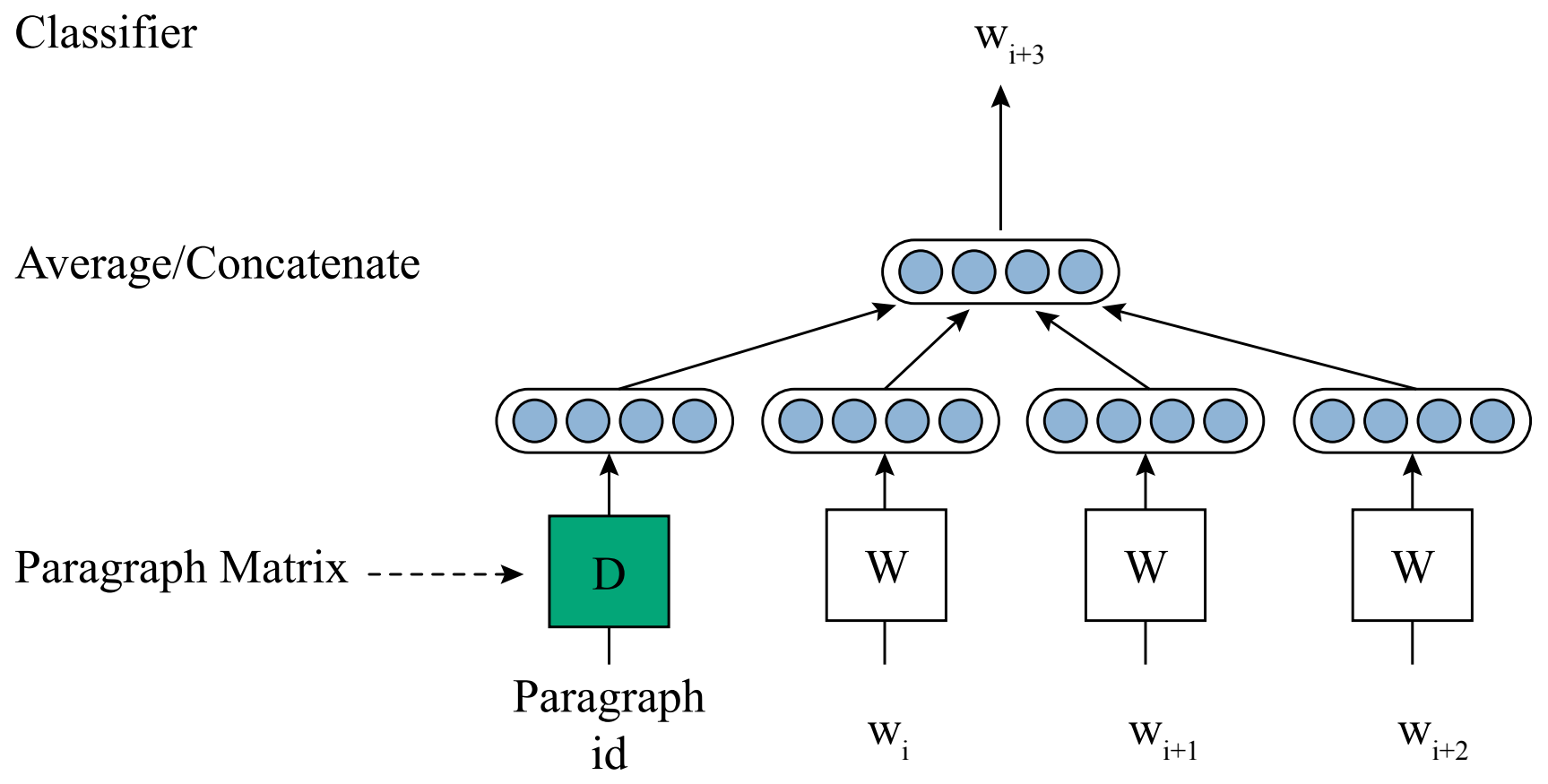

**Fig. 5.2** The architecture of PV-DM model

The other part of this model is that given a sequence of training words $w_1, w_2, w_3$, ..., $w_l$, the objective of the paragraph vector model is to maximize the average log probability:

$$\mathscr{O} = \frac{1}{l} \sum_{i=k}^{l-k} \log P(w_i \mid w_{i-k}, \ldots, w_{i+k}). \tag{5.11}$$

And the prediction task is typically done via a multi-class classifier, such as softmax. Thus, the probability equation is

$$P(w_i \mid w_{i-k}, \ldots, w_{i+k}) = \frac{e^{y_{w_i}}}{\sum_j e^{y_j}}. \tag{5.12}$$

The paragraph token can be thought of as another word. It acts as a memory that remembers what is missing from the current context, or the topic of the paragraph. For this reason, this model is often called the Distributed Memory Model of Paragraph Vectors (PV-DM).

The above method considers the concatenation of the paragraph vector with the word vectors to predict the next word in a text window. Another way is to ignore the context words in the input, but force the model to predict words randomly sampled from the paragraph in the output. In reality, what this means is that at each iteration of stochastic gradient descent, we sample a text window, then sample a random word from the text window and form a classification task given the Paragraph Vector. This technique is shown in Fig. 5.3. This version is named the Distributed Bag-of-Words version of Paragraph Vector (PV-DBOW), as opposed to the Distributed Memory version of Paragraph Vector (PV-DM) in the previous section.

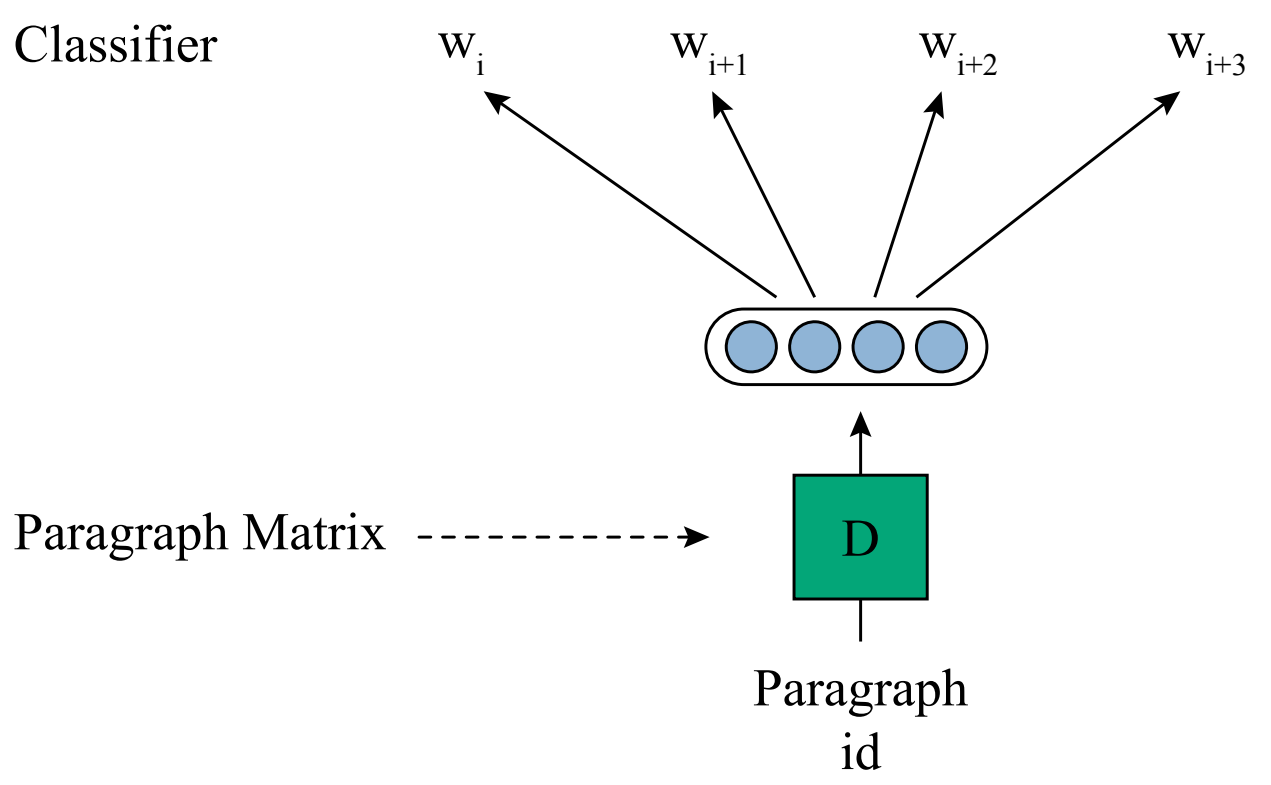

**Fig. 5.3** The architecture of PV-DBOW model

In addition to being conceptually simple, this model requires to store fewer data. The data only needed to be stored is the softmax weights as opposed to both softmax weights and word vectors in the previous model. This model is also similar to the Skip-gram model in word vectors.

## *5.4.2 Neural Document Representation*

In this part, we introduce two main kinds of neural networks for document representation including document-context language model and hierarchical document autoencoder.

### 5.4.2.1 Document-Context Language Model

Recurrent architectures can be used to combine local and global information in document language modeling. The simplest such model would be to train a single RNN, ignoring sentence boundaries as mentioned above; the last hidden state from the previous sentence $t-1$ is used to initialize the first hidden state in sentence $t$. In such an architecture, the length of the RNN is equal to the number of tokens in the document; in typical genres such as news texts, this means training RNNs from sequences of several hundred tokens, which introduces two problems: (1) **Information decay** In a sentence with thirty tokens (not unusual in news text), the contextual information from the previous sentence must be propagated through the recurrent dynamics thirty times before it can reach the last token of the current sentence. Meaningful document-level information is unlikely to survive such a long pipeline. (2) **Learning** It is notoriously difficult to train recurrent architectures that involve many time steps.

In the case of an RNN trained on an entire document, back-propagation would have to run over hundreds of steps, posing severe numerical challenges.

To address these two issues, [28] proposes to use multilevel recurrent structures to represent documents, thereby successfully efficiently leveraging document-level context in language modeling. They first proposed Context-to-Context Document-Context Language Model (ccDCLM), which assumes that contextual information from previous sentences needs to be able to "short-circuit" the standard RNN, so as to more directly impact the generation of words across longer spans of text. Formally, we have

$$\mathbf{c}_{t-1} = \mathbf{h}_{t-1,l}, \tag{5.13}$$

where $l$ is the length of sentence $t-1$. The ccDCLM model then creates additional paths for this information to impact each hidden representation in the current sentence $t$. Writing $\mathbf{w}_{t,n}$ for the word representation of the $n$th word in the $t$th sentence, we have

$$\mathbf{h}_{t,n} = g_\theta(\mathbf{h}_{t,n-1}, f(\mathbf{w}_{t,n}, \mathbf{c}_{t-1}), \tag{5.14}$$

where $g_\theta(\cdot)$ is the activation function parameterized by $\theta$ and $f(\cdot)$ is a function that combines the context vector with the input $\mathbf{x}_{t,n}$ for the hidden state. Here we simply concatenate the representations,

$$f(\mathbf{x}_{t,n}, \mathbf{c}_{t-1}) = [\mathbf{x}_{t,n}; \mathbf{c}_{t-1}]. \tag{5.15}$$

The emission probability for $\mathbf{y}_{t,n}$ is then computed from $\mathbf{h}_{t,n}$ as in the standard RNNLM. The underlying assumption of this model is that contextual information should impact the generation of each word in the current sentence. The model, therefore, introduces computational "short-circuits" for cross-sentence information, as illustrated in Fig. 5.4.

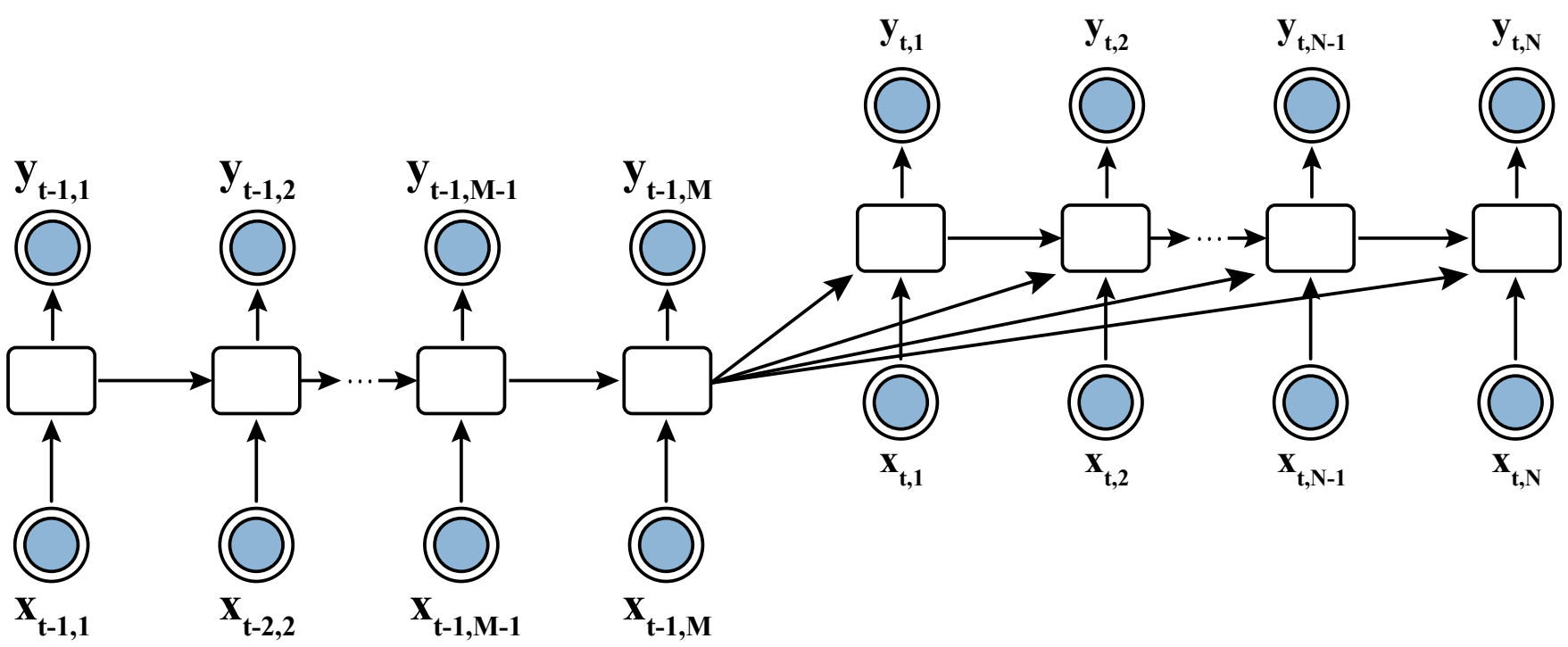

**Fig. 5.4** The architecture of ccDCLM model

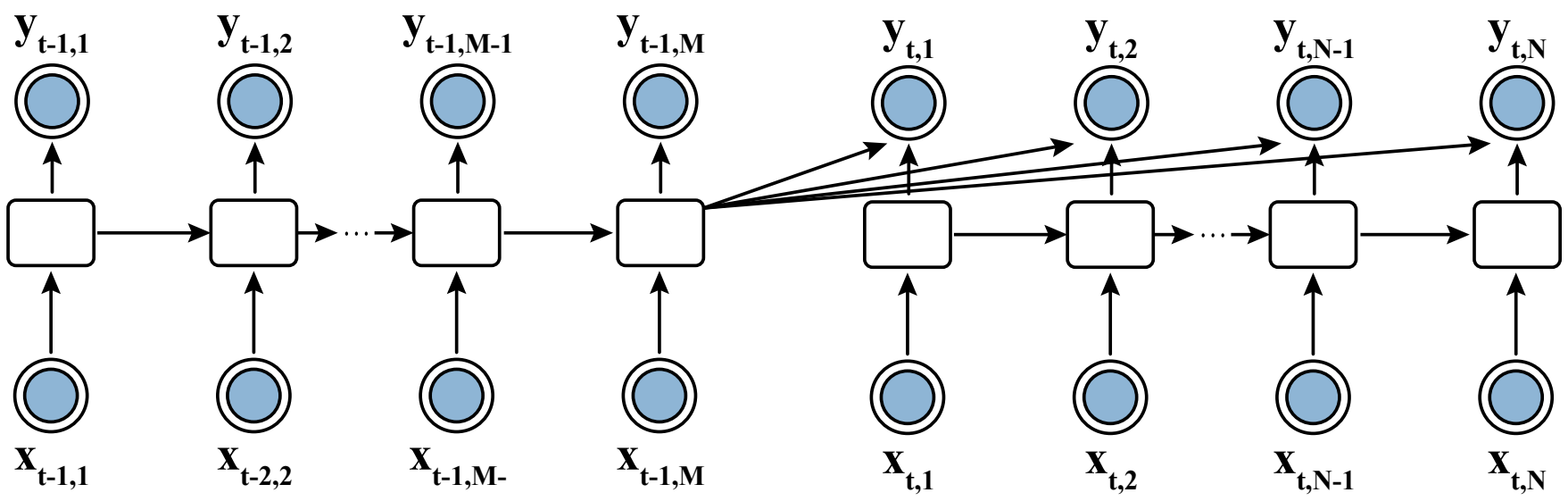

**Fig. 5.5** The architecture of coDCLM model

Besides, they also proposed Context-to-Output Document-Context Language Model (coDCLM). Rather than incorporating the document context into the recurrent definition of the hidden state, the coDCLM model pushes it directly to the output, as illustrated in Fig. 5.5. Let $\mathbf{h}_{t,n}$ be the hidden state from a conventional RNNLM of sentence $t$,

$$\mathbf{h}_{t,n} = g_\theta(\mathbf{h}_{t,n-1}, \mathbf{x}_{t,n}). \tag{5.16}$$

Then, the context vector $\mathbf{c}_{t-1}$ is directly used in the output layer as

$$\mathbf{y}_{t,n} \sim \text{Softmax}(\mathbf{W}_h \mathbf{h}_{t,n} + \mathbf{W}_c \mathbf{c}_{t-1} + \mathbf{b}). \tag{5.17}$$

#### 5.4.2.2 Hierarchical Document Autoencoder

Reference [33] also proposes hierarchical document autoencoder to represent documents. The model draws on the intuition that just as the juxtaposition of words creates a joint meaning of a sentence, the juxtaposition of sentences also creates a joint meaning of a paragraph or a document.

They first obtain representation vectors at the sentence level by putting one layer of LSTM (denoted as $\text{LSTM}^{word}_{encode}$) on top of its containing words:

$$h^w_t(\text{enc}) = \text{LSTM}^{word}_{encode}(\mathbf{w}_t, h^v_{t-1}(\text{enc})). \tag{5.18}$$

The vector output at the ending time step is used to represent the entire sentence as

$$\mathbf{s} = h^w_{end_s}. \tag{5.19}$$

To build representation $e_D$ for the current document/paragraph, another layer of LSTM (denoted as $\text{LSTM}^{sentence}_{encode}$) is placed on top of all sentences, computing representations sequentially for each time step:

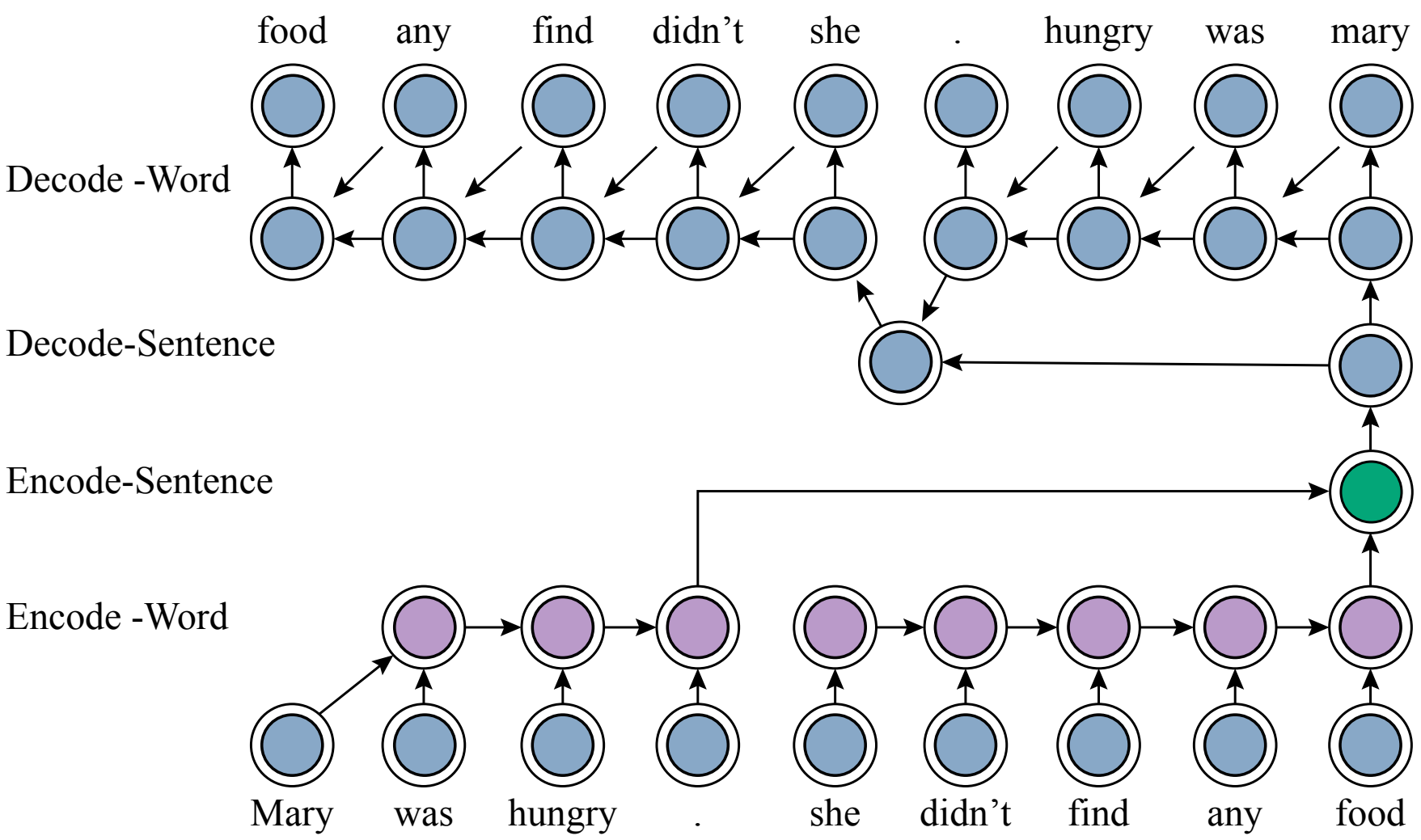

**Fig. 5.6** The architecture of hierarchical document autoencoder

$$h_t^s(\text{enc}) = \text{LSTM}_{encode}^{sentence}(\mathbf{s}, h_{t-1}^s(\text{enc})). \tag{5.20}$$

Representation $h_{end_D}^s$ computed at the final time step is used to represent the entire document: $\mathbf{d} = h_{end_D}^s$.

Thus one LSTM operates at the token level, leading to the acquisition of sentence-level representations that are then used as inputs into the second LSTM that acquires document-level representations, in a hierarchical structure.

As with encoding, the decoding algorithm operates on a hierarchical structure with two layers of LSTMs. LSTM outputs at sentence level for time step $t$ are obtained by

$$h_t^s(\text{dec}) = \text{LSTM}_{decode}^{sentence}(\mathbf{s}_t, h_{t-1}^s(\text{dec})). \tag{5.21}$$

The initial time step $h_0^s(d) = e_D$, the end-to-end output from the encoding procedure $h_t^s(d)$ is used as the original input into $\text{LSTM}_{decode}^{word}$ for subsequently predicting tokens within sentence $t + 1$. $\text{LSTM}_{decode}^{word}$ predicts tokens at each position sequentially, the embedding of which is then combined with earlier hidden vectors for the next time-step prediction until the $end_s$ token is predicted. The procedure can be summarized as follows:

$$h_t^w(\text{dec}) = \text{LSTM}_{decode}^{sentence}(\mathbf{w}_t, h_{t-1}^w(\text{dec})), \tag{5.22}$$

$$P(w|\cdot) = \text{Softmax}(\mathbf{w}, h_{t-1}^w(\text{dec})). \tag{5.23}$$

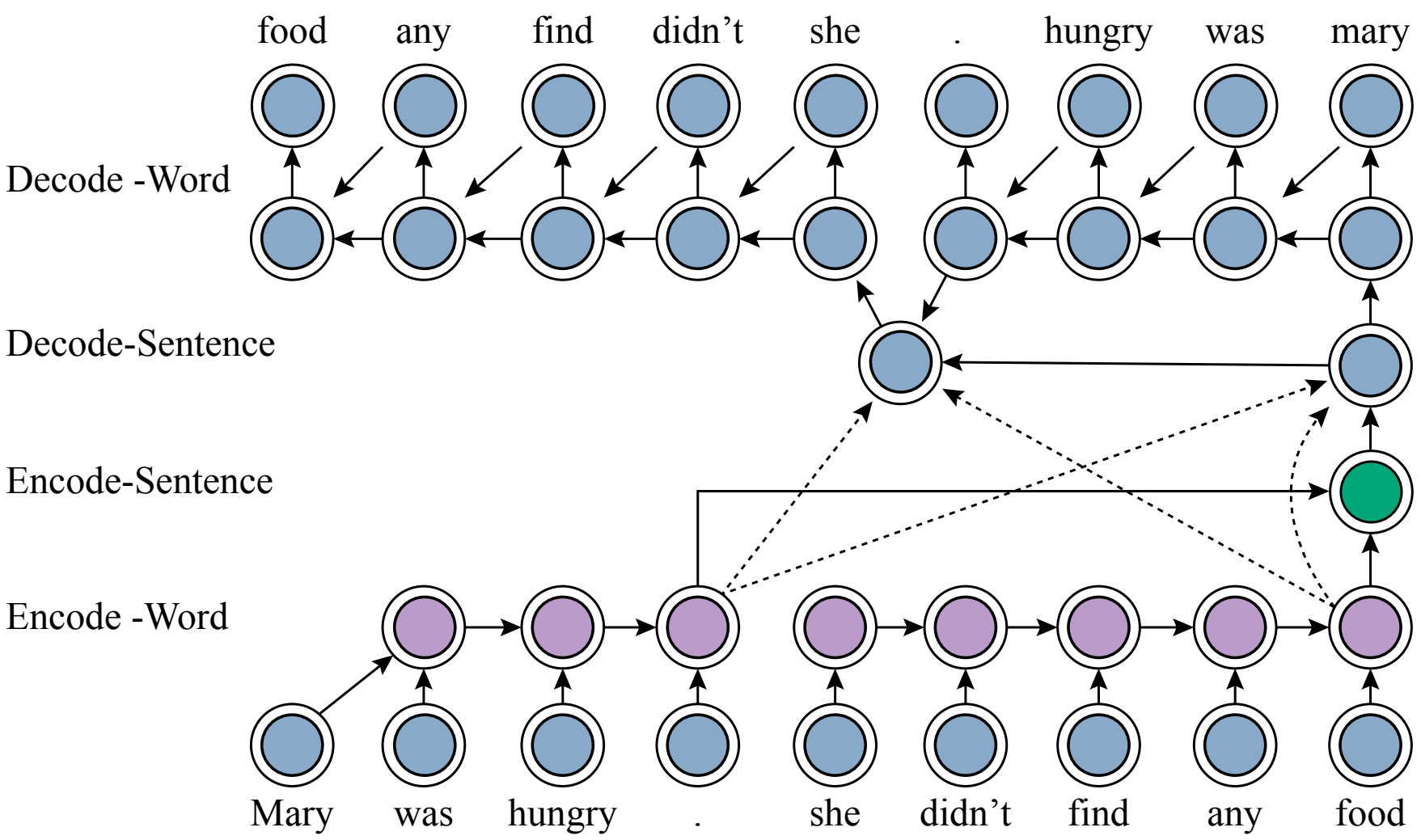

**Fig. 5.7** The architecture of hierarchical document autoencoder with attentions

During decoding, $\text{LSTM}^{word}_{decode}$ generates each word token $w$ sequentially and combines it with earlier LSTM-outputted hidden vectors. The LSTM hidden vector computed at the final time step is used to represent the current sentence.

This is passed to $\text{LSTM}^{sentence}_{decode}$, combined with $h^s_t$ for the acquisition of $h_{t+1}$, and outputted to the next time step in sentence decoding. For each time step $t$, $\text{LSTM}^{sentence}_{decode}$ has to first decide whether decoding should proceed or come to a full stop: we add an additional token $end_D$ to the vocabulary. Decoding terminates when token $end_D$ is predicted. Details are shown in Fig. 5.6.

Attention models adopt a look-back strategy by linking the current decoding stage with input sentences in an attempt to consider which part of the input is most responsible for the current decoding state (Fig. 5.7).

Let $H = \{h^s_1(e), h^s_2(e), \ldots, h^s_N(e)\}$ be the collection of sentence-level hidden vectors for each sentence from the inputs, outputted from $\text{LSTM}^{sentence}_{encode}$. Each element in H contains information about input sequences with a strong focus on the parts surrounding each specific sentence (time step). During decoding, suppose that $e^s_t$ denotes the sentence-level embedding at current step and that $h^s_{t-1}(\text{dec})$ denotes the hidden vector outputted from $\text{LSTM}^{sentence}_{decode}$ at previous time step $t-1$. Attention models would first link the current-step decoding information, i.e., $h^s_{t-1}(\text{dec})$ which is outputted from $\text{LSTM}^{sentence}_{dec}$ with each of the input sentences $i \in [1, N]$, characterized by a strength indicator $v_i$:

$$v_i = \mathbf{U}^\top f(\mathbf{W}_1 \cdot h^s_{t-1}(\text{dec}) + \mathbf{W}_2 \cdot h^s_i(\text{enc})), \tag{5.24}$$

where $\mathbf{W}_1, \mathbf{W}_2 \in \mathbb{R}^{K \times K}$, $\mathbf{U} \in \mathbb{R}^{K \times 1}$. $v_i$ is then normalized

$$\alpha_i = \frac{\exp(v_i)}{\sum_j \exp(v_j)}. \tag{5.25}$$

The attention vector is then created by averaging weights over all input sentences:

$$\mathbf{m}_t = \sum_{i=1}^{N_D} \alpha_i h_i^s(\text{enc}) \tag{5.26}$$

## 5.5 Applications

In this section, we will introduce several applications on document level analysis based on representation learning.

### *5.5.1 Neural Information Retrieval*

Information retrieval aims to obtain relevant resources from a large-scale collection of information resources. As shown in Fig. 5.8, given the query "Steve Jobs" as input, the search engine (a typical application of information retrieval) provides relevant web pages for users. Traditional information retrieval data consists of search queries and document collections $D$. And the ground truth is available through explicit human judgments or implicit user behavior data such as click-through rate.

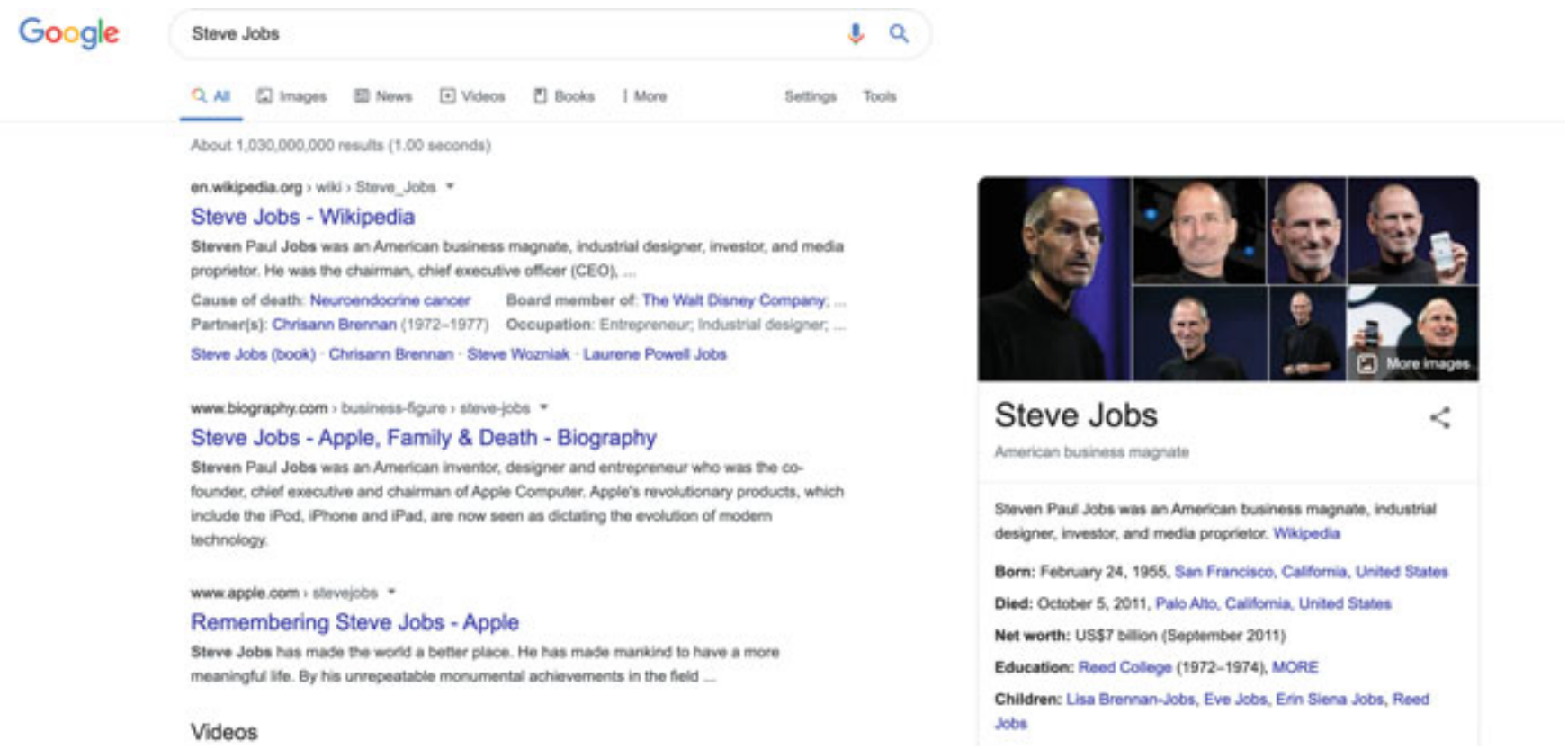

**Fig. 5.8** An example of information retrieval

For the given query $q$ and document $d$, traditional information retrieval models estimate their relevance through lexical matches. Neural information retrieval models pay more attention to garner the query and document relevance from semantic matches. Both lexical and semantic matches are essential for neural information retrieval. Thriving from neural network black magic, it helps information retrieval models catch more sophisticated matching features and have achieved the state of the art in the information retrieval task [17].

Current neural ranking models can be categorized into two groups: representation-based and interaction-based [23]. The earlier works mainly focus on representation-based models. They learn good representations and match them in the learned representation space of queries and documents. Interaction-based methods, on the other hand, model the query-document matches from the interactions of their terms.

#### 5.5.1.1 Representation-Based Neural Ranking Models

The representation-based methods directly match the query and documents by learning two distributed representations, respectively, and then compute the matching score based on the similarity between them. In recent years, several deep neural models have been explored based on such Siamese architecture, which can be done by feedforward layers, convolutional neural networks, or recurrent neural networks.

Reference [26] proposes Deep Structured Semantic Models (DSSM) first to hash words to the letter-trigram-based representation. And then use a multilayer fully connected neural network to encode a query (or a document) as a vector. The relevance between the query and document can be simply calculated with the cosine similarity. Reference [26] trains the model by minimizing the cross-entropy loss on click-through data where each training sample consists of a query $q$, a positive document $d^+$, and a uniformly sampled negative document set $D^-$:

$$\mathscr{L}_{DSSM}(q, d^+, D^-) = -\log\left(\frac{e^{r\cdot\cos(\mathbf{q},\mathbf{d}^+)}}{\sum_{d\in D} e^{r\cdot\cos(\mathbf{q},\mathbf{d})}}\right), \quad (5.27)$$

where $D = d^+ \cup D^-$.

Furthermore, CDSSM [54] and ARC-I [25] utilize convolutional neural network (CNN), while LSTM-RNN [44] adopts recurrent neural network with Long Short-Term Memory (LSTM) units to represent a sentence better. Reference [53] also comes up with a more sophisticated similarity function by leveraging additional layers of the neural network.

#### 5.5.1.2 Interaction-Based Neural Ranking Models

The interaction-based neural ranking models learn word-level interaction patterns from query-document pairs, as shown in Fig. 5.9. And they provide an opportunity to compare different parts of the query with different parts of the document individually and aggregate the partial evidence of relevance. ARC-II [25] and MatchPyramind [45] utilize convolutional neural network to capture complicated patterns from word-level interactions. The Deep Relevance Matching Model (DRMM) uses pyramid pooling (histogram) to summarize the word-level similarities into ranking models [23]. There are also some works establishing position-dependent interactions for ranking models [27, 46].

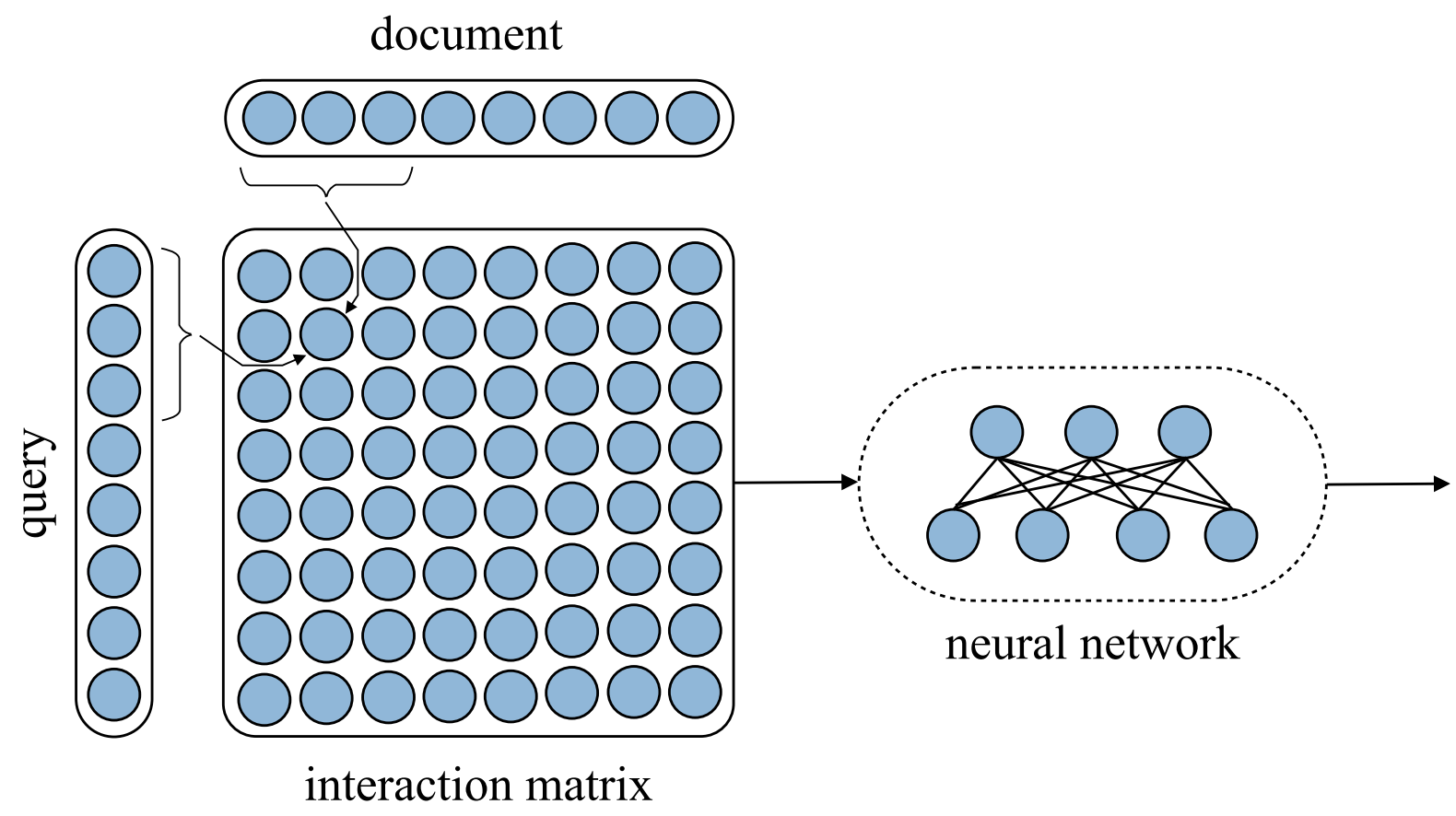

**Fig. 5.9** The architecture of interaction-based neural ranking models

Kernel-based Neural Ranking Model (K-NRM) [66] and its convolutional version Conv-KNRM [17] achieve the state of the art in neural information retrieval. K-NRM first establishes a translation matrix $\mathbf{M}$ in which each element $\mathbf{M}_{ij}$ is the cosine similarity of $i$th word in $q$ and $j$th word in $d$. Then K-NRM utilizes kernels to convert translation matrix $\mathbf{M}$ to ranking features $\phi(\mathbf{M})$ :

$$\phi(\mathbf{M}) = \sum_{i=1}^{n} \log \mathbf{K}(\mathbf{M}_i), \tag{5.28}$$

$$\mathbf{K}(\mathbf{M}_i) = \{K_1(\mathbf{M}_i), \ldots, K_K(\mathbf{M}_i)\}. \tag{5.29}$$

Each RBF kernel $K_k$ calculates how word pair similarities are distributed:

$$K_k(\mathbf{M}_i) = \sum_j \exp\left(-\frac{(\mathbf{M}_{ij} - \mu_k)^2}{2\sigma_K^2}\right). \tag{5.30}$$

Then the relevance of $q$ and $d$ is calculated by a ranking layer:

$$f(q, d) = \tanh(\mathbf{w}^\top \phi(\mathbf{M}) + b), \tag{5.31}$$

where $\mathbf{w}$ and $b$ are trainable parameters.

Reference [66] trains the model by minimizing pair-wise loss on click-through data:

$$\mathscr{L} = \sum_{q} \sum_{d^+, d^- \in D^{+,-}} \max(0, 1 - f(q, d^+) + f(q, d^-)). \tag{5.32}$$

For the given query $q$, $D^{+,-}$ are the pair-wise preferences from the ground truth. $d^+$ and $d^-$ are two documents such that $d^+$ is more relevant with $q$ than $d^-$. Conv-KNRM extends K-NRM to model $n$-gram semantic matches based on the convolutional neural network which can leverage snippet information.

#### 5.5.1.3 Summary

Representation-based models and interaction-based models extract match features from overall and local aspects, respectively. They can also be combined for further improvements [40].

Recently, large-scale knowledge graphs such as DBpedia, Yago, and Freebase have emerged. Knowledge graphs contain human knowledge about real-world entities and become an opportunity for search systems to understand queries and documents better. The emergence of large-scale knowledge graphs has motivated the development of entity-oriented search, which brings in entities and semantics from the knowledge graphs and has dramatically improved the effectiveness of feature-based search systems.

Entity-oriented search and neural ranking models push the boundary of matching from two different perspectives. Reference [36] incorporates semantics from knowledge graphs into the neural ranking, such as entity descriptions and entity types. This work significantly improves the effectiveness and generalization ability of interaction-based neural ranking models. However, how to fully leverage semi-structured knowledge graphs and establish semantic relevance between queries and documents remains an open question.

Information retrieval has been widely used in many natural language processing tasks such as reading comprehension and question answering. Therefore, it is no doubt that neural information retrieval will lead to a new tendency for these tasks.

### *5.5.2 Question Answering*

Question Answering (QA) is one of the most important tasks and so are document-level applications in NLP. Many efforts have been invested in QA, especially in machine reading comprehension and open-domain QA. In this section, we will introduce the advances in these two tasks, respectively.

#### 5.5.2.1 Machine Reading Comprehension

As shown in Fig. 5.10, machine reading comprehension aims to determine the answer $a$ to the question $q$ given a passage $p$. The task could be viewed as a supervised learning problem: given a collection of training examples $\{(p_i, q_i, a_i)\}_{i=1}^{n}$, we want to learn a mapping $f(\cdot)$ that takes the passage $p_i$ and corresponding question $q_i$ as inputs and outputs $\hat{a}_i$, where $evaluate(\hat{a}_i, a_i)$ is maximized. The evaluation metric is typically correlated with the answer type, which will be discussed in the following.

Generally, the current machine reading comprehension task could be divided into four categories depending on the answer types according to [10], i.e., cloze style, multiple choices, span prediction, and free-form answer.

The cloze style task such as CNN/DAILY MAIL [24] consists of fill-in-the-blank sentences where the question contains a placeholder to be filled in. The answer $a$ is either chosen from a predefined candidate set $|A|$ or from the vocabulary $|V|$. The multiple-choice task such as RACE [30] and MCTEST [50] aims to select the best answer from a set of answer choices. It is typical to use accuracy to measure the performance on these two tasks: the percentage of correctly answered questions in the whole example set, since the question could be either correctly answered or not from the given hypothesized answer set.

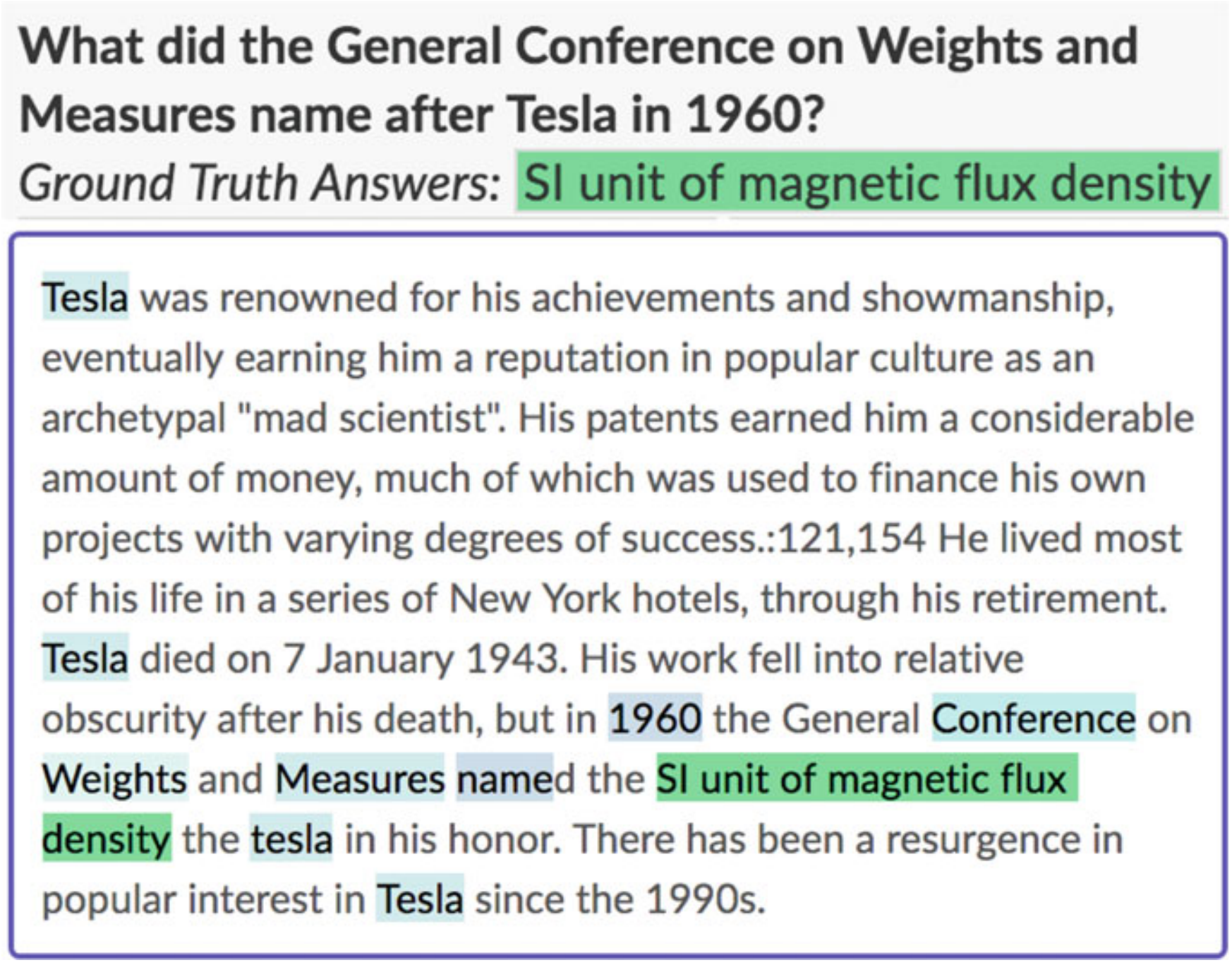

**Fig. 5.10** An example of machine reading comprehension from SQuAD [49]

The span prediction task such as SQuAD [49] is perhaps the most widely adopted task among all, since it takes compromises between flexibility and simplicity. The task is to extract a most likely text span from the passage as the answer to the question, which is usually modeled as predicting the start position $idx_{start}$ and end position $idx_{end}$ of the answer span. To evaluate the predicted answer span $\hat{a}$, we typically use two evaluation metrics proposed by [49]. Exact match assigns full score 1.0 to the predicted answer span $\hat{a}$ if it exactly equals the ground truth answer $a$, otherwise 0.0. F1 score measures the degree of overlap between $\hat{a}$ and $a$ by computing a harmonic mean of the precision and recall.

The free-form answer task such as MS MARCO [43] does not restrict the answer form or length and is also referred to as *generative question answering*. It is practical to model the task as a sequence generation problem, where the discrete token-level prediction was made. Currently, a consensus on what is the ideal evaluation metrics has not been achieved. It is common to adopt standard metrics in machine translation and summarization, including ROUGE [34] and BLEU [57].

As a critical component in the question answering system, the surging neural-based machine reading comprehension models have greatly boosted the task of question answering in the last decades.

The first attempt [24] to apply neural networks on machine reading comprehension constructs bidirectional LSTM reader models along with attention mechanisms. The work introduces two reader models, i.e., the attentive reader and the impatient reader, as shown in Fig. 5.11. After encoding the passage and the query into hidden states using LSTMs, the attentive reader computes a scalar distribution $s(t)$ over the passage tokens and uses it to compute the weighted sum of the passage hidden states $r$. The impatient reader extends this idea further by recurrently updating the weighted sum of passage hidden states after it has seen each query token.

The attention mechanisms used in reading comprehension could be viewed as a variant of Memory Networks [64]. Memory Networks use long-term memory units to store information for inference dynamically. Typically, given an input $x$, the model first converts it into an internal feature representation $F(x)$. Then, the model can update the designated memory units $m_i$ given the new input: $m_i = g(m_i, F(x), m)$, or generate output features $o$ given the new input and the memory states: $o = f(F(x), m)$. Finally, the model converts the output into the response with the desired format: $r = R(o)$. The key takeaway of Memory Networks is the retaining and updating of some internal memories that captivate global information. We will see how this idea is further extended in some sophisticated models.

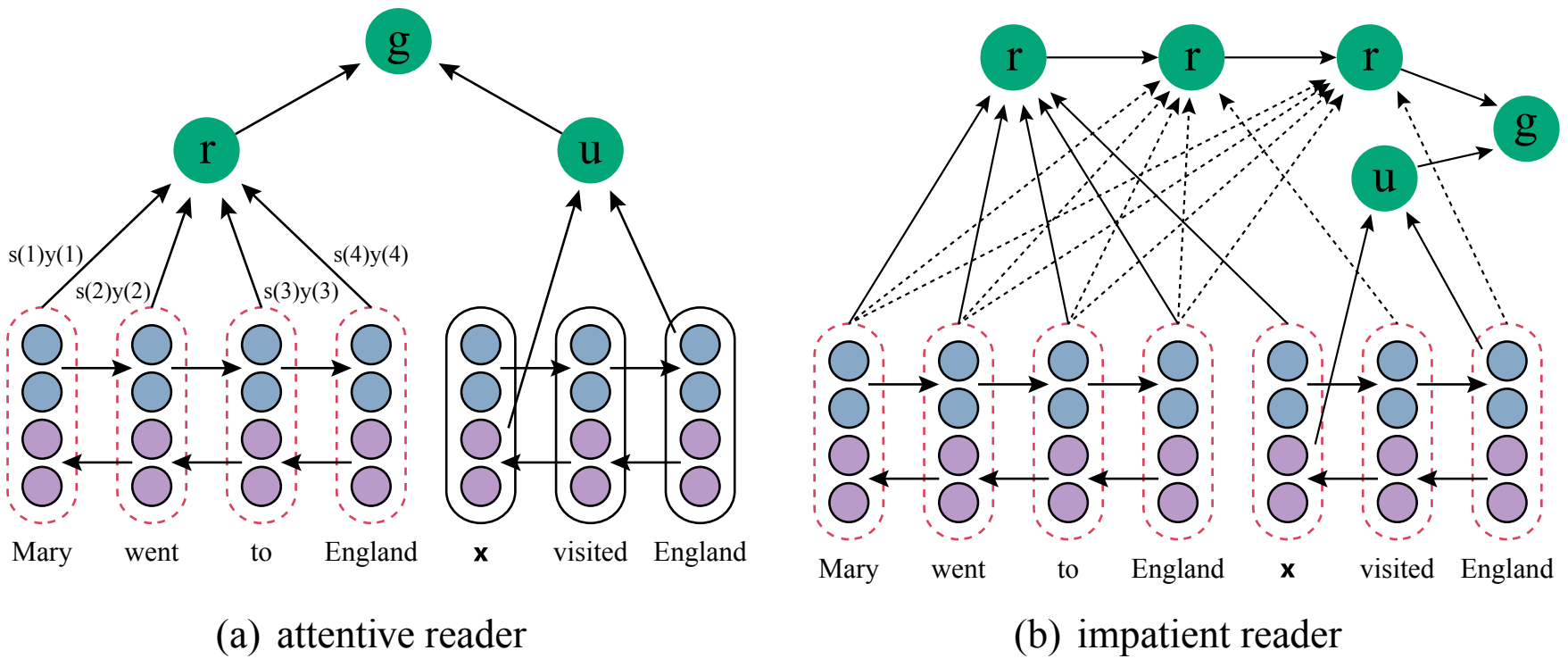

**Fig. 5.11** The architecture of bidirectional LSTM reader model

It is no doubt that the application of attention to machine reading comprehension greatly promotes researches in this field. Following [11], the work [24] modifies the method to compute attention and simplify the prediction layer in the attentive reader. Instead of using $tanh(\cdot)$ to compute the relevance between the passage representations $\{\tilde{\mathbf{p}}_i\}_{i=1}^n$ and the query hidden state $\mathbf{q}$ (see Eq. 5.33), Chen et al. use the bilinear terms to directly capture the passage-query alignment (see Eq. 5.34).

$$\alpha_i = \text{Softmax}_i(\tanh(\mathbf{W}_1\tilde{\mathbf{p}}_i + \mathbf{W}_2\mathbf{q})), \tag{5.33}$$

$$\alpha_i = \text{Softmax}_i(\mathbf{q}^\top \mathbf{W}_3\tilde{\mathbf{p}}_i). \tag{5.34}$$

Most machine reading comprehension models follow the same paradigm to locate the start and endpoint of the answer span. As shown in Fig. 5.12, while encoding the passage, the model retains the length of the sequence and encodes the question into a fixed-length hidden representation $\mathbf{q}$. The question's hidden vector is then used as a pointer to scan over the passage representation $\{\mathbf{p}_i\}_{i=1}^n$ and compute scores on every position in the passage. While maintaining this similar architecture, most machine reading comprehension models vary in the interaction methods between the passage and the question. In the following, we will introduce several classic reading comprehension architectures that follow this paradigm.

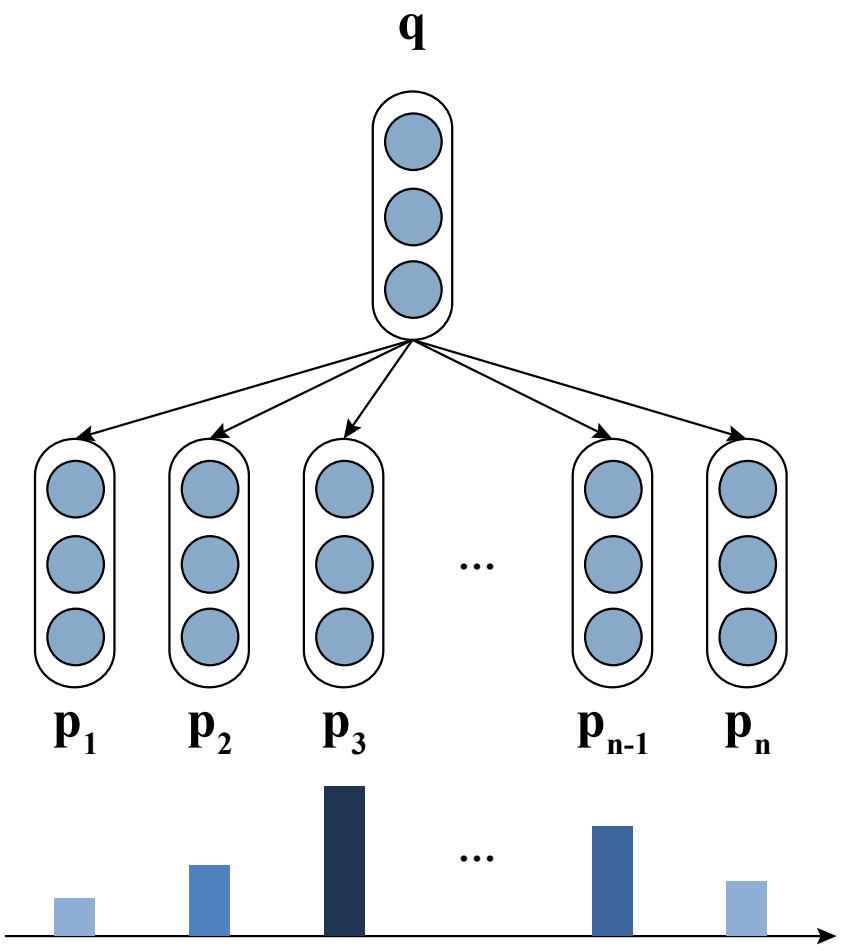

**Fig. 5.12** The architecture of classic machine reading comprehension models

First, we introduce BiDAF, which is short for *Bi-Directional Attention Flow* [52]. The BiDAF network consists of the token embedding layer, the contextual embedding layer, the bi-directional attention flow layer, the LSTM modeling layer, and the softmax output layer, as shown in Fig. 5.13.

The token embedding layer consists of two levels. First, the character embedding layer encodes each word in character level by adopting a 1D convolutional neural network (CNN). Specifically, for each word, characters are embedded into fixed-length vectors, which are considered as 1D input for CNNs. The outputs are then max-pooled along the embedding dimension to obtain a single fixed-length vector. Second, the word embedding layer uses pretrained word vectors, i.e., GloVe [47], to map each word into a high-dimensional vector directly.

Then the concatenation of the two vectors is fed into a two-layer Highway Network [56]. Equation 5.35 shows one layer of the highway network used in the paper, where $H_1(\cdot)$ and $H_2(\cdot)$ represent two affine transformations:

$$\mathbf{g} = \text{Sigmoid}(H_1(\mathbf{x})), \tag{5.35}$$

$$\mathbf{y} = \mathbf{g} \odot \text{ReLU}(H_2(\mathbf{x})) + (1 - \mathbf{g}) \odot \mathbf{x}. \tag{5.36}$$

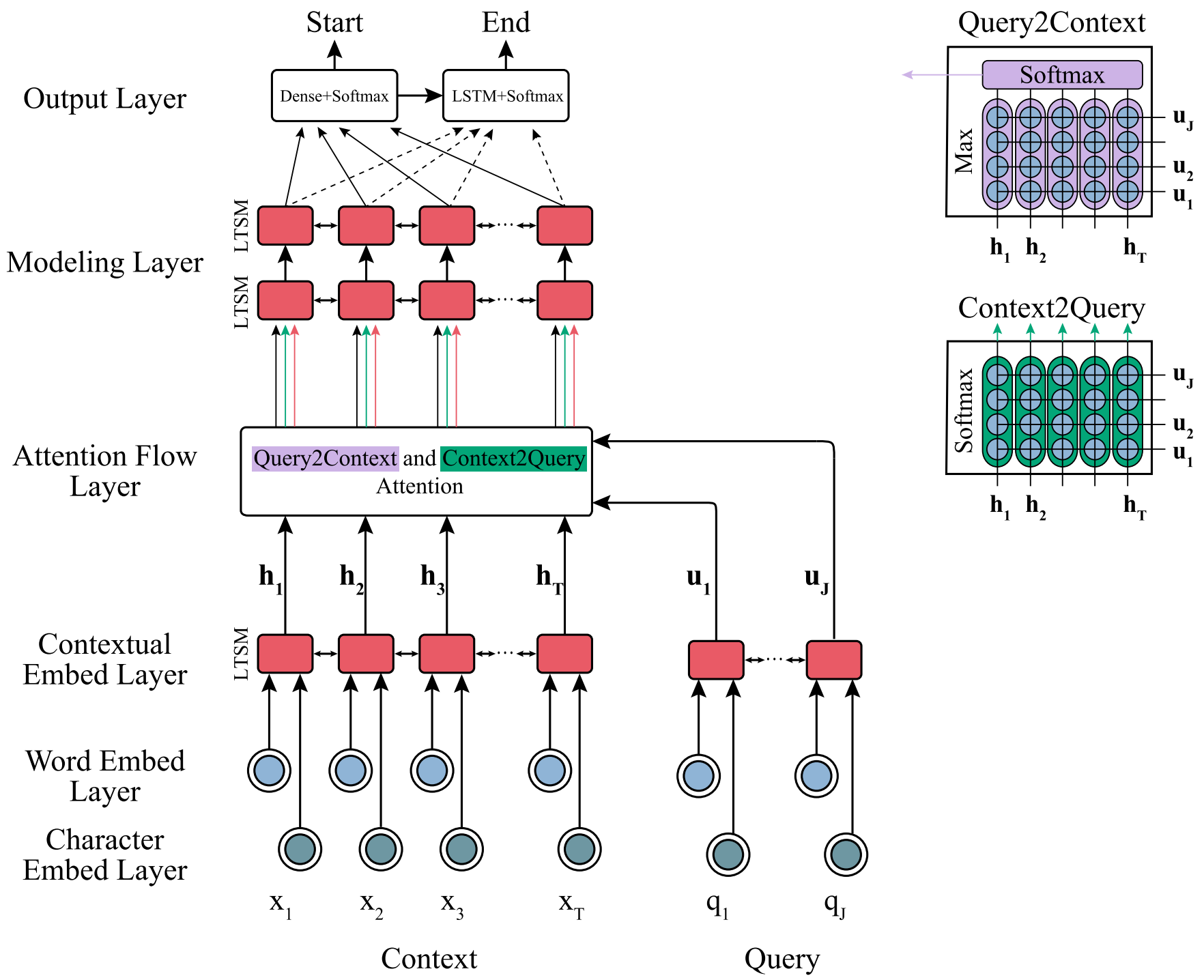

Fig. 5.13 The architecture of BiDAF model

After feeding the context and the query to the token embedding layer, we obtain $\mathbf{X} \in \mathbb{R}^{d \times T}$ for the context and $\mathbf{Q} \in \mathbb{R}^{d \times J}$ for the query, respectively. Afterward, the contextual embedding layer, which is a bidirectional LSTM, model the temporal interaction between words for both the context and the query.

Then, come to the attention flow layer. In this layer, the attention dependency is computed in both directions, i.e., the context-to-query (C2Q) attention and the query-to-context (Q2C) attention. For both kinds of attention, we first compute a similarity matrix $\mathbf{S} \in \mathbb{R}^{T \times J}$ using the contextual embeddings of the context $\mathbf{H}$ and the query $\mathbf{U}$ obtained from the last layer (Eq. 5.37). In the equation, $\alpha(\cdot)$ computes the scalar similarity of the given two vectors and $\mathbf{m}$ is a trainable weight vector.

$$\mathbf{S}_{tj} = \alpha(\mathbf{H}_{:,t}, \mathbf{U}_{:,j}) \tag{5.37}$$

$$\alpha(\mathbf{h}, \mathbf{u}) = \mathbf{m}^\top[\mathbf{h}; \mathbf{u}; \mathbf{h} \odot \mathbf{u}], \tag{5.38}$$

where $\odot$ indicates element-wise product.

For the C2Q attention, a weighted sum of contextual query embeddings is computed given each context word. The attention distribution over the query is obtained by $\mathbf{a}_j = \text{Softmax}(\mathbf{S}_{j,:}) \in \mathbb{R}^J$. The final attended query vector is therefore $\tilde{\mathbf{U}}_{:,t} = \sum_j a_{tj} \mathbf{U}_{:,j}$ for each context word.

For the Q2C attention, the context embeddings are merged into a single fixed length hidden vector $\tilde{\mathbf{h}}$. The attention distribution over the context is computed by $\mathbf{b}_t = \text{Softmax}(\max_j \mathbf{S}_{tj})$, and $\tilde{\mathbf{h}} = \sum_t \mathbf{b}_t \mathbf{H}_{:,t}$. Lastly, the merged context embeddings are tiled $T$ times along the column to produce $\tilde{\mathbf{H}}$.

Finally, the attended outputs are combined to yield $\mathbf{G}$, which is defined by Eq. 5.39

$$\mathbf{G}_{:,t} = \phi(\mathbf{H}_{:,t}, \tilde{\mathbf{U}}_{:,t}, \tilde{\mathbf{H}}_{:,t}) \tag{5.39}$$

$$\beta(\mathbf{h}, \tilde{\mathbf{u}}, \tilde{\mathbf{h}}) = [\mathbf{h}; \tilde{\mathbf{u}}; \mathbf{h} \odot \tilde{\mathbf{u}}; \mathbf{h} \odot \tilde{\mathbf{h}}]. \tag{5.40}$$

Afterward, the LSTM modeling layer takes $\mathbf{G}$ as input and encodes it using a two-layer bidirectional LSTM. The output $\mathbf{M} \in \mathbb{R}^{2d \times T}$ is combined with $\mathbf{G}$ to yield the final start and end probability distributions over the passage.

$$P^1 = \text{Softmax}(\mathbf{u}_1^\top[\mathbf{G}; \mathbf{M}]), \tag{5.41}$$

$$P^2 = \text{Softmax}(\mathbf{u}_2^\top[\mathbf{G}; \text{LSTM}(\mathbf{M})]), \tag{5.42}$$

where $\mathbf{u}_1,\ \mathbf{u}_2$ are two trainable weight vectors.

To train the model, the negative log likelihood loss is adopted and the goal is to maximize the probability of the golden start index $idx_{start}$ and end index $idx_{end}$ being selected by the model,

$$\mathscr{L} = -\frac{1}{N}\sum_{i=1}^{N}\left(\log(P^1_{idx^i_{start}}) + \log(P^2_{idx^i_{start}})\right). \tag{5.43}$$

Besides BiDAF, where attention dependencies are computed in two directions, we will also briefly introduce other interaction methods between the query and the passage. The Gated-Attention Reader proposed by [19] adopts the gated attention module, where each token representation of the passage $d_i$ is scaled by the attended query vector $\mathbf{Q}$ after each Bi-GRU layer (Eq. 5.44).

$$\alpha_i = \text{Softmax}(\mathbf{Q}^\top \mathbf{d}_i) \tag{5.44}$$

$$\tilde{q}_i = \mathbf{Q}\alpha_i \tag{5.45}$$

$$\mathbf{x}_i = \mathbf{d}_i \odot \tilde{\mathbf{q}}_i. \tag{5.46}$$

This gated attention mechanism allows the query to directly interact with the token embeddings of the passage at the semantic level. And such layer-wise interaction enables the model to learn conditional token representation given the question at different representation levels.

The Attention-over-Attention Reader [16] takes another path to model the interaction. The attention-over-attention mechanism involves calculating the attention between the passage attention $\alpha(t)$ and the averaged question attention $\beta$ after obtaining the similarity matrix $\mathbf{M} \in \mathbb{R}^{n\times m}$ (Eq. 5.47). This operation is considered to learn the contributions of individual question words explicitly.

$$\alpha(t) = \text{Softmax}(\mathbf{M}_{:,t}),$$
$$\beta = \frac{1}{N}\sum_{t=1}^{N}\text{Softmax}(\mathbf{M}_{t,:}). \tag{5.47}$$

#### 5.5.2.2 Open-Domain Question Answering

Open-domain QA (OpenQA) has been first proposed by [21]. The task aims to answer open-domain questions using external resources such as collections of documents [58], web pages [14, 29], structured knowledge graphs [3, 7] or automatically extracted relational triples [20].

Recently, with the development of machine reading comprehension techniques [11, 16, 19, 55, 63], researchers attempt to answer open-domain questions via performing reading comprehension on plain texts. Reference [12] proposes to employ neural-based models to answer open-domain questions. As illustrated in Fig. 5.14, neural-based OpenQA system usually retrieves relevant texts of the question from a large-scale corpus and then extracts answers from these texts using reading comprehension models.

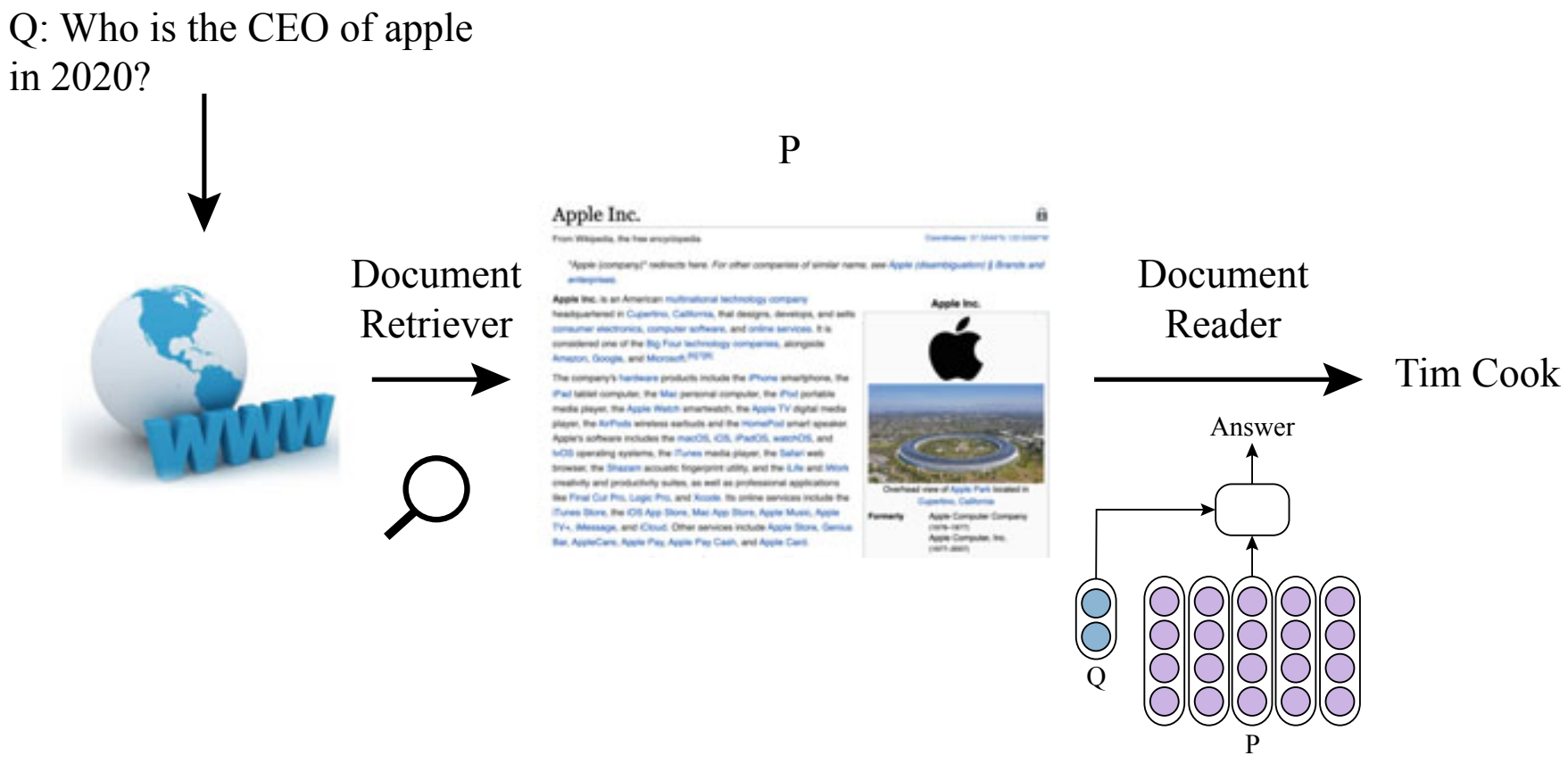

**Fig. 5.14** An example of open-domain question answering

The DrQA system consists of two components: (1) The document retriever module for finding relevant articles and (2) the document reader model for extracting answers from given contexts.

The document retriever is used as a first quick skim to narrow the searching space and focus on documents that are likely to be relevant. The retriever builds TF-IDF weighted bag-of-words vectors for the documents and the questions, and computes similarity scores for ranking. To further utilize local word order information, the retriever uses bigram counts with hash while preserving both the speed and memory efficiency.

The document reader model takes in the top 5 Wikipedia articles yielded by the document retriever and extracts the final answer to the question. For each article, the document reader predicts an answer span with a confidence score. The final prediction is made by maximizing the unnormalized exponential of prediction scores across the documents.

Given each document $d$, the document reader first builds feature representation $\tilde{\mathbf{d}}_i$ for each word in the document. The feature representation $\tilde{\mathbf{d}}$ is made up by the following components.

1. **Word embeddings**: The word embeddings $f_{emb}(d)$ are obtained from large-scale GloVe embeddings pretrained on Wikipedia.
2. **Manual features**: The manual features $f_{token}(d)$ combined part-of-speech (POS) and named entity recognition tags and normalized Term Frequencies (TF).
3. **Exact match**: This feature indicates whether $d_i$ can be exactly matched to one question word in $q$.
4. **Aligned question embeddings**: This feature aims to encode a soft alignment between words in the document and the question in the word embedding space.

$$f_{align}(d_i) = \sum_j \alpha_{ij} \mathbf{E}(q_j) \tag{5.48}$$

$$\alpha_{ij} = \frac{\exp(\text{MLP}(\mathbf{E}(d_i))^\top \text{MLP}(\mathbf{E}(q_j)))}{\sum_{j'} \exp(\text{MLP}(\mathbf{E}(d_i))^\top \text{MLP}(\mathbf{E}(q_{j'})))} \tag{5.49}$$

where $\text{MLP}(\mathbf{x}) = \max(0, \mathbf{Wx} + \mathbf{b})$ and $E(q_j)$ indicates the word embedding of the $j$th word in the question.

Finally, the feature representation is obtained by concatenating the above features:

$$\tilde{\mathbf{d}}_i = (f_{emb}(d_i), f_{token}(d_i), f_{exact_match}(d_i), f_{align}(d_i)). \tag{5.50}$$

Then the feature representation of the document is fed into a multilayer bidirectional LSTM (BiLSTM) to encode the contextual representation.

$$\mathbf{d}_1, \ldots, \mathbf{d}_n = \text{BiLSTM}(\tilde{\mathbf{d}}_1, \ldots, \tilde{\mathbf{d}}_n). \tag{5.51}$$

For the question, the contextual representation is simply obtained by encoding the word embeddings using a multilayer BiLSTM.

$$\mathbf{q}_1, \ldots, \mathbf{q}_m = \text{BiLSTM}(\tilde{\mathbf{q}}_1, \ldots, \tilde{\mathbf{q}}_m) \tag{5.52}$$

After that, the contextual representation is aggregated into a fixed-length vector using self-attention.

$$b_j = \frac{\exp(\mathbf{u}^\top \mathbf{q}_j)}{\sum_{j'} \exp(\mathbf{u}^\top \mathbf{q}_{j'})} \tag{5.53}$$

$$\mathbf{q} = \sum_j b_j \mathbf{q}_j. \tag{5.54}$$

In the answer prediction phase, the start and end probability distributions are calculated following the paradigm mentioned in the Reading Comprehension Model section (Sect. 5.5.2.1).

$$P^{start}(i) = \frac{\exp(\mathbf{d}_i^\top \mathbf{W}^{start} \mathbf{q})}{\sum_{i'} \exp(\mathbf{d}_{i'}^\top \mathbf{W}^{start} \mathbf{q})} \tag{5.55}$$

$$P^{end}(i) = \frac{\exp(\mathbf{d}_i^\top \mathbf{W}^{end} \mathbf{q})}{\sum_{i'} \exp(\mathbf{d}_{i'}^\top \mathbf{W}^{end} \mathbf{q})}. \tag{5.56}$$

Despite its success, the DrQA system is prone to noise in retrieved texts which may hurt the performance of the system. Hence, [15] and [61] attempt to solve the noise problem in DrQA via separating the question answering into paragraph selection and answer extraction, and they both only select the most relevant paragraph among all retrieved paragraphs to extract answers. They lose a large amount of rich information contained in those neglected paragraphs. Hence, [62] proposes strength-based and coverage-based re-ranking approaches, which can aggregate the results extracted from each paragraph by the existing DS-QA system to determine the answer better. However, the method relies on the pre-extracted answers of existing DS-QA models and still suffers the noise issue in distant supervision data because it considers all retrieved paragraphs indiscriminately. To address this issue, [35] proposes a coarse-to-fine denoising OpenQA model, which employs a paragraph selector to filter out paragraphs and a paragraph reader to extract the correct answer from those denoised paragraphs.

## 5.6 Summary

In this chapter, we have introduced document representation learning, which encodes the semantic information of the whole document into a real-valued representation vector, providing an effective way of downstream tasks utilizing the document information and has significantly improved the performances of these tasks.

First, we introduce the one-hot representation for documents. Next, we extensively introduce topic models to represent both words and documents using latent topic distribution. Further, we give an introduction on distributed document representation including paragraph vector and neural document representations. Finally, we introduce several typical real-world applications of document representations, including information retrieval and question answering.

In the future, for better document representation, some directions are requiring further efforts:

(1) **Incorporating External Knowledge.** Current document representation approaches focus on representing documents with the semantic information of the whole document text. Moreover, knowledge bases provide external semantic information to better understand the real-world entities in the given document. Researchers have formed a consensus that incorporating entity semantics of knowledge bases into document representation is a potential way toward better document representation. Some existing work leverages various entity semantics to enhance the semantic information of document representation and achieves better performance in multiple applications such as document ranking [36, 65]. Explicitly modeling structural and textual semantic information as well as considering the entity importance for the given document also share some lights for a more interpretable and knowledgable document representation for downstream NLP tasks.

(2) **Considering Document Interactions.** The candidate documents in downstream NLP tasks are usually relevant to each other and may help for better modeling document semantic information. There is no doubt that the interactions among documents, no matter with implicit semantic relations or with explicit links, will provide additional semantic signals to enhance the document representations. Reference [32] preliminarily uses document interactions to extract important words and improve model performance. Nevertheless, it remains an unsolved problem of how to effectively and explicitly incorporate semantic information into document representations from other documents.

(3) **Pretraining for Document Representation.** Pretraining has shown effectiveness and thrives on downstream NLP tasks. Existing pre-trained language models such as Word2vec style word co-occurrence models [38] and BERT style mask language models [18, 48] focus on the representation learning at the sentence level, which cannot work well for document-level representation. It is still challenging to model cross-sentence relations, text coherence, and co-reference at the document level in document representation learning. Moreover, there are also some methods that leverage useful signals such as anchor-document information to supervise document representation learning [67]. How to pretrain document representation models with efficient and effective strategies is still a critical and challenging problem.

# Chapter 6
# Sememe Knowledge Representation

**Abstract** Linguistic Knowledge Graphs (e.g., WordNet and HowNet) describe linguistic knowledge in formal and structural language, which can be easily incorporated in modern natural language processing systems. In this chapter, we focus on the research about HowNet. We first briefly introduce the background and basic concepts of HowNet and sememe. Next, we introduce the motivations of sememe representation learning and existing approaches. At the end of this chapter, we review important applications of sememe representation.

## 6.1 Introduction

In the field of Natural Language Processing (NLP), words are generally the smallest objects of study because they are considered as the smallest meaningful units that can stand by themselves of human languages. However, the meanings of words can be further divided into smaller parts. For example, the meaning of `man` can be considered as the combination of the meanings of `human`, `male` and `adult`, and the meaning of `boy` is composed of the meanings of `human`, `male`, and `child`. In linguistics, the minimum indivisible units of meaning, i.e., semantic units, are defined as sememes [8]. And some linguists believe that meanings of all the words can be composed of a limited closed set of sememes.

However, sememes are implicit and as a result, it is hard to intuitively define the set of sememes and determine which sememes a word can have at a glance. Therefore, some researchers spend tens of years sifting sememes from all kinds of dictionaries and linguistic Knowledge Bases (KBs), and annotating words with these selected sememes to construct sememe-based linguistic KB. WordNet and HowNet [17] are the two most famous ones of such KBs. In this section, we focus on the representation of linguistic knowledge in HowNet.

## *6.1.1 Linguistic Knowledge Graphs*

### 6.1.1.1 WordNet

WordNet is a large lexical database for the English language and could also be viewed as a KG containing multi-relational data. It was first started in 1985, and created under the direction of George Armitage Miller, a psychology professor in the Cognitive Science Laboratory of Princeton University. Nowadays, WordNet is becoming the most popular lexicon dictionary in the world that could be available through the Web for free and is widely used in NLP applications such as text analysis, information retrieval, and relation extraction. There is also a Global WordNet Association aiming to provide a public and noncommercial platform for WordNets of all languages in the world.

Based on meanings, WordNet groups English nouns, verbs, adjectives, and adverbs into synsets (i.e., sets of cognitive synonyms), which represent a distinct concept. Each synset possesses a brief description, and in most cases, there are even some short sentences functioning as examples illustrating the use of words in this synset. The conceptual-semantic and lexical relations link the synsets and words. The main relation among words is synonymy, which indicates that the words share similar meanings and could be replaced by others in some contexts, while the main relation among synsets is hyperonymy/hyponymy (i.e., the ISA relation), which indicates the relationship between a more general synset and a more specific synset. There are also hierarchical structures for verb synsets, and the antonymy is describing the relation between adjectives with opposite meanings. To sum up, all WordNets' 117, 000 synsets are linked to each other by a small number of conceptual relations.

### 6.1.1.2 HowNet

HowNet was initially designed and constructed by Zhendong Dong and his son Qiang Dong in the 1990s. And it has been kept frequently updated since it was published in 1999.

The sememe set of HowNet is determined by extracting, analyzing, merging, and filtering semantics of thousands of Chinese characters. And the sememe set can also be adjusted or expanded in the subsequent process of annotating words. Each sememe in HowNet is represented by a word or phrase in Chinese and English such as (human | { 人 }) and (ProperName | { 专 }).

HowNet also builds a taxonomy for the sememes. All the sememes of HowNet can be classified as one of the following types: Thing, Part, Attribute, Time, Space, Attribute Value, and Event. In addition, to depict the semantics of words more precisely, HowNet incorporates relations between sememes, which are called "dynamic roles", into the sememe annotations of words.

Considering the polysemy, HowNet differentiates diverse senses of each word in the sememe annotations. And each sense is also expressed in both Chinese and

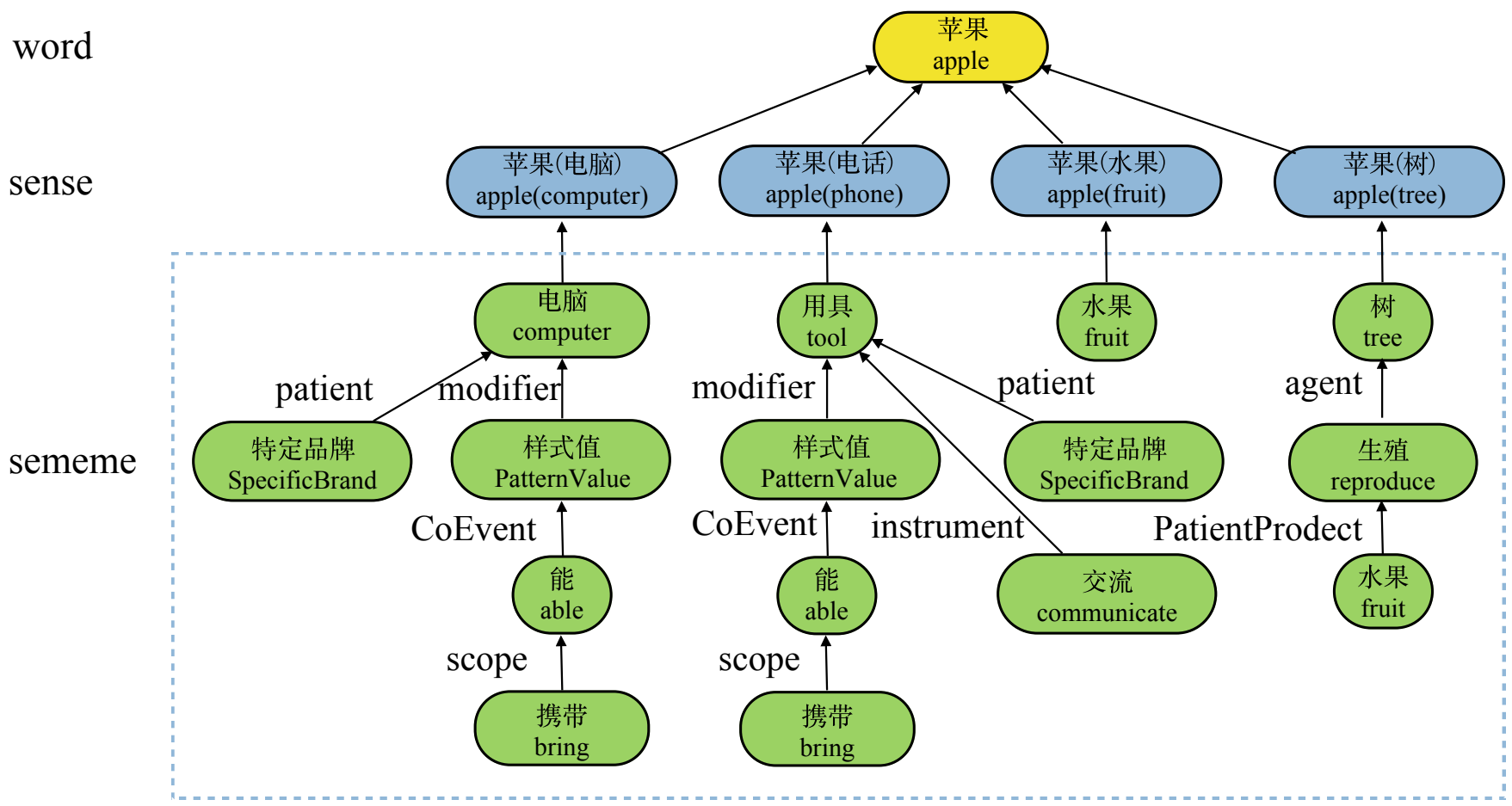

**Fig. 6.1** An example of word annotated with sememes in HowNet

**Table 6.1** Statistics of HowNet

| Type | Count |
|---|---|
| Sense | 229,767 |
| Distinct Chinese word | 127,266 |
| Distinct English word | 104,025 |
| Sememe | 2,187 |

English. An example of sememe annotation for a word is illustrated in Fig. 6.1. We can see from the figure that the word `apple` has four senses including `apple(computer)`, `apple(phone)`, `apple(fruit)`, and `apple(tree)`, and each sense is the root node of a "sememe tree" where each pair of father and son sememe nodes is multi-relational. Additionally, HowNet annotates the POS tag for each sense, and adds sentiment category as well as some usage examples for certain senses.

The latest version of HowNet was published in January 2019 and the statistics are shown in Table 6.1.

Since HowNet was published, it has attracted wide attention. People use HowNet and sememe in various NLP tasks including word similarity computation [40], word sense disambiguation [70], question classification [62], and sentiment analysis [16, 20]. Among these researches, [40] is one of the most influential works, in which the similarity of given two words is computed by measuring the degree of resemblance of their sememe trees.

Recent years also witnessed some works incorporating sememes into neural network models. Reference [49] proposes a novel word representation learning model named SST that reforms Skip-gram [43] by adding contextual attention to senses of the target word, which are represented with combinations of corresponding sememes' embeddings. Experimental results show that SST can not only improve the quality of word embeddings but also learn satisfactory sense embeddings to do word sense disambiguation.

Reference [23] incorporates sememes into the decoding phase of language modeling where sememes are predicted first, and then senses and words are predicted in succession. The proposed model shows enhancement in the perplexity of language modeling and the performance of the downstream task headline generation.

Besides, HowNet is also utilized in lexicon expansion [68], semantic rationality evaluation [41], etc.

Considering that human annotation is time-consuming and labor-intensive, some works attempt to employ machine learning methods to predict sememes for new words automatically. Reference [66] proposes the task firstly and presents two simple but effective models: SPWE, which is based on collaborative filtering, and SPSE, which is based on matrix factorization. Reference [30] further takes the internal information of words into account when predicting sememes and achieves a considerable boost of performance. And [38] takes advantage of definitions of words to predict sememes. As for [56], they propose the task of cross-lingual lexical sememe prediction and present a bilingual word representation learning and alignment-based model, which demonstrates effectiveness in predicting sememes for cross-lingual words.

## 6.2 Sememe Knowledge Representation

Word Representation Learning (WRL) is a fundamental and critical step in many NLP tasks such as language modeling [4] and neural machine translation [64]. There have been a lot of researches for learning word representations, among which Word2vec [43] achieves a nice balance between effectiveness and efficiency. In Word2vec, each word corresponds to one single embedding, ignoring the polysemy of most words. To address this issue, [29] introduces a multi-prototype model for WRL, conducting unsupervised word sense induction and embeddings according to context clusters. Reference [13] further utilizes the synset information in WordNet to instruct word sense representation learning.

These previous studies demonstrate that word sense disambiguation is critical for WRL, and the sememe annotation of word senses in HowNet can provide necessary semantic regularization for these tasks [63]. To explore its feasibility, we introduce the **S**ememe-**E**ncoded **W**ord **R**epresentation **L**earning (SE-WRL) model, which detects word senses and learns representations simultaneously. More specifically, this framework regards each word sense as a combination of its sememes, and iteratively performs word sense disambiguation according to their contexts and learns representations of sememes, senses, and words by extending Skip-gram in Word2vec [43]. In this framework, an attention-based method is proposed to select appropriate word senses according to contexts automatically. To take full advantage of sememes, we introduce three different learning and attention strategies SSA, SAC, and SAT for SE-WRL, which will be described in the following paragraphs.

### 6.2.1 Simple Sememe Aggregation Model

The Simple Sememe Aggregation model (SSA) is a straightforward idea based on Skip-gram model. For each word, SSA considers all sememes in all senses of the word together, and represents the target word using the average of all its sememe embeddings. Formally, we have

$$\mathbf{w} = \frac{1}{m} \sum_{s_i^{(w)} \in S^{(w)}} \sum_{x_j^{(s_i)} \in X_i^{(w)}} \mathbf{x}_j^{(s_i)}, \tag{6.1}$$

which means the word embedding of $w$ is composed by the average of all its sememe embeddings. Here, $S^{(w)}$ is the sense set of $w$ and $X_i^{(w)}$ is the sememe set of the $i$th sense of $w$. $m$ stands for the overall number of sememes belonging to $w$.

This model follows the assumption that the semantic meaning of a word is composed of the semantic units, i.e., sememes. As compared to the conventional Skip-gram model, since sememes are shared by multiple words, this model can utilize sememe information to encode latent semantic correlations between words. In this case, similar words that share the same sememes may finally obtain similar representations.

### 6.2.2 Sememe Attention over Context Model

The SSA Model replaces the target word embedding with the aggregated sememe embeddings to encode sememe information into word representation learning. However, each word in the SSA model still has only one single representation in different contexts, which cannot deal with the polysemy of most words. It is intuitive that we should construct distinct embeddings for a target word according to specific contexts, with the favor of word sense annotation in HowNet.

To address this issue, the Sememe Attention over Context model (SAC) is proposed. SAC utilizes the attention scheme to automatically select appropriate senses for context words according to the target word. That is, SAC conducts word sense disambiguation for context words to learn better representations of target words. The structure of the SAC model is shown in Fig. 6.2.

More specifically, SAC utilizes the original word embedding for target word $w$, and uses sememe embeddings to represent context word $w_c$ instead of the original context word embeddings. Suppose a word typically demonstrates some specific senses in one sentence. Here, the target word embedding is employed as attention to select the most appropriate senses to make up the context word embeddings. The context word embedding $\mathbf{w}_c$ can be formalized as follows:

$$\mathbf{w}_c = \sum_{j=1}^{|S^{(w_c)}|} \text{Att}(s_j^{(w_c)})\mathbf{s}_j^{(w_c)}, \tag{6.2}$$

where $\mathbf{s}_j^{(w_c)}$ stands for the $j$th sense embedding of $w_c$, and $\text{Att}(s_j^{(w_c)})$ represents the attention score of the $j$th sense with respect to the target word $w$, defined as follows:

$$\text{Att}(s_j^{(w_c)}) = \frac{\exp(\mathbf{w} \cdot \hat{\mathbf{s}}_j^{(w_c)})}{\sum_{k=1}^{|S^{(w_c)}|} \exp(\mathbf{w} \cdot \hat{\mathbf{s}}_k^{(w_c)})}. \tag{6.3}$$

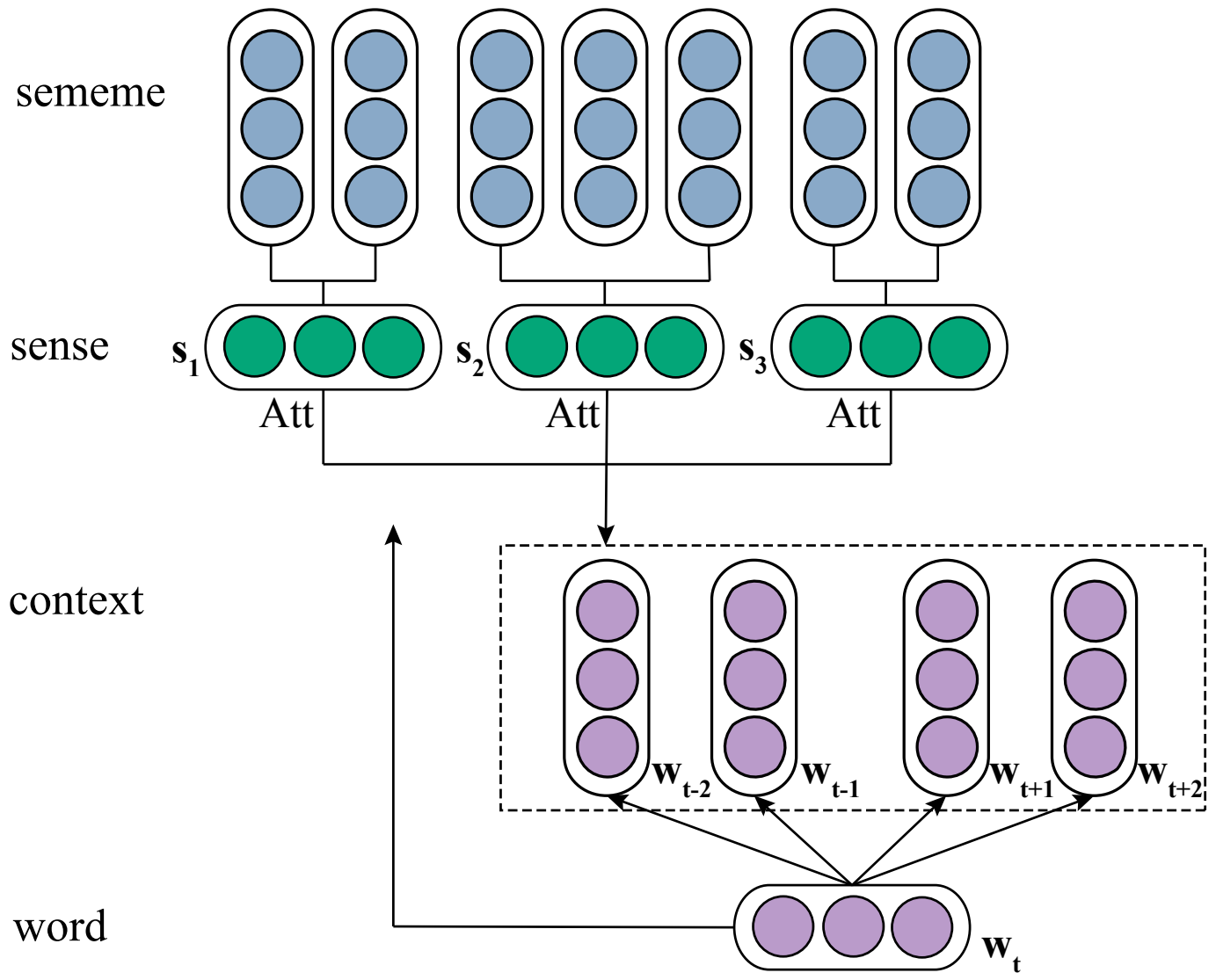

**Fig. 6.2** The architecture of SAC model

Note that, when calculating attention, the average of sememe embeddings is used to represent each sense $s_j^{(w_c)}$:

$$\hat{\mathbf{s}}_j^{(w_c)} = \frac{1}{|X_j^{(w_c)}|} \sum_{k=1}^{|X_j^{(w_c)}|} \mathbf{x}_k^{(s_j)}. \tag{6.4}$$

The attention strategy assumes that the more relevant a context word sense embedding is to the target word $\mathbf{w}$, the more this sense should be considered when building context word embeddings. With the favor of attention scheme, each context word can be represented as a particular distribution over its sense. This can be regarded as soft WSD and it helps learn better word representations.

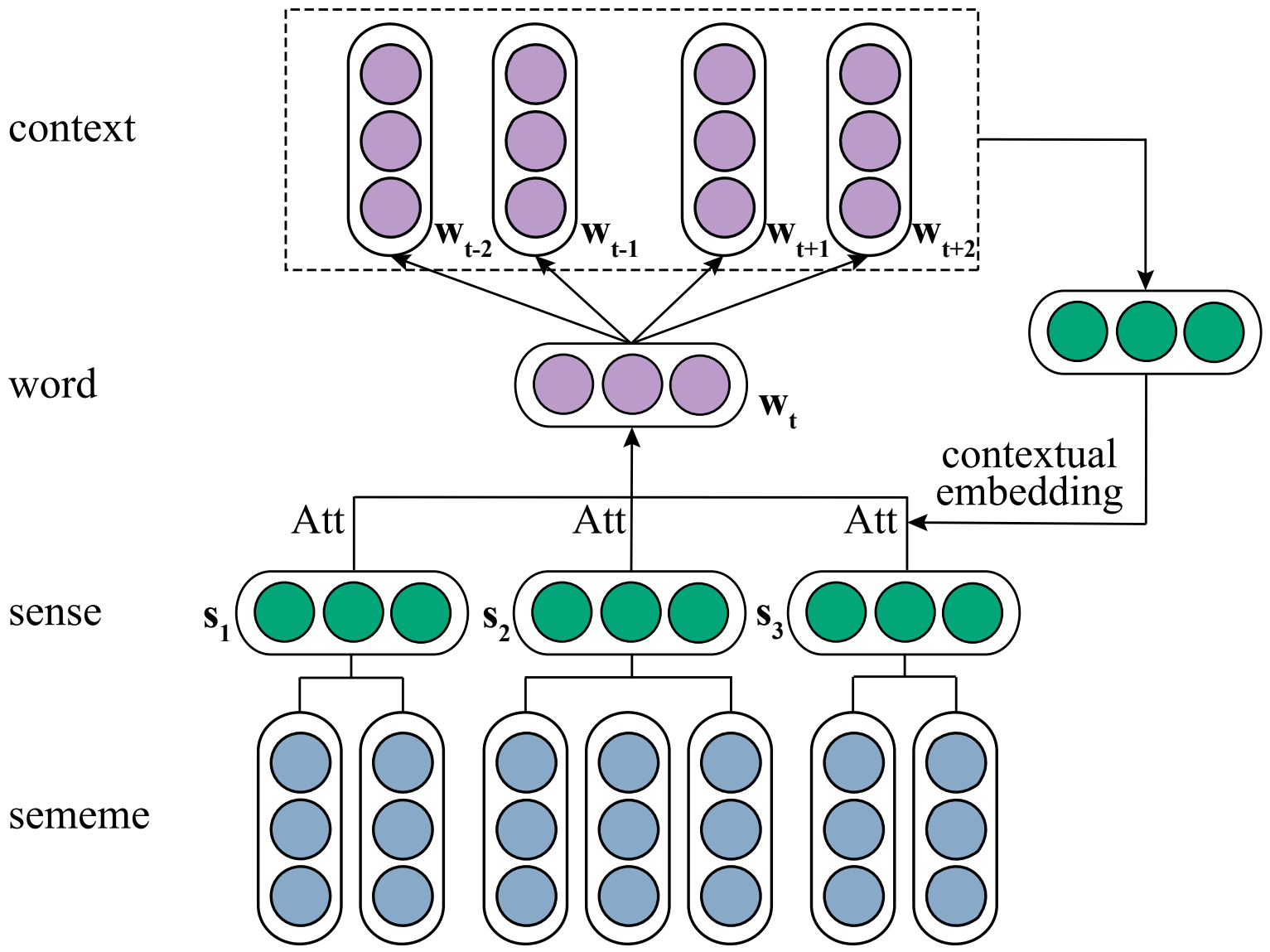

**Fig. 6.3** The architecture of SAT model

### *6.2.3 Sememe Attention over Target Model*

The Sememe Attention over Context Model can flexibly select appropriate senses and sememes for context words according to the target word. The process can also be applied to select appropriate senses for the target word by taking context words as attention. Hence, the Sememe Attention over Target model (SAT) is proposed, which is shown in Fig. 6.3.

Different from the SAC model, SAT learns the original word embeddings for context words and sememe embeddings for target words. Then SAT applies context words to perform attention over multiple senses of the target word $w$ to build the embedding of $w$, formalized as follows:

$$\mathbf{w} = \sum_{j=1}^{|S^{(w)}|} \text{Att}(s_j^{(w)})\mathbf{s}_j^{(w)}, \tag{6.5}$$

and the context-based attention is defined as follows:

$$\mathrm{Att}(s_j^{(w)}) = \frac{\exp(\mathbf{w}_c' \cdot \hat{\mathbf{s}}_j^{(w)})}{\sum_{k=1}^{|S^{(w)}|} \exp(\mathbf{w}_c' \cdot \hat{\mathbf{s}}_k^{(w)})}, \tag{6.6}$$

where the average of sememe embeddings $\hat{\mathbf{s}}_j^{(w)}$ is also used to represent each sense $s_j^{(w)}$. Here, $\mathbf{w}_c'$ is the context embedding, consisting of a constrained window of word embeddings in $C(w_i)$. We have

$$\mathbf{w}_c' = \frac{1}{2K'} \sum_{k=i-K'}^{k=i+K'} \mathbf{w}_k, \quad k \neq i. \tag{6.7}$$

Note that, since in experiments, the sense selection of the target word is found to be only dependent on more limited context words for calculating attention, hence a smaller $K'$ is selected as compared to $K$.

Recall that SAC only uses one target word as attention to select senses of context words whereas SAT uses several context words together as attention to select appropriate senses of target words. Hence SAT is expected to conduct more reliable WSD and result in more accurate word representations, which is explored in experiments.

## 6.3 Applications

In the previous section, we introduce HowNet and sememe representation. In fact, linguistic knowledge graphs such as HowNet contain rich information which could effectively help downstream applications. Therefore, in this section, we will introduce the major applications of sememe representation, including sememe-based word representation, linguistic knowledge graph construction, and language modeling.

### *6.3.1 Sememe-Guided Word Representation*

Sememe-Guided word representation is intended for improving word embeddings for sememe prediction by introducing the information of sememe-based linguistic KBs of the source language. Qi et al. [56] present two methods of the sememe-guided word representation.

#### 6.3.1.1 Relation-Based Word Representation

A simple and intuitive method is to let words with similar sememe annotations tend to have similar word embeddings, which is named as word relation-based approach. To begin with, a synonym list is constructed from sememe-based linguistic KBs, where words sharing a certain number of sememes are regarded as synonyms. Next, synonyms are forced to have closer word embeddings.

Formally, let $\mathbf{w}_i$ be the original word embedding of $w_i$ and $\hat{\mathbf{w}}_i$ be its adjusted word embedding. And let $\mathrm{Syn}(w_i)$ denote the synonym set of word $w_i$. Then the loss function is defined as

$$\mathscr{L}_{sememe} = \sum_{w_i \in V} \Big[\alpha_i \|\mathbf{w}_i - \hat{\mathbf{w}}_i\|^2 + \sum_{w_j \in \mathrm{Syn}(w_i)} \beta_{ij} \|\hat{\mathbf{w}}_i - \hat{\mathbf{w}}_j\|^2\Big], \tag{6.8}$$

where $\alpha$ and $\beta$ control the relative strengths of the two terms. It should be noted that the idea of forcing similar words to have close word embeddings is similar to the state-of-the-art retrofitting approach [19]. However, the retrofitting approach cannot be applied here because sememe-based linguistic KBs such as HowNet cannot directly provide its needed synonym list.

#### 6.3.1.2 Sememe Embedding-Based Word Representation

Simple and effective as the word relation-based approach is, it cannot make full use of the information of sememe-based linguistic KBs because it disregards the complicated relations between sememes and words as well as relations between different sememes. To address this limitation, the sememe embedding-based approach is proposed, which learns both sememe and word embeddings jointly.

In this approach, sememes are represented with distributed vectors as well and place them into the same semantic space as words. Similar to SPSE [66], which learns sememe embeddings by decomposing the word-sememe matrix and sememe-sememe matrix, the method utilizes sememe embeddings as regularizers to learn better word embeddings. Different from SPSE, the model described in [56] does not use pretrained word embeddings. Instead, it learns word embeddings and sememe embeddings simultaneously. More specifically, a word-sememe matrix $\mathbf{M}$ can be extracted from HowNet, where $\mathbf{M}_{ij} = 1$ indicates word $w_i$ is annotated with sememe $x_j$, otherwise $\mathbf{M}_{ij} = 0$. Hence by factorizing $\mathbf{M}$, the loss function can be defined as

$$\mathscr{L}_{sememe} = \sum_{w_i \in V, x_j \in X} (\mathbf{w}_i \cdot \mathbf{x}_j + b_s + b'_j - \mathbf{M}_{ij})^2, \tag{6.9}$$

where $b_i$ and $b'_j$ are the biases of $w_i$ and $x_j$, and $X$ denotes sememe set.

In this approach, word and sememe embeddings are obtained in a unified semantic space. The sememe embeddings bear all the information about the relationships between words and sememes, and they inject the information into word embeddings. Therefore, the word embeddings are expected to be more suitable for sememe prediction.

### 6.3.2 Sememe-Guided Semantic Compositionality Modeling

Semantic Compositionality (SC) is defined as the linguistic phenomenon that the meaning of a syntactically complex unit is a function of meanings of the complex unit's constituents and their combination rule [50]. Some linguists regard SC as the fundamental truth of semantics [51]. In the field of NLP, SC has proved effective in many tasks including language modeling [47], sentiment analysis [42, 61], syntactic parsing [59], etc.

Most literature on SC pays attention to using vector-based distributional models of semantics to learn representations of Multiword Expressions (MWEs), i.e., embeddings of phrases or compounds. Reference [46] conducts a pioneering work which introduces a general framework to formulate this task:

$$\mathbf{p} = f(\mathbf{w}_1, \mathbf{w}_2, \mathscr{R}, \mathscr{K}), \tag{6.10}$$

where[1] $f$ is the compositionality function, $\mathbf{p}$ denotes the embedding of an MWE, $\mathbf{w}_1$ and $\mathbf{w}_2$ represent the embeddings of the MWE's two constituents, $\mathscr{R}$ stands for the combination rule, and $\mathscr{K}$ refers to the additional knowledge which is needed to construct the semantics of the MWE.

Most of the proposed approaches ignore $\mathscr{R}$ and $\mathscr{K}$, centering on reforming compositionality function $f$ [3, 21, 60, 61]. Some try to integrate combination rule $R$ into SC models [7, 35, 65, 71]. A few works consider external knowledge $K$. Reference [72] tries to incorporate task-specific knowledge into an LSTM model for sentence-level SC.

Reference [55] proposes a novel sememe-based method to model semantic compositionality. They argue that sememes are beneficial to modeling SC. To verify this, they first design a simple SC degree (SCD) measurement experiment and find that the SCDs of MWEs computed by simple sememe-based formulae are highly correlated with human judgment. This result shows that sememes can finely depict meanings of MWEs and their constituents, and capture the semantic relations between the two sides. Moreover, they propose two sememe-incorporated SC models for learning embeddings of MWEs, namely Semantic Compositionality with Aggregated Sememe (SCAS) model and Semantic Compositionality with Mutual Sememe Attention (SCMSA) model. When learning the embedding of an MWE, the SCAS model concatenates the embeddings of the MWE's constituents and their sememes, while the SCMSA model considers the mutual attention between a constituent's sememes and the other constituent. Finally, they integrate the combination rule, i.e., $R$ in Eq. (6.10), into the two models. Their models achieve significant performance over the MWE similarity computation task and sememe prediction task compared with baseline methods.

In this section, we focus on the work conducted by [55]. We will first introduce sememe-based SC Degree (SCD) computation formulae, and then expand their Sememe-incorporated SC models.

[1]This formula only applies to two-word MWEs but can be easily extended to longer MWEs.

#### 6.3.2.1 Sememe-Based SCD Computation Formulae

Although SC widely exists in MWEs, not every MWE is fully semantically compositional. In fact, different MWEs show different degrees of SC. Reference [55] believes that sememes can be used to measure SCD conveniently.

To this end, based on the assumption that all the sememes of a word accurately depict the word's meaning, they intuitively design a set of SCD computation formulae, which are consistent with the principle of SCD.

The formulae are illustrated in Table 6.2. They define four SCDs denoted by numbers 3, 2, 1, and 0, where larger numbers mean higher SCDs. $S_{\boldsymbol{p}}$, $S_{\boldsymbol{w}_1}$, and $S_{\boldsymbol{w}_2}$ represent the sememe sets of an MWE, its first and second constituent, respectively. Here is a brief explanation for their SCD computation formulae:

(1) For SCD 3, the sememe set of an MWE is identical to the union of the two constituents' sememe sets, which means the meaning of the MWE is exactly the same as the combination of the constituents' meanings. Therefore, the MWE is fully semantically compositional and should have the highest SCD.

(2) For SCD 0, an MWE has totally different sememes from its constituents, which means the MWE's meaning cannot be derived from its constituents' meanings. Hence the MWE is completely non-compositional, and its SCD should be the lowest.

(3) As for SCD 2, the sememe set of an MWE is a proper subset of the union of its constituents' sememe sets, which means the meanings of the constituents cover the MWE's meaning but cannot precisely infer the MWE's meaning.

(4) Finally, for SCD 1, an MWE shares some sememes with its constituents, but both the MWE itself and its constituents have some unique sememes.

There is an example for each SCD in Table 6.2, including a Chinese MWE, its two constituents, and their sememes.[2]

#### 6.3.2.2 Evaluating SCD Computation Formulae

To evaluate their sememe-based SCD computation formulae, [55] constructs a human-annotated SCD dataset. They ask several native speakers to label SCDs for 500 Chinese MWEs, where there are four degrees to choose. Before labeling an MWE, they are shown the dictionary definitions of both the MWE and its constituents.

Each MWE is labeled by 3 annotators, and the average of the 3 SCDs given by them is the MWE's final SCD.

Eventually, they obtain a dataset containing 500 Chinese MWEs together with their human-annotated SCDs.

Then they evaluate the correlativity between SCDs of the MWEs in the dataset computed by sememe-based rules and those given by humans. They find Pearson's correlation coefficient is up to **0.75**, and Spearman's rank correlation coefficient is

[2]In Chinese, most MWEs are words consisting of more than two characters which are actually single-morpheme words.

**Table 6.2** Sememe-based semantic compositionality degree computation formulae and examples. Bold sememes of constituents are shared with the constituents' corresponding MWE

<table>
<tr><th rowspan="2">SCD</th><th rowspan="2">Our Computation Formulae</th><th colspan="2">Examples</th></tr>
<tr><th>MWEs and Constituents</th><th>Sememes</th></tr>
<tr><td>3</td><td>$S_{\boldsymbol{p}} = S_{\boldsymbol{w}_1} \cup S_{\boldsymbol{w}_2}$</td><td>农民起义(peasant uprising)<br>农民 (peasant)<br>起义(uprising)</td><td>事情 |fact, 职位 |occupation, 政 |politics, 暴动 |uprise, 人 |human, 农 |agricultural<br>**职位 |occupation, 人 |human, 农 |agricultural**<br>**暴动 |uprise, 事情 |fact, 政 |politics**</td></tr>
<tr><td>2</td><td>$S_{\boldsymbol{p}} \subsetneq (S_{\boldsymbol{w}_1} \cup S_{\boldsymbol{w}_2})$</td><td>几何图形(geometric figure)<br>几何 (geometry; how much)<br>图形(figure)</td><td>数学 |math, 图像 |image<br>**数学 |math**, 知识 |knowledge, 疑问 |question, 功能词 |funcword<br>**图像 |image**</td></tr>
<tr><td>1</td><td>$S_{\boldsymbol{p}} \cap (S_{\boldsymbol{w}_1} \cup S_{\boldsymbol{w}_2}) \neq \emptyset$<br>$\wedge S_{\boldsymbol{p}} \not\subset (S_{\boldsymbol{w}_1} \cup S_{\boldsymbol{w}_2})$</td><td>应考(engage a test)<br>应 (deal with; echo; agree)<br>考(quiz; check)</td><td>考试 |exam, 从事 |engage<br>处理 |handle, 回应 |respond, 同意 |agree, 遵循 |obey, 功能词 |funcword, 姓 |surname<br>**考试 |exam**, 查 |check</td></tr>
<tr><td>0</td><td>$S_{\boldsymbol{p}} \cap (S_{\boldsymbol{w}_1} \cup S_{\boldsymbol{w}_2}) = \emptyset$</td><td>画句号(end)<br>画 (draw)<br>句号(period)</td><td>完毕 |finish<br>画 |draw, 部件 |part, 图像 |image, 文字 |character, 表示 |express<br>符号 |symbol, 语文 |text</td></tr>
</table>

0.74. These results manifest remarkable capability of sememes to compute SCDs of MWEs and provide a proof that sememes of a word can finely represent the word's meaning.

#### 6.3.2.3 Sememe-Incorporated SC Models

In this section, we first introduce two basic sememe-incorporated SC models in detail, namely Semantic Compositionality with Aggregated Sememe (SCAS) and Semantic Compositionality with Mutual Sememe Attention (SCMSA). SCAS model simply concatenates the embeddings of the MWE's constituents and their sememes, while the SCMSA model takes account of the mutual attention between a constituent's sememes and the other constituent. Then we describe how to integrate combination rules into the two basic models.

**Incorporating Sememes Only**. Following the notations in Eq. (6.10), for an MWE $p = \{w_1, w_2\}$, its embedding can be represented as

$$\mathbf{p} = f(\mathbf{w}_1, \mathbf{w}_2, \mathscr{K}), \tag{6.11}$$

where $\mathbf{p}, \mathbf{w}_1, \mathbf{w}_2 \in \mathbb{R}^d$, and $d$ is the dimension of embeddings, $\mathscr{K}$ denotes the sememe knowledge here, and we assume that we only know the sememes of $w_1$ and $w_2$, considering that MWEs are normally not in the sememe KBs. $X$ indicates the set of all the sememes and $X_w = \{x_1, ..., x_{|X_w|}\} \subset X$ to signify the sememe set of $w$. In addition, $\mathbf{x} \in \mathbb{R}^d$ denotes the embedding of sememe $x$.

(1) **SCAS Model** The first model we introduce is the SCAS model, which is illustrated in Fig. 6.4. The idea of the SCAS model is straightforward, i.e., simply concatenating word embedding of a constituent and the aggregation of its sememes' embeddings. Formally, we have

$$\mathbf{w}_1' = \sum_{x_i \in X_{w_1}} \mathbf{x_i}, \quad \mathbf{w}_2' = \sum_{x_j \in X_{w_2}} \mathbf{x_j}, \tag{6.12}$$

where $\mathbf{w}_1'$ and $\mathbf{w}_2'$ represent the aggregated sememe embeddings of $w_1$ and $w_2$, respectively. Then $\mathbf{p}$ can be obtained by

$$\mathbf{p} = \tanh(\mathbf{W}_c[\mathbf{w}_1 + \mathbf{w}_2; \mathbf{w}_1' + \mathbf{w}_2'] + \mathbf{b}_c), \tag{6.13}$$

where $\mathbf{W}_c \in \mathbb{R}^{d \times 2d}$ is the composition matrix and $\mathbf{b}_c \in \mathbb{R}^d$ is a bias vector.

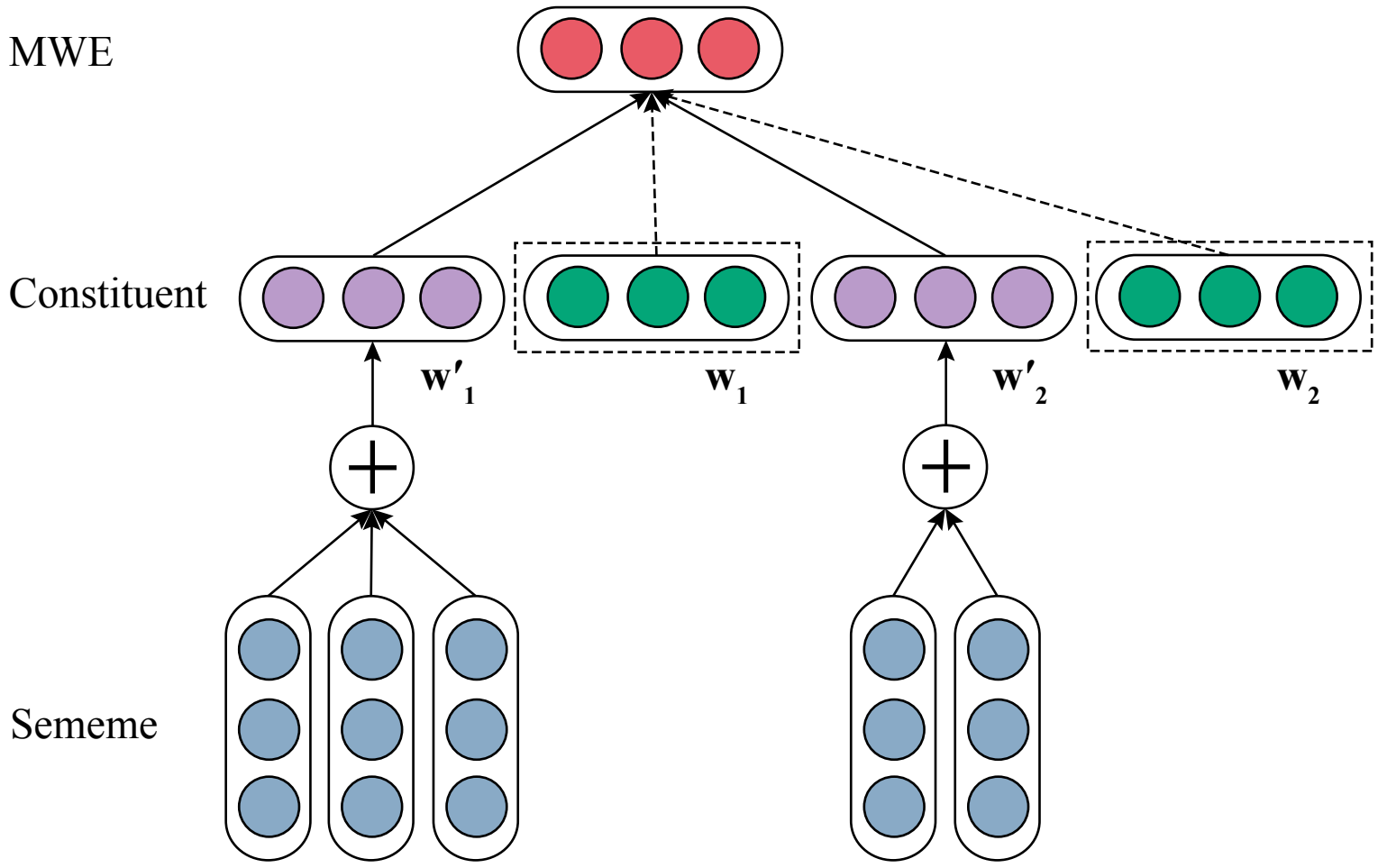

**Fig. 6.4** The architecture of SCAS model

(2) **SCMSA Model**

The SCAS model simply uses the sum of all the sememes' embeddings of a constituent as the external information. However, a constituent's meaning may vary with the other constituent, and accordingly, the sememes of a constituent should have different weights when the constituent is combined with different constituents (there is an example in the case study).

Correspondingly, we introduce the SCMSA model (Fig. 6.5), which adopts the mutual attention mechanism to dynamically endow sememes with weights. Formally, we have

$$\begin{aligned} \mathbf{e}_1 &= \tanh(\mathbf{W}_a \mathbf{w}_1 + \mathbf{b}_a), \\ a_{2,i} &= \frac{\exp(\mathbf{s}_i \cdot \mathbf{e}_1)}{\sum_{x_j \in X_{w_2}} \exp(\mathbf{x}_j \cdot \mathbf{e}_1)}, \\ \mathbf{w}_2' &= \sum_{x_i \in X_{w_2}} a_{2,i} \mathbf{x}_i, \end{aligned} \tag{6.14}$$

where $\mathbf{W}_a \in \mathbb{R}^{d \times d}$ is the weight matrix and $\mathbf{b}_a \in \mathbb{R}^d$ is a bias vector. Similarly, $\mathbf{w}_1'$ can be calculated. Then they still use Eq. (6.13) to obtain $\mathbf{p}$.

**Integrating Combination Rules**. Reference [55] further integrates combination rules into their sememe-incorporated SC models. In other words,

$$\mathbf{p} = f(\mathbf{w}_1, \mathbf{w}_2, K, R). \tag{6.15}$$

We can use totally different composition matrices for MWEs with different combination rules:

$$\mathbf{W}_c = \mathbf{W}_c^r, \quad r \in R_s, \tag{6.16}$$

where $\mathbf{W}_c^r \in \mathbb{R}^{d \times 2d}$ and $R_s$ refers to combination rule set containing syntax rules of MWEs, e.g., adjective-noun and noun-noun.

However, there are many different combination rules, and some rules have sparse instances which are not enough to train the corresponding composition matrices with $d \times 2d$ parameters. In addition, we believe that the composition matrix should contain common compositionality information except the combination rule-specific compositionality information. Hence, they let composition matrix $\mathbf{W}_c$ be the sum of a low-rank matrix containing combination rule information and a matrix containing common compositionality information:

$$\mathbf{W}_c = \mathbf{U}_1^r \mathbf{U}_2^r + \mathbf{W}_c^c, \tag{6.17}$$

where $\mathbf{U}_1^r \in \mathbb{R}^{d \times d_r}$, $\mathbf{U}_2^r \in \mathbb{R}^{d_r \times 2d}$, and $d_r \in \mathbb{N}_+$ is a hyperparameter and may vary with the combination rule, and $\mathbf{W}_c^c \in \mathbb{R}^{d \times 2d}$.

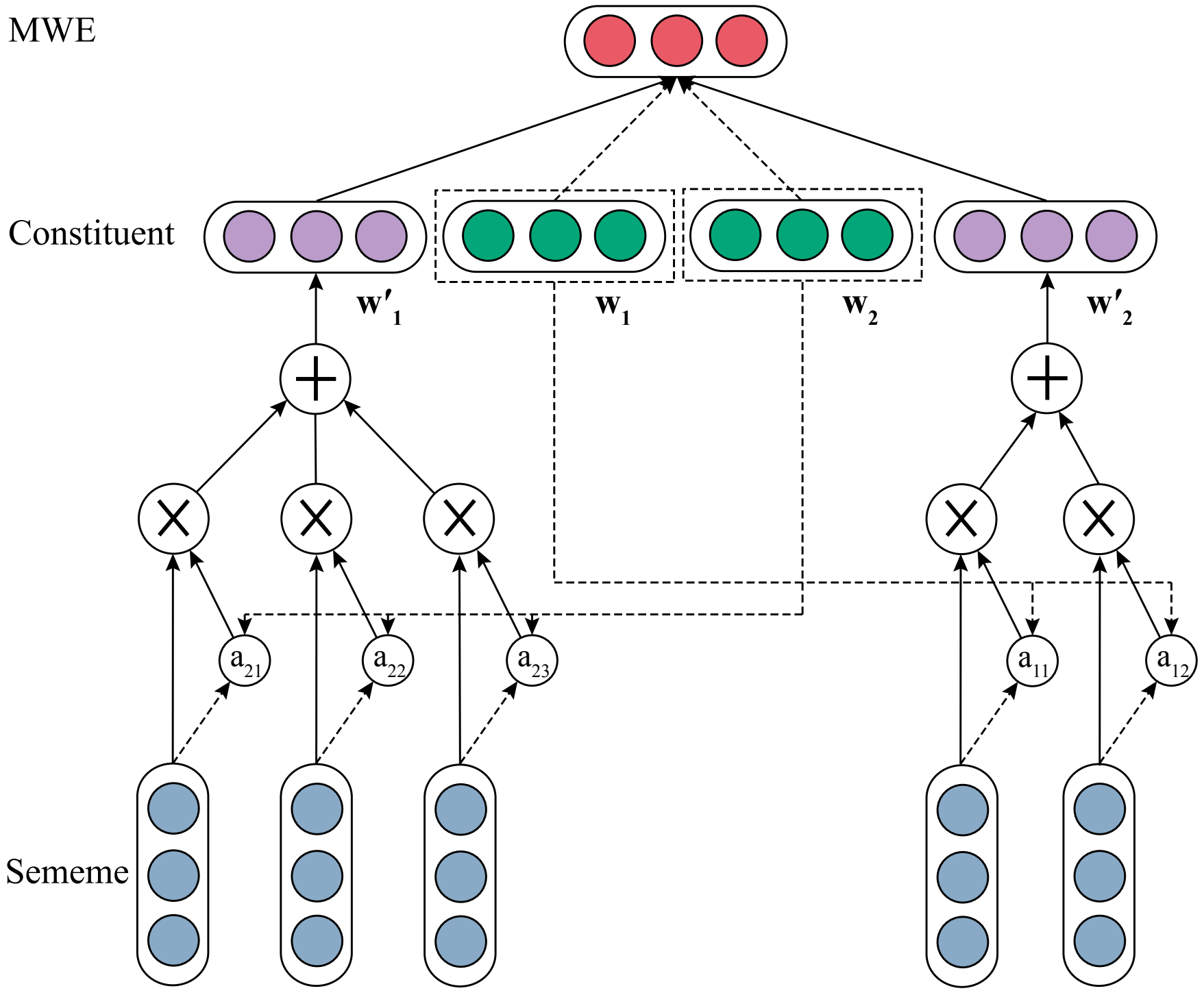

**Fig. 6.5** The architecture of SCMSA model

### *6.3.3 Sememe-Guided Language Modeling*

Language Modeling (LM) aims to measure the probability of a word sequence, reflecting its fluency and likelihood as a feasible sentence in a human language. Language Modeling is an essential component in a wide range of natural language processing (NLP) tasks, such as machine translation [9, 10], speech recognition [34], information retrieval [5, 24, 45, 54], and document summarization [2, 57].

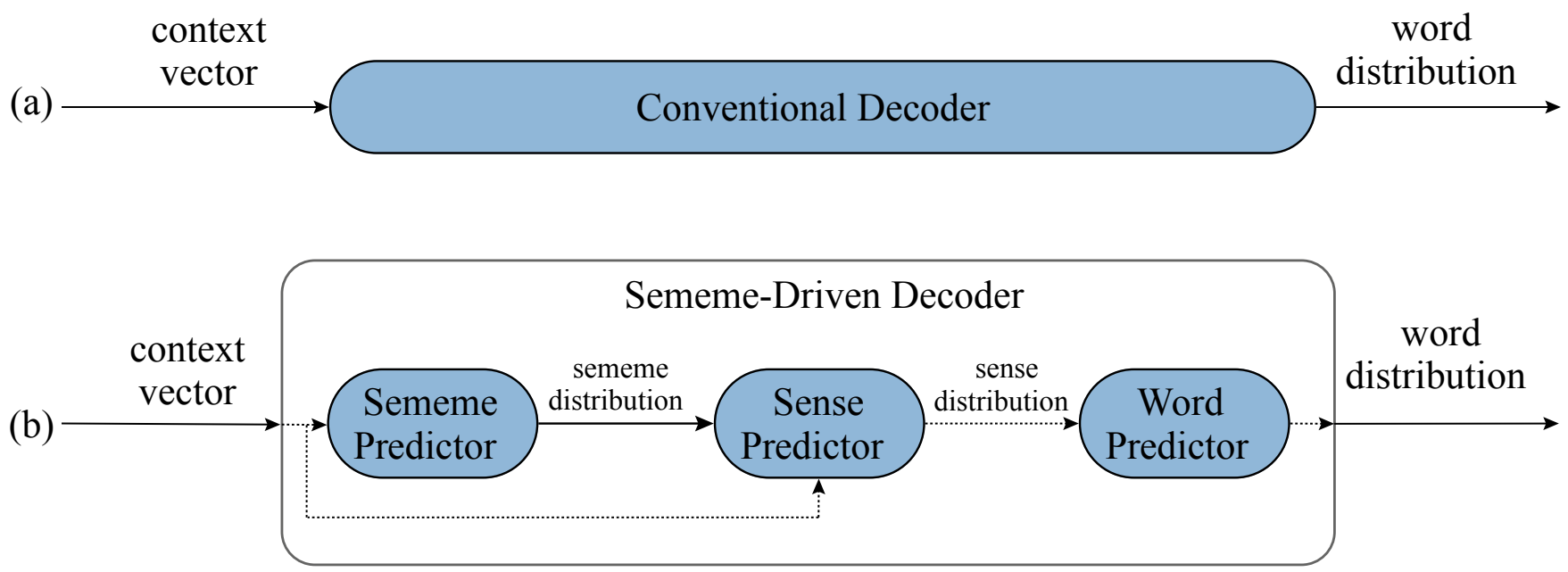

**Fig. 6.6** Decoders of **a** conventional LM, **b** sememe-driven LM

A probabilistic language model calculates the conditional probability of the next word given their contextual words, which are typically learned from large-scale text corpora. Taking the simplest language model, for example, $n$-gram estimates the conditional probabilities according to maximum likelihood over text corpora [31]. Recent years have witnessed the advances of Recurrent Neural Networks (RNNs) as the state-of-the-art approach for language modeling [44], in which the context is represented as a low-dimensional hidden state to predict the next word (Fig. 6.6).

Those conventional language models, including neural models, typically assume words as atomic symbols and model sequential patterns at the word level. However, this assumption does not necessarily hold to some extent. Consider the following example sentence for which people want to predict the next word in the blank,

`The U.S. trade deficit last year is initially estimated to be 40 billion___.`

People may first realize a `unit` should be filled in, then realize it should be a `currency unit`. Based on the country this sentence is talking about, the U.S., one may confirm it should be an `American currency unit` and predict the word `dollars`. Here, the `unit`, `currency`, and `American`, which are basic semantic units of the word `dollars`, are also the sememes of the word `dollars`. However, this process has not been explicitly taken into consideration by conventional language models. That is in most cases, words are atomic language units, words are not necessarily atomic semantic units for language modeling. Thus, explicit modeling of sememes could improve both the performance and the interpretability of language models. However, as far as we know, a few efforts have been devoted to exploring the effectiveness of sememes in language models, especially neural language models.

It is nontrivial for neural language models to incorporate discrete sememe knowledge, as it is not compatible with continuous representations in neural models. In this part, Sememe-Driven Language Model (SDLM) is proposed to leverage lexical sememe knowledge. In order to predict the next word, SDLM utilizes a novel sememe-sense-word generation process: (1) First, SDLM estimates sememes' distribution according to the context. (2) Regarding these sememes as experts, SDLM employs a sparse product of expert method to select the most probable senses. (3) Finally, SDLM calculates the distribution of words by marginalizing out the distribution of senses.

SDLM is composed of three modules in series: Sememe Predictor, Sense Predictor, and Word Predictor (Fig. 6.6). The Sememe Predictor first takes the context vector as input and assigns a weight to each sememe. Then each sememe is regarded as an expert and makes predictions about the probability distribution over a set of senses in the Sense Predictor. Finally, the probability of each word is obtained in the Word Predictor.

**Sememe Predictor**. The Sememe Predictor takes the context vector $\mathbf{g} \in \mathbb{R}^{H_1}$ as input and assigns a weight to each sememe. Assume that given the context $w^1, w^2, \ldots, w^{t-1}$, the events that word $w^t$ contains sememe $x_k$ ($k \in \{1, 2, \ldots, K\}$) are independent, since the sememe is the minimum semantic unit and there is no semantic overlap between any two different sememes. For simplicity, the superscript $t$ is ignored. The Sememe Predictor is designed as a linear decoder with the sigmoid activation function. Therefore, $p_k$, the probability that the next word contains sememe $x_k$, is formulated as

$$p_k = P(x_k|\mathbf{g}) = \text{Sigmoid}(\mathbf{g} \cdot \mathbf{v}_k + b_k), \tag{6.18}$$

where $\mathbf{v}_k \in \mathbb{R}^{H_1}$, $b_k \in \mathbb{R}$ are trainable parameters, and Sigmoid$(\cdot)$ denotes the sigmoid activation function.

**Sense Predictor and Word Predictor**. The architecture of the Sense Predictor is motivated by Product of Experts (PoE) [25]. Each sememe is regarded as an expert that only makes predictions on the senses connected with it. Let $S^{(x_k)}$ denote the set of senses that contain sememe $x_k$, the $k$th expert. Different from conventional neural language models, which directly use the inner product of the context vector $\mathbf{g} \in \mathbb{R}^{H_1}$ and the output embedding $\mathbf{w} \in \mathbb{R}^{H_2}$ for word $w$ to generate the score for each word, Sense Predictor uses $\phi^{(k)}(\mathbf{g}, \mathbf{w})$ to calculate the score given by expert $x_k$. And a bilinear function parameterized with a matrix $\mathbf{U}_k \in \mathbb{R}^{H_1 \times H_2}$ is chosen as a straight implementation of $\phi^{(k)}(\cdot, \cdot)$:

$$\phi^{(k)}(\mathbf{g}, \mathbf{w}) = \mathbf{g}^\top \mathbf{U}_k \mathbf{w}. \tag{6.19}$$

The score of sense $s$ provided by sememe expert $x_k$ can be written as $\phi^{(k)}(\mathbf{g}, \mathbf{s})$. Therefore, $P^{(x_k)}(s|\mathbf{g})$, the probability of sense $s$ given by expert $x_k$, is formulated as

$$P^{(x_k)}(s|\mathbf{g}) = \frac{\exp(q_k C_{k,s} \phi^{(k)}(\mathbf{g}, \mathbf{s}))}{\sum_{s' \in S^{(x_k)}} \exp(q_k C_{k,s'} \phi^{(k)}(\mathbf{g}, \mathbf{s}'))}, \tag{6.20}$$

where $C_{k,s}$ is a normalization constant because sense $s$ is not connected to all experts (the connections are sparse with approximately $\lambda N$ edges, $\lambda < 5$). Here we can choose either $C_{k,s} = 1/|X^{(s)}|$ (*left normalization*) or $C_{k,s} = 1/\sqrt{|X^{(s)}||S^{(x_k)}|}$ (*symmetric normalization*).

In the Sense Predictor, $q_k$ can be viewed as a gate which controls the magnitude of the term $C_{k,s}\phi^{(k)}(\mathbf{g}, \mathbf{s})$, thus controlling the flatness of the sense distribution provided by sememe expert $x_k$. Consider the extreme case when $p_k \to 0$, the prediction will converge to the discrete uniform distribution. Intuitively, it means that the sememe expert will refuse to provide any useful information when it is not likely to be related to the next word.

Finally, the predictions can be summarized on sense $s$ by taking the product of the probabilities given by relevant experts and then normalize the result; that is to say, $P(s|\mathbf{g})$, the probability of sense $s$, satisfies

$$P(s|\mathbf{g}) \propto \prod_{x_k \in X^{(s)}} P^{(x_k)}(s|\mathbf{g}). \tag{6.21}$$

Using Eqs. 6.19 and 6.20, $P(s|\mathbf{g})$ can be formulated as

$$P(s|\mathbf{g}) = \frac{\exp(\sum_{x_k \in X^{(s)}} q_k C_{k,s} \mathbf{g}^\top \mathbf{U}_k \mathbf{s})}{\sum_{s'} \exp(\sum_{x_k \in X^{(s')}} q_k C_{k,s'} \mathbf{g}^\top \mathbf{U}_k \mathbf{s}')}. \tag{6.22}$$

It should be emphasized that all the supervision information provided by HowNet is embodied in the connections between the sememe experts and the senses. If the model wants to assign a high probability to sense $s$, it must assign a high probability to some of its relevant sememes. If the model wants to assign a low probability to sense $s$, it can assign a low probability to its relevant sememes. Moreover, the prediction made by sememe expert $x_k$ has its own tendency because of its own $\phi^{(k)}(\cdot, \cdot)$. Besides, the sparsity of connections between experts and senses is also determined by HowNet itself.

As illustrated in Fig. 6.7, in the Word Predictor, $P(w|\mathbf{g})$, the probability of word $w$ is calculated by summing up probabilities of corresponding $s$ given by the Sense Predictor, that is

$$P(w|\mathbf{g}) = \sum_{s \in S^{(w)}} P(s|\mathbf{g}). \tag{6.23}$$

### *6.3.4 Sememe Prediction*

The manual construction of HowNet is actually time-consuming and labor-intensive, e.g., HowNet has been built for more than 10 years by several linguistic experts. However, as the development of communications and techniques, new words and phrases are emerging, the semantic meanings of existing words are also dynamically evolving. In this case, sustained manual annotation and updates are becoming much more overwhelmed. Moreover, due to the high complexity of sememe ontology and word meanings, it is also challenging to maintain annotation consistency among experts when they collaboratively annotate lexical sememes.

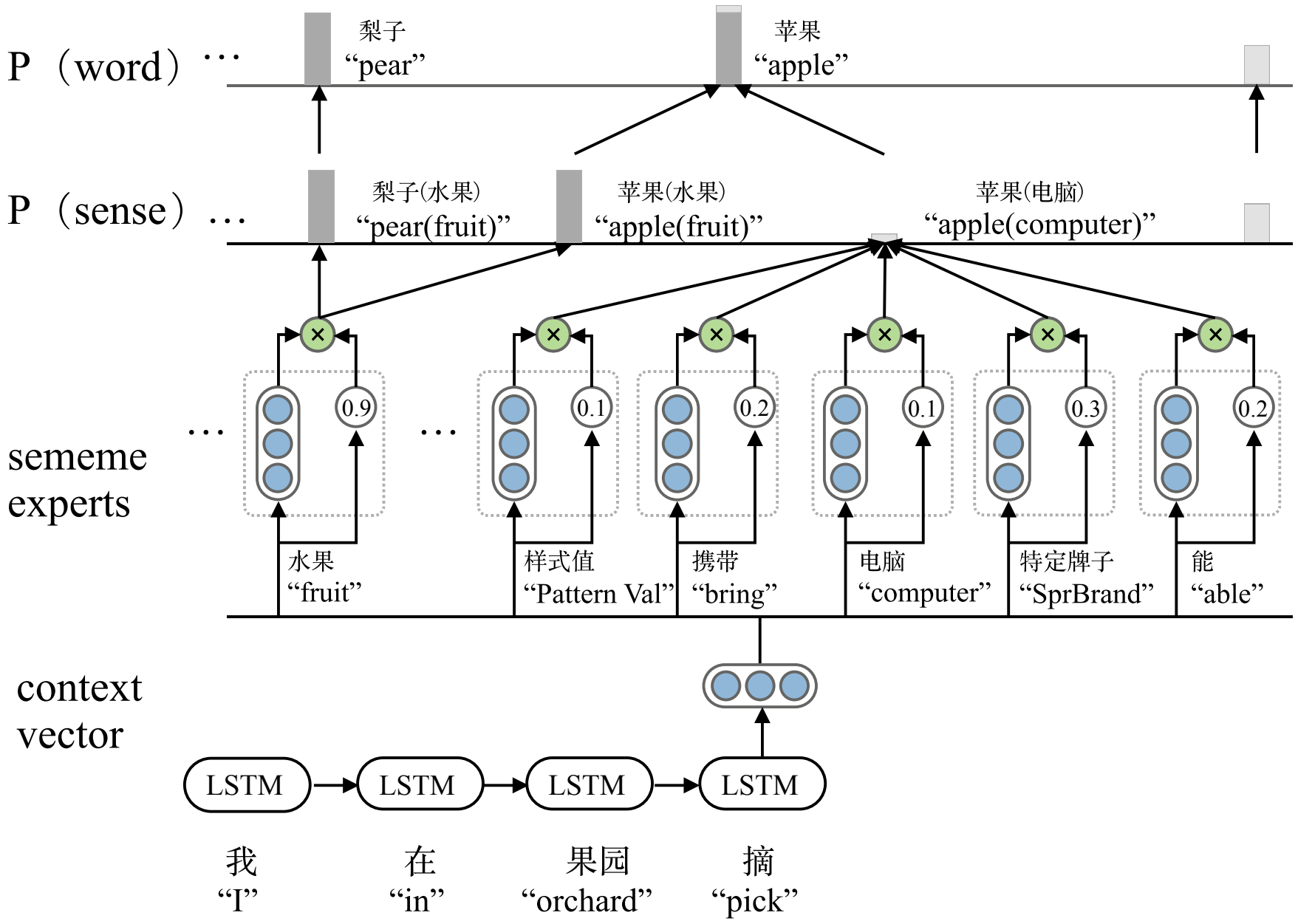

**Fig. 6.7** The architecture of SDLM model

To address the issues of inflexibility and inconsistency of manual annotation, the automatic lexical sememe prediction task is proposed, which is expected to assist expert annotation and reduce manual workload. Note that for simplicity, most works introduced in this part do not consider the complicated hierarchies of word sememes, and simply group all annotated sememes of each word as the sememe set for learning and prediction.

The basic idea of sememe prediction is that those words of similar semantic meanings may share overlapped sememes. Hence, the key challenge of sememe prediction is how to represent semantic meanings of words and sememes to model the semantic relatedness between them. In this part, we will focus on introducing the sememe prediction word accomplished by Xie et al. [66]. In their work, they propose to model the semantics of words and sememes using distributed representation learning [26]. Distributed representation learning aims to encode objects into a low-dimensional semantic space, which has shown its impressive capability of modeling semantics of human languages, e.g., word embeddings [43] have been widely studied and utilized in various tasks of NLP.

As shown in previous work [43], it is effective to measure word similarities using cosine similarity or Euclidean distance of their word embeddings learned from a large-scale text corpus. Hence, a straightforward method for sememe prediction is that, given an unlabeled word, we find its most related words in HowNet according to their word embeddings, and recommend the annotated sememes of these related words to the given word. The method is intrinsically similar to collaborative filtering [58] in recommendation systems, capable of capturing semantic relatedness between words and sememes based on their annotation co-occurrences.

Word embeddings can also be learned with techniques of matrix factorization [37]. Inspired by the successful practice of matrix factorization for personalized recommendation [36], a new model which factorizes the word-sememe matrix from HowNet and obtains sememe embeddings is proposed. In this way, the relatedness of words and sememes can be measured directly using dot products of their embeddings, according to which we could recommend the most related sememes to an unlabeled word.

The two methods are named as Sememe Prediction with Word Embeddings (SPWE) and with Sememe Embeddings (SPSE/SPASE), respectively.

#### 6.3.4.1 Sememe Prediction with Word Embeddings

Given an unlabeled word, it is straightforward to recommend sememes according to its most related words, assuming that similar words should have similar sememes. This idea is similar to collaborative filtering in the personalized recommendation, for in the scenario of sememe prediction words can be regarded as users and sememes as the items/products to be recommended. Inspired by this, Sememe Prediction with Word Embeddings (SPWE) model is proposed, which uses similarities of word embeddings to judge user distances.

Formally, the score function $P(x_j|w)$ of sememes $x_j$ given a word $w$ is defined as

$$P(x_j|w) = \sum_{w_i \in V} \cos(\mathbf{w}, \mathbf{w_i}) \mathbf{M}_{ij} c^{r_i}, \tag{6.24}$$

where $\cos(\mathbf{w}, \mathbf{w_i})$ is the cosine similarity between word embeddings of $w$ and $w_i$ pretrained by GloVe. $\mathbf{M}_{ij}$ indicates the annotation of sememe $x_j$ on word $w_i$, where $\mathbf{M}_{ij} = 1$ indicates the word $w_i$ which has the sememe $x_j$ in HowNet and otherwise has not. Higher the score function $P(x_j|w)$ is, more possible the word $w$ should be recommended with $x_j$.

Differing from classical collaborative filtering in recommendation systems, only the most similar words should be concentrated when predicting sememes for new words since irrelevant words have totally different sememes which may be noises for sememe prediction. To address this problem, a declined confidence factor $c^{r_i}$ is assigned for each word $w_i$, where $r_i$ is the descend rank of word similarity $\cos(\mathbf{w}, \mathbf{w_i})$, and $c \in (0, 1)$ is a hyperparameter. In this way, only a few top words that are similar to $w$ have strong influences on predicting sememes.

SPWE only uses word embeddings for word similarities and is simple and effective for sememe prediction. It is because, differing from the noisy and incomplete user-item matrix in most recommender systems, HowNet is carefully annotated by human experts, and thus the word-sememe matrix is with high confidence. Therefore, the word-sememe matrix can be confidently applied to collaboratively recommend reliable sememes based on similar words.

#### 6.3.4.2 Sememe Prediction with Sememe Embeddings

Sememe Prediction with Word Embeddings model follows the assumption that the sememes of a word can be predicted according to its related words' sememes. However, simply considering sememes as discrete labels may inevitably neglect the latent relations between sememes. To take the latent relations of sememes into consideration, Sememe Prediction with Sememe Embeddings (SPSE) model is proposed, which projects both words and sememes into the same semantic vector space, learning sememe embeddings according to the co-occurrences of words and sememes in HowNet.

Similar to GloVe [53] which decomposes co-occurrence matrix of words to learn word embeddings, sememe embeddings can be learned by factorizing word-sememe matrix and sememe-sememe matrix simultaneously. These two matrices are both constructed from HowNet. As for word embeddings, similar to SPWE, SPSE uses word embeddings pretrained from a large-scale corpus and fixes them during factorizing of the word-sememe matrix. With matrix factorization, both sememe and word embeddings can be encoded into the same low-dimensional semantic space, and then computed the cosine similarity between normalized embeddings of words and sememes for sememe prediction.

More specifically, similar to $\mathbf{M}$, a sememe-sememe matrix $\mathbf{C}$ can also be extracted, where $\mathbf{C}_{jk}$ is defined as point-wise mutual information that $\mathbf{C}_{jk} = \mathrm{PMI}(x_j, x_k)$ to indicate the correlations between two sememes $x_j$ and $x_k$. Note that, by factorizing $\mathbf{C}$, two distinct embeddings for each sememe $s$ will be obtained, denoted as $\mathbf{x}$ and $\bar{\boldsymbol{x}}$, respectively. The loss function of learning sememe embeddings is defined as follows:

$$\mathscr{L} = \sum_{w_i \in W, x_j \in X} \left(\mathbf{w}_i \cdot (\mathbf{x}_j + \bar{\mathbf{x}}_j) + \mathbf{b}_i + \mathbf{b}'_j - \mathbf{M}_{ij}\right)^2 + \lambda \sum_{x_j, x_k \in X} \left(\mathbf{x}_j \cdot \bar{\boldsymbol{x}}_k - \mathbf{C}_{jk}\right)^2, \tag{6.25}$$

where $\mathbf{b}_i$ and $\mathbf{b}'_j$ denote the bias of $w_i$ and $x_j$. These two parts correspond to the losses of factorizing matrices $\mathbf{M}$ and $\mathbf{C}$, adjusted by the hyperparameter $\lambda$. Since the sememe embeddings are shared by both factorizations, our SPSE model enables jointly encoding both words and sememes into a unified semantic space.

Since each word is typically annotated with 2–5 sememes in HowNet, most elements in the word-sememe matrix are zeros. If all zero elements and nonzero elements are treated equally during factorization, the performance will be much worse. To address this issue, different factorization strategies are assigned for zero and nonzero elements. For each zero element, the model chooses to factorize them with a small probability like 0.5%, and otherwise, the model chooses to ignore. While for nonzero elements, the model always chooses to factorize them. With the help of this strategy, the model can pay more attention to those annotated word-sememe pairs.

In SPSE, sememe embeddings are learned accompanying with word embeddings via matrix factorization into the unified low-dimensional semantic space. Matrix factorization has been verified as an effective approach in the personalized recommendation, because it can accurately model relatedness between users and items, and is highly robust to noises in user-item matrices. Using this model, we can flexibly compute semantic relatedness of words and sememes, which provides us an effective tool to manipulate and manage sememes, including but not limited to sememe prediction.

#### 6.3.4.3 Sememe Prediction with Aggregated Sememe Embeddings

Inspired by the characteristics of sememes, we assume that the word embeddings are semantically composed of sememe embeddings. In the word-sememe joint space, we can simply implement semantic composition as additive operations that each word embedding is expected to be the sum of its all sememes' embeddings. Following this assumption, Sememe Prediction with Aggregated Sememe Embeddings (SPASE) model is proposed. SPASE is also based on matrix factorization, and is formally denoted as

$$\mathbf{w}_i = \sum_{x_j \in X_{w_i}} \mathbf{M}'_{ij} \mathbf{x}_j, \tag{6.26}$$

where $X_{w_i}$ is the sememe set of the word $w_i$ and $\mathbf{M}'_{ij}$ represents the weight of sememe $x_j$ for word $w_i$, which only has value on nonzero elements of word-sememe labeled matrix $\mathbf{M}$. To learn sememe embeddings, we attempt to decompose the word embedding matrix $\mathbf{V}$ into $\mathbf{M}'$ and sememe embedding matrix $\mathbf{X}$, with pretrained word embeddings fixed during training, which could also be written as $\mathbf{V} = \mathbf{M}'\mathbf{X}$.

The contribution of SPASE is that it complies with the definition of sememes in HowNet that sememes are the semantic components of words. In SPASE, each sememe can be regarded as a tiny semantic unit, and all words can be represented by composing several semantic units, i.e., sememes, which make up an interesting semantic regularity. However, SPASE is difficult to train because word embeddings are fixed, and the number of words is much larger than the number of sememes. In the case of modeling complex semantic compositions of sememes into words, the representation capability of SPASE may be strongly constrained by limited parameters of sememe embeddings and excessive simplification of additive assumption.

#### 6.3.4.4 Lexical Sememe Prediction with Internal Information

In the previous section, we introduce the automatic lexical sememe prediction proposed by Xie et al. [66]. These methods ignore the internal information within words (e.g., the characters in Chinese words), which is also significant for word understanding, especially for words which are of low frequency or do not appear in the corpus

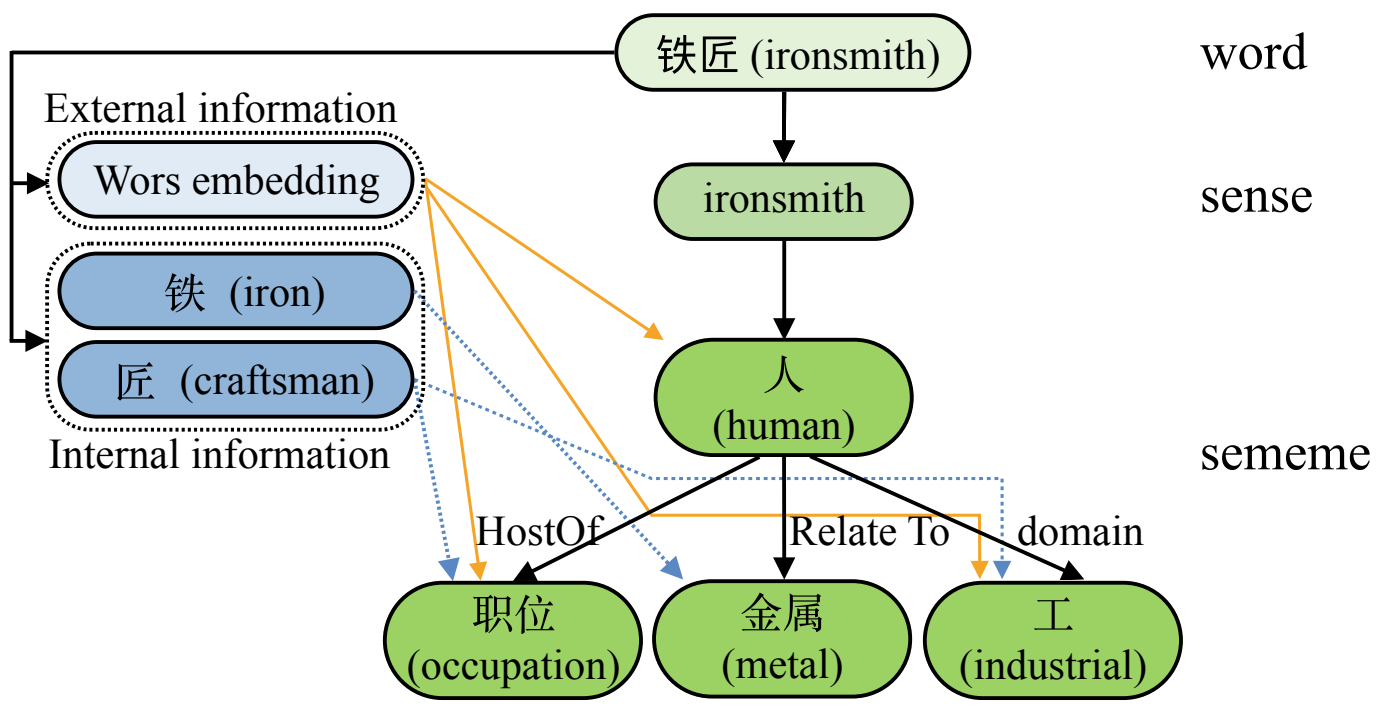

**Fig. 6.8** Sememes of the word 铁匠 (ironsmith) in HowNet, where occupation, human, and industrial can be inferred by both external (contexts) and internal (characters) information, while metal is well-captured only by the internal information within the character 铁 (iron)

at all. In this section, we introduce the work of Jin et al. [30], which takes Chinese as an example and explores methods of taking full advantage of both external and internal information of words for sememe prediction.

In Chinese, words are composed of one or multiple characters, and most characters have corresponding semantic meanings. As shown by [67], more than 90% of Chinese characters in modern Chinese corpora are morphemes. Chinese words can be divided into single-morpheme words and compound words, where compound words account for a dominant proportion. The meanings of compound words are closely related to their internal characters as shown in Fig. 6.8. Taking a compound word 铁匠 (ironsmith), for instance, it consists of two Chinese characters: 铁 (iron) and 匠 (craftsman), and the semantic meaning of 铁匠 can be inferred from the combination of its two characters (iron + craftsman → ironsmith). Even for some single-morpheme words, their semantic meanings may also be deduced from their characters. For example, both characters of the single-morpheme word 徘徊 (hover) represent the meaning of hover or linger. Therefore, it is intuitive to take the internal character information into consideration for sememe prediction.

Reference [30] proposes a novel framework for Character-enhanced Sememe Prediction (CSP), which leverages both internal character information and external context for sememe prediction. CSP predicts the sememe candidates for a target word from its word embedding and the corresponding character embeddings. Specifically, following SPWE and SPSE as introduced by [66] to model external information, Sememe Prediction with Word-to-Character Filtering (SPWCF) and Sememe Prediction with Character and Sememe Embeddings (SPCSE) are proposed to model internal character information.

**Sememe Prediction with Word-to-Character Filtering**. Inspired by collaborative filtering [58], Jin et al. [30] propose to recommend sememes for an unlabeled word according to its similar words based on internal information. And words are considered as *similar* if they contain the same characters at the same positions.

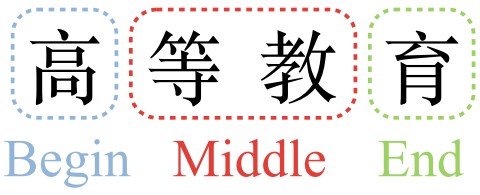

**Fig. 6.9** An example of the position of characters in a word

In Chinese, the meaning of a character may vary according to its position within a word [14]. Three positions within a word are considered: *Begin*, *Middle*, and *End*. For example, as shown in Fig. 6.9, the character at the *Begin* position of the word 火车站 (railway station) is 火 (fire), while 车 (vehicle) and 站 (station) are at the *Middle* and *End* position, respectively. The character 站 usually means station when it is at the *End* position, while it usually means stand at the *Begin* position like in 站立 (stand), 站岗哨兵 (standing guard), and 站起来 (stand up).

Formally, for a word $w = c_1 c_2 ... c_{|w|}$, we define $\pi_B(w) = \{c_1\}$, $\pi_M(w) = \{c_2, ..., c_{|w-1|}\}$, $\pi_E(w) = \{c_{|w|}\}$, and

$$P_p(x_j|c) \sim \frac{\sum_{w_i \in W \wedge c \in \pi_p(w_i)} \mathbf{M}_{ij}}{\sum_{w_i \in W \wedge c \in \pi_p(w_i)} |X_{w_i}|}, \tag{6.27}$$

that represents the score of a sememe $x_j$ given a character $c$ and a position $p$, where $\pi_p$ may be $\pi_B$, $\pi_M$, or $\pi_E$. $\mathbf{M}$ is the same matrix used in SPWE. Finally, the score function $P(x_j|w)$ of sememe $x_j$ given a word $w$ is defined as

$$P(x_j|w) \sim \sum_{p \in \{B,M,E\}} \sum_{c \in \pi_p(w)} P_p(x_j|c). \tag{6.28}$$

SPWCF is a simple and efficient method. It performs well because compositional semantics are pervasive in Chinese compound words, which makes it straightforward and effective to find similar words according to common characters.

**Sememe Prediction with Character and Sememe Embeddings (SPCSE).** The method Sememe Prediction with Word-to-Character Filtering (SPWCF) can effectively recommend the sememes that have strong correlations with characters. However, just like SPWE, it ignores the relations between sememes. Hence, inspired by SPSE, Sememe Prediction with Character and Sememe Embeddings (SPCSE) is proposed to take the relations between sememes into account. In SPCSE, the model instead learns the sememe embeddings based on internal character information, then computes the semantic distance between sememes and words for prediction.

Inspired by GloVe [53] and SPSE, matrix factorization is adopted in SPCSE, by decomposing the word-sememe matrix and the sememe-sememe matrix simultaneously. Instead of using pretrained word embeddings in SPSE, pretrained character embeddings are used in SPCSE. Since the ambiguity of characters is stronger than that of words, multiple embeddings are learned for each character [14]. The most representative character and its embedding are selected to represent the word meaning. Because low-frequency characters are much rare than those low-frequency words,

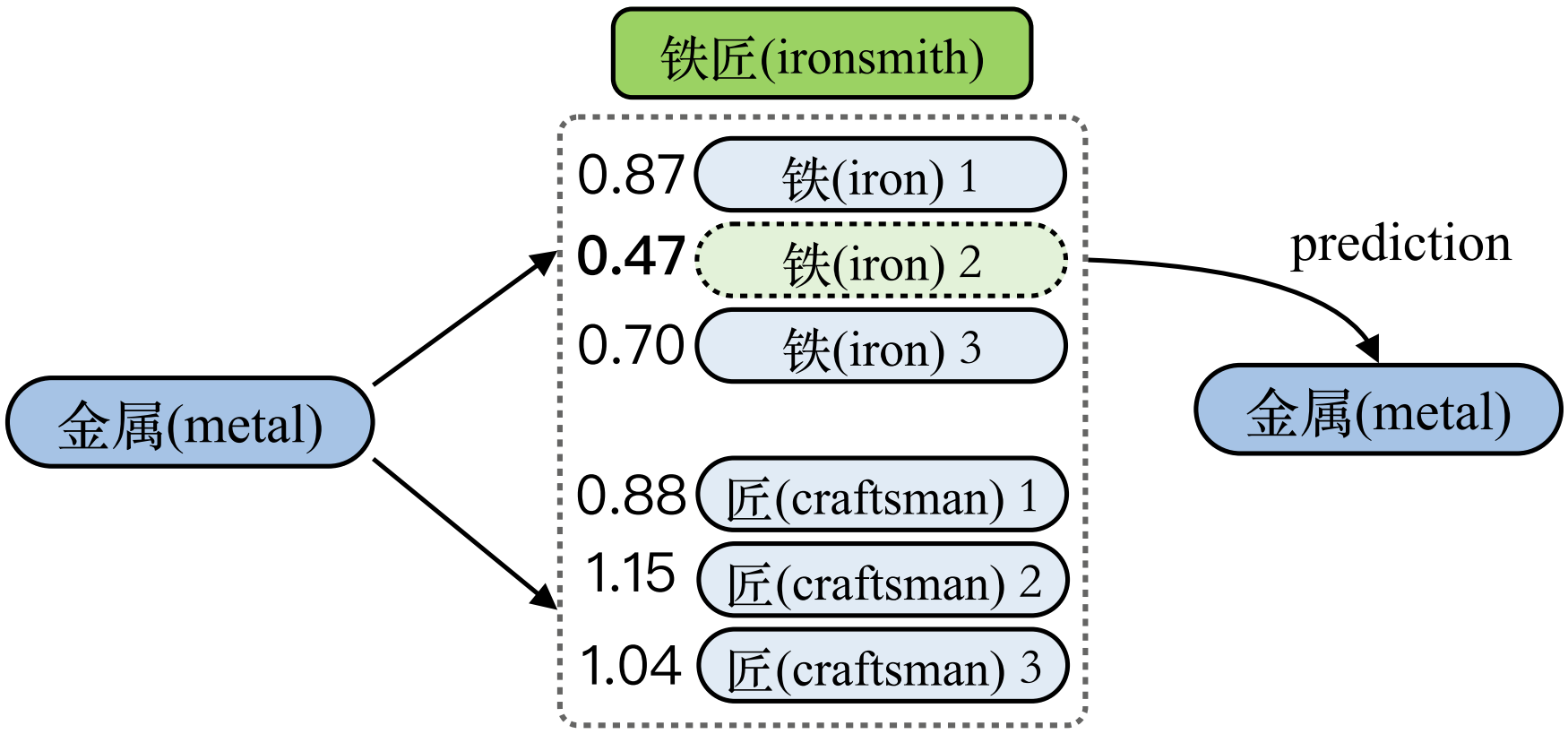

**Fig. 6.10** An example of adopting multiple-prototype character embeddings. The numbers are the cosine distances. The sememe 金属 (metal) is the closest to one embedding of 铁 (iron)

and even low-frequency words are usually composed of common characters, it is feasible to use pretrained character embeddings to represent rare words. During factorizing of the word-sememe matrix, the character embeddings are fixed.

$N_e$ is set as the number of embeddings for each character, and each character $c$ has $N_e$ embeddings $\mathbf{c}^1$, ..., $\mathbf{c}^{N_e}$. Given a word $w$ and a sememe $x$, the embedding of a character of $w$ closest to the sememe embedding by cosine distance is selected as the representation of the word $w$, as shown in Fig. 6.10. Specifically, given a word $w = c_1...c_{|w|}$ and a sememe $x_j$, we define

$$k^*, r^* = \arg\min_{k,r} \left[1 - \cos(\mathbf{c}_k^r, \mathbf{x}_j' + \bar{\boldsymbol{x}}_j')\right], \tag{6.29}$$

where $k^*$ and $r^*$ indicate the indices of the character and its embedding closest to the sememe $x_j$ in the semantic space. With the same word-sememe matrix $\mathbf{M}$ and sememe-sememe correlation matrix $\mathbf{C}$ in SPSE, the sememe embeddings are learned with the loss function:

$$\mathcal{L} = \sum_{w_i \in W, x_j \in X} \left(\mathbf{c}_{k^*}^{r^*} \cdot \left(\mathbf{x}_j' + \bar{\boldsymbol{x}}_j'\right) + \mathbf{b}_{k^*}^c + \mathbf{b}_j'' - \mathbf{M}_{ij}\right)^2 + \lambda' \sum_{x_j, x_q \in X} \left(\mathbf{x}_j' \cdot \bar{\boldsymbol{x}}_q' - \mathbf{C}_{jq}\right)^2, \tag{6.30}$$

where $\mathbf{x}_j'$ and $\bar{\boldsymbol{x}}_j'$ are the sememe embeddings for sememe $x_j$, and $\mathbf{c}_{k^*}^{r^*}$ is the embedding of the character that is the closest to sememe $x_j$ within $w_i$. Note that, as the characters and the words are not embedded into the same semantic space, new sememe embeddings are learned instead of using those learned in SPSE, hence different notations are used for the sake of distinction. $\mathbf{b}_k^c$ and $\mathbf{b}_j''$ denote the biases of $c_k$ and $x_j$, and $\lambda'$ is the hyperparameter adjusting the two parts. Finally, the score function of word $w = c_1...c_{|w|}$ is defined as

$$P(x_j|w) \sim \mathbf{c}_{k^*}^{r^*} \cdot \left(\mathbf{x}_j' + \bar{\boldsymbol{x}}_j'\right). \tag{6.31}$$

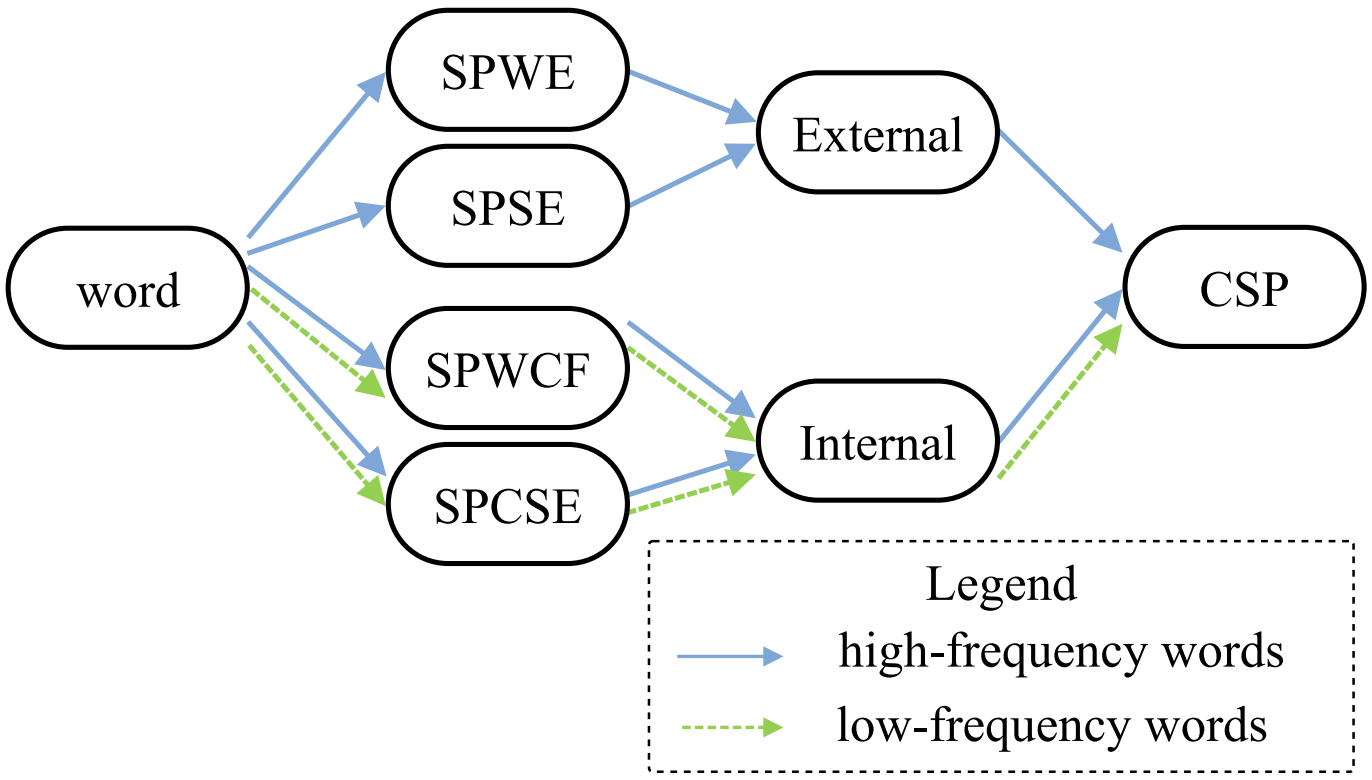

**Fig. 6.11** An illustration of model ensembling in sememe prediction

**Model Ensembling**. SPWCF/SPCSE and SPWE/SPSE take different sources of information as input, which means that they have different characteristics: SPWCF/SPCSE only have access to internal information, while SPWE/SPSE can only make use of external information. On the other hand, just like the difference between SPWE and SPSE, SPWCF originates from collaborative filtering, whereas SPCSE uses matrix factorization. All of those methods have in common that they tend to recommend the sememes of *similar* words, but they diverge in their interpretation of *similar*.

Therefore, to obtain better prediction performance, it is necessary to combine these models. We denote the ensemble of SPWCF and SPCSE as the *internal* model, and the ensemble of SPWE and SPSE as the *external* model. The ensemble of the *internal* and the *external* models is the novel framework CSP. In practice, for words with reliable word embeddings, i.e., high-frequency words, we can use the integration of the *internal* and the *external* models; for words with extremely low frequencies (e.g., having no reliable word embeddings), we can just use the *internal* model and ignore the *external* model, because the external information is noisy in this case. Figure 6.11 shows model ensembling in different scenarios. For the sake of comparison, we use the integration of SPWCF, SPCSE, SPWE, and SPSE as CSP in all experiments. And two models are integrated by simple weighted addition.

#### 6.3.4.5 Cross-Lingual Sememe Prediction

Most languages do not have sememe-based linguistic KBs such as HowNet, which prevents us from understanding and utilizing human languages to a greater extent. Therefore, it is important to build sememe-based linguistic KBs for various languages.

To address the issue of the high labor cost of manual annotation, Qi et al. [56] propose a new task, cross-lingual lexical sememe prediction (CLSP) which aims to automatically predict lexical sememes for words in other languages. There are two critical challenges for CLSP:

(1) There is not a consistent one-to-one match between words in different languages. For example, English word “beautiful” can refer to Chinese words of either 美丽 or 漂亮. Hence, we cannot simply translate HowNet into another language. And how to recognize the semantic meaning of a word in other languages becomes a critical problem.

(2) Since there is a gap between the semantic meanings of words and sememes, we need to build semantic representations for words and sememes to capture the semantic relatedness between them.

To tackle these challenges, Qi et al. [56] propose a novel model for CLSP, which aims to transfer sememe-based linguistic KBs from source language to target language. Their model contains three modules: (1) monolingual word embedding learning which is intended for learning semantic representations of words for source and target languages, respectively; (2) cross-lingual word embedding alignment which aims to bridge the gap between the semantic representations of words in two languages; (3) sememe-based word embedding learning whose objective is to incorporate sememe information into word representations.

They take Chinese as source language and English as the target language to show the effectiveness of their model. Experimental results show that the proposed model could effectively predict lexical sememes for words with different frequencies in other languages and their model has consistent improvements on two auxiliary experiments including bilingual lexicon induction and monolingual word similarity computation by jointly learning the representations of sememes, words in source and target languages.

The model consists of three parts: monolingual word representation learning, cross-lingual word embedding alignment, and sememe-based word representation learning. Hence, they define the objective function of our method corresponding to the three parts:

$$\mathscr{L} = \mathscr{L}_{mono} + \mathscr{L}_{cross} + \mathscr{L}_{sememe}. \tag{6.32}$$

Here, the monolingual term $\mathscr{L}_{mono}$ is designed for learning monolingual word embeddings from nonparallel corpora for source and target languages, respectively. The cross-lingual term $\mathscr{L}_{cross}$ aims to align cross-lingual word embeddings in a unified semantic space. And $\mathscr{L}_{sememe}$ can draw sememe information into word representation learning and conduce to better word embeddings for sememe prediction. In the following paragraphs, we will introduce the three parts in detail.

**Monolingual Word Representation**. Monolingual word representation is responsible for explaining regularities in monolingual corpora of source and target languages. Since the two corpora are nonparallel, $\mathscr{L}_{mono}$ comprises two monolingual submodels that are independent of each other:

$$\mathscr{L}_{mono} = \mathscr{L}^{S}_{mono} + \mathscr{L}^{T}_{mono}, \tag{6.33}$$

where the superscripts $S$ and $T$ denote source and target languages, respectively.

As a common practice, the well-established Skip-gram model is chosen to obtain monolingual word embeddings. The Skip-gram model is aimed at maximizing the predictive probability of context words conditioned on the centered word. Formally, taking the source side, for example, given a training word sequence $\{w_1^S, \ldots, w_n^S\}$, Skip-gram model intends to minimize

$$\mathcal{L}_{mono}^S = -\sum_{c=K+1}^{n-K} \sum_{-K \leq k \leq K, k \neq 0} \log P(w_{c+k}^S | w_c^S), \tag{6.34}$$

where $K$ is the size of the sliding window. $P(w_{c+k}^S | w_c^S)$ stands for the predictive probability of one of the context words conditioned on the centered word $w_c^S$, formalized by the following softmax function:

$$P(w_{c+k}^S | w_c^S) = \frac{\exp(\mathbf{w}_{c+k}^S \cdot \mathbf{w}_c^S)}{\sum_{w_s^S \in V^S} \exp(\mathbf{w}_s^S \cdot \mathbf{w}_c^S)}, \tag{6.35}$$

in which $V^s$ indicates the word vocabulary of source language. $\mathcal{L}_{mono}^T$ can be formulated similarly.

**Cross-lingual Word Embedding Alignment**. Cross-lingual word embedding alignment aims to build a unified semantic space for the words in source and target languages. Inspired by [69], the cross-lingual word embeddings are aligned with signals of a seed lexicon and self-matching.

Formally, $\mathcal{L}_{cross}$ is composed of two terms including alignment by seed lexicon $\mathcal{L}_{seed}$ and alignment by matching $\mathcal{L}_{match}$:

$$\mathcal{L}_{cross} = \lambda_s \mathcal{L}_{seed} + \lambda_m \mathcal{L}_{match}, \tag{6.36}$$

where $\lambda_s$ and $\lambda_m$ are hyperparameters for controlling relative weightings of the two terms.

(1) **Alignment by Seed Lexicon**

The seed lexicon term $\mathcal{L}_{seed}$ encourages word embeddings of translation pairs in a seed lexicon $\mathcal{D}$ to be close, which can be achieved via an $L_2$ regularizer:

$$\mathcal{L}_{seed} = \sum_{\langle w_s^S, w_t^T \rangle \in \mathcal{D}} \|\mathbf{w}_s^S - \mathbf{w}_t^T\|^2, \tag{6.37}$$

in which $w_s^S$ and $w_t^T$ indicate the words in source and target languages in the seed lexicon, respectively.

(2) **Alignment by Matching Mechanism**

As for the matching process, it is found on the assumption that each target word should be matched to a single source word or a special empty word, and vice versa. The goal of the matching process is to find the matched source (target) word for each target (source) word and maximize the matching probabilities for all the matched word pairs. The loss of this part can be formulated as

$$\mathscr{L}_{match} = \mathscr{L}_{match}^{T2S} + \mathscr{L}_{match}^{S2T}, \tag{6.38}$$

where $\mathscr{L}_{match}^{T2S}$ is the term for target-to-source matching and $\mathscr{L}_{match}^{S2T}$ is the term for source-to-target matching.

Next, a detailed explanation of target-to-source matching is given, and the source-to-target matching is defined in the same way. A latent variable $m_t \in \{0, 1, \ldots, |V^S|\}$ $(t = 1, 2, \ldots, |V^T|)$ is first introduced for each target word $w_t^T$, where $|V^S|$ and $|V^T|$ indicate the vocabulary size of source and target languages, respectively. Here, $m_t$ specifies the index of the source word that $w_t^T$ matches with, and $m_t = 0$ signifies the empty word is matched. Then we have $\mathbf{m} = \{m_1, m_2, \ldots, m_{|V^T|}\}$, and can formalize the target-to-source matching term:

$$\mathscr{L}_{match}^{T2S} = -\log P(\mathscr{C}^T|\mathscr{C}^S) = -\log \sum_{\mathbf{m}} P(\mathscr{C}^T, \mathbf{m}|\mathscr{C}^S), \tag{6.39}$$

where $\mathscr{C}^T$ and $\mathscr{C}^S$ denote the target and source corpus, respectively. Here, they simply assume that the matching processes of target words are independent of each other. Therefore, we have

$$P(\mathscr{C}^T, \mathbf{m}|\mathscr{C}^S) = \prod_{w^T \in \mathscr{C}^T} P(w^T, \mathbf{m}|\mathscr{C}^S) = \prod_{t=1}^{|V^T|} P(w_t^T|w_{m_t}^S)^{c(w_t^T)}, \tag{6.40}$$

where $w_{m_t}^S$ is the source word that $w_t^T$ matches with, and $c(w_t^T)$ is the number of times $w_t^T$ occurs in the target corpus.

## 6.3.5 Other Sememe-Guided Applications

### 6.3.5.1 Chinese LIWC Lexicon Expansion

Linguistic Inquiry and Word Count (LIWC) [52] has been widely used for computerized text analysis in social science. Not only can LIWC be used to analyze text for classification and prediction, but it has also been used to examine the underlying psychological states of a writer or speaker. In the beginning, LIWC was developed to address content analytic issues in experimental psychology. Nowadays, there is an increasing number of applications across fields such as computational linguistics [22], demographics [48], health diagnostics [11], and social relationship [32].

Chinese is the most spoken language in the world, but we cannot use the original LIWC to analyze Chinese text. Fortunately, Chinese LIWC [28] has been released to fill the vacancy. In this part, we mainly focus on Chinese LIWC and using LIWC to stand for Chinese LIWC if not otherwise specified.

While LIWC has been used in a variety of fields, its lexicon only contains less than 7,000 words. This is insufficient because according to [39], there are at least 56,008 common words in Chinese. Moreover, LIWC lexicon does not consider emerging words and phrases on the Internet. Therefore, it is reasonable and necessary to expand the LIWC lexicon so that it is more accurate and comprehensive for scientific research. One way to expand LIWC lexicon is to annotate the new words manually. However, it is too time-consuming and often requires language expertise to add new words. Hence, expanding LIWC lexicon automatically is proposed.

In LIWC lexicon, words are labeled with different categories and categories form a certain hierarchy. Therefore, hierarchical classification algorithms can be naturally applied to LIWC lexicon. Reference [15] proposes Hierarchical SVM (Support Vector Machine), which is a modified version of SVM based on the hierarchical problem decomposition approach. In [6], the authors presented a novel algorithm which can be used on both tree- and Directed Acyclic Graph (DAG)-structured hierarchies. Some recent works [12, 33] attempted to use neural networks in the hierarchical classification.

However, these methods are often too generic without considering the special properties of words and LIWC lexicon. Many words and phrases have multiple meanings and are thereby classified into multiple leaf categories. This is often referred to as polysemy. Additionally, many categories in LIWC are fine-grained, thus making it more difficult to distinguish them. To address these issues, we introduce several models to incorporate sememe information when expanding the lexicon, which will be discussed after the introduction of the basic model.

**Basic Decoder for Hierarchical Classification**. First, we introduce the basic model for Chinese LIWC lexicon expansion. The well-known Sequence-to-Sequence decoder [64] is exploited for hierarchical classification. The original Sequence-to-Sequence decoder is often trained to predict the next word $w_t$ with consideration of all the previously predicted words $\{w_1, \dots, w_{t-1}\}$. This is a useful feature since an important difference between flat multilabel classification and hierarchical classification is that there are explicit connections among hierarchical labels. This property is utilized by transforming hierarchical labels into a sequence. Let $Y$ denote the label set and $\pi \colon Y \rightarrow Y$ denote the parent relationship where $\pi(y)$ is the parent node of $y \in Y$. Given a word $w$, its labels form a tree structure hierarchy. We then choose each path from the root node to the leaf node, and transform it into a sequence $\{y_1, y_2, \dots, y_L\}$ where $\pi(y_i) = y_{i-1}$, $\forall i \in [2, L]$ and $L$ is the number of levels in the hierarchy. In this way, when the model predicts a label $y_i$, it takes into consideration the probability of parent label sequence $\{y_1,\dots,y_{i-1}\}$. Formally, the decoder defines a probability over the label sequence:

$$P(y_1, y_2, \dots, y_L) = \prod_{i=1}^{L} P(y_i|(y_1, \dots, y_{i-1}), w). \tag{6.41}$$

A common approach for decoder is to use LSTM [27] so that each conditional probability is computed as

$$P(y_i|(y_1, \ldots, y_{i-1}), w) = g(\mathbf{y}_{i-1}, \mathbf{s}_i) = \mathbf{o}_i \odot \tanh(\mathbf{h}_i), \tag{6.42}$$

where

$$\begin{aligned}
\mathbf{h}_i &= \mathbf{f}_i \odot \mathbf{h}_{i-1} + \mathbf{i}_i \odot \tilde{\mathbf{h}}_i, \\
\tilde{\mathbf{h}}_i &= \tanh(\mathbf{W}_h[\mathbf{h}_{i-1}; \mathbf{y}_{i-1}] + \mathbf{b}_h), \\
\mathbf{o}_i &= \text{Sigmoid}(\mathbf{W}_o[\mathbf{h}_{i-1}; \mathbf{y}_{i-1}] + \mathbf{b}_o), \\
\mathbf{z}_i &= \text{Sigmoid}(\mathbf{W}_z[\mathbf{h}_{i-1}; \mathbf{y}_{i-1}] + \mathbf{b}_z), \\
\mathbf{f}_i &= \text{Sigmoid}(\mathbf{W}_f[\mathbf{h}_{i-1}; \mathbf{y}_{i-1}] + \mathbf{b}_f),
\end{aligned} \tag{6.43}$$

where $\odot$ is an element-wise multiplication and $\mathbf{h}_i$ is the $i$th hidden state of the RNN. $\mathbf{W}_h$, $\mathbf{W}_o$, $\mathbf{W}_z$, $\mathbf{W}_f$ are weights and $\mathbf{b}_h$, $\mathbf{b}_o$, $\mathbf{b}_z$, $\mathbf{b}_f$ are biases. $\mathbf{o}_i$, $\mathbf{z}_i$, and $\mathbf{f}_i$ are known as output gate layer, input gate layer, and forget gate layer, respectively.

To take advantage of word embeddings, the initial state $\mathbf{h}_0 = \mathbf{w}$ is defined where $\mathbf{w}$ represents the embedding of the word. In other words, the word embeddings are applied as the initial state of the decoder.

Specifically, the inputs of our model are word embeddings and label embeddings. First, raw words are transformed into word embeddings by an embedding matrix $\mathbf{E} \in \mathbb{R}^{|V| \times d_w}$, where $d_w$ is the word embedding dimension. Then, at each time step, label embeddings $\mathbf{y}$ are fed to the model, which is obtained by a label embedding matrix $\mathbf{Y} \in \mathbb{R}^{|Y| \times d_y}$, where $d_y$ is the label embedding dimension. Here word embeddings are pretrained and fixed during training.

Generally speaking, the decoder is expected to decode word labels hierarchically based on word embeddings. At each time step, the decoder will predict the current label depending on previously predicted labels.

**Hierarchical Decoder with Sememe Attention**. The basic decoder uses word embeddings as the initial state, then predicts word labels hierarchically as sequences. However, each word in the basic decoder model has only one representation. This is insufficient because many words are polysemous and many categories are fine-grained in the LIWC lexicon. It is difficult to handle these properties using a single real-valued vector. Therefore, Zeng et al. [68] propose to incorporate sememe information.

Because different sememes represent different meanings of a word, they should have different weights when predicting word labels. Moreover, we believe that the same sememe should have different weights in different categories. Take the word `apex` in Fig. 6.12, for example. The sememe `location` should have a relatively higher weight when the decoder chooses among the subclasses of `relative`. When choosing among the subclasses of `PersonalConcerns`, `location` should have a lower weight because it represents a relatively irrelevant sense `vertex`.

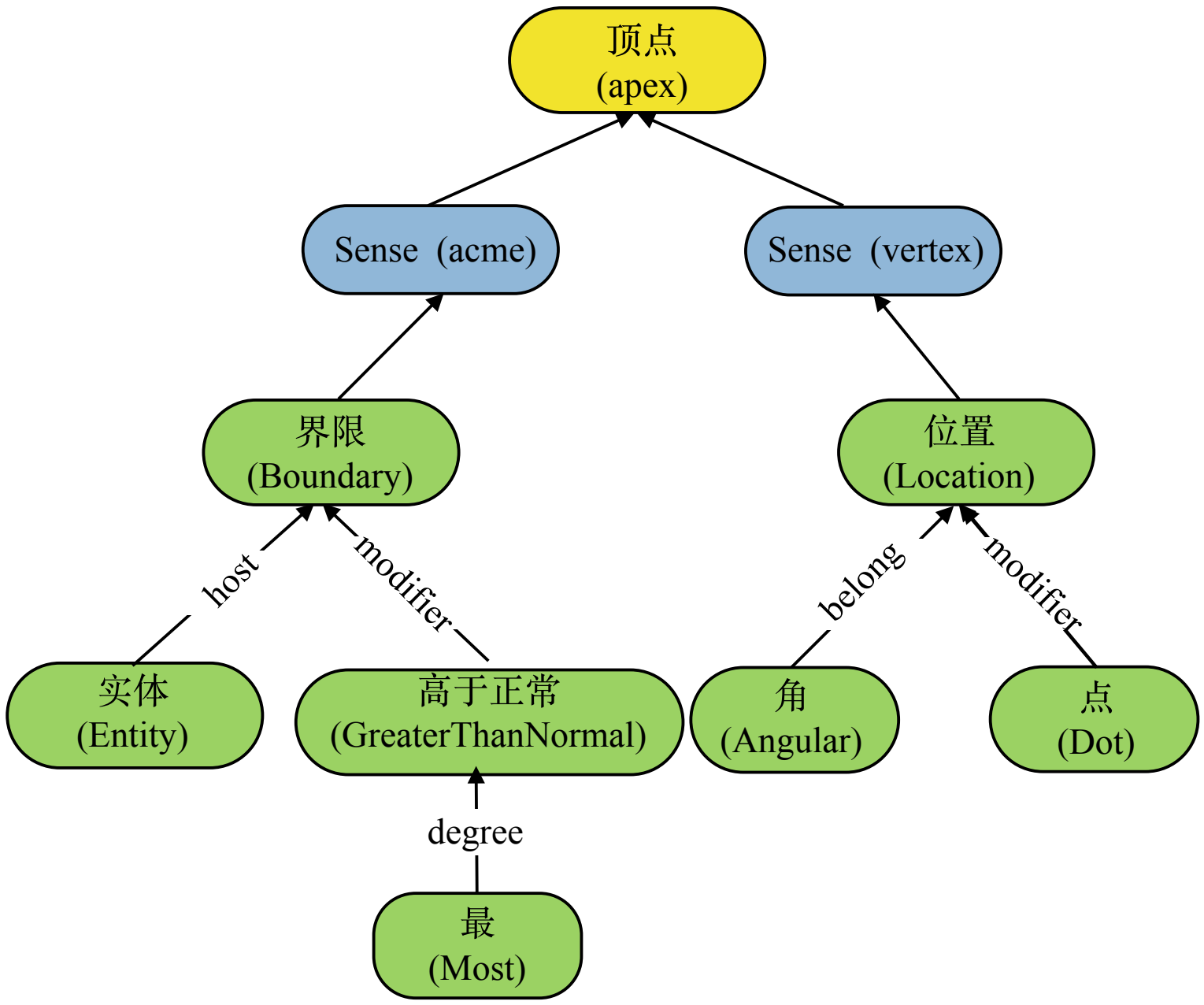

**Fig. 6.12** Example word `apex` and its senses and sememes in HowNet annotation

To achieve these goals, the utilization of attention mechanism [1] is proposed to incorporate sememe information when decoding the word label sequence. The structure of the model is illustrated in Fig. 6.13.

Similar to the basic decoder approach, word embeddings are applied as the initial state of the decoder. The primary difference is that the conditional probability is defined as

$$P(y_i|(y_1, \ldots, y_{i-1}), w, c_i) = g([\mathbf{y}_{i-1}; \mathbf{c}_i], \mathbf{h}_i), \tag{6.44}$$

where $\mathbf{c}_i$ is known as context vector. The context vector $\mathbf{c}_i$ depends on a set of sememe embeddings $\{\mathbf{x}_1, \ldots, \mathbf{x}_N\}$, acquired by a sememe embedding matrix $\mathbf{X} \in \mathbb{R}^{|S| \times d_s}$, where $d_s$ is the sememe embedding dimension.

To be more specific, the context vector $\mathbf{c}_i$ is computed as a weighted sum of the sememe embedding $\mathbf{x}_j$:

$$\mathbf{c}_i = \sum_{j=1}^{N} \alpha_{ij} \mathbf{x}_j. \tag{6.45}$$

The weight $\alpha_{ij}$ of each sememe embedding $\mathbf{x}_j$ is defined as

$$\alpha_{ij} = \frac{\exp(\mathbf{v} \cdot \tanh(\mathbf{W}_1 \mathbf{y}_{i-1} + \mathbf{W}_2 \mathbf{x}_j))}{\sum_{k=1}^{N} \exp(\mathbf{v} \cdot \tanh(\mathbf{W}_1 \mathbf{y}_{i-1} + \mathbf{W}_2 \mathbf{x}_k))}, \tag{6.46}$$

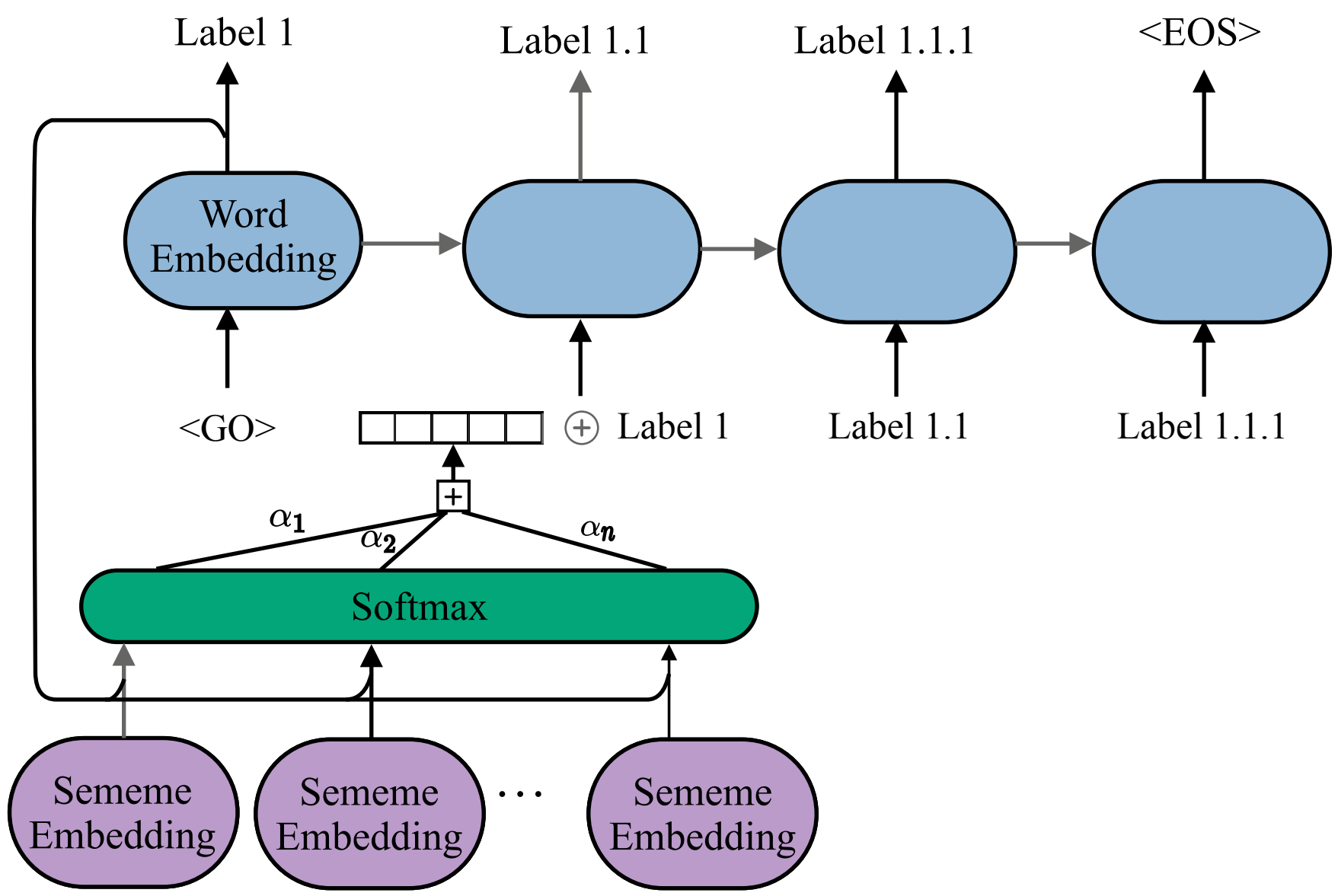

**Fig. 6.13** The architecture of sememe attention decoder with word embeddings as the initial state

where $\mathbf{v} \in \mathbb{R}^a$ is a trainable parameter, $\mathbf{W}_1 \in \mathbb{R}^{a\times d_y}$ and $\mathbf{W}_2 \in \mathbb{R}^{a\times d_s}$ are weight matrices, and $a$ is the number of hidden units in attention model.

Intuitively, at each time step, the decoder chooses which sememes to pay attention to when predicting the current word label. In this way, different sememes can have different weights, and the same sememe can have different weights in different categories. With the support of sememe attention, the decoder can differentiate multiple meanings in a word and the fine-grained categories and thus can expand a more accurate and comprehensive lexicon.

## 6.4 Summary

In this chapter, we first give an introduction to the most well-known sememe knowledge base, HowNet, which uses about 2, 000 predefined sememes to annotate over 100, 000 Chinese and English words and phrases. Different from other linguistic knowledge bases like WordNet, HowNet is based on the minimum semantics units (sememes) and captures the compositional relations between sememes and words. To learn the representations of sememe knowledge, we elaborate on three models, namely Simple Sememe Aggregation model (SSA), Sememe Attention over Context model (SAC), and Sememe Attention over Target model (SAT). These models not only learn the representations of sememes but also help improve the representations of words. Next, we describe some applications of sememe knowledge, including word representation, semantic composition, and language modeling. We also detail how to automatically predict sememes for both monolingual and cross-lingual unannotated words.

For further learning of sememe knowledge-based NLP, you can read the book written by the authors of HowNet [18]. You can also find more related papers in this paper list https://github.com/thunlp/SCPapers. You can use the open source API OpenHowNet https://github.com/thunlp/OpenHowNet to access HowNet data.

In the future, there are some research directions worth exploring:

(1) **Utilizing Structures of Sememe Annotations**. The sememe annotations in HowNet are hierarchical, and sememes annotated to a word are actually organized as a tree. However, existing studies still do not utilize the structural information of sememes. Instead, in current methods, sememes are simply regarded as semantic labels. In fact, the structures of sememes also incorporate abundant semantic information and will be helpful to the deep understanding of lexical semantics. Besides, existing sememe prediction studies also predict unstructured sememes only, and it is an interesting task to conduct structured sememe predictions.

(2) **Leveraging Sememes in Low-data Regimes**. One of the most important and typical characteristics of sememes is that limited sememes can represent unlimited semantics, which can play an important and positive role in tackling the low-data regimes. In word representation learning, the representations of low-frequency words can be improved by their sememes, which have been well learned with the high-frequency words they annotate. We believe sememes will be beneficial to other low-data regimes, e.g., low-resource language NLP tasks.

(3) **Building Sememe Knowledge Bases for Other Languages**. Original HowNet annotates sememes for only two languages: Chinese and English. As far as we know, there are not sememe knowledge bases like HowNet in other languages. Since HowNet and its sememe knowledge have been verified helpful for better understanding human languages, it will be of great significance to annotate sememes for words and phrases in other languages. In the section, we have described a study on cross-lingual sememe prediction. And we think it is promising to make efforts toward this direction.

# Chapter 7
# World Knowledge Representation

**Abstract** World knowledge representation aims to represent entities and relations in the knowledge graph in low-dimensional semantic space, which have been widely used in large knowledge-driven tasks. In this chapter, we first introduce the concept of the knowledge graph. Next, we introduce the motivations and give an overview of the existing approaches for knowledge graph representation. Further, we discuss several advanced approaches that aim to deal with the current challenges of knowledge graph representation. We also review the real-world applications of knowledge graph representation, such as language modeling, question answering, information retrieval, and recommender systems.

## 7.1 Introduction

Knowledge Graph (KG), which is also named as Knowledge Base (KB), is a significant multi-relational dataset for modeling concrete entities and abstract concepts in the real world. It provides useful structured information and plays a crucial role in lots of real-world applications such as web search and question answering. It is not exaggerated to say that knowledge graphs teach us how to model the entities as well as the relationships among them in this complicated real world.

To encode knowledge into a real-world application, knowledge graph representation, which represents entities and relations in knowledge graphs with distributed representations, has been proposed and applied to various real-world artificial intelligence fields including question answering, information retrieval, and dialogue system. That is, knowledge graph representation learning plays a vital role as a bridge between knowledge graphs and knowledge-driven tasks.

In this section, we will introduce the concept of knowledge graph, several typical knowledge graphs, knowledge graph representation learning, and several typical knowledge-driven tasks.

## 7.1.1 World Knowledge Graphs

In ancient times, knowledge was stored and inherited through books and letters written on parchment or bamboo slip. With the Internet thriving in the twenty-first century, millions of thousands of messages have flooded into the World Wide Web, and knowledge was transferred to the semi-structured textual information on the web. However, due to the information explosion, it is not easy to extract knowledge we want from the significant, noisy plain text on the Internet. To obtain knowledge effectively, people notice that the world is not only made of strings but also made of entities and relations. Knowledge Graph, which arranges structured multi-relational data of concrete entities and abstract concepts in the real world, is blooming in recent years and attracts wide attention in both academia and industry.

KGs are usually constructed from existing Semantic Web datasets in Resource Description Framework (RDF) with the help of manual annotation, while it can also be automatically enriched by extracting knowledge from large plain texts on the Internet. A typical KG usually contains two elements, including entities (i.e., concrete entities and abstract concepts in the real world) and relations between entities. It usually represents knowledge with large quantities of triple facts in the triple form of $\langle$*head entity*, *relation*, *tail entity*$\rangle$ abridged as $\langle h, r, t\rangle$. For example, `William Shakespeare` is a famous English poet and playwright, who is widely regarded as the greatest writer in the English language, and `Romeo and Juliet` is one of his masterpieces. In knowledge graph, we will represent this knowledge as $\langle$`William Shakespeare`, `works_written`, `Romeo and Juliet`$\rangle$. Note that in the real world, the same head entity and relation may have multiple tail entities (e.g., `William Shakespeare` also wrote `Hamlet` and `A Midsummer Night's Dream`), and reversely the same situation will happen when tail entity and relation are fixed. Even it is possible when both the head entity and tail entity are multiple (e.g., in relations like `actor_in_movie`). However, in KG, all knowledge can be represented in triple facts regardless of the types of entities and relations. Through these triples, we can generate a huge directed graph whose nodes correspond to entities and edges correspond to relations to model the real world. With the well-structured united knowledge representation, KGs are widely used in a variety of applications to enhance their system performance.

There are several KGs widely utilized nowadays in applications of information retrieval and question answering. In this subsection, we will introduce some famous KGs such as Freebase, DBpedia, Yago, and WordNet. In fact, there are also lots of comparatively smaller KGs in specific fields of knowledge functioned in vertical search.

#### 7.1.1.1 Freebase

Freebase is one of the most popular knowledge graphs in the world. It is a large community-curated database consisting of well-known people, places, and things, which is composed of existing databases and its community members. Freebase was first developed by Metaweb, an American software company, and ran since March 2007. In July 2010, Metaweb was acquired by Google, and Freebase was combined to power up Google's Knowledge Graph. In December 2014, the Freebase team officially announced that the website, as well as the API of Freebase, would be shut down by June 30, 2015. While the data in Freebase would be transferred to Wikidata, which is another collaboratively edited knowledge base operated by Wikimedia Foundation. Up to March 24, 2016, Freebase arranged 58,726,427 topics and 3,197,653,841 facts.

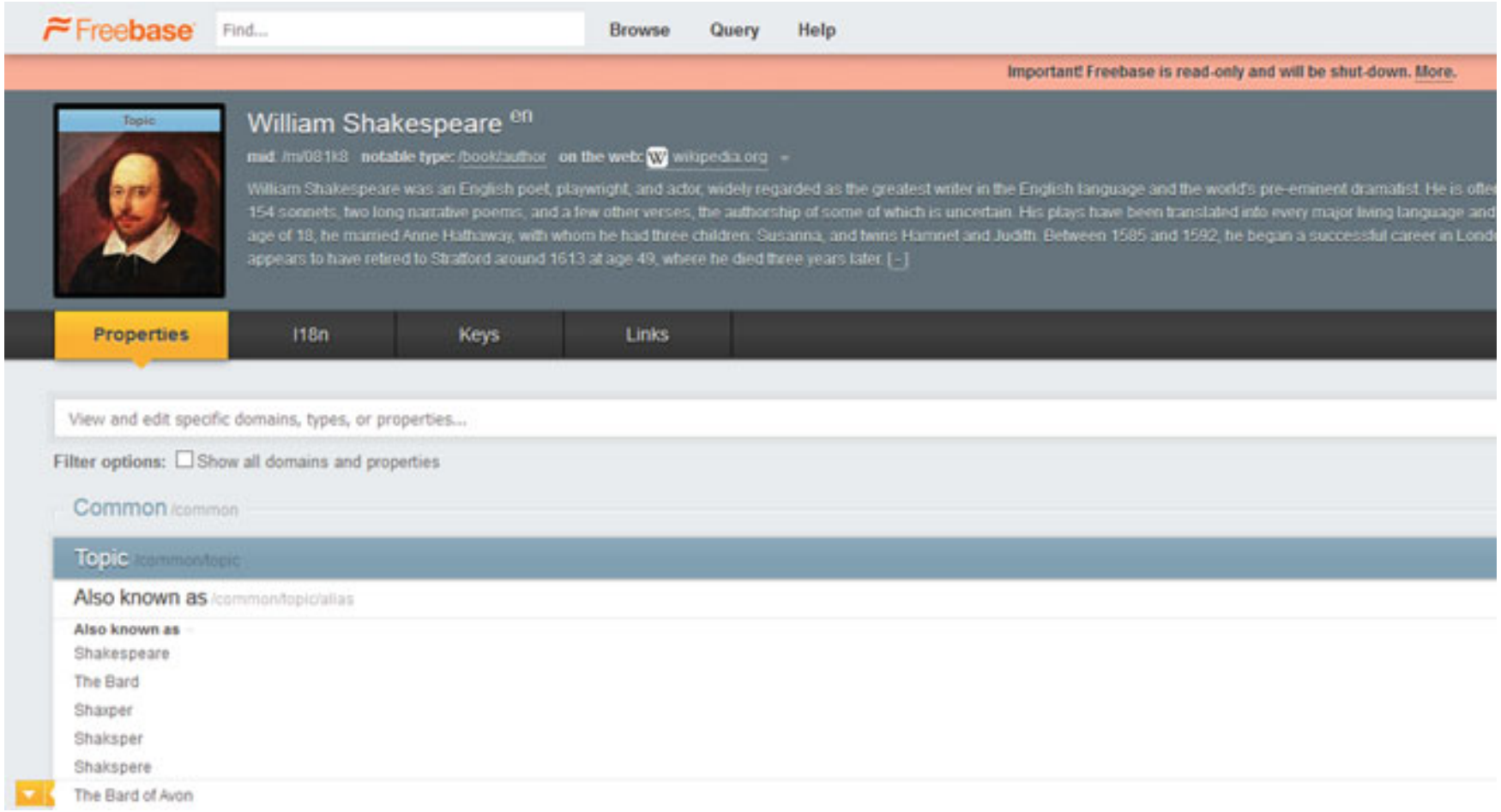

**Fig. 7.1** An example of search results in Freebase

Freebase contains well-structured data representing relationships between entities as well as the attributes of entities in the form of triple facts (Fig. 7.1). Data in Freebase was mainly harvested from various sources, including Wikipedia, Fashion Model Directory, NNDB, MusicBrainz, and so on. Moreover, the community members also contributed a lot to Freebase. Freebase is an open and shared database that aims to construct a global database which encodes the world's knowledge. It announced an open API, RDF endpoint, and a database dump for its users for both commercial and noncommercial use. As described by Tim O'Reilly, Freebase is the bridge between the bottom-up vision of Web 2.0 collective intelligence and the more structured world of the Semantic Web.

#### 7.1.1.2 DBpedia

DBpedia is a crowd-sourced community effort aiming to extract structured content from Wikipedia and make this information accessible on the web. It was started by researchers at Free University of Berlin, Leipzig University and OpenLink Software, initially released to the public in January 2007. DBpedia allows users to ask semantic queries associated with Wikipedia resources, even including links to other related datasets, which makes it easier for us to fully utilize the massive amount of information in Wikipedia in a novel and effective way. DBpedia is also an essential part of the Linked Data effort described by Tim Berners-Lee.

The English version of DBpedia describes 4.58 million entities, out of which 4.22 million are classified in a consistent ontology, including 1,445,000 persons, 735,000 places, 411,000 creative works, 251,000 species, 241,000 organizations, and 6,000 diseases. There are also localized versions of DBpedia in 125 languages, all of which contain 38.3 million entities. Besides, DBpedia also contains a great number of internal and external links, including 80.9 million links to Wikipedia categories, 41.2 million links to YAGO categories, 25.2 million links to images, and 29.8 million links to external web pages. Moreover, DBpedia maintains a hierarchical, cross-domain ontology covering overall 685 classes, which has been manually created based on the commonly used infoboxes in Wikipedia.

DBpedia has several advantages over other KGs. First, DBpedia has a close connection to Wikipedia and can automatically evolve as Wikipedia changes. It makes the update process of DBpedia more efficient. Second, DBpedia is multilingual that is convenient for users over the world with their native languages.

#### 7.1.1.3 YAGO

YAGO, which is short for Yet Another Great Ontology, is a high-quality KG developed by Max Planck Institute for Computer Science in Saarbruücken initially released in 2008. Knowledge in YAGO is automatically extracted from Wikipedia, WordNet, and GeoNames, whose accuracy has been manually evaluated and proves a confirmed accuracy of 95%. YAGO is special not only because of the confidence value every fact possesses depending on the manual evaluation but also because that YAGO is anchored in space and time, which can provide a spatial dimension or temporal dimension to part of its entities.

Currently, YAGO has more than 10 million entities, including persons, organizations, and locations, with over 120 million facts about these entities. YAGO also combines knowledge extracted from Wikipedias of 10 different languages and classifies them into approximately 350,000 classes according to the Wikipedia category system and the taxonomy of WordNet. YAGO has also joined the linked data project and been linked to the DBpedia ontology and the SUMO ontology (Fig. 7.2).

## 7.2 Knowledge Graph Representation

Knowledge Graphs provide us with a novel aspect to describe the world with entities and triple facts, which attract growing attention from researchers. Large KGs such as Freebase, DBpedia, and YAGO have been constructed and widely used in an enormous amount of applications such as question answering and Web search.

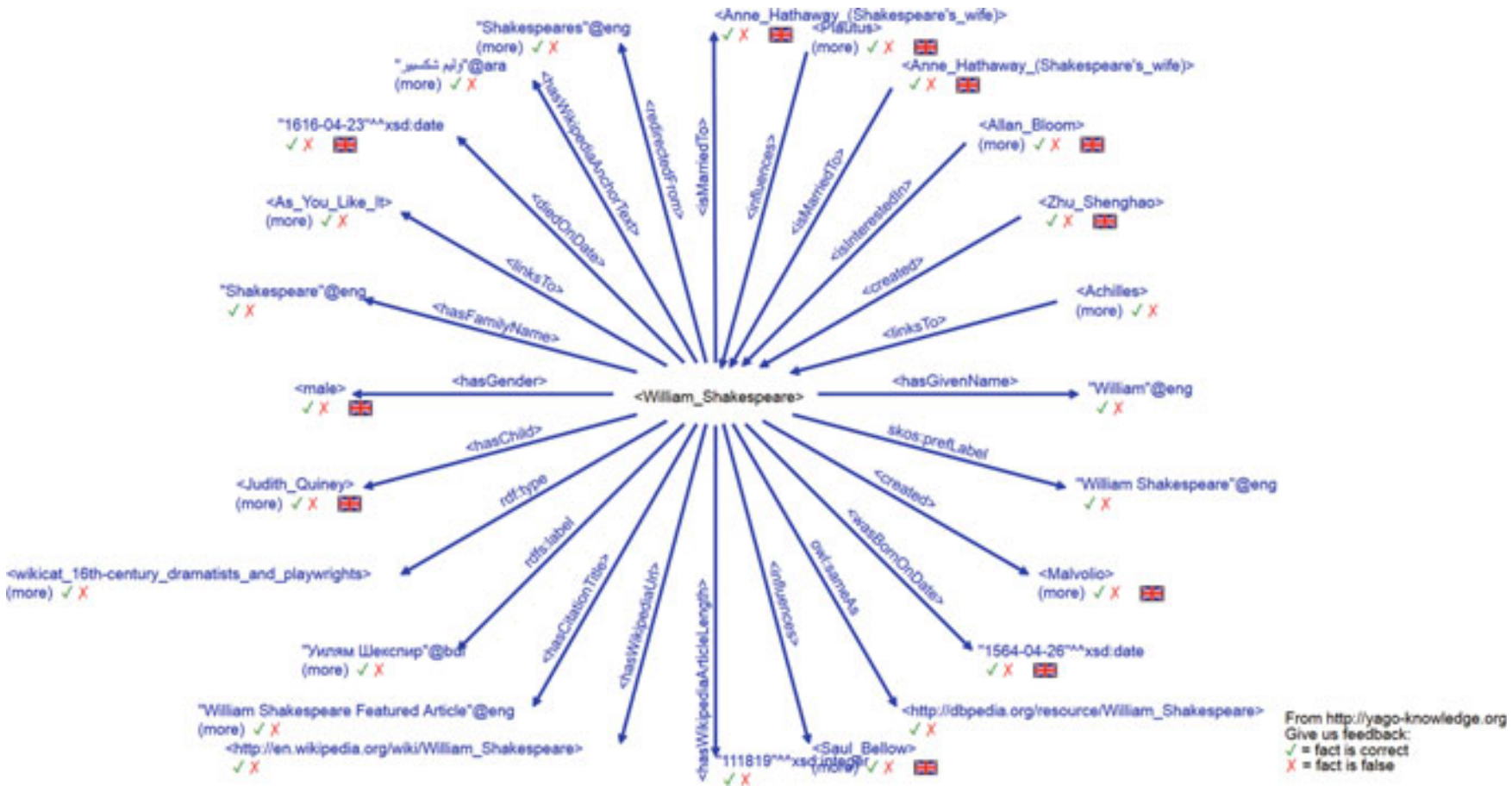

**Fig. 7.2** An example of search results in YAGO

However, with KG size increasing, we are facing two main challenges: data sparsity and computational inefficiency. Data sparsity is a general problem in lots of fields like social network analysis or interest mining. It is because that there are too many nodes (e.g., users, products, or entities) in a large graph, while too few edges (e.g., relationships) between these nodes, since the number of relations of a node is limited in the real world. Computational efficiency is another challenge we need to overcome with the increasing size of knowledge graphs.

To tackle these problems, representation learning is introduced to knowledge representation. Representation learning in KGs aims to project both entities and relations into a low-dimensional continuous vector space to get their distributed representations, whose performance has been confirmed in word representation and social representation. Compared with the traditional one-hot representation, distributed representation has much fewer dimensions, and thus lowers the computational complexity. What is more, distributed representation can explicitly show the similarity between entities through some distance calculated by the low-dimensional embeddings, while all embeddings in one-hot representation are orthogonal, making it difficult to tell the potential relations between entities.

With the advantages above, knowledge graph representation learning is blooming in knowledge applications, significantly improving the ability of KGs on the task of knowledge completion, knowledge fusion, and reasoning. It is considered as the bridge between knowledge construction, knowledge graphs, and knowledge-driven applications. Up till now, a high number of methods have been proposed using a distributed representation for modeling knowledge graphs, with the learned knowledge representations widely utilized in various knowledge-driven tasks like question answering, information retrieval, and dialogue system.

In summary, Knowledge graph Representation Learning (KRL) aims to construct distributed knowledge representations for entities and relations, projecting knowledge into low-dimensional semantic vector spaces. Recent years have witnessed significant advances in knowledge graph representation learning with a large amount of KRL methods proposed to construct knowledge representations, among which the translation-based methods achieve state-of-the-art performance in many KG tasks, with a right balance in both effectiveness and efficiency.

In this section, we will first describe the notations that we will use in KRL. Then, we will introduce TransE, which is the fundamental version of translation-based methods. Next, we will explore the various extension methods of TransE in detail. At last, we will take a brief look over other representation learning methods utilized in modeling knowledge graphs.

### *7.2.1 Notations*

First, we introduce the general notations used in the rest of this section. We use $G = (E, R, T)$ to denote the whole KG, in which $E = \{e_1, e_2, \ldots, e_{|E|}\}$ stands for the entity set, $R = \{r_1, r_2, \ldots, r_{|R|}\}$ stands for the relation set, and $T$ stands for the triple set. $|E|$ and $|R|$ are the corresponding entity and relation numbers in their overall sets. As stated above, we represent knowledge in the form of triple fact $\langle h, r, t\rangle$, where $h \in E$ means the head entity, $t \in E$ means the tail entity, and $r \in R$ means the relation between $h$ and $t$.

### *7.2.2 TransE*

TransE [7] is a translation-based model for learning low-dimensional embeddings of entities and relations. It projects entities as well as relations into the same semantic embedding space, and then considers relations as translations in the embedding space. First, we will start with the motivations of this method, and then discuss the details in how knowledge representations are trained under TransE. Finally, we will explore the advantages and disadvantages of TransE for a deeper understanding.

#### 7.2.2.1 Motivation

There are three main motivations behind the translation-based knowledge graph representation learning method. The primary motivation is that it is natural to consider relationships between entities as translating operations. Through distributed representations, entities are projected to a low-dimensional vector space. Intuitively, we agree that a reasonable projection should map entities with similar semantic meanings to the same field, while entities with different meanings should belong to distinct clusters in the vector space. For example, `William Shakespeare` and `Jane Austen` may be in the same cluster of writers, `Romeo, and Juliet` and `Pride and Prejudice` may be in another cluster of books. In this case, they share the same relation `works_written`, and the translations between writers and books in the vector space are similar.

The secondary motivation of TransE derives from the breakthrough in word representation by Word2vec [49]. Word2vec proposes two simple models, Skip-gram and CBOW, to learn word embeddings from large-scale corpora, significantly improving the performance in word similarity and analogy. The word embeddings learned by Word2vec have some interesting phenomena: if two word-pairs share the same semantic or syntactic relationships, their subtraction embeddings in each word pair will be similar. For instance, we have

$$\mathbf{w}(\texttt{king}) - \mathbf{w}(man) \approx \mathbf{w}(\texttt{queen}) - \mathbf{w}(\texttt{woman}), \tag{7.1}$$

which indicates that the latent semantic relation between `king` and `man`, which is similar to the relation between `queen` and `woman`, is successfully embedded in the word representation. This approximate relation could be found not only with the semantic relations but also with the syntactic relations. We have

$$\mathbf{w}(\texttt{bigger}) - \mathbf{w}(\texttt{big}) \approx \mathbf{w}(\texttt{smaller}) - \mathbf{w}(\texttt{small}). \tag{7.2}$$

The phenomenon found in word representation strongly implies that there may exist an explicit method to represent relationships between entities as translating operations in vector space.

The last motivation comes from the consideration of computational complexity. On the one hand, a substantial increase in model complexity will result in high computational costs and obscure model interpretability. Moreover, a complex model may lead to overfitting. On the other hand, experimental results on model complexity demonstrate that the simpler models perform almost as good as more expressive models in most KG applications, in the condition that there are sizeable multi-relational dataset and a relatively large amount of relations. As KG size increases, computational complexity becomes the primary challenge in the knowledge graph representation. The intuitive assumption of translation leads to a better trade-off between accuracy and efficiency.

#### 7.2.2.2 Methodology

As illustrated in Fig. 7.3, TransE projects entities and relations into the same low-dimensional space. All embeddings take values in $\mathbb{R}^d$, where $d$ is a hyperparameter indicating the dimension of embeddings. With the translation assumption, for each triple $\langle h, r, t\rangle$ in $T$, we want the summation embedding $\mathbf{h} + \mathbf{r}$ to be the nearest neighbor of tail embedding $\mathbf{t}$. The score function of TransE is then defined as follows:

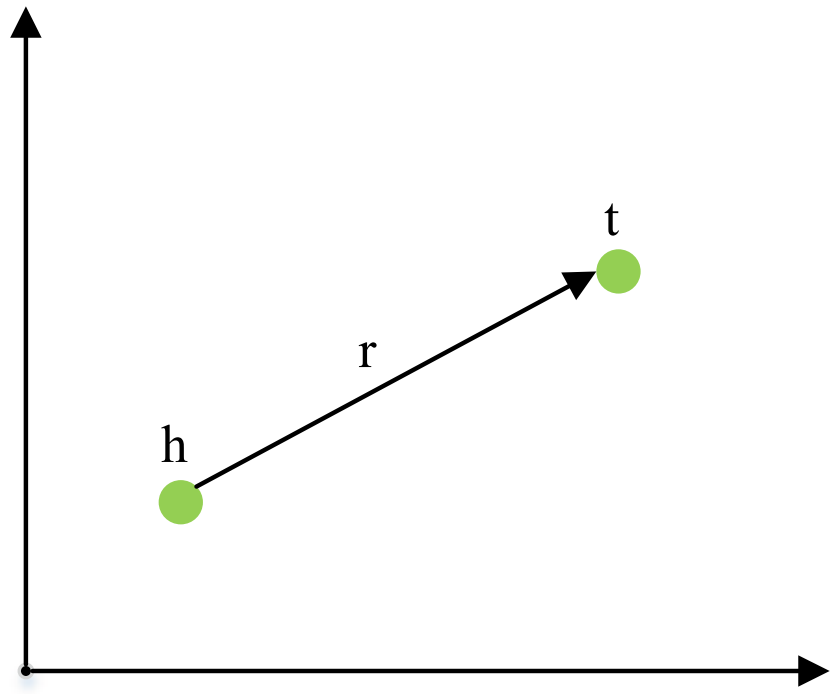

**Fig. 7.3** The architecture of TransE model [47]

$$\mathcal{E}(h, r, t) = \|\mathbf{h} + \mathbf{r} - \mathbf{t}\|. \tag{7.3}$$

More specifically, to learn such embeddings of entities and relations, TransE formalizes a margin-based loss function with negative sampling as objective for training. The pair-wise function is defined as follows:

$$\mathcal{L} = \sum_{\langle h,r,t\rangle \in T} \sum_{\langle h',r',t'\rangle \in T^-} \max(\gamma + \mathcal{E}(h, r, t)) - \mathcal{E}(h', r', t'), 0), \tag{7.4}$$

in which $\mathcal{E}(h, r, t)$ is the score of energy function for a positive triple (i.e., triple in $T$) and $\mathcal{E}(h', r', t')$ is that of a negative triple. The energy function $\mathcal{E}$ can be either measured by $L_1$ or $L_2$ distance. $\gamma > 0$ is a hyperparameter of margin and a bigger $\gamma$ means a wider gap between positive and the corresponding negative scores. $T^-$ is the negative triple set with respect to $T$.

Since there are no explicit negative triples in knowledge graphs, we define $T^-$ as follows:

$$T^- = \{\langle h', r, t\rangle | h' \in E\} \cup \{\langle h, r', t\rangle | r' \in R\} \cup \{\langle h, r, t'\rangle | t' \in E\}, \quad \langle h, r, t\rangle \in T, \tag{7.5}$$

which means the negative triple set $T^-$ is composed of the positive triple $\langle h, r, t\rangle$ with head entity, relation, or tail entity randomly replaced by any other entities or relations in KG. Note that the new triple generated after replacement will not be considered as a negative sample if it has already been in $T$.

TransE is optimized using mini-batch stochastic gradient descent (SGD), with entities and relations randomly initialized. Knowledge completion, which is a link prediction task aiming to predict the third element in a triple (could be either entity or relation) with the given rest two elements, is designed to evaluate the learned knowledge representations.

#### 7.2.2.3 Disadvantages and Challenges

TransE is effective and efficient and has shown its power on link prediction. However, it still has several disadvantages and challenges to be further explored.

First, in knowledge completion, we may have multiple correct answers with the given two elements in a triple. For instance, with the given head entity `William Shakespeare` and the relation `works_written`, we will get a list of masterpieces including `Romeo and Juliet`, `Hamlet` and `A Midsummer Night's Dream`. These books share the same information in the writer while differing in many other fields such as theme, background, and famous roles in the book. However, with the translation assumption in TransE, every entity has only one embedding in all triples, which significantly limits the ability of TransE in knowledge graph representations. In [7], the authors categorize all relations into four classes, 1-to-1, 1-to-Many, Many-to-1, Many-to-Many, according to the cardinalities of their head and tail arguments. A relation is considered as 1-to-1 if most heads appear with one tail, 1-to-Many if a head can appear with many tails, Many-to-1 if a tail can appear with many heads, and Many-to-Many if multiple heads appear with multiple tails. Statistics demonstrate that the 1-to-Many, Many-to-1, Many-to-Many relations occupy a large proportion. TransE does well in 1-to-1, but it has issues when dealing with 1-to-Many, Many-to-1, Many-to-Many relations. Similarly, TransE may also struggle with reflexive relations.

Second, the translating operation is intuitive and effective, only considering the simple one-step translation, which may limit the ability to model KGs. Taking entities as nodes and relations as edges, we can construct a huge knowledge graph with the triple facts. However, TransE focuses on minimizing the energy function $\mathcal{E}(h, r, t) = \|\mathbf{h} + \mathbf{r} - \mathbf{t}\|$, which only utilize the one-step relation information in knowledge graphs, regardless of the latent relationships located in long-distance paths. For example, if we know the triple fact that ⟨`The forbidden city`, `locate_in`, `Beijing`⟩ and ⟨`Beijing`, `capital_of`, `China`⟩, we can infer that `The forbidden city` locates in `China`. TransE can be further enhanced with the favor of multistep information.

Third, the representation and the dissimilarity function in TransE are oversimplified for the consideration of efficiency. Therefore, TransE may not be capable enough of modeling those complicated entities and relations in knowledge graphs. There still exist challenges on how to balance the effectiveness and efficiency, avoiding both overfitting and underfitting.

Besides the disadvantages and challenges stated above, multisource information such as textual information and hierarchical type/label information is of great significance, which will be further discussed in the following.

### *7.2.3 Extensions of TransE*

There are lots of extension methods following TransE to address the challenges above. Specifically, TransH, TransR, TransD, and TranSparse are proposed to solve the challenges in modeling 1-to-Many, Many-to-1, and Many-to-Many relations, PTransE is proposed to encode long-distance information located in multistep paths, and CTransR, TransA, TransG, and KG2E further extend the oversimplified model of TransE. We will discuss these extension methods in detail.

#### 7.2.3.1 TransH

With distributed representation, entities are projected to the semantic vector space, and similar entities tend to be in the same cluster. However, it seems that `William Shakespeare` should be in the neighborhood of `Isaac Newton` when talking about nationality, while it should be next to `Mark Twain` when talking about occupation. To accomplish this, we want entities to show different preferences in different situations, that is, to have multiple representations in different triples.

To address the issue when modeling 1-to-Many, Many-to-1, Many-to-Many, and reflexive relations, TransH [77] enables an entity to have multiple representations when involved in different relations. As illustrated in Fig. 7.4, TransH proposes a relation-specific hyperplane $\mathbf{w}_r$ for each relation, and judge dissimilarities on the hyperplane instead of the original vector space of entities. Given a triple $\langle h, r, t\rangle$, TransH first projects $\mathbf{h}$ and $\mathbf{t}$ to the corresponding hyperplane $\mathbf{w}_r$ to get the projection $\mathbf{h}_\perp$ and $\mathbf{t}_\perp$, and the translation vector $\mathbf{r}$ is used to connect $\mathbf{h}_\perp$ and $\mathbf{t}_\perp$ on the hyperplane. The score function is defined as follows:

$$\mathscr{E}(h, r, t) = \|\mathbf{h}_\perp + \mathbf{r} - \mathbf{t}_\perp\|, \tag{7.6}$$

in which we have

$$\mathbf{h}_\perp = \mathbf{h} - \mathbf{w}_r^\top \mathbf{h}\mathbf{w}_r, \quad \mathbf{t}_\perp = \mathbf{t} - \mathbf{w}_r^\top \mathbf{t}\mathbf{w}_r, \tag{7.7}$$

where $\mathbf{w}_r$ is a vector and $\|\mathbf{w}_r\|_2$ is restricted to 1. As for training, TransH also minimizes the margin-based loss function with negative sampling which is similar to TransE, and use mini-batch SGD to learn representations.

#### 7.2.3.2 TransR/CTransR

TransH enables entities to have multiple representations in different relations with the favor of hyperplanes, while entities and relations are still restricted in the same semantic vector space, which may limit the ability for modeling entities and relations. TransR [39] assumes that entities and relations should be arranged in distinct spaces, that is, entity space for all entities and relation space for each relation.

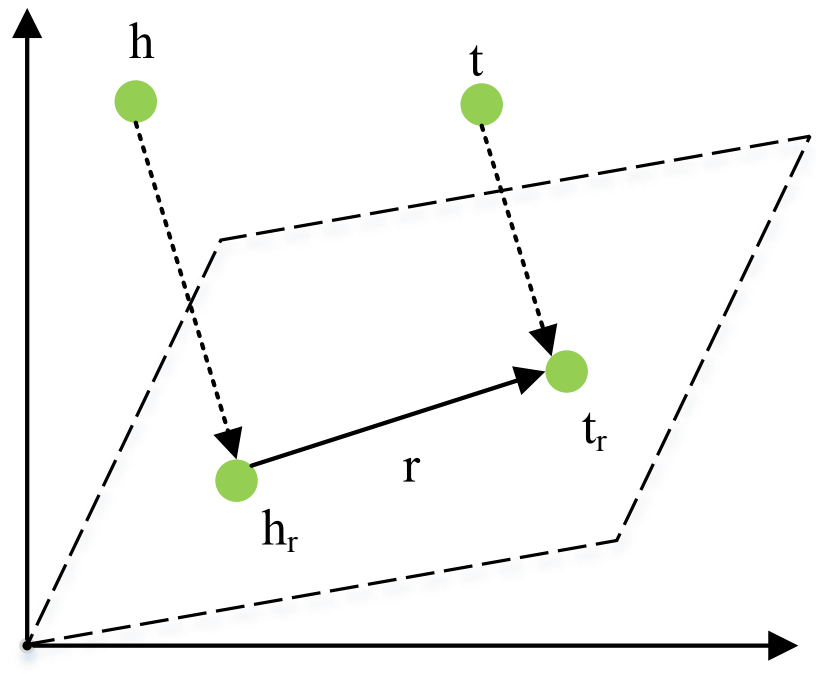

**Fig. 7.4** The architecture of TransH model [47]

As illustrated in Fig. 7.5, For a triple $\langle h, r, t\rangle$, $\mathbf{h}, \mathbf{t} \in \mathbb{R}^k$ and $\mathbf{r} \in \mathbb{R}^d$, TransR first projects $\mathbf{h}$ and $\mathbf{t}$ from entity space to the corresponding relation space of $r$. That is to say, every entity has a relation-specific representation for each relation, and the translating operation is processed in the specific relation space. The energy function of TransR is defined as follows:

$$\mathcal{E}(h, r, t) = \|\mathbf{h}_r + \mathbf{r} - \mathbf{t}_r\|, \tag{7.8}$$

where $\mathbf{h}_r$ and $\mathbf{t}_r$ stand for the relation-specific representation for $\mathbf{h}$ and $\mathbf{t}_r$ in the corresponding relation space of $r$. The projection from entity space to relation space is

$$\mathbf{h}_r = \mathbf{h}\mathbf{M}_r, \quad \mathbf{t}_r = \mathbf{t}\mathbf{M}_r, \tag{7.9}$$

where $\mathbf{M}_r \in \mathbb{R}^{k\times d}$ is a projection matrix mapping entities from the entity space to the relation space of $r$. TransR also constrains the norms of the embeddings and has $\|\mathbf{h}\|_2 \le 1$, $\|\mathbf{t}\|_2 \le 1$, $\|\mathbf{r}\|_2 \le 1$, $\|\mathbf{h}_r\|_2 \le 1$, $\|\mathbf{t}_r\|_2 \le 1$. As for training, TransR shares the same margin-based score function as TransE.

Furthermore, the author found that some relations in knowledge graphs could be divided into a few sub-relations that give more precise information. The differences between those sub-relations can be learned from corresponding entity pairs. For instance, the relation `location_contains` has head-tail patterns like `city-street`, `country-city`, and even `country-university`, showing different attributes in cognition. With the sub-relations being considered, entities may be projected to more precise positions in the semantic vector space.

Cluster-based TransR (CTransR), which is an enhanced version of TransR with the sub-relations into consideration, is then proposed. More specifically, for each relation $r$, all entity pairs $(h, t)$ are first clustered into several groups. The clustering of entity pairs depends on the subtraction result of $\mathbf{t} - \mathbf{h}$, in which $\mathbf{h}$ and $\mathbf{t}$ are pretrained by TransE. Next, we learn a distinct sub-relation vector $\mathbf{r}_c$ for each cluster according to the corresponding entity pairs, and the original energy function is modified as

$$\mathscr{E}(h, r, t) = \|\mathbf{h}_r + \mathbf{r}_c - \mathbf{t}_r\| + \alpha\|\mathbf{r}_c - \mathbf{r}\|, \tag{7.10}$$

where $\|\mathbf{r}_c - \mathbf{r}\|$ wants the sub-relation vector $\mathbf{r}_c$ not to be too distinct from the unified relation vector $\mathbf{r}$.

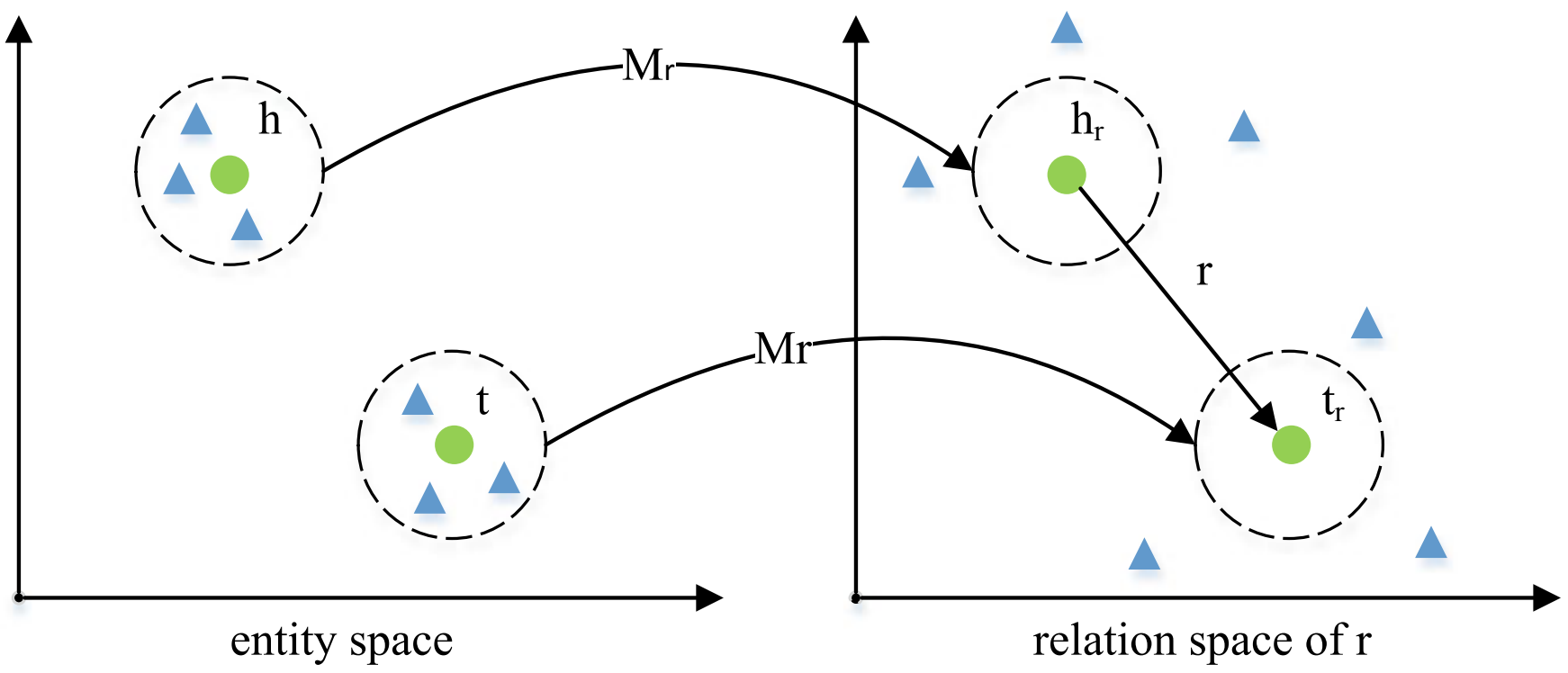

**Fig. 7.5** The architecture of TransR model [47]

#### 7.2.3.3 TransD

TransH and TransR focus on the multiple representations of entities in different relations, improving the performance on knowledge completion and triple classification. However, both models only project entities according to the relations in triples, ignoring the diversity of entities. Moreover, the projection operation with matrix-vector multiplication leads to a higher computational complexity compared to TransE, which is time consuming when applied on large-scale graphs. To address this problem, TransD [32] proposes a novel projection method with a dynamic mapping matrix depending on both entity and relation, which takes the diversity of entities as well as relations into consideration.

TransD defines two vectors for each entity and relation, i.e., the original vector that is also used in TransE, TransH, and TransR for distributed representation of entities and relations, and the projection vector that is used in constructing projection matrices for mapping entities from entity space to relation space. As illustrated in Fig. 7.6, TransD uses $\mathbf{h}$, $\mathbf{t}$, $\mathbf{r}$ to represent the original vectors, while $\mathbf{h}_p$, $\mathbf{t}_p$, and $\mathbf{r}_p$ are used to represent the projection vectors. There are two projection matrices $\mathbf{M}_{rh}, \mathbf{M}_{rt} \in \mathbb{R}^{m \times n}$ used to project from entity space to relation space, and the projection matrices are dynamically constructed as follows:

$$\mathbf{M}_{rh} = \mathbf{r}_p\mathbf{h}_p^\top + \mathbf{I}_{m\times n}, \quad \mathbf{M}_{rt} = \mathbf{r}_p\mathbf{t}_p^\top + \mathbf{I}_{m\times n}, \tag{7.11}$$

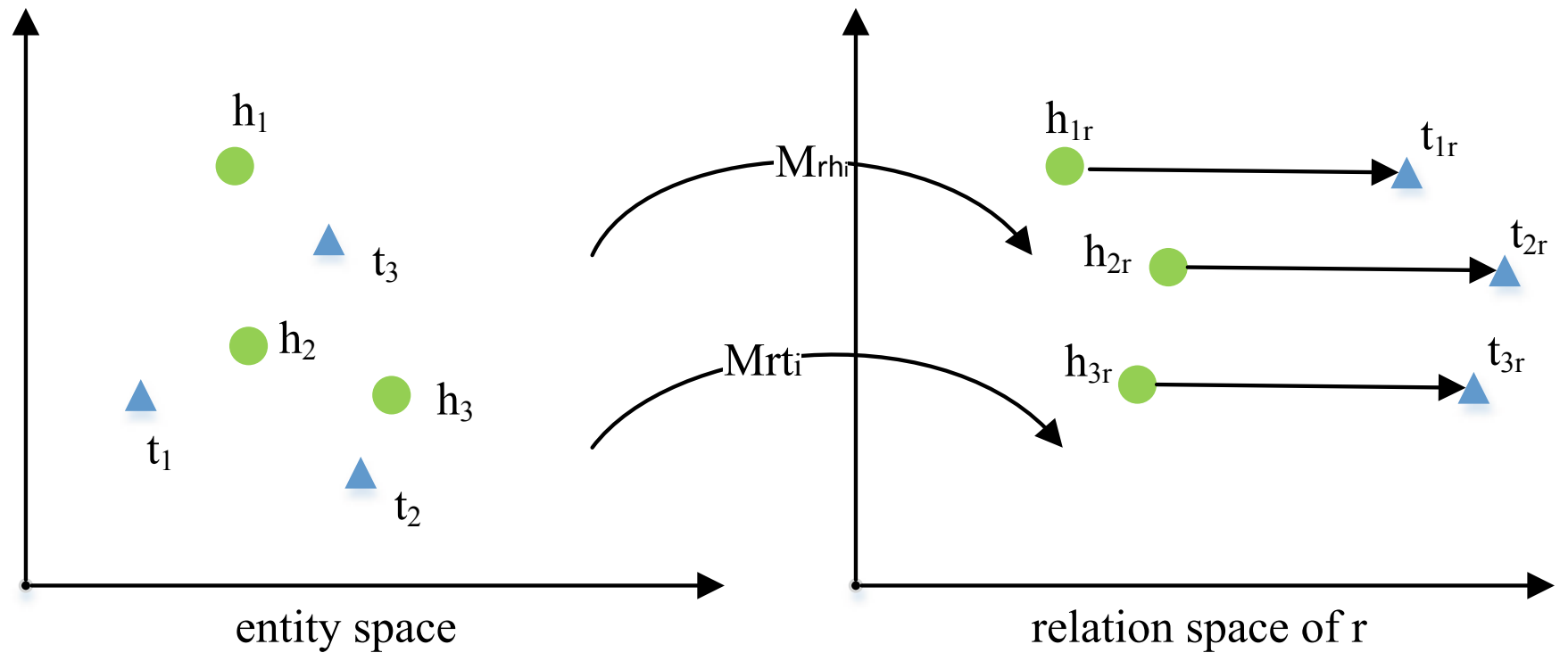

**Fig. 7.6** The architecture of TransD model [47]

which means the projection vectors of entity and relation are combined to determine the dynamic projection matrix. The score function is then defined as

$$\mathscr{E}(h, r, t) = \|\mathbf{M}_{rh}\mathbf{h} + \mathbf{r} - \mathbf{M}_{rt}\mathbf{t}\|. \tag{7.12}$$

The projection matrices are initialized with identity matrices, and there are also some normalization constraints as in TransR.

TransD proposes a dynamic method to construct projection matrices with the consideration of diversity in both entities and relations, achieving better performance compared to existing methods in link prediction and triple classification. Moreover, it lowers both computational and spatial complexity compared to TransR.

#### 7.2.3.4 TranSparse

The extension methods of TransE stated above focus on the multiple representations for entities in different relations and entity pairs. However, there are still two challenges ignored: (1) The heterogeneity. Relations in knowledge graphs differ in granularity. Some complex relations may link to many entity pairs, while some relatively simple relations not. (2) The unbalance. Some relations may have more links to head entities and fewer links to tail entities, and vice versa. The performance will be further improved if we consider these rather than merely treat all relations equally.

Existing methods like TransR build projection matrices for each relation, while these projection matrices have the same number of parameters, regardless of the variety in the complexity of relations. TranSparse [33] is then proposed to address the issues. The underlying assumption of TranSparse is that complex relations should have more parameters to learn while simple relations have fewer, where the complexity of a relation is judged from the number of triples or entities linked by the relation.

To accomplish this, two models, i.e., TranSparse(share) and TranSparse(separate), are proposed for avoiding overfitting and underfitting.

Inspired by TransR, TranSparse(share) builds a projection matrix $\mathbf{M}_r(\theta_r)$ for each relation $r$. This projection matrix is sparse, and the sparse degree $\theta_r$ mainly depends on the number of entity pairs linked to $r$. Suppose $N_r$ is the number of linked entity pairs, $N_r^*$ represents the maximum number of $N_r$, and $\theta_{min}$ denotes the minimum sparse degree of projection matrix $\mathbf{M}_r$ that $0 \leq \theta_{min} \leq 1$. The sparse degree of relation $r$ is defined as follows:

$$\theta_r = 1 - (1 - \theta_{min}) N_r / N_r^*. \tag{7.13}$$

Both head and tail entities share the same sparse projection matrix $\mathbf{M}_r(\theta_r)$ in translation. The score function is

$$\mathscr{E}(h, r, t) = \|\mathbf{M}_r(\theta_r)\mathbf{h} + \mathbf{r} - \mathbf{M}_r(\theta_r)\mathbf{t}\|. \tag{7.14}$$

Differing from TranSparse(share), TranSparse(separate) builds two different sparse matrices $\mathbf{M}_{rh}(\theta_{rh})$ and $\mathbf{M}_{rt}(\theta_{rt})$ for head and tail entities. The sparse degree $\theta_{rh}$ (or $\theta_{rt}$) then depends on the number of head (or tail) entities linked by relation $r$. We have $N_{rh}$ (or $N_{rt}$) to represent the number of head (or tail) entities, as well as $N_{rh}^*$ (or $N_{rt}^*$) to represent the maximum number of $N_{rh}$ (or $N_{rt}$). And $\theta_{min}$ will also be set as the minimum sparse degree of projection matrices that $0 \leq \theta_{min} \leq 1$. We have

$$\theta_{rh} = 1 - (1 - \theta_{min}) N_{rh} / N_{rh}^*, \quad \theta_{rt} = 1 - (1 - \theta_{min}) N_{rt} / N_{rt}^*. \tag{7.15}$$

The score function of TranSparse(separate) is

$$\mathscr{E}(h, r, t) = \|\mathbf{M}_{rh}(\theta_{rh})\mathbf{h} + \mathbf{r} - \mathbf{M}_{rt}(\theta_{rt})\mathbf{t}\|. \tag{7.16}$$

Through the sparse projection matrix, TranSparse solves the heterogeneity and the unbalance simultaneously.

#### 7.2.3.5 PTransE

The extension models of TransE stated above are mainly focused on the challenge of multiple representations of entities in different scenarios. However, those extension models only consider the simple one-step paths (i.e., relation) in translating operation, ignoring the rich global information located in the whole knowledge graphs. Considering the multistep relational path is a potential method to utilize the global information. For instance, if we notice the multistep relational path that ⟨`The forbidden city`, `locate_in`, `Beijing`⟩ → ⟨`Beijing`, `capital_of`, `China`⟩, we can inference with confidence that the triple ⟨`The forbidden city`, `locate_in`, `China`⟩ may exist. The relational path provides us with a powerful way to con-struct better knowledge graph representations and even get a better understanding of knowledge reasoning.

There are two main challenges when encoding the information in multistep relational paths. First, how to select reliable and meaningful relational paths among enormous path candidates in KGs, since there are lots of relation sequence patterns which do not indicate reasonable relationships. Let us just consider the relational path ⟨`The forbidden city`, `locate_in`, `Beijing`⟩ → ⟨`Beijing`, `held`, `2008 Summer Olympics`⟩, it is hard to describe the relationship between `The forbidden city` and `2008 Summer Olympics`. Second, how to model those meaningful relational paths once we get them since it is difficult to solve this composition semantic problem in relational paths.

PTransE [38] is then proposed to model the multistep relational paths. To select meaningful relational paths, the authors propose a Path-Constraint Resource Allocation (PCRA) algorithm to judge the relation path reliability. Suppose there is information (or resource) in head entity $h$ which will flow to tail entity $t$ through some certain relational paths. The basic assumption of PCRA is that: the reliability of path $\ell$ depends on the resource amount that finally flows from head to tail. Formally, we set $\ell = (r_1, \ldots, r_l)$ for a certain path between $h$ and $t$. The resource travels from $h$ to $t$ and the path could be represented as $S_0/h \xrightarrow{r_1} S_1 \xrightarrow{r_2} \ldots \xrightarrow{r_l} S_l/t$. For an entity $m \in S_i$, the resource amount of $m$ is defined as follows:

$$R_\ell(m) = \sum_{n \in S_{i-1}(\cdot,m)} \frac{1}{|S_i(n,\cdot)|} R_\ell(n), \tag{7.17}$$

where $S_{i-1}(\cdot, m)$ indicates all direct predecessors of entity $m$ along with relation $r_i$ in $S_{i-1}$, and $S_i(n, \cdot)$ indicates all direct successors of $n \in S_{i-1}$ with relation $r$. Finally, the resource amount of tail $R_\ell(t)$ is used to measure the reliability of $\ell$ in the given triple $\langle h, \ell, t\rangle$.

Once we have learned the reliability and select those meaningful relational path candidates, the next challenge is how to model the meaning of those multistep paths. PTransE proposes three types of composition operation, namely, Addition, Multiplication, and recurrent neural networks, to get the representation $\mathbf{l}$ of $\ell = (r_1, \ldots, r_l)$ through those relations. The score function of the path triple $\langle h, \ell, t\rangle$ is defined as follows:

$$\mathscr{E}(h, \ell, t) = \|\mathbf{l} - (\mathbf{t} - \mathbf{h})\| \approx \|\mathbf{l} - \mathbf{r}\| = \mathscr{E}(\ell, r), \tag{7.18}$$

where $r$ indicates the golden relation between $h$ and $t$. Since PTransE wants to meet the assumption in TransE that $\mathbf{r} \approx \mathbf{t} - \mathbf{h}$ simultaneously, PTransE directly utilizes $\mathbf{r}$ in training. The optimization objective of PTransE is

$$\mathscr{L} = \sum_{(h,r,t)\in S} [\mathscr{L}(h, r, t) + \frac{1}{Z} \sum_{\ell \in P(h,t)} R(\ell|h, t)\mathscr{L}(\ell, r)], \tag{7.19}$$

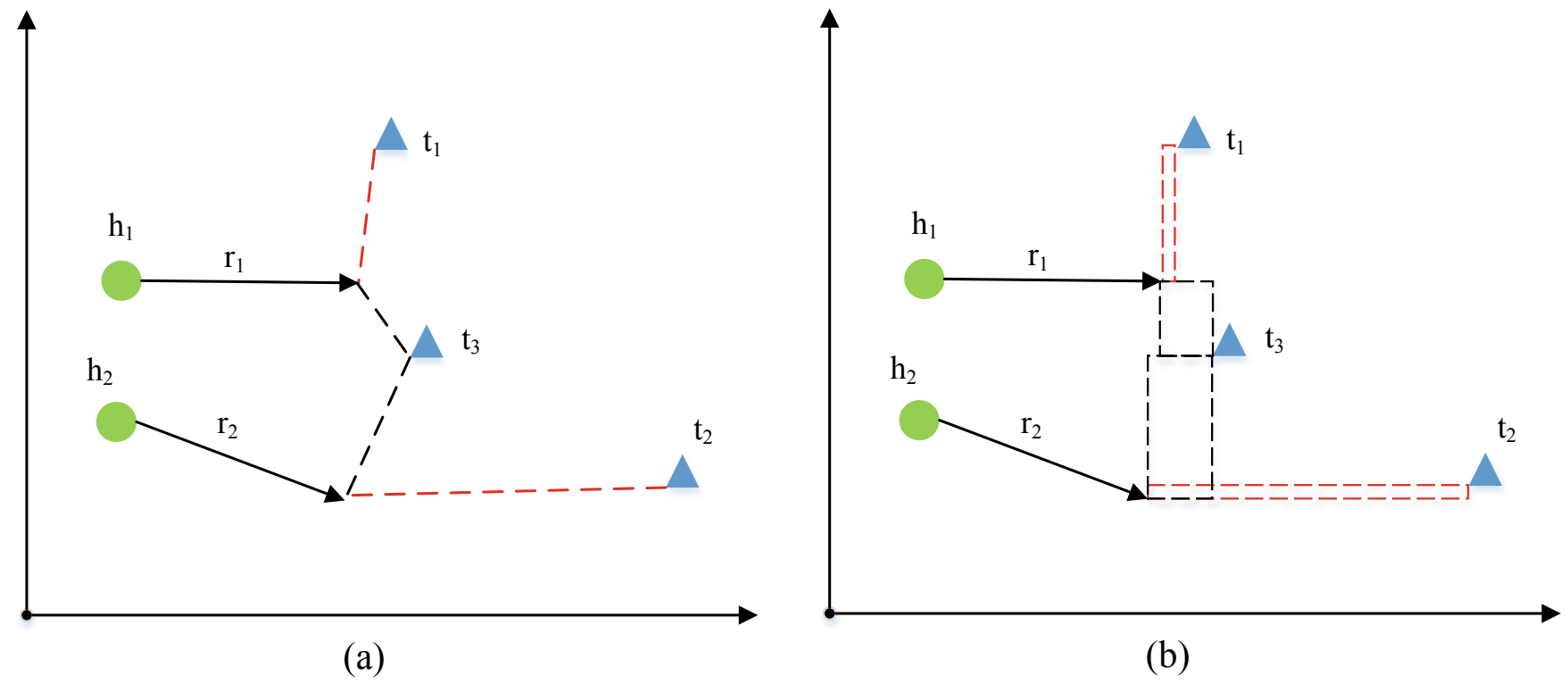

**Fig. 7.7** The architecture of TransA model [47]

where $\mathscr{L}(h, r, t)$ is the margin-based score function with $\mathscr{E}(h, r, t)$ and $\mathscr{L}(\ell, r)$ is the margin-based score function with $\mathscr{E}(\ell, r)$. The reliability $R(\ell|h, t)$ of $\ell$ in $(h, \ell, t)$ is well considered in the overall loss function.

Besides PTransE, similar ideas such as [21, 22] also consider the multistep relational paths on different tasks such as knowledge completion and question answering successfully. These works demonstrate that there is plentiful information located in multi-step relational paths, which could significantly improve the performance of knowledge graph representation, and further explorations on more sophisticated models for relational paths are still promising.

#### 7.2.3.6 TransA

TransA [78] is proposed to solve the following problems in TransE and other extensions: (1) TransE and its extensions only consider the Euclidean distance in their energy functions, which seems to be less flexible. (2) Existing methods regard each dimension in the semantic vector space identically whatever the triple is, which may bring in errors when calculating dissimilarities. To solve these problems, as illustrated in Fig. 7.7, TransA replaces the inflexible Euclidean distance with adaptive Mahalanobis distance, which is more adaptive and flexible. The energy function of TransA is as follows:

$$\mathscr{E}(h, r, t) = (|\mathbf{h} + \mathbf{r} - \mathbf{t}|)^\top \mathbf{W}_r (|\mathbf{h} + \mathbf{r} - \mathbf{t}|), \tag{7.20}$$

where $\mathbf{W}_r$ is a relation-specific nonnegative symmetric matrix corresponding to the adaptive matric. Note that the $|\mathbf{h} + \mathbf{r} - \mathbf{t}|$ stands for a nonnegative vector that each dimension is the absolute value of the translating operation. We have

$$(|\mathbf{h} + \mathbf{r} - \mathbf{t}|) \triangleq (|h_1 + r_1 - t_1|, |h_2 + r_2 - t_2|, \ldots |h_n + r_n - t_n|). \tag{7.21}$$

#### 7.2.3.7 KG2E

Existing translation-based models usually consider entities and relations as vectors embedded in low-dimensional semantic spaces. However, as explained above, entities and relations in KGs are various with different granularities. Therefore, the margin in the margin-based score function that is used to distinguish positive triples from negative triples should be more flexible due to the diversity, and the uncertainties of entities and relations should be taken into consideration.

To solve this, KG2E [30] is proposed, introducing the multidimensional Gaussian distributions to KG representations. As illustrated in Fig. 7.8, KG2E represents each entity and relation with a Gaussian distribution. Specifically, the mean vector denotes the entity/relation's central position, and the covariance matrix denotes its uncertainties. To learn the Gaussian distributions for entities and relations, KG2E also follows the score function proposed in TransE. For a triple $\langle h, r, t\rangle$, the Gaussian distributions of entity and relation are defined as follows:

$$\mathbf{h} \sim \mathcal{N}(\boldsymbol{\mu}_h, \boldsymbol{\Sigma}_h), \quad \mathbf{t} \sim \mathcal{N}(\boldsymbol{\mu}_t, \boldsymbol{\Sigma}_t), \quad \mathbf{r} \sim \mathcal{N}(\boldsymbol{\mu}_r, \boldsymbol{\Sigma}_r). \tag{7.22}$$

Note that the covariances are diagonal for the consideration of efficiency. KG2E hypothesizes that the head and tail entity are independent with specific relations, then the translation $h - t$ could be defined as

$$\mathbf{h} - \mathbf{t} = \mathbf{e} \sim \mathcal{N}(\boldsymbol{\mu}_h - \boldsymbol{\mu}_t, \boldsymbol{\Sigma}_h + \boldsymbol{\Sigma}_t). \tag{7.23}$$

To measure the dissimilarity between **e** and **r**, KG2E proposes two methods considering both asymmetric similarity and symmetric similarity.

The asymmetric similarity is based on the KL divergence between **e** and **r**, which is a straightforward method to measure the similarity between two probability distributions. The energy function is as follows:

$$\begin{aligned}\mathcal{E}(h, r, t) &= D_{\mathrm{KL}}(\mathbf{e}\|\mathbf{r})\\ &= \int_{x \in R^{k_e}} \mathcal{N}(x; \boldsymbol{\mu}_r, \boldsymbol{\Sigma}_r) \log \frac{\mathcal{N}(x; \boldsymbol{\mu}_e, \boldsymbol{\Sigma}_e)}{\mathcal{N}(x; \boldsymbol{\mu}_r, \boldsymbol{\Sigma}_r)} dx\\ &= \frac{1}{2}\left\{\mathrm{tr}(\boldsymbol{\Sigma}_r^{-1}\boldsymbol{\Sigma}_r) + (\boldsymbol{\mu}_r - \boldsymbol{\mu}_e)^\top \boldsymbol{\Sigma}_r^{-1}(\boldsymbol{\mu}_r - \boldsymbol{\mu}_e) - \log \frac{\det(\boldsymbol{\Sigma}_e)}{\det(\boldsymbol{\Sigma}_r)} - k_e\right\},\end{aligned} \tag{7.24}$$

where $\mathrm{tr}(\boldsymbol{\Sigma})$ indicates the trace of $\boldsymbol{\Sigma}$, and $\boldsymbol{\Sigma}^{-1}$ indicates the inverse.

The symmetric similarity is based on the expected likelihood or probability product kernel. KE2G takes the inner product between $P_e$ and $P_r$ as the measurement of similarity. The logarithm of energy function is

$$\begin{aligned}\mathscr{E}(h, r, t) &= \int_{x \in R^{k_e}} \mathscr{N}(x; \boldsymbol{\mu}_e, \boldsymbol{\Sigma}_e)\mathscr{N}(x; \boldsymbol{\mu}_r, \boldsymbol{\Sigma}_r)dx \\ &= \log \mathscr{N}(0; \boldsymbol{\mu}_e - \boldsymbol{\mu}_r, \boldsymbol{\Sigma}_e + \boldsymbol{\Sigma}_r) \\ &= \frac{1}{2}\left\{(\boldsymbol{\mu}_e - \boldsymbol{\mu}_r)^\top(\boldsymbol{\Sigma}_e + \boldsymbol{\Sigma}_r)^{-1}(\boldsymbol{\mu}_e - \boldsymbol{\mu}_r) + \log\det(\boldsymbol{\Sigma}_e + \boldsymbol{\Sigma}_r) + k_e \log(2\pi)\right\}.\end{aligned} \tag{7.25}$$

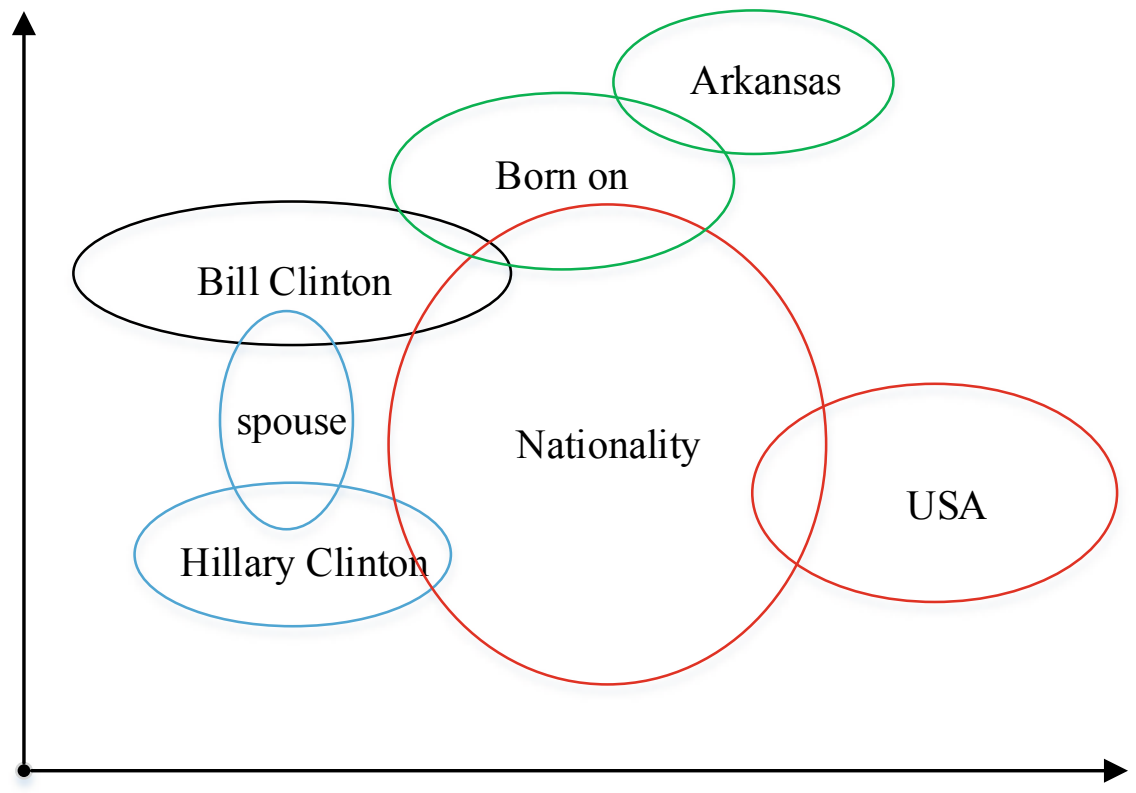

**Fig. 7.8** The architecture of KG2E model [47]

The optimization objective of KG2E is also margin-based similar to TransE. Both asymmetric and symmetric similarities are constrained by some regularization to avoid overfitting:

$$\forall l \in E \cup R, \quad \|\boldsymbol{\mu}_l\|_2 \le 1, \quad c_{min}\mathbf{I} \le \boldsymbol{\Sigma}_l \le c_{max}\mathbf{I}, \quad c_{min} > 0. \tag{7.26}$$

Figure 7.8 shows a brief example of representations in KG2E.

#### 7.2.3.8 TransG

We have discussed the problem of TransE in the session of TransR/CTransR that some relations in knowledge graphs such as `location_contains` or `has_part` may have multiple sub-meanings. These relations are more likely to be some combinations that could be divided into several more precise relations. To address this issue, CTransR is proposed with a preprocess of clustering for each relation $r$ depending on the entity pairs $(h, t)$. TransG [79] also focuses on this issue more elegantly by introducing a generative model. As illustrated in Fig. 7.9, it assumes that different semantic component embeddings should follow a Gaussian Mixture Model. The generative process is as follows:

1. For each entity $e \in E$, TransG sets a standard normal distribution: $\boldsymbol{\mu}_e \sim \mathscr{N}(\mathbf{0}, \mathbf{I})$.
2. For a triple $\langle h, r, t\rangle$, TransG uses Chinese Restaurant Process to automatically detect semantic components (i.e., sub-meanings in a relation): $\pi_{r,n} \sim \text{CRP}(\beta)$.

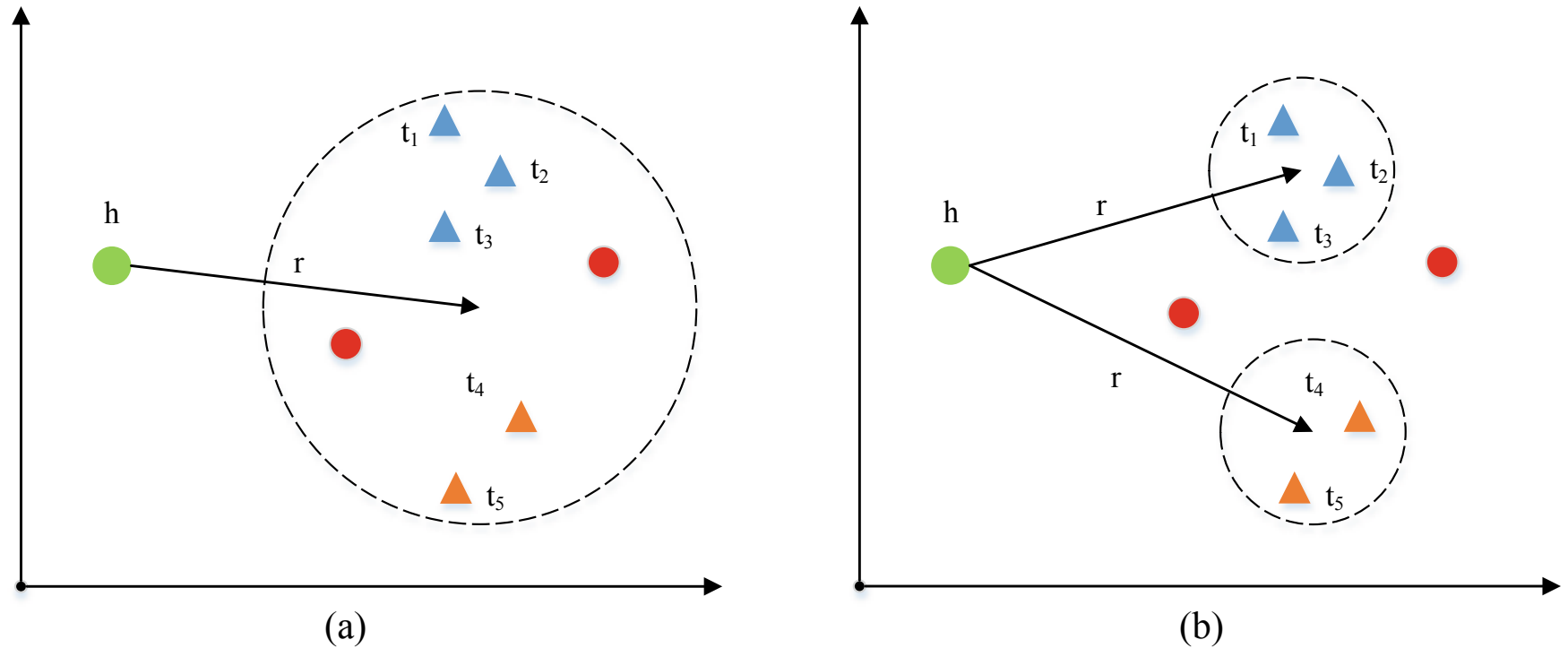

**Fig. 7.9** The architecture of TransG model [47]

3. Draw the head embedding to form a standard normal distribution: $\mathbf{h} \sim \mathcal{N}(\boldsymbol{\mu}_h, \sigma_h^2 \mathbf{I})$.
4. Draw the tail embedding to form a standard normal distribution: $\mathbf{t} \sim \mathcal{N}(\boldsymbol{\mu}_t, \sigma_t^2 \mathbf{I})$.
5. Draw the relation embedding for this semantic component: $\boldsymbol{\mu}_{r,n} = \mathbf{t} - \mathbf{h} \sim \mathcal{N}(\boldsymbol{\mu}_t - \boldsymbol{\mu}_h, (\sigma_h^2 + \sigma_t^2)\mathbf{I})$.

$\boldsymbol{\mu}$ is the mean embedding and $\sigma$ is the variance. Finally, the score function is

$$\mathcal{E}(h, r, t) \propto \sum_{n=1}^{N_r} \pi_{r,n} \mathcal{N}(\boldsymbol{\mu}_t - \boldsymbol{\mu}_h, (\sigma_h^2 + \sigma_t^2)\mathbf{I}), \tag{7.27}$$

in which $N_r$ is the number of semantic components of $r$, and $\pi_{r,n}$ is the weight of $i$th component generated by the Chinese Restaurant Process.

Figure 7.9 shows the advantages of the generative Gaussian Mixture Model.

#### 7.2.3.9 ManifoldE

KG2E and TransG introduce Gaussian distributions to knowledge graph representation learning, improving the flexibility and diversity with the various forms of entity and relation representation. However, TransE and its most extensions view the golden triples as almost points in the low-dimensional vector space, following the assumption of translation. This point assumption may lead to two problems: being an ill-posed algebraic system and being over-strict with the geometric form.

ManifoldE [80] is proposed to address this issue, considering the possible position of the golden candidate in vector space as a manifold instead of one point. The overall score function of ManifoldE is defined as follows:

$$\mathcal{E}(h, r, t) = \|\mathcal{M}(h, r, t) - D_r^2\|^2, \tag{7.28}$$

in which $D_r^2$ is a relation-specific manifold parameter indicating the bias. Two kinds of manifolds are then proposed in ManifoldE. ManifoldE(Sphere) is a straightforward manifold that supposes $\mathbf{t}$ should be located in the sphere which has $\mathbf{h} + \mathbf{r}$ to be the center and $D_r$ to be the radius. We have

$$\mathscr{M}(h, r, t) = \|\mathbf{h} + \mathbf{r} - \mathbf{t}\|_2^2. \tag{7.29}$$

The second manifold utilized is the hyperplane for it is much easier for two hyperplanes to intersect. The function of ManifoldE(Hyperplane) is

$$\mathscr{M}(h, r, t) = (\mathbf{h} + \mathbf{r}_h)^\top(\mathbf{t} + \mathbf{r}_t), \tag{7.30}$$

in which $r_h$ and $r_t$ represent the two relation embeddings. This indicates that for a triple $\langle h, r, t\rangle$, the tail entity $\mathbf{t}$ should locate in the hyperplane whose direction is $\mathbf{h} + \mathbf{r}_h$ with the bias to be $D_r^2$. Furthermore, ManifoldE(Hyperplane) considers absolute values in $\mathscr{M}(h, r, t)$ as $|\mathbf{h} + \mathbf{r}_h|^\top|\mathbf{t} + \mathbf{r}_t|$ to double the solution number of possible tails. For both manifolds, the author applies kernel forms on Reproducing Kernel Hilbert Space.

## 7.2.4 *Other Models*

Translation-based methods such as TransE are simple but effective, whose power has been consistently verified on various tasks like knowledge graph completion and triple classification, achieving state-of-the-art performance. However, there are also some other representation learning methods performing well on knowledge graph representation. In this part, we will take a brief look at these methods as inspiration.

### 7.2.4.1 Structured Embeddings

Structured Embeddings (SE) [8] is a classical representation learning method for KGs. In SE, each entity is projected to a $d$-dimensional vector space. SE designs two relation-specific matrices $\mathbf{M}_{r,1}, \mathbf{M}_{r,2} \in \mathbb{R}^{d\times d}$ for each relation $r$, projecting both head and tail entities with these relation-specific matrices when calculating the similarities. The score function of SE is defined as follows:

$$\mathscr{E}(h, r, t) = \|\mathbf{M}_{r,1}\mathbf{h} - \mathbf{M}_{r,2}\mathbf{t}\|_1, \tag{7.31}$$

in which both $\mathbf{h}$ and $\mathbf{t}$ are transformed into a relation-specific vector space with those projection matrices. The assumption of SE is that the projected head and tail embeddings should be as similar as possible according to the loss function. Different from the translation-based methods, SE models entities as embeddings and relations as projection matrices. In training, SE considers all triples in the training set and minimizes the overall loss function.

#### 7.2.4.2 Semantic Matching Energy

Semantic Matching Energy (SME) [5, 6] proposes a more complicated representation learning method. Differing from SE, SME considers both entities and relations as low-dimensional vectors. For a triple $\langle h, r, t\rangle$, $\mathbf{h}$ and $\mathbf{r}$ are combined using a projection function $g$ to get a new embedding $\mathbf{l}_{h,r}$, and the same with $\mathbf{t}$ and $\mathbf{r}$ to get $\mathbf{l}_{t,r}$. Next, a point-wise multiplication function is used on the two combined embeddings $\mathbf{l}_{h,r}$ and $\mathbf{l}_{t,r}$ to get the score of this triple. SME proposes two different projection functions in the second step, among which the linear form is

$$\mathscr{E}(h, r, t) = (\mathbf{M}_1\mathbf{h} + \mathbf{M}_2\mathbf{r} + \mathbf{b}_1)^\top(\mathbf{M}_3\mathbf{t} + \mathbf{M}_4\mathbf{r} + \mathbf{b}_2), \tag{7.32}$$

and the bilinear form is:

$$\mathscr{E}(h, r, t) = ((\mathbf{M}_1\mathbf{h} \odot \mathbf{M}_2\mathbf{r}) + \mathbf{b}_1)^\top((\mathbf{M}_3\mathbf{t} \odot \mathbf{M}_4\mathbf{r}) + \mathbf{b}_2), \tag{7.33}$$

where $\odot$ is the element-wise (Hadamard) product. $\mathbf{M}_1$, $\mathbf{M}_2$, $\mathbf{M}_3$, $\mathbf{M}_4$ are weight matrices in the projection function, and $\mathbf{b}_1$ and $\mathbf{b}_2$ are the bias. Bordes et al. [6] is based on SME and improves the bilinear form with three-way tensors instead of matrices.

#### 7.2.4.3 Latent Factor Model

Latent Factor Model (LFM) is proposed for modeling large multi-relational datasets. LFM is based on a bilinear structure, which models entities as embeddings and relations as matrices. It could share sparse latent factors among different relations, significantly reducing the model and computational complexity. The score function of LFM is defined as follows:

$$\mathscr{E}(h, r, t) = \mathbf{h}^\top\mathbf{M}_r\mathbf{t}, \tag{7.34}$$

in which $\mathbf{M}_r$ is the representation of the relation $r$. Moreover, [92] proposes DIST-MULT model, which restricts $\mathbf{M}_r$ to be a diagonal matrix. This enhanced model not only reduces the parameter number of LFM and thus lowers the model's computational complexity, but also achieves better performance.

#### 7.2.4.4 RESCAL

RESCAL is a knowledge graph representation learning method based on matrix factorization [54, 55]. In RESCAL, to represent all triple facts in knowledge graphs, the authors employ a three-way tensor $\overrightarrow{\mathbf{X}} \in \mathbb{R}^{d\times d\times k}$ in which $d$ is the dimension of entities and $k$ is that of relations. In the three-way tensor $\overrightarrow{\mathbf{X}}$, two modes stand for the head and tail entities while the third mode represents the relations. The entries of $\overrightarrow{\mathbf{X}}$ are based on the existence of the corresponding triple facts. That is, $\overrightarrow{\mathbf{X}}_{ijm} = 1$ if the triple $\langle ith\ entity, mth\ relation, jth\ entity\rangle$ holds in the training set, and otherwise $\overrightarrow{\mathbf{X}}_{ijm} = 0$ if the triple is nonexisting.

To capture the inherent structure of all triples, a tensor factorization model named RESCAL is then proposed. Suppose $\overrightarrow{\mathbf{X}} = \{\mathbf{X}_1, \dots, \mathbf{X}_k\}$, for each slice $\mathbf{X}_n$, we have the following rank-$r$ factorization:

$$\mathbf{X}_n \approx \mathbf{A}\mathbf{R}_n\mathbf{A}^\top, \tag{7.35}$$

where $\mathbf{A} \in \mathbb{R}^{d\times r}$ stands for the $r$-dimensional entity representations, and $\mathbf{R}_n \in \mathbb{R}^{r\times r}$ represents the interactions of the $r$ latent components for $n$-th relation. The assumption in this factorization is similar to LFM, while RESCAL also optimizes the nonexisting triples where $\overrightarrow{\mathbf{X}}_{ijm} = 0$ instead of only considering the positive instances.

Following this tensor factorization assumption, the loss function of RESCAL is defined as follows:

$$\mathscr{L} = \frac{1}{2}\left(\sum_n \|\mathbf{X}_n - \mathbf{A}\mathbf{R}_n\mathbf{A}^\top\|_F^2\right) + \frac{1}{2}\lambda\left(\|\mathbf{A}\|_F^2 + \sum_n \|\mathbf{R}_n\|_F^2\right), \tag{7.36}$$

in which the second term is a regularization term and $\lambda$ is a hyperparameter.

#### 7.2.4.5 HOLE

RESCAL works well with multi-relational data but suffers from high computational complexity. To leverage both effectiveness and efficiency, Holographic Embeddings (HOLE) is proposed as an enhanced version of RESCAL [53].

HOLE employs an operation named circular correlation to generate compositional representations, which is similar to those holographic models of associative memory. The circular correlation operation $\star : \mathbb{R}^d \times \mathbb{R}^d \to \mathbb{R}^d$ between two entities $h$ and $t$ is as follows:

$$\mathbf{h} \star \mathbf{t}t_k = \sum_{i=0}^{d-1} h_i t_{(k+i) mod\ d}. \tag{7.37}$$

Figure 7.10a also demonstrates a simple instance of this operation. The probability of a triple $\langle h, r, t\rangle$ is then defined as

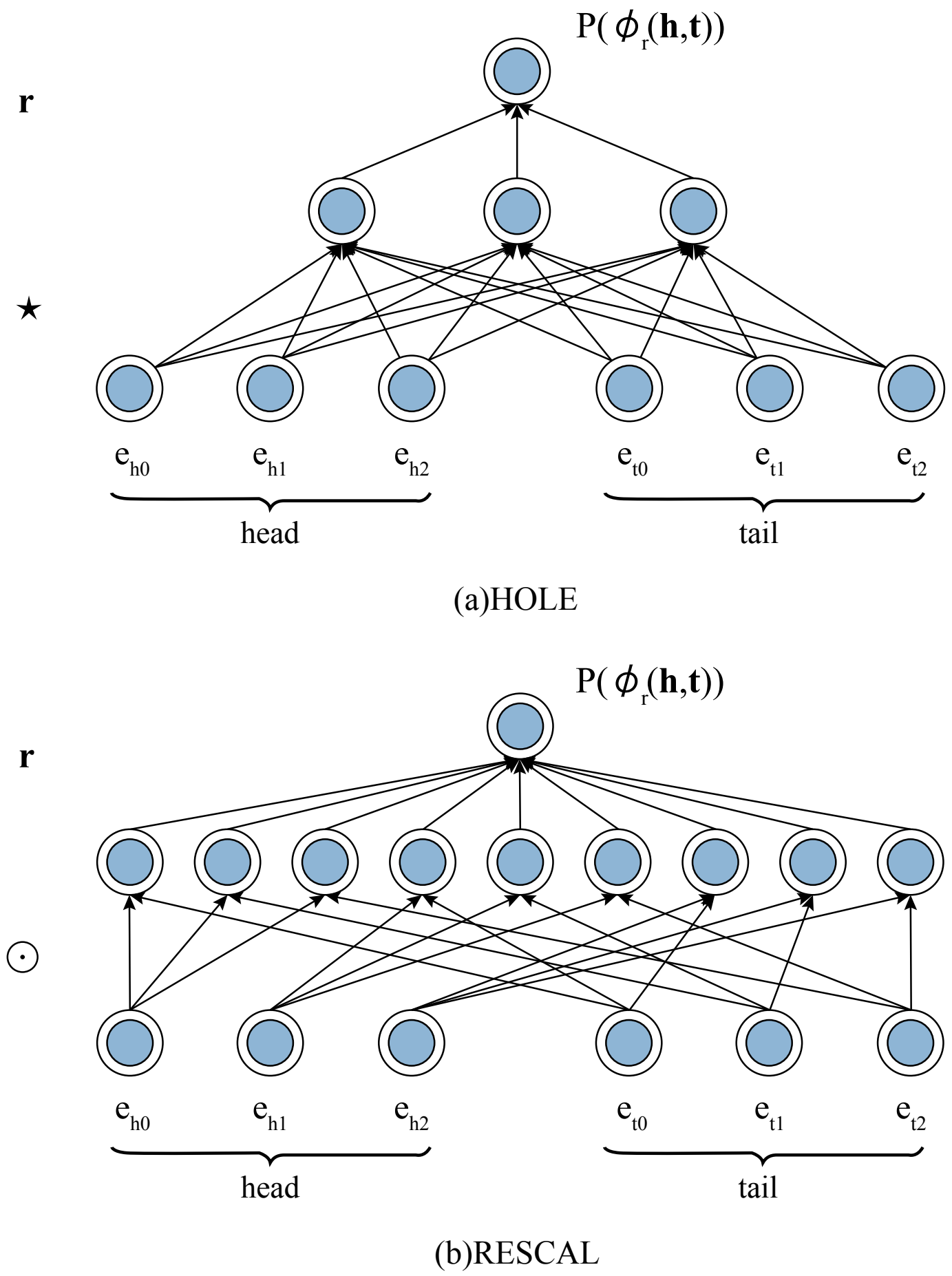

**Fig. 7.10** The architecture of RESCAL and HOLE models

$$P(\phi_r(h,t)=1)=\text{Sigmoid}(\mathbf{r}^\top(\mathbf{h}\star\mathbf{t})). \tag{7.38}$$

Considering circular correlation brings in lots of advantages: (1) unlike other operations like multiplication or convolution, circular correlation is noncommutative (i.e., $\mathbf{h}\star\mathbf{t}\neq\mathbf{t}\star\mathbf{h}$), which is capable of modeling asymmetric relations in knowledge graphs. (2) Circular correlation has lower computational complexity compared to tensor product in RESCAL. What's more, the circular correlation could further speed up with the help of Fast Fourier Transform (FFT), which is formalized as follows:

$$\mathbf{h}\star\mathbf{t}=\mathscr{F}^{-1}(\overline{\mathscr{F}(\mathbf{h})}\odot\mathscr{F}(\mathbf{b})). \tag{7.39}$$

$\mathscr{F}(\cdot)$ and $\mathscr{F}(\cdot)^{-1}$ represent the FFT and its inverse, while $\overline{\mathscr{F}(\cdot)}$ denotes the complex conjugate in $\mathbb{C}^d$, and $\odot$ stands for the element-wise (Hadamard) product. Due to FFT, the computational complexity of circular correlation is $O(d \log d)$, which is much lower than that of tensor product.

#### 7.2.4.6 Complex Embedding (ComplEx)

ComplEx [70] employs an eigenvalue decomposition model, which makes use of complex valued embeddings. The composition of complex embeddings can handle a large variety of binary relations, among the symmetric and antisymmetric relations. Formally, the log-odd of the probability that the fact $\langle h, r, t \rangle$ is true is

$$f_r(h, t) = \text{Sigmoid}(X_{hrt}), \tag{7.40}$$

where $f_r(h, t)$ is expected to be 1 when $(h, r, t)$ holds, otherwise $-1$. Here, $X_{hrt}$ is calculated as follows:

$$\begin{aligned} X_{hrt} &= \text{Re}(\langle \mathbf{r}, \mathbf{h}, \mathbf{t} \rangle) \\ &= \langle \text{Re}(\mathbf{r}), \text{Re}(\mathbf{h}), \text{Re}(\mathbf{t}) \rangle + \langle \text{Re}(\mathbf{r}), \text{Im}(\mathbf{h}), \text{Im}(\mathbf{t}) \rangle \\ &\quad - \langle \text{Im}(\mathbf{r}), \text{Re}(\mathbf{h}), \text{Im}(\mathbf{t}) \rangle - \langle \text{Im}(\mathbf{r}), \text{Im}(\mathbf{h}), \text{Re}(\mathbf{t}) \rangle, \end{aligned} \tag{7.41}$$

where $\langle \mathbf{x}, \mathbf{y}, \mathbf{z} \rangle = \sum_i x_i y_i z_i$ denotes the trilinear dot product, $\text{Re}(x)$ and $\text{Im}(x)$ indicate the real part and the imaginary part of the number $x$ respectively. In fact, ComplEx can be viewed as an extension of RESCAL, which assigns complex embedding of the entities and relations.

Besides, [29] has proved that HolE is mathematically equivalent to ComplEx recently.

#### 7.2.4.7 Convolutional 2D Embeddings (ConvE)

ConvE [16] uses 2D convolution over embeddings and multiple layers of nonlinear features to model knowledge graphs. It is the first nonlinear model that significantly outperforms previous linear models.

Specifically, ConvE uses convolutional and fully connected layers to model the interactions between input entities and relationships. After that, the obtained features are flattened, transformed through a fully connected layer, and the inner product is taken with all object entity vectors to generate a score for each triple.

For each triple $\langle h, r, t \rangle$, ConvE defines its score function as

$$f_r(h, t) = f(\text{vec}(f([\bar{\mathbf{h}}; \bar{\mathbf{r}}] * \omega))\mathbf{W})\mathbf{t}, \tag{7.42}$$

where $*$ denotes the convolution operator, and $\text{vec}(\cdot)$ means compressing a matrix into a vector. $\mathbf{r} \in \mathbb{R}^k$ is a relation parameter depending on $r$, $\bar{\mathbf{h}}$ and $\bar{\mathbf{r}}$ denote a 2D reshaping of $\mathbf{h}$ and $\mathbf{r}$, respectively: if $\mathbf{h}, \mathbf{r} \in \mathbb{R}^k$, then $\bar{\mathbf{h}}, \bar{\mathbf{r}} \in \mathbb{R}^{k_a \times k_b}$, where $k = k_a k_b$.

ConvE can be seen as an improvement on HolE. Compared with HolE, it learns multiple layers of nonlinear features, and thus theoretically more expressive than HolE.

#### 7.2.4.8 Rotation Embeddings (RotatE)

RotatE [67] defines each relation as a rotation from the head entity to the tail entity in the complex vector space. Thus, it is able to model and infer various relation patterns, including symmetry/antisymmetry, inversion, and composition. Formally, the score function of the fact $\langle h, r, t \rangle$ of RotatE is defined as

$$f_r(h, t) = \|\mathbf{h} \odot \mathbf{r} - \mathbf{t}\|, \tag{7.43}$$

where $\odot$ denotes the element-wise (Hadamard) product, $\mathbf{h}, \mathbf{r}, \mathbf{t} \in \mathbb{C}^k$ and $|r_i| = 1$.

RotatE is simple but achieves quite good performance. Compared with previous work, it is the first model that is capable of modeling and inferring all the three relation patterns above.

#### 7.2.4.9 Neural Tensor Network

Socher et al. [65] propose Neural Tensor Network (NTN) as well as Single Layer Model (SLM), while NTN is an enhanced version of SLM. Inspired by the previous attempts in KRL, SLM represents both entities and relations as low-dimensional vectors, and also designs relation-specific projection matrices to map entities from entity space to relation space. Similar to SE, the score function of SLM is as follows:

$$\mathscr{E}(h, r, t) = \mathbf{r}^\top \tanh(\mathbf{M}_{r,1}\mathbf{h} + \mathbf{M}_{r,2}\mathbf{t}), \tag{7.44}$$

where $\mathbf{h}$, $\mathbf{t} \in \mathbb{R}^d$ represent head and tail embeddings, $\mathbf{r} \in \mathbb{R}^k$ represents relation embedding, and $\mathbf{M}_{r,1}, \mathbf{M}_{r,2} \in \mathbb{R}^{d \times k}$ stand for the relation-specific matrices.

Although SLM has introduced relation embeddings as well as a nonlinear layer into the score function, the model representation capability is still restricted. Neural tensor network is then proposed with tensors being introduced into the SLM framework. Besides the original linear neural network layer that projects entities to the relation space, NTN also adds another tensor-based neural layer which combines head and tail embeddings with a relation-specific tensor, as illustrated in Fig. 7.11. The score function of NTN is then defined as follows:

$$\mathscr{E}(h, r, t) = \mathbf{r}^\top \tanh(\mathbf{h}^\top \overrightarrow{\mathbf{M}}_r \mathbf{t} + \mathbf{M}_{r,1}\mathbf{h} + \mathbf{M}_{r,2}\mathbf{t} + \mathbf{b}_r), \tag{7.45}$$

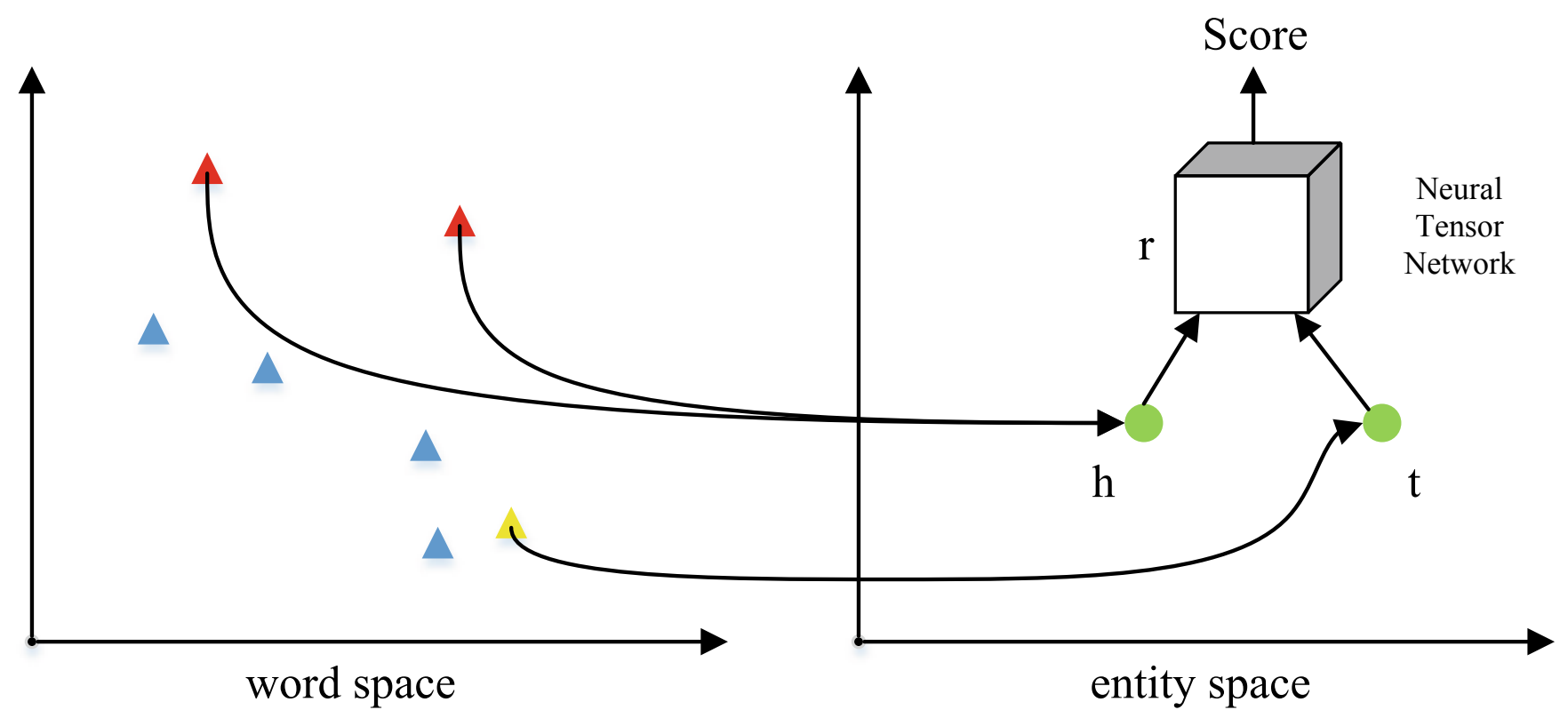

**Fig. 7.11** The architecture of NTN model [47]

where $\overrightarrow{\mathbf{M}}_r \in \mathbb{R}^{d\times d\times k}$ is a 3-way relation-specific tensor, $\mathbf{b}_r$ is the bias, and $\mathbf{M}_{r,1}$, $\mathbf{M}_{r,2} \in \mathbb{R}^{d\times k}$ is the relation-specific matrices similar to SLM. Note that SLM is the simplified version of NTN if the tensor and bias are set to zero.

Besides the improvements in score function, NTN also attempts to utilize the latent textual information located in entity names and successfully achieves significant improvements. Differing from previous RL models that provide each entity with a vector, NTN represents each entity as the average of its entity name's word embeddings. For example, the entity `Bengal tiger` will be represented as the average word embeddings of `Bengal` and `tiger`. It is apparent that the entity name will provide valuable information for understanding an entity, since `Bengal tiger` may come from Bengal and be related to other tigers. Moreover, the number of words is far less than that of entities. Therefore, using the average word embeddings of entity names will also lower the computational complexity and alleviate the issue of data sparsity.

NTN utilizes tensor-based neural networks to model triple facts and achieves excellent successes. However, the overcomplicated method will lead to higher computational complexity compared to other methods, and the vast number of parameters will limit the performance on rather sparse and large-scale KGs.

#### 7.2.4.10 Neural Association Model (NAM)

NAM [43] adopts multilayer nonlinear activations in the deep neural network to model the conditional probabilities between head and tail entities. NAM studies two model structures Deep Neural Network (DNN) and Relation Modulated Neural Network (RMNN).

NAM-DNN feeds the head and tail entities' embeddings into an MLP with $L$ fully connected layers, which is formalized as follows:

$$\mathbf{z}^{(l)} = \text{Sigmoid}(\mathbf{M}^{l}\mathbf{z}^{(l-1)} + b^{(l)}), \;\; l = 1, \ldots, L, \tag{7.46}$$

where $\mathbf{z}^{(0)} = [\mathbf{h}; \mathbf{r}]$, $\mathbf{M}^{(l)}$ and $\mathbf{b}^{(l)}$ is the weight matrix and bias vector for the $l$-th fully connected layer, respectively. And finally the score function of NAM-DNN is defined as

$$f_r(h, t) = \text{Sigmoid}(\mathbf{t}^{\top}\mathbf{z}^{(L)}). \tag{7.47}$$

Different from NAM-DNN, NAM-RMNN feds the relation embedding $\mathbf{r}$ into each layer of the deep neural network as follows:

$$\mathbf{z}^{(l)} = \text{Sigmoid}(\mathbf{M}^{(l)}\mathbf{z}^{(l-1)} + \mathbf{B}^{(l)}\mathbf{r}), \;\; l = 1, \ldots, L, \tag{7.48}$$

where $\mathbf{z}^{(0)} = [\mathbf{h}; \mathbf{r}]$, $\mathbf{M}^{(l)}$ and $\mathbf{B}^{(l)}$ indicate the weight matrices. The score function of NAM-RMNN is defined as

$$f_r(h, t) = \text{Sigmoid}(\mathbf{t}^{\top}\mathbf{z}^{(L)} + \mathbf{B}^{(l+1)}\mathbf{r}). \tag{7.49}$$

## 7.3 Multisource Knowledge Graph Representation

We are living in a complicated pluralistic real world, in which we can get information through all senses and learn knowledge not only from structured knowledge graphs but also from plain texts, categories, images, and videos. This cross-modal information is considered as multisource information. Besides the structured knowledge graph which is well utilized in previous KRL methods, we will introduce the other kinds of KRL methods utilizing multisource information:

1. **Plain text** is one of the most common information we deliver, receive, and analyze every day. There are vast amounts of plain texts we possess remaining to be detected, in which the significant knowledge that structured knowledge graphs may not include locates. **Entity description** is a special kind of textual information that describes the corresponding entity within a few sentences or a short paragraph. Usually, entity descriptions are maintained by some knowledge graphs (i.e., Freebase) or could be automatically extracted from huge databases like Wikipedia.

2. **Entity type** is another important structured information for building knowledge representations. To learn new objects within our prior knowledge systems, human beings tend to systemize those objects into existing categories. An entity type is usually represented with hierarchical structures, which consist of different granularities of entity subtypes. It is natural that entities in the real world usually have multiple entity types. Most of the existing famous knowledge graphs own their customized hierarchical structures of entity types.

3. **Images** provide intuitive visual information to describe what the entity looks like, which is confirmed to be the most significant information we receive and process every day. The latent information located in images helps a lot, especially when dealing with concrete entities. For instance, we may find out the potential relationship between Cherry and Plum (there are both plants belonging to Rosaceae) from their appearances. Images could be downloaded from websites, and there are also substantial image datasets like ImageNet.

Multisource information learning provides a novel method to learn knowledge representations not only from the internal information of structured knowledge graphs but also from the external information of plain texts, hierarchical types, and images. Moreover, the exploration in multisource information learning helps to further understand human cognition with all senses in the real world. The cross-modal representations learned based on knowledge graphs will also provide possible relationships between different kinds of information.

## 7.3.1 Knowledge Graph Representation with Texts

Textual information is one of the most common and widely used information these days. There are large plain texts generated every day on the web and easy to be extracted. Words are compressed symbols of our thoughts and can provide the connections between entities, which are of great significance in KRL.

### 7.3.1.1 Knowledge Graph and Text Joint Embedding

Wang et al. [76] attempt to utilize textual information by jointly embedding entities, relations, and words into the same low-dimensional continuous vector space. Their joint model contains three parts, namely, the knowledge model, the text model, and the alignment model. More specifically, the knowledge model is learned based on the triple facts in KGs by translation-based models, while the text model is learned based on the concurrences of words in the large corpus by Skip-gram. As for the alignment model, two methods are proposed utilizing Wikipedia anchors and entity names. The main idea of alignment by Wikipedia anchors is replacing the word-word pair $(w, v)$ with the word-entity pair $(w, e_v)$ according to the anchors in Wiki pages, while the main idea of alignment by entity names is replacing the entities in original triple $\langle h, r, t \rangle$ with the corresponding entity names $\langle w_h, r, t \rangle$, $\langle h, r, w_t \rangle$, and $\langle w_h, r, w_t \rangle$.

Modeling entities and words into the same vector space are capable of encoding both information in knowledge graphs and that in plain texts, while the performance of this joint model depends on the completeness of Wikipedia anchors and may suffer from the weak interactions merely based on entity names. To address this issue, [101] proposes a new joint embedding based on [76] and improves the alignment model with entity descriptions into consideration, assuming that entities should be similar to all words in their descriptions. These joint models learn knowledge and text joint embeddings, improving evaluation performance in both word and knowledge representations.

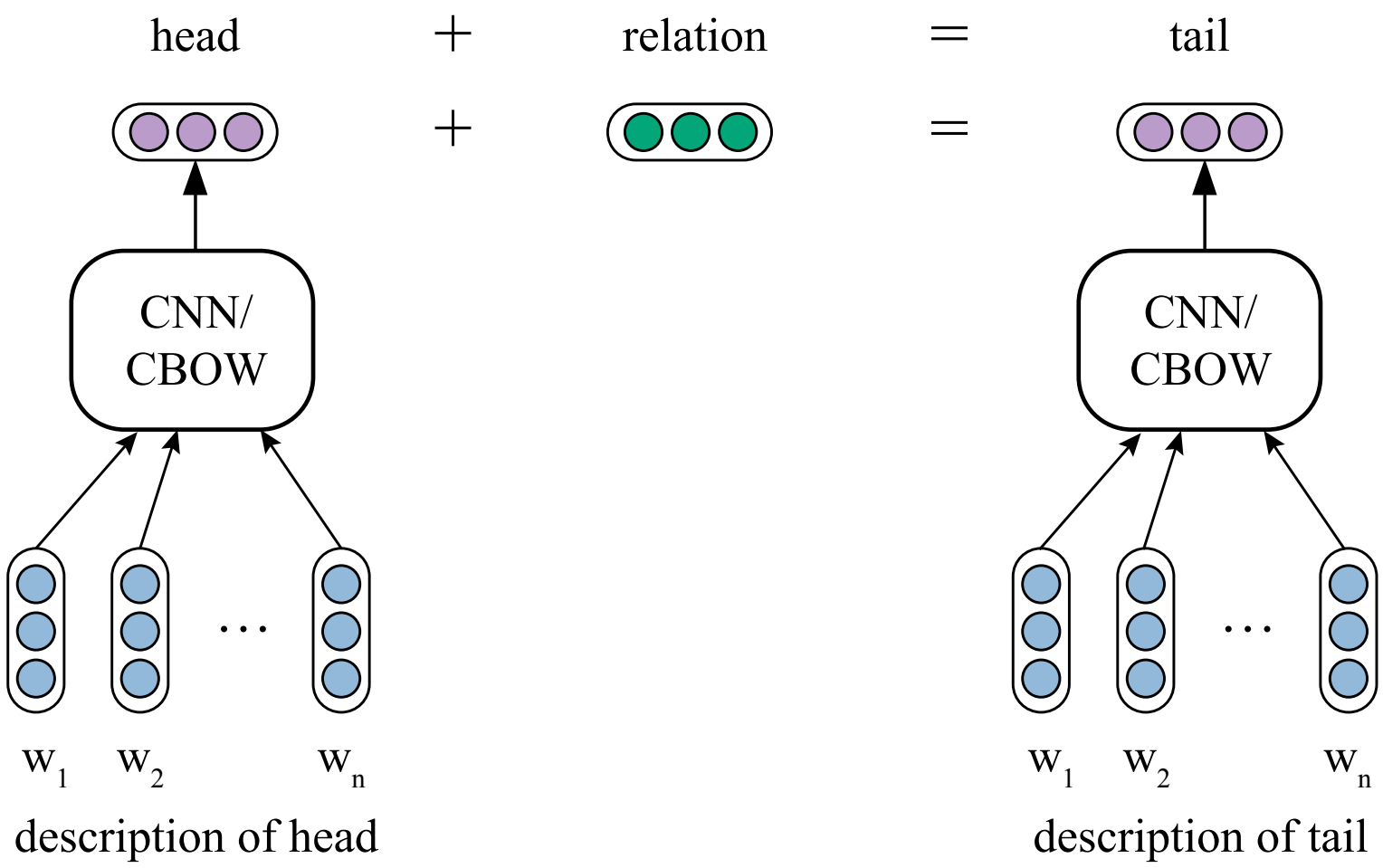

**Fig. 7.12** The architecture of DKRL model

#### 7.3.1.2 Description-Embodied Knowledge Graph Representation

Another way of utilizing textual information is directly constructing knowledge representations from entity descriptions instead of merely considering the alignments. Xie et al. [82] proposes Description-embodied Knowledge Graph Representation Learning (DKRL) that provides two kinds of knowledge representations: the first is the structure-based representation $\mathbf{h}_S$ and $\mathbf{t}_S$, which can directly represent entities widely used in previous methods, and the second is the description-based representation $\mathbf{h}_D$ and $\mathbf{t}_D$ which derives from entity descriptions. The energy function derives from translation-based framework:

$$\mathscr{E}(h, r, t) = \|\mathbf{h}_S + \mathbf{r} - \mathbf{t}_S\| + \|\mathbf{h}_S + \mathbf{r} - \mathbf{t}_D\| + \|\mathbf{h}_D + \mathbf{r} - \mathbf{t}_S\| + \|\mathbf{h}_D + \mathbf{r} - \mathbf{t}_D\|. \tag{7.50}$$

The description-based representation is constructed via CBOW or CNN encoders that encode rich textual information from plain texts into knowledge representations. The architecture of DKRL is shown in Fig. 7.12.

Compared to conventional translation-based methods, the two types of entity representations in DKRL are constructed with both structural information and textual information, and thus could get better performance in knowledge graph completion and type classification. Besides, DKRL could represent an entity even if it is not in the training set, as long as there are a few sentences to describe the entity. As their millions of new entities come up every day, DKRL is capable of handling zero-shot learning.

## 7.3.2 *Knowledge Graph Representation with Types*

Entity types, which serve as a kind of category information of entities and are usually arranged with hierarchical structures, could provide structured information to understand entities in KRL better.

### 7.3.2.1 Type-Constraint Knowledge Graph Representation

Krompaß et al. [36] take type information as type constraints, and improves existing methods like RESCAL and TransE via type constraints. It is intuitive that in a particular relation, the head or tail entities should belong to some specific types. For example, the head entities of the relation `write_books` should be a human (or more precisely an author), and the tail entities should be a book.

Specifically, in RESCAL, the original factorization $\mathbf{X}_r \approx \mathbf{A}\mathbf{R}_r\mathbf{A}^\top$ is modified to

$$\mathbf{X}'_r \approx \mathbf{A}_{[head_r,:]}\mathbf{R}_r\mathbf{A}^\top_{[tail_r,:]}, \tag{7.51}$$

in which $head_r, tail_r$ are the set of entities fitting the type constraints of head or tail and $\mathbf{X}'_r$ is a sparse adjacency matrix of shape $|head_r| \times |tail_r|$. In the enhanced version, only the entities that fit type constraints will be considered during factorization.

In TransE, type constraints are utilized in negative sampling. The margin-based score functions of translation-based methods need negative instances, which are generated through randomly replacing head or tail entities with another entity in triples. With type constraints, the negative samples are chosen by

$$h' \in E_{[head_r]} \subseteq E, \quad t' \in E_{[tail_r]} \subseteq E, \tag{7.52}$$

where $E_{[head_r]}$ is the subset of entities following type constraints for head in relation $r$, and $E_{[tail_r]}$ is that for tail.

### 7.3.2.2 Type-Embodied Knowledge Graph Representation

Considering type information as constraints is simple but effective, while the performance is still limited. Instead of merely viewing type information as type constraints, Xie et al. [83] propose Type-embodied Knowledge Graph Representation Learning (TKRL), utilizing hierarchical-type structures to instruct the construction of projection matrices. Inspired by TransR that every entity should have multiple representations in different scenarios, the energy function of TKRL is defined as follows:

$$\mathscr{E}(h, r, t) = \|\mathbf{M}_{rh}\mathbf{h} + \mathbf{r} - \mathbf{M}_{rt}\mathbf{t}\|, \tag{7.53}$$

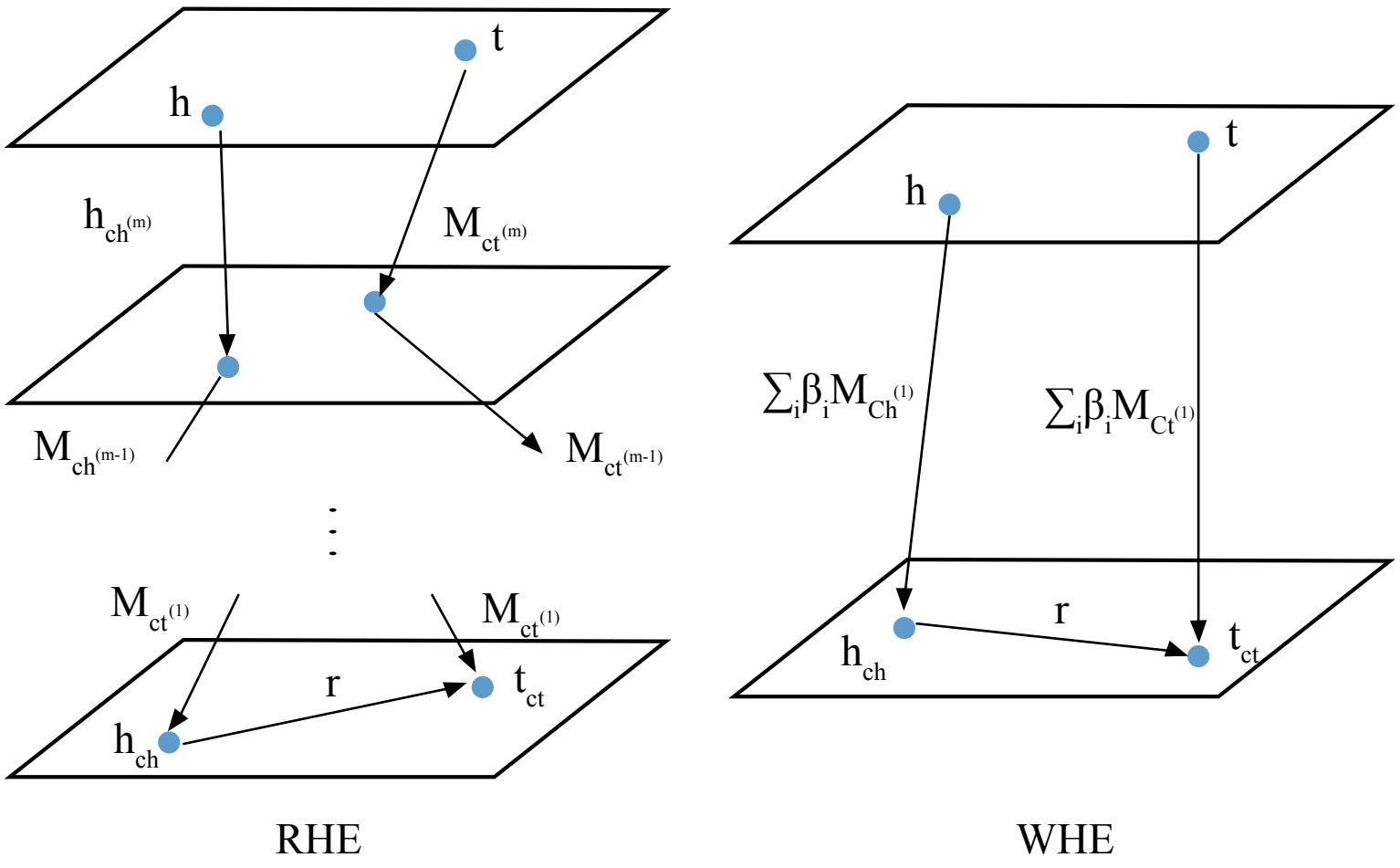

**Fig. 7.13** The architecture of TKRL model

in which $\mathbf{M}_{rh}$ and $\mathbf{M}_{rt}$ are two projection matrices for $h$ and $t$ that depend on their corresponding hierarchical types in this triple. Two hierarchical-type encoders are proposed to learn the projection matrices, regarding all subtypes in the hierarchy as projection matrices, in which Recursive Hierarchy Encoder is based on matrix multiplication, while Weighted Hierarchy Encoder is based on matrix summation:

$$\mathbf{M}_{RHE_c} = \prod_{i=1}^{m} \mathbf{M}_{c^{(i)}} = \mathbf{M}_{c^{(1)}} \mathbf{M}_{c^{(2)}} \ldots \mathbf{M}_{c^{(m)}}, \tag{7.54}$$

$$\mathbf{M}_{WHE_c} = \sum_{i=1}^{m} \beta_i \mathbf{M}_{c^{(i)}} = \beta_1 \mathbf{M}_{c^{(1)}} + \cdots + \beta_m \mathbf{M}_{c^{(m)}}, \tag{7.55}$$

where $\mathbf{M}_{c^{(i)}}$ stands for the projection matrix of the $i$th subtype of the hierarchical type $c$, $\beta_i$ is the corresponding weight of the subtype. Figure 7.13 demonstrates a simple illustration of TKRL. Taking RHE, for instance, given an entity `William Shakespeare`, it is first projected to a rather general sub-type space like `human`, and then sequentially projected to a more precise subtype like `author` or `English author`. Moreover, TKRL also proposes an enhanced soft-type constraint to alleviate the problems caused by type information incompleteness.

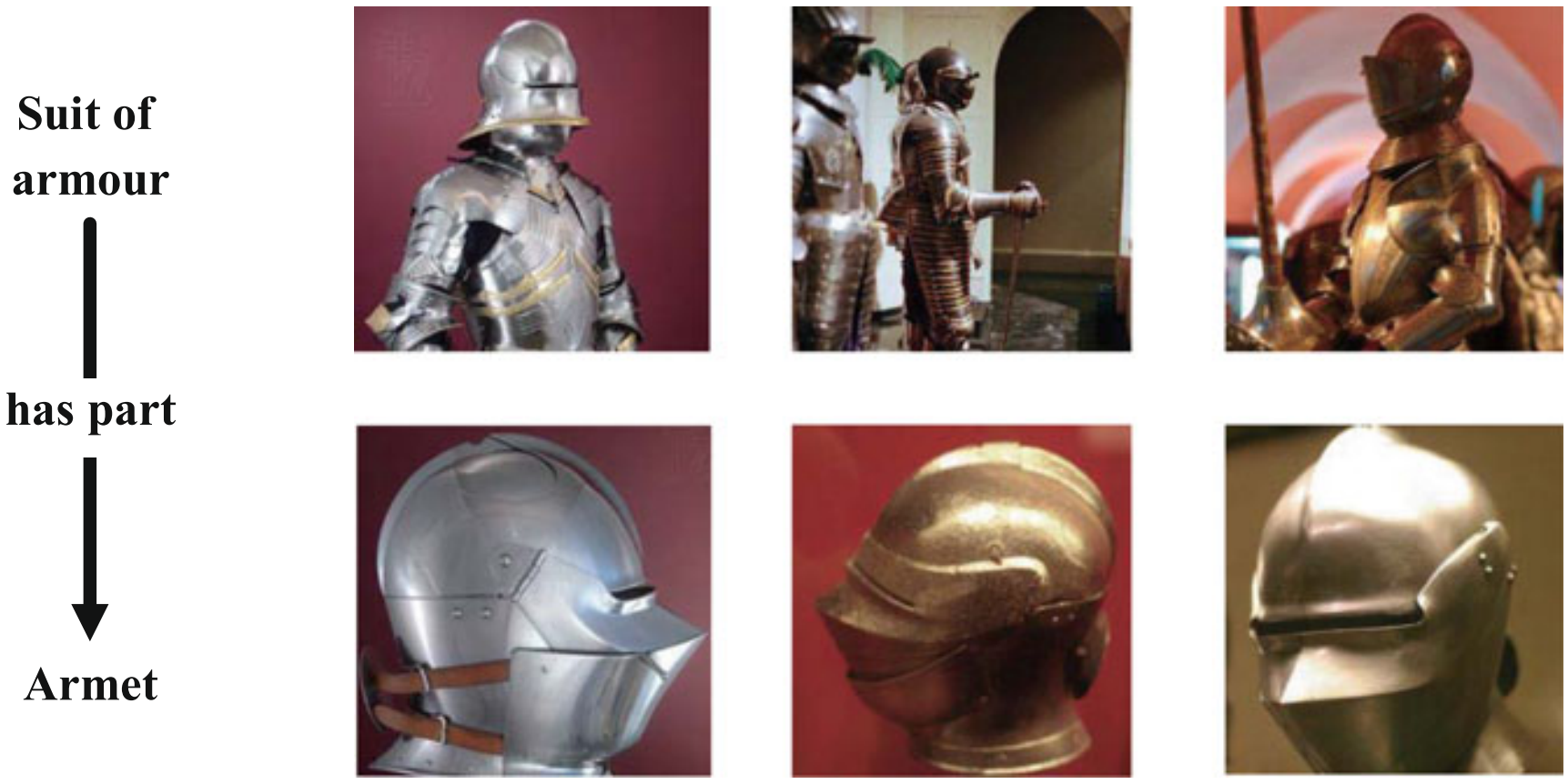

**Fig. 7.14** Examples of entity images [81]

## 7.3.3 *Knowledge Graph Representation with Images*

Images could provide intuitive visual information of their corresponding entities' outlook, which may give significant hints suggesting some latent attributes of entities from certain aspects. For instance, Fig. 7.14 demonstrates some examples of entity images of their corresponding entities `Suit of armour` and `Armet`. The left side shows the triple facts that ⟨`Suit of armour`, `has_a_part`, `Armet`⟩, and surprisingly, we can infer this knowledge directly from the images.

### 7.3.3.1 Image-Embodied Knowledge Graph Representation

Xie et al. [81] propose Image-embodied Knowledge Graph Representation Learning (IKRL) to take visual information into consideration when constructing knowledge representations. Inspired by the multiple entity representations in [82], IKRL also proposes the image-based representation $\mathbf{h}_I$ and $\mathbf{t}_I$ besides the original structure-based representation, and jointly learn both two types of entity representations simultaneously within the translation-based framework.

$$\mathscr{E}(h, r, t) = \|\mathbf{h}_S + \mathbf{r} - \mathbf{t}_S\| + \|\mathbf{h}_S + \mathbf{r} - \mathbf{t}_I\| + \|\mathbf{h}_I + \mathbf{r} - \mathbf{t}_S\| + \|\mathbf{h}_I + \mathbf{r} - \mathbf{t}_I\|. \tag{7.56}$$

More specifically, IKRL first constructs the image representations for all entity images with neural networks, and then project these image representations from image space to entity space via a projection matrix. Since most entities may have multiple images with different qualities, IKRL selects the more informative and discriminative images via an attention-based method. The evaluation results of IKRL not only confirm the significance of visual information in understanding entities but also show the possibility of a joint heterogeneous semantic space. Moreover, the authors also find some interesting semantic regularities such as $\mathbf{w}(\texttt{man}) - \mathbf{w}(\texttt{king}) \approx \mathbf{w}(\texttt{woman}) - \mathbf{w}(\texttt{queen})$ found in word space, which are shown in Fig. 7.15.

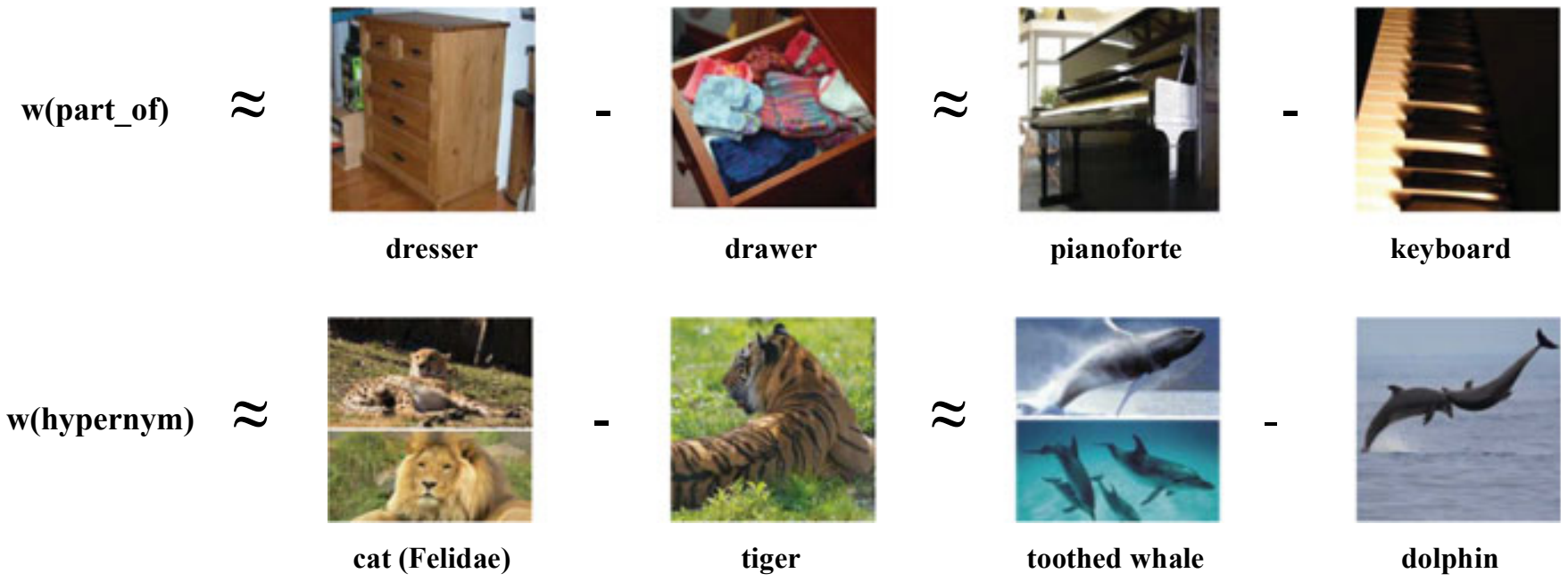

**Fig. 7.15** An example of semantic regularities in word space [81]

### *7.3.4 Knowledge Graph Representation with Logic Rules*

Typical knowledge graphs store knowledge in the form of triple facts with one relation linking two entities. Most existing KRL methods only consider the information within triple facts separately, ignoring the possible interactions and correlations between different triples. Logic rules, which are certain kinds of summaries deriving from human beings' prior knowledge, could help us with knowledge inference and reasoning. For instance, if we know the triple fact that $\langle$ `Beijing`, `is_capital_of`, `China`$\rangle$, we can easily infer with high confidence that $\langle$`Beijing`, `located_in`, `China`$\rangle$, since we know the logic rule that the relation `is_capital_of` $\Rightarrow$ `located_in`.

Some works are focusing on introducing logic rules to knowledge acquisition and inference, among which Markov Logic Networks are intuitively utilized to address this challenge [3, 58, 75]. The path-based TransE [38] stated above also implicitly considers the latent logic rules between different relations via relation paths.

#### 7.3.4.1 KALE

KALE is a translation-based KRL method that jointly learns knowledge representations with logic rules [24]. The joint learning consists of two parts, namely, the triple modeling and the rule modeling. For triple modeling, KALE follows the translation assumption with minor alteration in scoring function as follows:

$$\mathscr{E}(h, r, t) = 1 - \frac{1}{3\sqrt{d}}\|\mathbf{h} + \mathbf{r} - \mathbf{t}\|, \tag{7.57}$$

in which $d$ stands for the dimension of knowledge embeddings. $\mathscr{E}(h, r, t)$ takes value in [0, 1] for the convenience of joint learning.

For the newly added rule modeling, KALE uses the t-norm fuzzy logics proposed in [25] that represent the truth value of a complex formula with the truth values of its constituents. Specially, KALE focuses on two typical types of logic rules. The first is $\forall h, t : \langle h, r_1, t\rangle \Rightarrow \langle h, r_2, t\rangle$ (e.g., given ⟨`Beijing`, `is_capital_of`, `China`⟩, we can infer that ⟨`Beijing`, `located_in`, `China`⟩). KALE represents the scoring function of this logic rule $f_1$ via specific t-norm based logical connectives as follows:

$$\mathscr{E}(f_1) = \mathscr{E}(h, r_1, t)\mathscr{E}(h, r_2, t) - \mathscr{E}(h, r_1, t) + 1. \tag{7.58}$$

The second is $\forall h, e, t : \langle h, r_1, e\rangle \wedge \langle e, r_2, t\rangle \Rightarrow \langle h, r_3, t\rangle$ (e.g., given ⟨`Tsinghua`, `located_in`, `Beijing`⟩) and ⟨`Beijing`, `located_in`, `China`⟩, we can infer that ⟨`Tsinghua`, `located_in`, `China`⟩). And KALE defines the second scoring function as

$$\mathscr{E}(f_2) = \mathscr{E}(h, r_1, e)\mathscr{E}(e, r_2, t)\mathscr{E}(h, r_3, t) - \mathscr{E}(h, r_1, e)\mathscr{E}(e, r_2, t) + 1. \tag{7.59}$$

The joint training contains all positive formulae, including triple facts as well as logic rules. Note that for the consideration of logic rule qualities, KALE ranks all possible logic rules by their truth values with pretrained TransE and manually filters some rules ranked at the top.

## 7.4 Applications

Recent years have witnessed the great thrive in knowledge-driven artificial intelligence, such as QA systems and chatbot. AI agents are expected to accurately and deeply understand user demands, and then appropriately and flexibly give responses and solutions. Such kind of work cannot be done without certain forms of knowledge.

To introduce knowledge to AI agents, researchers first extract knowledge from heterogeneous information like plain texts, images, and structured knowledge bases. These various kinds of heterogeneous information are then fused and stored with certain structures like knowledge graphs. Next, the knowledge is projected to a low-dimensional semantic space following some KRL methods. And finally, these learned knowledge representations are utilized in various knowledge applications like information retrieval and dialogue system. Figure 7.16 demonstrates a brief pipeline of knowledge-driven applications from scratch.

From the illustration, we can observe that knowledge graph representation learning is the critical component in the whole knowledge-driven application's pipeline. It bridges the gap between knowledge graphs that store knowledge and knowledge applications that use knowledge. Knowledge representations with distributed methods, compared to those with symbolic methods, are able to solve the data sparsity and modeling the similarities between entities and relations. Moreover, embedding-based methods are convenient to be used with deep learning methods and are naturally fit for the combination with heterogeneous information.

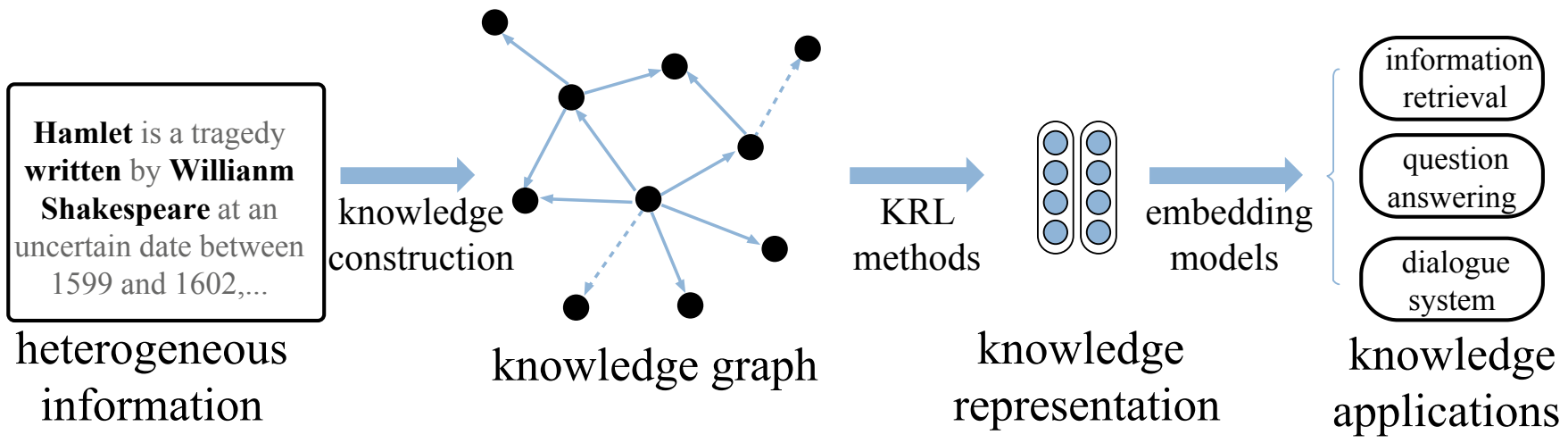

**Fig. 7.16** An illustration of knowledge-driven applications

In this section, we will introduce possible applications of knowledge representations mainly from two aspects. First, we will introduce the usage of knowledge representations for knowledge-driven applications, and then we will show the power of knowledge representations for knowledge extraction and construction.

## *7.4.1 Knowledge Graph Completion*

Knowledge graph completion aims to build structured knowledge bases by extracting knowledge from heterogeneous sources such as plain texts, existing knowledge bases, and images. Knowledge construction consists of several subtasks like relation extraction and information extraction, making the fundamental step in the whole knowledge-driven framework.

Recently, automatic knowledge construction has attracted considerable attention since it is incredibly time consuming and labor intensive to deal with enormous existing and new information. In the following section, we will introduce some explorations on neural relation extraction, and concentrate on the combination of knowledge representations.

### 7.4.1.1 Knowledge Representations for Relation Extraction

Relation extraction focuses on predicting the correct relation between two entities given a short plain text containing the two entities. Generally, all relations to predict are predefined, which is different to open information extraction. Entities are usually marked with named entity recognition systems or extracted according to anchor texts, or automatically generated via distance supervision [50].

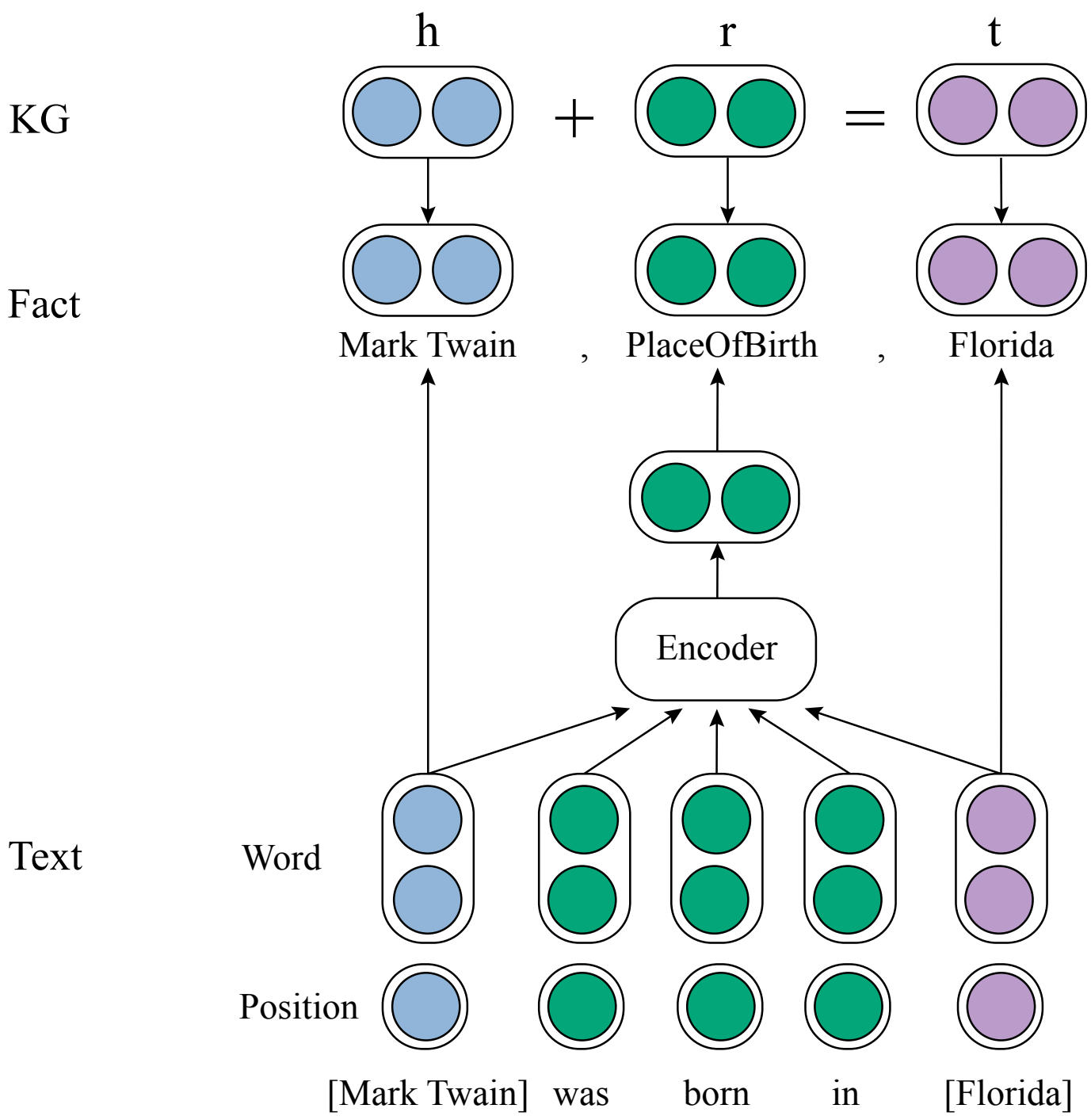

**Fig. 7.17** The architecture of joint representation learning framework for knowledge acquisition

Conventional methods for relation extraction and classification are mainly based on statistical machine learning, which strongly depends on the qualities of extracted features. Zeng et al. [96] first introduce CNN to relation classification and achieve great improvements. Lin et al. [40] further improves neural relation extraction models with attention-based models over instances.

Han et al. [27, 28] propose a novel joint representation learning framework for knowledge acquisition. The key idea is that the joint model learns knowledge and text representations within a unified semantic space via KG-text alignments. Figure 7.17 shows the brief framework of the KG-text joint model. For the text part, the sentence with two entities `Mark Twain` and `Florida` is regarded as the input for a CNN encoder, and the output of CNN is considered to be the latent relation `place_of_birth` of this sentence. While for the KG part, entity and relation representations are learned via translation-based methods. The learned representations of KG and text parts are aligned during training. This work is the first attempt to encode knowledge representations from existing KGs to knowledge construction tasks and achieves improvements in both knowledge completion and relation extraction.

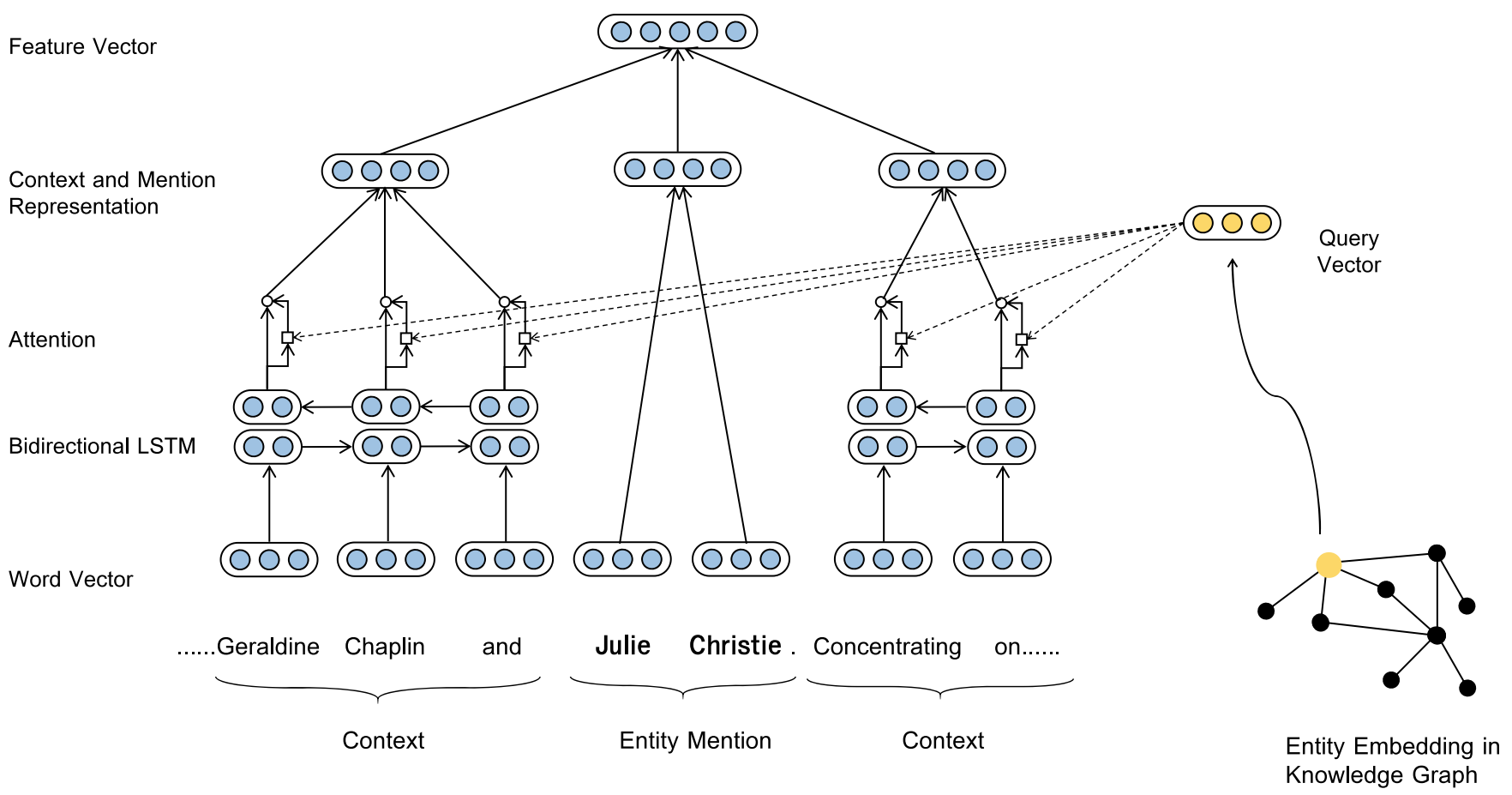

**Fig. 7.18** The architecture of KNET model

### *7.4.2 Knowledge-Guided Entity Typing*

Entity typing is the task of detecting semantic types for a named entity (or entity mention) in plain text. For example, given a sentence `Jordan played 15 seasons in the NBA`, entity typing aims to infer that `Jordan` in this sentence is a `person`, an `athlete`, and even a `basketball player`. Entity typing is important for named entity disambiguation since it can narrow down the range of candidates for an entity mention [10]. Moreover, entity typing also benefits massive Natural Language Processing (NLP) tasks such as relation extraction [46], question answering [90], and knowledge base population [9].

Conventional named entity recognition models [69, 73] typically classify entity mentions into a small set of coarse labels (e.g., `person`, `organization`, `location`, and `others`). Since these entity types are too coarse grained for many NLP tasks, a number of works [15, 41, 94, 95] have been proposed to introduce a much larger set of fine-grained types, which are typically subtypes of those coarse-grained types. Previous fine-grained entity typing methods usually derive features using NLP tools such as POS tagging and parsing, and inevitably suffer from error propagation. Dong et al. [18] make the first attempt to explore deep learning in entity typing. The method only employs word vectors as features, discarding complicated feature engineering. Shimaoka et al. [63] further introduce the attention scheme into neural models for fine-grained entity typing.

Neural models have achieved state-of-the-art performance for fine-grained entity typing. However, these methods face the following nontrivial challenges:

(1) **Entity-Context Separation.** Existing methods typically encode context words without utilizing crucial correlations between entity and context. However, it is intuitive that the importance of words in the context for entity typ-ing is significantly influenced by which entity mentions we concern about. For example, in a sentence `In 1975, Gates and Paul Allen co-founded Microsoft, which became the world's largest PC software company`, the word `company` is much more important for determining the type of `Microsoft` than for the type of `Gates`.

(2) **Entity-Knowledge Separation.** Existing methods only consider text information of entity mentions for entity typing. In fact, Knowledge Graphs (KGs) provide rich and effective additional information for determining entity types. For example, in the above sentence `In 1975, Gates ... Microsoft ... company`, even if we have no type information of `Microsoft` in KG, entities similar to `Microsoft` (such as `IBM`) will also provide supplementary information.

In order to address the issues of entity-context separation and entity-knowledge separation, we propose **K**nowledge-guided **A**ttention (KNET) **N**eural **E**ntity **T**yping. As illustrated in Fig. 7.18, KNET mainly consists of two parts. Firstly, KNET builds a neural network, including a Long Short-Term Memory (LSTM) and a fully connected layer, to generate context and named entity representations. Secondly, KNET introduces knowledge attention to emphasize those critical words and improve the quality of context representations. Here we introduce the knowledge attention in detail.

Knowledge graphs provide rich information about entities in the form of triples $\langle h, r, t \rangle$, where $h$ and $t$ are entities and $r$ is the relation between them. Many KRL works have been devoted to encoding entities and relations into real-valued semantic vector space based on triple information in KGs. KRL provides us with an efficient way to exploit KG information for entity typing.

KNET employs the most widely used KRL method TransE to obtain entity embedding $\mathbf{e}$ for each entity $e$. During the training scenario, it is known that the entity mention $m$ indicates the corresponding $e$ in KGs with embedding $\mathbf{e}$, and hence, KNET can directly compute knowledge attention as follows:

$$\alpha_i^{\text{KA}} = f\left(\mathbf{e}\mathbf{W}_{\text{KA}} \begin{bmatrix} \overrightarrow{\mathbf{h}_i} \\ \overleftarrow{\mathbf{h}_i} \end{bmatrix}\right), \tag{7.60}$$

where $\mathbf{W}_{\text{KA}}$ is a bilinear parameter matrix, and $\alpha_i^{\text{KA}}$ is the attention weight for the $i$th word.

**Knowledge Attention in Testing.** The challenge is that, in the testing scenario, we do not know the corresponding entity in the KG of a certain entity mention. A solution is to perform entity linking, but it will introduce linking errors. Besides, in many cases, KGs may not contain the corresponding entities for many entity mentions.

To address this challenge, we build an additional text-based representation for entities in KGs during training. Concretely, for an entity $e$ and its context sentence $s$, we encode its left and right context into $\mathbf{c}_l$ and $\mathbf{c}_r$ using an one-directional LSTM, and further learn the text-based representation $\hat{\mathbf{e}}$ as follows:

$$\hat{\mathbf{e}} = \tanh\left(\mathbf{W}\begin{bmatrix}\mathbf{m}\\ \mathbf{c}_l\\ \mathbf{c}_r\end{bmatrix}\right), \tag{7.61}$$

where $\mathbf{W}$ is the parameter matrix, and $\mathbf{m}$ is the mention representation. Note that, LSTM used here is different from those in context representation in order to prevent interference. In order to bridge text-based and KG-based representations, in the training scenario, we simultaneously learn $\hat{\mathbf{e}}$ by putting an additional component in the objective function:

$$\mathscr{O}_{\mathrm{KG}}(\theta) = -\sum_{e} \|\mathbf{e} - \hat{\mathbf{e}}\|^2. \tag{7.62}$$

In this way, in the testing scenario, we can directly use Eq. 7.61 to obtain the corresponding entity representation and compute knowledge attention using Eq. 7.60.

### 7.4.3 Knowledge-Guided Information Retrieval

The emergence of large-scale knowledge graphs has motivated the development of *entity-oriented search*, which utilizes knowledge graphs to improve search engines. Recent progresses in entity-oriented search include better text representations with entity annotations [61, 85], richer ranking features [14], entity-based connections between query and documents [45, 84], and soft-match query and documents through knowledge graph relations or embeddings [19, 88]. These approaches bring in entities and semantics from knowledge graphs and have greatly improved the effectiveness of feature-based search systems.

Another frontier of information retrieval is the development of neural ranking models (*neural-IR*). Deep learning techniques have been used to learn distributed representations of queries and documents that capture their relevance relations (*representation-based*) [62], or to model the query-document relevancy directly from their word-level interactions (*interaction-based*) [13, 23, 87]. Neural-IR approaches, especially the *interaction-based* ones, have greatly improved the ranking accuracy when large-scale training data are available [13].

Entity-oriented search and neural-IR push the boundary of search engines from two different aspects. Entity-oriented search incorporates human knowledge from entities and knowledge graph semantics. It has shown promising results on feature-based ranking systems. On the other hand, neural-IR leverages distributed representations and neural networks to learn more sophisticated ranking models form large-scale training data. Entity-Duet Neural Ranking Model (EDRM), as shown in Fig. 7.19, incorporates entities in interaction-based neural ranking models. EDRM first learns the distributed representations of entities using their semantics from knowledge graphs: descriptions and types. Then it follows a recent state-of-the-art entity-oriented search framework, the word-entity duet [86], and matches documents to queries with both bag-of-words and bag-of-entities. Instead of manual features, EDRM uses interaction-based neural models [13] to match the query and documents with word-entity duet representations. As a result, EDRM combines entity-oriented search and the interaction-based neural-IR; it brings the knowledge graph semantics to neural-IR and enhances entity-oriented search with neural networks.

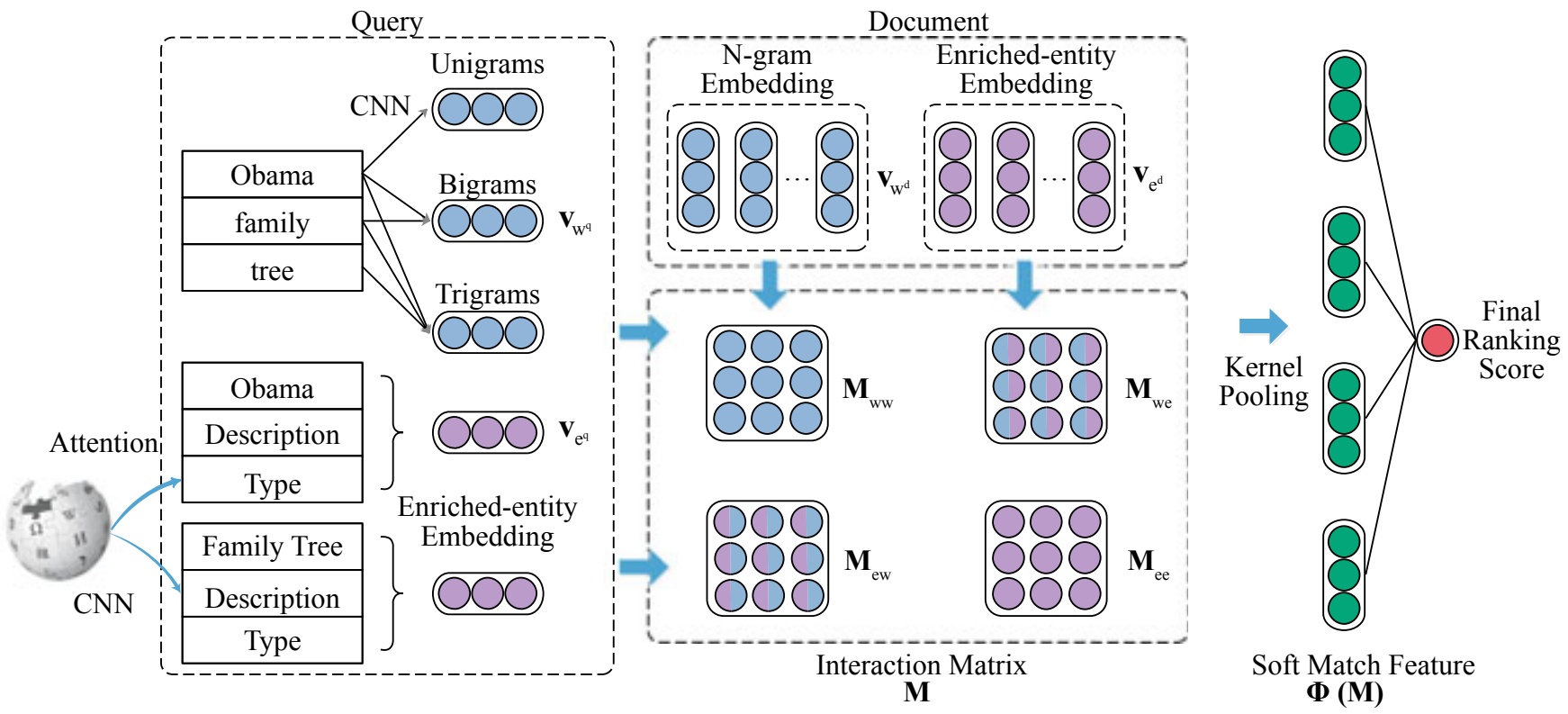

**Fig. 7.19** The architecture of EDRM model

#### 7.4.3.1 Interaction-Based Ranking Models

Given a query $q$ and a document $d$, interaction-based models first build the word-level translation matrix between $q$ and $d$. The translation matrix describes word-pairs similarities using word correlations, which are captured by word embedding similarities in interaction-based models.

Typically, interaction-based ranking models first map each word $w$ in $q$ and $d$ to an $L$-dimensional embedding $\mathbf{v}_w$.

$$\mathbf{v}_w = \text{Emb}_w(w). \tag{7.63}$$

It then constructs the interaction matrix $\mathbf{M}$ based on query and document embeddings. Each element $\mathbf{M}_{ij}$ in the matrix, compares the $i$th word in $q$ and the $j$th word in $d$, e.g., using the cosine similarity of word embeddings:

$$\mathbf{M}_{ij} = \cos(\mathbf{v}_{w_i^q}, \mathbf{v}_{w_j^d}). \tag{7.64}$$

With the translation matrix describing the term level matches between query and documents, the next step is to calculate the final ranking score from the matrix. Many approaches have been developed in interaction-based neural ranking models, but in general, that would include a feature extractor on $\mathbf{M}$ and then one or several ranking layers to combine the features to the ranking score.

#### 7.4.3.2 Semantic Entity Representation

EDRM incorporates the semantic information about an entity from the knowledge graphs into its representation. The representation includes three embeddings: entity embedding, description embedding, and type embedding, all in $L$ dimension and are combined to generate the semantic representation of the entity.

**Entity Embedding** uses an $L$-dimensional embedding layer $\text{Emb}_e$ to get the entity embedding **e** for $e$:

$$\mathbf{v}_e = \text{Emb}_e(e). \tag{7.65}$$

**Description Embedding** encodes an entity description which contains $m$ words and explains the entity. EDRM first employs the word embedding layer $\text{Emb}_v$ to embed the description word $v$ to $\mathbf{v}$. Then it combines all embeddings in the text to an embedding matrix $\mathbf{V}$. Next, it leverages convolutional filters to slide over the text and compose the $l$ length $n$-gram as $\mathbf{g}_e^j$:

$$\mathbf{g}_e^j = \text{ReLU}(\mathbf{W}_{\text{CNN}} \cdot \mathbf{V}_w^{j:j+h} + \mathbf{b}_{\text{CNN}}), \tag{7.66}$$

where $W_{\text{CNN}}$ and $\mathbf{b}_{\text{CNN}}$ are two parameters of the convolutional filter.

Then we use max pooling after the convolution layer to generate the description embedding $\mathbf{v}_e^{des}$:

$$\mathbf{v}_e^{des} = \max(\mathbf{g}_e^1, ..., \mathbf{g}_e^j, ..., \mathbf{g}_e^m). \tag{7.67}$$

**Type Embedding** encodes the categories of entities. Each entity $e$ has $n$ kinds of types $F_e = \{f_1, ..., f_j, ..., f_n\}$. EDRM first gets the $f_j$ embedding $\mathbf{v}_{f_j}$ through the type embedding layer $\text{Emb}_{type}$:

$$\mathbf{v}_{f_j}^{emb} = \text{Emb}_{type}(e). \tag{7.68}$$

Then EDRM utilizes an attention mechanism to combine entity types to the type embedding $\mathbf{v}_e^{type}$:

$$\mathbf{v}_e^{type} = \sum_j^n \alpha_j \mathbf{v}_{f_j}, \tag{7.69}$$

where $\alpha_j$ is the attention score, calculated as:

$$\alpha_j = \frac{\exp(y_j)}{\sum_l^n \exp(y_l)}, \tag{7.70}$$

$$y_j = \left(\sum_i \mathbf{W}_{bow} \mathbf{v}_{t_i}\right) \cdot \mathbf{v}_{f_j}, \tag{7.71}$$

where $y_j$ is the dot product of the query or document representation and type embedding $f_j$. We leverage bag-of-words for query or document encoding. $\mathbf{W}_{bow}$ is a parameter matrix.

**Combination.** The three embeddings are combined by a linear layer to generate the semantic representation of the entity:

$$\mathbf{v}_e^{sem} = \mathbf{v}_e^{emb} + \mathbf{W}_e[\mathbf{v}_e^{des}; \mathbf{v}_e^{type}]^\top + \mathbf{b}_e, \tag{7.72}$$

in which $\mathbf{W}_e$ is an $L \times 2L$ matrix and $\mathbf{b}_e$ is an $L$-dimensional vector.

#### 7.4.3.3 Neural Entity-Duet Framework

Word-entity duet [86] is a recently developed framework in entity-oriented search. It utilizes the duet representation of bag-of-words and bag-of-entities to match question $q$ and document $d$ with handcrafted features. This work introduces it to neural-IR.

They first construct bag-of-entities $q^e$ and $d^e$ with entity annotation as well as bag-of-words $q^w$ and $d^w$ for $q$ and $d$. The duet utilizes a four-way interaction: query words to document words ($q^w$-$d^w$), query words to documents entities ($q^w$-$d^e$), query entities to document words ($q^e$-$d^w$), and query entities to document entities ($q^e$-$d^e$).

Instead of features, EDRM uses a translation layer that calculates the similarity between a pair of query-document terms: ($\mathbf{v}_{w^q}^i$ or $\mathbf{v}_{e^q}^i$) and ($\mathbf{v}_{w^d}^j$ or $\mathbf{v}_{e^d}^j$). It constructs the interaction matrix $\mathbf{M} = \{\mathbf{M}_{ww}, \mathbf{M}_{we}, \mathbf{M}_{ew}, \mathbf{M}_{ee}\}$. And $\mathbf{M}_{ww}, \mathbf{M}_{we}, \mathbf{M}_{ew}, \mathbf{M}_{ee}$ denote interactions of $q^w$-$d^w$, $q^w$-$d^e$, $q^e$-$d^w$, $q^e$-$d^e$ respectively. And elements in them are the cosine similarities of corresponding terms:

$$\begin{aligned} \mathbf{M}_{ww}^{ij} &= \cos(\mathbf{v}_{w^q}^i, \mathbf{v}_{w^d}^j); \mathbf{M}_{ee}^{ij} = \cos(\mathbf{v}_{e^q}^i, \mathbf{v}_{e^d}^j) \\ \mathbf{M}_{ew}^{ij} &= \cos(\mathbf{v}_{e^q}^i, \mathbf{v}_{w^d}^j); \mathbf{M}_{we}^{ij} = \cos(\mathbf{v}_{w^q}^i, \mathbf{v}_{e^d}^j). \end{aligned} \tag{7.73}$$

The final ranking feature $\Phi(\mathbf{M})$ is a concatenation of four cross matches ($\phi(\mathbf{M})$):

$$\Phi(\mathbf{M}) = [\phi(\mathbf{M}_{ww}); \phi(\mathbf{M}_{we}); \phi(\mathbf{M}_{ew}); \phi(\mathbf{M}_{ee})], \tag{7.74}$$

where the $\phi$ can be any function used in interaction-based neural ranking models.

The entity-duet presents an effective way to crossly match query and document in entity and word spaces. In EDRM, it introduces the knowledge graph semantics representations into neural-IR models.

The duet translation matrices provided by EDRM can be plugged into any standard interaction-based neural ranking models such as K-NRM [87] and Conv-KNRM [13]. With sufficient training data, the whole model is optimized end-to-end with back-propagation. During the process, the integration of the knowledge graph semantics, entity embedding, description embeddings, type embeddings, and matching with entities is learned jointly with the ranking neural network.

### 7.4.4 *Knowledge-Guided Language Models*

Knowledge is an important external information for language modeling. It is because the statistical co-occurrences cannot instruct the generation of all kinds of knowledge, especially for those named entities with low frequencies. Researchers try to incorporate external knowledge into language models for better performance on generation and representation.

#### 7.4.4.1 NKLM

Language models aim to learn the probability distribution over sequences of words, which is a classical and essential NLP task widely studied. Recently, sequence to sequence neural models (seq2seq) are blooming and widely utilized in sequential generative tasks like machine translation [68] and image caption generation [72]. However, most seq2seq models have significant limitations when modeling and using background knowledge.

To address this problem, Ahn et al. [1] propose a Neural Knowledge Language Model (NKLM) that considers knowledge provided by knowledge graphs when generating natural language sequences with RNN language models. The key idea is that NKLM has two ways to generate a word. The first is the same way as conventional seq2seq models that generate a "vocabulary word" according to the probabilities of softmax, and the second is to generate a "knowledge word" according to the external knowledge graphs.

Specifically, the NKLM model takes LSTM as the framework of generating "vocabulary word". For external knowledge graph information, NKLM denotes the topic knowledge as $\mathscr{K} = \{a_1, \ldots a_{|\mathscr{K}|}\}$, in which $a_i$ represents the entities (i.e., named as "topic" in [1]) that appear in the same triple of a certain entity. At each step $t$, NKLM takes both "vocabulary word" $w^v_{t-1}$ and "knowledge word" $w^o_{t-1}$ as well as the fact $a_{t-1}$ predicted at step $t-1$ as the inputs of LSTM. Next, the hidden state of LSTM $h_t$ is combined with the knowledge context $e$ to get the fact key $k_t$ via an MLP module. The knowledge context $e_k$ derives from the mean embeddings of all related facts of fact $k$. The fact key $k_t$ is then used to extract the most appropriate fact $a_t$ from the corresponding topic knowledge. And finally, the selected fact $a_t$ is combined with hidden state $h_t$ to predict (1) both "vocabulary word" $w^v_t$ and "knowledge word" $w^o_t$, and (2) which word to generate at this step. The architecture of NKLM is shown in Fig. 7.20.

The NKLM model explores a novel neural model that combines the symbolic knowledge information in external knowledge graphs with seq2seq language models. However, the topic of knowledge is given when generating natural languages, which makes NKLM less practical and scalable for more general free talks. Nevertheless, we still believe that it is promising to encode knowledge into language models with such methods.

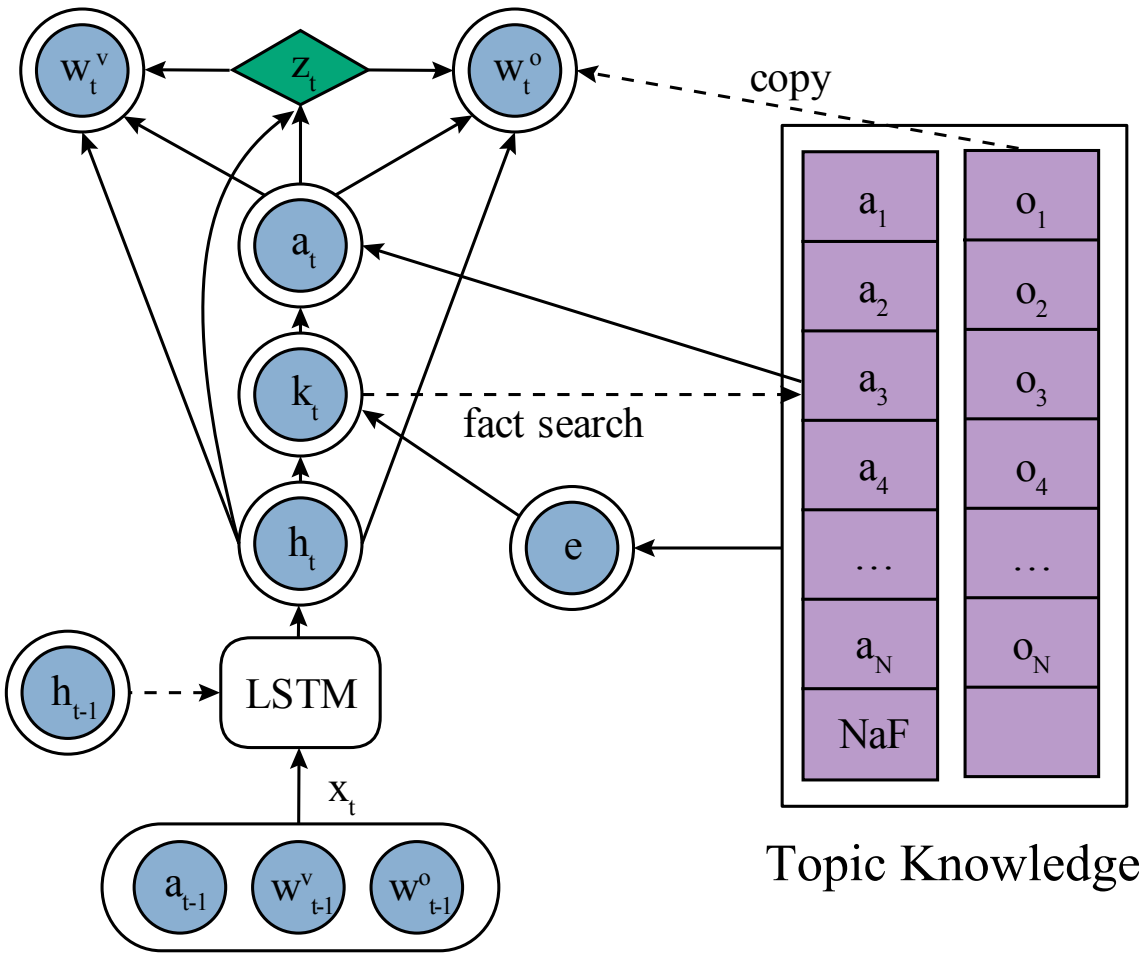

**Fig. 7.20** The architecture of NKLM model

#### 7.4.4.2 ERNIE

Pretrained language models like BERT [17] have a strong ability to represent language information from text. With rich language representation, pretrained models obtain state-of-the-art results on various NLP applications. However, the existing pretrained language models rarely consider incorporating external knowledge to provide related background information for better language understanding. For example, given a sentence `Bob Dylan wrote Blowin' in the Wind and Chronicles: Volume One`, without knowing `Blowin' in the Wind` and `Chronicles: Volume One` are `song` and `book` respectively, it is difficult to recognize the two occupations of `Bob Dylan`, i.e., `songwriter` and `writer`.

To enhance language representation models with external knowledge, Zhang et al. [100] propose an enhanced language representation model with informative entities (ERNIE). Knowledge Graphs (KGs) are important external knowledge resources, and they think informative entities in KGs can be the bridge to enhance language representation with knowledge. ERNIE considers overcoming two main challenges for incorporating external knowledge: *Structured Knowledge Encoding* and *Heterogeneous Information Fusion*.

For extracting and encoding knowledge information, ERNIE firstly recognizes named entity mentions in text and then aligns these mentions to their corresponding entities in KGs. Instead of directly using the graph-based facts in KGs, ERNIE encodes the graph structure of KGs with knowledge embedding algorithms like TransE [7], and then takes the informative entity embeddings as input. Based on the alignments between text and KGs, ERNIE integrates entity representations in the knowledge module into the underlying layers of the semantic module.

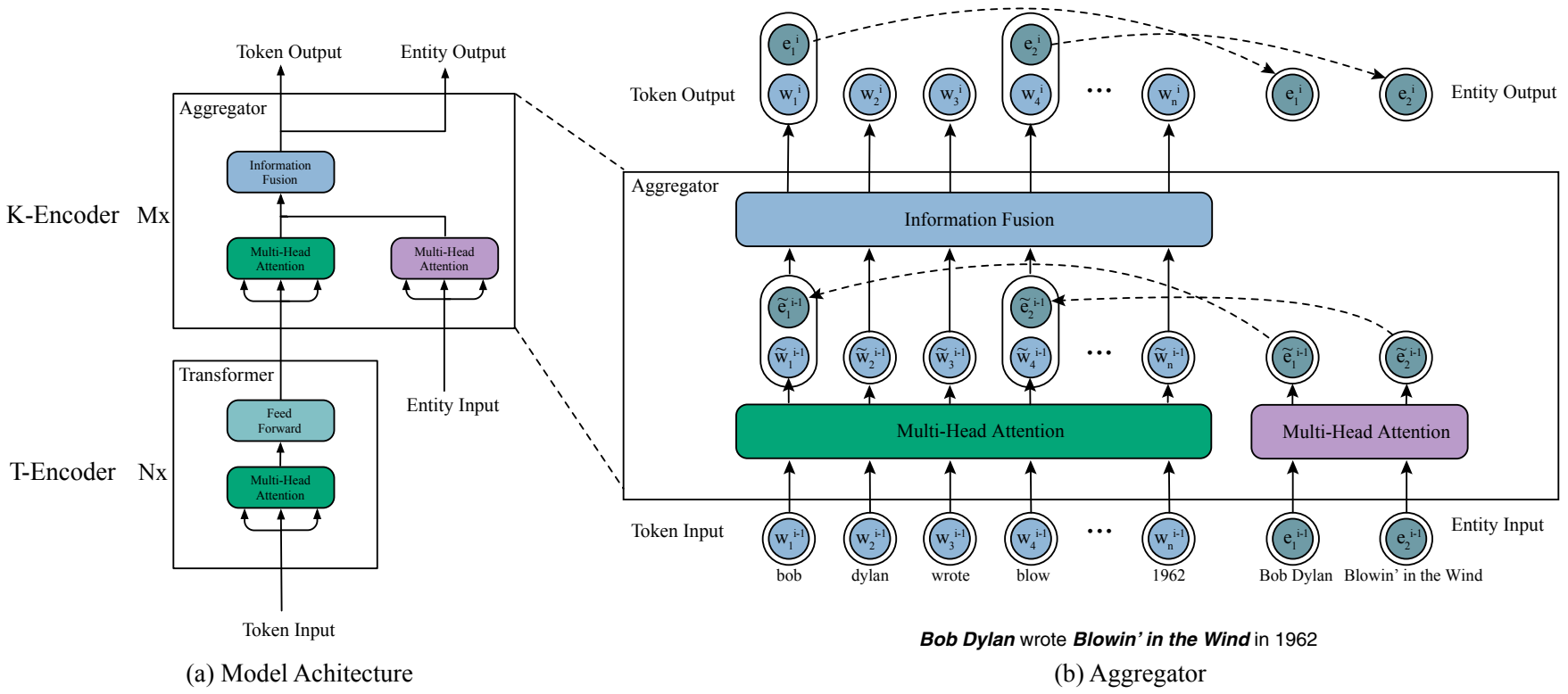

**Fig. 7.21** The architecture of ERNIE model

Similar to BERT, ERNIE adopts the masked language model and the next sentence prediction as the pretraining objectives. Besides, for the better fusion of textual and knowledge features, ERNIE uses a new pretraining objective (denoising entity auto-encoder) by randomly masking some of the named entity alignments in the input text and training to select appropriate entities from KGs to complete the alignments. Unlike the existing pre-trained language representation models only utilizing local context to predict tokens, these objectives require ERNIE to aggregate both context and knowledge facts for predicting both tokens and entities, and lead to a knowledgeable language representation model.

Figure 7.21 is the overall architecture. The left part shows that ERNIE consists of two encoders (T-Encoder and K-Encoder), where T-Encoder is stacked by several classical transformer layers and K-Encoder is stacked by the new aggregator layers designed for knowledge integration. The right part is the detail of the aggregator layer. In the aggregator layer, the input token embeddings and entity embeddings from the preceding aggregator are fed into two multi-head self-attention, respectively. Then, the aggregator adopts an information fusion layer for the mutual integration of the token and entity sequence and computes the output embedding for each token and entity.

ERNIE explores how to incorporate knowledge information into language representation models. The experimental results demonstrate that ERNIE has more powerful abilities of both denoising distantly supervised data and fine-tuning on limited data than BERT.

#### 7.4.4.3 KALM

Pre-trained language models can do many tasks without supervised training data, like reading comprehension, summarization, and translation [60]. However, traditional language models are unable to efficiently model entity names observed in text. To solve this problem, Liu et al. [42] propose a new language model architecture, called Knowledge-Augmented Language Model (KALM), to use the entity types of words for better language modeling.

KALM is a language model with the option to generate words from a set of entities from a knowledge database. An individual word can either come from a general word dictionary as in the traditional language model or be generated as a name of an entity from a knowledge database. The training objectives just supervise the output and ignore the decision of the word type. Entities in the knowledge database are partitioned by type and they use the database to build the types of words. According to the context observed so far, the model decides whether the word is a general term or a named entity in a given type. Thus, KALM learns to predict whether the context observed is indicative of a named entity and what tokens are likely to be entities of a given type.

With the language modeling, KALM learns a named entity recognizer without any explicit supervision by using only plain text and the potential types of words. And, it achieves a comparable performance with the state-of-the-art supervised methods.

### *7.4.5 Other Knowledge-Guided Applications*

Knowledge enables AI agents to understand, infer, and address user demands, which is essential in most knowledge-driven applications like information retrieval, question answering, and dialogue system. The behavior of AI agents will be more reasonable and accurate with the favor of knowledge representations. In the following subsections, we will introduce the great improvements made by knowledge representation in question answering.

#### 7.4.5.1 Knowledge-Guided Question Answering

Question answering aims to give correct answers according to users' questions, which needs the capabilities of both natural language understanding of questions and inference on answer selection. Therefore, combining knowledge with question answering is a straightforward application for knowledge representations. Most conventional question answering systems directly utilize knowledge graphs as certain databases, ignoring the latent relationships between entities and relations. Recently, with the thriving in deep learning, explorations have focused on neural models for understanding questions and even generating answers.

Considering the flexibility and diversity of generated answers in natural languages, Yin et al. [93] propose a neural Generative Question Answering model (GENQA), which explores on generating answers to simple factoid questions in natural languages. Figure 7.22 demonstrates the workflow of GENQA. First, a bidirectional RNN is regarded as the Interpreter to transform question $q$ from natural language to compressed representation $\mathbf{H}_q$. Next, Enquirer takes $\mathbf{H}_Q$ as the key to rank rel-evant triples facts of $q$ in knowledge graphs and retrieves possible entities in $\mathbf{r}_q$. Finally, Answerer combines $\mathbf{H}_q$ and $\mathbf{r}_q$ to generate answers in the form of natural languages. Similar to [1], at each step, Answerer first decides whether to generate common words or knowledge words according to a logistic regression model. For common words, Answerer acts in the same way as RNN decoders with $\mathbf{H}_q$ selected by attention-based methods. As for knowledge words, Answerer directly generates entities with higher ranks.

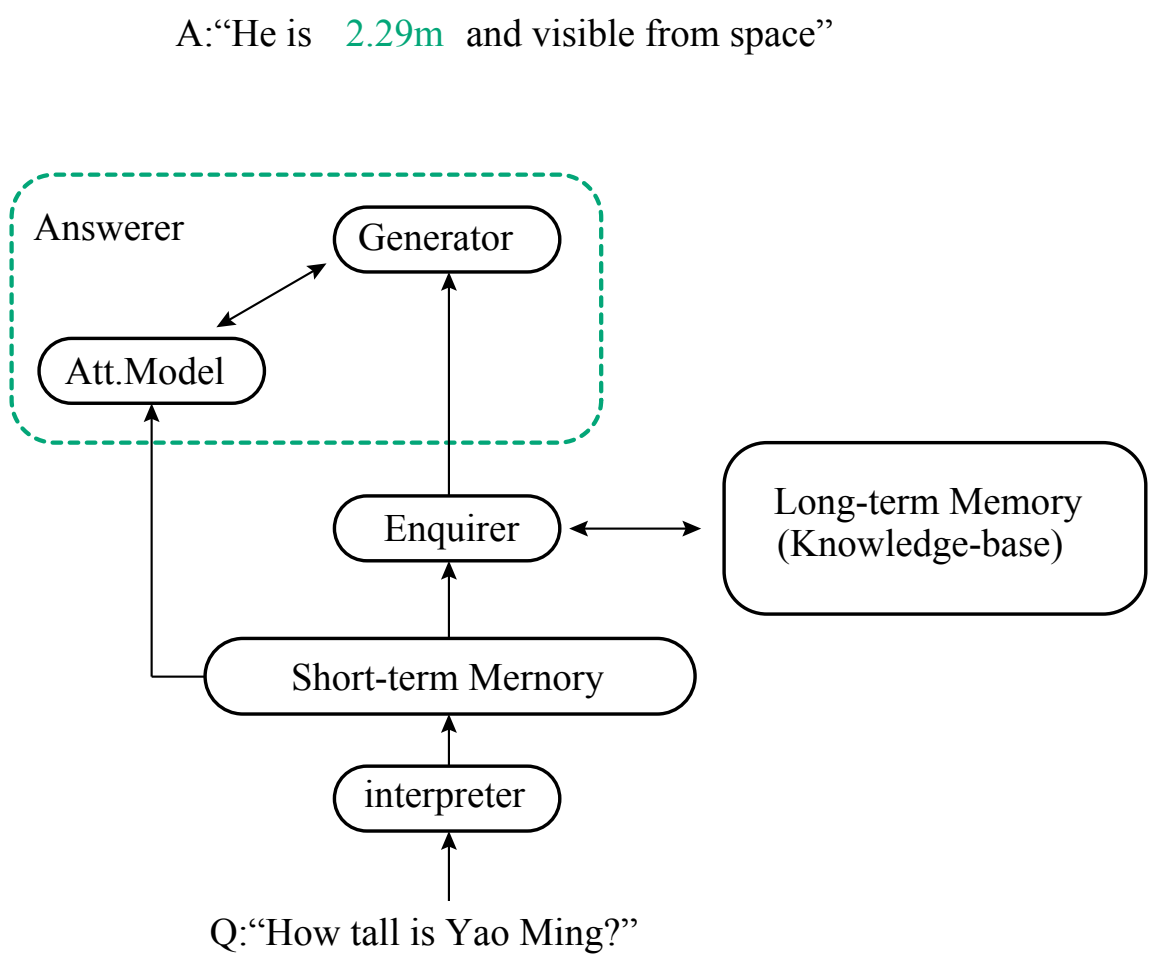

**Fig. 7.22** The architecture of GENQA model

There are gradually more efforts focusing on encoding knowledge representations into knowledge-driven tasks like information retrieval and dialogue systems. However, how to flexibly and effectively combine knowledge with AI agents remains to be explored in the future.

#### 7.4.5.2 Knowledge-Guided Recommendation System

Due to the rapid growth of web information, recommendation systems have been playing an essential role in the web application. The recommendation system aims to predict the "rating" or "preference" that users may give to items. And since KGs can provide rich information, including both structured and unstructured data, recommendation systems have utilized more and more knowledge from KGs to enrich their contexts.

Cheekula et al. [11] explore to utilize the hierarchical knowledge from the DBpedia category structure in the recommendation system and employs the spreading activation algorithm to identify entities of interest to the user. Besides, Passant [56] measures the semantic relatedness of the artist entity in a KG to build music recommendation systems. However, most of these systems mainly investigate the problem by leveraging the structure of KGs. Recently, with the development of representation learning, [98] proposes to jointly learn the latent representations in a collaborative filtering recommendation system as well as entities' representations in KGs.

Except the tasks stated above, there are gradually more efforts focusing on encoding knowledge graph representations into other tasks such as dialogue system [37, 103], entity disambiguation [20, 31], knowledge graph alignment [12, 102], dependency parsing [35], etc. Moreover, the idea of KRL has also motivated the research on visual relation extraction [2, 99] and social relation extraction [71].

## 7.5 Summary

In this chapter, we first introduce the concept of the knowledge graph. Knowledge graph contains both entities and the relationships among them in the form of triple facts, providing an effective way of human beings learning and understanding the real world. Next, we introduce the motivations of knowledge graph representation, which is considered as a useful and convenient method for a large amount of data and is widely explored and utilized in multiple knowledge-based tasks and significantly improves the performance. And we describe existing approaches for knowledge graph representation. Further, we discuss several advanced approaches that aim to deal with the current challenges of knowledge graph representation. We also review the real-world applications of knowledge graph representation such as language modeling, question answering, information retrieval, and recommendation systems.

For further understanding of knowledge graph representation, you can find more related papers in this paper list https://github.com/thunlp/KRLPapers. There are also some recommended surveys and books including:

- Bengio et al. Representation learning: A review and new perspectives [4].
- Liu et al. Knowledge representation learning: A review [47].
- Nickel et al. A review of relational machine learning for knowledge graphs [52].
- Wang et al. Knowledge graph embedding: A survey of approaches and applications [74].
- Ji et al. A survey on knowledge graphs: representation, acquisition and applications [34].

In the future, for better knowledge graph representation, there are some directions requiring further efforts:

(1) **Utilizing More Knowledge**. Current KRL approaches focus on representing triple-based knowledge from world knowledge graphs such as Freebase, Wikidata, etc. In fact, there are various kinds of knowledge in the real world such as factual knowledge, event knowledge, commonsense knowledge, etc. What's more, the knowledge is stored with different formats, such as attributions, quantifier, text, and so on. The researchers have formed a consensus that utilizing more knowledge is a potential way toward more interpretable and intelligent NLP. Some existing works [44, 82] have made some preliminary attempts of utilizing more knowledge in KRL. Beyond these works, is it possible to represent different knowledge in a unified semantic space, which can be easily applied in downstream NLP tasks?

(2) **Performing Deep Fusion** of knowledge and language. There is no doubt that the joint learning of knowledge and language information can further benefit downstream NLP tasks. Existing works [76, 89, 97] have preliminarily verified the effectiveness of joint learning. Recently, ERINE [100] and KnowBERT [57] further provide us a novel perspective to fuse knowledge and language in pretraining. Soares et al. [64] learn the relational similarity in text with the guidance of KGs, which is also a pioneer of knowledge fusion. Besides designing novel pretraining objectives, we could also design novel model architectures for downstream tasks, which are more suitable to utilize KRL, such as memory-based models [48, 91] and graph network-based models [66]. Nevertheless, it still remains an unsolved problem for effectively performing the deep fusion of knowledge and language.

(3) **Orienting Heterogeneous Modalities**. With the fast development of the World Wide Web, the data size of audios, images, and videos on the Web have become larger and larger, which are also important resources for KRL besides texts. Some pioneer works [51, 81] explore to learn knowledge representations on a multi-modal knowledge graph, but are still preliminary attempts. Intuitively, audio and visual knowledge can provide complementary information, which benefits related NLP tasks. To the best of our knowledge, there still lacks research on applying multi-modal KRL in downstream tasks. How to efficiently and effectively integrate multi-modal knowledge is becoming a critical and challenging problem for KRL.

(4) **Exploring Knowledge Reasoning**. Most of the existing KRL methods represent knowledge information in low-dimensional semantic space, which is feasible for the computation of complex knowledge graphs in neural-based NLP models. Although benefiting from the usability of low-dimensional embeddings, KRL cannot perform explainable reasoning such as symbolic rules, which is of great importance for downstream NLP tasks. Recently, there has been increasing interest in the combination of embedding methods and symbolic reasoning methods [26, 59], aiming at taking both advantages of them. Beyond these works, there remain lots of unsolved problems for developing better knowledge reasoning ability for KRL.

# Chapter 8
# Network Representation

**Abstract** Network representation learning aims to embed the vertexes in a network into low-dimensional dense representations, in which similar vertices in the network should have "close" representations (usually measured by cosine similarity or Euclidean distance of their representations). The representations can be used as the feature of vertices and applied to many network study tasks. In this chapter, we will introduce network representation learning algorithms in the past decade. Then we will talk about their extensions when applied to various real-world networks. Finally, we will introduce some common evaluation tasks of network representation learning and relevant datasets.

## 8.1 Introduction

As a natural way to represent objects and their relationships, the network is ubiquitous in our daily lives. The rapid development of social networks like Facebook and Twitter encourage researchers to design effective and efficient algorithms on network structure. A key problem of network study is how to represent the network information properly. Traditional representations of networks are usually high dimensional and sparse, which becomes a weakness when people apply statistical learning to networks. With the development of machine learning, feature learning of vertices in a network is becoming an emerging task. Therefore, network representation learning algorithms turn network information into low-dimensional dense real-valued vectors, which can be used as input for existing machine learning algorithms. For example, the representations of vertices can be fed to a classifier like Support Vector Machine (SVM) for the vertex classification task. Also, the representations can be used for visualization by taking the representations as points in Euclidean space. In this section, we will formalize the network representation learning problem.

Denote a network as $G = (V, E)$ where $V$ is the vertex set and $E$ is the edge set. An edge $e = (v_i, v_j) \in E$ where $v_i, v_j \in V$ is a directed edge from vertex $v_i$ to $v_j$. The outdegree of vertex $v_i$ is defined as $\deg_O(v_i) = |\{v_j|(v_i, v_j) \in E\}|$. Similarly, the indegree of vertex $v_i$ is $\deg_I(v_i) = |\{v_j|(v_j, v_i) \in E\}|$. For undirected network, we have $\deg(v_i) = \deg_O(v_i) = \deg_I(v_i)$. Taking social network as an example, a vertex represents a user and an edge represents the friendship between two users. The indegree and outdegree represent the number of followers and followees of a user, respectively.

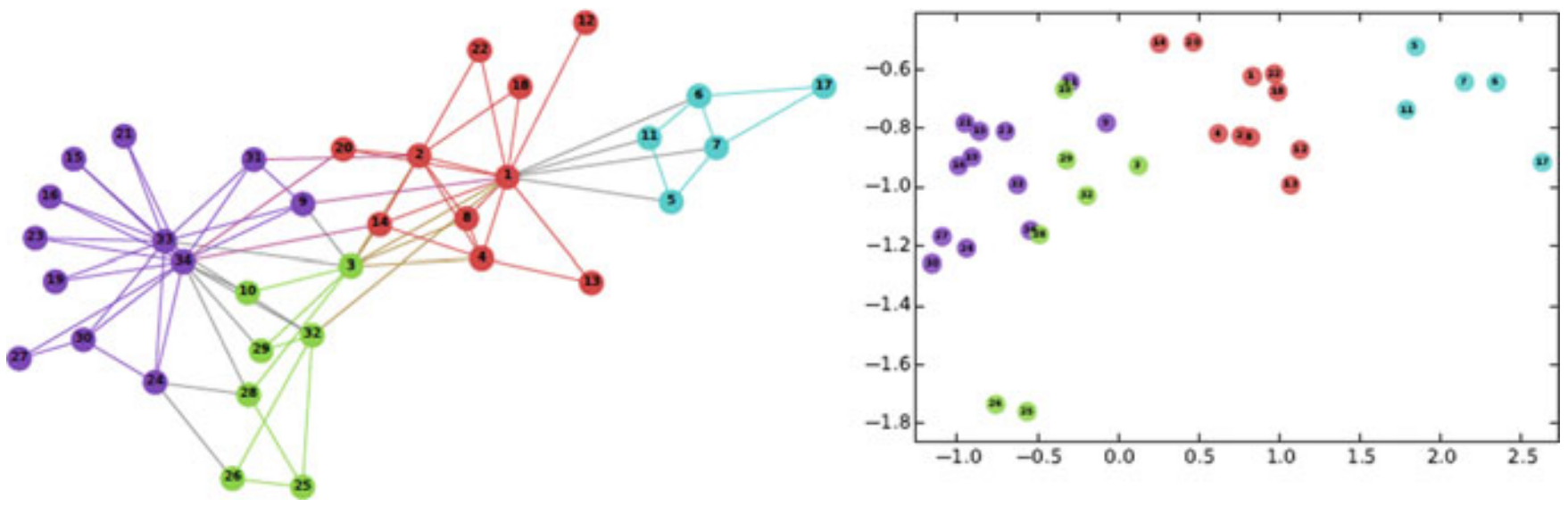

**Fig. 8.1** A visualization of vertex embeddings learned by DeepWalk model [93]

Adjacency matrix $A \in \mathbb{R}^{|V|\times|V|}$ is a matrix where $A_{ij} = 1$ if $(v_i, v_j) \in E$ and $A_{ij} = 0$ otherwise. We can easily generalize adjacency matrix to weighted network by setting $A_{ij}$ to the weight of edge $(v_i, v_j)$. The adjacency matrix is a simple and straightforward representation of the network. Each row of adjacency matrix $A$ denotes the relationship between a vertex and other vertices and can be seen as the representation of the corresponding vertex.

Though convenient and straightforward, the representation of the adjacency matrix suffers from the scalability problem. Adjacency matrix $A$ takes $|V| \times |V|$ space to store, and it is usually unacceptable when $|V|$ grows large. Also, the adjacency matrix is very sparse, which means most of its entries are zeros. The data sparsity makes discrete algorithms applicable, but it is still hard to develop efficient algorithms for statistic learning [93].

Therefore, people come up with the idea to learn low-dimensional dense representations for vertices in a network. Formally, the goal of network representation learning is to learn a real-valued vector $\mathbf{v} \in \mathbb{R}^d$ for vertex $v \in V$ where dimension $d$ is much smaller than the number of vertices $|V|$. The idea is that similar vertices should have close representations as shown in Fig. 8.1. Network representation learning can be unsupervised or semi-supervised. The representations are automatically learned without feature engineering and can be further used for specific tasks like classifications once they are learned. These representations are low dimensional, which enables efficient algorithms to be designed over the representations without considering the network structure itself. We will discuss more details about the evaluation of network representations later in this chapter.

## 8.2 Network Representation

In this section, we will introduce several kinds of network representation learning algorithms in detail.

### 8.2.1 Spectral Clustering Based Methods

Spectral clustering based methods are a group of algorithms that compute first $k$ eigenvectors or singular vectors of an affinity matrix, such as adjacency or Laplacian matrix of the network. These methods depend heavily on the construction of the affinity matrix. The evaluation result of different affinity matrices varies a lot. Generally speaking, spectral clustering based methods have a high complexity because the computations of eigenvectors and singular vectors have a nonlinear time complexity.

On the other hand, spectral clustering based methods need to save an affinity matrix in the memory during the computation. Thus the space complexity cannot be ignored, either. These disadvantages limit the large-scale and online generalization of these methods. Now we will present several algorithms based on spectral clustering.

Locally Linear Embedding (LLE) [98] assumes that the representations of vertices are sampled from a manifold. More specifically, LLE supposes that the representations of a vertex and its neighbors lie in a locally linear patch of the manifold. That is to say, a vertex's representation can be approximated by a linear combination of the representation of its neighbors. LLE uses the linear combination of neighbors to reconstruct the center vertex. Formally, the reconstruction error of all vertices can be expressed as

$$\mathscr{L}(\mathbf{W}, \mathbf{V}) = \sum_{i=1}^{|V|} \left\| \mathbf{v}_i - \sum_{j=1}^{|V|} \mathbf{W}_{ij} \mathbf{v}_j \right\|^2, \tag{8.1}$$

where $\mathbf{V} \in \mathbb{R}^{|V| \times d}$ is the vertex embedding matrix and $\mathbf{W}_{ij}$ is the contribution coefficient of vertex $v_j$ to $v_i$. LLE enforces $\mathbf{W}_{ij} = 0$ if $v_i$ and $v_j$ are not connected, i.e., $(v_i, v_j) \notin E$. Further, the summation of a row of matrix $\mathbf{W}$ is set to 1, i.e., $\sum_{j=1}^{|V|} \mathbf{W}_{ij} = 1$.

Equation 8.1 is solved by alternatively optimizing weight matrix $\mathbf{W}$ and representation $\mathbf{V}$. The optimization over $\mathbf{W}$ can be solved as a least-squares problem. The optimization over representation $\mathbf{V}$ leads to the following optimization problem:

$$\mathscr{L}(\mathbf{W}, \mathbf{V}) = \sum_{i=1}^{|V|} \left\| \mathbf{v}_i - \sum_{j=1}^{|V|} \mathbf{W}_{ij} \mathbf{v}_j \right\|^2, \tag{8.2}$$

$$s.t. \sum_{i=1}^{|V|} \mathbf{v}_i = \mathbf{0}, \tag{8.3}$$

$$\text{and } |V|^{-1} \sum_{i=1}^{|V|} \mathbf{v}_i^\top \mathbf{v}_i = \mathbf{I}_d, \tag{8.4}$$

where $\mathbf{I}_d$ denotes $d \times d$ identity matrix. The conditions Eqs. 8.3 and 8.4 ensure the uniqueness of the solution. The first condition enforces the center of all vertex embeddings to zero point and the second condition guarantees different coordinates have the same scale, i.e., equal contribution to the reconstruction error.

The optimization problem can be formulated as the computation of eigenvectors of matrix $(\mathbf{I}_{|V|} - \mathbf{W}^\top)(\mathbf{I}_{|V|} - \mathbf{W})$, which is an easily solvable eigenvalue problem. More details can be found in the note [22].

Laplacian Eigenmap [8] algorithm simply follows the idea that the representations of two connected vertices should be close. Specifically, the "closeness" is measured by the square of Euclidean distance. We use $D$ to denote diagonal degree matrix where $D$ is a $|V| \times |V|$ diagonal matrix and the $i$th diagonal entry $D_{ii}$ is the degree of vertex $v_i$. The Laplacian matrix $L$ of a graph is defined as the difference of diagonal matrix $D$ and adjacency matrix $A$, i.e., $L = D - A$.

Laplacian Eigenmap algorithm wants to minimize the following cost function:

$$\mathscr{L}(\mathbf{V}) = \sum_{\{i,j|(v_i,v_j)\in E\}} \|\mathbf{v}_i - \mathbf{v}_j\|^2, \tag{8.5}$$

$$s.t. \mathbf{V}^\top D\mathbf{V} = \mathbf{I}_d. \tag{8.6}$$

The cost function is the summation of square loss of all connected vertex pairs and the condition prevents the trivial all-zero solution caused by arbitrary scale. Equation 8.5 can be reformulated in matrix form as

$$\mathbf{V}^* = \arg \min_{\mathbf{V}^\top D\mathbf{V}=\mathbf{I}_d} \operatorname{tr}(\mathbf{V}^\top L\mathbf{V}). \tag{8.7}$$

Algebraic knowledge tells us that the optimal solution $\mathbf{V}^*$ of Eq. 8.7 is the corresponding eigenvectors of $d$ smallest nonzero eigenvalues of Laplacian matrix $L$. Note that the Laplacian Eigenmap algorithm can be easily generalized to the weighted graph.

Both LLE and Laplacian Eigenmap have a symmetric cost function which indicates that both algorithms cannot be applied to the directed graph. Directed Graph Embedding (DGE) [17] was proposed to generalize Laplacian Eigenmap.

For both directed and undirected graph, we can define a transition probability matrix $\mathbf{P} \in \mathbb{R}^{|V|\times|V|}$, where $\mathbf{P}_{ij}$ denotes the probability that vertex $v_i$ walks to $v_j$.

**Table 8.1** Applicability of LLE, Laplacian Eigenmap, and DGE algorithms on undirected, weighted, and directed graph

| Algorithm | Capability | | |
|---|---|---|---|
| | Undirected | Weighted | Directed |
| LLE | ✓ | – | – |
| Laplacian Eigenmap | ✓ | ✓ | – |
| DGE | ✓ | ✓ | ✓ |

The transition matrix defines a Markov random walk through the graph. We denote the stationary value of vertex $v_i$ as $\pi_i$ where $\sum_i \pi_i = 1$. The stationary distribution of random walk is commonly used in many ranking algorithms such as PageRank. DGE designs a new cost function which emphasizes the important vertices, which have a higher stationary value:

$$\mathscr{L}(\mathbf{V}) = \sum_{i=1}^{|V|} \pi_i \sum_{j=1}^{|V|} \mathbf{P}_{ij} \|\mathbf{v}_i - \mathbf{v}_j\|^2. \tag{8.8}$$

By denoting $\mathbf{M} = \text{diag}(\pi_1, \pi_2, \ldots, \pi_{|V|})$, the cost function Eq. 8.8 can be reformulated as

$$\mathscr{L}(\mathbf{V}) = 2\text{tr}(\mathbf{V}^\top \mathbf{B} \mathbf{V}), \tag{8.9}$$

$$s.t.\ \mathbf{V}^\top \mathbf{M} \mathbf{V} = \mathbf{I}_d, \tag{8.10}$$

where

$$\mathbf{B} = \mathbf{M} - \frac{\mathbf{M}\mathbf{P} - \mathbf{P}^\top \mathbf{M}}{2}. \tag{8.11}$$

The condition Eq. 8.10 is added to remove an arbitrary scaling factor. Similar to Laplacian Eigenmap, the optimization problem can also be solved as a generalized eigenvector problem.

For comparisons between the above three network embedding learning algorithms, we conclude the following table to illustrate their applicability (Table 8.1).

Unlike previous works which minimize the distance between vertex representations, Tang and Liu [112] introduces modularity [85] into the cost function instead. Modularity is a measurement which characterizes how far the graph is away from a uniform random graph. Given graph $G = (V, E)$, we assume that vertices $V$ are divided into $k$ nonoverlapping communities. By "uniform random graph", we mean vertices connect to each other based on a uniform distribution given their degrees. Then the expected edges between $v_i$ and $v_j$ is $\frac{\deg(v_i)\deg(\mathrm{v_j})}{2|E|}$. Then the modularity of a graph $Q$ is defined as

$$Q = \frac{1}{2|E|} \sum_{i,j} \left[ A_{ij} - \frac{\deg(v_i)\deg(v_j)}{2|E|} \right] \delta(v_i, v_j), \tag{8.12}$$

where $\delta(v_i, v_j) = 1$ if $v_i$ and $v_j$ belong to the same community and $\delta(v_i, v_j) = 0$ otherwise. A larger modularity indicates that the subgraphs inside communities are denser, which follows the intuition that a community is a dense well-connected cluster. Then the problem is to find a partition that maximizes the modularity $Q$.

However, a hard clustering on modularity maximization is proved to be NP hard. Therefore, they relax the problem to a soft case. Let $\mathbf{d} \in \mathbb{Z}_+^{|V|}$ denotes the degree of all vertices and $\mathbf{1} \in \{0, 1\}^{|V| \times k}$ denotes the community indicator matrix where

$$\mathbf{1}_{ij} = \begin{cases} 1 & \text{if vertex } i \text{ belongs to community } j, \\ 0 & \text{otherwise.} \end{cases} \tag{8.13}$$

Then we define modularity matrix $\mathbf{B}$ as

$$\mathbf{B} = A - \frac{\mathbf{d}\mathbf{d}^T}{2|E|}, \tag{8.14}$$

and modularity $Q$ can be reformulated as

$$Q = \frac{1}{2|E|} \text{tr}(\mathbf{1}^\top \mathbf{B} \mathbf{1}). \tag{8.15}$$

By relaxing $\mathbf{1}$ to a continuous matrix, it has been proved that the optimal solution $\mathbf{1}$ is the top-$k$ eigenvectors of modularity matrix $\mathbf{B}$ [84].

As an alternatively cost function, Tang and Liu also proposed another algorithm [113] by optimizing over normalized cut of the graph. Similarly, the algorithm turns to the computation of top-$k$ eigenvectors of normalized graph Laplacian $\widetilde{L}$:

$$\widetilde{L} = D^{-\frac{1}{2}} L D^{-\frac{1}{2}} = I - D^{-\frac{1}{2}} A D^{-\frac{1}{2}}. \tag{8.16}$$

Then the community indicator matrix $\mathbf{1}$ is taken as a $k$-dimensional vertex representation.

To conclude spectral clustering methods for network representation learning, these methods often define a cost function that is linear or quadratic to the vertex embedding. Then they reformulate the cost function as a matrix form and figure out that the optimal solutions are eigenvectors of a particular matrix according to algebra knowledge. The major drawback of spectral clustering methods is the complexity: the computation of eigenvectors for large-scale matrices is both time consuming and space consuming.

### 8.2.2 DeepWalk

As shown in previous subsections, accurate computation of the optimal solution, such as eigenvector computation, is not very efficient for large-scale problems. Meantime, neural network approaches have proved their effectiveness in many areas such as natural language and image processing. Though the gradient descent method cannot always guarantee an optimal solution of the neural network models, the implementation and learning of neural networks are relatively fast, and they usually have good performances. On the other hand, neural network models can let people get rid of feature engineering and are mostly data driven. Thus, the exploration of the neural network approach on representation learning is becoming an emerging task.

DeepWalk [93] proposes a novel approach that introduces deep learning techniques into network representation learning for the first time. The benefits of modeling truncated random walks instead of the adjacency matrix are twofold: first, random walks need only local information and thus enable discrete and online algorithms on it while modeling of adjacency matrix may need to store everything in memory and thus be space consuming; second, modeling random walks can alleviate the variance and uncertainty of modeling original binary adjacency matrix. We will look insight into DeepWalk in the next subsection.

Unsupervised representation learning algorithms have been widely studied and applied in the natural language processing area. The authors show that the vertex frequency in short random walks also follows the power law as words in documents do. Showing the connection between vertex to the word and random walks to sentences, the authors adapted a well-known word representation learning algorithm word2vec [80] into vertex representation learning. Now, we will introduce DeepWalk algorithms in detail.

Given graph $G = (V, E)$, we denote a random walk started at vertex $v_i$ as $\ell_{v_i}$. We use $\ell_{v_i}^k$ to represent the $k$th vertex in the random walk $\ell_{v_i}$. The next vertex $\ell_{v_i}^{k+1}$ is generated by uniformly random selection from neighbors of vertex $\ell_{v_i}^k$. Random walk sequences have been used for many network analysis tasks, such as similarity measurement and community detection [2, 32].

DeepWalk follows the idea of language modeling to model short random walk sequences. That is to estimate the likelihood of observing vertex $v_i$ given all previous vertices in the random walk:

$$P(v_i|(v_1, v_2, \ldots, v_{i-1})). \tag{8.17}$$

To the extent of vertex representation learning, we turn to predict vertex $v_i$ given the representations of all previous vertices:

$$P(v_i|(\mathbf{v}_1, \mathbf{v}_2, \ldots, \mathbf{v}_{i-1})). \tag{8.18}$$

A relaxation of this formula in language modeling turns to use vertex $v_i$ to predict its neighboring vertices $v_{i-w}, \ldots, v_{i-1}, v_{i+1}, \ldots, v_{i+w}$ where $w$ is the window size.

This part of model is named as Skip-gram model in word embedding learning. The neighboring vertices are also called context vertices of the center vertex. As another simplification, DeepWalk ignores the order and offset of the vertices and thus predict $v_{i-w}$ and $v_{i-1}$ in the same way. The optimization function of a single vertex of a random walk can be formulated as

$$\min_{\mathbf{V}} -\log P(\{v_{i-w}, \ldots, v_{i-1}, v_{i+1}, \ldots, v_{i+w}\}|\mathbf{v}_i). \tag{8.19}$$

Based on independent assumption, the loss function can be rewritten as

$$\min_{\mathbf{V}} \sum_{k=-w, k\neq 0}^{w} -\log P(v_{i+k}|\mathbf{v}_i). \tag{8.20}$$

The overall loss function can be obtained by adding up over every vertex in every random walk.

Now we talk about how to predict a single vertex $v_j$ given center vertex $v_i$. In DeepWalk, each vertex $v_i$ has two representations with the same dimension: vertex representation $\mathbf{v}_i \in \mathbb{R}^d$ and context representation $\mathbf{c}_i \in \mathbb{R}^d$. The probability of prediction $P(v_j|\mathbf{v}_i)$ is defined by a softmax function over all vertices:

$$P(v_j|\mathbf{v}_i) = \frac{\exp(\mathbf{v}_i \mathbf{c}_j^\top)}{\sum_{k=1}^{|V|} \exp(\mathbf{v}_i \mathbf{c}_k^\top)}. \tag{8.21}$$

Here we come to the parameter learning phase of DeepWalk. We first present the pseudocode of the DeepWalk framework in Algorithm 8.1.

**Algorithm 8.1** DeepWalk algorithm

Given graph $G = (V, E)$, window size $w$, embedding size $d$, walks per vertex $n$ and walk length $l$
**for** $i = 1, 2, \ldots, n$ **do**
  **for** $v_i \in V$ **do**
    $\ell_{v_i}$ =RandomWalk$(G, v_i, l)$
    Skip-gram$(\mathbf{V}, \ell_{v_i}, w)$
  **end for**
**end for**

where RandomWalk$(G, v_i, l)$ generates a random walk rooted at $v_i$ with length $l$ and Skip-gram$(\mathbf{V}, \ell_{v_i}, w)$ function is defined in Algorithm 8.2, where $\alpha_l$ is the learning rate of stochastic gradient descent.

Note that the parameter updating rule $\mathbf{V} = \mathbf{V} - \alpha_l \frac{\partial J}{\partial \mathbf{V}}$ in Skip-gram has a complexity of $O(|V|)$ because in the computation of the gradient of $P(v_k|\mathbf{v}_j)$ (as shown in Eq. 8.21), the denominator has $|V|$ terms to compute. This complexity is unacceptable for large-scale networks.

**Algorithm 8.2** Skip-gram($R$, $W_{v_i}$, $w$)

**for** $v_j \in \ell_{v_i}$ **do**
  **for** $v_k \in \ell_{v_i}[j-w : j+w]$ **do**
    **if** $v_k \neq v_j$ **then**
      $J(\mathbf{V}) = -\log P(v_k|\mathbf{V}_j)$
      $\mathbf{V} = \mathbf{V} - \alpha_l \frac{\partial J}{\partial \mathbf{V}}$
    **end if**
  **end for**
**end for**

**Table 8.2** Analogy of DeepWalk and word2vec

| Method | Object | Input | Output |
|---|---|---|---|
| Word2vec | Word | Sentence | Word embedding |
| DeepWalk | Vertex | Random walk | Vertex embedding |

To address this problem, people proposed Hierarchical Softmax as a variant of original softmax function. The core idea is to map the vertices to a balanced binary tree, where each vertex corresponds to a leaf of the tree. Then the prediction of a vertex turns to the prediction of the path from the root to the corresponding leaf. Assume that the path from root to vertex $v_k$ is denoted by a sequence of tree nodes $b_1, b_2 \ldots, b_{\lceil \log |V| \rceil}$ and then we have

$$\log P(v_k|\mathbf{v}_j) = \sum_{i=1}^{\lceil \log |V| \rceil} \log P(b_i|\mathbf{v}_j). \tag{8.22}$$

A logistic function can easily implement a binary decision on a tree node. Hence, the time complexity reduces to $O(\log |V|)$ from $O(|V|)$. We can accelerate the algorithm by using Huffman coding to map frequent vertices to the tree nodes that are close to the root. We can also use negative sampling which is used in word2vec to replace hierarchical softmax for speeding up.

So far, we have finished the introduction of the DeepWalk algorithm. DeepWalk introduces efficient deep learning techniques into network embedding learning. Table 8.2 gives an analogy between DeepWalk and Word2vec. DeepWalk outperforms traditional network representation learning methods on network classification tasks and is also efficient for large-scale networks. Besides, the generation of random walks can be generalized to nonrandom walk, such as the information propagation streams. In the next subsection, we will give a detailed proof to demonstrate the correlation between DeepWalk and matrix factorization.

#### 8.2.2.1 Matrix Factorization Comprehension of DeepWalk

Perozzi et al. introduced the Skip-gram model into the study of social network for the first time, and designed an algorithm named DeepWalk [93] for learning vertex representation on a graph. In this subsection, we prove that the DeepWalk algorithm with Skip-gram and softmax model is actually factoring a matrix $\mathbf{M}$ where each entry $\mathbf{M}_{ij}$ is the logarithm of the average probability that vertex $v_i$ randomly walks to vertex $v_j$ in fix steps. We will explain it later.

Since the Skip-gram model does not consider the offset of context vertex and predict context vertices independently, we can regard the random walks as a set of vertex-context pairs. The useful information on random walks is the co-occurrence of vertex pairs inside a window. Given network $G = (V, E)$, we suppose that vertex-context set $D$ is generated from random walks, where each piece of $D$ is a vertex-context pair $(v, c)$. Let $V$ be the set of nodes, and $V_C$ be the set of context nodes. In most cases, $V = V_C$.

Consider a vertex-context pair $(v, c)$:

$N_{(v,c)}$ denotes the number of times $(v, c)$ appears in $D$. $N_v = \sum_{c' \in V_C} N_{(v,c')}$ and $N_c = \sum_{v' \in V} N_{(v',c)}$ denotes the number of times $v$ and $c$ appears in $D$. Note that $|D| = \sum_{v' \in V} \sum_{c' \in V_C} N_{(v',c')}$.

A context vertex $c \in V_C$ is represented by a $d$-dimension vector $\mathbf{c} \in \mathbb{R}^d$ and $\mathbf{C}$ is a $|V_C| \times d$ matrix, where row $j$ is vector $\mathbf{c_j}$. Our goal is to figure out a matrix $\mathbf{M} = \mathbf{V}\mathbf{C}^\top$.

Perozzi et al. implemented the DeepWalk algorithm with the Skip-gram and Hierarchical Softmax model. Note that Hierarchical Softmax is a variant of softmax for speeding the training time. In this subsection, we give proofs for both negative sampling and softmax with the Skip-gram model.

Negative sampling approximately maximizes the probability of softmax function by randomly choosing $k$ negative samples from the context set. Levy and Goldberg showed that Skip-gram with the Negative Sampling model (SGNS) is implicitly factorizing a word-context matrix [69] by assuming that dimensionality $d$ is sufficiently large. In other words, we can assign each product $\mathbf{v} \cdot \mathbf{c}$ a value independent of the others.

In SGNS model, we have

$$P((v, c) \in D) = \text{Sigmoid}(\mathbf{v} \cdot \mathbf{c}) = \frac{1}{1 + e^{-\mathbf{v} \cdot \mathbf{c}}}. \tag{8.23}$$

Suppose we choose $k$ negative samples for each vertex-context pair $(v, c)$ according to the distribution $P_D(c_N) = \frac{N_{c_N}}{|D|}$. Then, the objective function for SGNS can be written as

$$\begin{aligned}\mathscr{O} &= \sum_{v\in V}\sum_{c\in V_C} N_{(v,c)}(\log \text{Sigmoid}(\mathbf{v}\cdot\mathbf{c}) + k\mathbb{E}_{c_N\sim P_D}[\log \text{Sigmoid}(-\mathbf{v}\cdot\mathbf{c})]) \\ &= \sum_{v\in V}\sum_{c\in V_C} N_{(v,c)} \log \text{Sigmoid}(\mathbf{v}\cdot\mathbf{c}) + k\sum_{v\in V} N_v \sum_{c_N\in V_C} \frac{N_{c_N}}{|D|} \log \text{Sigmoid}(-\mathbf{v}\cdot\mathbf{c}) \\ &= \sum_{v\in V}\sum_{c\in V_C} N_{(v,c)} \log \text{Sigmoid}(\mathbf{v}\cdot\mathbf{c}) + kN_v \frac{N_c}{|D|} \log \text{Sigmoid}(-\mathbf{v}\cdot\mathbf{c}).\end{aligned} \tag{8.24}$$

Denote $x = \mathbf{v}\cdot\mathbf{c}$. By solving $\frac{\partial\mathscr{O}}{\partial x} = 0$, we have

$$\mathbf{v}\cdot\mathbf{c} = x = \log\frac{N_{(v,c)}|D|}{N_v N_c} - \log k. \tag{8.25}$$

Thus we have $\mathbf{M}_{ij} = \log \frac{\frac{N_{(v_i,c_j)}}{|D|}}{\frac{N_{v_i}}{|D|}\frac{N_{c_j}}{|D|}} - \log k$. $\mathbf{M}_{ij}$ can be interpreted as Point-wise Mutual Information(PMI) of vertex-context pair $(v_i, c_j)$ shifted by $\log k$.

Since both negative sampling and hierarchical softmax are variants of softmax, we pay more attention to the softmax model and give a further discussion on it. We also assume that the values of $\mathbf{v}\cdot\mathbf{c}$ are independent.

In softmax model,

$$P((v,c)\in D) = \frac{e^{\mathbf{v}\cdot\mathbf{c}}}{\sum_{c'\in V_C} e^{\mathbf{v}\cdot\mathbf{c}'}}. \tag{8.26}$$

And the objective function is

$$\mathscr{O} = \sum_{v\in V}\sum_{c\in V_C} N_{(v,c)} \log \frac{e^{\mathbf{v}\cdot\mathbf{c}}}{\sum_{c'\in V_C} e^{\mathbf{v}\cdot\mathbf{c}'}}. \tag{8.27}$$

After extracting all terms associated to $\mathbf{v}\cdot\mathbf{c}$ as $\mathscr{O}(v,c)$, we have

$$\mathscr{O}(v,c) = N_{(v,c)} \log \frac{e^{\mathbf{v}\cdot\mathbf{c}}}{\sum_{c'\in V_C, c'\neq c} e^{\mathbf{v}\cdot\mathbf{c}'} + e^{\mathbf{v}\cdot\mathbf{c}}} + \sum_{\tilde{c}\in V_C, \tilde{c}\neq c} N_{(v,\tilde{c})} \log \frac{e^{\mathbf{v}\cdot\tilde{\mathbf{c}}}}{\sum_{c'\in V_C, c'\neq c} e^{\mathbf{v}\cdot\mathbf{c}'} + e^{\mathbf{v}\cdot\mathbf{c}}}. \tag{8.28}$$

Note that $\mathscr{O} = \frac{1}{|V_C|}\sum_{v\in V}\sum_{c\in V_C}\mathscr{O}(v,c)$. Denote $x = \mathbf{v}\cdot\mathbf{c}$. By solving $\frac{\partial\mathscr{O}}{\partial x} = 0$ for all such x, we have

$$\mathbf{v}\cdot\mathbf{c} = x = \log\frac{N_{(v,c)}}{N_v} + b_v, \tag{8.29}$$

where $b_v$ can be any real constant since it will be canceled when we compute $P((v,c)\in D)$. Thus, we have $\mathbf{M}_{ij} = \log\frac{N_{(v_i,c_j)}}{N_{(v_i)}} + b_{v_i}$. We will discuss what $\mathbf{M}_{ij}$ represents in next section.

It is clear that the method of sampling vertex-context pairs, i.e., random walks generation, will affect matrix $\mathbf{M}$. In this section, we will discuss $\frac{N_v}{|D|}$, $\frac{N_c}{|D|}$ and $\frac{N_{(v,c)}}{N_v}$ based on an ideal sampling method for DeepWalk algorithm.

Assume the graph is connected and undirected, and the window size is $w$. The sampling algorithm is illustrated in Algorithm 8.3. We can easily generalize this sampling method to the directed graph by only adding $(RW_i, RW_j)$ into $D$.

**Algorithm 8.3** Ideal vertex-context pair sampling algorithm

Generate an infinite long random walk $\ell$.
Denote $\ell_i$ as the vertex on position i of $\ell$, where $i = 0, 1, 2, \ldots$
**for** $i = 0, 1, 2, \ldots$ **do**
  **for** $j \in [i + 1, i + w]$ **do**
    add $(\ell_i, \ell_j)$ into $D$
    add $(\ell_j, \ell_i)$ into $D$
  **end for**
**end for**

Each appearance of vertex $i$ will be recorded $2w$ times in $D$ for undirected graph and $w$ times for directed graph. Thus, we can figure out that $\frac{N_{v_i}}{|D|}$ is the frequency of $v_i$ that appears in the random walk, which is exactly the PageRank value of $v_i$. Also note that $\frac{N_{(v_i,v_j)}}{N_{v_i}/2w}$ is the expectation times that $v_j$ is observed in left/right $w$ neighbors of $v_i$.

Denote the transition matrix in PageRank algorithm be $\mathbf{P}$. More formally, let $\deg(v_i)$ be the degree of vertex $i$. $\mathbf{P}_{ij} = \frac{1}{\deg(v_i)}$ if $(i, j) \in E$ and $\mathbf{P}_{ij} = 0$ otherwise. We use $\mathbf{e}_i$ to denote a $|V|$-dimension row vector, where all entries are zero except the $i$th entry is 1.

Suppose that we start a random walk from vertex $i$ and use $\mathbf{e}_i$ to denote the initial state. Then $\mathbf{e}_i\mathbf{P}$ is the distribution over all the vertices where $j$th entry is the probability that vertex $v_i$ walks to vertex $v_j$. Hence, $j$th entry of $\mathbf{e}_i\mathbf{P}^w$ is the probability that vertex $v_i$ walks to vertex $v_j$ at exactly $w$ steps. Thus $[\mathbf{e}_i(\mathbf{P} + \mathbf{P}^2 + \cdots + \mathbf{P}^w)]_j$ is the expectation times that $v_j$ appears in right $w$ neighbors of $v_i$.

Hence

$$
\begin{aligned}
\frac{N_{(v_i,v_j)}}{N_{v_i}/2w} &= 2[\mathbf{e}_i(\mathbf{P} + \mathbf{P}^2 + \cdots + \mathbf{P}^w)]_j, \\
\frac{N_{(v_i,v_j)}}{N_{v_i}} &= \frac{[\mathbf{e}_i(\mathbf{P} + \mathbf{P}^2 + \cdots + \mathbf{P}^w)]_j}{w}.
\end{aligned}
\tag{8.30}
$$

This equality also holds for a directed graph.

By setting $b_{v_i} = \log 2w$ for all $i$, $\mathbf{M}_{ij} = \log \frac{N_{(v_i,v_j)}}{N_{v_i}/2w}$ is logarithm of the expectation times that $v_j$ appears in left/right $w$ neighbors of $v_i$.

By setting $b_{v_i} = 0$ for all $i$, $\mathbf{M}_{ij} = \log \frac{N_{(v_i,v_j)}}{N_{v_i}} = \log \frac{[\mathbf{e}_i(A+A^2+\cdots+A^w)]_j}{w}$ is logarithm of the average probability that vertex $v_i$ randomly walks to vertex $v_j$ in $w$ steps.

#### 8.2.2.2 Discussion

So far we have seen many different network representation learning algorithms and we can figure out some patterns that how network representation methods share. Then we will move forward and see how these patterns match some recent network embedding algorithms.

Most network representation algorithms try to reconstruct a data matrix generated from the graph with vertex embeddings. The simplest matrix would be the adjacency matrix. However, recovering the adjacency matrix may not be the best choice. First, real-world networks are mostly very sparse which means $O(|E|) = O(|V|)$. Therefore, the adjacency matrix will be very sparse as well. Though the sparseness enables an efficient algorithm, it can harm the performance of vertex representation learning because of the deficiency of useful information. Second, the adjacency matrix may be noisy and sensitive. A single missing link can completely change the correlation between two vertices.

Hence people seek to find an alternative matrix to replace the adjacency matrix though implicitly. Take DeepWalk as an example, DeepWalk models the following matrix based on matrix factorization comprehension of DeepWalk:

$$\mathbf{M} = \mathbf{P} + \mathbf{P}^2 + \cdots + \mathbf{P}^w, \tag{8.31}$$

where

$$\mathbf{P}_{ij} = \begin{cases} 1/\deg(v_i) & \text{if } (v_i, v_j) \in E, \\ 0 & \text{otherwise.} \end{cases} \tag{8.32}$$

Compared with the adjacency matrix $A$, the matrix $\mathbf{M}$ modeled by DeepWalk is much denser. Furthermore, the window size parameter $w$ can adjust the density: a larger window size models a denser matrix but will slow down the algorithm. Hence, the window size $w$ works as a harmonic factor to balance efficiency and effectiveness. On the other hand, the matrix $\mathbf{M}$ can alleviate the noises in the adjacency matrix. Consider two similar vertices $v_i$ and $v_j$, even though the edge between them is missing, they can still have many co-occurrences by appearing inside a window size of the same random walks.

In a real-world application, direct computation of $\mathbf{M}$ may have a high time complexity when window size $w$ grows. Thus, it is essential to choose a proper $w$. However, window size $w$ is a discrete parameter, and thus the matrix $M$ may grow from too sparse to too dense by changing $w$ by 1. Here, we can see another benefit of random walks. Random walks used by DeepWalk serve as Monte Carlo simulations for approximating matrix $\mathbf{M}$. The more random walks you walk, the more likely you can approximate the matrix.

After we choose a matrix to model, we need to correlate the matrix entry with vertex representations pairs. There are two widely used measurements of vertices pairs: Euclidean distance and inner product. Assume that we want to model the entry $M_{ij}$ given vertex representations $\mathbf{v}_i$ and $\mathbf{v}_j$, we can employ

$$\mathbf{M}_{ij} = f(\|\mathbf{v}_i - \mathbf{v}_j\|_2),$$
$$\mathbf{M}_{ij} = f(\mathbf{v}_i \cdot \mathbf{v}_j), \quad (8.33)$$

where function $f$ can be any reasonable matching functions such as sigmoid function or linear function for our propose. Actually, the inner product $\mathbf{v}_i \cdot \mathbf{v}_j$ is used more widely and would correspond to equivalent matrix factorization methods.

The next phase is to design a proper loss function between $\mathbf{M}_{ij}$ and $f(\mathbf{v}_i \cdot \mathbf{v}_j)$. Several loss functions such as square loss and hinge loss can be employed. You can also design a generative model and maximize the likelihood of matrix $\mathbf{M}$.

The final step of a network representation learning algorithm would be parameter learning. The most frequently used parameter learning method would be Stochastic Gradient Descent (SGD). Other variants of SGD such as AdaGrad and AdaDelta can make the learning phase converge faster. In the next subsection, we will see some recent network representation learning algorithms which follow DeepWalk. We will find that their models can match all these phases above and have some innovations on building matrix $\mathbf{M}$, modifying function $f$, and changing loss function.

## *8.2.3 Matrix Factorization Based Methods*

We will focus on two network representation learning algorithms LINE and GraRep [13, 111] in this subsection. They both follow the framework introduced in the last subsection.

### 8.2.3.1 LINE

Tang et al. [111] proposed a network embedding model named as LINE. LINE algorithm can handle large-scale networks with arbitrary types: (un)directed or weighted. To model the interaction between vertices, LINE models first-order proximity which is represented by observed links and second-order proximity which is determined by shared neighbors but not links between vertices.

Before we introduce the details of the algorithm, we can move one step back and see how the idea works. The modeling of first-order proximity, i.e., observed links, is the modeling of the adjacency matrix. As we said in the last subsection, the adjacency matrix is usually too sparse. Hence the modeling of second-order proximity, i.e., vertices with shared neighbors, can serve as complement information to enrich the adjacency matrix and make it denser. The enumeration of all vertex pairs which have common neighbors is time consuming. Thus, it is necessary to design a sampling phase to handle large-scale networks. The sampling phase works like Monte Carlo simulation to approximate the ideal matrix.

Now we only have two questions: how to define first-order and second-order proximity and how to define the loss function. In other words, it is equal to how to define $\mathbf{M}$ and loss function.

First-order proximity between vertex $u$ and $v$ is defined as the weight $w_{uv}$ on edge $(u, v)$. If there is no edge between vertex $u$ and $v$, then the first-order proximity between them is 0.

Second-order proximity between vertex $u$ and $v$ is defined as the similarity between their neighborhood network. Let $p_u = (w_{u,1}, \ldots, w_{u,|V|})$ denote the first-order proximity between vertex $u$ and all other vertices. Then the second-order proximity between $u$ and $v$ is defined as the similarity of $p_u$ and $p_v$. If they have no shared neighbors, then the second-order proximity is zero.

Then we can introduce LINE model more specifically. The joint probability between $v_i$ and $v_j$ is

$$p_1(v_i, v_j) = \frac{1}{1 + \exp(-\mathbf{v}_i \cdot \mathbf{v}_j)}, \tag{8.34}$$

where $\mathbf{v}_i$ and $\mathbf{v}_j$ are $d$-dimensional row vectors which indicate the representations of vertex $v_i$ and $v_j$.

To supervise the probabilities, empirical probability is defined as $\hat{p}_1(i, j) = \frac{w_{ij}}{W}$, where $W = \sum_{(v_i, v_j) \in E} w_{ij}$. Thus our goal is to find vertex embeddings to approximate $\frac{w_{ij}}{W}$ with $\frac{1}{1+\exp(-\mathbf{v}_i \cdot \mathbf{v}_j)}$. Following the idea in last subsection, it is equivalent to say $\mathbf{v}_i \cdot \mathbf{v}_j = \mathbf{M}_{ij} = -\log(\frac{W}{w_{ij}} - 1)$.

The loss function between joint probability $p_1$ and its empirical probability $\hat{p}_1$ is

$$\mathscr{L}_1 = D_{\mathrm{KL}}(\hat{p}_1 \,||\, p_1), \tag{8.35}$$

where $D_{\mathrm{KL}}(\cdot \,||\, \cdot)$ is KL-divergence of two probability distributions.

On the other hand, we define the probability that vertex $v_j$ appears in $v_i$'s context:

$$p_2(v_j | v_i) = \frac{\exp(\mathbf{c}_j \cdot \mathbf{v}_i)}{\sum_{k=1}^{|V|} \exp(\mathbf{c}_k \cdot \mathbf{v}_i)}. \tag{8.36}$$

Similarly, the empirical probability is defined as $\hat{p}_2(v_j | v_i) = \frac{w_{ij}}{d_i}$ where $d_i = \sum_k w_{ik}$ and the loss function is

$$\mathscr{L}_2 = \sum_i d_i D_{\mathrm{KL}}(\hat{p}_2(\cdot, v_i) \,||\, p_2(\cdot, v_i)). \tag{8.37}$$

The first-order and second-order proximity embeddings are trained separately, and we concatenate the embeddings together after the training phase as vertex representations.

#### 8.2.3.2 GraRep

Now we turn to another network representation learning algorithm, GraRep, which directly follows the proof of matrix factorization form of DeepWalk. Recall that we prove DeepWalk is actually factorizing a matrix $\mathbf{M}$ where $\mathbf{M} = \log \frac{A+A^2+\cdots+A^w}{w}$. GraRep algorithm can be divided into 3 steps:

- Get $k$-step transition probability matrix $A^k$ for each $k = 1, 2, \ldots, K$.
- Get each $k$-step representation.
- Concatenate all $k$-step representations.

GraRep uses a simple idea, i.e., SVD decomposition on $A^k$, in the second step to get embeddings. As $K$ gets large, the matrix $\mathbf{M}$ gets denser and thus outputs a better representation. However, this algorithm is not very efficient especially when $K$ gets large.

### *8.2.4 Structural Deep Network Methods*

Different from previous methods that use a shallow neural network model to characterize the network representations, Structural Deep Network Embedding (SDNE) [125] employs the deeper neural model to model the nonlinearity between vertex embeddings. As shown in Fig. 8.2, the whole model can be divided into two parts: (1) the first part is supervised by Laplacian Eigenmaps, which models the first-order proximity; (2) the second part is unsupervised deep neural autoencoder which characterizes the second-order proximity. Finally, the algorithm takes the intermediate layer which is used for the supervised part as the network representation.

First, we will give a brief introduction to deep neural autoencoder. A neural autoencoder requires that the output vector should be as similar to the input vector. Generally speaking, the output cannot be the same with the input vector because the dimension of intermediate layers of the autoencoder is much smaller than that of the input and output layer. That is to say, a deep autoencoder first compresses the input into a low-dimensional intermediate vector and then tries to reconstruct the original input vector from the low-dimensional intermediate vector. Once the deep autoencoder is trained, we can say that the intermediate layer is an excellent low-dimensional representation of the original inputs since we can recover the input vector from it.

More formally, we assume the input vector is $\mathbf{x}_i$. Then the hidden representation of each layer is defined as

$$
\begin{aligned}
\mathbf{y}_i^{(1)} &= \text{Sigmoid}(\mathbf{W}^{(1)}\mathbf{x}_i + \mathbf{b}^{(1)}), \\
\mathbf{y}_i^{(k)} &= \text{Sigmoid}(\mathbf{W}^{(k)}\mathbf{y}_i^{(k-1)} + \mathbf{b}^{(k)}), k = 2, 3 \ldots,
\end{aligned}
\tag{8.38}
$$

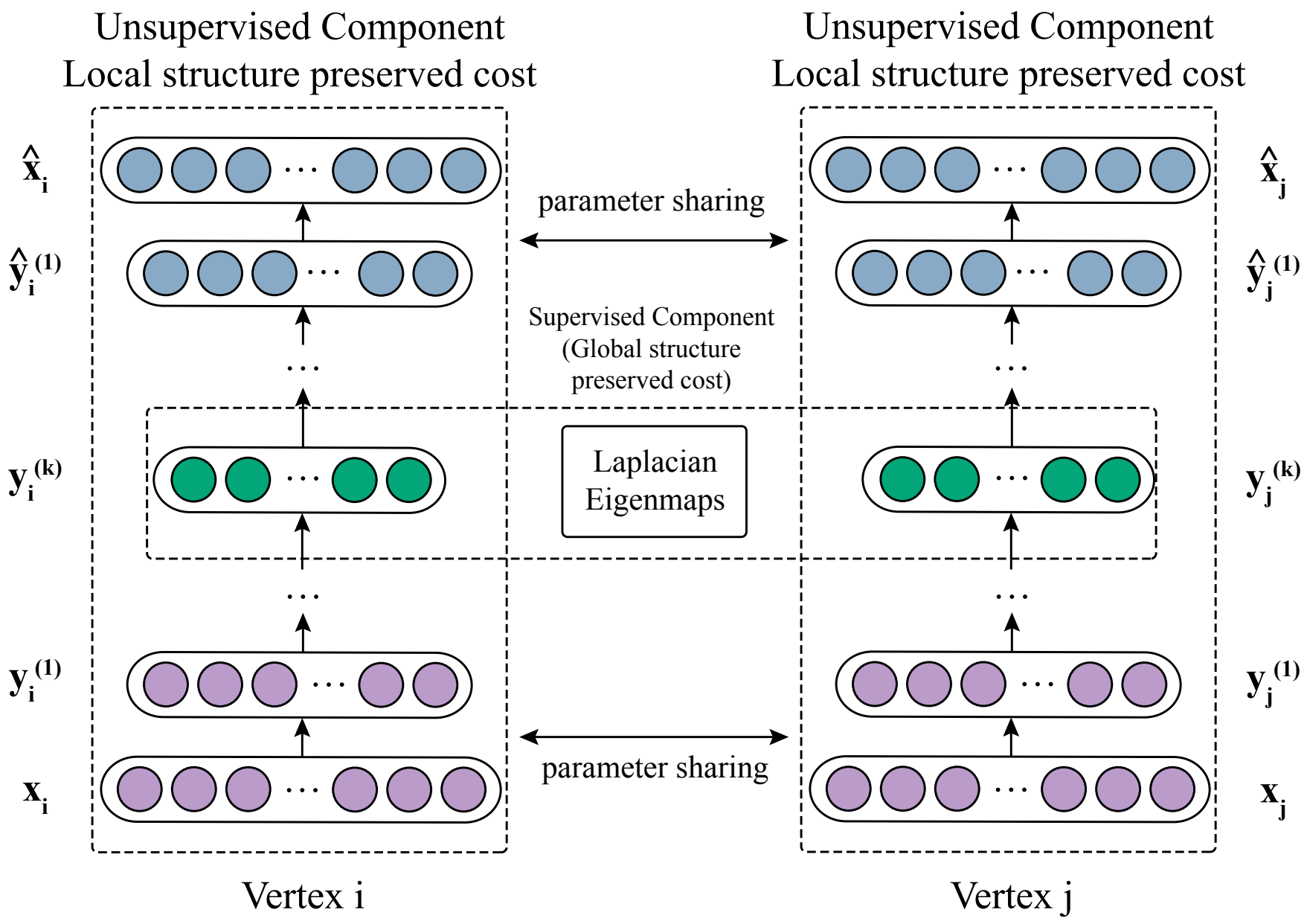

**Fig. 8.2** The architecture of structural deep network embedding model

where $\mathbf{W}^{(k)}$ and $\mathbf{b}^{(k)}$ are weighted matrix and bias vector of $k$th layer. We assume that the hidden representation of the $K$th layer has the minimum dimension. After obtaining $\mathbf{y}_i^{(K)}$, we can get the output $\hat{\mathbf{x}}_i$ by reversing the calculation process. Then the optimization objective of autoencoder is to minimize the difference between input vector $\mathbf{x}_i$ and output vector $\hat{\mathbf{x}}_i$:

$$\mathscr{L}(\mathbf{W}, \mathbf{b}) = \sum_{i=1}^{n} \|\hat{\mathbf{x}}_i - \mathbf{x}_i\|^2, \tag{8.39}$$

where $n$ is the number of input instances.

Back to the network representation problem, SDNE applies the autoencoder to every vertex. The input vector $\mathbf{x}_i$ of each vertex $v_i$ is defined as follows: if vertex $v_i$ and $v_j$ are connected, then the $j$th entry $\mathbf{x}_{ij} > 0$, otherwise $\mathbf{x}_{ij} = 0$. For unweighed graph, if vertex $(v_i, v_j) \in E$, $\mathbf{x}_{ij} = 1$. Then the intermediate layer $\mathbf{y}_i^{(K)}$ can be seen as the low-dimension representation of vertex $v_i$. Also note that there are much more zero entries in input vectors than positive entries due to the sparity of real-world network. Therefore, the loss of positive entries should be emphasized. Therefore, the final optimization objective of second proximity modeling can be written as

$$\mathscr{L}_{2nd} = \sum_{i=1}^{|V|} \|(\hat{\mathbf{x}}_i - \mathbf{x}_i) \odot \mathbf{b}_i\|^2, \tag{8.40}$$

where $\odot$ denotes element-wise multiplication and $\mathbf{b}_{ij} = 1$ if $\mathbf{x}_{ij} = 0$ while $\mathbf{b}_{ij} = \beta > 1$ if $\mathbf{x}_{ij} > 0$.

We have introduced the unsupervised part modeled by deep autoencoder. Now we turn to the supervised part. The supervised part simply requires that the representation of connected vertices should be close to each other. Thus, the loss function of this part is

$$\mathscr{L}_{1st} = \sum_{i,j=1}^{|V|} \mathbf{x}_{ij} \|\mathbf{y}_i^{(K)} - \mathbf{y}_j^{(K)}\|^2. \tag{8.41}$$

Finally, the overall loss function included regularization term is

$$\mathscr{L} = \mathscr{L}_{2nd} + \alpha \mathscr{L}_{1st} + \lambda \mathscr{L}_{reg}, \tag{8.42}$$

where $\alpha$ and $\lambda$ are harmonic hyperparameter and regularization loss $\mathscr{L}_{reg}$ is the sum of the square of all parameters. The model can be optimized by back-propagation in a standard neural network way. After the training process, $\mathbf{y}_i^{(K)}$ is taken as the representation of vertex $v_i$.

### *8.2.5 Extensions*

#### 8.2.5.1 Network Representation with Internal Information

**Asymmetric Transitivity Preserving Network Representation**. Existing network representation learning algorithms mostly focus on an undirected graph. Most of the methods cannot handle the directed graph well because they do not accurately characterize the asymmetric property. High-Order Proximity preserved Embedding (HOPE) [89] is proposed to preserve high-order proximities of large-scale graphs and capture the asymmetric transitivity. The algorithm further derives a general formulation that covers multiple popular high-order proximity measurements and provides an approximate algorithm with an upper bound of RMSE (Root Mean Squared Error).

Network embedding assumes that the more and the shorter paths from $v_i$ to $v_j$, the more similar should be their representation vectors. In particular, the algorithm assigns two vectors, i.e., source and target vectors for each vertex. We denote adjacency matrix as $A$ and the user representations as $\mathbf{U} = [\mathbf{U}^s, \mathbf{U}^t]$, where $\mathbf{U}^s \in \mathbb{R}^{|V|\times d}$ and $\mathbf{U}^t \in \mathbb{R}^{|V|\times d}$ are source and target vertex embeddings, respectively. We define a high-order proximity matrix as $\mathbf{S}$, where $\mathbf{S}_{ij}$ is the proximity between $v_i$ and $v_j$. Then our goal is to approximate the matrix $\mathbf{S}$ with the product of $\mathbf{U}^s$ and $\mathbf{U}^t$. The optimization objective can be written as

$$\min_{\mathbf{U}^s,\mathbf{U}^t} \|\mathbf{S} - \mathbf{U}^s {\mathbf{U}^t}^\top\|_F^2. \tag{8.43}$$

Many high-order proximity measurements which characterize the asymmetric transitivity share a general formulation which can be used for the approximation of the proximities:

$$\mathbf{S} = \mathbf{M}_g^{-1}\mathbf{M}_l, \tag{8.44}$$

where $\mathbf{M}_g$ and $\mathbf{M}_l$ are both polynomials of matrices. Now we will take three commonly used high-order proximity measurements to illustrate the formula.

- Katz Index Katz Index is a weighted summation over the path set between two vertices. The computation of the Katz Index can be written recurrently:

$$\mathbf{S} := \beta A\mathbf{S} + \beta A, \tag{8.45}$$

  where the decay parameter $\beta$ represents how fast the weight decreases when the length of paths grows.
- Rooted PageRank For rooted PageRank, $\mathbf{S}_{ij}$ is the probability that a random walk from vertex $v_i$ will locate at $v_j$ in the stable state. The formula can be written as

$$\mathbf{S} := \alpha\mathbf{SP} + (1-\alpha)\mathbf{I}, \tag{8.46}$$

  where $\alpha$ is the probability that a random walk returns to its start point and $\mathbf{P}$ is the transition matrix.
- Common Neighbors $\mathbf{S}_{ij}$ is the number of vertexes which is the target of an edge from $v_i$ and the source of an edge to $v_j$. The matrix $\mathbf{S}$ can be expressed as

$$\mathbf{S} = A^2. \tag{8.47}$$

For the three high-order proximity measurements introduced above, we summarize their equivalent form $\mathbf{S} = \mathbf{M}_g^{-1}\mathbf{M}_l$ in the following table (Table 8.3).

A simple idea of approximating $\mathbf{S}$ with the product of matrices is SVD decomposition. However, the direct computation of SVD decomposition of matrix $\mathbf{S}$ has a complexity of $O(|V|^3)$. By writing matrix $\mathbf{S}$ as $\mathbf{M}_g^{-1}\mathbf{M}_l$, we do not need to compute matrix $\mathbf{S}$ directly. Instead, we can do JDGSVD decomposition on $\mathbf{M}_g$ and $\mathbf{M}_l$ independently and then use their results to derive the decomposition of $\mathbf{S}$. The complexity reduces to $|E|d^2$ for each iteration of JDGSVD.

**Community Preserving Network Representation**. While previous methods aim at preserving the microscopic structure of a network such as first- and second-order proximities. Wang et al. [127] proposed Modularized Nonnegative Matrix Factorization (M-NMF), which encodes the mesoscopic community structure information into the network representations. The basic idea is to consider the modularity as part of the optimization function. Recall that the modularity is formulated in Eq. 8.15 and $\mathbf{S}$ is the community indicator matrix. Then the loss function of modularity part is to minimize $-\text{tr}(\mathbf{S}^\top\mathbf{B}\mathbf{S})$.

**Table 8.3** General formula for high-order proximity measurements

| Measurement | $\mathbf{M}_g$ | $\mathbf{M}_l$ |
|---|---|---|
| Katz index | $\mathbf{I} - \beta A$ | $\beta A$ |
| Rooted PageRank | $\mathbf{I} - \alpha\mathbf{P}$ | $(1-\alpha)\mathbf{I}$ |
| Common neighbors | $\mathbf{I}$ | $A^2$ |

Similar to previous methods, M-NMF also factorizes an affinity matrix which encodes first-order and second-order proximities. Specifically, M-NMF takes adjacency matrix $A$ as the first-order proximity matrix $A_1$ and computes the cosine similarity of corresponding rows of adjacency matrix $A$ as the second-order proximity matrix $A_2$. M-NMF uses a mixture of $A_1$ and $A_2$ as the similarity matrix. To conclude, the overall optimization function of M-NMF is

$$\min_{\mathbf{M},\mathbf{U},\mathbf{S},\mathbf{C}} \left\|A_1 + \eta A_2 - \mathbf{M}\mathbf{U}^\top\right\|_F^2 + \alpha \left\|\mathbf{S} - \mathbf{U}\mathbf{C}^\top\right\|_F^2 - \beta\text{tr}(\mathbf{S}^\top\mathbf{B}\mathbf{S}), \tag{8.48}$$

where $\mathbf{S} \in \mathbb{R}^{|V|\times k}$, $\mathbf{M}, \mathbf{U}\in\mathbb{R}^{|V|\times m}$, $\mathbf{C}\in\mathbb{R}^{k\times m}$, $\mathbf{M}_{ij}, \mathbf{U}_{ij}, \mathbf{S}_{ij}, \mathbf{C}_{ij} \geq 0, \forall i \forall j, \text{tr}(\mathbf{S}^\top\mathbf{S}) = |V|$ and $\alpha, \beta, \eta > 0$ are harmonic hyperparameters. Subscript $F$ denotes Frobenius norm. Here similarity matrix $A_1 + \eta A_2$ is factorized into two nonnegative matrices $\mathbf{M}$ and $\mathbf{U}$. Then community representation matrix $\mathbf{C}$ in the second term bridges the matrix factorization part and the modularity part.

A concurrent algorithm Community-enhanced NRL (CNRL) [116, 117] is a pipeline algorithm that learns node-community assignment at first and then reforms the DeepWalk algorithm to incorporate community information. Specifically, in the first phase, CNRL made an analogy between community detection and topic modeling. Then CNRL started by generating random walks and fed these vertex sequences into Latent Dirichlet Allocation (LDA) algorithm. By taking a vertex as a word and a topic as a community, CNRL can get a soft-assignment of vertex-community membership. Then in the second phase, both the embedding of a center node and the embedding of its community are used to predict the neighborhood vertices in the random walk sequences. The illustration figure is shown in Fig. 8.3.

#### 8.2.5.2 Network Representation with External Information

**Network Representation with Text Information**. We will present the network embedding algorithm TADW, which further generalizes the matrix factorization framework to take advantage of text information. Text-Associated DeepWalk (TADW) [136] incorporates text features of vertices into network representation learning under the framework of matrix factorization. The matrix factorization view of DeepWalk enables the introduction of text information into matrix factorization for network representation learning. Figure 8.4 shows the main idea of TADW: factorize vertex affinity matrix $\mathbf{M} \in \mathbb{R}^{|V|\times|V|}$ into the product of three matrices: $\mathbf{W} \in \mathbb{R}^{k\times|V|}$, $\mathbf{H} \in \mathbb{R}^{k\times f_t}$, and text features $\mathbf{T} \in \mathbb{R}^{f_t\times|V|}$. Then TADW concatenates $\mathbf{W}$ and $\mathbf{HT}$ as $2k$-dimensional representations of vertices.

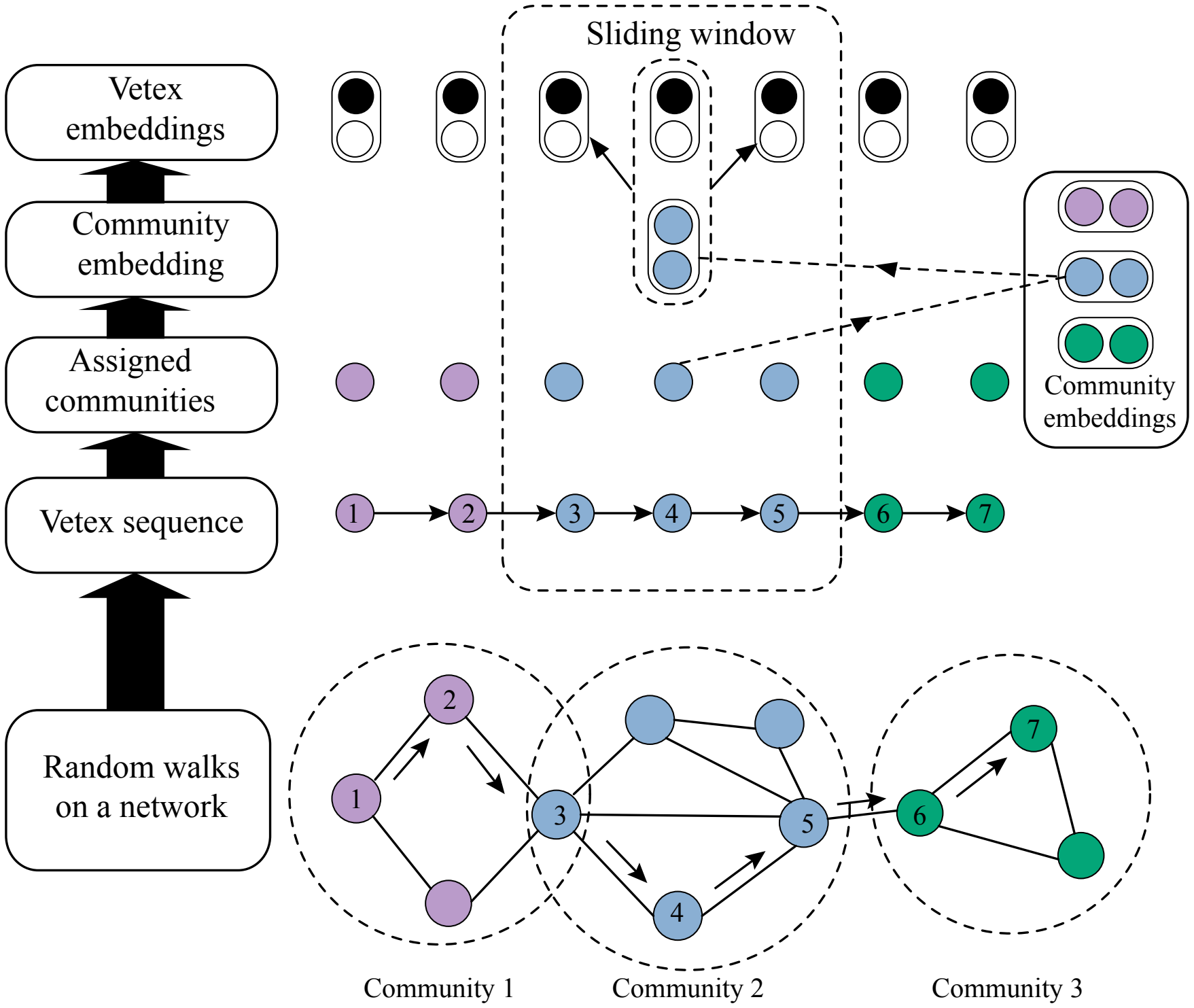

**Fig. 8.3** The architecture of community preserving network embedding model

**Fig. 8.4** The architecture of text-associated DeepWalk model

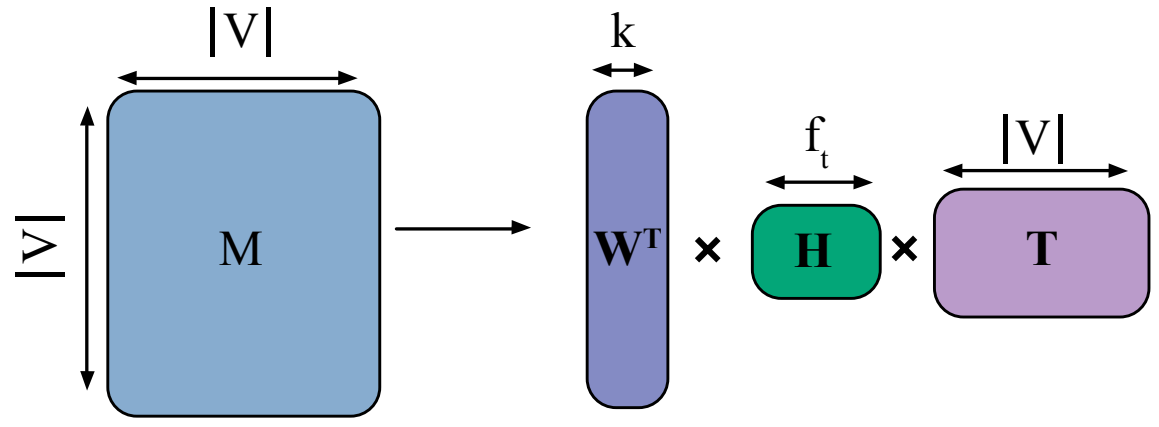

Then the question is how to build vertex affinity matrix **M** and how to extract text feature **T** from the text information. Following the proof of matrix factorization form of DeepWalk, TADW set vertex affinity matrix **M** to a tradeoff between speed and accuracy: factorize the matrix $\mathbf{M} = (A + A^2)/2$ where $A$ is the row-normalized adjacency matrix. For text feature matrix **T**, TADW first constructs the TF-IDF matrix from the text and then reduces the dimension of the TF-IDF matrix to 200 via SVD decomposition.

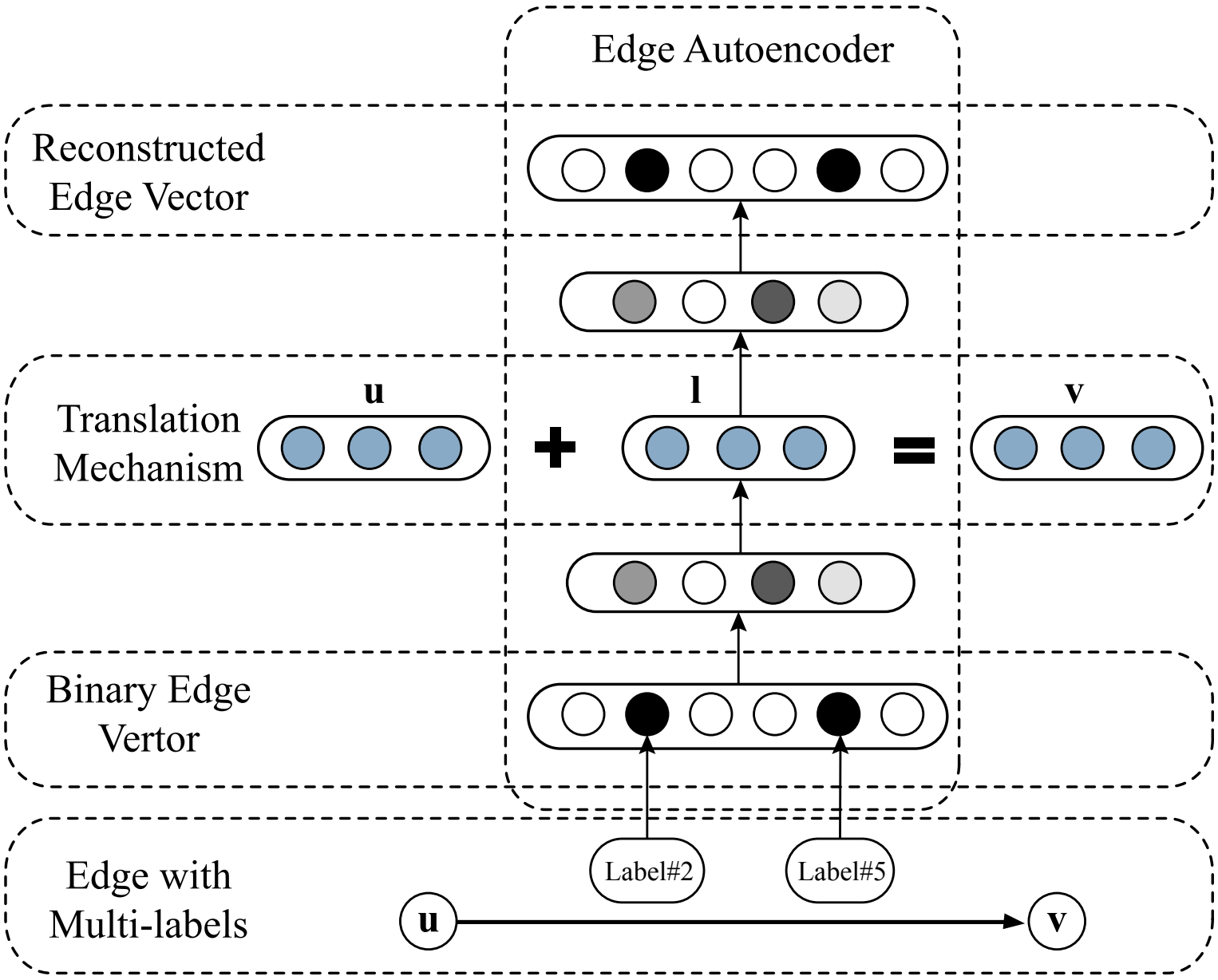

**Fig. 8.5** The architecture of TransNet model

Formally, the model of TADW minimizes the following optimization function:

$$\min_{\mathbf{W},\mathbf{H}} \|\mathbf{M} - \mathbf{W}^\top \mathbf{H}\mathbf{T}\|_F^2 + \frac{\lambda}{2}(\|\mathbf{W}\|_F^2 + \|\mathbf{H}\|_F^2), \tag{8.49}$$

where $\lambda$ is the regularization factor. The optimization of parameters are processed by updating $\mathbf{W}$ and $\mathbf{H}$ iteratively via conjugate gradient descent.

**TransNet**. Most existing NRL methods neglect the semantic information of edges and simplify the edge as a binary or continuous value. TransNet algorithm [119] considers the label information on the edges instead of nodes. In particular, TransNet is based on translation mechanism shown in Fig. 8.5.

In the settings of TransNet, each edge has a number of binary labels on it. Then the loss function of TransNet consists of two parts: one part is the translation loss which measures the distance between $\mathbf{u} + \mathbf{e}$ and $\mathbf{v}$ where $\mathbf{u}$, $\mathbf{e}$, $\mathbf{v}$ stand for the embeddings of head vertex, edge, and tail vertex; another part is the reconstruction loss of the autoencoder which encodes the labels of an edge into its embedding $\mathbf{e}$ and restore the labels from the embedding. After the learning phase, we can compute the edge embedding by subtracting two vertices and use the decoder part of the autoencoder to predict the labels of an unobserved edge.

**Semi-supervised Network Representation**. In this part, we introduce several semi-supervised network representation learning methods that are applied to heterogeneous networks. All methods learn vertex embeddings and their classification labels simultaneously.

(1) **LSHM** The first algorithm LSHM (Latent Space Heterogeneous Model) [52], follows the manifold assumption which assumes that two connected nodes tend to have similar node embeddings. Thus, the regularization loss which forces connected nodes to have similar representations can be formulated as

$$\sum_{i,j} w_{ij} \|\mathbf{v}_i - \mathbf{v}_j\|^2, \tag{8.50}$$

where $w_{ij}$ is the weight of edge $(v_i, v_j)$.

As a semi-supervised representation learning algorithm, LHSM also needs to predict the classification labels for unlabeled vertices. To train the classifiers, LHSM computes the loss of observed labels as

$$\sum_{i} \Delta(f_\theta(\mathbf{v}_i), \mathbf{y}_i), \tag{8.51}$$

where $f_\theta(\mathbf{v}_i)$ is the predicted label for vertex $v_i$, $y_i$ is the observed label for $v_i$ and $\Delta(\cdot, \cdot)$ is the loss function between predicted label and ground truth label. Specifically, $f_\theta(\cdot)$ is a linear function and $\Delta(\cdot, \cdot)$ is set to hinge loss.

Finally, the objective function is

$$\mathscr{L}(\mathbf{V}, \theta) = \sum_{i} \Delta(f_\theta(\mathbf{v}_i), y_i) + \lambda \sum_{i,j} w_{ij} \|\mathbf{v}_i - \mathbf{v}_j\|^2, \tag{8.52}$$

where $\lambda$ is a harmonic hyperparameter. The algorithm is optimized via stochastic gradient descent.

(2) **node2vec** Node2vec [38] modifies DeepWalk by changing the generation of random walks. As shown in previous subsections, DeepWalk generates rooted random walks by choosing the next vertex according to a uniform distribution, which could be improved by using a well-designed random walk generation strategy.

Node2vec first considers two extreme cases of vertex visiting sequences: Breadth-First Search (BFS) and Depth-First Search (DFS). By restricting the search to nearby nodes, BFS characterizes the nearby neighborhoods of center vertices and obtains a microscopic view of the neighborhood of every node. Vertices in the sampled neighborhoods of BFS tend to repeat many times, which can reduce the variance in characterizing the distribution of neighboring vertices of the source node. In contrast, the sampled nodes in DFS reflect a macro-view of the neighborhood which is essential in inferring communities based on homophily.

Node2vec designs a neighborhood sampling strategy which can smoothly interpolate between BFS and DFS. More specifically, consider a random walk that just walks through edge $(t, v)$ and now stays at vertex $v$. The walk evaluates the transition probabilities of edge $(v, x)$ to decide the next step. Node2vec sets the unnormalized transition probability to $\pi_{vx} = \alpha_{pq}(t, x) \cdot w_{vx}$, where

$$\alpha_{pq}(t,x) = \begin{cases} \frac{1}{p} \text{ if } d_{tx} = 0, \\ 1 \text{ if } d_{tx} = 1, \\ \frac{1}{q} \text{ if } d_{tx} = 2, \end{cases} \tag{8.53}$$

and $d_{tx}$ denotes the shortest path distance between vertices $t$ and $x$. $p$ and $q$ are parameters that guide the random walk and control how fast the walk explores and leaves the neighborhood of starting vertex. A low $p$ will increase the probability of revisiting a vertex and make the random walk focus on local neighborhoods while a low $q$ will encourage the random walk to explore further vertices. After the generation of the random walks, the rest of the algorithm is almost the same as that of DeepWalk.

(3) **MMDW** Max-Margin DeepWalk (MMDW) [118] utilizes the max-margin strategy in SVM to generalize DeepWalk algorithm for semi-supervised learning. Specifically, MMDW employs the matrix factorization form of DeepWalk proved in TADW [136] and further add the max-margin constraint which requires that the embeddings of nodes from different labels should be far from each other. The optimization function can be written as

$$\begin{aligned} \min_{\mathbf{X},\mathbf{Y},\mathbf{W},\xi} \mathscr{L} &= \min_{\mathbf{X},\mathbf{Y},\mathbf{W},\xi} \mathscr{L}_{DW} + \frac{1}{2}\|\mathbf{W}\|^2 + C\sum_{i=1}^{T}\xi_i, \\ s.t.\ w_{l_i}^\top x_i &- w_j^\top x_i \geq e_i^j - \xi_i, \forall i, j, \end{aligned} \tag{8.54}$$

where $W = [w_1, w_2, \dots, w_m]^T$ is the weight matrix of SVM, $\xi$ is the slack variables, $e_i^j = 1$ if $l_i \neq j$ and $e_i^j = 0$ otherwise, and $\mathscr{L}_{DW}$ is the matrix factorization form DeepWalk loss function:

$$\mathscr{L}_{DW} = \|\mathbf{M} - \mathbf{X}^\top\mathbf{Y}\|_2^2 + \frac{\lambda}{2}(\|\mathbf{X}\|_2^2 + \|\mathbf{Y}\|_2^2), \tag{8.55}$$

which is introduced in previous sections.

Figure 8.6 shows the visualization result of the DeepWalk and MMDW algorithm on the Wiki dataset [103]. We can see that the embeddings of nodes from different classes are more separable with the help of semi-supervised max-margin representation learning.

(4) **PTE** Another algorithm called PTE (Predictive Text Embedding) [110] focuses on text network such as the bibliography network where a paper is a vertex, and the citation relationship between papers forms the edges. PTE considers network structure together with plain text and observed vertex labels. PTE proposes a semi-supervised framework to learn vertex representation and predict unobserved vertex labels.

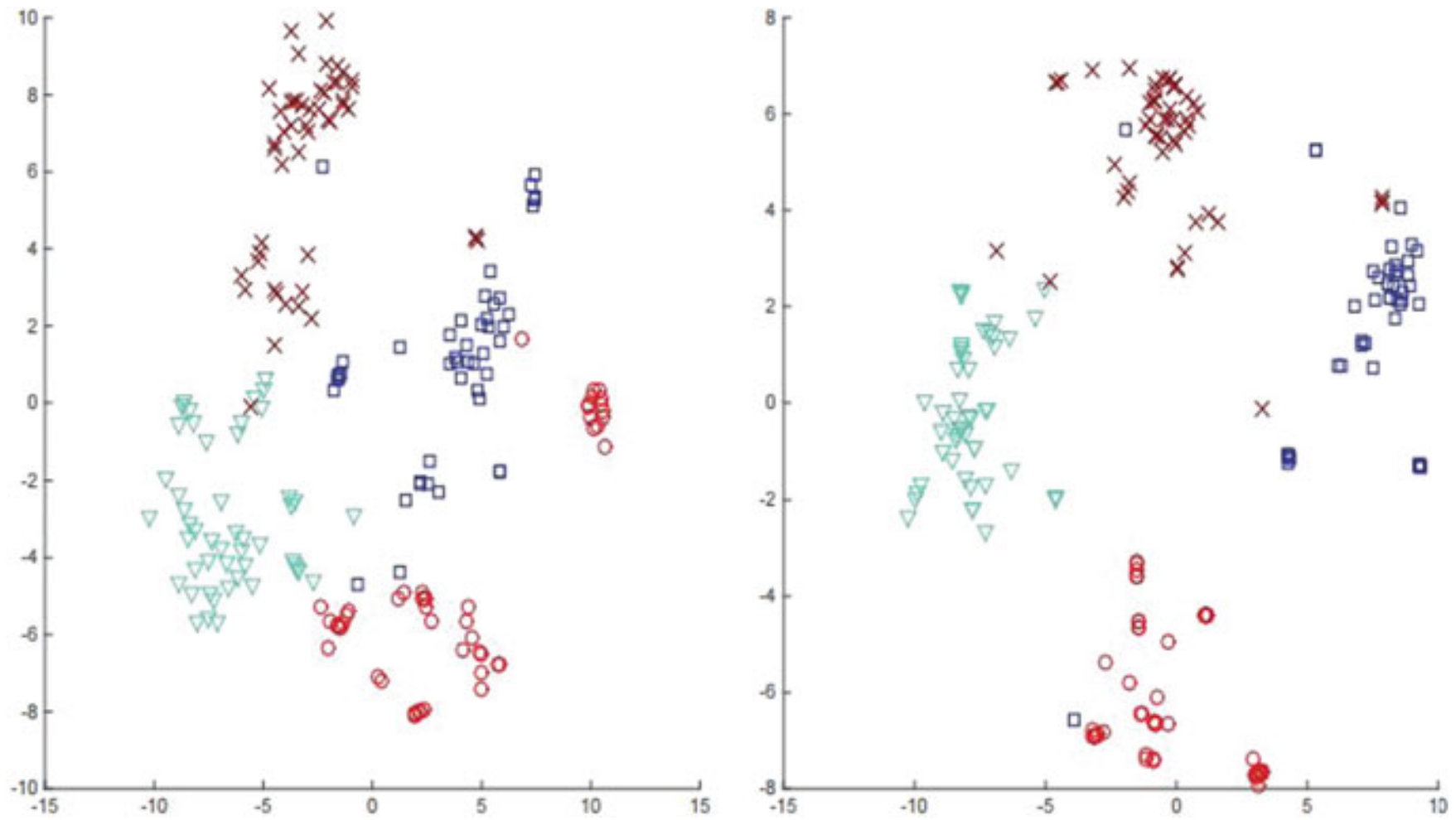

**Fig. 8.6** A visualization of t-SNE 2D representations on Wiki dataset (left: DeepWalk, right: MMDW) [118]

A text network is divided into three bipartite networks: word-word, word-document, and word-label networks. We will introduce the definition of the three networks in more detail.

For the word-word network, the weight $w_{ij}$ of the edge between word $v_i$ and $v_j$ is defined as the number of times that the two words co-occur in the same context windows. For word-document network, the weight $w_{ij}$ between word $v_i$ and document $d_j$ is defined as the number of times $v_i$ appears in document $d_j$. For the word-label network, the weight $w_{ij}$ of the edge between word $v_i$ and class $c_j$ is defined as: $w_{ij} = \sum_{d:l_d=j} n_{di}$, where $n_{di}$ is the term frequency of word $v_i$ in document $d$, and $l_d$ is the class label of document $d$.

Then following previous work LINE, given bipartite network $G = (V_A \cup V_B, E)$, the conditional probability of generating $v_i \in V_A$ from $v_j \in V_B$ is defined as

$$P(v_i|v_j) = \frac{\exp(\mathbf{v}_j \cdot \mathbf{v}_i)}{\sum_{k=1}^{|V|} \exp(\mathbf{v}_k \cdot \mathbf{v}_i)}. \tag{8.56}$$

Similar to LINE model, the loss function is defined as the KL-divergence between empirical distribution and conditional distribution. The optimization objective can be further formulated as

$$\mathscr{L} = -\sum_{(v_i,v_j)\in E} w_{ij} \log P(v_i|v_j). \tag{8.57}$$

Then the final objective can be obtained by summing all three bipartite networks:

$$\mathscr{L}_{pte} = \mathscr{L}_{ww} + \mathscr{L}_{wd} + \mathscr{L}_{wl}, \tag{8.58}$$

where

$$\mathscr{L}_{ww} = -\sum_{(v_i,v_j)\in E_{ww}} w_{ij} \log P(v_i|v_j), \tag{8.59}$$

$$\mathscr{L}_{wd} = -\sum_{(v_i,v_j)\in E_{wd}} w_{ij} \log P(v_i|d_j), \tag{8.60}$$

$$\mathscr{L}_{wl} = -\sum_{(v_i,v_j)\in E_{wl}} w_{ij} \log P(v_i|l_j). \tag{8.61}$$

Then the optimization can be done by stochastic gradient descent.

#### 8.2.5.3 Task-Specific Network Representation

**Network Representation for Community Detection**. As shown in spectral clustering methods, people make their effort to learn community indicator matrix based on modularity and normalized graph cut. The continuous community indicator matrix can be seen as a $k$-dimensional vertex representation, where $k$ is the number of communities. Note that modularity and graph cut is defined for nonoverlapping communities. By alternating a cost function for overlapping communities, the idea can also work for overlapping community detection. In this subsection, we will introduce several community detection algorithms. These community detection algorithms start by learning a $k$-dimensional nonnegative vertex-community affinity matrix and then derive a hard community assignment for vertices based on the matrix. Therefore, the key procedure of these algorithms can be regarded as an unsupervised $k$-dimensional nonnegative vertex embedding learning.

BIGCLAM [140] is an overlapping community detection method. It assumes that matrix $\mathbf{F} \in \mathbb{R}^{|V|\times k}$ is the user-community affinity matrix, where $\mathbf{F}_{vc}$ is the strength between vertex $v$ and community $c$. Matrix $\mathbf{F}$ is nonnegative and $\mathbf{F}_{vc} = 0$ indicates no affiliation. BIGCLAM builds a generative model by modeling the probability that vertex $v_i$ connects $v_j$ given user-community affinity matrix $\mathbf{F}$. More specifically, given matrix $\mathbf{F}$, BIGCLAM generates an edge between vertex $v_i$ and $v_j$ with a probability

$$P(v_i, v_j) = 1 - \exp(-\mathbf{F}_{v_i} \cdot \mathbf{F}_{v_j}), \tag{8.62}$$

where $\mathbf{F}_{v_i}$ is the corresponding row of matrix $\mathbf{F}$ for vertex $v_i$ and can be seen as the representation of $v_i$. Note that the probability $P(v_i, v_j)$ has an increasing relationship with $\mathbf{F}_{v_i} \cdot \mathbf{F}_{v_j}^\top = \sum_c \mathbf{F}_{v_i,c}\mathbf{F}_{v_j,c}$, which indicates that the more communities a pair of nodes shared, the more likely they are connected.

For the case that $\mathbf{F}_{v_i} \cdot \mathbf{F}_{v_j} = 0$, BIGCLAM adds a background probability $\epsilon = \frac{2|E|}{|V|(|V|-1)}$ to the pair of nodes to avoid a zero probability.

Then BIGCLAM tries to maximize the log-likelihood of the graph $G = (V, E)$:

$$\mathscr{O}(\mathbf{F}) = \sum_{i,j:(v_i,v_j)\in E} \log P(v_i, v_j) + \sum_{i,j:(v_i,v_j)\notin E} \log(1 - P(v_i, v_j)), \tag{8.63}$$

which can be reformulated as

$$\mathscr{O}(\mathbf{F}) = \sum_{i,j:(v_i,v_j)\in E} \log(1 - \exp(-\mathbf{F}_{v_i} \cdot \mathbf{F}_{v_j})) - \sum_{i,j:(v_i,v_j)\notin E} \mathbf{F}_{v_i} \cdot \mathbf{F}_{v_j}. \tag{8.64}$$

The parameters $\mathbf{F}$ are learned by projected gradient descent. Note that the training objective can be regarded as a variant of nonnegative matrix factorization. The maximization of log-likelihood function is an approximation of adjacency matrix $A$ by $\mathbf{F}\mathbf{F}^\top$. Compared with L2-norm loss function, the gradient of Eq. 8.64 can be computed more efficiently for a sparse matrix $A$ which is the most case in the real-world dataset.

The model can also be generalized to asymmetric case [141]. That is to replace Eq. 8.62 by

$$P(v_i, v_j) = 1 - \exp(-\mathbf{F}_{v_i} \cdot \mathbf{H}_{v_j}), \tag{8.65}$$

where $\mathbf{H}$ is another matrix that has the same size with the matrix $\mathbf{F}$. The generative model can also consider attributes of vertices by adding attribute terms to Eq. 8.62 [79].

#### 8.2.5.4 Network Representation for Visualization

Different from previous algorithms that focus on machine learning tasks, the algorithms introduced in this subsection are designed for visualization. As a commonly used data structure, the visualization of networks is an important task. The dimensions of representations of vertices are usually 2 or 3 to draw the graph.

Representation learning for network visualization generally follows the following aesthetic criteria [30]:

- Distribute the vertices evenly in the frame.
- Minimize edge crossings.
- Make edge lengths uniform.
- Reflect inherent symmetry.
- Conform to the frame.

Following these criteria, graph visualization algorithms build a force-directed graph drawing framework. The basic assumption is that there is a spring between each pair of vertices. Then the optimization objective is to minimize the energy of the graph according to Hooke's law:

$$\mathscr{E} = \sum_{i,j} \frac{1}{2} k_{ij} (\|\mathbf{v}_i - \mathbf{v}_j\| - l_{ij})^2, \tag{8.66}$$

where $k_{ij}$ is spring constant, $\mathbf{v}_i$ is the position of vertex $v_i$ and $l_{ij}$ is the length of shortest path between vertex $v_i$ and $v_j$. The intuition is straightforward: close vertices should have close positions in the drawing. Several algorithms have been proposed to improve this framework [34, 54, 60] by changing the setting of spring constant $k_{ij}$ or the energy function. The parameters can be easily learned via gradient descent.

#### 8.2.5.5 Embedding Enhancement via High-Order Proximity Approximation

Yang et al. [137] summarize several existing NRL methods into a unified two-step framework, including proximity matrix construction and dimension reduction. They conclude that an NRL method can be improved by exploring higher order proximities when building the proximity matrix. Then they propose Network Embedding Update (NEU) algorithm, which implicitly approximates higher order proximities with theoretical approximation bound and can be applied to any NRL methods to enhance their performances. NEU can make a consistent and significant improvement over some NRL methods with almost negligible running time.

The two-step framework is summarized as follows:

**Step 1: Proximity Matrix Construction.** Compute a proximity matrix $\mathbf{M} \in \mathbb{R}^{|V|\times|V|}$, which encodes the information of $k$-order proximity matrix where $k = 1, 2 \ldots, K$. For example, $\mathbf{M} = \frac{1}{K}A + \frac{1}{K}A^2 \cdots + \frac{1}{K}A^K$ stands for an average combination of $k$-order proximity matrix for $k = 1, 2 \ldots, K$. The proximity matrix $M$ is usually represented by a polynomial of normalized adjacency matrix $A$ of degree $K$, and we denote the polynomial as $f(A) \in \mathbb{R}^{|V|\times|V|}$. Here the degree $K$ of polynomial $f(A)$ corresponds to the maximum order of proximities encoded in the proximity matrix. Note that the storage and computation of proximity matrix $M$ doesn't necessarily take $O(|V|^2)$ time because we only need to save and compute the nonzero entries.

**Step 2: Dimension Reduction.** Find network embedding matrix $\mathbf{V} \in \mathbb{R}^{|V|\times d}$ and context embedding $\mathbf{C} \in \mathbb{R}^{|V|\times d}$ so that the product $\mathbf{V}\mathbf{C}^\top$ approximates proximity matrix $\mathbf{M}$. Here different algorithms may employ different distance functions to minimize the distance between $\mathbf{M}$ and $\mathbf{V}\mathbf{C}^\top$. For example, we can naturally use the norm of matrix $\mathbf{M} - \mathbf{V}\mathbf{C}^\top$ to measure the distance and minimize it.

Spectral Clustering, DeepWalk, and GraRep can be formalized into the two-step framework. Now we focus on the first step and study how to define the right proximity matrix for NRL.

We summarize the comparisons among Spectral Clustering (SC), DeepWalk, and GraRep in Table 8.4 and conclude the following observations.

**Table 8.4** Comparisons among three NRL methods

| | SC | DeepWalk | GraRep |
|---|---|---|---|
| Proximity matrix | $L$ | $\sum_{k=1}^{K} \frac{A^k}{K}$ | $A^k, k = 1 \dots K$ |
| Computation | Accurate | Approximate | Accurate |
| Scalability | Yes | Yes | No |
| Performance | Low | Middle | High |

**Observation 8.1** Modeling higher order and accurate proximity matrix can improve the quality of network representation. In other words, NRL can benefit from exploring a polynomial proximity matrix $f(A)$ of a higher degree.

From the development of NRL methods, it can be seen that DeepWalk outperforms Spectral Clustering because DeepWalk considers higher order proximity matrices, and the higher order proximity matrices can provide complementary information for lower order proximity matrices. GraRep outperforms DeepWalk because GraRep accurately calculates the $k$-order proximity matrix rather than approximating it by Monte Carlo simulation as DeepWalk does.

**Observation 8.2** Accurate computation of high-order proximity matrix is not feasible for large-scale networks.

The major drawback of GraRep is the computation complexity of calculating the accurate $k$-order proximity matrix. In fact, the computation of high-order proximity matrix takes $O(|V|^2)$ time and the time complexity of SVD decomposition also increases as $k$-order proximity matrix gets dense when $k$ grows. In summary, the time complexity of $O(|V|^2)$ is too expensive to handle large-scale networks.

The first observation provides the motivation to explore higher order proximity matrices in NRL models, but the second observation indicates that an accurate inference of higher order proximity matrices isn't acceptable. Therefore, how to learn network embeddings from approximate higher order proximity matrices efficiently becomes important. To be more efficient, the network representations which encode the information of lower order proximity matrices can be used as our basis to avoid repeated computations. The problem is formalized below.

**Problem Formalization**. Assume that we have normalized adjacency matrix $A$ as the first-order proximity matrix, network embedding $\mathbf{V}$, and context embedding $\mathbf{C}$, where $\mathbf{V}, \mathbf{C} \in \mathbb{R}^{|V|\times d}$. Suppose that the embeddings $\mathbf{V}$ and $\mathbf{C}$ are learned by the above NRL framework which indicates that the product $\mathbf{V}\mathbf{C}^\top$ approximates a polynomial proximity matrix $f(A)$ of degree $K$. The goal is to learn a better representation $\mathbf{V}'$ and $\mathbf{C}'$, which approximates a polynomial proximity matrix $g(A)$ with higher degree than $f(A)$. Also, the algorithm should be efficient in the linear time of $|V|$. Note that the lower bound of time complexity is $O(|V|d)$ which is the size of embedding matrix $R$.

There is a simple, efficient, and effective iterative updating algorithm to solve the above problem.

**Method**. Given hyperparameter $\lambda \in (0, \frac{1}{2}]$, normalized adjacency matrix $A$, we update $\mathbf{V}$ and $\mathbf{C}$ as follows:

$$\begin{aligned}\mathbf{V}' &= \mathbf{V} + \lambda A\mathbf{V},\\ \mathbf{C}' &= \mathbf{C} + \lambda A^\top \mathbf{C}.\end{aligned} \tag{8.67}$$

The time complexity of computing $A\mathbf{V}$ and $A^\top\mathbf{C}$ is $O(|V|d)$ because matrix $A$ is sparse and has $O(|V|)$ nonzero entries. Thus the overall time complexity of one iteration of operation (Eq. 8.67) is $O(|V|d)$.

Recall that product of previous embedding $\mathbf{V}$ and $\mathbf{C}$ approximates polynomial proximity matrix $f(A)$ of degree $K$. It can be proved that the algorithm can learn better embeddings $\mathbf{V}'$ and $\mathbf{C}'$, where the product $\mathbf{V}'\mathbf{C}'^\top$ approximates a polynomial proximity matrix $g(A)$ of degree $K+2$ bounded by matrix infinite norm.

**Theorem** *Denote the network and context embedding by* $\mathbf{V}$ *and* $\mathbf{C}$*, and suppose that the approximation between* $\mathbf{V}\mathbf{C}^\top$ *and proximity matrix* $\mathbf{M} = f(A)$ *is bounded by* $r = \|f(A) - \mathbf{V}\mathbf{C}^\top\|_\infty$ *and* $f(\cdot)$ *is a polynomial of degree* $K$*. Then the product of updated embeddings* $\mathbf{V}'$ *and* $\mathbf{C}'$ *from Eq. 8.67 approximates a polynomial* $g(A) = f(A) + 2\lambda A f(A) + \lambda^2 A^2 f(A)$ *of degree* $K+2$ *with approximation bound* $r' = (1 + 2\lambda + \lambda^2)r \le \frac{9}{4}r$*.*

***Proof*** Assume that $\mathbf{S} = f(A) - \mathbf{V}\mathbf{C}^\top$ and thus $r = \|\mathbf{S}\|_\infty$.

$$\begin{aligned}&\|g(A) - \mathbf{V}'\mathbf{C}'^\top\|_\infty = \|g(A) - (\mathbf{V} + \lambda A\mathbf{V})(\mathbf{C}^\top + \lambda \mathbf{C}^\top A)\|_\infty\\ &= \|g(A) - \mathbf{V}\mathbf{C}^\top - \lambda A\mathbf{V}\mathbf{C}^\top - \lambda\mathbf{V}\mathbf{C}^\top A - \lambda^2 A\mathbf{V}\mathbf{C}^\top A\|_\infty\\ &= \|\mathbf{S} + \lambda A\mathbf{S} + \lambda\mathbf{S}A + \lambda^2 A\mathbf{S}A\|_\infty\\ &\le \|\mathbf{S}\|_\infty + \lambda\|A\|_\infty\|\mathbf{S}\|_\infty + \lambda\|\mathbf{S}\|_\infty\|A\|_\infty + \lambda^2\|\mathbf{S}\|_\infty\|A\|_\infty^2\\ &= r + 2\lambda r + \lambda^2 r,\end{aligned} \tag{8.68}$$

where the second last equality replaces $g(A)$ and $f(A) - \mathbf{V}\mathbf{C}^\top$ by the definitions of $g(A)$ and $\mathbf{S}$ and the last equality uses the fact that $\|A\|_\infty = \max_i \sum_j |A_{ij}| = 1$ because the summation of each row of $A$ equals to 1.

In the experimental settings, it is assumed that the weight of lower order proximities should be larger than higher order proximities because they are more directly related to the original network. Therefore, given $g(A) = f(A) + 2\lambda A f(A) + \lambda^2 A^2 f(A)$, we have $1 \ge 2\lambda \ge \lambda^2 > 0$ which indicates that $\lambda \in (0, \frac{1}{2}]$. The proof indicates that the updated embedding can implicitly approximate a polynomial $g(A)$ of 2 more degrees within $\frac{9}{4}$ times matrix infinite norm of previous embeddings. ■

**Algorithm**. The update Eq. 8.67 can be further generalized in two directions. First we can update embeddings $\mathbf{V}$ and $\mathbf{C}$ according to Eq. 8.69:

$$\begin{aligned}\mathbf{V}' &= \mathbf{V} + \lambda_1 A\,\mathbf{V} + \lambda_2 A\,(A\,\mathbf{V}),\\ \mathbf{C}' &= \mathbf{C} + \lambda_1 A^\top\,\mathbf{C} + \lambda_2 A^\top\,(A^\top\,\mathbf{C}).\end{aligned} \tag{8.69}$$

The time complexity is still $O(|V|d)$ but Eq. 8.69 can obtain higher proximity matrix approximation than Eq. 8.67 in one iteration. More complex update formulas that explore further higher proximities than Eq. 8.69 can also be applied but Eq. 8.69 is used in current experiments as a cost-effective choice.

Another direction is that the update equation can be processed for $T$ rounds to obtain higher proximity approximation. However, the approximation bound would grow exponentially as the number of rounds $T$ grows and thus the update cannot be done infinitely. Note that the update operation of $\mathbf{V}$ and $\mathbf{C}$ are completely independent. Therefore, only updating network embedding $\mathbf{V}$ is enough for NRL. The above algorithm (NEU) avoids an accurate computation of high-order proximity matrix but can yield network embeddings that actually approximate high-order proximities. Hence, this algorithm can improve the quality of network embeddings efficiently. Intuitively, Eqs. 8.67 and 8.69 allow the learned embeddings to further propagate to their neighbors. Hence, the proximities of longer distances between vertices will be embedded.

### *8.2.6 Applications*

In this part, we will introduce common applications for network representation learning and their evaluation metrics.

#### 8.2.6.1 Multi-label Classification

A multi-label classification task is the most widely used network representation learning evaluation task. The representations of vertices are considered as vertex features and applied to classifiers to predict vertex labels. More formally, we assume that there are $K$ labels in total. The vertex-label relationship can be expressed as a binary matrix $\mathbf{M} \in \{0, 1\}^{|V|\times K}$ where $\mathbf{M}_{ij} = 1$ indicates that vertex $v_i$ has $j$th label and $\mathbf{M}_{ij} = 0$ otherwise. Specifically, for the multiclass classification problem, each vertex has exactly one label, which means there is only an "1" in each row of matrix $\mathbf{M}$. For the evaluation task, we set a training ratio which indicates how much percent of vertices have observed labels. Then our goal is to predict the labels for the vertices in the test set.

For unsupervised network representation learning algorithms, the labels of the training set are not used for embedding learning. The network representations are fed to classifiers like SVM or logistic regression. Each label will have its classifier. For semi-supervised learning methods, they take the observed vertex labels into account in the representation learning period. These algorithms will have their specific classifiers for label prediction.

Once the label prediction is done, we can move to compute the evaluation metrics. For multiclass classification, we assume that the number of correctly predicted vertices is $|V_r|$. Then the classification accuracy is defined as the ratio of correctly predicted vertices which can be formulated as $|V_r|/|V|$. For multi-label classification, the precision, recall, and F1 are the most popular metrics, which are computed as follows:

$$\begin{aligned}\text{Precision} &= \frac{N_{\text{correctly predicted labels}}}{N_{\text{predicted labels}}},\\ \text{Recall} &= \frac{N_{\text{correctly predicted labels}}}{N_{\text{unobserved labels}}},\\ \text{F1-Score} &= \frac{2\text{Precision} \times \text{Recall}}{\text{Precision} + \text{Recall}}.\end{aligned} \tag{8.70}$$

#### 8.2.6.2 Link Prediction

Link prediction is another important evaluation task for network representation learning because a good network embedding should have the ability to model the affinity between vertices. For evaluation, we randomly pick up edges as training set and leave the rest as test set. Cross-validation can also be employed for training and testing.

To make link prediction given the vertex representations, we first need to evaluate the strength of a pair of vertices. The strength between two vertices is evaluated by computing the similarity between their representations. This similarity is usually computed by cosine similarity, inner product, or square loss, which depends on the algorithm. For example, if an algorithm uses $\|\mathbf{V}_i - \mathbf{C}_j\|_2^2$ in their objective function, then square loss should be used to measure the similarity between vertex representations. Then after we get the similarity of all unobserved links, we can rank them for link prediction. There are two significant metrics for link prediction: area under the receiver operating characteristic curve (AUC) and precision.

**AUC**. The AUC value is the probability that a randomly chosen missing link has a higher score than a randomly chosen nonexistent link. For implementation, we randomly select a missing link and a nonexistent link and compare their similarity score. Assume that there are $n_1$ times that missing link having a higher score and $n_2$ times they have the same score among $n$ independent comparisons. Then the AUC value is

$$\text{AUC} = \frac{n_1 + 0.5n_2}{n}. \tag{8.71}$$

Note that for a random network representation, the AUC value should be 0.5.

**Precision**. Given the ranking of all the non-observed links, we predict the links with top-$L$ highest score as predicted ones. Assume that there are $L_r$ links that are missing links, then the precision is defined as $L_r/L$.

#### 8.2.6.3 Community Detection

For the network representation based community detection algorithm, we first need to convert the nonnegative vertex representation into the hard assignment of commu-nities. Assume that we have network representation matrix $\mathbf{V} \in \mathbb{R}^{+|V| \times k}$ where row $i$ of $\mathbf{V}$ is the nonnegative embedding of vertex $v_i$. For community detection, we regard each dimension of the embeddings as a community. That is to say, $\mathbf{V}_{ij}$ denotes the affinity between vertex $v_i$ and community $c_j$. For each column of matrix $\mathbf{V}$, we set a threshold $\Delta$ and the vertices with affinity score higher than the threshold will be considered as a member of the corresponding community. The threshold can be set in various ways. For example, we can set $\delta$ so that a vertex belongs to a community $c$ if the node is connected to other members of $c$ with an edge probability higher than $1/N$: [140]

$$\frac{1}{N} \leq 1 - \exp(-\Delta^2), \tag{8.72}$$

which indicates that $\Delta = \sqrt{-\log(1 - 1/N)}$.

For evaluation metrics, we have two choices: modularity and matching score.

**Modularity**. Recall that the modularity of a graph $Q$ is defined as

$$Q = \frac{1}{2|E|} \sum_{i,j} \left[ A_{ij} - \frac{\deg(v_i)\deg(v_j)}{2|E|} \right] \delta(v_i, v_j), \tag{8.73}$$

where $\delta(v_i, v_j) = 1$ if $v_i$ and $v_j$ belong to the same community and $\delta(v_i, v_j) = 0$ otherwise. A larger modularity indicates a better community detection algorithm.

**Matching Score**. This is a more sophisticated evaluation metric for community detection. To compare a set of ground truth communities $C^*$ to a set of detected communities $C$, we first need to match each detected community to the most similar ground truth community. On the other side, we also find the most similar detected community for each ground truth community. Then the final performance is evaluated by the average of both sides:

$$\frac{1}{2|C^*|} \sum_{c_i^* \in C^*} \max_{c_j \in C} \delta(c_i^*, c_j) + \frac{1}{2|C|} \sum_{c_j \in C} \max_{c_i^* \in C^*} \delta(c_i^*, c_j), \tag{8.74}$$

where $\delta(c_i^*, c_j)$ is a similarity measurement of ground truth community $c_i^*$ and detected community $c_j$, such as Jaccard similarity. The score is between 0 and 1, where 1 indicates a perfect matching of ground truth communities.

#### 8.2.6.4 Recommender System

Recommender systems aim at recommending items (e.g., products, movies, or locations) for users and cover a wide range of applications. In many cases, an application comes with an associated social network between users. Now we will present an example to show how to use the idea of network representation for building recommender systems in location-based social networks.

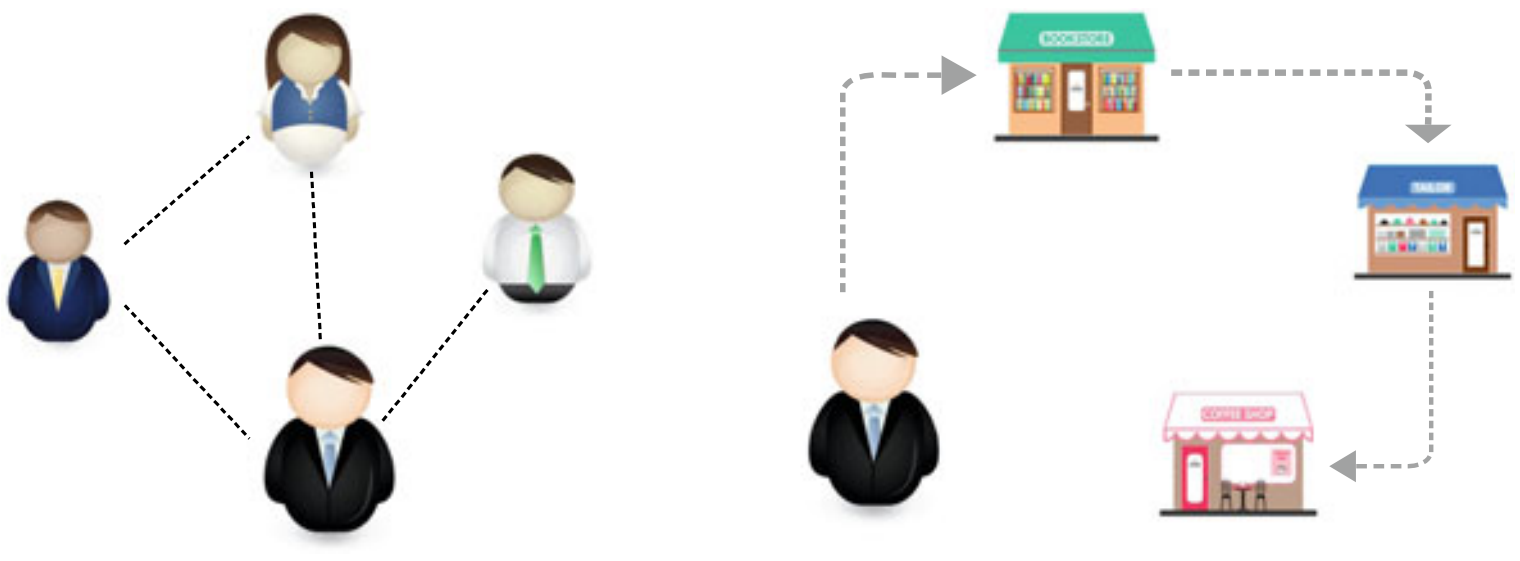

(a) Friendship Network (b) User Trajectory

**Fig. 8.7** An illustrative example for the data in LBSNs: **a** Link connections represent the friendship between users. **b** A trajectory generated by a user is a sequence of chronologically ordered check-in records [138]

The accelerated growth of mobile trajectories in location-based services brings valuable data resources to understand users' moving behaviors. Apart from recording the trajectory data, another major characteristic of these location-based services is that they also allow the users to connect whomever they like or are interested in. As shown in Fig. 8.7, a combination of social networking and location-based services is called as Location-Based Social Networks (LBSN). As shown in [21], locations that are frequently visited by socially related persons tend to be correlated, which indicates the close association between social connections and trajectory behaviors of users in LBSNs. In order to better analyze and mine LBSN data, we need to have a comprehensive view to analyze and mine the information from the two aspects, i.e., the social network and mobile trajectory data.

Specifically, JNTM [138] is proposed to model both social networks and mobile trajectories jointly. The model consists of two components: the construction of social networks and the generation of mobile trajectories. First, JNTM adopts a network embedding method for the construction of social networks where a networking representation can be derived for a user. Secondly, JNTM considers four factors that influence the generation process of mobile trajectories, namely, user visit preference, influence of friends, short-term sequential contexts, and long-term sequential contexts. Then JNTM uses real-valued representations to encode the four factors and set two different user representations to model the first two factors: a visit interest representation and a network representation. To characterize the last two contexts, JNTM employs the RNN and GRU models to capture the sequential relatedness in mobile trajectories at different levels, i.e., short term or long term. Finally, the two components are tied by sharing user network representations. The overall model is illustrated in Fig. 8.8.

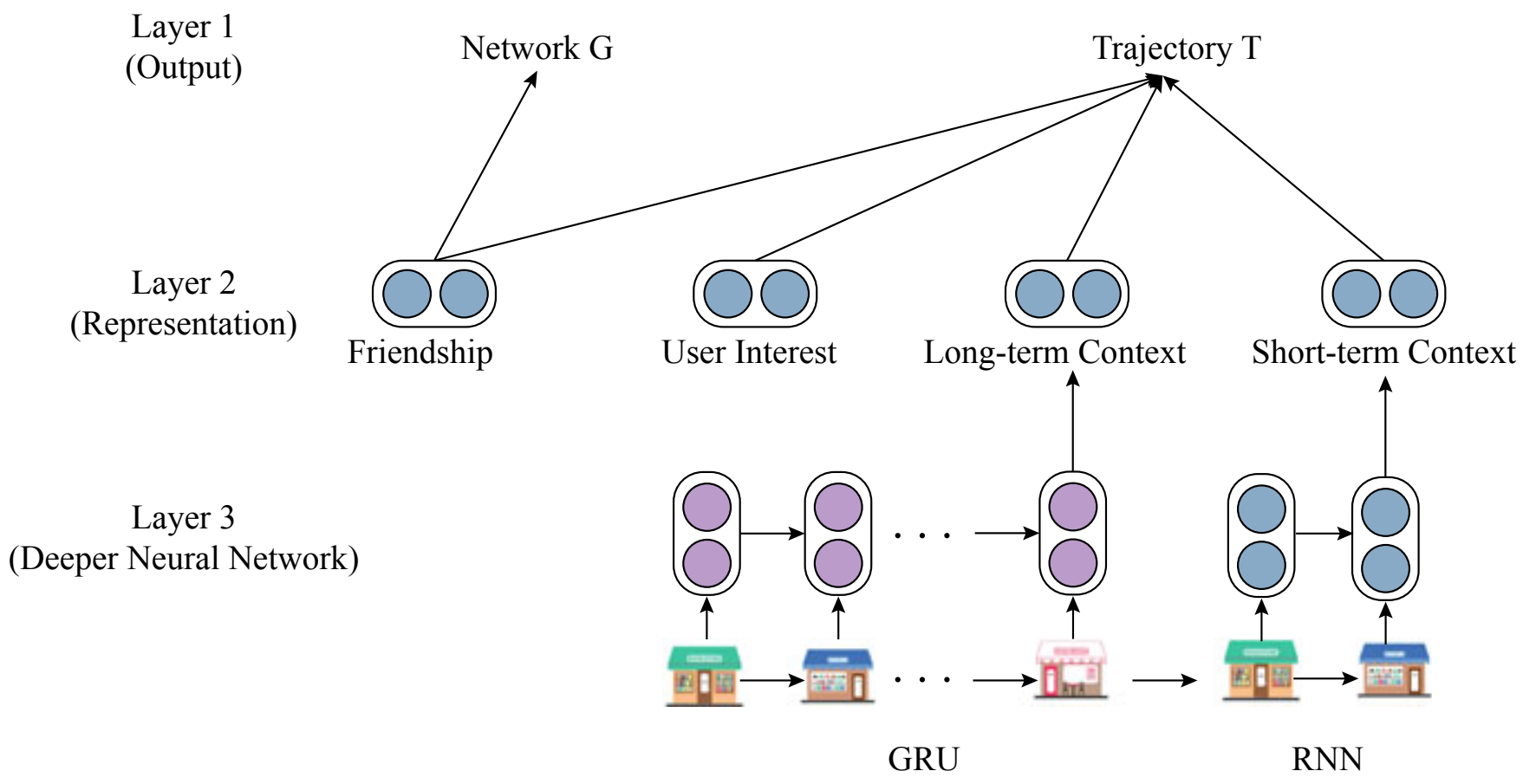

**Fig. 8.8** The architecture of JNTM model

#### 8.2.6.5 Information Diffusion Prediction

Information diffusion prediction is an important task which studies how information items spread among users. The prediction of information diffusion, also known as *cascade* prediction, has been studied over a wide range of applications, such as product adoption [67], epidemiology [124], social networks [63], and the spread of news and opinions [68].

As shown in Fig. 8.9, microscopic diffusion prediction aims at guessing the next infected user, while macroscopic diffusion prediction estimates the total numbers of infected users during the diffusion process. Also, an underlying social graph among users will be available when information diffusion occurs on a social network service. The social graph will be considered as additional structural inputs for diffusion prediction.

FOREST [139] is the first work to address both microscopic and macroscopic predictions. As shown in Fig. 8.10, FOREST proposes a structural context extrac-tion algorithm that was originally introduced for accelerating graph convolutional networks [41] to build an RNN-based microscopic cascade model. For each user $v$, we first sample $Z$ users $\{u_1, u_2 \ldots, u_Z\}$ from $v$ and its neighbors $\mathscr{N}(v)$. Then we update its feature vector by aggregating the neighborhood features. The updated user feature vector encodes structural information by aggregating features from $v$'s first-order neighbors. The operation can also be processed recursively to explore a larger neighborhood of user $v$. Empirically, a two-step neighborhood exploration is time efficient and enough to give promising results.

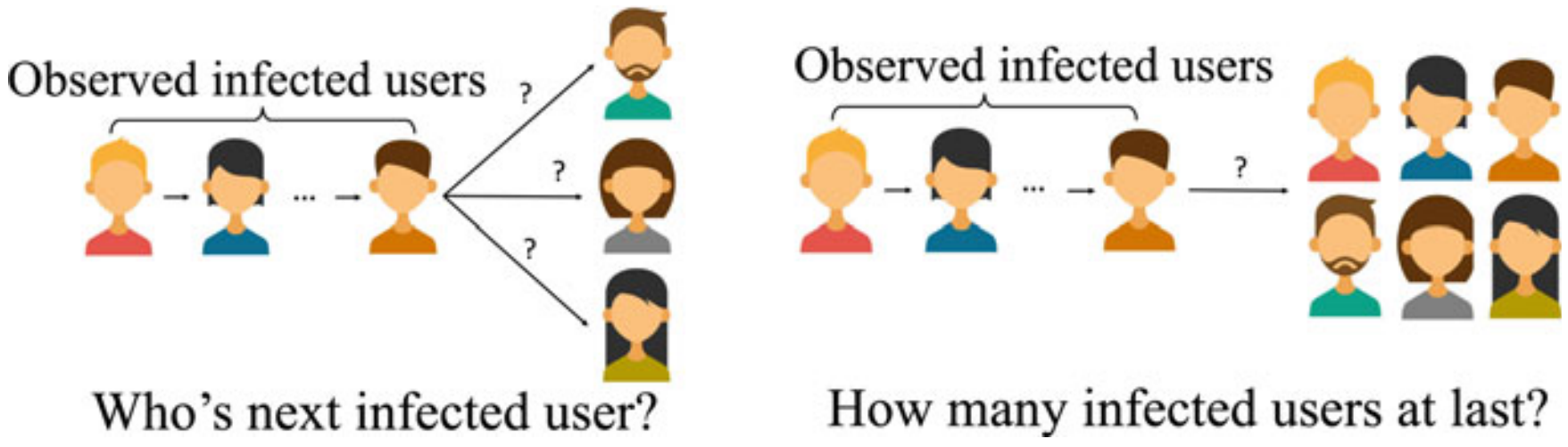

**Fig. 8.9** Illustrative examples for microscopic next infected user prediction (left) and macroscopic cascade size prediction (right) [139]

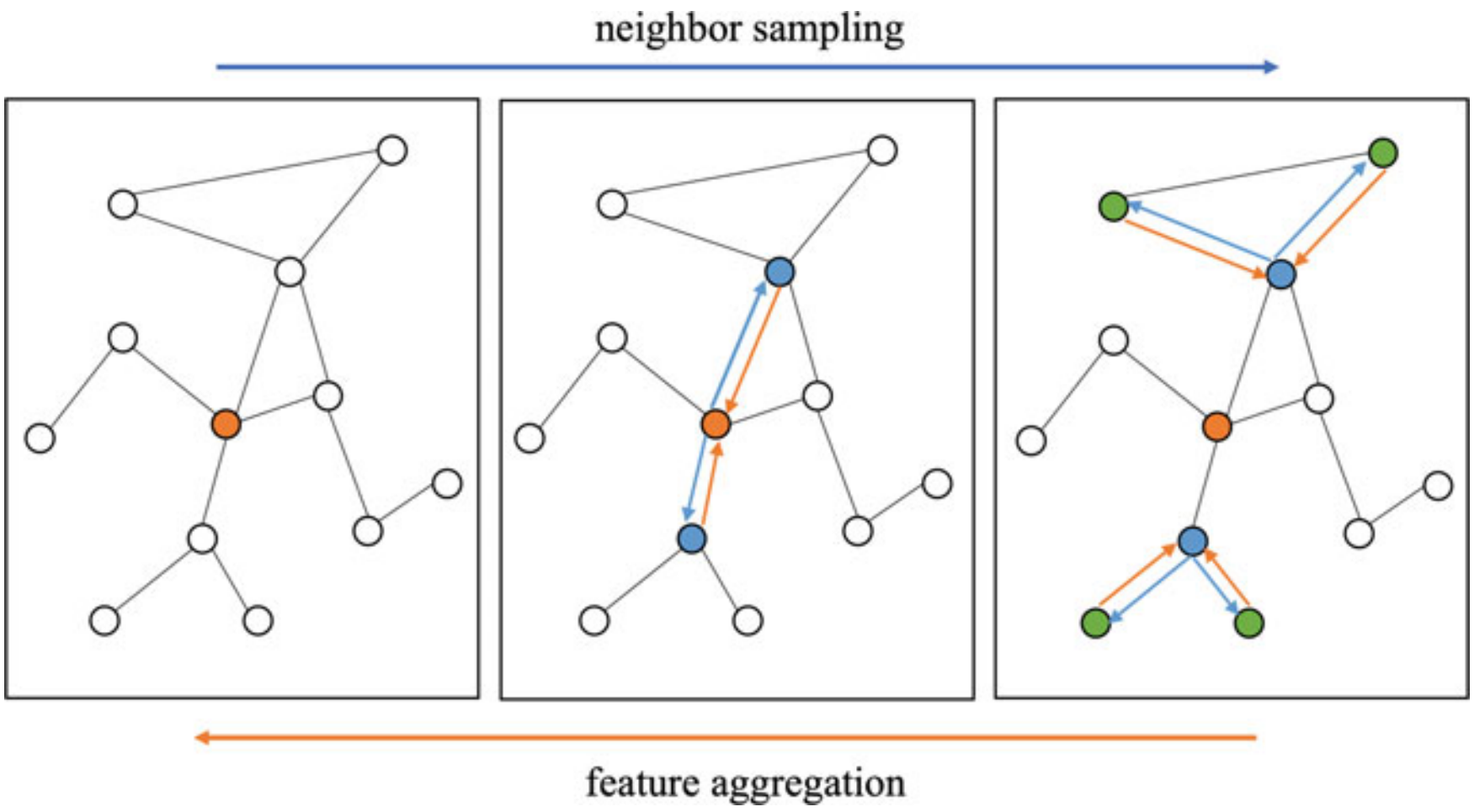

**Fig. 8.10** An illustrative example of structural context extraction of the orange node by neighbor sampling and feature aggregation [139]

FOREST further incorporates the ability of macroscopic prediction, i.e., estimating the eventual size of a cascade into the model by reinforcement learning. The method can be divided into four steps: (a) encode observed $K$ users by a microscopic cascade model; (b) enable the microscopic cascade model to predict the size of a cascade by cascade simulations; (c) use Mean-Square Log-Transformed Error (MSLE) as the supervision signal for macroscopic predictions; and (d) employ a reinforcement learning framework to update parameters through policy gradient algorithm. The overall workflow is illustrated in Fig. 8.11.

## 8.3 Graph Neural Networks

In this section, we will introduce another kind of method for network representation learning, which is called Graph Neural Networks (GNNs) [101]. These methods aim to utilize neural networks to model graph data and have shown their strong capabilities in many applications.

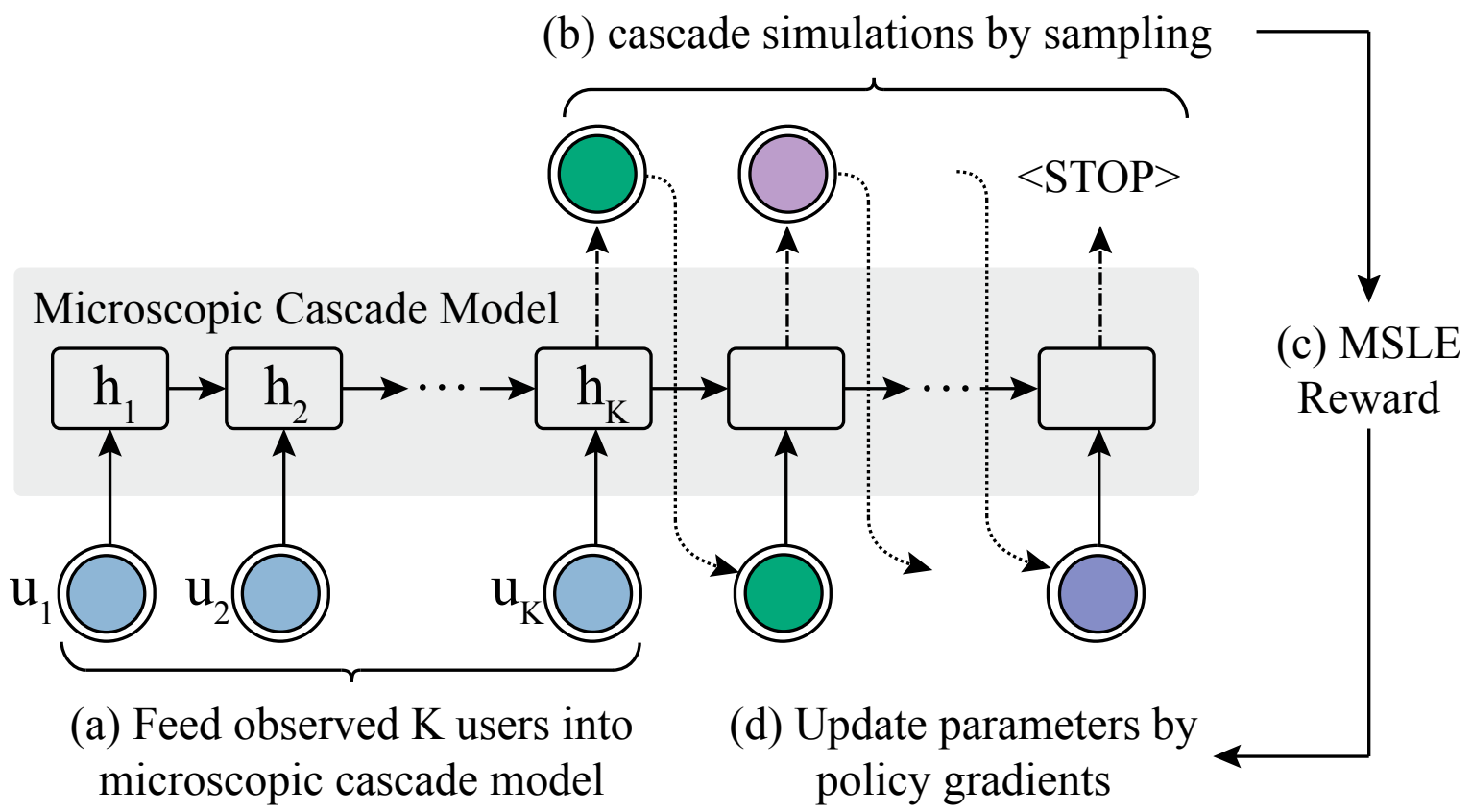

**Fig. 8.11** The workflow of adopting microscopic cascade model for macroscopic size prediction by reinforcement learning

## *8.3.1 Motivations*

Graph Neural Networks (GNNs) are deep learning based methods that operate on graph domain. Due to its convincing performance and high interpretability, GNN has been a widely applied graph analysis method recently. In this subsection, we will illustrate the fundamental motivations of graph neural networks.

In recent years, CNNs [65] have made breakthroughs in various machine learning areas, especially in the area of computer vision, and started the revolution of deep learning [64]. CNNs are capable of extracting multiscale localized features and these features are used to generate more expressive representations. As we are going deeper into CNNs and graphs, we found the keys of CNNs: local connection, shared weights, and the use of multilayer [64]. These are also of great importance in solving problems of graph domain, because (1) graphs are the most typical locally connected structure, (2) shared weights reduce the computational cost compared with traditional spectral graph theory [23], and (3) multilayer structure is the key to deal with hierarchical patterns, which captures the features of various sizes. However, CNNs can only operate on regular Euclidean data like images (2D grid) and text (1D sequence) while these data structures can be regarded as instances of graphs. Therefore, it is straightforward to think of finding the generalization of CNNs to graphs. As shown in Fig. 8.12, it is hard to define localized convolutional filters and pooling operators, which hinders the transformation of CNN from Euclidean domain to non-Euclidean domain.

The other motivation comes from *network embedding* [12, 24, 37, 42, 149]. In the field of graph analysis, traditional machine learning approaches usually rely on hand-engineered features and are limited by its inflexibility and high cost. Following the idea of *representation learning* and the success of word embedding [81], DeepWalk [93], which is regarded as the first graph embedding method based on representation learning, applies Skip-gram model [81] on the generated random walks. Similar approaches such as node2vec [38], LINE [111], and TADW [136] also achieved breakthroughs. However, these methods suffer from two severe drawbacks [42]. First, no parameters are shared between nodes in the encoder, which leads to computational inefficiency, since it means the number of parameters grows linearly with the number of nodes. Second, the direct embedding methods lack the ability of generalization, which means they cannot deal with dynamic graphs or generalize to new graphs.

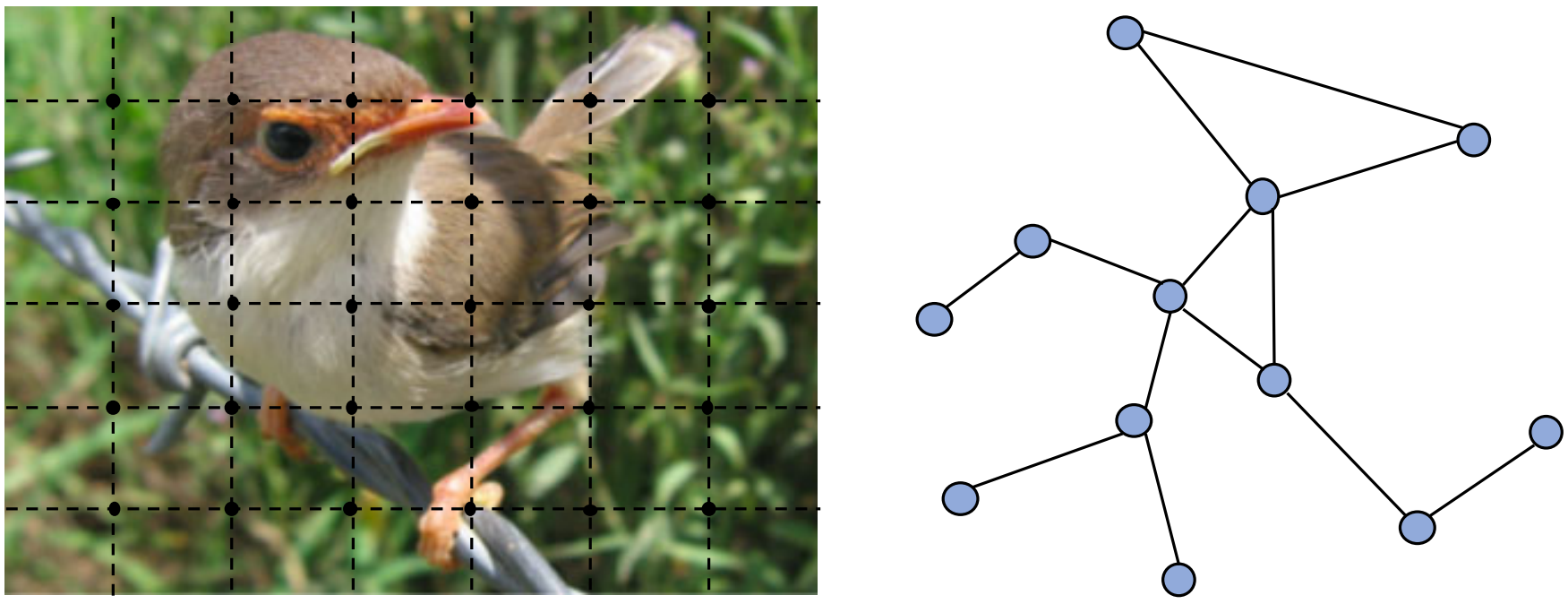

**Fig. 8.12** Left: image in Euclidean space. Right: graph in non-Euclidean space [155]

Based on CNNs and network embedding, Graph Neural Networks (GNNs) are proposed to collectively aggregate information from graph structure. Thus, they can model input and/or output consisting of elements and their dependency. Further, the graph neural networks can simultaneously model the diffusion process on the graph with the RNN kernel.

In the rest of this section, we will first introduce several typical variants of graph neural networks such as Graph Convolutional Networks (GCNs), Graph Attention Networks (GATs), and Graph Recurrent Networks (GRNs). Then we will introduce several extensions to the original model and finally, we will give some examples of applications that utilize graph neural networks.

### *8.3.2 Graph Convolutional Networks*

Graph Convolutional Networks (GCNs) aim to generalize convolutions to the graph domain. Advances in this direction are often categorized as spectral approaches and spatial (nonspectral) approaches.

#### 8.3.2.1 Spectral Approaches

Spectral approaches work with a spectral representation of the graphs.

**Spectral Network.** Bruna et al. [11] proposes the spectral network. The convolution operation is defined in the Fourier domain by computing the eigendecomposition of the graph Laplacian. The operation can be defined as the multiplication of a signal $\mathbf{x} \in \mathbb{R}^N$ (a scalar for each node) with a filter $g_\theta$ =diag($\boldsymbol{\theta}$) parameterized by $\boldsymbol{\theta} \in \mathbb{R}^N$:

$$g_\theta \star \mathbf{x} = U g_\theta(\Lambda) U^T \mathbf{x}, \tag{8.75}$$

where $U$ is the matrix of eigenvectors of the normalized graph Laplacian $L = \mathbf{I}_N - D^{-\frac{1}{2}} A D^{-\frac{1}{2}} = U \Lambda U^T$ ($D$ is the degree matrix and $A$ is the adjacency matrix of the graph), with a diagonal matrix of its eigenvalues $\Lambda$.

This operation results in potentially intense computations and non-spatially localized filters. Henaff et al. [47] attempts to make the spectral filters spatially localized by introducing a parameterization with smooth coefficients.

**ChebNet.** Hammond et al. [43] suggests that $g_\theta(\Lambda)$ can be approximated by a truncated expansion in terms of Chebyshev polynomials $T_k(x)$ up to $K$th order. Thus, the operation is

$$g_\theta \star \mathbf{x} \approx \sum_{k=0}^{K} \theta_k T_k(\tilde{L}) \mathbf{x}, \tag{8.76}$$

with $\tilde{L} = 2/\lambda_{max} L - \mathbf{I}_N$. $\lambda_{max}$ denotes the largest eigenvalue of $L$. $\theta \in \mathbb{R}^K$ is now a vector of Chebyshev coefficients. The Chebyshev polynomials are defined as $T_k(x) = 2x T_{k-1}(x) - T_{k-2}(x)$, with $T_0(x) = 1$ and $T_1(x) = x$. It can be observed that the operation is $K$-localized since it is a $K$th-order polynomial in the Laplacian. Defferrard et al. [28] proposes the ChebNet. It uses this $K$-localized convolution to define a convolutional neural network, which could remove the need to compute the eigenvectors of the Laplacian.

**GCN.** Kipf and Welling [59] limits the layer-wise convolution operation to $K = 1$ to alleviate the problem of overfitting on local neighborhood structures for graphs with very wide node degree distributions. It further approximates $\lambda_{max} \approx 2$ and the equation simplifies to

$$g_{\theta'} \star \mathbf{x} \approx \theta'_0 \mathbf{x} + \theta'_1 (L - \mathbf{I}_N) \mathbf{x} = \theta'_0 \mathbf{x} - \theta'_1 D^{-\frac{1}{2}} A D^{-\frac{1}{2}} \mathbf{x}, \tag{8.77}$$

with two free parameters $\theta'_0$ and $\theta'_1$. After constraining the number of parameters with $\theta = \theta'_0 = -\theta'_1$, we can obtain the following expression:

$$g_\theta \star \mathbf{x} \approx \theta \left( \mathbf{I}_N + D^{-\frac{1}{2}} A D^{-\frac{1}{2}} \right) \mathbf{x}. \tag{8.78}$$

Note that stacking this operator could lead to numerical instabilities and exploding/vanishing gradients, [59] introduces the *renormalization trick*:

$\mathbf{I}_N + D^{-\frac{1}{2}} A D^{-\frac{1}{2}} \rightarrow \tilde{D}^{-\frac{1}{2}} \tilde{A} \tilde{D}^{-\frac{1}{2}}$, with $\tilde{A} = A + \mathbf{I}_N$ and $\tilde{D}_{ii} = \sum_j \tilde{A}_{ij}$. Finally, [59] generalizes the definition to a signal $X \in \mathbb{R}^{N \times C}$ with $C$ input channels and $F$ filters for feature maps as follows:

$$\mathbf{H} = f(\tilde{D}^{-\frac{1}{2}} \tilde{A} \tilde{D}^{-\frac{1}{2}} \mathbf{XW}), \tag{8.79}$$

where $\mathbf{W} \in \mathbb{R}^{C \times F}$ is a matrix of filter parameters, $\mathbf{H} \in \mathbb{R}^{N \times F}$ is the convolved signal matrix and $f(\cdot)$ is the activation function.

The GCN layer can be stacked for multiple times so that we have the equation:

$$\mathbf{H}^{(t)} = f(\tilde{D}^{-\frac{1}{2}} \tilde{A} \tilde{D}^{-\frac{1}{2}} \mathbf{H}^{(t-1)} \mathbf{W}^{(t-1)}), \tag{8.80}$$

where the superscripts $t$ and $t-1$ denote the layers of the matrices, the initial matrix $\mathbf{H}^{(0)}$ could be $\mathbf{X}$. After L layers, we can use the final embedding matrix $\mathbf{H}^{(L)}$ and a readout function to get the final output matrix $\mathbf{Z}$:

$$\mathbf{Z} = \text{Readout}(H^{(L)}), \tag{8.81}$$

where the readout function can be any machine learning methods, such as MLP.

Finally, as a semi-supervised algorithm, GCN uses the feature matrix at the top layer $\mathbf{Z}$ which has the same dimension with the total number of labels to predict the labels of all observed labels. The loss function can be written as

$$\mathscr{L} = -\sum_{l \in y_L} \sum_f Y_{lf} \ln \mathbf{Z}_{lf}, \tag{8.82}$$

where $y_L$ is the set of node indices that have observed labels. Figure 8.13 shows the algorithm of GCN.

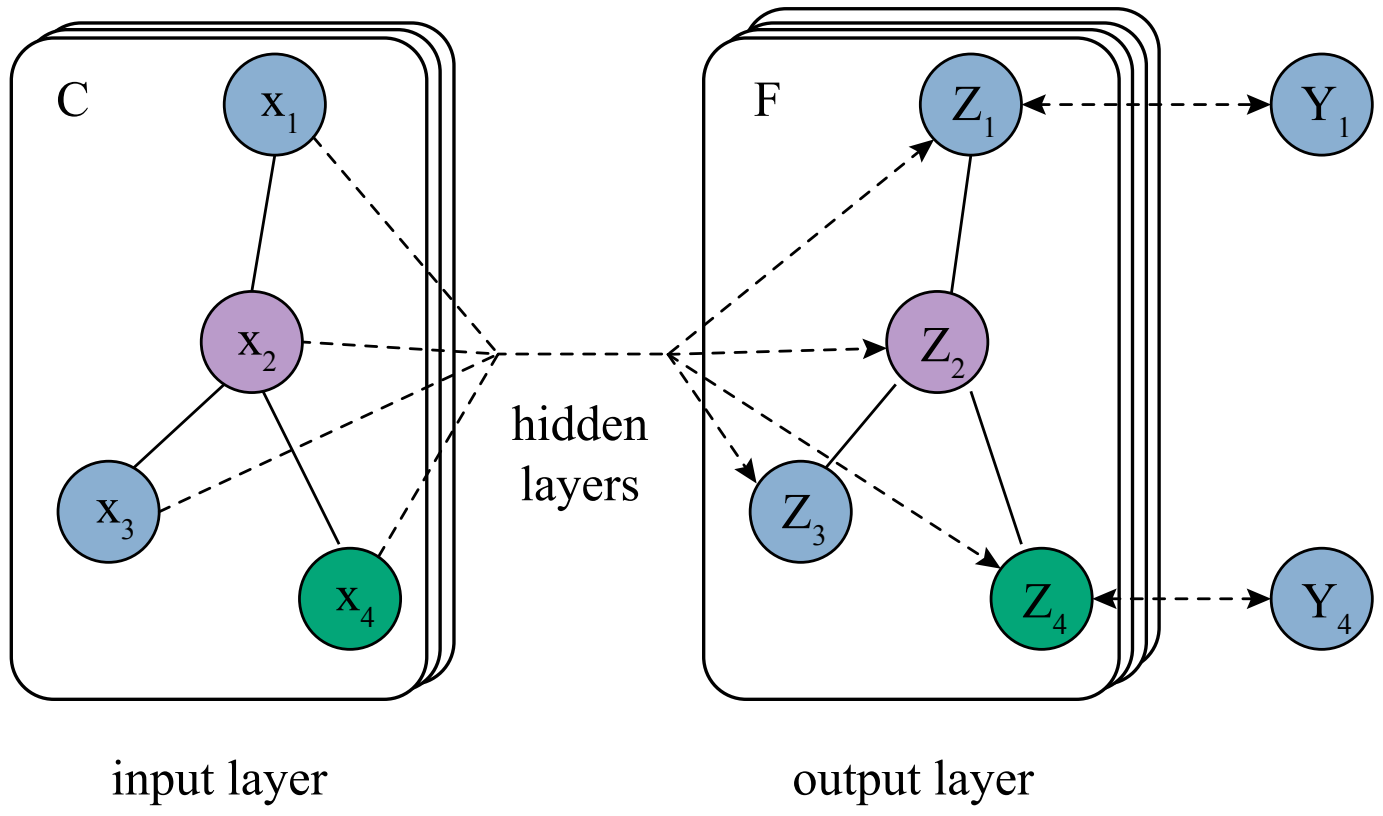

**Fig. 8.13** The architecture of graph convolutional network model

### 8.3.2.2 Spatial Approaches

In all of the spectral approaches mentioned above, the learned filters depend on the Laplacian eigenbasis, which depends on the graph structure, that is, a model trained on a specific structure could not be directly applied to a graph with a different structure.

Spatial approaches define convolutions directly on the graph, operating on spatially close neighbors. The major challenge of spatial approaches is defining the convolution operation with differently sized neighborhoods and maintaining the local invariance of CNNs.

**Neural FPs.** Duvenaud et al. [31] uses different weight matrices for nodes with different degrees

$$\begin{aligned} \mathbf{x}^{(t)} &= \mathbf{h}_v^{(t-1)} + \sum_{i=1}^{|N_v|} \mathbf{h}_i^{(t-1)}, \\ \mathbf{h}_v^{(t)} &= f(\mathbf{W}_{|N_v|}^{(t)} \mathbf{x}^{(t)}), \end{aligned} \tag{8.83}$$

where $\mathbf{W}_{|N_v|}^{(t)}$ is the weight matrix for nodes with degree $|N_v|$ at layer $t$. And the main drawback of the method is that it cannot be applied to large-scale graphs with more node degrees.

In the following description of other models, we use $h_v^{(t)}$ to denote the hidden state of node $v$ at layer $t$. $N_v$ denotes the neighbor set of node $v$ and $|N_v|$ denotes the size of the set.

**DCNN.** Atwood and Towsley [4] proposes the Diffusion-Convolutional Neural Networks (DCNNs). Transition matrices are used to define the neighborhood for nodes in DCNN. For node classification, it has

$$\mathbf{H} = f\left(\mathbf{W}^c \odot \overrightarrow{\mathbf{P}}\mathbf{X}\right), \tag{8.84}$$

where $\odot$ is the element-wise multiplication and $\mathbf{X}$ is an $N \times F$ matrix of input features. $\overrightarrow{\mathbf{P}}$ is an $N \times K \times N$ tensor which contains the power series $\{\mathbf{P}, \mathbf{P}^2, \ldots, \mathbf{P}^K\}$ of matrix $\mathbf{P}$. And $\mathbf{P}$ is the degree-normalized transition matrix from the graphs adjacency matrix $A$. Each entity is transformed to a diffusion-convolutional representation, which is a $K \times F$ matrix defined by $K$ hops of graph diffusion over $F$ features. And then it will be defined by a $K \times F$ weight matrix and a nonlinear activation function $f$. Finally $\mathbf{H}$ (which is $N \times K \times F$) denotes the diffusion representations of each node in the graph.

**DGCN.** Zhuang and Ma [158] proposes the Dual Graph Convolutional Network (DGCN) to consider the local consistency and global consistency of graphs jointly. It uses two convolutional networks to capture the local/global consistency and adopts an unsupervised loss to ensemble them. The first convolutional network is the same as Eq. 8.80. And the second network replaces the adjacency matrix with Positive Point-wise Mutual Information (PPMI) matrix:

$$\mathbf{H}^{(t)} = f(D_P^{-\frac{1}{2}} X_P D_P^{-\frac{1}{2}} H^{(t-1)} \mathbf{W}), \tag{8.85}$$

where $X_P$ is the PPMI matrix and $D_P$ is the diagonal degree matrix of $X_P$.

**GraphSAGE.** Hamilton et al. [41] proposes the GraphSAGE, a general inductive framework. The framework generates embeddings by sampling and aggregating features from a node's local neighborhood.

$$\begin{aligned}\mathbf{h}_{N_v}^{(t)} &= \text{AGGREGATE}^{(t)}(\{\mathbf{h}_u^{(t-1)}, \forall u \in N_v\}),\\ \mathbf{h}_v^{(t)} &= f(\mathbf{W}^{(t)}[\mathbf{h}_v^{(t-1)}; \mathbf{h}_{N_v}^{(t)}]).\end{aligned} \tag{8.86}$$

However, [41] does not utilize the full set of neighbors in Eq. 8.86 but a fixed-size set of neighbors by uniformly sampling. And [41] suggests three aggregator functions.

- Mean aggregator. It could be viewed as an approximation of the convolutional operation from the transductive GCN framework [59], so that the inductive version of the GCN variant could be derived by

$$\mathbf{h}_v^{(t)} = f\left(\mathbf{W} \cdot \text{MEAN}\left(\{\mathbf{h}_v^{(t-1)}\} \cup \{\mathbf{h}_u^{(t-1)} | \forall u \in N_v\}\right)\right). \tag{8.87}$$

  The mean aggregator is different from other aggregators because it does not perform the concatenation operation which concatenates $\mathbf{h}_v^{t-1}$ and $\mathbf{h}_{N_v}^t$ in Eq. 8.86. It can be viewed as a form of "skip connection" [46] and can achieve better performance.
- LSTM aggregator. Hamilton et al. [41] also uses an LSTM-based aggregator which has a larger expressive capability. However, LSTMs process inputs in a sequential manner so that they are not permutation invariant. Hamilton et al. [41] adapts LSTMs to operate on an unordered set by permutating node's neighbors.
- Pooling aggregator. In the pooling aggregator, each neighbor's hidden state is fed through a fully connected layer and then a max-pooling operation is applied to the set of the node's neighbors.

$$\mathbf{h}_{N_v}^{(t)} = \max(\{f(\mathbf{W}_{pool}\mathbf{h}_u^{(t-1)} + \mathbf{b}), \forall u \in N_v\}). \tag{8.88}$$

  Note that any symmetric functions could be used in place of the max-pooling operation here.

**Other methods.** There are still many other spatial methods. The PATCHY-SAN model [86] first extracts exactly $k$ nodes for each node and normalizes them. Then the convolutional operation is applied to the normalized neighborhood. LGCN [35] leverages CNNs as aggregators. It performs max-pooling on nodes' neighborhood matrices to get top-k feature elements and then applies 1-D CNN to compute hidden representations. Monti et al. [82] proposes a spatial-domain model (MoNet) on non-Euclidean domains which could generalize several previous techniques.

The Geodesic CNN (GCNN) [78] and Anisotropic CNN (ACNN) [10] on manifolds or GCN [59] and DCNN [4] on graphs could be formulated as particular instances of MoNet. Our readers can refer to their papers for more details.

### 8.3.3 Graph Attention Networks

The attention mechanism has been successfully used in many sequence-based tasks such as machine translation [5, 36, 121], machine reading [19], etc. Many works focus on generalizing the attention mechanism to the graph domain.

**GAT.** Velickovic et al. [122] proposes a Graph Attention Network (GAT) which incorporates the attention mechanism into the propagation step. Specifically, it uses the *self-attention* strategy and each node's hidden state is computed by attending over its neighbors.

Velickovic et al. [122] defines a single *graph attentional layer* and constructs arbitrary graph attention networks by stacking this layer. The layer computes the coefficients in the attention mechanism of the node pair $(i, j)$ by:

$$\alpha_{ij} = \frac{\exp\left(\text{LeakyReLU}\left(\mathbf{a}^\top[\mathbf{W}\mathbf{h}_i^{(t-1)}; \mathbf{W}\mathbf{h}_j^{(t-1)}]\right)\right)}{\sum_{k\in N_i}\exp\left(\text{LeakyReLU}\left(\mathbf{a}^\top[\mathbf{W}\mathbf{h}_i^{(t-1)}; \mathbf{W}\mathbf{h}_k^{(t-1)}]\right)\right)}, \tag{8.89}$$

where $\alpha_{ij}$ is the attention coefficient of node $j$ to $i$. $\mathbf{W} \in \mathbb{R}^{F' \times F}$ is the *weight matrix* of a shared linear transformation which applied to every node, $\mathbf{a} \in \mathbb{R}^{2F'}$ is the weight vector. It is normalized by a softmax function and the LeakyReLU nonlinearity (with negative input slop 0.2) is applied.

Then the final output features of each node can be obtained by (after applying a nonlinearity $f$):

$$\mathbf{h}_i^{(t)} = f\left(\sum_{j\in N_i}\alpha_{ij}\mathbf{W}\mathbf{h}_j^{(t-1)}\right). \tag{8.90}$$

Moreover, the layer utilizes the *multi-head attention* similarly to [121] to stabilize the learning process. It applies $K$ independent attention mechanisms to compute the hidden states and then concatenates their features(or computes the average), resulting in the following two output representations:

$$\mathbf{h}_i^{(t)} = \|_{k=1}^{K} f\left(\sum_{j\in N_i}\alpha_{ij}^k\mathbf{W}^k\mathbf{h}_j^{(t-1)}\right), \tag{8.91}$$

$$\mathbf{h}_i^{(t)} = f\left(\frac{1}{K}\sum_{k=1}^{K}\sum_{j\in N_i}\alpha_{ij}^k\mathbf{W}^k\mathbf{h}_j^{(t-1)}\right), \tag{8.92}$$

where $\alpha_{ij}^k$ is normalized attention coefficient computed by the $k$th attention mechanism, $\|$ is the concatenation operation.

The attention architecture in [122] has several properties: (1) the computation of the node-neighbor pairs is parallelizable thus the operation is efficient; (2) it can deal with nodes that have different degrees by assigning reasonable weights to their neighbors; (3) it can be applied to the inductive learning problems easily.

**GAAN.** Besides GAT, Gated Attention Network (GAAN) [150] also uses the multi-head attention mechanism. However, it uses a self-attention mechanism to gather information from different heads to replace the average operation of GAT.

### *8.3.4 Graph Recurrent Networks*

Several works are attempting to use the gate mechanism like GRU [20] or LSTM [48] in the propagation step to release the limitations induced by the vanilla GNN architecture and improve the effectiveness of the long-term information propagation across the graph. We call these methods Graph Recurrent Networks (GRNs) and we will introduce some variants of GRNs in this subsection.

**GGNN.** Li et al. [72] proposes the gated graph neural network (GGNN) which uses the Gate Recurrent Units (GRU) in the propagation step. It follows the computation steps from recurrent neural networks for a fixed number of $L$ steps, then it back-propagates through time to compute gradients.

Specifically, the basic recurrence of the propagation model is

$$\begin{aligned}
\mathbf{a}_v^{(t)} &= A_v^\top [\mathbf{h}_1^{(t-1)} \dots \mathbf{h}_N^{(t-1)}]^\top + \mathbf{b}, \\
\mathbf{z}_v^{(t)} &= \text{Sigmoid}\left(\mathbf{W}^z \mathbf{a}_v^{(t)} + \mathbf{U}^z \mathbf{h}_v^{(t-1)}\right), \\
\mathbf{r}_v^{(t)} &= \text{Sigmoid}\left(\mathbf{W}^r \mathbf{a}_v^{(t)} + \mathbf{U}^r \mathbf{h}_v^{(t-1)}\right), \\
\widetilde{\mathbf{h}}_v^{(t)} &= \tanh\left(\mathbf{W}\mathbf{a}_v^{(t)} + \mathbf{U}\left(\mathbf{r}_v^{(t)} \odot \mathbf{h}_v^{(t-1)}\right)\right), \\
\mathbf{h}_v^{(t)} &= \left(1 - \mathbf{z}_v^{(t)}\right) \odot \mathbf{h}_v^{(t-1)} + \mathbf{z}_v^{(t)} \odot \widetilde{\mathbf{h}}_v^{(t)}.
\end{aligned} \tag{8.93}$$

The node $v$ first aggregates message from its neighbors, where $A_v$ is the sub-matrix of the graph adjacency matrix $A$ and denotes the connection of node $v$ with its neighbors. Then the hidden state of the node is updated by the GRU-like function using the information from its neighbors and the hidden state from the previous timestep. $\mathbf{a}$ gathers the neighborhood information of node $v$, $\mathbf{z}$ and $\mathbf{r}$ are the update and reset gates.

LSTMs are also used similarly as GRU through the propagation process based on a tree or a graph.

**Tree-LSTM.** Tai et al. [109] proposes two extensions to the basic LSTM architecture: the Child-Sum Tree-LSTM and the N-ary Tree-LSTM. Like in standard LSTM units, each Tree-LSTM unit (indexed by $v$) contains input and output gates $\mathbf{i}_v$ and $\mathbf{o}_v$, a memory cell $\mathbf{c}_v$ and hidden state $\mathbf{h}_v$. The Tree-LSTM unit replaces the single forget gate by a forget gate $\mathbf{f}_{vk}$ for each child $k$, allowing node $v$ to select information from its children accordingly. The equations of the Child-Sum Tree-LSTM are

$$
\begin{aligned}
\widetilde{\mathbf{h}}_v^{t-1} &= \sum_{k \in N_v} \mathbf{h}_k^{t-1}, \\
\mathbf{i}_v^t &= \text{Sigmoid}\left(\mathbf{W}^i \mathbf{x}_v^t + \mathbf{U}^i \widetilde{\mathbf{h}}_v^{t-1} + \mathbf{b}^i\right), \\
\mathbf{f}_{vk}^t &= \text{Sigmoid}\left(\mathbf{W}^f \mathbf{x}_v^t + \mathbf{U}^f \mathbf{h}_k^{t-1} + \mathbf{b}^f\right), \\
\mathbf{o}_v^t &= \text{Sigmoid}\left(\mathbf{W}^o \mathbf{x}_v^t + \mathbf{U}^o \widetilde{\mathbf{h}}_v^{t-1} + \mathbf{b}^o\right), \\
\mathbf{u}_v^t &= \tanh\left(\mathbf{W}^u \mathbf{x}_v^t + \mathbf{U}^u \widetilde{\mathbf{h}}_v^{t-1} + \mathbf{b}^u\right), \\
\mathbf{c}_v^t &= \mathbf{i}_v^t \odot \mathbf{u}_v^t + \sum_{k \in N_v} \mathbf{f}_{vk}^t \odot \mathbf{c}_k^{t-1}, \\
\mathbf{h}_v^t &= \mathbf{o}_v^t \odot \tanh(\mathbf{c}_v^t),
\end{aligned}
\tag{8.94}
$$

where $\mathbf{x}_v^t$ is the input vector at time $t$ in the standard LSTM setting.

In a specific case, if each node's number of children is at most $K$ and these children can be ordered from 1 to $K$, then the $N$-ary Tree-LSTM can be applied. For node $v$, $\mathbf{h}_{vk}^t$ and $\mathbf{c}_{vk}^t$ denote the hidden state and memory cell of its $k$th child at time $t$ respectively. The transition equations are the following:

$$
\begin{aligned}
\mathbf{i}_v^t &= \text{Sigmoid}\left(\mathbf{W}^i \mathbf{x}_v^t + \sum_{l=1}^{K} \mathbf{U}_l^i \mathbf{h}_{vl}^{t-1} + \mathbf{b}^i\right), \\
\mathbf{f}_{vk}^t &= \text{Sigmoid}\left(\mathbf{W}^f \mathbf{x}_v^t + \sum_{l=1}^{K} \mathbf{U}_{kl}^f \mathbf{h}_{vl}^{t-1} + \mathbf{b}^f\right), \\
\mathbf{o}_v^t &= \text{Sigmoid}\left(\mathbf{W}^o \mathbf{x}_v^t + \sum_{l=1}^{K} \mathbf{U}_l^o \mathbf{h}_{vl}^{t-1} + \mathbf{b}^o\right), \\
\mathbf{u}_v^t &= \tanh\left(\mathbf{W}^u \mathbf{x}_v^t + \sum_{l=1}^{K} \mathbf{U}_l^u \mathbf{h}_{vl}^{t-1} + \mathbf{b}^u\right), \\
\mathbf{c}_v^t &= \mathbf{i}_v^t \odot \mathbf{u}_v^t + \sum_{l=1}^{K} \mathbf{f}_{vl}^t \odot \mathbf{c}_{vl}^{t-1}, \\
\mathbf{h}_v^t &= \mathbf{o}_v^t \odot \tanh(\mathbf{c}_v^t).
\end{aligned}
\tag{8.95}
$$

Compared to the Child-Sum Tree-LSTM, the $N$-ary Tree-LSTM introduces separate parameters for each child $k$. These parameters allow the model to learn more fine-grained representations conditioning on each node's children.

**Graph LSTM.** The two types of Tree-LSTMs can be easily adapted to the graph. The graph-structured LSTM in [148] is an example of the $N$-ary Tree-LSTM applied to the graph. However, it is a simplified version since each node in the graph has at most 2 incoming edges (from its parent and sibling predecessor). Peng et al. [92] proposes another variant of the Graph LSTM based on the relation extraction task. The main difference between graphs and trees is that edges of graphs have their labels, and [92] utilizes different weight matrices to represent different labels.

$$
\begin{aligned}
\mathbf{i}_v^t &= \text{Sigmoid}\left(\mathbf{W}^i\mathbf{x}_v^t + \sum_{k\in N_v}\mathbf{U}^i_{m(v,k)}\mathbf{h}_k^{t-1} + \mathbf{b}^i\right),\\
\mathbf{f}_{vk}^t &= \text{Sigmoid}\left(\mathbf{W}^f\mathbf{x}_v^t + \mathbf{U}^f_{m(v,k)}\mathbf{h}_k^{t-1} + \mathbf{b}^f\right),\\
\mathbf{o}_v^t &= \text{Sigmoid}\left(\mathbf{W}^o\mathbf{x}_v^t + \sum_{k\in N_v}\mathbf{U}^o_{m(v,k)}\mathbf{h}_k^{t-1} + \mathbf{b}^o\right),\\
\mathbf{u}_v^t &= \tanh\left(\mathbf{W}^u\mathbf{x}_v^t + \sum_{k\in N_v}\mathbf{U}^u_{m(v,k)}\mathbf{h}_k^{t-1} + \mathbf{b}^u\right),\\
\mathbf{c}_v^t &= \mathbf{i}_v^t \odot \mathbf{u}_v^t + \sum_{k\in N_v}\mathbf{f}_{vk}^t \odot \mathbf{c}_k^{t-1},\\
\mathbf{h}_v^t &= \mathbf{o}_v^t \odot \tanh(\mathbf{c}_v^t),
\end{aligned}
\tag{8.96}
$$

where $m(v, k)$ denotes the edge label between node $v$ and $k$.

Besides, [74] proposes a Graph LSTM network to address the semantic object parsing task. It uses the confidence-driven scheme to adaptively select the starting node and determine the node updating sequence. It follows the same idea of generalizing the existing LSTMs into the graph-structured data but has a specific updating sequence while the methods we mentioned above are agnostic to the order of nodes.

**Sentence LSTM.** Zhang et al. [152] proposes the Sentence LSTM (S-LSTM) for improving text encoding. It converts text into a graph and utilizes the Graph LSTM to learn the representation. The S-LSTM shows strong representation power in many NLP problems.

### *8.3.5 Extensions*

In this subsection, we will talk about some extensions of graph neural networks.

#### 8.3.5.1 Skip Connection

Many applications unroll or stack the graph neural network layer aiming to achieve better results as more layers (i.e., $k$ layers) make each node aggregate more information from neighbors $k$ hops away. However, it has been observed in many experiments that deeper models could not improve the performance and deeper models could even perform worse [59]. This is mainly because more layers could also propagate the noisy information from an exponentially increasing number of expanded neighborhood members.

A straightforward method to address the problem, the residual network [45], can be found from the computer vision community. Nevertheless, even with residual connections, GCNs with more layers do not perform as well as the 2-layer GCN on many datasets [59].

**Highway Network.** Rahimi et al. [96] borrows ideas from the highway network [159] and uses layer-wise gates to build a Highway GCN. The input of each layer is multiplied by the gating weights and then summed with the output:

$$\begin{aligned} T(\mathbf{h}^{(t)}) &= \text{Sigmoid}\left(\mathbf{W}^{(t)}\mathbf{h}^{(t)} + \mathbf{b}^{(t)}\right), \\ \mathbf{h}^{(t+1)} &= \mathbf{h}^{(t+1)} \odot T(\mathbf{h}^{(t)}) + \mathbf{h}^{(t)} \odot (1 - T(\mathbf{h}^{(t)})). \end{aligned} \tag{8.97}$$

By adding the highway gates, the performance peaks at four layers in a specific problem discussed in [96]. The Column Network (CLN) proposed in [94] also utilizes the highway network. However, it has a different function to compute the gating weights.

**Jump Knowledge Network.** Xu et al. [134] studies properties and resulting limitations of neighborhood aggregation schemes. It proposes the Jump Knowledge Network which could learn adaptive, structure-aware representations. The Jump Knowledge Network selects from all of the intermediate representations (which"jump" to the last layer) for each node at the last layer, which enables the model to select effective neighborhood information for each node. Xu et al. [134] uses three approaches of **concatenation**, **max-pooling**, and **LSTM-attention** in the experiments to aggregate information. The Jump Knowledge Network performs well on the experiments in social, bioinformatics, and citation networks. It can also be combined with models like Graph Convolutional Networks, GraphSAGE, and Graph Attention Networks to improve their performance.

#### 8.3.5.2 Hierarchical Pooling

In the area of computer vision, a convolutional layer is usually followed by a pooling layer to get more general features. Similar to these pooling layers, much work focuses on designing hierarchical pooling layers on graphs. Complicated and large-scale graphs usually carry rich hierarchical structures that are of great importance for node-level and graph-level classification tasks.

To explore such inner features, Edge-Conditioned Convolution (ECC) [106] designs its pooling module with the recursively downsampling operation. The downsampling method is based on splitting the graph into two components by the sign of the largest eigenvector of the Laplacian.

DIFFPOOL [144] proposes a learnable hierarchical clustering module by training an assignment matrix in each layer:

$$\mathbf{S}^{(l)} = \text{Softmax}(\text{GNN}_{l,pool}(A^{(l)}, \mathbf{V}^{(l)})), \tag{8.98}$$

where $\mathbf{V}^{(l)}$ is node features and $A^{(l)}$ is coarsened adjacency matrix of layer $l$.

#### 8.3.5.3 Neighborhood Sampling

The original graph convolutional neural network has several drawbacks. Specifically, GCN requires the full graph Laplacian, which is computationally consuming for large graphs. Furthermore, the embedding of a node at layer $L$ is computed recursively by the embeddings of all its neighbors at layer $L-1$. Therefore, the receptive field of a single node grows exponentially with respect to the number of layers, so computing gradient for a single node costs a lot. Finally, GCN is trained independently for a fixed graph, which lacks the ability for inductive learning.

**GraphSAGE** [41] is a comprehensive improvement of the original GCN. To solve the problems mentioned above, GraphSAGE replaced full graph Laplacian with learnable aggregation functions, which are crucial to perform message passing and generalize to unseen nodes. As shown in Eq. 8.86, they first aggregate neighborhood embeddings, concatenate with target node's embedding, then propagate to the next layer. With learned aggregation and propagation functions, GraphSAGE could generate embeddings for unseen nodes. Also, GraphSAGE uses neighbor sampling to alleviate the receptive field expansion.

**PinSage** [143] proposes importance-based sampling method. By simulating random walks starting from target nodes, this approach chooses the top $T$ nodes with the highest normalized visit counts.

**FastGCN** [16] further improves the sampling algorithm. Instead of sampling neighbors for each node, FastGCN directly samples the receptive field for each layer. FastGCN uses importance sampling, which the important factor is calculated as below:

$$q(v) \propto \frac{1}{|N_v|} \sum_{u \in N_v} \frac{1}{|N_u|}. \tag{8.99}$$

**Adapt.** In contrast to fixed sampling methods above, [51] introduces a parameterized and trainable sampler to perform layer-wise sampling conditioned on the former layer. Furthermore, this adaptive sampler could find optimal sampling importance and reduce variance simultaneously.

#### 8.3.5.4 Various Graph Types

In the original GNN [101], the input graph consists of nodes with label information and undirected edges, which is the simplest graph format. However, there are many variants of graphs in the world. In the following, we will introduce some methods designed to model different kinds of graphs.

**Directed Graphs.** The first variant of the graph is directed graphs. Undirected edge which can be treated as two directed edges shows that there is a relation between two nodes. However, directed edges can bring more information than undirected edges. For example, in a knowledge graph where the edge starts from the head entity and ends at the tail entity, the head entity is the parent class of the tail entity, which suggests we should treat the information propagation process from parent classes and child classes differently. DGP [55] uses two kinds of the weight matrix, $\mathbf{W}_p$ and $\mathbf{W}_c$, to incorporate more precise structural information. The propagation rule is shown as follows:

$$\mathbf{H}^{(t)} = f(D_p^{-1}A_p f(D_c^{-1}A_c\mathbf{H}^{(t-1)}\mathbf{W}_c)\mathbf{W}_p), \tag{8.100}$$

where $D_p^{-1}A_p$, $D_c^{-1}A_c$ are the normalized adjacency matrix for parents and children, respectively.

**Heterogeneous Graphs.** The second variant of the graph is a heterogeneous graph, where there are several kinds of nodes. The simplest way to process the heterogeneous graph is to convert the type of each node to a one-hot feature vector which is concatenated with the original feature.

What's more, GraphInception [151] introduces the concept of metapath into the propagation on the heterogeneous graph. With metapath, we can group the neighbors according to their node types and distances. For each neighbor group, GraphInception treats it as a subgraph in a homogeneous graph to do propagation and concatenates the propagation results from different homogeneous graphs to do a collective node representation. Recently, [128] proposes the Heterogeneous graph Attention Network (HAN) which utilizes node-level and semantic-level attention. And the model has the ability to consider node importance and metapaths simultaneously.

**Graphs with Edge Information.** In another variant of graph, each edge has additional information like the weight or the type of the edge. We list two ways to handle this kind of graphs:

Firstly, we can convert the graph to a bipartite graph where the original edges also become nodes and one original edge is split into two new edges which means there are two new edges between the edge node and begin/end nodes. The encoder of G2S [7] uses the following aggregation function for neighbors:

$$\mathbf{h}_v^{(t)} = f\left(\frac{1}{|N_v|}\sum_{u\in N_v}\mathbf{W}_r\left(\mathbf{r}_v^{(t)}\odot\mathbf{h}_u^{(t-1)}\right)+\mathbf{b}_r\right), \tag{8.101}$$

where $\mathbf{W}_r$ and $\mathbf{b}_r$ are the propagation parameters for different types of edges (relations).

Secondly, we can adapt different weight matrices for the propagation of different kinds of edges. When the number of relations is huge, r-GCN [102] introduces two kinds of regularization to reduce the number of parameters for modeling amounts of relations: *basis*- and *block-diagonal*-decomposition. With the basis decomposition, each $W_r$ is defined as follows:

$$\mathbf{W}_r = \sum_{b=1}^{B} \alpha_{rb} \mathbf{M}_b. \tag{8.102}$$

Here each $\mathbf{W}_r$ is a linear combination of basis transformations $\mathbf{M}_b \in \mathbb{R}^{d_{in} \times d_{out}}$ with coefficients $\alpha_{rb}$. In the block-diagonal decomposition, r-GCN defines each $\mathbf{W}_r$ through the direct sum over a set of low-dimensional matrices, which needs more parameters than the first one.

**Dynamic Graphs.** Another variant of the graph is dynamic graph, which has a static graph structure and dynamic input signals. To capture both kinds of information, DCRNN [71] and STGCN [147] first collect spatial information by GNNs, then feed the outputs into a sequence model like sequence-to-sequence model or CNNs. Differently, Structural-RNN [53] and ST-GCN [135] collect spatial and temporal messages at the same time. They extend static graph structure with temporal connections so they can apply traditional GNNs on the extended graphs.

## *8.3.6 Applications*

Graph neural networks have been explored in a wide range of problem domains across supervised, semi-supervised, unsupervised, and reinforcement learning settings. In this section, we simply divide the applications into three scenarios: (1) Structural scenarios where the data has explicit relational structure, such as physical systems, molecular structures, and knowledge graphs; (2) Nonstructural scenarios where the relational structure is not explicit include image, text, etc; (3) Other application scenarios such as generative models and combinatorial optimization problems. Note that we only list several representative applications instead of providing an exhaustive list. We further give some examples of GNNs in the task of fact verification and relation extraction. Figure 8.14 illustrates some application scenarios of graph neural networks.

### 8.3.6.1 Structural Scenarios

In the following, we will introduce GNN's applications in structural scenarios, where the data are naturally performed in the graph structure. For example, GNNs are widely being used in social network prediction [41, 59], traffic prediction [25, 96], recommender systems [120, 143], and graph representation [144]. Specifically, we are discussing how to model real-world physical systems with object-relationship graphs, how to predict the chemical properties of molecules and biological interaction properties of proteins and the applications of GNNs on knowledge graphs.

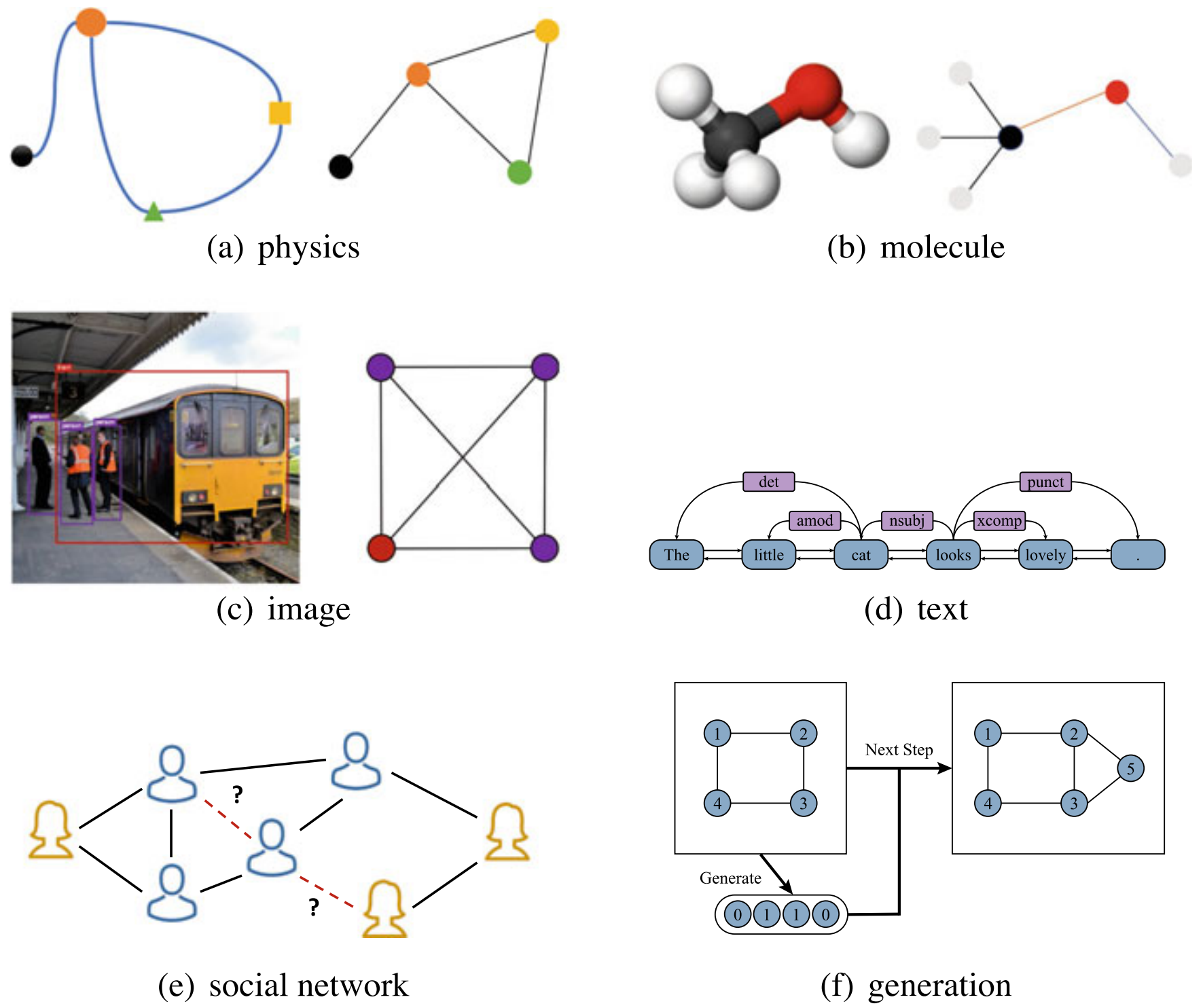

**Fig. 8.14** Application scenarios of graph neural network [155]

**Physics.** Modeling real-world physical systems is one of the most fundamental aspects of understanding human intelligence. By representing objects as nodes and relations as edges, we can perform GNN-based reasoning about objects, relations, and physics in a simplified but effective way.

Battaglia et al. [6] proposes Interaction Networks to make predictions and inferences about various physical systems. Objects and relations are first fed into the model as input. Then the model considers the interactions and physical dynamics to predict new states. They separately model relation-centric and object-centric models, making it easier to generalize across different systems. In CommNet [107], interactions are not modeled explicitly. Instead, an interaction vector is obtained by averaging all other agents' hidden vectors. VAIN [49] further introduced attentional methods into the agent interaction process, which preserves both the complexity advantages and computational efficiency as well.

Visual Interaction Networks [132] can make predictions from pixels. It learns a state code from two consecutive input frames for each object. Then, after adding their interaction effect by an Interaction Net block, the state decoder converts state codes to the next step's state.

Sanchez-Gonzalez et al. [99] proposes a Graph Network based model which could either perform state prediction or inductive inference. The inference model takes partially observed information as input and constructs a hidden graph for implicit system classification.

**Molecular Fingerprints.** Molecular fingerprints are feature vectors representing molecules, which are important in computer-aided drug design. Traditional molecular fingerprint discovering relies on heuristic methods which are hand-crafted. And GNNs can provide more flexible approaches for better fingerprints.

Duvenaud et al. [31] propose neural graph fingerprints (Neural FPs) that calculate substructure feature vectors via GCN and sum to get overall representation. The aggregation function is introduced in Eq. 8.83.

Kearnes et al. [56] further explicitly models atom and atom pairs independently to emphasize atom interactions. It introduces edge representation $\mathbf{e}_{uv}^{(t)}$ instead of aggregation function, i.e., $\mathbf{h}_{N_v}^{(t)} = \sum_{u \in N_v} \mathbf{e}_{uv}^{(t)}$. The node update function is

$$\mathbf{h}_v^{(t+1)} = \text{ReLU}(\mathbf{W}_1[\text{ReLU}(\mathbf{W}_0\mathbf{h}_u^{(t)}); \mathbf{h}_{N_v}^{(t)}]), \tag{8.103}$$

while the edge update function is

$$\mathbf{e}_{uv}^{(t+1)} = \text{ReLU}(\mathbf{W}_4[\text{ReLU}(\mathbf{W}_2\mathbf{e}_{uv}^{(t)}); \text{ReLU}(\mathbf{W}_3[\mathbf{h}_v^{(t)}; \mathbf{h}_u^{(t)}])]). \tag{8.104}$$

**Protein Interface Prediction.** Fout et al. [33] focuses on the task named protein interface prediction, which is a challenging problem with critical applications in drug discovery and design. The proposed GCN-based method, respectively, learns ligand and receptor protein residue representation and merges them for pair-wise classification.

GNN can also be used in biomedical engineering. With Protein-Protein Interaction Network, [97] leverages graph convolution and relation network for breast cancer subtype classification. Zitnik et al. [160] also suggest a GCN-based model for polypharmacy side effects prediction. Their work models the drug and protein interaction network and separately deals with edges in different types.

**Knowledge Graph.** Hamaguchi et al. [40] utilizes GNNs to solve the Out-Of-Knowledge-Base (OOKB) entity problem in Knowledge Base Completion (KBC). The OOKB entities in [40] are directly connected to the existing entities thus the embeddings of OOKB entities can be aggregated from the existing entities. The method achieves satisfying performance both in the standard KBC setting and the OOKB setting.

Wang et al. [130] utilize GCNs to solve the cross-lingual knowledge graph alignment problem. The model embeds entities from different languages into a unified embedding space and aligns them based on the embedding similarity.

#### 8.3.6.2 Nonstructural Scenarios

In this section we will talk about applications on nonstructural scenarios such as image, text, programming source code [1, 72], and multi-agent systems [49, 58, 107]. We will only give a detailed introduction to the first two scenarios due to the length limit. Roughly, there are two ways to apply the graph neural networks on nonstructural scenarios: (1) Incorporate structural information from other domains to improve the performance, for example, using information from knowledge graphs to alleviate the zero-shot problems in image tasks; (2) Infer or assume the relational structure in the scenario and then apply the model to solve the problems defined on graphs, such as the method in [152] which models text as graphs.

**Image classification.** Image classification is a fundamental and essential task in the field of computer vision, which attracts much attention and has many famous datasets like ImageNet [62]. Recent progress in image classification benefits from big data and the strong power of GPU computation, which allows us to train a classifier without extracting structural information from images. However, **zero-shot** and **few-shot learning** become more and more popular in the field of image classification, because most models can achieve similar performance with enough data. There are several works leveraging graph neural networks to incorporate structural information in image classification.

First, knowledge graphs can be used as extra information to guide zero-shot recognition classification [55, 129]. Wang et al. [129] builds a knowledge graph where each node corresponds to an object category and takes the word embeddings of nodes as input for predicting the classifier of different categories. As the over-smoothing effect happens with the deep depth of convolution architecture, the 6-layer GCN used in [129] will wash out much useful information in the representation. To solve the smoothing problem in the propagation of GCN, [55] uses single-layer GCN with a larger neighborhood, which includes both one-hop and multi-hop nodes in the graph. And it proved effective in building a zero-shot classifier with existing ones.

Except for the knowledge graph, the similarity between images in the dataset is also helpful for few-shot learning [100]. Satorras and Estrach [100] propose to build a weighted fully connected image network based on the similarity and do message passing in the graph for few-shot recognition. As most knowledge graphs are large for reasoning, [77] selects some related entities to build a subgraph based on the result of object detection and apply GGNN to the extracted graph for prediction. Besides, [66] proposes to construct a new knowledge graph where the entities are all the categories. And, they defined three types of label relations: super-subordinate, positive correlation, and negative correlation and propagate the confidence of labels in the graph directly.

**Visual reasoning.** Computer vision systems usually need to perform reasoning by incorporating both spatial and semantic information. So it is natural to generate graphs for reasoning tasks.

A typical visual reasoning task is Visual Question Answering (VQA), [114], respectively, constructs image scene graph and question syntactic graph. Then it applies GGNN to train the embeddings for predicting the final answer. Despite spatial connections among objects, [87] builds the relational graphs conditioned on the questions. With knowledge graphs, [83, 131] can perform finer relation exploration and a more interpretable reasoning process.

Other applications of visual reasoning include object detection, interaction detection, and region classification. In object detection [39, 50], GNNs are used to calculate RoI features; In interaction detection [53, 95], GNNs are message-passing tools between human and objects; In region classification [18], GNNs perform reasoning on graphs which connects regions and classes.

**Text Classification.** Text classification is an essential and classical problem in natural language processing. The classical GCN models [4, 28, 41, 47, 59, 82] and GAT model [122] are applied to solve the problem, but they only use the structural information between the documents and they do not use much text information.

Peng et al. [91] propose a graph-CNN-based deep learning model. It first turns texts to graph-of-words, then the graph convolution operations in [347] are used on the word graph. Zhang et al. [152] propose the Sentence LSTM to encode text. The whole sentence is represented in a single state which contains a global sentence-level state and several substates for individual words. It uses the global sentence-level representation for classification tasks.

These methods either view a document or a sentence as a graph of word nodes or rely on the document citation relation to construct the graph. Yao et al. [142] regard the documents and words as nodes to construct the corpus graph (hence heterogeneous graph) and uses the Text GCN to learn embeddings of words and documents. Sentiment classification could also be regarded as a text classification problem and a Tree-LSTM approach is proposed by [109].

**Sequence Labeling.** As each node in GNNs has its hidden state, we can utilize the hidden state to address the sequence labeling problem if we consider every word in the sentence as a node. Zhang et al. [152] utilize the Sentence LSTM to label the sequence. It has conducted experiments on POS-tagging and NER tasks and achieves promising performance.

Semantic role labeling is another task of sequence labeling. Marcheggiani and Titov [76] propose a Syntactic GCN to solve the problem. The Syntactic GCN which operates on the direct graph with labeled edges is a special variant of the GCN [59]. It uses edge-wise gates that enable the model to regulate the contribution of each dependency edge. The Syntactic GCNs over syntactic dependency trees are used as sentence encoders to learn latent feature representations of words in the sentence. Marcheggiani and Titov [76] also reveal that GCNs and LSTMs are functionally complementary in the task.

#### 8.3.6.3 Other Scenarios

Besides structural and nonstructural scenarios, there are some other scenarios where graph neural networks play an important role. In this subsection, we will introduce generative graph models and combinatorial optimization with GNNs.

**Generative Models.** Generative models for real-world graphs have drawn significant attention for its essential applications, including modeling social interactions, discovering new chemical structures, and constructing knowledge graphs. As deep learning methods have a powerful ability to learn the implicit distribution of graphs, there is a surge in neural graph generative models recently.

NetGAN [104] is one of the first works to build a neural graph generative model, which generates graphs via random walks. It transformed the problem of graph generation to the problem of walk generation, which takes the random walks from a specific graph as input and trains a walk generative model using GAN architecture. While the generated graph preserves essential topological properties of the original graph, the number of nodes is unable to change in the generating process, which is as same as the original graph. GraphRNN [146] generate the adjacency matrix of a graph by generating the adjacency vector of each node step by step, which can output required networks having different numbers of nodes.

Instead of generating adjacency matrix sequentially, MolGAN [27] predict a discrete graph structure (the adjacency matrix) at once and utilizes a permutation-invariant discriminator to solve the node variant problem in the adjacency matrix. Besides, it applies a reward network for RL-based optimization towards desired chemical properties. What is more, [75] proposes constrained variational autoencoders to ensure the semantic validity of generated graphs. Moreover, GCPN [145] incorporates domain-specific rules through reinforcement learning.

Li et al. [73] propose a model that generates edges and nodes sequentially and utilize a graph neural network to extract the hidden state of the current graph, which is used to decide the action in the next step during the sequential generative process.

**Combinatorial optimization.** Combinatorial optimization problems over graphs are a set of NP-hard problems that attract much attention from scientists of all fields. Some specific problems like Traveling Salesman Problem (TSP) have got various heuristic solutions. Recently, using a deep neural network for solving such problems has been a hotspot, and some of the solutions further leverage graph neural networks because of their graph structure.

Bello et al. [9] first propose a deep learning approach to tackle TSP. Their method consists of two parts: a Pointer Network [123] for parameterizing rewards and a policy gradient [108] module for training. This work has been proved to be comparable with traditional approaches. However, Pointer Networks are designed for sequential data like texts, while order-invariant encoders are more appropriate for such work.

Khalil et al. [57] and Kool and Welling [61] improve the above method by including graph neural networks. The former work first obtains the node embeddings from structure2vec [26] then feeds them into a Q-learning module for making decisions. The latter one builds an attention-based encoder-decoder system. By replacing the reinforcement learning module with an attention-based decoder, it is more efficient for training. These work achieved better performance than previous algorithms, which proved the representation power of graph neural networks.

Nowak et al. [88] focus on Quadratic Assignment Problem, i.e., measuring the similarity of two graphs. The GNN-based model learns node embeddings for each graph independently and matches them using an attention mechanism. Even in situations where traditional relaxation-based methods may perform not well, this model still shows satisfying performance.

#### 8.3.6.4 Example: GNNs for Fact Verification

Due to the rapid development of Information Extraction (IE), huge volumes of data have been extracted. How to automatically verify the data becomes a vital problem for various data-driven applications, e.g., knowledge graph completion [126] and open domain question answering [15]. Hence, many recent research efforts have been devoted to Fact Verification (FV), which aims to verify given claims with the evidence retrieved from plain text. More specifically, given a claim, an FV system is asked to label it as “SUPPORTED”, “REFUTED”, or “NOT ENOUGH INFO”, which indicates that the evidence can support, refute, or is not sufficient for the claim. An example of the FV task is shown in Table 8.5.

Existing FV methods formulate FV as a Natural Language Inference (NLI) [3] task. However, they utilize simple evidence combination methods such as concatenating the evidence or just dealing with each evidence-claim pair. These methods are unable to grasp sufficient relational and logical information among the evidence. In fact, many claims require to simultaneously integrate and reason over several pieces of evidence for verification. As shown in Table 8.5, for this particular example, we cannot verify the given claim by checking any evidence in isolation. The claim can be verified only by understanding and reasoning over multiple evidence.

**Table 8.5** A case of the claim that requires integrating multiple evidence to verify. The representation for evidence “{*DocName*, *LineNum*}” means the evidence is extracted from the document “*DocName*” and of which the line number is *LineNum*

| |
|---|
| **Claim:**<br>***Al Jardine*** is an ***American rhythm guitarist*** |
| **Truth evidence:**<br>{Al Jardine, 0}, {Al Jardine, 1} |
| **Retrieved evidence:**<br>{Al Jardine, 1}, {Al Jardine, 0}, {Al Jardine, 2}, {Al Jardine, 5}, {Jardine, 42} |
| **Evidence:**<br>(1) ***He is best known as the band’s rhythm guitarist***, and for occasionally singing lead vocals on singles such as “Help Me, Rhonda” (1965), “Then I Kissed Her” (1965), and “Come Go with Me” (1978)<br>(2) ***Alan Charles Jardine*** (born September 3, 1942) is ***an American musician***, singer and songwriter who co-founded the Beach Boys<br>(3) In 2010, Jardine released his debut solo studio album, A Postcard from California<br>(4) In 1988, Jardine was inducted into the Rock and Roll Hall of Fame as a member of the Beach Boys<br>(5) Ray Jardine American rock climber, lightweight backpacker, inventor, author, and global adventurer |
| **Label:** SUPPORTED |

To integrate and reason over information from multiple pieces of evidence, [156] proposes a graph-based evidence aggregating and reasoning (GEAR) framework. Specifically, [156] first builds a fully connected evidence graph and encourages information propagation among the evidence. Then, GEAR aggregates the pieces of evidence and adopts a classifier to decide whether the evidence can support, refute, or is not sufficient for the claim. Intuitively, by sufficiently exchanging and reasoning over evidence information on the evidence graph, the proposed model can make the best of the information for verifying claims. For example, by delivering the information "Los Angeles County is the most populous county in the USA" to "the Rodney King riots occurred in Los Angeles County" through the evidence graph, the synthetic information can support "The Rodney King riots took place in the most populous county in the USA". Furthermore, we adopt an effective pretrained language representation model BERT [29] to better grasp both evidence and claim semantics.

Zhou et al. [156] employ a three-step pipeline with components for document retrieval, sentence selection, and claim verification to solve the task. In the document retrieval and sentence selection stages, they simply follow the method from [44] since their method has the highest score on evidence recall in the former FEVER-shared task. And they propose the GEAR framework in the final claim verification stage. The full pipeline is illustrated in Fig. 8.15.

Given a claim and the retrieved evidence, GEAR first utilizes a **sentence encoder** to obtain representations for the claim and the evidence. Then it builds a fully connected evidence graph and uses an **Evidence Reasoning Network** (ERNet) to propagate information among evidence and reason over the graph. Finally, it utilizes an **evidence aggregator** to infer the final results.

**Sentence Encoder.** Given an input sentence, GEAR employs BERT [29] as the sentence encoder by extracting the final hidden state of the [CLS] token as the representation, where [CLS] is the special classification token in BERT.

$$\begin{aligned} \mathbf{e}_i &= \text{BERT}\left(e_i, c\right), \\ \mathbf{c} &= \text{BERT}\left(c\right). \end{aligned} \tag{8.105}$$

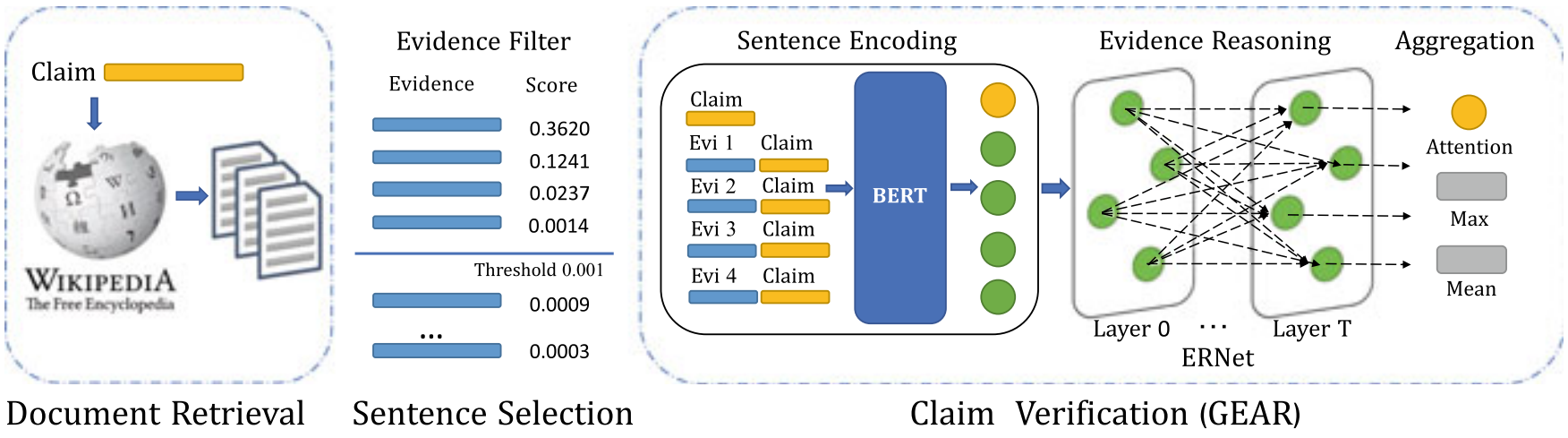

**Fig. 8.15** The pipeline used in [156]. The GEAR framework is illustrated in the claim verification section

**Evidence Reasoning Network.** To encourage the information propagation among evidence, GEAR builds a fully connected evidence graph where each node indicates a piece of evidence. It also adds *self-loop* to every node because each node needs the information from itself in the message propagation process. We use $h^t = \{h_1^t, h_2^t, \ldots, h_N^t\}$ to represent the hidden states of nodes at layer $t$. The initial hidden state of each evidence node $h_i^0$ is initialized by the evidence presentation: $h_i^0 = e_i$.

Inspired by recent work on semi-supervised graph learning and relational reasoning [59, 90, 122], Zhou et al. [156] propose an Evidence Reasoning Network (ERNet) to propagate information among the evidence nodes. It first uses an MLP to compute the attention coefficients between a node $i$ and its neighbor $j$ ($j \in \mathscr{N}_i$),

$$y_{ij} = \mathbf{W}_1^{(t-1)}(\text{ReLU}(\mathbf{W}_0^{(t-1)}[\mathbf{h}_i^{(t-1)}; \mathbf{h}_j^{(t-1)}])), \tag{8.106}$$

where $\mathscr{N}_i$ denotes the set of neighbors of node $i$, $\mathbf{W}_0^{(t-1)}$ and $\mathbf{W}_1^{(t-1)}$ are weight matrices, and $[\cdot;\cdot]$ denotes concatenation operation.

Then, it normalizes the coefficients using the softmax function

$$\alpha_{ij} = \text{Softmax}_j(y_{ij}) = \frac{\exp(y_{ij})}{\sum_{k \in N_i} \exp(y_{ik})}. \tag{8.107}$$

Finally, the normalized attention coefficients are used to compute a linear combination of the neighbor features and thus we obtain the features for node $i$ at layer $t$,

$$\mathbf{h}_i^{(t)} = \sum_{j \in N_i} \alpha_{ij} \mathbf{h}_j^{(t-1)}. \tag{8.108}$$

By stacking $T$ layers of ERNet, [156] assumes that each evidence could grasp enough information by communicating with other evidence.

**Evidence Aggregator.** Zhou et al. [156] employ an evidence aggregator to gather information from different evidence nodes and obtain the final hidden state $o$. The aggregator may utilize different aggregating strategies and [156] suggests three aggregators:

Attention Aggregator. Zhou et al. [156] use the representation of the claim $\mathbf{c}$ to attend the hidden states of evidence and get the final aggregated state $\mathbf{o}$.

$$\begin{aligned} y_j &= \mathbf{W}_1'(\text{ReLU}(\mathbf{W}_0'[\mathbf{c}; \mathbf{h}_j^\top])), \\ \alpha_j &= \text{Softmax}(y_j) = \frac{\exp(y_j)}{\sum_{k=1}^N \exp(y_k)}, \\ \mathbf{o} &= \sum_{k=1}^N \alpha_k \mathbf{h}_k^\top. \end{aligned} \tag{8.109}$$

Max Aggregator. The max aggregator performs the *element-wise* max operation among hidden states.

$$\mathbf{o} = \mathrm{Max}(\mathbf{h}_1^\top, \mathbf{h}_2^\top, \ldots, \mathbf{h}_N^\top). \tag{8.110}$$

Mean Aggregator. The mean aggregator performs the *element-wise* mean operation among hidden states.

$$\mathbf{o} = \mathrm{Mean}(\mathbf{h}_1^\top, \mathbf{h}_2^\top, \ldots, \mathbf{h}_N^\top). \tag{8.111}$$

Once the final state $o$ is obtained, GEAR employs a one-layer MLP to get the final prediction $\mathbf{l}$.

$$\mathbf{l} = \mathrm{Softmax}(\mathrm{ReLU}(\mathbf{Wo} + \mathbf{b})), \tag{8.112}$$

where $\mathbf{W}$ and $\mathbf{b}$ are parameters.

Zhou et al. [156] conduct experiments on the large-scale benchmark dataset for Fact Extraction and VERification (FEVER) [115]. Experimental results show that the proposed framework outperforms recent state-of-the-art baseline systems. The further case study indicates that the framework could better leverage multi-evidence information and reason over the evidence for FV.

## 8.4 Summary

In this chapter, we have introduced network representation learning, which turns the network structure information into the continuous vector space and make deep learning techniques possible on network data.

Unsupervised network representation learning comes first during the development of NRL. Spectral Clustering, DeepWalk, LINE, GraRep, and other methods utilize the network structure for vertex embedding learning. Afterward, TADW incorporates text information into NRL under the framework of matrix factorization. The NEU algorithm then moves one step forward and proposes a general method to improve the quality of any learned network embeddings. Other unsupervised methods also consider preserving specific properties of the network topology, e.g., community and asymmetry.

Recently, semi-supervised NRL algorithms have attracted much attention. This kind of methods focus on a specific task such as classification and use the labels of the training set to improve the quality of network embeddings. Node2vec, MMDW, and many other methods including the family of Graph Neural Networks (GNNs) are proposed for this end. Semi-supervised algorithms can achieve better results as they can take advantage of more information from the specific task.

For further understanding of network representation learning, you can also find more related papers in this paper list https://github.com/thunlp/GNNPapers. There are also some recommended surveys and books including the following:

- Cui et al. A survey on network embedding [24].
- Goyal and Ferrara. Graph embedding techniques, applications, and performance: A survey [37].
- Zhang et al. Network representation learning: A survey [149].
- Wu et al. A comprehensive survey on graph neural networks [133].
- Zhou et al. Graph neural networks: A review of methods and applications [155].
- Zhang et al. Deep learning on graphs: A survey [154].

In the future, for better network representation learning, some directions are requiring further efforts:

(1) **More Complex and Realistic Networks**. An intriguing direction would be the representation of learning on heterogeneous and dynamic networks where most real-world network data fall into this category. The vertices and edges in a heterogeneous network may belong to different types. Networks in real life are also highly dynamic, e.g., the friendship between Facebook users may establish and disappear. These characteristics require the researchers to design specific algorithms for them. Network embedding learning on dynamic network structures is, therefore, an important task. There have been some works proposed [14, 105] for much more complex and realistic settings.

(2) **Deeper Model Architectures**. Conventional deep neural networks can stack hundreds of layers to get better performance because the deeper structure has more parameters and may improve the expressive power significantly. However, NRL and GNN models are usually shallow. In fact, most of them have no more than three layers. Taking GCN as an example, as experiments in [70] show, stacking multiple GCN layers will result in over-smoothing: the representations of all vertices will converge to the same. Although some researchers have managed to tackle this problem [70, 125] to some extents, it remains to be a limitation of NRL. Designing deeper model architectures is an exciting challenge for future research, and will be a considerable contribution to the understanding of NRL.

(3) **Scalability**. Scalability determines whether an algorithm is able to be applied to practical use. How to apply NRL methods in real-world web-scale scenarios such as social networks or recommendation systems has been an essential problem for most network embedding algorithms. Scaling up NRL methods especially GNN is difficult because many core steps are computationally consuming in a big data environment. For example, network data are not regular Euclidean, and each node has its own neighborhood structure. Therefore, batch tricks cannot be easily applied. Moreover, computing graph Laplacian is also unfeasible when there are millions or even billions of nodes and edges. Several works has proposed their solutions to this problem [143, 153, 157] and we are paying close attention to the progress.

# Chapter 9
# Cross-Modal Representation

**Abstract** Cross-modal representation learning is an essential part of representation learning, which aims to learn latent semantic representations for modalities including texts, audio, images, videos, etc. In this chapter, we first introduce typical cross-modal representation models. After that, we review several real-world applications related to cross-modal representation learning including image captioning, visual relation detection, and visual question answering.

## 9.1 Introduction

As introduced in Wikipedia, *a modality is the classification of a single independent channel of sensory input/output between a computer and a human*. To be more general, modalities are different means of information exchange between human beings and the real world. The classification is usually based on the form in which information is presented to a human. Typical modalities in the real world include texts, audio, images, videos, etc.

Cross-modal representation learning is an important part of representation learning. In fact, artificial intelligence is inherently a multi-modal task [30]. Human beings are exposed to multi-modal information every day, and it is normal for us to integrate information from different modalities and make comprehensive judgments. Furthermore, different modalities are not independent, but they have correlations more or less. For example, the judgment of a syllable is made by not only the sound we hear but also the movement of the lips and tongue of the speaker we see. An experiment in [48] shows that a voiced /ba/ with a visual /ga/ is perceived by most people as a /da/. Another example is human beings' ability to consider the 2D image and 3D scan of the same object together and reconstruct its structure: correlations between image and scan can be found based on the fact that a discontinuity of depth in the scan usually indicates a sharp line in the image [52]. Inspired by this, it is natural for us to consider the possibility of combining inputs from multi-modalities in our artificial intelligence systems and generate cross-modal representation.

Ngiam et al. [52] explore the probability of merging multi-modalities into one learning task. The authors divide a typical machine learning task into three stages:

feature learning, supervised learning, and prediction. And they further propose four kinds of learning settings for multi-modalities:

(1) **Single-modal learning**: all stages are all done on just one modality.

(2) **Multi-modal fusion**: all stages are all done with all modalities available.

(3) **Cross-modal learning**: in the feature learning stage, all modalities are available, but in supervised learning and prediction, only one modality is used.

(4) **Shared representation learning**: in the feature learning stage, all modalities are available. In supervised learning, only one modality is used, and in prediction, a different modality is used.

Experiments show promising results for these multi-modal tasks. When more modalities are provided (such as multi-modal fusion, cross-modal learning, and shared representation learning), the performance of the system is generally better.

In the following part of this chapter, we will first introduce cross-modal representation models, which are fundamental parts of cross-modal representation learning in NLP. And then, we will introduce several critical applications, such as image captioning, visual relationship detection, and visual question answering.

## 9.2 Cross-Modal Representation

Cross-modal representation learning aims to build embeddings using information from multiple modalities. Existing cross-modal representation models involving text modality can be generally divided into two categories: (1) [30, 77] try to fuse information from different modalities into unified embeddings (e.g., visually grounded word representations). (2) Researchers also try to build embeddings for different modalities in a common semantic space, which allows the model to compute cross-modal similarity. Such cross-modal similarity can be further utilized for downstream tasks, such as zero-shot recognition [5, 14, 18, 53, 65] and cross-media retrieval [23, 55]. In this section, we will introduce these two kinds of cross-modal representation models, respectively.

### *9.2.1 Visual Word2vec*

Computing word embeddings is a fundamental task in representation learning for natural language processing. Typical word embedding models (like Word2vec [49]) are trained on a text corpus. These models, while being extremely successful, cannot discover implicit semantic relatedness between words that could be expressed in other modalities. Kottur et al. [30] provide an example: even though `eat` and `stare at` seem are unrelated from text, images might show that when people are `eating` something they would also tend to `stare at` it. This implies that considering other modalities when constructing word embeddings may help capture more implicit semantic relatedness.

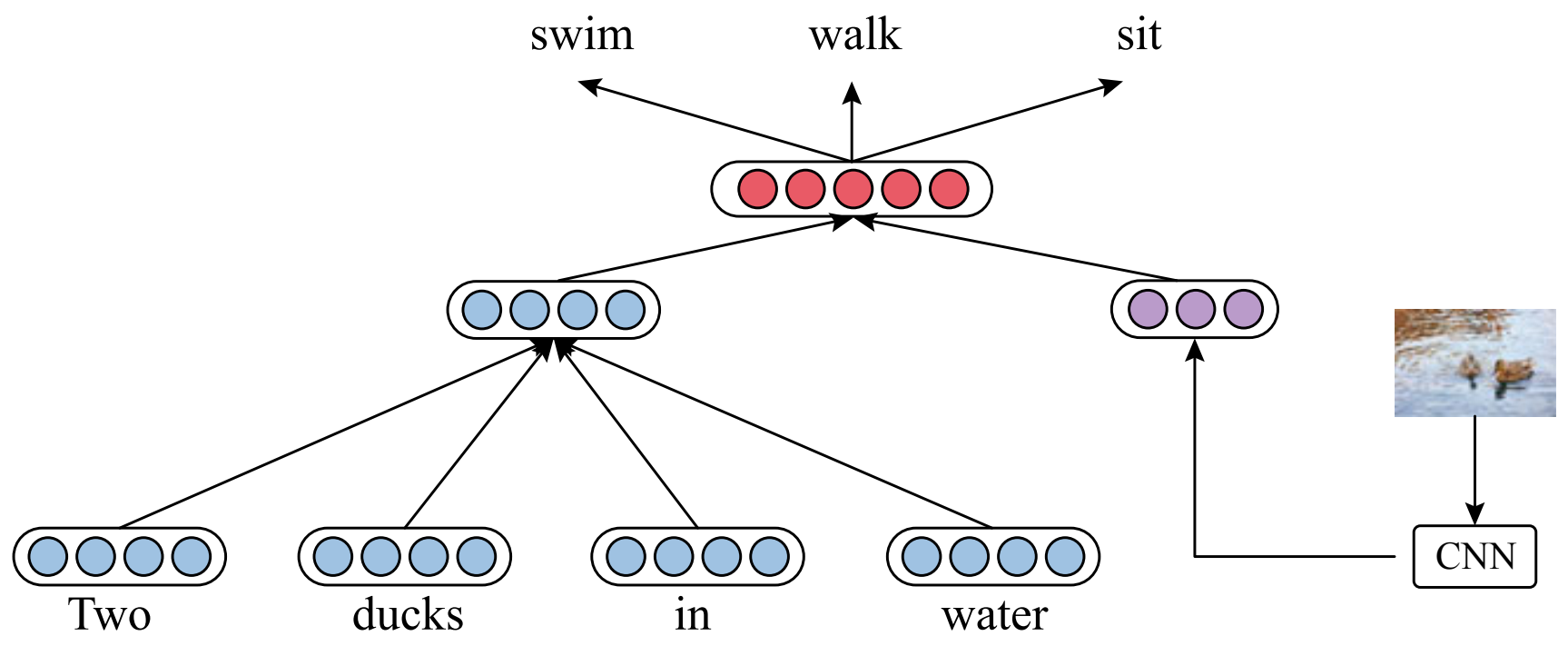

**Fig. 9.1** The architecture for word embedding with global visual context

Vision, being one of the most critical modalities, has attracted attention from researchers seeking to improve word representation. Several models that incorporate visual information and improve word embeddings with vision have been proposed. We introduce two typical word representation models incorporating visual information in the following.

#### 9.2.1.1 Word Embedding with Global Visual Context

Xu et al. [77] propose a model that makes a natural attempt to incorporate visual features. It claims that in most word representation models, only local context information (e.g., trying to predict a word using neighboring words and phrases) is considered. Global text information (e.g., the topic of the passage), on the other hand, is often neglected. This model extends a simple local context model by using visual information as global features (see Fig. 9.1).

The input of the model is an image $I$ and a sequence describing it. It is based on a simple local context language model: when we consider a certain word $w_t$ in a sequence, its local feature is the average of embeddings of words in a window, i.e., $\{w_{t-k}, \ldots, w_{t-1}, w_{t+1}, \ldots, w_{t+k}\}$. The visual feature is computed directly from the image $I$ using a CNN and then used as the global feature. The local feature and the global feature are then concatenated into a vector $\mathbf{f}$. The predicted probability of a word $w_t$ (in this blank) is the softmax normalized product of $\mathbf{f}$ and the word embedding $\mathbf{w}_t$:

$$o_{w_t} = \mathbf{w}_t^T \mathbf{f}, \tag{9.1}$$

$$P(w_t | w_{t-k}, \ldots, w_{t-1}, w_{t+1}, \ldots, w_{t+k}; I) = \frac{\exp(o_{w_t})}{\sum_i \exp(o_{w_i})}. \tag{9.2}$$

The model is optimized by maximizing the average of log probability:

$$\mathscr{L} = \frac{1}{T}\sum_{t=k}^{T-k} \log P(w_t | w_{t-k}, \ldots, w_{t-1}, w_{t+1}, \ldots, w_{t+k}; I). \tag{9.3}$$

The classification error will be back-propagated to local text vector (i.e., word embeddings), visual vector, and all model parameters. This accomplishes jointly learning for a set of word embeddings, a language model, and the model used for visual encoding.

#### 9.2.1.2 Word Embedding with Abstract Visual Scene

Kottur et al. [30] also propose a neural model to capture fine-grained semantics from visual information. Instead of focusing on literal pixels, the abstract scene behind the vision is considered. The model takes a pair of the visual scene and a related word sequence $(I, w)$ as input. At each training step, a window is used upon the word sequence $w$, forming a subsequence $S_w$. All the words in $S_w$ will be fed into the input layer using one-hot encoding, and therefore the dimension of the input layer is $|\mathscr{V}|$, which is also the size of the vocabulary. The words are then transformed into their embeddings, and the hidden layer is the average of all these embeddings. The size of the hidden layer is $N_H$, which is also the dimension of the word embeddings. The hidden layer and the output layer are connected by a full connection matrix of dimension $N_H * N_K$ and a softmax function. The output layer can be regarded as a probability distribution over a discrete-valued function $g(\cdot)$ of the visual scene $I$ (details will be given in the following paragraph). The entire model is optimized by minimizing the objective function:

$$\mathscr{L} = -\log P(g(w)|S_w). \tag{9.4}$$

The most important part of the model is the function $g(\cdot)$. It maps the visual scene $I$ into the set $\{1, 2, \ldots, N_K\}$, which indicates what kind of abstract scene it is. In practice, it is learned offline using K-means clustering, and each cluster represents the semantics of one kind of visual scenes, consequently, the word sequence $w$, which is designed to be related to the scene.

### 9.2.2 *Cross-Modal Representation for Zero-Shot Recognition*

Large-scale datasets partially support the success of deep learning methods. Even though the scales of datasets continue to grow larger, and more categories are involved, the annotation of datasets is expensive and time-consuming. For many categories, there are very limited or even no instances, which restricts the scalability of recognition systems.

Zero-shot recognition is proposed to solve the problem as mentioned above, which aims to classify instances of categories that have not been seen during training. Many works propose to utilize cross-modal representation for zero-shot image classification [5, 14, 18, 53, 65]. Specifically, image representation and category representation are embedded into a common semantic space, where similarities between image and category representations can serve for further classification. For example, in such a common semantic space, the embedding of an image of `cat` is expected to be closer to the embedding of category `cat` than the embedding of category `truck`.

#### 9.2.2.1 Deep Visual-Semantic Embedding

The challenge of zero-shot learning lies in the absence of instances of unseen categories, which makes it challenging to obtain well-performed classifiers of unseen categories. Frome et al. [18] present a model that utilizes both labeled images and information from the large-scale plain text for zero-shot image classification. They try to leverage semantic information from word embeddings and transfer it to image classification systems.

Their model is motivated by the fact that word embeddings incorporate semantic information of concepts or categories, which can be potentially utilized as classifiers of corresponding categories. Similar categories cluster well in semantic space. For example, in word embedding space, the nearest neighbors of the term `tiger shark` are similar kinds of sharks, such as `bull shark, blacktip shark, sandbar shark, and oceanic whitetip shark`. In addition, boundaries between different clusters are clear. The aforementioned properties indicate that word embeddings can be further utilized as classifiers for recognition systems.

Specifically, the model first pretrains word embeddings using the Skip-gram text model on large-scale Wikipedia articles. For visual feature extraction, the model pretrains a deep convolutional neural network for 1, 000 object categories on ImageNet. The pretrained word embeddings and the convolutional neural network are used to initialize the proposed Deep Visual-Semantic Embedding model (DeViSE).

To train the proposed model, they replace the softmax layer of the pretrained convolutional neural network with a linear projection layer. The model is trained to predict the word embeddings of categories for images using a hinge ranking loss:

$$\mathscr{L}(I, y) = \sum_{j \neq y} \max[0, \gamma - \mathbf{w}_y \mathbf{M}\mathbf{I} + \mathbf{w}_j \mathbf{M}\mathbf{I}], \tag{9.5}$$

where $\mathbf{w}_y$ and $\mathbf{w}_j$ are the learned word embeddings of the positive label and sampled negative label, respectively, $\mathbf{I}$ denotes the feature of the image obtained from the convolutional neural network, $\mathbf{M}$ is the trainable parameters in linear projection layer, and $\gamma$ is a hyperparameter in hinge ranking loss. Given an image, the objective requires the model to produce a higher score for the correct label than randomly chosen labels, where the score is defined as the dot product of the projected image feature and word embedding of terms.

At test time, given a test image, the score of each possible category is obtained using the same approach during training. Note that a crucial difference at test time is that the classifiers (word embeddings) are expanded to all possible categories, including unseen categories. Thus the model is capable of predicting unseen categories.

Experiment results show that DeViSE can make zero-shot predictions with more semantically reasonable errors, which means that even if the prediction is not exactly correct, it is semantically related to the ground truth class. However, a drawback is that although the model can utilize semantic information in word embeddings to make zero-shot image classification, using word embeddings as classifiers restricts the flexibility of the model, which results in inferior performance in the original 1, 000 categories compared to the original softmax classifier.

#### 9.2.2.2 Convex Combination of Semantic Embeddings

Inspired by DeViSE, [53] proposes a model ConSE that tries to utilize semantic information from word embeddings for zero-shot classification. A vital difference to DeViSE is that they obtain the semantic embedding of test image using a convex combination of word embeddings of seen categories. The score of the corresponding category determines the weights of the composing word embeddings.

Specifically, they train a deep convolutional neural network on seen categories. At test time, given a test image $I$ (possibly from unseen categories), they obtain the top $T$ confident predictions of seen categories, where $T$ is a hyperparameter. Then the semantic embedding $f(I)$ of $I$ is determined by the convex combination of semantic embeddings of the top $T$ confident categories, which can be formally defined as follows:

$$f(I) = \frac{1}{Z} \sum_{t=1}^{T} P(\hat{y}_0(I,t)|I) \cdot \mathbf{w}(\hat{y}_0(I,t)), \tag{9.6}$$

where $\hat{y}_0(I,t)$ is the $t$th most confident training label for $I$, $\mathbf{w}(\hat{y}_0(I,t))$ is the semantic embedding (word embedding) of $\hat{y}_0(I,t)$, and $Z$ is a normalization factor given by

$$Z = \sum_{t=1}^{T} P(\hat{y}_0(I,t)|I). \tag{9.7}$$

After obtaining the semantic embedding $f(I)$, the score of the category $m$ is given by the cosine similarity of $f(I)$ and $\mathbf{w}(m)$.

The motivation of ConSE is that they assume novel categories can be modeled as the convex combination of seen categories. If the model is highly confident about a prediction, (i.e., $P(\hat{y}_0(I,1)|I) \approx 1$), the semantic embedding $f(I)$ will be close to $\mathbf{w}(\hat{y}_0(I,1))$. If the predictions are ambiguous, (e.g., $P(\texttt{tiger}|I) = 0.5$, $P(\texttt{lion}|I) = 0.5$), the semantic embedding $f(I)$ will be between $\mathbf{w}(\texttt{lion})$ and $\mathbf{w}(\texttt{tiger})$. And they expect the semantic embedding $f(I) = 0.5\mathbf{w}(\texttt{lion}) +$ $0.5\mathbf{w}(\texttt{tiger})$ to be close to the semantic embedding $\mathbf{w}(\texttt{liger})$ (a hybrid cross between `lions` and `tigers`).

Although ConSE and DeViSE share many similarities, there are also some crucial differences. DeViSE replaces the softmax layer of the pretrained visual model with a projection layer, while ConSE preserves the softmax layer. ConSE does not need to be further trained and uses a convex combination of semantic embeddings to perform zero-shot classification at test time. Experiment results show that ConSE outperforms DeViSE on unseen categories, indicating better generalization capability. However, the performance of ConSE on seen categories is not as competitive as DeViSE and the original softmax classifier.

#### 9.2.2.3 Cross-Modal Transfer

Socher et al. [65] present a cross-modal representation model for zero-shot recognition. In their model, all word vectors are initialized with pretrained 50-dimensional word vectors and are kept fixed during training. Each image is represented by a vector $\mathbf{I}$ constructed by a deep convolutional neural network. They first project an image into semantic word spaces by minimizing

$$\mathscr{L}(\Theta) = \sum_{y \in Y_s} \sum_{I^{(i)} \in X_y} \|\mathbf{w}_y - \theta^{(2)} f(\theta^{(1)} \mathbf{I}^{(i)})\|^2, \tag{9.8}$$

where $Y_s$ denotes the set of images' classes which can be seen in training data, $X_y$ denotes the set of images' vectors of class $y$, $\mathbf{w}_y$ denotes the word vector of class $y$, and $\Theta = (\theta^{(1)}, \theta^{(2)})$ denotes parameters of the 2-layer neural network with $f(\cdot) = \tanh(\cdot)$ as activation function.

They observe that instances from unseen categories are usually outliers of the complete data manifold. Following this observation, they first classify an instance into seen and unseen categories via outlier detection methods. Then the instance is classified using corresponding classifiers.

Formally, they marginalize a binary random variable $V \in \{s, u\}$ which denotes whether an instance belongs to seen categories or unseen categories separately, which means probability is given as

$$P(y|I) = \sum_{V \in \{s,u\}} P(y|V, I) P(V|I). \tag{9.9}$$

For seen image classes, they simply use softmax classifier to determine $P(y|s, I)$, while for unseen classes, they assume an isometric Gaussian distribution around each of the novel class word vectors and assign classes based on their likelihood. To detect novelty, they calculate a *Local Outlier Probability* by Gaussian error function.

## 9.2.3 Cross-Modal Representation for Cross-Media Retrieval

Learning cross-modal representation from different modalities in a common semantic space allows one to easily compute cross-modal similarities, which can facilitate many important cross-modal tasks, such as cross-media retrieval. With the rapid growth of multimedia data such as text, image, video, and audio on the Internet, the need to retrieve information across different modalities has become stronger. Cross-media retrieval is an important task in the multimedia area, which aims to perform retrieval across different modalities such as text and image. For example, a user may submit an image of a white horse, and retrieve relevant information from different modalities, such as textual descriptions of horses, and vice versa.

A significant challenge of cross-modal retrieval is the domain discrepancies between different modalities. Besides, for a specific area of interest, cross-modal data can be insufficient, which limits the performance of existing cross-modal retrieval methods. Many works have focused on the challenges as mentioned above in cross-modal retrieval [23, 24].

### 9.2.3.1 Cross-Modal Hybrid Transfer Network

Huang et al. [24] present a framework that tries to relieve the cross-modal data sparsity problem by transfer learning. They propose to leverage knowledge from a large-scale single-modal dataset to boost the model training on the small-scale dataset. The massive auxiliary dataset is denoted as the source domain, and the small-scale dataset of interest is denoted as the target domain. In their work, they adopt ImageNet [12], a large-scale image database as the source domain.

Formally, a training set consists of data from source domain $Src = \{I_s^p, y_s^p\}_{p=1}^P$ and target domain $Tar_{tr} = \{(I_s^j, t_s^j), y_s^j\}_{j=1}^J$, where $(I, t)$ is the image/text pair with label $y$. Similarly, a test set can be denoted as $Tar_{te} = \{(I_s^m, t_s^m), y_s^m\}_{m=1}^M$. The goal of their model is to transfer knowledge from $Src$ to boost the model performance on $Tar_{te}$ for cross-media retrieval.

Their model consists of a modal-sharing transfer subnetwork and a layer-sharing correlation subnetwork. In modal-sharing transfer subnetwork, they adopt the convolutional layers of AlexNet [32] to extract image features for source and target domains, and use word vectors to obtain text features. The image and text features pass through two fully connected layers, where single-modal and cross-modal knowledge transfer are performed.

Single-modal knowledge transfer aims to transfer knowledge from images in the source domain to images in the target domain. The main challenge is the domain discrepancy between the two image datasets. They propose to solve the domain discrepancy problem by minimizing the Maximum Mean Discrepancy (MMD) of image modality between the source and target domains. MMD is calculated in a layer-wise style in the fully connected layers. By minimizing MMD in reproduced kernel Hilbert space, the image representations from source and target domains are encouraged to have the same distribution, so knowledge from images in the source domain is expected to transfer to images in the target domain. Besides, the image encoder in the source domain is also fine-tuned by optimizing softmax loss on labeled image instances.

Cross-modal knowledge transfer aims to transfer knowledge between image and text in the target domain. Text and image representations from an annotated pair in the target domain are encouraged to be close to each other by minimizing their Euclidean distance. The cross-modal transfer loss of image and text representations is also computed in a layer-wise style in the fully connected layers. The domain discrepancy between image and text modalities is expected to be reduced in high-level layers.

In layer-sharing correlation subnetwork, representations from modal-sharing transfer subnetwork in the target domain are fed into shared fully connected layers to obtain the final common representation for both image and text. As the parameters are shared between two modalities, the last two fully connected layers are expected to capture the cross-modal correlation. Their model also utilizes label information in the target domain by minimizing softmax loss on labeled image/text pairs. After obtaining the final common representations, cross-media retrieval can be achieved by simply computing the nearest neighbors in semantic space.

#### 9.2.3.2 Deep Cross-Media Knowledge Transfer

As an extension of [23, 24] also focuses on dealing with domain discrepancy and insufficient cross-modal data for cross-media retrieval in specific areas, Huang and Peng [23] present a framework that transfers knowledge from a large-scale cross-media dataset (source domain) to boost the model performance on another small-scale cross-media dataset (target domain).

A crucial difference from [24] is that the dataset in the source domain also consists of image/text pairs with label annotations instead of a single-modal setting in [24]. Since both domains contain image and text media types, domain discrepancy comes from the media-level discrepancy in the same media type, and correlation-level discrepancy in image/text correlation patterns between different domains. They propose to transfer intra-media semantic and inter-media correlation knowledge by jointly reducing domain discrepancies on media-level and correlation-level.

To extract the distributed features for different media types, they adopt VGG19 [63] for image encoder and Word CNN [29] for text encoder. The two domains have the same architecture but do not share parameters. The extracted image/text features pass through two fully connected layers, respectively, where the media-level transfer is performed. Similar to [24], they reduce domain discrepancies within the same modalities by minimizing Maximum Mean Discrepancy (MMD) between the source and target domains. The MMD is computed in a layer-wise style to transfer knowledge within the same modalities. They also minimize Euclidean distance between image/text representations pairs in both source and target domains to preserve the semantic information across modalities.

Correlation-level transfer aims to reduce domain discrepancy in image/text correlation patterns in different domains. In two domains, both image and text representations share the last two fully connected layers to obtain the common representation for each domain. They optimize layer-wise MMD loss between the shared fully connected layers in different domains for correlation-level knowledge transfer, which encourages source and target domains to have the same image/text correlation patterns. Finally, both domains are trained with label information of image/text pairs. Note that the source domain and target domain do not necessarily share the same label set.

In addition, they propose a progressive transfer mechanism, which is a curriculum learning method aiming to promote the robustness of the model training. This is achieved by selecting easy samples for model training in the early period, and gradually increases the difficulty during the training. The difficulty of training samples is measured according to the bidirectional cross-media retrieval consistency.

## 9.3 Image Captioning

Image captioning is the task of automatically generating natural language descriptions for images. It is a fundamental task in artificial intelligence, which connects natural language processing and computer vision. Compared with other computer vision tasks, such as image classification and object detection, image captioning is significantly harder for two reasons: first, not only objects but also relationships between them have to be detected; second, besides basic judgments and classification, natural language sentences have to be generated.

Traditional methods for image captioning are usually using retrieval models or generation models, of which the ability to generalize is comparatively weaker compared with that of novel deep neural network models. In this section, we will introduce several typical models of both genres in the following.

### *9.3.1 Retrieval Models for Image Captioning*

The primary pipeline of retrieval models is (1) represent images and/or sentences using special features; (2) for new images and/or sentences, search for probable candidates according to the similarity of features.

Linking words to images has a rich history, and [50] (a retrieval model) is the first image annotation system. This paper tries to build a keyword assigning system for images from labeled data. The pipeline is as follows:

(1) Image segmentation. Every image is divided into several parts, using the simplest rectangular division. The reason for doing so is that an image is typically annotated with multiple labels, each of which often corresponds to only a part of it. Segmentation would help reduce noises in labeling.

(2) Feature extraction. Features of every part of the image are extracted.

(3) Clustering. Feature vectors of image segments are divided into several clusters. Each cluster accumulates word frequencies and thereby calculates word likelihood. Concretely,

$$P(w_i|c_j) = \frac{P(c_j|w_i)P(w_i)}{\sum_k P(c_j|w_k)P(w_k)} = \frac{n_{ji}}{N_j}, \tag{9.10}$$

where $n_{ji}$ is the number of times word $w_i$ appears in cluster $j$, and $N_j$ is the number of times that all words appear in cluster $j$. The calculation is based on using frequencies as probabilities.

(4) Inference. For a new image, the model divides it into segments, extracts features for every part, and finally, aggregates keywords assigned to every part to obtain the final prediction.

The key idea of this model is image segmentation. Take a landscape picture, for instance, there are two parts: `mountain` and `sky`, and both parts will be annotated with both labels. However, if another picture has two parts `mountain` and `river`, the two `mountain` parts would hopefully be in the same cluster and discover that they share the same label `mountain`. In this way, labels can be assigned to the correct part of the image, and noises could be alleviated.

Another typical retrieval model is proposed by [17], which can assign a linking score between an image and a sentence. An intermediate space of meaning calculates this score of linking. The representation of the meaning space is a triple in the form of $\langle$*object*, *action*, *scene* $\rangle$. Each slot of the triple has a finite discrete candidate set. The problem of mapping images and sentences into the meaning space involves solving a Markov random field.

Different from the previous model, this system can do not only image caption, but also do the inverse, that is, given a sentence, the model provides certain probable associated images. At the inference stage, the image (sentence) is first mapped to the intermediate meaning space, then we search in the pool for the sentence (image) that has the best matching score.

After that, researchers also proposed a lot of retrieval models which consider different kinds of characteristics of the images, such as [21, 28, 34].

### 9.3.2 Generation Models for Image Captioning

Different from the retrieval-based model, the basic pipeline of generation models is (1) use computer vision techniques to extract image features, (2) generate sentences from these features using methods such as language models or sentence templates.

Kulkarni et al. [33] propose a system that makes a tight connection between the particular image and the sentence generating process. The model uses visual detectors to detect specific objects, as well as attributes of a single object and relationships between multiple objects. Then it constructs a conditional random field to incorporate unary image potentials and higher order text potentials and thereby predicts labels for the image. Labels predicted by conditional random fields (CRF) is arranged as a triple, e.g., ⟨⟨`white`, `cloud`⟩, `in`, ⟨`blue`, `sky`⟩⟩.

Then sentences are generated according to the labels. There are two ways to build a sentence based on the triple skeleton. (1) The first is to use an $n$-gram language model. For example, when trying to decide whether or not to put a glue word $x$ between a pair of meaningful words (which means they are inside the triple) $a$ and $b$, the probabilities $\hat{p}(axb)$ and $\hat{p}(ab)$ are compared for the decision. $\hat{p}$ is the standard length-normalized probability of the $n$-gram language model. (2) The second is to use a set of descriptive language templates, which alleviates the problem of grammar mistakes in the language model.

Further, [16] proposes a novel framework to explicitly represent the relationship between image structure and its caption sentence's structure. The method, Visual Dependency Representation, detects objects in the image, and detects the relationship between these objects based on the proposed Visual Dependency Grammar, which includes eight typical relations like `beside` or `above`. Then the image can be arranged as a dependency graph, where nodes are objects and edges are relations. This image dependency graph can be aligned with the syntactic dependency representation of the caption sentence. The paper further provides four templates to generating descriptive sentences from the extracted dependency representation.

Besides these two typical works, there are massive generation models for image captioning, such as [15, 35, 78].

### *9.3.3 Neural Models for Image Captioning*

In [33], it was claimed in 2011 that in image captioning tasks: *Natural language generation still remains an open research problem. Most previous work is based on retrieval and summarization*. From 2015, inspired by advances in neural language model and neural machine translation, a number of end-to-end neural image captioning models based on the encoder-decoder system have been proposed. These new models significantly improve the ability to generate natural language descriptions.

#### 9.3.3.1 The Basic Model

Traditional machine translation models typically stitch many subtasks together, such as individual word translation and reordering, to perform sentence and paragraph translation. Recent neural machine translation models, such as [8], use a single encoder-decoder model, which can be optimized by stochastic gradient descent conveniently. The task of image captioning is inherently analogous to machine translation because it can also be regarded as a translation task, where the source "language" is an image. The encoders and decoders used for machine translations are typically RNNs, which is a natural selection for sequences of words. For image captioning, CNN is chosen to be the encoder, and RNN is still used as the decoder.

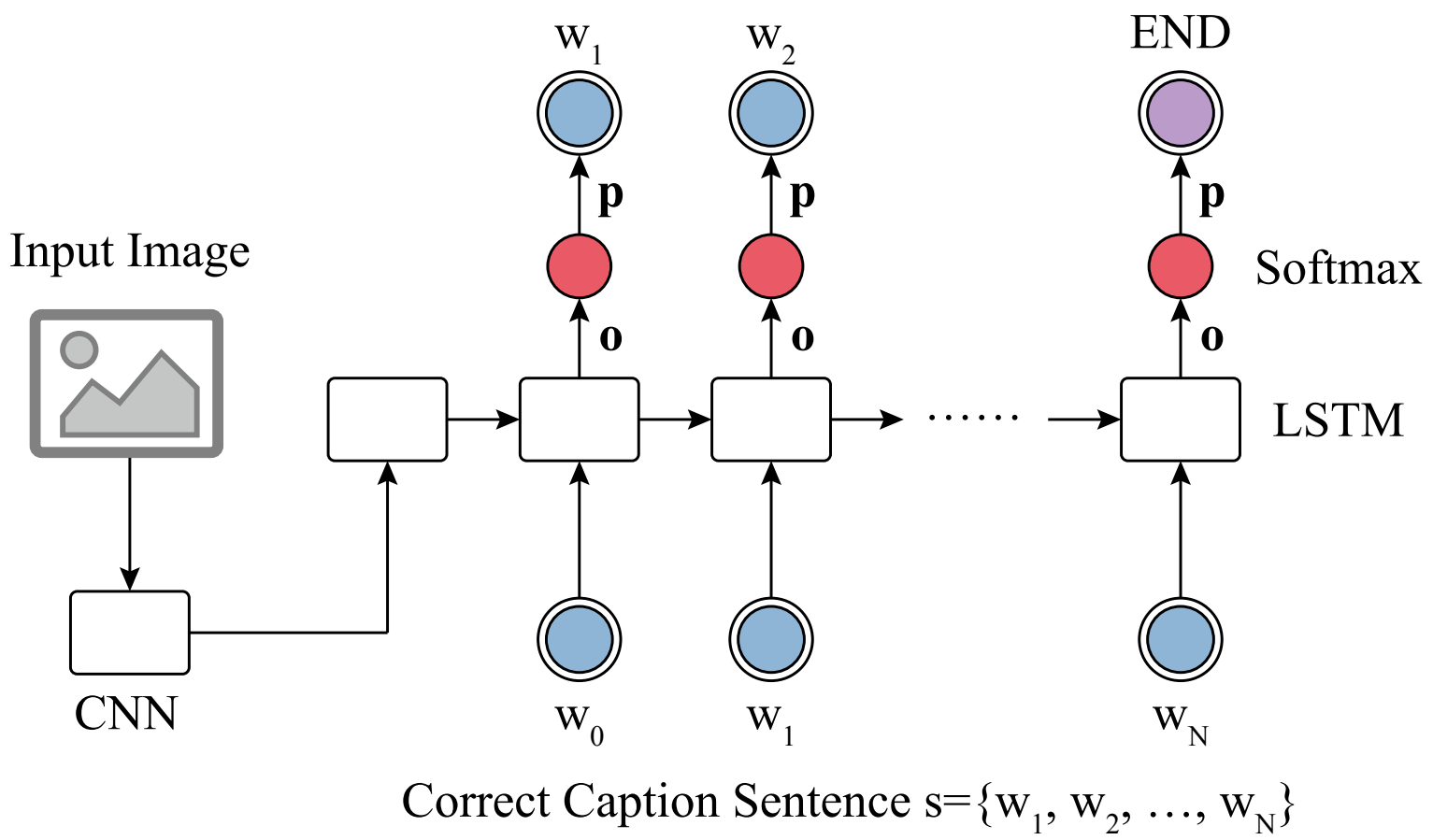

**Fig. 9.2** The architecture of encoder-decoder framework for image captioning

Vinyals et al. [70] is the most typical model which uses encoder-decoder for image captioning (see Fig. 9.2). Concretely, a CNN model is used to encode the image into a fix length vector, which is believed to contain the necessary information for captioning. With this vector, an RNN language model is used to generate natural language descriptions, and this is the decoder. Here, the decoder is similar to the LSTM used for machine translation. The first unit takes the image vector as the input vector, and the rest units take the previous word embedding as input. Each unit outputs a vector **o** and passes a vector to the next unit. **o** is further fed into a softmax layer, whose output **p** is the probability of each word within the vocabulary. The ways to deal with these calculated probabilities are different in training and testing:

**Training.** These probabilities **p** are used to calculate the likelihood of the provided description sentences. Considering the nature of RNNs, it is easy to model the joint probability into conditional probabilities.

$$\log P(s|I) = \sum_{t=0}^{N} \log P(w_t|I, w_0, \ldots, w_{t-1}), \tag{9.11}$$

where $s = \{w_1, w_2, ..., w_N\}$ is the sentence and its words, $w_0$ is a special START token, and $I$ is the image. Stochastic gradient descent can thereby be performed to optimize the model.

**Testing.** There are multiple approaches to generate sentences given an image. The first one is called Sampling. For each step, the single word with the highest probability in **p** is chosen, and used as the input of the next unit until the END token is generated or a maximal length is reached. The second one is called Beam Search. For each step (now the length of sentences is $t$), $k$ best sentences are kept. Each of them generates several new sentences of length $t + 1$, and again, only $k$ new

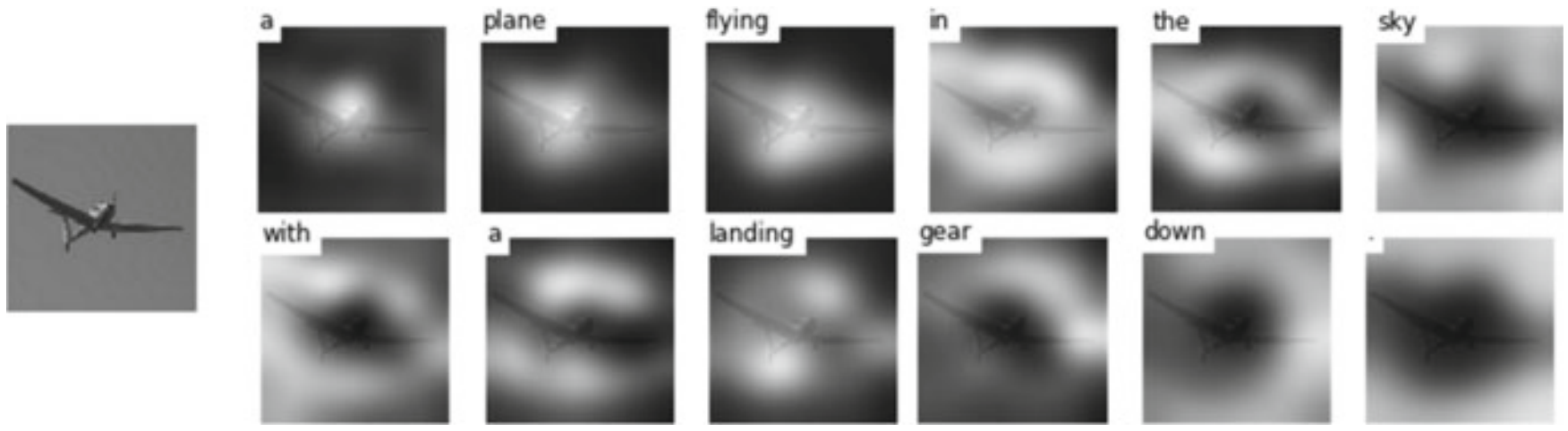

**Fig. 9.3** An example of image captioning with attention mechanism

sentences are kept. Beam Search provides a better approximation for

$$s^* = \arg\max_s \log P(s|I). \tag{9.12}$$

#### 9.3.3.2 Variants of the Basic Model

The research on image captioning tightly follows that on machine translation. Inspired by [6], which uses attention mechanism in machine translation, [76] introduces visual attention into the encoder-decoder image captioning model.

The major bottleneck of [70] is the fact that information from the image is shown to the LSTM decoder *only* at the first decoding unit, which actually requires the encoder to squeeze all useful information into one fixed-length vector. In contrast, [76] does not require such compression. The CNN encoder does not produce one vector for the entire image; instead, it produces $L$ region vectors $\mathbf{I}_i$, each of which is the representation of a part of the image. At every step of decoding, the inputs include standard LSTM inputs (i.e., output and hidden state of last step $\mathbf{o}_{t-1}$ and $\mathbf{h}_{t-1}$), and an input vector $\mathbf{z}$ from the encoder. Here, $\mathbf{z}$ is the weighted sum of image vectors $\mathbf{I}_i$: $\mathbf{z} = \sum_i \alpha_i \mathbf{I}_i$, where $\alpha_i$ is the weight computed from $\mathbf{I}_i$ and $\mathbf{h}_{t-1}$. Throughout the training process, the model learns to focus on parts of the image for generating the next word by producing larger weights $\alpha$ on more relevant parts, as shown in Fig. 9.3.[1]

While the above paper uses soft attention for the image, [27] makes explicit alignment between image fragments and sentence fragments before generating a description for the image. In the first stage, the alignment stage, sentence and image fragments are aligned by being mapped to a shared space. Concretely, sentence fragments (i.e., $n$ consecutive words) are encoded using a bidirectional LSTM into the embeddings $\mathbf{s}$, and image fragments (i.e., part of the image, and also the entire image) are encoded using a CNN into the embeddings $\mathbf{I}$. The similarity score between image $I$ and sentence $s$ is computed as

[1]The example is obtained from the implementation of Yunjey Choi (https://github.com/yunjey/show-attend-and-tell).

$$\text{sim}(I, s) = \sum_{t \in g_s} \max_{i \in g_I}(0, \mathbf{I}_i^\top \mathbf{s}_t), \tag{9.13}$$

where $g_s$ is the sentence fragment set of sentence $s$, and $g_I$ is the image fragment set of image $I$. The alignment is then optimized by minimizing the ranking loss $\mathscr{L}$ for both sentences and images:

$$\mathscr{L} = \sum_I \left[ \sum_s \max(0, \text{sim}(I, s) - \text{sim}(I, I) + 1) + \sum_s \max(0, \text{sim}(s, I) - \text{sim}(I, I) + 1) \right]. \tag{9.14}$$

The assumption for this alignment procedure is similar to [50] (see Sect. 9.3.1): all description sentences are regarded as (possibly noisy) labels for every image section and are based on the massive training data, the model would hopefully be trained to align caption sentences to their corresponding image fragments. The second stage is similar to the basic model in [70], but the alignment results are used to provide more precise training data.

As mentioned above, [76] makes the decoder have the ability to focus attention on the different parts of the image for different words. However, there are some nonvisual words in the decoding process. For example, words such as `the` and `of` are more dependent on semantic information than visual information. Furthermore, words such as `phone` followed by `cell` or `meter` before `near the parking` are usually generated by the language model. To avoid the gradient of a nonvisual word decreasing the effectiveness of visual attention, in the process of generating captions, [43] adopts an adaptive attention model with a visual sentinel. At each time step, the model needs to determine that it depends on an image region or a visual sentinel.

Adaptive attention model [43] uses attention in the process of generating a word rather than updating the LSTM state; it utilizes "visual sentinel" vector $\mathbf{x}_t$ and image region vectors $\mathbf{I}_i$. Here, $\mathbf{x}_t$ is produced by the inputs and states of LSTM at time step t, while $\mathbf{I}_i$ is provided from CNN encoder. Then the adaptive context vector $\hat{\mathbf{c}}_t$ is the weighted sum of $L$ image region vectors $\mathbf{I}_i$ and visual sentinel $\mathbf{x}_t$:

$$\hat{\mathbf{c}}_t = \sum_i^L \alpha_i \mathbf{I}_i + \alpha_{L+1} \mathbf{x}_t, \tag{9.15}$$

where $\alpha_i$ are the weights computed by $\mathbf{I}_i$, $\mathbf{x}_t$, and the LSTM hidden state $\mathbf{h}_t$. We have $\sum_{i=1}^{L+1} \alpha_i = 1$. Finally the probability of a word in vocabulary at time t can be calculated as a residual form:

$$\mathbf{p}_t = \text{Softmax}(\mathbf{W}_p(\hat{\mathbf{c}}_t + \mathbf{h}_t)), \tag{9.16}$$

where $\mathbf{W}_p$ is a learned weight parameter.

Many existing image captioning models with attention allocate attention over image's regions, whose size is often $7 \times 7$ or $14 \times 14$ decided by the last pooling layer in CNN encoder. Anderson et al. [2] first calculate attention at the level of objects. It first employs Faster R-CNN [58] which is trained on ImageNet [60] and Genome [31] to predict attribute class, such as an open oven, green bottle, floral dress, and so on. After that, it applies attention over valid bounding boxes to get fine-grained attention for helping the caption generation.

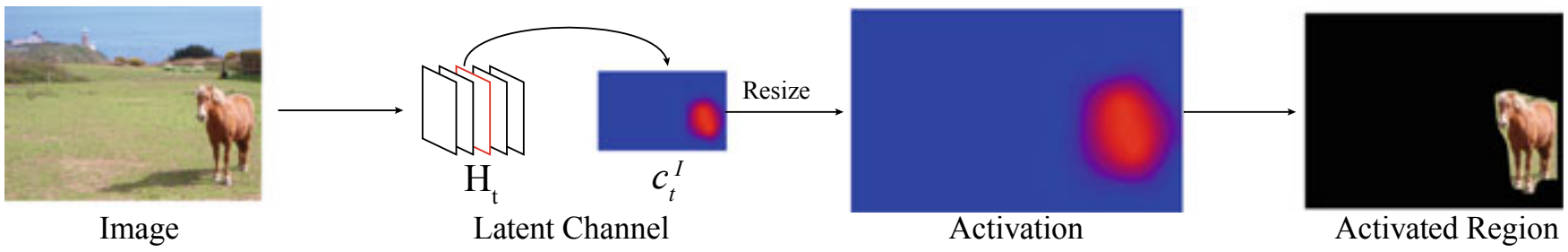

**Fig. 9.4** An example of the activated region of a latent channel

Besides, [11] rethinks the form of latent states in image captioning, which usually compresses two-dimensional visual feature maps encoded by CNN to a one-dimensional vector as the input of the language model. They find that the language model with 2D states can preserve the spatial locality, which can link the input visual domain and output linguistic domain observed by visualizing the transformation of hidden states.

Word embeddings and hidden states in [11] are 3D tensors of size $C \times H \times W$, which means $C$ channels, each of size $H \times W$. The encoded features maps will be directly inputted to the 2D language model instead of going through an average pooling layer. In the 2D language model, the convolution operator takes the place of matrix multiplication in the 1D model, and mean pooling will be used to generate the output word probability distribution from 2D hidden states. Figure 9.4 shows activated region of a latent channel at the $t$th step. When we set a threshold for the activated regions, it is revealed that the special channels are associated with specific nouns in the decoding process, which help get a better understanding of the process of generating captions.

Traditional methods train the caption model by maximizing the likelihood of training examples, which forms a gap between the optimization objective and evaluating metrics. To alleviate the problem, [59] uses reinforcement learning to directly maximize the CIDEr metric [69]. CIDEr reflects the diversity of generated captions by giving high weights to the low-frequency $n$-grams in the training set, which demonstrates that people prefer detailed captions rather than universal ones, like `a boy is playing a game`. To encourage the distinctiveness of captions, [10] adopts contrastive learning. Their model learns to discriminate the caption of a given image and the caption of an alike image by maximizing the difference between ground truth positive pair and mismatch negative pair. The experiment shows that contrastive learning increases the diversity of captions significantly.

Furthermore, automatic evaluation metrics, such as BLEU [54], METEOR [13], ROUGE [38], CIDEr [69], SPICE [1], and so on, may neglect some novel expressions restrained by the ground truth captions. To better evaluate the naturalness and diversity of captions, [9] proposes a framework based on Conditional Generative Adversarial Networks, whose generator tries to achieve a higher score in the evalua-tor, while the evaluator tries to distinguish between the generated caption and human descriptions for a given image, as well as between the given image and the mismatch description. The user study shows that the trained generator can generate natural and diverse captions than the model trained by maximum likelihood estimate, while the trained evaluator is more consistent with human's evaluation.

Besides the works we introduced above, there are also a mass of variants of the basic encoder-decoder model such as [20, 26, 40, 45, 51, 71, 73].

## 9.4 Visual Relationship Detection

Visual relationship detection is the task of detecting objects in an image and understanding the relationship between them. While detecting the objects is always based on semantic segmentation or object detection methods, such as R-CNN, understanding the relationship is the key challenge of this task. While detecting visual relation with image information is intuitive and effective [25, 62, 84], leveraging information from language can further boost the model performance [37, 41, 82].

### *9.4.1 Visual Relationship Detection with Language Priors*

Lu et al. [41] propose a model that uses language priors to enhance the performance on infrequent relationships for which sufficient training instances are hard to obtain solely from images. The overall architecture is shown in Fig. 9.5.

They first train a CNN to calculate the unnormalized relations' probability obtained from visual inputs by

$$P_V(R_{\langle i,j,k\rangle}, \Theta | \langle O_1, O_2\rangle) = P_i(O_1)(\mathbf{z}_k^\top \text{CNN}(O_1, O_2) + s_k)P_j(O_2), \quad (9.17)$$

where $P_i(O_j)$ denotes the probability that bounding box $O_j$ is entity $i$, and $CNN(O_1, O_2)$ is the joint feature of box $O_1$ with box $O_2$. $\Theta = \{\mathbf{z}_k, s_k\}$ is the set of parameters.

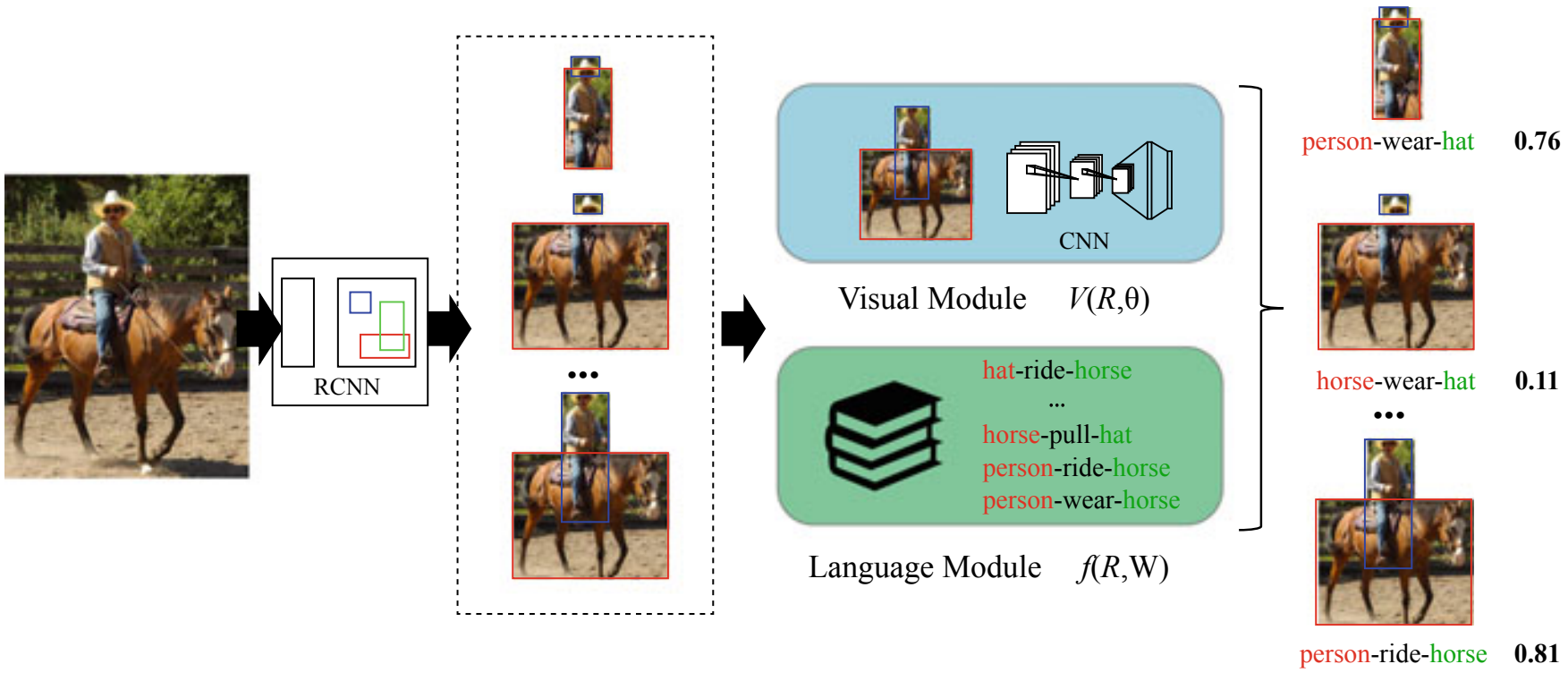

**Fig. 9.5** The architecture of visual relationship detection with language prior

Besides, language prior is considered in this model by calculating the unnormalized probability that the entity pair $\langle i, j\rangle$ has the relation $k$:

$$P_f(R, \mathbf{W}) = \mathbf{r}_k^\top [\mathbf{w}_i; \mathbf{w}_j] + b_k, \quad (9.18)$$

where $\mathbf{w}_i$ and $\mathbf{w}_j$ are the word embeddings of the text of subject and object, respectively, $\mathbf{r}_k$ is the learned relational embedding of the relation $k$.

Given the probabilities of a relation from visual and textual inputs, respectively, the authors combine them into the integrated probability of a relation. The final prediction is the one with maximal integrated probability:

$$R^* = \max_R P_V(R_{\langle i,j,k\rangle} | \langle O_1, O_2\rangle) P_f(R, \mathbf{W}). \quad (9.19)$$

The rank of the ground truth relationship $R$ with bounding boxes $O_1$ and $O_2$ is maximized using the following rank loss function:

$$\begin{aligned} C(\Theta, \mathbf{W}) = &\sum_{\langle O_1, O_2\rangle, R} \max\{1 - P_V(R, \Theta | \langle O_1, O_2\rangle) P_f(R, \mathbf{W}) \\ &+ \max_{\langle O_1', O_2'\rangle \neq \langle O_1, O_2\rangle, R' \neq R} P_V(R', \Theta | \langle O_1', O_2'\rangle) P_f(R', \mathbf{W}), 0\}. \end{aligned} \quad (9.20)$$

In addition to the loss that optimizes the rank of the ground truth relationships, the authors also propose two regularization functions for language priors. The final loss function of this model is defined as

$$\mathscr{L} = C(\Theta, \mathbf{W}) + \lambda_1 L(\mathbf{W}) + \lambda_2 K(\mathbf{W}). \quad (9.21)$$

$K(\mathbf{W})$ is a variance function to make the similar relationships' corresponding $f(\cdot)$ function closer:

$$K(\mathbf{W}) = Var\{\frac{[P_f(R, \mathbf{W}) - P_f(R', \mathbf{W})]^2}{d(R, R')}\}, \forall R, R', \quad (9.22)$$

where $d(R, R')$ is the sum of the cosine distances (in Word2vec space) between the two objects and the predicates of the two relationships $R$ and $R'$.

$L(\mathbf{W})$ is a function to encourage less-frequent relation to have a lower $f()$ score. When $R$ occurs more frequently than $R'$, we have

$$L(\mathbf{W}) = \sum_{R, R'} \max\{P_f(R', \mathbf{W}) - P_f(R, \mathbf{W}) + 1, 0\}. \quad (9.23)$$

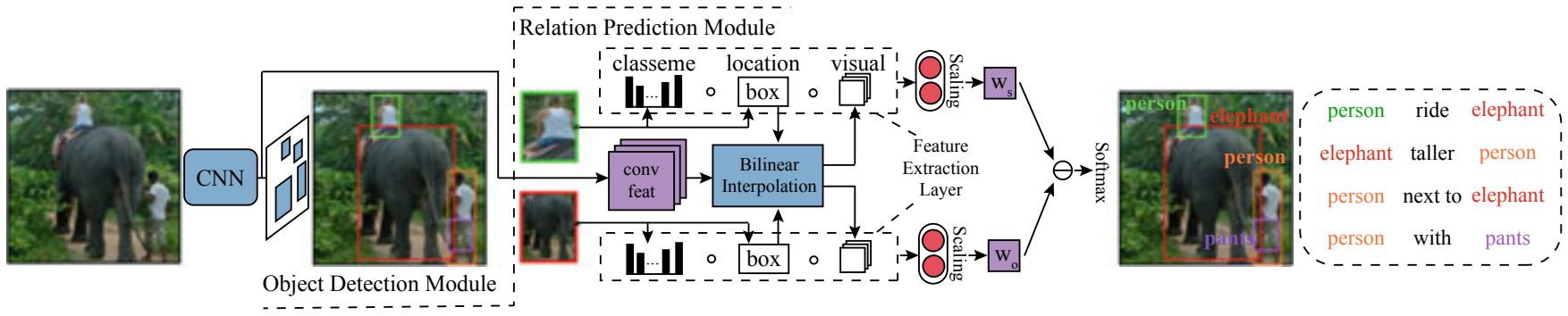

**Fig. 9.6** The architecture of VTransE model

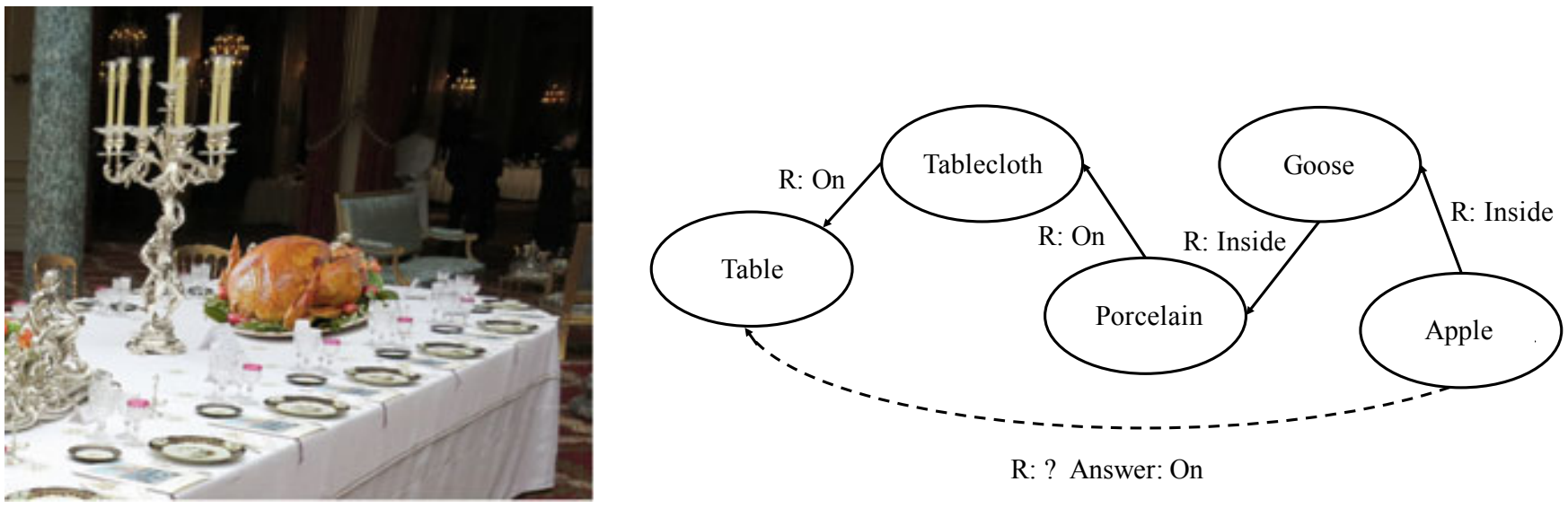

(a) A scene. (b) The corresponding scene graph.

**Fig. 9.7** An illustration for scene graph generation

## 9.4.2 Visual Translation Embedding Network

Inspired by recent progress in knowledge representation learning, [82] proposes VTransE, a visual translation embedding network. Objects and the relationship between objects are modeled as TransE [7] like vector translation. VTransE first projects *subject* and *object* into the same space as relation translation vector $\mathbf{r} \in \mathbb{R}^r$. *Subject* and *object* could be denoted as $\mathbf{x}_s, \mathbf{x}_o \in \mathbb{R}^M$ in the feature space, where $M \gg r$. Similar to TransE relationship, VTransE establishes a relationship as

$$\mathbf{W}_s\mathbf{x}_s + \mathbf{r} \sim \mathbf{W}_o\mathbf{x}_o, \tag{9.24}$$

where $\mathbf{W}_s$ and $\mathbf{W}_o$ are projection matrices. The overall architecture is shown in Fig. 9.6.

## 9.4.3 Scene Graph Generation

Li et al. [37] further formulate visual relation detection as a scene graph generation task, where nodes correspond to objects and directed edges correspond to visual relations between objects, as shown in Fig. 9.7.

This formulation allows [37] to leverage different levels of context information, such as information from objects, phrases (i.e., ⟨*subject*, *predicate*, *object*⟩ triples),

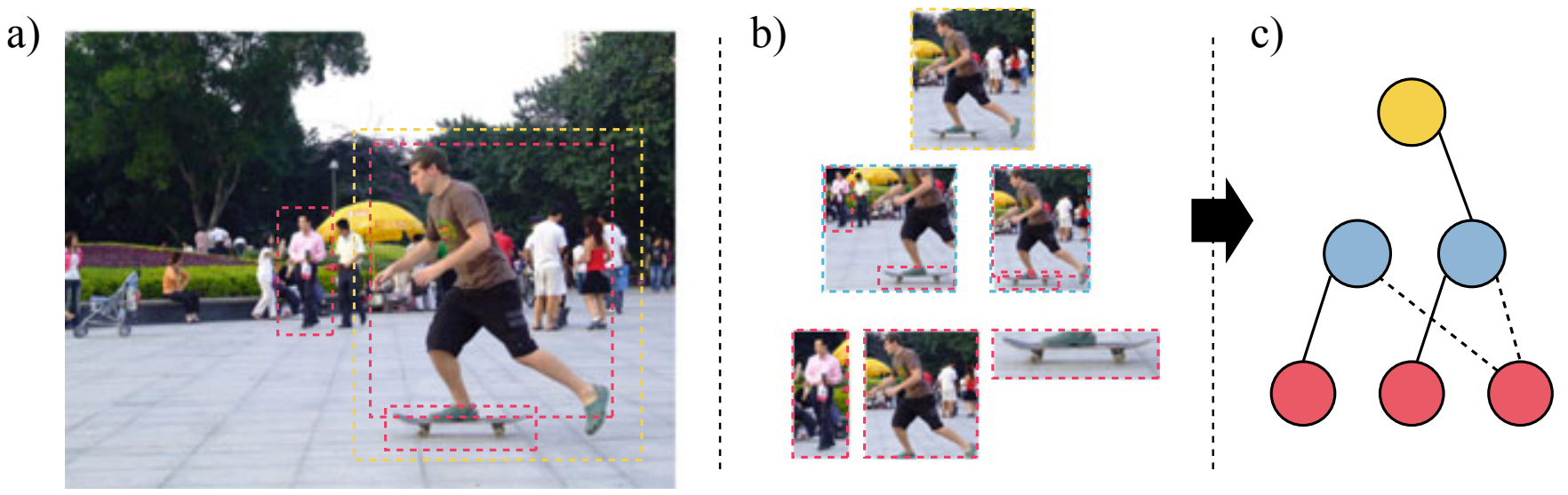

**Fig. 9.8** Dynamical graph construction. **a** The input image. **b** Object (bottom), phrase (middle), and caption region (top) proposals. **c** The graph modeling connections between proposals. Some of the phrase boxes are omitted

and region captions, to boost the performance of visual relation detection. Specifically, [37] proposes to construct a graph that aligns these three levels of information and perform feature refinement via message passing, as shown in Fig. 9.8. By leveraging complementary information from different levels, the performances of different tasks are expected to be mutually improved.

**Dynamic Graph Construction.** Given an image, they first generate three kinds of proposals that correspond to three kinds of nodes in the proposed graph structure. The proposals include object proposals, phrase proposals, and region proposals. The object and region proposals are generated using Region Proposal Network (RPN) [57] trained with ground truth bounding boxes. Given $N$ object proposals, phrase proposals are constructed based on $N(N-1)$ object pairs that fully connect the object proposals with direct edges, where each direct edge represents a potential phrase between an object pair.

Each phrase proposal is connected to the corresponding subject and object with two directed edges. A phrase proposal and a region proposal are connected if their overlap exceeds a certain fraction (e.g., 0.7) of the phrase proposal. There are no direct connections between objects and regions since they can be indirectly connected via phrases.

**Feature Refinement.** After obtaining the graph structure of different levels of nodes, they perform feature refinement by iterative message passing. The message passing procedure is divided into three parallel stages, including object refinement, phrase refinement, and region refinement.

In object feature refinement, the object proposal feature is updated with gated features from adjacent phrases. Given an object $i$, the aggregated feature from regions that are linked to object $i$ via subject-predicate edges $\hat{\mathbf{x}}_i^{p\to s}$ can be defined as follows:

$$\hat{\mathbf{x}}_i^{p\to s} = \frac{1}{\|E_{i,p}\|} \sum_{(i,j)\in E_{s,p}} f_{\langle o,p\rangle}(\mathbf{x}_i^{(o)}, \mathbf{x}_j^{(p)})\mathbf{x}_j^{(p)}, \tag{9.25}$$

where $E_{s,p}$ is the set of subject predicate connections, and $E_{i,p}$ denotes the number of phrases connected with the object $i$ as the subject predicate pairs. $f_{\langle o,p \rangle}$ is a learnable gate function that controls the weights of information from different sources:

$$f_{\langle o,p\rangle}(\mathbf{x}_i^{(o)}, \mathbf{x}_j^{(p)}) = \sum_{k=1}^{K} \text{Sigmoid}(\omega_{\langle o,p\rangle}^{(k)} \cdot [\mathbf{x}_i^{(o)}; \mathbf{x}_j^{(p)}]), \tag{9.26}$$

where $\omega_{\langle o,p\rangle}^{(k)}$ is a gate template used to calculate the importance of the information from a subject-predicate edge and $K$ is the number of templates. The aggregated feature from object-predicate edges $\hat{\mathbf{x}}_i^{p\rightarrow o}$ can be similarly computed.

After obtaining information $\hat{\mathbf{x}}_i^{p\rightarrow s}$ and $\hat{\mathbf{x}}_i^{p\rightarrow o}$ from adjacent phrases, the object refinement at time step $t$ can be defined as follows:

$$\mathbf{x}_{i,t+1}^{(o)} = \mathbf{x}_{i,t}^{(o)} + f^{(p\rightarrow s)}(\hat{\mathbf{x}}_i^{p\rightarrow s}) + f^{(p\rightarrow o)}(\hat{\mathbf{x}}_i^{p\rightarrow o}), \tag{9.27}$$

where $f(\cdot) = \mathbf{W}\text{ReLU}(\cdot)$, $\mathbf{W}$ is a learnable parameter and not shared between $f^{(p\rightarrow s)}(\cdot)$ and $f^{(p\rightarrow o)}(\cdot)$.

The refinement scheme of phrases and regions is similar to objects. The only difference is the information sources: Phrase proposals receive information from adjacent objects and regions, and region proposals receive information from phrases.

After feature refinement via iterative message passing, the feature of different levels of nodes can be used for corresponding tasks. Region features can be used as the initial state of a language model to generate region captions. Phrase features can be used to predict visual relation between objects, which composes of the scene graph of the image.

In comparison with scene graph generation methods that model the dependencies between relation instances by attention mechanism or message passing, [47] decomposes the scene graph task into a mixture of two phases: extracting primary relations from input, and completing the scene graph with reasoning. The authors propose a Hybrid Scene Graph generator (HRE) that combines these two phases in a unified framework and generates scene graphs from scratch.

Specifically, HRE first encodes the object pair into representations and then employs a neural relation extractor resolving primary relations from inputs and a differentiable inductive logic programming model that iteratively completes the scene graph. As shown in Fig. 9.9, HRE contains two units, a pair selector and a relation predictor, and runs in an iterative way.

At each time step, the pair selector takes a look at all object pairs $P^-$ that have not been associated with a relation and chooses the next pair of entities whose relation is to be determined. The relation predictor utilizes the information contained in all pairs $P^+$ whose relations have been determined, and the contextual information of the pair to make the prediction on the relation. The prediction result is then added to $P^+$ and benefits future predictions.

To encode object pair into representations, HRE extends the union box encoder proposed by [41] by adding the object features (what are the objects) and their locations (where are the objects) into the object pair representation, as shown in Fig. 9.10.

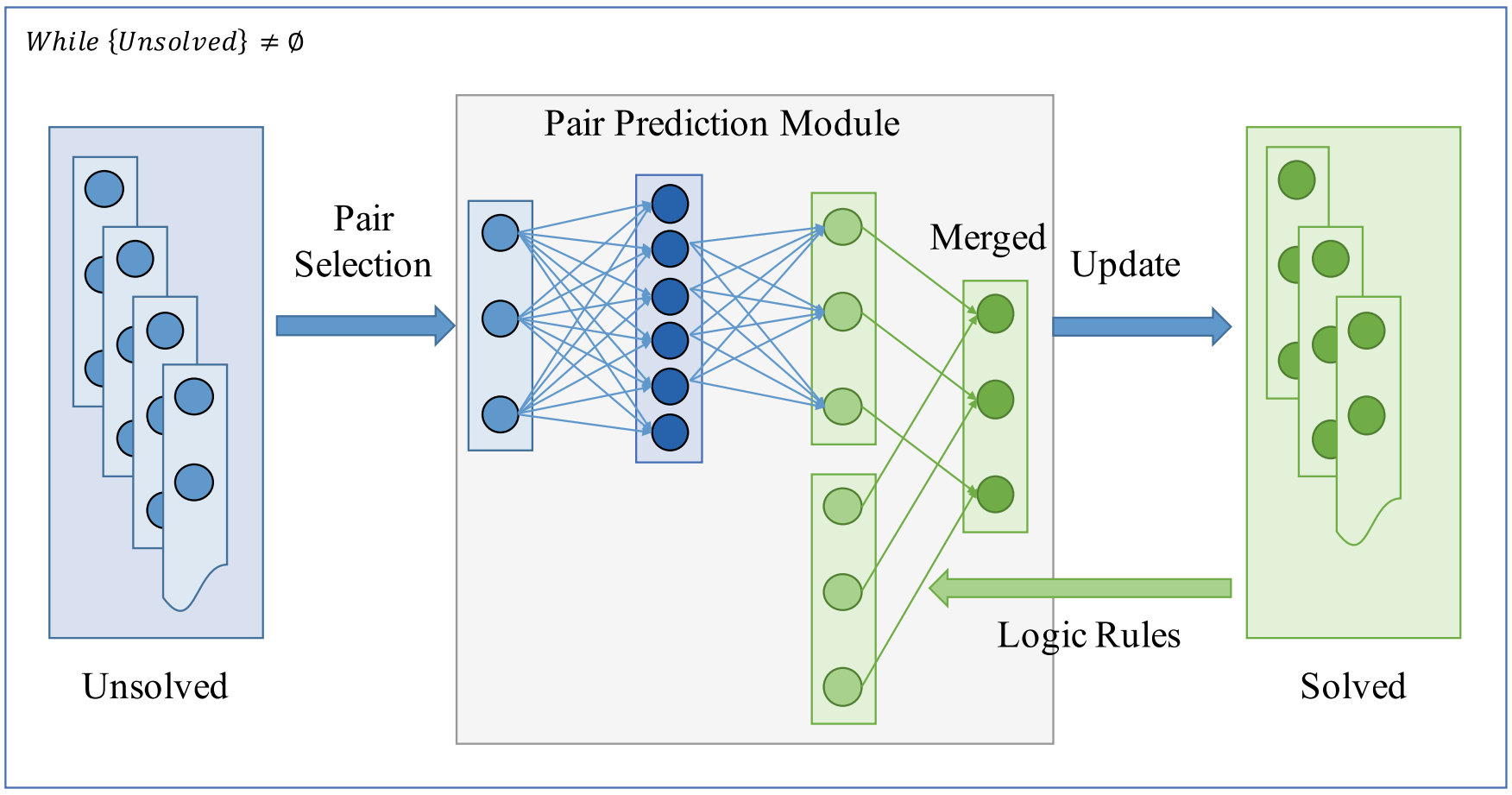

**Fig. 9.9** Framework of HRE that detects primary relations from inputs and iteratively completes the scene graph via inductive logic programming

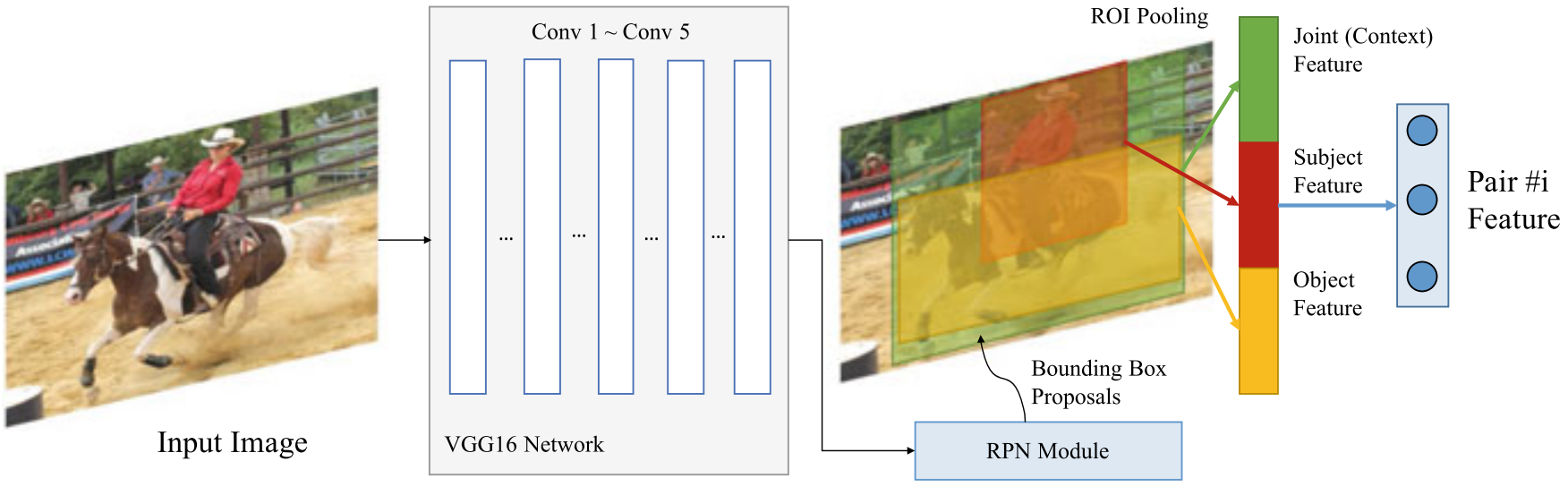

**Fig. 9.10** Object pair encoder of HRE

**Relation Predictor.** The relation predictor is composed of two modules: a neural module predicting the relations between entities based on the given context (i.e., a visual image) and a differentiable inductive logic module performing reasoning on $P^+$. Both modules predict the relation score between a pair of objects individually. The relation scores from the two modules are finally integrated by multiplication.

**Pair Selector.** The selector works as the predictor's collaborator with the goal to figure out the next relation which should be determined. Ideally, the choice $p^*$ made by the selector should satisfy the condition that all relations that will affect the predictor's prediction on $p^*$ should be sent to the predictor ahead of $p^*$. HRE implements the pair selector as a greedy selector which always chooses the entity pair from $P^-$ to be added to $P^+$ as the entity pair of which the relation predictor is most confident in its prediction.

It is worth noting that the task of scene graph generation resembles document-level relation extraction in many aspects. Both tasks seek to extract structured graphs consisting of entities and relations. Also, they need to model the complex dependencies between entities and relations in a rich context. We believe both tasks are worthy to explore for future research.

## 9.5 Visual Question Answering

Visual Question Answering (VQA) aims to answer natural language questions about an image, and can be seen as a single turn of dialogue about a picture. In this section, we will introduce widely used VQA datasets and several typical VQA models.

### 9.5.1 VQA and VQA Datasets

VQA was first proposed in [46]. They first propose a single-world approach by modeling the probability of an answer $a$ given question $q$ and a world $w$ by

$$P(a|q,w) = \sum_{z} P(a|z,w)P(z|q), \tag{9.28}$$

where $z$ is a latent variable associated with the question and the world $w$ is a representation of the image. They further extend the single-world approach to a multi-world approach by marginalizing over different segments $s$ of the given image. The probability of an answer $a$ given question $q$ and a world $w$ is given by

$$P(a|q,s) = \sum_{w}\sum_{z} P(a|w,z)P(w|s)P(z|q). \tag{9.29}$$

They also release the first dataset of VQA named as DAQUAR in their paper.

Besides DAQUAR, researchers also release a lot of VQA datasets with various characteristics. The most widely used dataset was released in [4], where the authors provided cases and experimental evidence to demonstrate that to answer these questions, a human or an algorithm should use features of the image and external knowledge. Figure 9.11 shows examples of VQA dataset released in [4]. It is also demonstrated that this problem cannot be solved by converting images to captions and answering questions according to captions. Experiment results show that the performance of vanilla methods is still far from human.

In fact, there are also other existing datasets for Visual QA such as Visual7W [85], Visual Madlibs [80], COCO-QA [56], and FM-IQA [19].

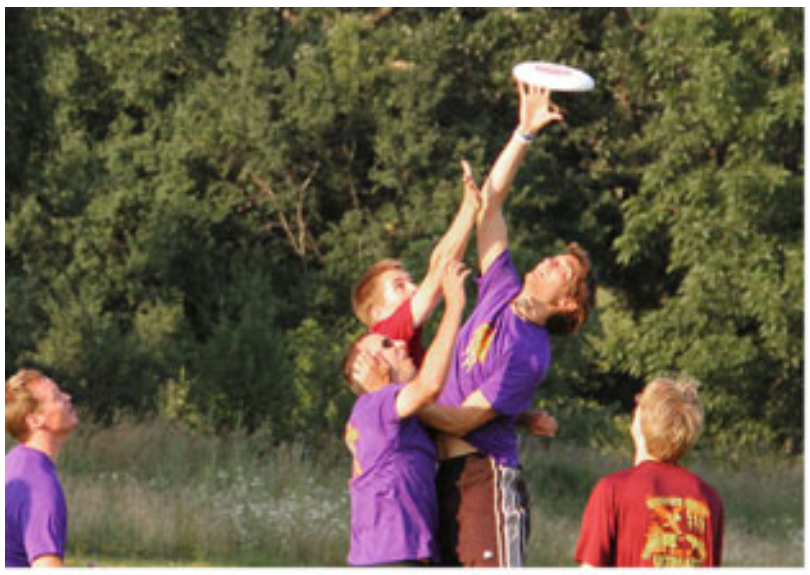

Question: Why are the men jumping?
Answer: to catch frisbee

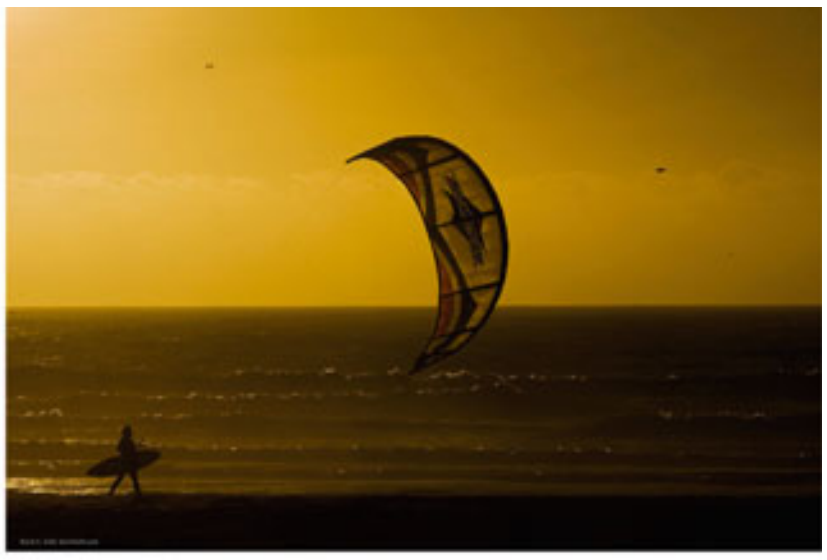

Question: Is the water still?
Answer: no

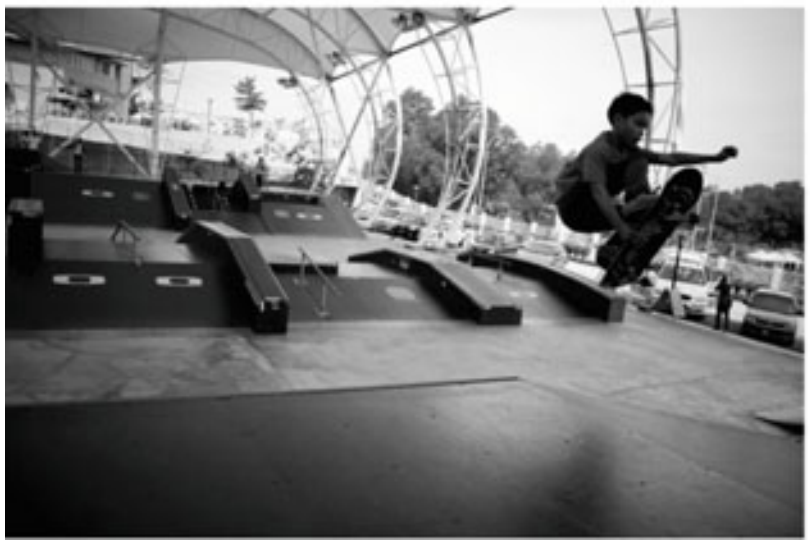

Question: What is the kid doing?
Answer: skateboarding

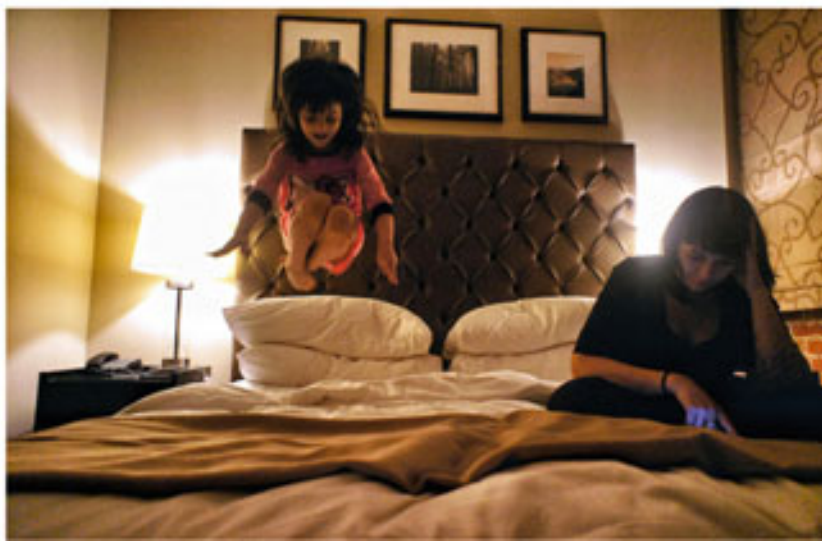

Question: What is hanging on the wall above the headboard?
Answer: pictures

**Fig. 9.11** Examples of VQA dataset

### *9.5.2 VQA Models*

Besides, [4, 46] further investigate approaches to solve specific types of questions in VQA. Moreover, [83] proposes an approach to solve "YES/NO" questions. Note that the model is an ensemble model of two similar models: Q-model and Tuple-model, the difference between which will be described later. The overall approach can be divided into two steps: (1) Language Parsing and (2) Visual Verification. In the former step, they extract ⟨P, R, S⟩ tuples from questions first by parsing it and assigning an entity to each word. Then they summarize the parsed sentences through removing "stop words", auxiliary verbs, and all words before a nominal subject or passive nominal subject, and further split the summary into PRS arguments according to the part of speech of phrases. The difference between Q-Model and Tuple-model is that the Q-model is the one used in their previous work [4], embedding the question into a dense 256-dim vector by LSTM, while Tuple-model is to convert ⟨P, R, S⟩ tuples into 256-dim embeddings by MLP. As for the Visual Verification step, they use the same feature of images as in [39] which was encoded into the dense 256-dim vector by an inner-product layer followed by a *tanh* layer. These two vectors are passed through an MLP to produce the final output ("Yes" or "No").

Moreover, [61] proposes a method to calculate attention $\alpha_j$ by the set of image features $\mathbf{I} = (\mathbf{I}_1, \mathbf{I}_2, \ldots, \mathbf{I}_K)$ and the question embedding $\mathbf{q}$ by

$$\alpha_j = (\mathbf{W}_1\mathbf{I}_j + \mathbf{b}_1)^\top(\mathbf{W}_2\mathbf{q} + \mathbf{b}_2), \tag{9.30}$$

where $\mathbf{W}_1$,$\mathbf{W}_2$,$\mathbf{b}_1$,$\mathbf{b}_2$ are trainable parameters.

Attention based techniques are quite efficient for filtering noises that are irrelevant to the question. However, some questions are only related to some small regions, which encourages researchers to use stacked attention to further filtering noises. We refer readers to Fig. 1b in [79] for an example of stacked attention.

Yang et al. [79] further extend the attention-based model used in [61], which employs LSTMs to predict the answer. They take the question as input and attend to different regions in the image to obtain additional input. The key idea is to gradually filter out noises and pinpoint the regions that are highly relevant to the answer by reasoning through multiple stacked attention layers progressively. The stacked attention could be calculated by stacking:

$$\mathbf{h}_A^k = \tanh(\mathbf{W}_1^k\mathbf{I} \oplus (\mathbf{W}_2^k\mathbf{u}^{k-1} + \mathbf{b}_A^k)). \tag{9.31}$$

Note that we denote the addition of a matrix and a vector by $\oplus$. The addition between a matrix and a vector is performed by adding each column of the matrix by the vector. $\mathbf{u}$ is a refined query vector that combines information from the question and image regions. $\mathbf{u}^0$ (i.e, $\mathbf{u}$ from the first attention layer with $k = 0$) could be initialized as the feature vector of the question. $\mathbf{h}_A^k$ is then used to compute $\mathbf{p}_I^k$, which corresponds to the attention probability of each image region,

$$\mathbf{p}_I^k = \text{Softmax}(\mathbf{W}_3^k\mathbf{h}_A^k + \mathbf{b}_P^k). \tag{9.32}$$

$\mathbf{u}^k$ could be iterated by

$$\tilde{\mathbf{I}}^k = \sum_i \mathbf{p}_i^k\mathbf{I}_i, \tag{9.33}$$

$$\mathbf{u}^k = \mathbf{u}^{k-1} + \tilde{\mathbf{I}}^k. \tag{9.34}$$

That is, in every layer, the model progressively uses the combined question and image vector $\mathbf{u}^{k-1}$ as the query vector for attending the image region to obtain the new query.

The above models attend only on images, but questions should also be attended. [44] calculates co-attention by

$$\mathbf{Z} = \tanh(\mathbf{Q}^\top\mathbf{W}\mathbf{I}), \tag{9.35}$$

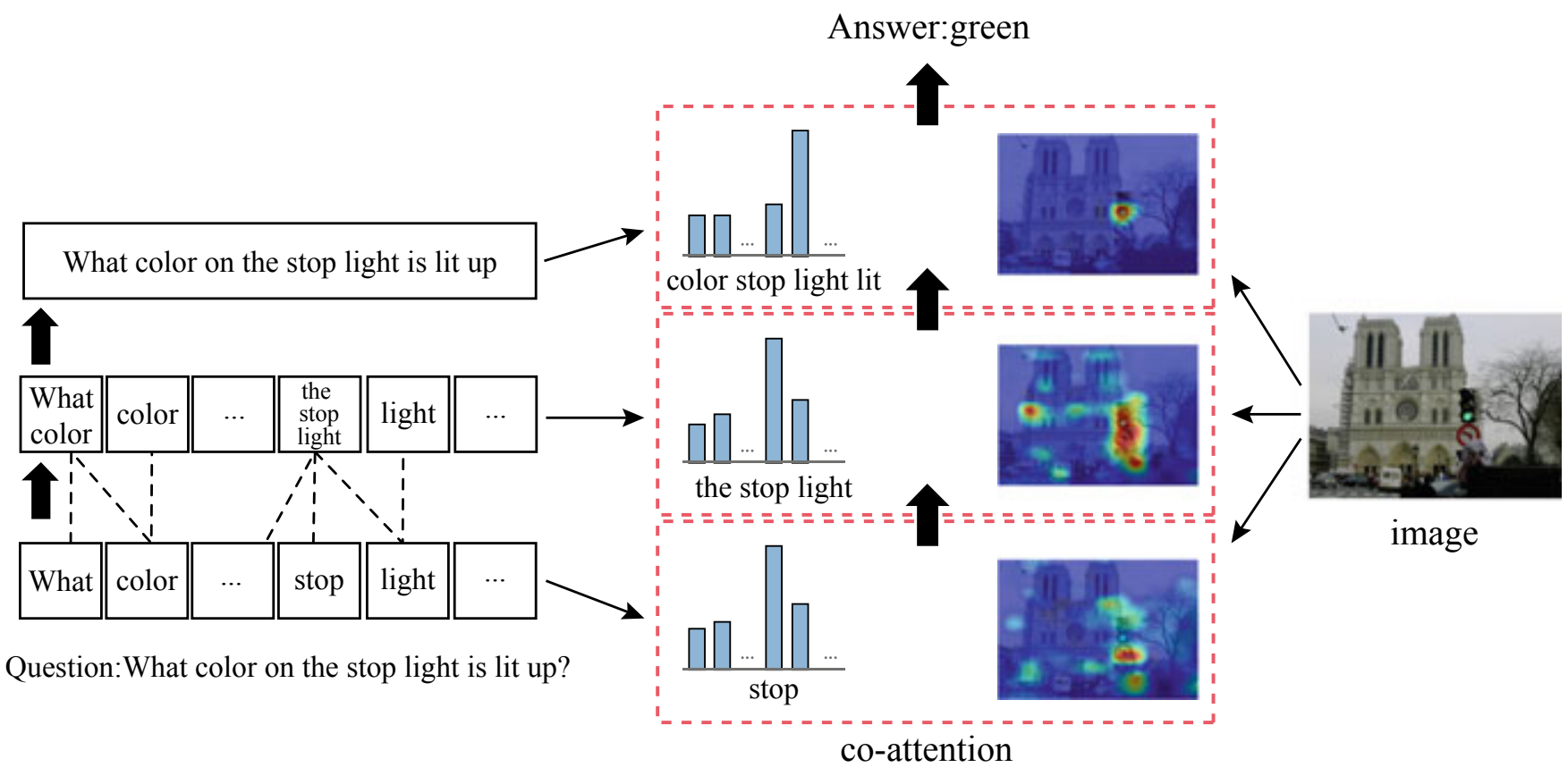

**Fig. 9.12** The architecture of hierarchical co-attention model

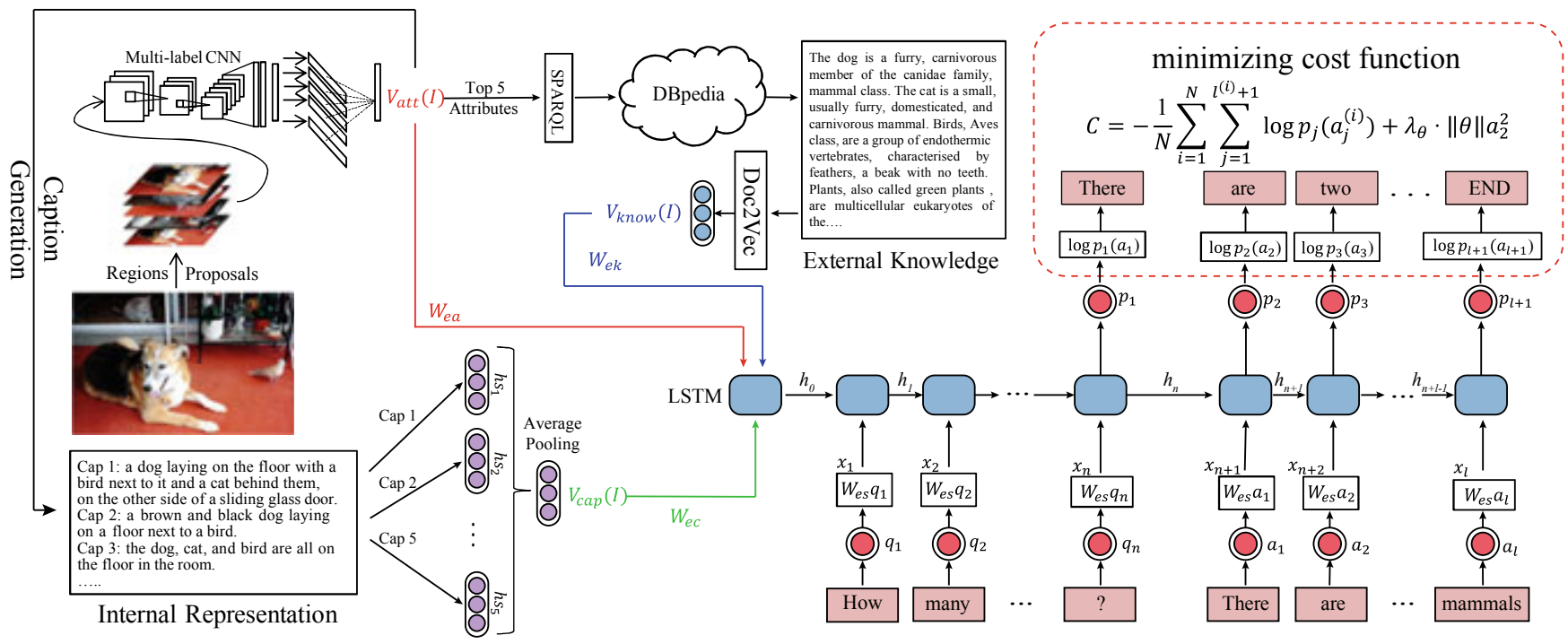

**Fig. 9.13** The architecture of VQA incorporating external knowledge bases

where $\mathbf{Z}_{ij}$ represents the affinity of the $i$th word and $j$th region. Figure 9.12 shows the hierarchical co-attention model.

Another intuitive approach is to use external knowledge from knowledge bases, which will help us better explain the implicit information hiding behind the image. Such an approach was proposed in [75], which first encodes the image into captions and vectors representing different attributes of the image to retrieve documents about a different part of the images from knowledge bases. Documents are encoded through doc2vec [36]. The representation of captions, attributes, and documents are transformed and concatenated to form the initial vector of an LSTM, which is trained in Seq2seq fashion. Details of the model are shown in Fig. 9.13.

Neural Module Network is a framework for constructing deep networks with a dynamic computational structure, which was first proposed in [3]. In such a framework, every input is associated with a layout that provides a template for assembling an instance-specific network from a collection of shallow network fragments called modules. The proposed method processes the input question through two separate ways: (1) parsing and laying out several modules, and (2) encoding by an LSTM. The corresponding picture is processed by the modules laid out according to the question, the types of which are predefined, `find`, `transform`, `combine`, `describe`, and `measure`. The authors defined `find` to be a transformation from *Image* to *Attention* map, `transform` to be a mapping from one *Attention* to another, `combine` to be a combination of two *Attention*, `describe` to be a description relying on *Image* and *Attention*, and `Measure` to be a measure only relying on *Attention*. The model is shown in Fig. 9.14.

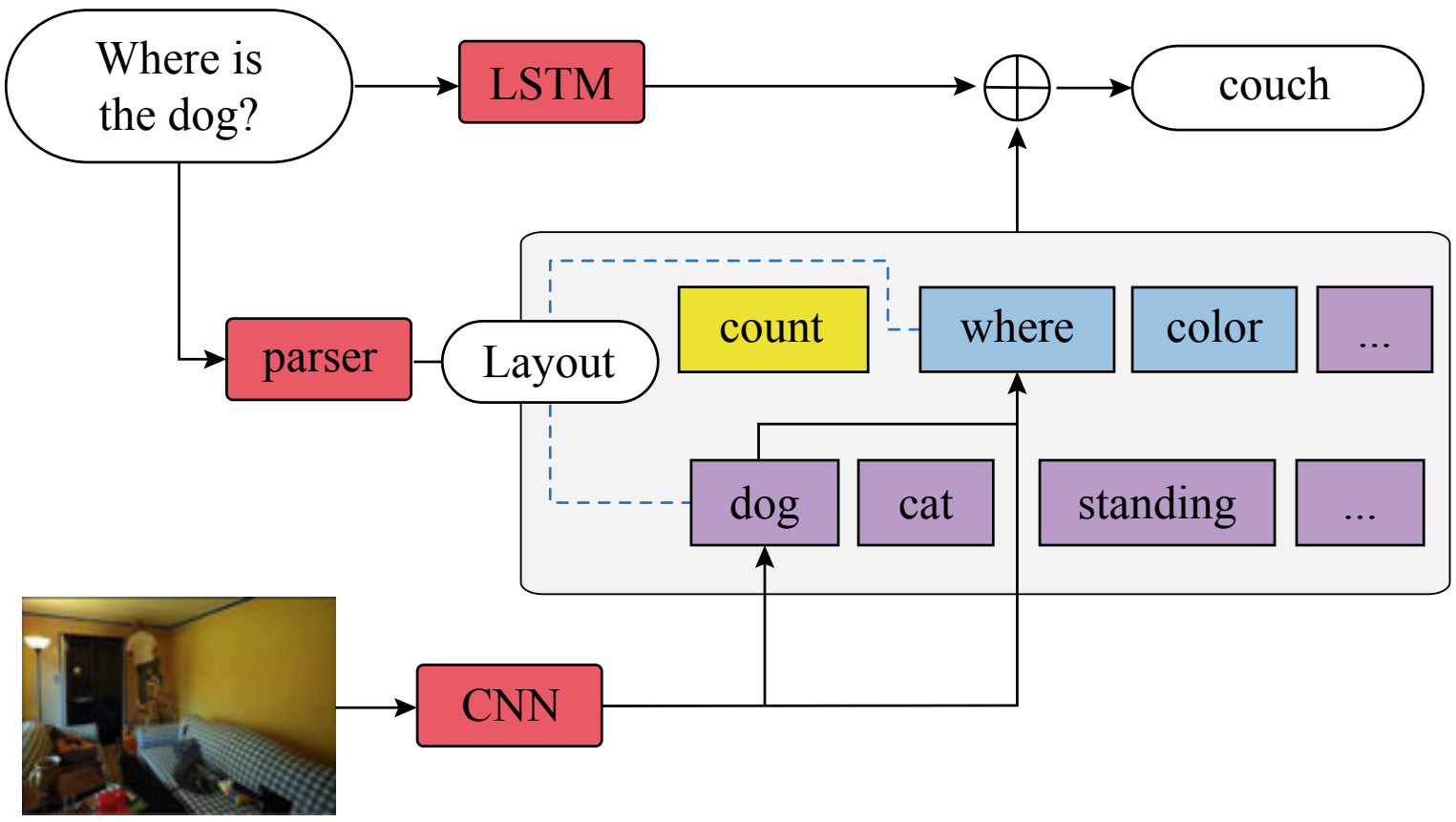

**Fig. 9.14** The architecture of the neural module network model

A key drawback of [3] is that it relies on the parser to generate modules. [22] proposes an end-to-end model to generate a sequence of Reverse-Polish expression to describe the module network, as shown in Fig. 9.15. And the overall architecture is shown in Fig. 9.16.

Graph Neural Networks (GNNs) have also been applied to VQA tasks. [68] tries to build graphs about both the scene and the question. The authors described a deep neural network to take advantage of such a structured representation. As shown in Fig. 9.17, the GNN-based VQA model could capture the relationships between words and objects.

## 9.6 Summary

In this chapter, we first introduce the concept of cross-modal representation learning. Cross-modal learning is essential since many real-world tasks require the ability to understand the information from different modalities, such as text and image. Next, we introduce the concept of cross-modal representation learning, which aims to exploit the links and enable better utilization of information from different modali-ties. And we overview existing cross-modal representation learning methods for several representative cross-modal tasks, including zero-shot recognition, cross-media retrieval, image captioning, and visual question answering. These cross-modal learning methods either try to fuse information from different modalities into unified embeddings, or try to build embeddings for different modalities in a common semantic space, allowing the model to compute cross-modal similarity. Cross-modal representation learning is drawing more and more attention and can serve as a promising connection between different research areas.

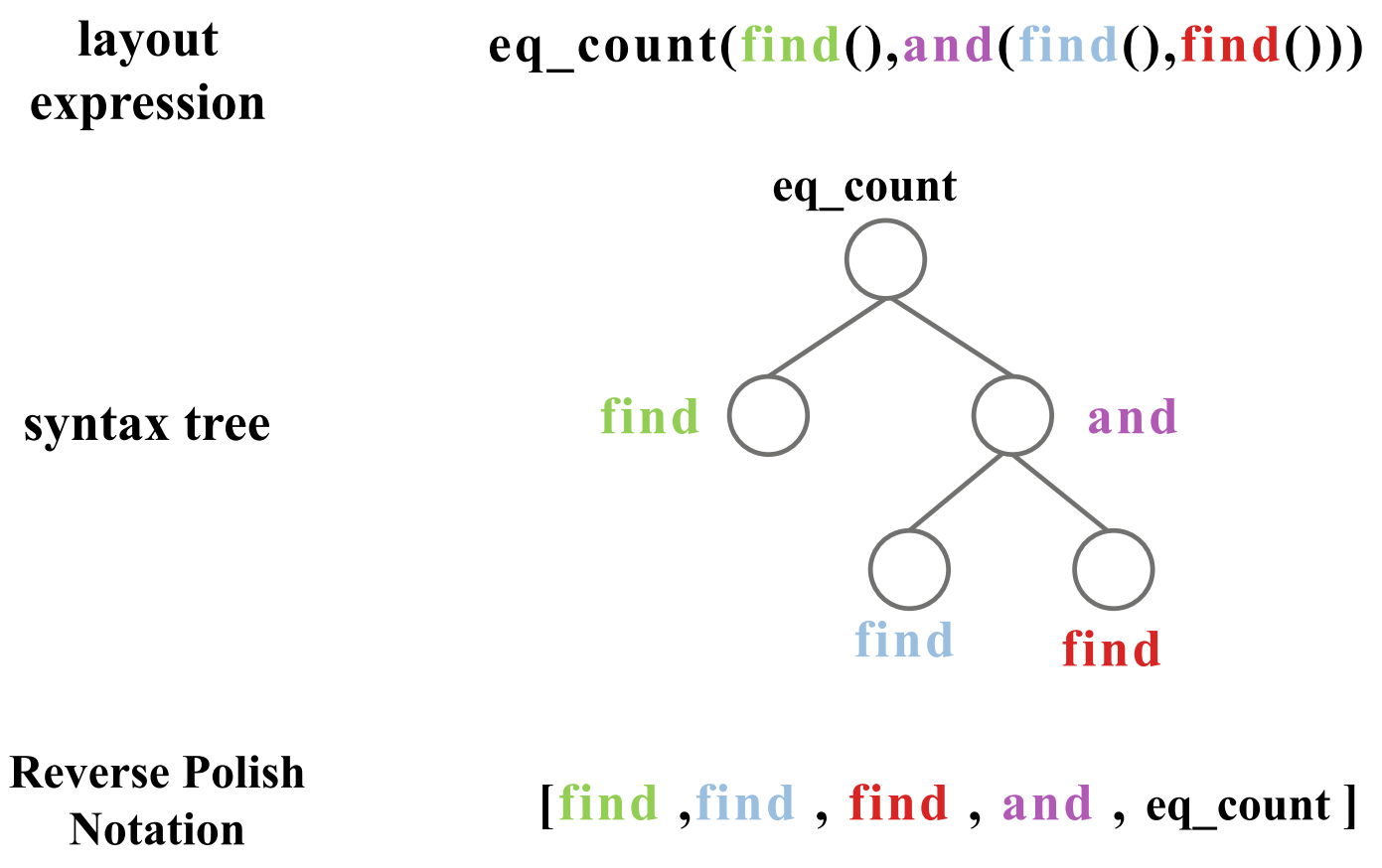

**Fig. 9.15** The architecture of Reverse-Polish expression and corresponding module network model

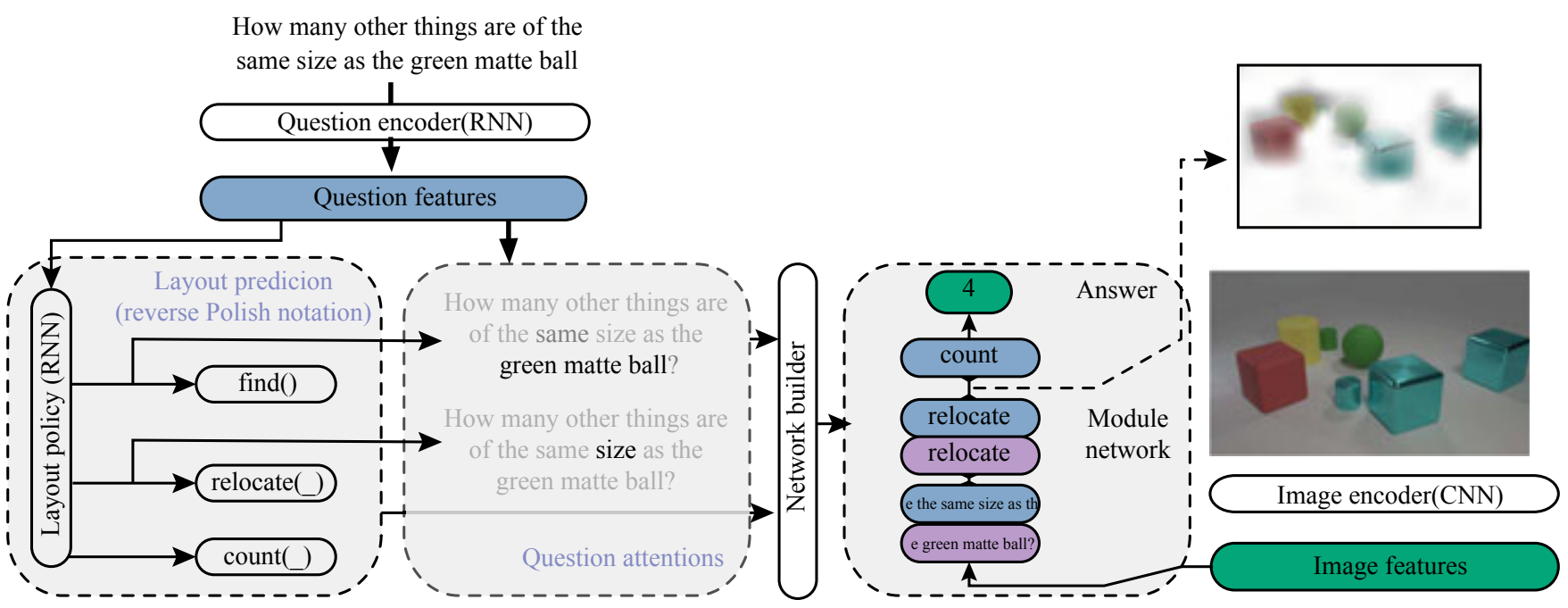

**Fig. 9.16** The architecture of end-to-end module network model

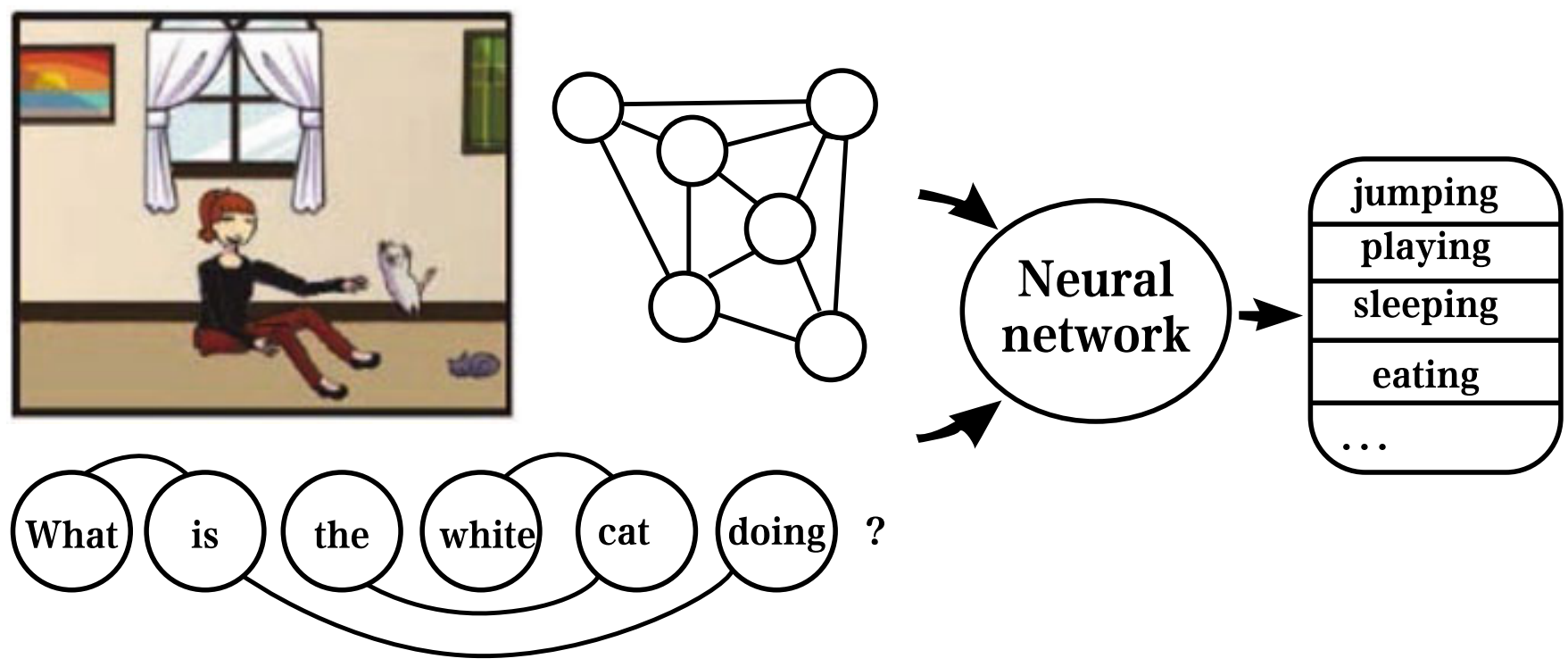

**Fig. 9.17** The architecture of GNN-based VQA models

For further understanding of cross-modal representation learning, there are also some recommended surveys and books including:

- Skocaj et al., Cross-modal learning [64].
- Spence, Crossmodal correspondences: A tutorial review [66].
- Wang et al., A comprehensive survey on cross-modal retrieval [72].

In the future, for better cross-modal representation learning, some directions are requiring further efforts:

(1) **Fine-grained Cross-modal Grounding**. Cross-modal grounding is a fundamental ability in solving cross-modal tasks, which aims to align semantic units in different modalities. For example, visual grounding aims to ground textual symbols (e.g., words or phrases) into visual objects or regions. Many existing works [27, 74, 76] have been devoted to cross-modal grounding, which mainly focuses on coarse-grained semantic unit grounding (e.g., grounding of sentences and images). Better fine-grained cross-modal grounding (e.g., grounding of words and objects) could promote the development of a broad variety of cross-modal tasks.

(2) **Cross-modal Reasoning**. In addition to recognizing and grounding semantic units in different modalities, understanding and inferring the relationship between semantic units are also crucial to cross-modal tasks. Many existing works [37, 41, 82] have investigated detecting visual relation between objects. However, most visual relations in existing visual relation detection datasets do not require complex reasoning. Some works [81] have made preliminary attempts on cross-modal commonsense reasoning. Inferring the latent semantic relationships in cross-modal context is critical for cross-modal understanding and modeling.

(3) **Utilizing Unsupervised Cross-modal Data**. Most current cross-modal learning approaches rely on human-annotated datasets. The scale of such supervised datasets is usually limited, which also limits the capability of data-hungry neural models. With the rapid development of the World Wide Web, cross-modal data on the Web have become larger and larger. Some existing works [42, 67] have leveraged unsupervised cross-modal data for representation learning. They first pretrained cross-modal models on large-scale image-caption pairs, and then fine-tuned the models on those downstream tasks, which shows significant improvement in a broad variety of cross-modal tasks. It is thus promising to better leverage the vast amount of unsupervised cross-modal data for representation learning.

# Chapter 10
# Resources

**Abstract** Deep learning has been shown as a powerful method for a variety of artificial intelligence tasks, including some critical tasks in NLP. However, training a deep neural network is usually a very time-intensive process and requires lots of code to build related models. To alleviate these issues, some deep learning frameworks have been developed and released, which incorporate some existing and necessary arithmetic operators for neural network constructions. And these frameworks exploit hardware features such as multi-core CPUs and many-core GPUs to shorten the training time. Each framework has its advantages and disadvantages. In this chapter, we aim to exhibit features and running performance of these frameworks so that users can select an appropriate framework for their usage.

## 10.1 Open-Source Frameworks for Deep Learning

In this section, we will introduce several typical open-source frameworks for deep learning including Caffe, Theano, TensorFlow, Torch, PyTorch, Keras, and MXNet. In fact, as the rapid development of the deep learning community, these open-source frameworks are updating every day, and therefore the information in this section may not be up to date. In fact, this section mainly focuses on introducing the special features of these frameworks and lets the readers have a preliminary understanding of them. To know the latest features of these deep learning frameworks, please refer to their official sites.

### *10.1.1 Caffe*

Caffe[1] is a well-known framework and is widely used for computer vision tasks. It was created by Yangqing Jia and developed by Berkeley AI Research (BAIR). Caffe uses a layer-wise approach to make building models become easy, and it is also convenient to fine-tune the existing neural networks without writing too much code via its simple interfaces. The underlying designs of Caffe are for the fast construction of convolutional neural networks, which make it efficient and effective.

[1] http://caffe.berkeleyvision.org/.

On the other hand, as normal pictures often have a fixed size, the interfaces of Caffe are fixed and hard to be extended. It is thus difficult to use Caffe for other tasks with a variable input length, such as text, sound, or other time-series data. Recurrent neural networks are also not well supported by Caffe. Although users can easily build an existing network architecture with the layer-wise framework, it is not flexible when dealing with big and complex networks. If users want to design a new layer, the users need to use C/C++ and CUDA for the underlying coding of the new layer.

### *10.1.2 Theano*

Theano[2] is the typical framework developed to use symbolic tensor graphs for model specification. Any neural networks or other machine-learning models can be represented as symbolic tensor graphs. Forward, backward, and gradient updates can be calculated based on the flow between tensors. Hence, Theano provides more flexibility than Caffe using a layer-wise approach to build models. In Caffe, to define a new layer that is not already in the existing repository of layers is complicated, which needs to implement its forward, backward, and gradient update functions before. In Theano, you only need to use basic operators to define the customized layer following the order of operations.

Theano is a platform and is easy to configure as compared with other frameworks. And some high-level frameworks are built on top of Theano such as Keras, which further makes Theano easier to use. Theano supports cross-platform configuration well, which means it works on not only Linux but also Windows. Because of this, many researchers and engineers use Theano to build their models and then release these projects. Rich open resources based on Theano attract some more users.

Though Theano uses Python syntax to define symbolic tensor graphs, its graph processor will compile the graphs into high-performance C++ or CUDA code for computing. Owing to this, Theano can run very fast and make programmers code mode simply. Only one deficiency is that the compilation process is slow and needs some time. If a neural network does not need to be trained for several days, it is not a good idea to select Theano. Compiling too often in Theano is maddening and annoying. As a comparison, the later framework like TensorFlow uses the compiled package for the symbolic tensor operations, which seems a little more relaxing.

Theano has some other serious disadvantages. Theano cannot support many-core GPUs very well, which makes it hard to train big neural networks. Besides the compilation process, importing Theano is also slow. When you run your code in Theano, you will be stuck for a long time with a preconfigured device. If you want to improve and contribute to Theano itself, this will also be maddening and annoying.

[2]http://www.deeplearning.net/software/theano/.

In fact, Theano is no longer maintained, but it is still worth introducing as a landmark work in the history of deep learning frameworks, which inspires many subsequent frameworks.

### 10.1.3 TensorFlow

TensorFlow[3] is mainly developed and used by Google based on the experience on Theano and DistBelief [1]. TensorFlow and Theano are in fact quite similar to some extent. Both of them allow building a symbolic graph of the neural network architecture via the Python interface. Different from Theano, TensorFlow allows implementing new operations or machine-learning algorithms using C/C++ and Java. With building symbolic graphs, the auto-gradient can be easily used to train complicated models. Hence, TensorFlow is more than a deep learning framework. Its flexibility enables it to solve various complex computing problems such as reinforcement learning.

In TensorFlow, both code development and deployment is fast and convenient. Trained models can be deployed quickly on a variety of devices, including servers and mobile devices, without the need to implement a separate model setting code or load Python/LuaJIT interpreter. Caffe also allows easy deployment of models. However, Caffe has trouble running on devices without a GPU, which is a prevalent situation of smartphones. TensorFlow supports model decoding using ARM/NEON instructions and does not need too many operations to choose training devices.

TensorBoard of TensorFlow provides a platform for visualization of the model architectures, which is beautiful and also useful. By visualizing the symbolic graph, it is not difficult to find bugs in the source code. To debug models on other deep learning frameworks is relatively bothering. TensorBoard can also log and generate real-time visualization of variables during training, which is a pleasant way to monitor the training process.

Though customizing operations in TensorFlow is convenient, it usually changes a lot of function interfaces in every new release which is challenging for developers to keep their code compatible with different TensorFlow versions. And mastering TensorFlow is also not easy. As TensorFlow 2.0 has been released recently, TensorFlow may gradually handle these issues in the predictable future.

### 10.1.4 Torch

Torch[4] is a computational framework mainly developed and used by Facebook and Twitter. Torch provides an API written in Lua to support the implementation of some machine-learning algorithms, especially convolutional neural networks. A temporal convolutional layer implemented by Torch can have a variable input length, which is extremely useful for NLP tasks and not designed in Theano and TensorFlow. Torch also contains the 3D convolutional layer, which can be easily used in video recognition tasks. Besides its various flexible convolutional layers, Torch is light and speedy. The above reasons attract lots of researchers in universities and companies to customize their own deep learning platforms.

---

[3] https://www.tensorflow.org/.

[4] http://torch.ch/.

However, the negative aspects of Torch are also apparent. Though Torch is powerful, it is not designed to be widely accessible to the Python-based academic community. And there are not any other interfaces but Lua. Lua is a multi-paradigm scripting language, which was developed in Brazil in the early 1990s and is not a popular mainstream programming language. Hence, it needs some time to learn Lua before you use Torch to construct models. Different from convolutional neural networks, there is no official support for recurrent neural networks. There are some open resources about recurrent neural networks implemented by Torch, but they are not yet integrated to the main repository. And it is difficult to distinguish the effectiveness of these implementations.

Similar to Caffe, Torch is not a framework based on symbolic tensor graphs, it also uses the layer-wise approach. This means that your models in Torch are a graph of layers and not a graph of mathematical functions. The mechanism is convenient to build a network whose layers are stable and hierarchical. If you want to design a new connection layer or change an existing neural model, you need lots of code to implement new layers with full forward, backward, and gradient update functions. However, those frameworks based on symbolic tensor graphs, such as Theano and TensorFlow, give more flexibility to do this. In fact, these issues are handled as PyTorch has been released, which we will introduce then.

### *10.1.5 PyTorch*

PyTorch[5] is a Python package built over Torch, developed by Facebook and other companies. However, it is not just an interface, and PyTorch has amounts of improvements over Torch. The most important one is that PyTorch can use a symbolic graph to define neural networks, and then use automatic differentiation following the graph to automate the computation of backward passes in neural networks. Meanwhile, PyTorch maintains some characteristics of the layer-wise approach in Torch, which means coding with PyTorch is easy. Moreover, PyTorch has minimal framework overhead and custom memory allocators for the GPU, which means PyTorch is faster and memory-efficient than Torch.

[5] http://pytorch.org/.

Compared with other deep learning frameworks, PyTorch has two main advantages. First, most frameworks like TensorFlow are based on static computational graphs (define-and-run), while PyTorch uses dynamic computational graphs (define-by-run). What it means is that with dynamic computational graphs, you can change the network architecture based on the data flowing through the network. There is a way to do something similar in TensorFlow, but your static computational graphs must contain all possible branches in advance, which will limit the performance. Second, PyTorch is built to be deeply integrated into Python and has a seamless connection with other popular Python packages, such as Numpy, Spicy, and Cython. Thus, it is easy to extend your model when needed.

After Facebook released it, PyTorch has drawn considerable attention from the deep learning community, and many past Torch users switch to this new package. For now, PyTorch already has a thriving community which contributes to its increasing popularity among researchers. It is no exaggeration to say that PyTorch is one of the most popular frameworks at present.

### *10.1.6 Keras*

Keras[6] is a top-design deep learning framework that is based on Theano and TensorFlow. Interestingly, Keras sits atop Theano and TensorFlow, however, its interfaces are similar to Torch. To use Keras needs Python code, and there are lots of detailed documents and examples for a quick start. There is also a very active community of developers, and they make Keras fastly updated. Hence, it is a very fast-growing framework.

Because Theano and TensorFlow are the backends of Keras, disadvantages of Keras are most similar to Theano and TensorFlow. With TensorFlow as the backend, it will run even slower than the pure TensorFlow code. Because it is a high-level framework, to customize a new neural layer is not easy, though you can easily use existing layers under Keras. The package is too advanced, and it hides too many training parameters. You cannot touch and change all details of your own models unless you use Theano, TensorFlow, or PyTorch.

### *10.1.7 MXNet*

MXNet[7] is an effective and efficient open-source machine-learning framework, mainly pushed by Amazon. It supports APIs with multiple languages, including C++, Python, R, Scala, Julia, Perl, MATLAB, and JavaScript, some of which can be adopted for Amazon Web Services. Some interfaces of MXNet are also reserved for future mobile devices, just like TensorFlow. MXNet is built on a dynamic dependency scheduler that automatically parallelizes both symbolic and imperative operations on the fly. A graph optimization layer on top of that makes symbolic execution fast and memory efficient. The MXNet library is portable and lightweight, and it scales to multiple GPUs and multiple machines. The main problem of MXNet is the lack of detailed and well-organized documentation. The user groups are also smaller than other frameworks, especially as compared with TensorFlow and PyTorch. It is more challenging to grasp MXNet for newbies. The MXNet is developing fastly, and these problems may be solved in the future.

---

[6] https://keras.io/.

[7] http://mxnet.io/.

## 10.2 Open Resources for Word Representation

### *10.2.1 Word2Vec*

Word2vec[8] is a widely used toolkit for word representation learning, which provides an effective and efficient implementation of the continuous bag-of-words and Skip-gram architectures. The word representations learned by Word2vec can be used in many natural language processing fields. Empirically, To use pretrained word vectors as the model inputs can be a good way to enhance model performances.

Word2vec takes free text corpus as input and constructs the vocabulary list from the training data. Then it uses simple predictive models based on neural networks to learn the language model, which encode the co-occurrence information between words into the resulting word representations.

The resulting representations showcase interesting linear substructures of the word vector space. The Euclidean distance (or cosine similarity) between two-word vectors provides an effective method for measuring the linguistic or semantic similarity of the corresponding words. Sometimes, the nearest neighbors, according to this metric, reveal rare but relevant words that lie outside an average human's vocabulary.

Words frequently appearing together in the text will have representations with close distance within the embedding space. Word2vec also provides a tool to find the closest words for a user-specified word via the learned representations and distances between representation embeddings.

### *10.2.2 GloVe*

GloVe[9] is a widely used toolkit, which supports an unsupervised learning method for word representation learning. Similar to Word2vec, GloVe also trains on text corpus and captures the aggregated global word-word co-occurrence information for word embeddings. However, GloVe uses count-based models instead of predictive models, which are different from Word2vec.

[8] https://code.google.com/archive/p/word2vec/.

[9] https://nlp.stanford.edu/projects/glove/.

The GloVe model first builds a global word-word co-occurrence matrix, which can show how frequently words co-occur with one another in a given text. Then word representations are trained on the nonzero entries of the matrix. To construct this matrix requires the entire corpus traversal for the statistics collection. For large corpora, this pass can be computationally expensive, but it is a one-time up-front cost. Subsequent training iterations are much faster because the number of nonzero matrix entries is typically much smaller than the total number of words in the corpus.

## 10.3 Open Resources for Knowledge Graph Representation

### *10.3.1 OpenKE*

OpenKE[10] [2] is an open-source toolkit for Knowledge Embedding (KE), which provides a unified framework and various fundamental KE models. OpenKE prioritizes operational efficiency to support quick model validation and large-scale knowledge representation learning. Meanwhile, OpenKE maintains sufficient modularity and extensibility to incorporate new models easily. Besides the toolkit, the embeddings of some existing large-scale knowledge graphs pretrained by OpenKE are also available. The toolkit, documentation, and pretrained embeddings are all released on http://openke.thunlp.org/.

As compared to other implementations, OpenKE has five advantages. First, OpenKE has implemented nine classical knowledge embedding algorithms, including RESCAL, TransE, TransH, TransR, TransD, ComplEx, DistMult, HolE, and Analogy, which are verified effective and stable. Second, OpenKE shows high performance due to memory optimization, multi-threading acceleration, and GPU learning. OpenKE supports multiple computing devices and provides interfaces to control CPU/GPU modes. Third, system encapsulation makes OpenKE easy to train and test KE models. Users just need to set hyperparameters via interfaces of the platform to construct KE models. Fourth, it is easy to construct new KE models. All specific models are implemented by inheriting the base class by designing their own scoring functions and loss functions. Fifth, besides the toolkit, OpenKE also provides the embeddings of some existing large-scale knowledge graphs pretrained by OpenKE, which can be directly applied for many applications, including information retrieval, personalized recommendation, and question answering.

[10] https://github.com/thunlp/OpenKE.

### 10.3.2 Scikit-Kge

Scikit-kge[11] is an open-source Python library for knowledge representation learning. The library supports different building blocks to train and develop models for knowledge graph embeddings. The primary purpose of Scikit-kge is to compute the embeddings of knowledge graphs for the method HolE; meanwhile, it also provides some other methods. Besides HolE, RESCAL, TransE, TransR, and ER-MLP can also be trained in Scikit-kge. The library contains some parameter update methods, not only the basic SGD but also AdaGrad. It also implements different negative sampling strategies to select negative samples.

## 10.4 Open Resources for Network Representation

### 10.4.1 OpenNE

OpenNE[12] is an open-source standard NE/NRL (Network Representation Learning) training and testing framework. It unifies the input and output interfaces of different NE models and provides scalable options for each model. Moreover, typical NE models under this framework are based on TensorFlow, which enables these models to be trained with GPUs. The implemented or modified models include DeepWalk, LINE, node2vec, GraRep, TADW, GCN, HOPE, GF, SDNE, and LE. The framework also provides classification and embedding visualization modules for evaluating the result of NRL.

### 10.4.2 GEM

GEM (Graph Embedding Methods)[13] is a Python package that offers a general framework for graph embedding methods. It implements many state-of-the-art embedding techniques including Locally Linear Embedding, Laplacian Eigenmaps, Graph Factorization, High-Order Proximity preserved Embedding (HOPE), Structural Deep Network Embedding (SDNE), and node2vec. Furthermore, the framework implements several functions to evaluate the quality of the obtained embeddings including graph reconstruction, link prediction, visualization, and node classification. For faster execution, C++ backend is integrated using Boost for supported methods.

[11] https://github.com/mnick/scikit-kge.

[12] https://github.com/thunlp/OpenNE.

[13] https://github.com/palash1992/GEM.

### 10.4.3 GraphVite

GraphVite[14] is a general and high-performance graph embedding system for various applications including node embedding, knowledge graph embedding, and graph high-dimensional data visualization.

GraphVite provides a complete pipeline for users to implement and evaluate graph embedding models. For reproducibility, the system integrates several commonly used models and benchmarks, and you can also develop your own models with the flexible interface. Additionally, for semantic tasks, GraphVite releases a bunch of pretrained knowledge graph embedding models to enhance language understanding. There are two core advantages of GraphVite over other toolkits: fast and large-scale training. GraphVite accelerates graph embedding with multiple CPUs and GPUs. It takes around one minute to learn node embeddings for graphs with one million nodes. Moreover, GraphVite is designed to be scalable. Even with limited memory, GraphVite can process node embedding task on billion-scale graphs.

### 10.4.4 CogDL

CogDL[15] is another graph representation learning toolkit that allows researchers and developers to easily train and evaluate baseline or custom models for node classification, link prediction, and other tasks on graphs. It provides implementations of many popular models, including non-GNN models and GNN-based ones.

CogDL benefits from several unique techniques. First, utilizing sparse matrix operation, CogDL is capable of performing fast network embedding on large-scale networks. Second, CogDL has the ability to deal with different types of graph structures attributed, multiplex, and heterogeneous networks. Third, CogDL supports parallel training. With different seeds and different models, CogDL performs training on multiple GPUs and reports the result table automatically. Finally, CogDL is extendable. New datasets, models, and tasks can be added without difficulty.

## 10.5 Open Resources for Relation Extraction

### 10.5.1 OpenNRE

OpenNRE[16] [3] is an open-source framework for neural relation extraction, which aims to easily build relation extraction (RE) models.

[14] https://graphvite.io/.

[15] http://keg.cs.tsinghua.edu.cn/cogdl/index.html.

[16] https://github.com/thunlp/OpenNRE.

Compared with other implementations, OpenNRE has four advantages. First, OpenNRE has implemented various state-of-the-art RE models, including attention mechanism, adversarial learning, and reinforcement learning. Second, OpenNRE enjoys great system encapsulation. It divides the pipeline of relation extraction into four parts, namely, embedding, encoder, selector (for distant supervision), and classifier. For each part, it has implemented several methods. System encapsulation makes it easy to train and test models by changing hyperparameters or appoint model architectures by using Python arguments. Third, OpenNRE is extendable. Users can construct new RE models by choosing specific blocks provided in four parts as mentioned above and combining them freely, with only a few lines of codes. Fourth, the framework has implemented multi-GPU learning, which is efficient.

# Chapter 11
# Outlook

**Abstract** The aforementioned representation learning models and methods have shown their effectiveness in various NLP scenarios and tasks. With the rapid growth of data scales and the development of computation devices, there are also new challenges and opportunities for next-stage researches of deep learning techniques. In the last chapter, we will look into the future directions of representation learning techniques for NLP. To be more specific, we will consider the following directions including using more unsupervised data, utilizing few labeled data, employing deeper neural architectures, improving model interpretability and fusing the advantages of other areas.

## 11.1 Introduction

We have used ten chapters to introduce the advances of representation learning for NLP, covering both multi-grained language entries including words, phrases, sentences and documents, and closely related objects including world knowledge, sememe knowledge, networks, and cross-modal data. Those mentioned models and methods of representation learning for NLP have shown their effectiveness in various NLP scenarios and tasks.

As shown by the unsatisfactory performance of most NLP systems in open domains, and recent great advances of pre-trained language models, representation learning for NLP is far from perfect. With the rapid growth of data scales and the development of computation devices, we are facing new challenges and opportunities for next-stage researches of representation learning and deep learning techniques.

In this last chapter, we will look into the future research and exploration directions of representation learning techniques for NLP. Since we have summarized the future work of each individual part in the summary section of each previous chapter, here we focus on discussing the general and important issues that should be addressed by representation learning for NLP.

For general representation learning for NLP, we conclude the following directions, including using more unsupervised data, utilizing a few labeled data, employing deeper neural architectures, improving model interpretability, and fusing the advances from other areas.

## 11.2 Using More Unsupervised Data

The rapid development of Internet technology and the popularization of information digitization have brought massive text data for NLP researches and applications. For example, the whole corpus of Wikipedia already contains more than 50 million articles (including 6 million articles in English)[1] and is growing rapidly every day contributed by collaborative work all over the world. The amount of user-generated content on many social platforms such as Twitter, Weibo, and Facebook also increases quickly by billions of users. It is worth considering these massive text data for learning better NLP models. However, due to the expensive cost of expert annotations, it is impossible to label such massive amounts of data for specific NLP tasks.

Hence, an essential direction of NLP is how to take better advantages of unlabeled data for efficient unsupervised representation learning. Though without labeled annotations, unsupervised data can help initialize the randomized neural network parameters and thus improve the performances of those downstream NLP tasks.

This line of work usually employs a pipeline strategy: first, pretrain the model parameters and then fine-tune these parameters in specific downstream NLP tasks. Recurrent language model [7], word embeddings [6], and pre-trained language models (PLM) such as BERT [3], all utilize unsupervised plain text to pretrain neural parameters and then benefit downstream supervised tasks via fine-tuning.

Current state-of-the-art PLM models still can only learn from limited plain text due to limited learning efficiency and computation power. Moreover, there are various types of large-scale data online with abundant informative signals and labels, such as HTML tags, anchor text, keywords, document meta-information, and other structured and semi-structured data. How to take full advantage of the large-scale Web text data has not been extensively studied. In the future, with better computation devices (e.g., GPUs) and data resources, we are expected to develop more advanced methods to utilize more unsupervised data.

## 11.3 Utilizing Fewer Labeled Data

As NLP technologies become more powerful, people can explore more complicated and fine-grained problems. Taking text classification as an example, early work targeted on flat classification with limited categories, and now researchers are more interested in classification with hierarchical structure and a large number of classes. However, when a problem gets more complicated, it requires more knowledge from experts to annotate training instances for fine-grained tasks and increases the cost of data labeling.

[1] http://en.wikipedia.org/wiki.

Therefore, we expect the models or systems can be developed efficiently with (very) few labeled data. When each class has only one or a few labeled instances, the problem becomes a one/few-shot learning problem. The few-shot learning problem is derived from computer vision and has also been studied in NLP recently. For example, researchers have explored few-shot relation extraction [5] where each relation has a few labeled instances, and low-resource machine translation [11] where the size of the parallel corpus is limited.

A promising approach to few-shot learning is to compare the semantic similarity between the test instance and those labeled ones (i.e., the support set), and then make the prediction. The idea is similar to $k$-nearest neighbor classification ($k$NN) [10]. Since the key is to represent the semantic meanings of each instance for measuring their semantic similarity, it has been verified that language models pretrained on unsupervised data and fine-tuned on the target few-shot domain are very effective for few-shot learning.

Another approach to few-shot learning is to transfer the models from some related domains into the target domain with the few-shot problem [2]. This is usually named as transfer learning or domain adaptation. For these methods, representation learning can also help the transfer or adaptation process by learning joint representations of both domains.

In the future, one may go beyond the abovementioned frameworks and design more appropriate methods according to the characteristics of NLP tasks and problems. The goal is to develop effective NLP methods with as less annotated data in the target domain as possible, by better utilizing unsupervised data that are much cheaper to get from the Web and existing supervised data from other domains. The exploration of the few-shot learning problem in NLP will help us develop data-efficient methods for language learning.

## 11.4 Employing Deeper Neural Architectures

As the amount of available text data rapidly increases, the size of the training corpus for NLP tasks grows as well. With more training data, a natural way to boost model performances is to employ deeper neural architectures for modeling. Intuitively, deeper neural models that have more sophisticated architecture and parameters can better fit the increasing data. Another motivation for using deeper architectures for modeling comes from the development of computation devices (e.g., GPUs). Current state-of-the-art methods are usually a compromise between efficiency and effectiveness. As the computation devices operate faster, the time/space complexities of complicated models become acceptable, which motivate researchers to design more complex but effective models. To summarize, employing deeper neural architectures would be one of the definite orientations for representation learning in NLP.

Very deep neural network architectures have been widely used in computer vision. For example, the well-known VGG [8] network which was proposed in the famous ImageNet contest has 16 layers of convolutional and fully connected layers. In NLP, the depths of neural architectures were relatively shallow until the Transformer [9] structure was proposed. Specifically, as compared with word embedding [6] which is based on shallow models, the state-of-the-art pre-trained language model BERT [3] can be regarded as a giant model that stacks 12 self-attention layers and each layer has 8 attention heads. BERT has demonstrated its effectiveness in a number of NLP tasks. Besides the well-designed model architecture and training objectives, the success of BERT also benefits from TPUs which is one of the most powerful devices for parallel computations. In contrast, it may take months or years for a single CPU to finish the training process of BERT. When these computation devices go popular, we can expect more deep neural architectures to be developed for NLP as well.

## 11.5 Improving Model Interpretability

Model transparency and interpretability are hot topics in artificial intelligence and machine learning. Human interpretable predictions are very important for decision-critical applications related to ethics, privacy, and safety. However, neural network models or deep learning techniques are short of model transparency for human interpretable predictions and thus are often treated as black boxes.

Most NLP techniques based on neural networks and distributed representation are also hard to be interpreted except for the attention mechanism where the attention weights can be interpreted as the importance of corresponding inputs. For the sake of employing representation learning techniques for decision-critical applications, there is a need to improve model interpretability and transparency of current representation learning and neural network models.

A recent survey [1] classifies interpretable machine learning methods into two main categories: interpretable models and post-hoc explainability techniques. Models that are understandable by themselves are called interpretable models. For example, linear models, decision trees, and rule-based systems are such transparent models. However, in most cases, we have to probe into the model by a second one for explanations, namely post-hoc explainability techniques. In NLP, there have been some researches to visualize neural models such as neural machine translation [4] for interpretable explanations. However, the understanding of most neural-based models remains unsolved. We are looking forward to more studies on improving model interpretability to facilitate the extensive use of representation learning methods for NLP.

## 11.6 Fusing the Advances from Other Areas

During the development of deep learning techniques, mutual learning between different research areas has never stopped.

For example, Word2vec aims to learn word embeddings from large-scale text corpus published in 2013 and can be regarded as a milestone of representation learning for NLP. In 2014, the idea of Word2vec was adopted for learning node embeddings in a network/graph by treating random walks over the network as sentences, named as DeepWalk; the analogical reasoning phenomenon learned by Word2vec, i.e., king − man = queen − woman also inspired the representation learning of world knowledge, named as TransE. Meanwhile, graph convolutional networks were first proposed for semi-supervised graph learning in 2016, and have been widely applied on many NLP tasks such as relation extraction and text classification recently. Another example is the Transformer model which was proposed for neural machine translation at first and then transferred to computer vision, data mining, and many other areas.

The fusion also appears between two quite distant disciplines. We should recall again that, the idea of distributed representation proposed in the 1980s is inspired by the neural computation scheme of humans and other animals. It takes about 40 years to see the development of distributed representation and deep learning come to fruition. In fact, many ideas such as convolution in CNN and the attention mechanism are inspired by the computation scheme of human cognition.

Therefore, an intriguing direction of representation learning for NLP is to fuse the advances from other areas, including not only those closely related areas in AI such as machine learning, computer vision, and data mining, but also those distant areas to some extent such as linguistics, brain science, psychology, and sociology. This line of work requires researchers to have sufficient knowledge of other fields.